\documentclass[oneside,11pt, a4paper, footinclude=true, headinclude=true, cleardoublepage=empty]{scrbook}
\usepackage[linedheaders,parts,pdfspacing,dottedtoc]{classicthesis}
\usepackage{amsmath}
\usepackage{acronym}
\usepackage[a4paper, hmargin={2.8cm, 2.8cm}, vmargin={2.5cm, 2.5cm}]{geometry}
\usepackage{eso-pic} 
\usepackage{graphicx} 
\usepackage[utf8]{inputenc}
\usepackage[english]{babel}
\usepackage{csquotes}
\usepackage{bookmark}
\usepackage{longtable}
\usepackage{array}
\usepackage{multirow}
\usepackage{times}
\usepackage{lingmacros}
\usepackage{color, colortbl}
\usepackage{tabularx}
\usepackage{pdfpages}
\usepackage{footnote}
\definecolor{mygray}{rgb}{0.86,0.86,0.86}
\makesavenoteenv{tabular}
\makesavenoteenv{table}
\DeclareUnicodeCharacter{0301}{\'{e}}
\DeclareUnicodeCharacter{0308}{{\"o}}

\def\signed #1{{\leavevmode\unskip\nobreak\hfil\penalty50\hskip2em
  \hbox{}\nobreak\hfil(#1)%
    \parfillskip=0pt \finalhyphendemerits=0 \endgraf}}

    \newsavebox\mybox

\usepackage[backend=biber,style=authoryear,dashed=false,sorting=nty,maxcitenames=2,maxbibnames=99,uniquelist=false]{biblatex}

\setkomafont{disposition}{\normalfont\bfseries\small}
\setkomafont{chapter}{}
\setkomafont{section}{}
\addtokomafont{chapterentry}{\mdseries}
 
\clearpairofpagestyles
\ihead*{\pagemark}
 
\usepackage{comment}
\usepackage{enumitem}
\DeclareOldFontCommand{\bf}{\normalfont\bfseries}{\mathbf}
\usepackage{xparse}

\NewDocumentCommand{\rot}{O{45} O{1em} m}{\makebox[#2][l]{\rotatebox{#1}{#3}}}%

\usepackage{adjustbox}

\newcolumntype{R}[2]{%
    >{\adjustbox{angle=#1,lap=\width-(#2)}\bgroup}%
    l%
    <{\egroup}%
}

\newcommand\blfootnote[1]{%
  \begingroup
  \renewcommand\thefootnote{}\footnote{#1}%
  \addtocounter{footnote}{-1}%
  \endgroup
}

\preto\fullcite{\AtNextCite{\defcounter{maxnames}{99}}}

\usepackage[htt]{hyphenat}

\usepackage{latexsym}

\usepackage{url}
\usepackage{tikz}
\usepackage{xcolor}
\usetikzlibrary{arrows}
\usetikzlibrary{decorations.text}

\usepackage{amssymb}
\usepackage{booktabs}

\definecolor{nice-red}{HTML}{E41A1C}
\definecolor{nice-orange}{HTML}{FF7F00}
\definecolor{nice-yellow}{HTML}{FFC020}
\definecolor{nice-green}{HTML}{4DAF4A}
\definecolor{nice-blue}{HTML}{377EB8}
\definecolor{nice-purple}{HTML}{984EA3}

\usepackage{lineno}

\usepackage{framed}
\usepackage{scrextend}
\usepackage{bm}
\usepackage[mathscr]{eucal}
\usepackage{txfonts}
\usepackage{changepage}
\usepackage{color,soul}

\usepackage{bm}
\usepackage{amsfonts}
\usepackage{subcaption}

\newcolumntype{L}{>{\centering\arraybackslash}m{0.96\linewidth}}

\usepackage{algorithmic}
\usepackage{pgfplots}
\usepackage{rotating}

\pgfplotsset{compat=1.14}

\newcommand{\claimNum}{43837}
\newcommand{\docNum}{257982}

\usepackage{cleveref}       
\usepackage{nicefrac}       
\usepackage{microtype}

\newcolumntype{P}[1]{>{\centering\arraybackslash}p{#1}}

\usepackage{afterpage}
\usepackage{wrapfig}
\usepackage{float,pdflscape}
\usepackage{tablefootnote}
\def\hyphenateAndTtWholeString #1{\xHyphenate#1$\wholeString\unskip}
\def\xHyphenate#1#2\wholeString {\if#1$%
    \else\transform{#1}%
    \takeTheRest#2\ofTheString\fi}

\def\takeTheRest#1\ofTheString\fi
{\fi \xHyphenate#1\wholeString}

\def\transform#1{\url{#1}\hskip 0pt plus 1pt}

\def\urlx #1{\href{#1}{\hyphenateAndTtWholeString{#1}}}
\usetikzlibrary{calc}
\usetikzlibrary{decorations.pathmorphing}
\definecolor{crandom}{HTML}{0173B2}
\definecolor{cshap}{HTML}{DE8F05}
\definecolor{clime}{HTML}{029E73}
\definecolor{cocc}{HTML}{D55E00}
\definecolor{csalm}{HTML}{CC78BC}
\definecolor{csall2}{HTML}{CA9161}
\definecolor{cinputm}{HTML}{FBAFE4}
\definecolor{cinputl2}{HTML}{949494}
\definecolor{cguidedm}{HTML}{ECE133}
\definecolor{cguidedl2}{HTML}{56B4E9}

\definecolor{myblue}{RGB}{204, 229, 255}
\definecolor{myred}{RGB}{255, 205, 205}
\definecolor{myyellow}{RGB}{253, 253, 153}
\definecolor{mygreen}{RGB}{179, 253, 179}
\DeclareRobustCommand{\hlred}[1]{{\sethlcolor{myred}\hl{#1}}}
\DeclareRobustCommand{\hlblue}[1]{{\sethlcolor{myblue}\hl{#1}}}
\DeclareRobustCommand{\hlyellow}[1]{{\sethlcolor{myyellow}\hl{#1}}}
\DeclareRobustCommand{\hlviolet}[1]{{\sethlcolor{mygreen}\hl{#1}}}

\newcommand{\salmean}{$\textit{Saliency}^{\mu}$}
\newcommand{\salnorm}{$\textit{Saliency}^{\ell2}$}
\newcommand{\inputxmean}{$\textit{InputXGrad}^{\mu}$}
\newcommand{\inputxnorm}{$\textit{InputXGrad}^{\ell2}$}
\newcommand{\guidedmean}{$\textit{GuidedBP}^{\mu}$}
\newcommand{\guidednorm}{$\textit{GuidedBP}^{\ell2}$}
\newcommand{\occlusion}{\textit{Occlusion}}
\newcommand{\shapsamp}{\textit{ShapSampl}}
\newcommand{\lime}{\textit{LIME}}
\newcommand{\rand}{\textit{Random}}

\newcommand{\trans}{$\mathtt{Transformer}$}
\newcommand{\transrand}{$\mathtt{Transformer^{RI}}$}
\newcommand{\cnn}{$\mathtt{CNN}$}
\newcommand{\cnnrand}{$\mathtt{CNN^{RI}}$}
\newcommand{\lstm}{$\mathtt{LSTM}$}
\newcommand{\lstmrand}{$\mathtt{LSTM^{RI}}$}

\newcommand{\salscores}{$\omega_{x_i,c}^M$}
\newcommand{\property}[0]{diagnostic property}
\newcommand{\propertyplural}[0]{diagnostic properties}
\newcommand{\salmap}[0]{\ensuremath{\textrm{SD}}}

\definecolor{bad_res}{HTML}{800000}
\definecolor{gr}{HTML}{18109B}

\usepackage{lscape}

\usepackage{breqn}

\def \GreyLogo   {Images/ucph.pdf}

\subject{ 
  \vspace{3.5cm}
  \scriptsize{A thesis presented to the Faculty of Science in partial fulfillment of the requirements for the degree of} \\
  \Large{Doctor Scientiarum (Dr. Scient.)} \\}

\title{
  \Huge{Towards Explainable Fact Checking}\\
}

\author{
  \Large{Isabelle Augenstein} \\
  \texttt{augenstein@di.ku.dk}
}
\date{January 2021}

\begin{document} 
\pagenumbering{roman}
\AddToShipoutPicture*{\put(0,0){\includegraphics*{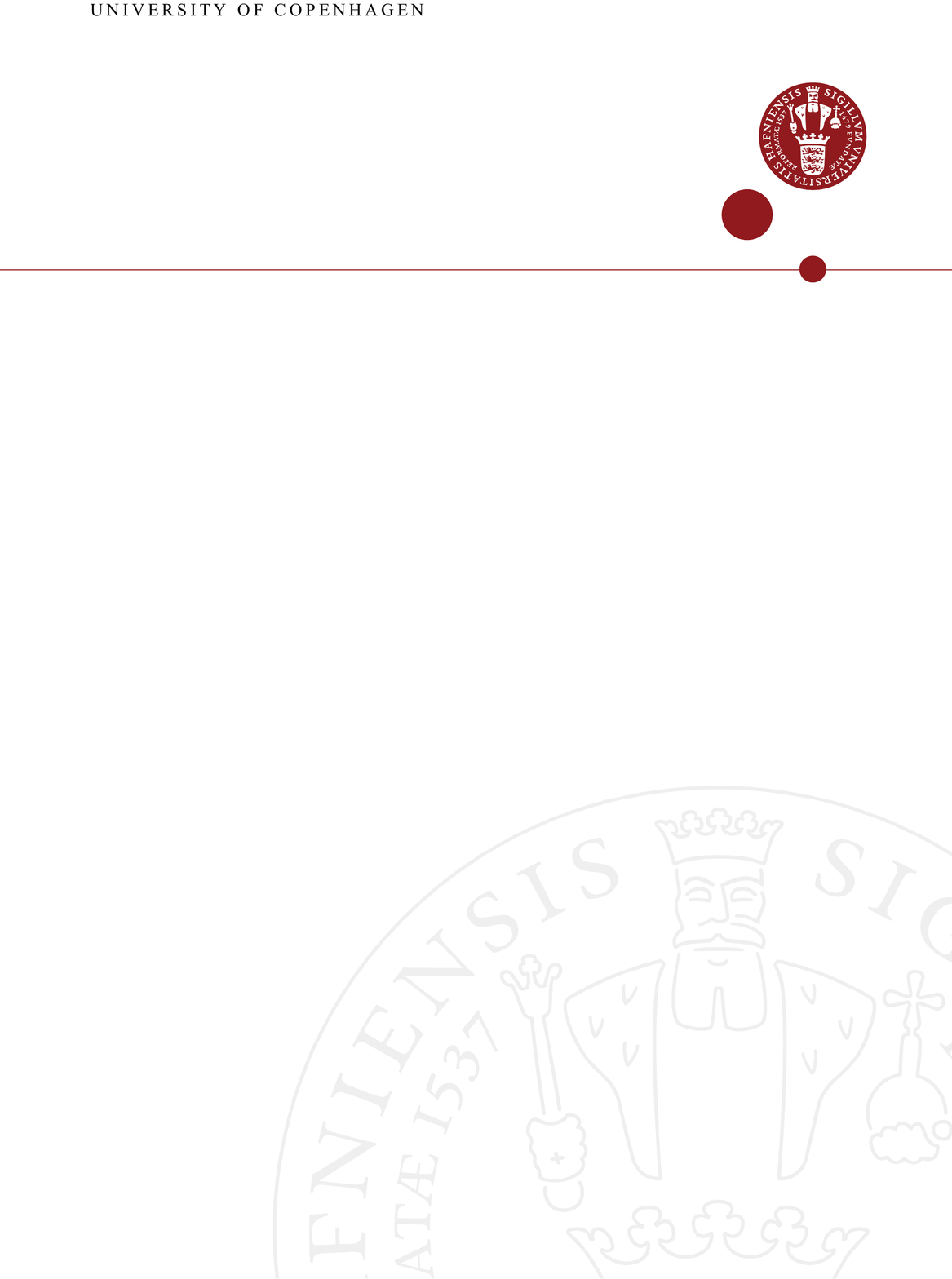}}}  


  \usetikzlibrary{shapes,shadows}
  \tikzstyle{abstractbox} = [draw=black, fill=white, rectangle, 
  inner sep=15pt, style=rounded corners, drop shadow={fill=black,
  opacity=1}]
  \tikzstyle{abstracttitle} =[fill=white]
 
  \newcommand{\boxabstract}[2][fill=white]{
    \begin{center}
      \begin{tikzpicture}
        \node [abstractbox, #1] (box)
        {\begin{minipage}{0.80\linewidth}
            \normalsize #2
          \end{minipage}};
        \node[abstracttitle, right=6pt] at (box.north west) {ABSTRACT};
      \end{tikzpicture}
    \end{center}
  }


\clearpage\maketitle
\thispagestyle{empty}

\newpage

\pdfbookmark[1]{Preface}{preface}
\chapter*{Preface}

This document is a \textit{doktordisputats} - a thesis within the Danish academic system required to obtain the degree of \textit{Doctor Scientiarum}, in form and function equivalent to the French and German Habilitation and the Higher Doctorate of the Commonwealth.

This thesis documents and discusses my research in the field of content-based automatic fact checking, conducted in the period from 2016 to 2020. The first part of the thesis offers an executive summary, which provides a brief introduction to the field of fact checking for a broader computer science audience; a summary of this thesis' contributions to science; a positioning of the contributions of this thesis in the broader research landscape of content-based automatic fact checking; and finally, perspectives for future work.

As this thesis is a \textit{cumulative} one as opposed to a monograph, the remainder of this document contains in total 10 technical sections organised in 3 technical chapters, which are reformatted versions of previously published papers. While the chapters are all self-contained, they are arranged to follow the logical steps of a fact checking pipeline. Moreover, the methodology presented in Section 8 builds on Section 6, which in turn builds on Section 4. Thus, it is advised to read the chapters in the order they occur in.
\pdfbookmark[1]{Acknowledgements}{acknowledgements}
\chapter*{Acknowledgements}

This doktordisputats documents research within the area of Applied Computer Science, more specifically within Natural Language Processing. As such, to get from having an idea, including a detailed plan for the experimental setup, to results, which can be interpreted and written up in a paper, is a long and winding road, involving writing and debugging code, and starting and monitoring experiments on the computing cluster. I could therefore not have produced this thesis on my own, while also fulfilling my duties as a faculty member at the University of Copenhagen. I am therefore immensely grateful to all the early-career researchers for choosing to work with me and being inspired by my research vision.

In a similar vein, this research would not have been possible without external research funding. I should therefore like to give my sincere thanks to the funding programmes of the European Commission, which have funded most of the research in this thesis. The very first project I worked on related to fact checking was funded by the Commission, namely the PHEME FP7 project (grant No. 611233). Later on, two of my PhD students were supported by the Marie Sk\l{}odowska-Curie grant agreement No 801199. Moreover, two of my collaborators for research documented in this thesis were supported by EU grants, namely ERC Starting Grant Number 313695 and QUARTZ (721321, EU H2020 MSCA-ITN).

Next, I would like to thank my external collaborators and mentors over the years who supported and guided me. Working out how to succeed in academia is anything but easy, and would be even harder without positive role models and fruitful collaborations.

Last but not least, I would like to thank my partner, Barry, for his unwavering support, having read and provided useful comments on more of my academic writing than anyone else, including on this thesis.
\pdfbookmark[1]{Abstract}{Abstract}
\chapter*{Abstract}

The past decade has seen a substantial rise in the amount of mis- and disinformation online, from targeted disinformation campaigns to influence politics, to the unintentional spreading of misinformation about public health. This development has spurred research in the area of automatic fact checking, from approaches to detect check-worthy claims and determining the stance of tweets towards claims, to methods to determine the veracity of claims given evidence documents.

These automatic methods are often content-based, using natural language processing methods, which in turn utilise deep neural networks to learn higher-order features from text in order to make predictions. As deep neural networks are black-box models, their inner workings cannot be easily explained. At the same time, it is desirable to explain how they arrive at certain decisions, especially if they are to be used for decision making. While this has been known for some time, the issues this raises have been exacerbated by models increasing in size, and by EU legislation requiring models to be used for decision making to provide explanations, and, very recently, by legislation requiring online platforms operating in the EU to provide transparent reporting on their services. Despite this, current solutions for explainability are still lacking in the area of fact checking.

A further general requirement of such deep learning based method is that they require large amounts of in-domain training data to produce reliable explanations. As automatic fact checking is a very recently introduced research area, there are few sufficiently large datasets. As such, research on how to learn from limited amounts of training data, such as how to adapt to unseen domains, is needed.

This thesis presents my research on automatic fact checking, including claim check-worthiness detection, stance detection and veracity prediction. Its contributions go beyond fact checking, with the thesis proposing more general machine learning solutions for natural language processing in the area of learning with limited labelled data. Finally, the thesis presents some first solutions for explainable fact checking.

Even so, the contributions presented here are only a start on the journey towards what is possible and needed. Future research should focus on more holistic explanations by combining instance- and model-based approaches, by developing large datasets for training models to generate explanations, and by collective intelligence and active learning approaches for using explainable fact checking models to support decision making.
\pdfbookmark[1]{Resume}{resume}
\chapter*{Resume}

I det forløbne årti har der været en betydelig stigning i mængden af mis- og desinformation online, fra målrettede desinformationskampagner til at påvirke politik til utilsigtet spredning af misinformation om folkesundhed. Denne udvikling har ansporet forskning inden for automatisk faktatjek, fra tilgange til at opdage kontrolværdige påstande og bestemmelse af tweets holdning til påstande til metoder til at bestemme rigtigheden af påstande, givet bevisdokumenter.  

Disse automatiske metoder er ofte indholdsbaserede ved hjælp af naturlige sprogbehandlingsmetoder, som bruger dybe neurale netværk til at lære abstraktioner i data på højt niveau til klassificering. Da dybe neurale netværk er `black-box'-modeller, er der ikke direkte indsigt i, hvorfor modellerne når frem til deres forudsigelser. Samtidig er det ønskeligt at forklare, hvordan de når frem til bestemte forudsigelser, især hvis de skal benyttes til at træffe beslutninger. 
Selv om dette har været kendt i nogen tid, er problemerne, som dette rejser, blevet forværret af modeller, der stiger i størrelse, og af EU-lovgivning, der kræver, at modeller bruges til beslutningstagning for at give forklaringer, og for nylig af lovgivning, der kræver online platforme, der opererer i EU til at levere gennemsigtig rapportering om deres tjenester. På trods af dette mangler de nuværende løsninger til forklarlighed stadig inden for faktatjek.  

Et yderligere generelt krav til en sådan `deep learning’ baseret metode er, at de kræver store mængder af i domæne træningsdata for at producere pålidelige forklaringer. Da automatisk faktatjek er et meget nyligt introduceret forskningsområde, er der få tilstrækkeligt store datasæt. Derfor er der behov for forskning i, hvordan man lærer af begrænsede mængder træningsdata, såsom hvordan man tilpasser sig til usete domæner.  

Denne doktordisputats præsenterer min forskning om automatisk faktatjek, herunder opdagelse af påstande, og om de bør tjekkes (`claim check-worthiness detection’), opdagelse af holdninger (`stance detection’) og forudsigelse af sandhed (`veracity prediction’). Dens bidrag går ud over faktatjek, idet afhandlingen foreslår mere generelle maskinindlæringsløsninger til naturlig sprogbehandling inden for læring med begrænsede mærkede data. Endelig præsenterer denne doktordisputats nogle første løsninger til forklarlig faktatjek.  

Alligevel er bidragene, der præsenteres her, kun en start på rejsen mod hvad der er muligt og nødvendigt. Fremtidig forskning bør fokusere på mere holistiske forklaringer ved at kombinere instans- og modelbaserede tilgange, ved at udvikle store datasæt til træningsmodeller til generering af forklaringer og ved kollektiv intelligens og aktive læringsmetoder til brug af forklarlige faktatjekmodeller til støtte for beslutningstagning. 

\pdfbookmark[1]{Publications}{publications}
\chapter*{Publications}

The following published papers are included in the text of this dissertation, listed in the order of their appearance:

\begin{enumerate}
\item \fullcite{wright-augenstein-2020-claim}
\item \fullcite{wright-augenstein-2020-transformer}
\item \fullcite{augenstein-etal-2016-stance}\\An earlier version of this work appeared as:
    \begin{itemize}
    \item \fullcite{augenstein-etal-2016-usfd}
    \end{itemize}
\item \fullcite{journals/ipm/ZubiagaKLPLBCA18}\\An earlier version of this work appeared as:
    \begin{itemize}
    \item \fullcite{kochkina-etal-2017-turing}
    \end{itemize}
\item \fullcite{augenstein-etal-2018-multi}
\item \fullcite{Bjerva_Kouw_Augenstein_2020} 
\item \fullcite{atanasova-etal-2020-diagnostic}
\item \fullcite{augenstein-etal-2019-multifc}
\item \fullcite{atanasova-etal-2020-generating}
\item \fullcite{atanasova-etal-2020-generating-fact}
\end{enumerate}

The papers listed below were published concurrently with my research on content-based automatic fact checking, and are unrelated or marginally related to the topic of this dissertation. Therefore, they are not included in it. They are listed here to indicate the broadness of my research and to contextualise the research presented in this dissertation. In chronological order, those papers are:

\begin{enumerate}
\item \fullcite{derczynski-etal-2015-usfd} 
\item \fullcite{lendvai-etal-2016-monolingual} 
\item \fullcite{spithourakis-etal-2016-numerically} 
\item \fullcite{eisner-etal-2016-emoji2vec} 
\item \fullcite{series/synthesis/2016Maynard}
\item \fullcite{augenstein-sogaard-2017-multi}
\item \fullcite{collins-etal-2017-supervised} 
\item \fullcite{augenstein-etal-2017-semeval} 
\item \fullcite{journals/semweb/ZhangGBAC17} 
\item \fullcite{bjerva-augenstein-2018-tracking} 
\item \fullcite{bjerva-augenstein-2018-phonology} 
\item \fullcite{weissenborn-etal-2018-jack} 
\item \fullcite{kann-etal-2018-character} 
\item \fullcite{nyegaard-signori-etal-2018-ku} 
\item \fullcite{ws-2018-representation} 
\item \fullcite{kementchedjhieva-etal-2018-copenhagen} 
\item \fullcite{de-lhoneux-etal-2018-parameter} 
\item \fullcite{gonzalez-etal-2018-strong} 
\item \fullcite{sogaard-etal-2018-nightmare} 
\item \fullcite{bjerva-etal-2019-probabilistic} 
\item \fullcite{hartmann-etal-2019-issue} 
\item \fullcite{hoyle-etal-2019-combining} 
\item \fullcite{bjerva-etal-2019-language} 
\item \fullcite{conf/aaai/RuderBAS19} 
\item \fullcite{bjerva-etal-2019-uncovering} 
\item \fullcite{hoyle-etal-2019-unsupervised} 
\item \fullcite{ws-2019-representation} 
\item \fullcite{gonzalez2019retrieval} 
\item \fullcite{bingel2019domain} 
\item \fullcite{abdou-etal-2019-x} 
\item \fullcite{bjerva-etal-2019-transductive} 
\item \fullcite{hartmann-etal-2019-mapping} 
\item \fullcite{pmlr-v124-rethmeier20a} 
\item \fullcite{10.1145/3366424.3383758} 
\item \fullcite{a-augenstein-2020-2kenize}
\item \fullcite{bjerva-etal-2020-sigtyp} 
\item \fullcite{muttenthaler-etal-2020-unsupervised}
\item \fullcite{rogers-augenstein-2020-improve}
\item \fullcite{bjerva-etal-2020-subjqa}
\item \fullcite{nooralahzadeh-etal-2020-zero}

\end{enumerate}

Finally, the following are, at the time of writing, unpublished manuscripts available on pre-print servers, which have been written concurrently with the research papers included in this dissertation. Some of them are strongly related to the topic of this dissertation, whereas others are not. As they are not yet accepted for publication, they do not fulfill the formal critera for inclusion into a doktordisputats at the Faculty of Science, University of Copenhagen. They are listed here to demonstrate my continued research efforts in the area of content-based automatic fact checking and related fields.

\begin{enumerate}
    \item \fullcite{journals/corr/RiedelASR17} 
    \item \fullcite{journals/corr/abs-1911-08782} 
    \item \fullcite{waseem2020disembodied} 
    \item \fullcite{journals/corr/abs-2010-01061} 
    \item \fullcite{journals/corr/abs-2009-06401} 
    \item \fullcite{journals/corr/abs-2009-06402} 
    \item \fullcite{journals/corr/abs-2008-09112} 
    \item \fullcite{journals/corr/abs-2012-05742} 
    \item \fullcite{journals/corr/abs-2012-05776} 
\end{enumerate}
\pdfbookmark[1]{\contentsname}{tableofcontents}

\setcounter{tocdepth}{2} 
\setcounter{secnumdepth}{3} 

\tableofcontents 


\pdfbookmark[1]{\listfigurename}{lof}
\listoffigures

\pdfbookmark[1]{\listtablename}{lot}
\listoftables
  
   

\pagenumbering{arabic}

\part{Executive Summary}\label{part:I}
\pdfbookmark[1]{Executive Summary}{executive summary}
\chapter{Executive Summary}\label{ch:summary}

\section{Introduction}


False information online is one of the greatest problems of the past decade, both from a societal and an individual decision making perspective. Among others, deliberate spreading of false information (\textit{disinformation}) is being used to influence elections across the world (\cite{Bovet2019,juhasz2017political,howard2016bots,derczynski2019misinformation,ncube2019digital}), and accidental spreading of false information (\textit{misinformation}) about public health in the wake of COVID-19 has led to what has been coined an `infodemic' (\cite{cuan2020misinformation,kouzy2020coronavirus,brennen2020types}).

While manual journalistic efforts to curb false information are crucial (\cite{waisbord2018truth}), they cannot scale to fact-checking hundreds of millions of daily Twitter posts.
Consequently, detecting false information has become an important task to automate.

What follows hereafter is an introduction to the topic of automatic fact checking, covering fact checking sub-tasks, machine learning methods, as well as explainability methods. Where appropriate, this section cross-references the papers contained in the later methodological chapters. A more detailed introduction to the contributions of each paper can be subsequently found in Section \ref{sec:summary:contrib}.

\subsection{Automatic Fact Checking}\label{sec:summary:fc}

One typically differentiates automatic fact checking approaches by whether they are based on network or content information; we consider each of these in turn below.

\subsubsection{Network-Based Approaches}

Network-based approaches aim to identify false information purely based on interactions between people on e.g. social media platforms. Each social media user is connected to others in different ways, e.g. through following their posts, being followed or quoting users in posts. All these interactions together then form networks. Research has found that users in the same such networks often share common beliefs and, crucially here, that users spreading disinformation often tend to part of the same such networks.

\begin{figure}
\centering
\resizebox{\linewidth}{!}{
\centering
\includegraphics[]{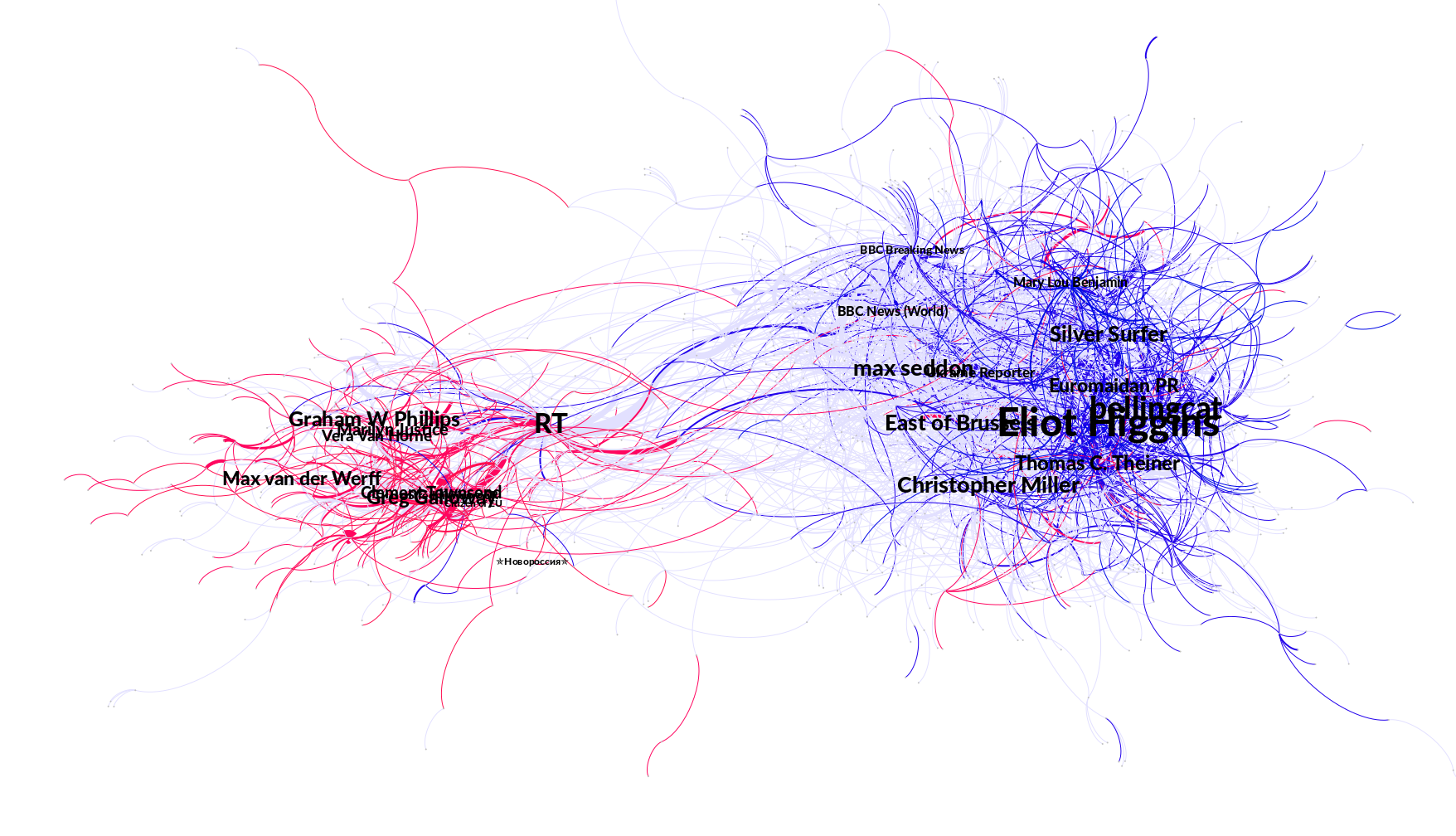} 
}
\caption{This plot shows the joint retweet network from \cite{hartmann-etal-2019-mapping}. Pro-Russian edges are colored in red, pro-Ukrainian edges are colored in dark blue and neutral edges are colored in grey. Both plots were made using The Force Atlas 2 layout in gephi (\cite{gephi}).}\label{f:orig_k10}
\end{figure}

To illustrate this, please find an example from a paper of mine (\cite{hartmann-etal-2019-mapping}) in Figure \ref{f:orig_k10}. The use case here is the visualisation of tweets around the crash of Malaysian Airlines (MH17) flight on 17 July 2014, on its way from Amsterdam to Kuala Lumpur over Ukrainian territory, resulting in the death of 298 civilians. The crash resulted in competing narratives around who was to blame -- whether the plane was shot down by the Ukrainian military, or by Russian separatists in Ukraine supported by the Russian government (\cite{oates2016}). The latter theory is the one subsequently confirmed by an international investigations team. 
The figure shows a retweet network, a graph that contains users as nodes and an edge between two users if at least one of the users retweets the other in a tweet that is detected to be on topic. Each edge is semi-automatically labelled as either pro-Russian, pro-Ukrainian or neutral, depending on the prevailing polarity of the content of the tweet being retweeted between users. As can be seen, the two networks corresponding to the two competing narratives are largely disconnected, i.e. they are being put forward by different groups of people. Based on such an analysis, disinformation can easily be detected, e.g. by flagging users\footnote{At this point, it might be worth mentioning that not all such users are real people -- many of them are bots. Bot detection, though, is beyond the scope of this thesis.} who are part of disinformation networks.

\subsubsection{Content-Based Approaches}

\begin{figure}
\centering
\resizebox{0.8\linewidth}{!}{
\centering
\includegraphics[]{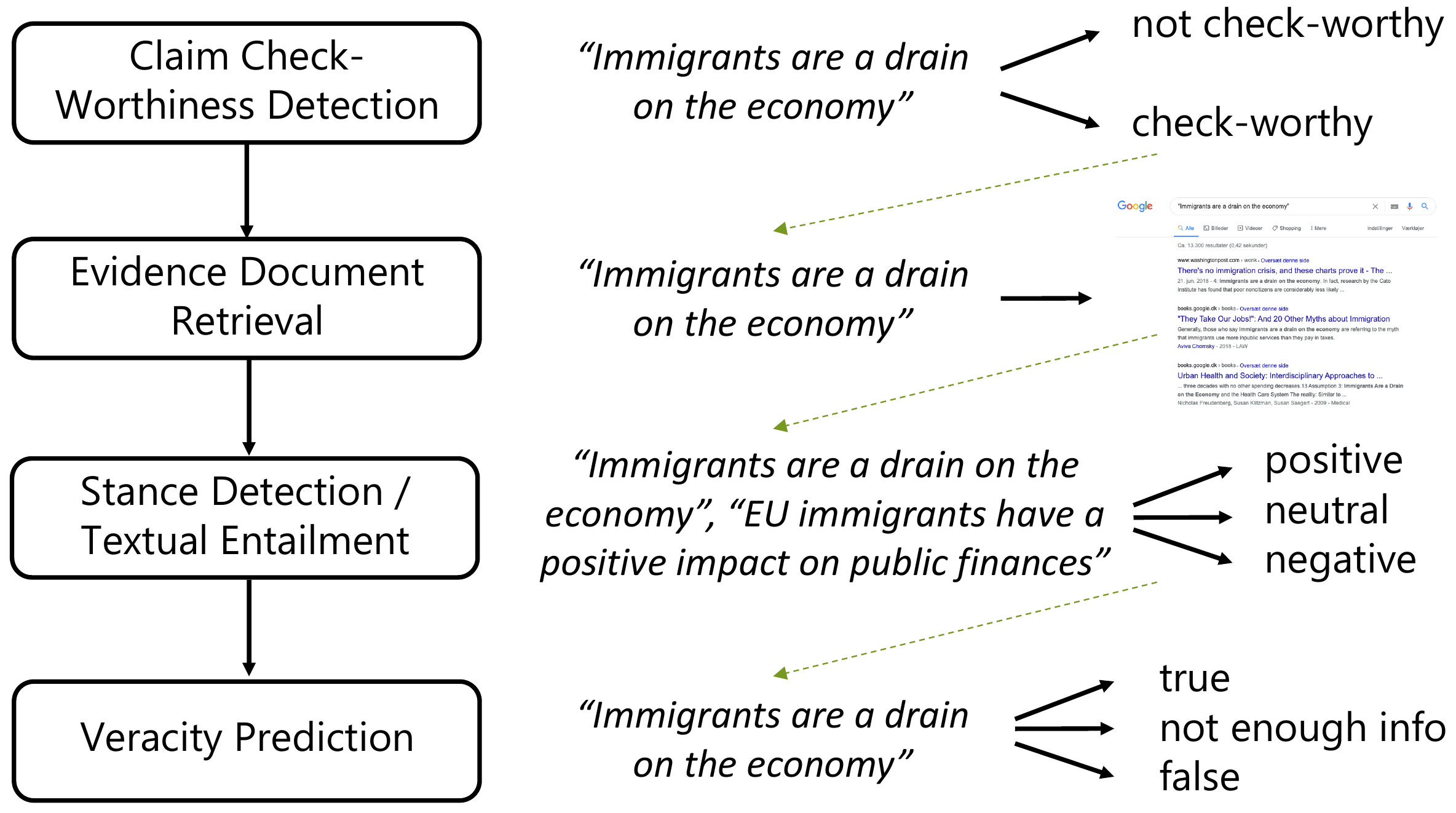} 
}
\caption{This illustrates a typical content-based fact checking pipeline, starting with the detection of check-worthy claims, and ending with the verification of a claim's veracity.}\label{fig:fc_pipeline}
\end{figure}

A content-based automatic fact checking pipeline can be decomposed into the following tasks:
\begin{enumerate}[noitemsep]
    \item Detection of check-worthy claims;
    \item Evidence retrieval and ranking;
    \item Stance detection;
    \item Veracity prediction.
\end{enumerate}

\subsubsection{Detection of check-worthy claims}
An example of how this works for a given input is shown in Figure \ref{fig:fc_pipeline}. Assume the statement `Immigrants are a drain on the economy' is given. The first step would then be to determine if this statement constitutes a claim or an opinion and, if it constitutes a claim, whether or not that claim is worth fact checking (\textit{claim check-worthiness detection}). Whether or not a claim is determined to be worth fact-checking is influenced by claim importance, which is subjective. Typically, only sentences unlikely to be believed without verification are marked as check-worthy. Furthermore, domain interests skew what is deemed check-worthy, which might e.g. only be certain political claims, or only celebrity gossip, depending on the application at hand.
Lastly, claims can be very different in nature (\cite{francis2016fast}). The main categories of claims are: 1) numerical claims involving numerical properties about entities and comparisons among them; 2) entity and event properties such as professional qualifications and event participants; 3) position statements such as whether a political entity supported a certain policy; 4) quote verification assessing whether the claim states precisely the source of a quote, its content, and the event at which it supposedly occurred.
All of this already makes the first step in the fact checking pipeline, the detection of check-worthy claims, a surprisingly non-trivial task. A more in-depth discussion of these challenges can be found in Paper 1 (Section \ref{ch:pu_learning}).

\subsubsection{Evidence retrieval and ranking}
Following this, if the statement indeed is a check-worthy claim, evidence documents which can be used to confirm or refute the claim are retrieved from the Web and ranked by their relevance to the claim (\textit{evidence retrieval and ranking}). Note that the source of documents for retrieval can be restricted to certain domains, e.g. Wikipedia only, as done for the FEVER shared task and dataset (\cite{thorne-etal-2018-fever}), the setup of which is used in Paper 9, contained in Section \ref{ch:adversarial}. Alternatively, the whole Web can be used as a source, in which case a search engine is often used in this part of the pipeline, as done for the MultiFC dataset presented in Paper 8, Section \ref{ch:multifc}. Moreover, if the statement is a social media post, replies to this post can be used as evidence documents, as done in RumourEval \cite{derczynski2017semeval,rumour:19}, the setting of which is used in Papers 4 (Section \ref{ch:discourse} and 6 (Section \ref{ch:sequential}). For simplicity, some experimental settings skip this step altogether, assuming all relevant evidence is identified manually, as done in Paper 10 (Section \ref{ch:fc_explanations}). A less stark simplification, but a simplification nonetheless, is to assume all relevant evidence documents are provided, and merely have to be reranked, which is e.g. explored in \cite{journals/corr/abs-2009-06402} (which has not been included in this thesis).

\subsubsection{Stance detection}
The next step is to determine if the evidence documents retrieved in the previous step agree with, disagree with or are neutral towards the claim. This step is typically referred to as either \textit{stance detection} or \textit{textual entailment}. More generally, the task can be called \textit{pairwise sequence classification}, where a label corresponding to a pair of sequences is learned. The first sequence can be a claim, a headline, a premise, a hypothesis, or a target such as a person or a topic.
There are many different labelling schemes for this task, ranging from simply `positive' vs. `negative' to a very fine-grained task with many different labels. This challenge is explored in more detail in Paper 5 (Section \ref{ch:disparate}). Intuitively, this step is easier the more directly an evidence document discusses a claim -- the less textual overlap there is between the two texts, the harder it to determine the stance automatically -- which is discussed in Papers 3 (Section \ref{ch:conditional}) and 8 (Section \ref{ch:multifc}).

\subsubsection{Veracity prediction}
Lastly, given the predicted labels from the stance classifier for $<$claim, evidence$>$ pairs obtained in Step 3, as well as a ranking of evidence pages obtained in Step 2, the overall veracity of the claim in question can be determined. As in stance detection, there exist many different labelling schemes for this step. In addition to fine-grained judgements about where on the scale from `completely true' to `completely false' a claim lies, labels such as `not enough information' or `spins the truth' are used. Solutions to this are discussed in great detail in Paper 8 (Section \ref{ch:multifc}).

It should be noted at this point that what automatic fact checking models aim to predict is not the irrefutable, objective truth, but rather the veracity of a claim with respect to certain evidence documents. If there exists counter evidence, or important evidence is missing, but they are not available to the fact checking model in a machine-readable format, the model has no way of telling this. Thus, it is up to the person utilising the fact checking model to carefully examine how the model has arrived at its prediction. Supporting humans in this endeavour is possible through models which, as is the goal explored in this thesis, produce an explanation for the automatic fact-check. This is further described in Section \ref{sec:summary:xai}.

An intuition behind some of the key machine learning challenges with building such automatic fact checking methods and how they are addressed in this thesis is given next.


\subsection{Learning with Limited and Heterogeneous Labelled Data}\label{sec:summary:lld}

A core methodological challenge addressed throughout this thesis is that most fact checking datasets are small in nature and apply different labelling schemes. Learning stable automatic fact checking models from those datasets thus presents a significant challenge.

This thesis proposes methods within the following areas to tackle this problem:
\begin{enumerate}[noitemsep]
    \item Output space modelling;
    \item Multi-task learning;
    \item Transfer learning.
\end{enumerate}

Automatic fact checking is, at its core, a classification task: each claim has to be categorised as belonging to one of a set of classes denoting the claim's veracity. Hence, methods popular for text classification ought to be suitable for automatic fact checking as well.
On a high level, standard text classification models take the input text (e.g. a claim, perhaps concatenated with an evidence sentence), transform it to higher-level abstract features using a feature transformation function, and output a class label.


\subsubsection{Output space modelling}
While this works well for most purposes, this formulation has several shortcomings, which can be addressed by different ways of \textit{modelling the output space}. First of all, with classes simply being represented by IDs, the inherent meaning of these classes is not directly represented by the model. Moreover, in fact checking, different veracity labels represent fine-grained nuances in meaning, e.g. `partly true', `mostly true', 'more evidence needed', i.e. there is a relationship between labels. Thus, this thesis proposes to not just learn input features, but also output features for fact checking, more concretely, label embeddings, which capture the semantic relationships between labels. Papers 5 (Section \ref{ch:disparate}) and 8 (Section \ref{ch:multifc}) show that this trick can be applied to not just improve the performance of pairwise sequence classification and veracity prediction models, respectively, but also to unify label schemes from different datasets, without having to attempt to do this unification manually.


Moreover, Paper 3 (Section \ref{ch:conditional}) applies a similar idea to the task of unseen target stance detection -- i.e. determining the stance towards a target (e.g. a person or topic) that is unseen at test time. Prior to that, ordinarily, approaches would consist of one model per stance target. By representing the targets as features as well and training a joint model, the resulting model can make stance predictions for any target, a formulation which resulted in state-of-the-art performance at the time.

Lastly, another form of output space modelling investigated in Paper 4 (Section \ref{ch:discourse}) concerns the context of the target instance. As mentioned before, one task formulation of fact checking is detecting the veracity of rumours, where a rumour is verified by determining the stance of social media posts towards a rumour. As social media posts do not appear in isolation but, more often than not, are replies to previous posts, modelling them in isolation would mean losing out on important context. Therefore, this thesis proposes to frame rumour stance detection as a structured prediction task. The conversational structure on Twitter can be viewed as a tree, in turn consisting of branches that are made up of sequences of posts. We propose to treat each such branch as an instance. The labels for posts are then predicted in a sequential fashion, such that the model which makes these predictions has access to the posts as well as the predicted labels in the same conversational thread. 


\subsubsection{Multi-task learning}
Different label schemes having been dealt with, another challenge concerns the generally low number of training instances of fact checking datasets -- especially the earlier datasets such as \cite{mohammad-etal-2016-semeval,PomerleauRao,derczynski2017semeval,P17-2067} only contain around a couple of thousand training instances. Deep learning based text classification models tend to perform better the more examples they have available at training time. Moreover, with a paucity of training instances, they tend to struggle to even outperform simply predicting a random label or the most frequent label (see e.g. \cite{hartmann-etal-2019-mapping}).
Two streams of research are explored in this thesis to tackle this problem -- multi-task and transfer learning -- both of which build on the general idea of obtaining more training data from other sources.

\textit{Multi-task learning} is the idea of, in addition to the \textit{target task}, obtaining training data for so-called \textit{auxiliary tasks}. The latter are tasks which are often related to the target task, be it in form (i.e. if the target task is a text classification task, those would also be text classification tasks); in domain (e.g. the target task and auxiliary task data could all be from the legal domain); or in the nature of the task (e.g. only taking into consideration different variants of sentiment analysis).
The idea is then to train a model on all such tasks at once, but to only utilise the predictions of the target task. 

Deep neural network based text classification models typically consist of an: 1) input layer, which maps inputs to features, 2) hidden layers, which learn more abstract higher-order features; and 3) an output layer, which maps the features from the hidden layer to output labels.
Since inputs as well as the type and number of classes tends to differ by task, in multi-task learning, only the hidden layers tend to be shared between tasks, whereas the input and output layers tend to be kept separate for each task.
Intuitively, sharing the hidden layer means the model can better learn abstract features by having seen more examples. Training on several tasks at the same time also has a regularisation effect -- it is more difficult for a model to overfit to spurious patterns for any one task if it is trained to perform several tasks at the same time. In this thesis, multi-task learning is used as a building block in several papers. In Papers 5 (Section \ref{ch:disparate}) and 8 (Section \ref{ch:multifc}), it is combined with the idea of label embeddings. In Papers 9 (Section \ref{ch:adversarial}) and 10 (Section \ref{ch:fc_explanations}), it is used as a way of specifically instilling different types of knowledge in a model -- about semantic coherence and how to generate adversarial claims (Paper 9), and about veracity and how to generate instance-level explanations (Paper 10).

\subsubsection{Transfer learning}
Finally, this thesis examines \textit{transfer learning} as a way of increasing the amount of training data for a task. Transfer learning can be seen as as special form of multi-task learning where a model is trained on several tasks, but instead of it being trained on several tasks at once, it is trained on several tasks sequentially. The last one of these tasks is typically the \textit{target task}, unless there is no training data available for the target task at all, in which case one typically speaks of \textit{unsupervised} or \textit{zero-shot} transfer learning. The other tasks are typically referred to as the \textit{source tasks}.

As with multi-task learning, the intuition that is typically applied is that the closer the source tasks are to the target task, be it in terms of form, domain, text type, or application task considered, the more likely it is to benefit the target task.
The current de-facto approach to natural language processing at the time of writing is the so-called `pre-train, then fine-tune' approach, where sentence encoding models are pre-trained on unsupervised tasks which require no manual annotation. 
The overall goal is for this pre-training procedure to result in a better initialisation of the hidden layers of a deep neural network, such that when it is fine-tuned for a target task, it converges more quickly and is less likely to get stuck in local minima.
There are many different variants of such pre-training tasks for NLP, such as simple next-word prediction (language modelling), next-sentence prediction or term frequency prediction (\cite{aroca-ouellette-rudzicz-2020-losses}). These pre-trained models, often pre-trained on large amounts of raw data, can then be re-used across different applications in a plug-and-play fashion, and a large number of such pre-trained models have been published in the last year alone \cite{journals/corr/abs-2002-12327}.

This thesis is not concerned with learning better unsupervised representations from large amounts of data as such -- though it does utilise such pre-trained models. Instead, it focuses on how to better fine-tune them for given target tasks. Three different types of fine-tuning settings are explored. Paper 1 (Section \ref{ch:pu_learning}) explores the common setting where relatively little target-task data is available, but noisy training data for the same task, albeit from a different domain, is available. The paper then deals with how best to fine-tune a claim check-worthiness detection model in two steps -- first on the out-of-domain data, then on the in-domain data.
Paper 2 (Section \ref{ch:domain}) assumes no training data is available at all for the target domain. Instead, multiple training datasets for the same task, though from other domains, are available. Thus, the task becomes one of unsupervised multi-source domain adaptation.
Lastly, Paper 6 (Section \ref{ch:sequential}) addresses the problem that language, and thus test data, changes over time. To reflect that change, the paper proposes to perform sequential temporal domain adaptation, where models are adapted to more recent data sequentially, and shows that this improves performance.

\subsection{Explainable Natural Language Understanding}\label{sec:summary:xai}

The last vertical of relevance to fact checking is how to make fact checking models more transparent, such that end users can understand: 1) what a model as a whole has learned (\textit{model interpretabilty} as well as 2) why a model produces a certain prediction for a certain instance (\textit{model explainability}). Note here that the terms `interpretability' and `explainability' are often conflated in the literature, not least because an explainable model is often also interpretable. 

The methods for explainable natural language understanding researched in this thesis can be divided into the following streams of approaches:
\begin{enumerate}[noitemsep]
    \item Generating natural language explanations;
    \item Generating adversarial examples;
    \item Post-hoc explainability methods.
\end{enumerate}

These are considered in turn below.

\subsubsection{Generating natural language explanations}
The inner workings of deep neural networks are, as already mentioned, relatively complex; especially since modern models have too many parameters to inspect them individually. One solution to this is to generate natural language explanations, based on the assumption that the easiest explanations to understand for users are those written in natural language. 

The overall aim to produce free text (typically a sentence or a paragraph) that succinctly explains how the model has arrived at a certain prediction.
Technically, this is achieved by training a model to both solve the main task and generate a textual explanation. In Paper 10 (Section \ref{ch:fc_explanations}), we approach this using multi-task learning.
Ideally, such a free text explanation would be directly generated from a model's hidden layers as an unsupervised task, however, this is extremely challenging to do. In Paper 10, we show that generating free-text explanations for fact checking is possible in a simplified setting. Namely, the model is given long articles written by journalists discussing evidence documents, and the model is trained to summarise those evidence documents, while also predicting the veracity of the corresponding claim. The summaries, in turn, represent justifications for the fact-check and thus an explanation for the respective fact checking label.

\subsubsection{Generating adversarial examples}\label{ch:Intro:Explain:Adversarial}
Another way of interpreting what a model has learned is to try to reveal systemic vulnerabilities of the model. Sometimes, usually because a model is trained on biased and/or small amounts of training data, it learns spurious correlations resulting in features that are red herrings -- which a model has only seen a handful of times at training time, which were always associated with only one label, but which, in reality, are not indicative of the label. An example from the fact checking domain could be if a model is only exposed to false claims mentioning certain people, then it will very likely learn to always predict the veracity of `false' whenever this name occurs in a claim.

The goal of generating adversarial examples is to identify such features, sometimes also called `universal adversarial triggers', and use them to  automatically generate instances which a fact checking model would predict an incorrect label for. This not only tells a user what a model would likely struggle with, but the automatically generated adversarial instances can in turn be used to improve models.

In Paper 10 (Section \ref{ch:adversarial}), we explore how to generate adversarial claims for fact checking. There are some additional challenges when researching this method for fact checking. First, the generated adversarial claims should contain these additional triggers, but without changing the meaning of the claim to the extent that they would change the ground truth label. To build on the example above, a trigger could be a certain name, which would not change the meaning of a claim, but could change what a fact checking model would predict for it. It could also be the word `not', which in most cases would change the underlying meaning.
The more different veracity labels are considered, the more difficult this becomes.
Morever, generating a syntactically valid and semantically coherent claim is also non-trivial. While previous work generated claims from templates, this restricts the range of claim types that can be generated. We instead research how to do this automatically, using large language models, which generate these claims from scratch.

\subsubsection{Post-hoc explainability methods}

Lastly, another explainability technique explored in this thesis is post-hoc explainability. Unlike the approaches presented for generating natural language explanations and discovering universal adversarial triggers, post-hoc explainability methods are methods that can be applied after a model for a certain application task has already been trained.
The general idea is to find regions of the input which best explain the predicted label for the corresponding instance. These regions of the input are typically called `rationales', and are portions of the input text -- words, phrases or sentences -- which are salient for the predicted label given the trained model. 

This can then very easily tell a user if a model focuses on the correct parts of the input or the incorrect parts of the input. Following up from the example above, in Section \ref{ch:Intro:Explain:Adversarial}, an undesired part of the input to focus on for fact checking could be a person's name in a claim, as this could be a reflection of the data selection process more than a reflection of the real world. 

At the time of running experiments, and also of writing this thesis, there are no fact checking datasets annotated with human rationales. Veracity prediction models take not only claims as their input, but also evidence pages. As such, rationales would have to relate portions of claims and evidence pages, requiring a conceptually different annotation scheme than currently used. 
Therefore, Paper 8 (Section \ref{ch:Diagnostic}) focuses only on the subtask of fact checking which determines the relationship between two input texts, i.e. stance detection / natural language inference.

Paper 8 addresses the highly challenging task of automatically evaluating such post-hoc explainability methods. First off, they should of course be in line with human annotations, but these are not always available, and besides, should not be the only thing taken into account -- for instance, an explanation should also be faithful to what the respective model has learned. The paper proposes different so-called `diagnostic properties' for evaluating post-hoc explanations, and uses them to compare different types of post-hoc explainability methods across different classification tasks and datasets. Some of the findings are that different explainability methods produce very different explanations, and that explanations further differ by model. Different explanations can all be correct in their own way, e.g. they might offer true alternative explanations, which is just one of the things that makes explainability research challenging.

\section{Scientific Contributions}\label{sec:summary:contrib}

Having given a general introduction to the topic of fact checking and the core challenges tackled in this thesis, this section now turns to describing the scientific contributions made by this thesis in more depth.

The core scientific contributions of models presented in thesis can be conceptualised along three axes:

\begin{enumerate}[noitemsep]
    \item the fact checking sub-task they address (see Sec. \ref{sec:summary:fc});
    \item the method they present for dealing with limited and heterogeneous limited data (see Sec. \ref{sec:summary:lld});
    \item the explainability method they propose (see Sec. \ref{sec:summary:xai}).
\end{enumerate}

Table \ref{tab:overview_contributions} indicates where along these three axes each of the ten papers that make up this thesis are located.
What follows next is a brief summary of the contributions of each paper, grouped by the `fact checking sub-task' axis, following the structure of this thesis.\footnote{The exception to this is evidence retrieval and reranking. As this is only a contribution in Paper 9, and there does not present the main contribution, this fact checking sub-task neither has its own subsection below, nor its own chapter in the thesis.}

\begin{table}[!t]
\fontsize{10}{10}\selectfont
\centering
\begin{tabular}{c|cccc@{\hspace{4.0\tabcolsep}}|ccc@{\hspace{4.0\tabcolsep}}|cccc@{\hspace{4.0\tabcolsep}}}
\toprule
 & \multicolumn{4}{c}{\bf FC Subtask} & \multicolumn{3}{c}{\bf LLD Method} & \multicolumn{4}{c}{\bf Explainability Method} \\
 & \rot{\bf{Claims}} & \rot{\bf{Evidence}} & \rot{\bf{Stance}} & \rot{\bf{Veracity}} & \rot{\bf{Output space}} & \rot{\bf{Multi-task}} & \rot{\bf{Transfer}} & \rot{\bf{Natural language}} & \rot{\bf{Adversarial}} & \rot{\bf{Post-hoc}} \\

\midrule
\cite{augenstein-etal-2016-stance} &  &  & X &  & X &  &  &  &  &   \\
\cite{journals/ipm/ZubiagaKLPLBCA18} &  &  & X &  & X &  &  &  &  &   \\
\cite{augenstein-etal-2018-multi} &  &  & X &  & X & X &  &  &  &   \\
\cite{augenstein-etal-2019-multifc} &  & X &  & X & X & X &  &  &  &   \\
\cite{Bjerva_Kouw_Augenstein_2020} &  &  & X &  &  &  & X &  &  &   \\
\cite{wright-augenstein-2020-claim} & X &  &  &  &  &  & X &  &  &   \\
\cite{wright-augenstein-2020-transformer} & X &  &  &  &  &  & X &  &  &   \\
\cite{atanasova-etal-2020-generating-fact} &  &  &  & X &  & X  &  & X &  &  \\
\cite{atanasova-etal-2020-diagnostic} &  &  & X &  &  &  &  &   &  & X \\
\cite{atanasova-etal-2020-generating} & X &  &  & X &  & X &   &  & X &  \\

\bottomrule
\end{tabular}
\caption{The ten approaches to automatic content-based fact checking presented in this thesis presented along three axes, representing their  core areas of contribution.}
\label{tab:overview_contributions}
\end{table}

\subsection{Detecting Check-Worthy Claims}

The first fact checking area in which contributions are made concerns the detecting of check-worthy claims.

\subsubsection{Paper 1: Positive Unlabelled Learning}

\begin{figure}[t]
  \centering
    \includegraphics[width=0.6\textwidth]{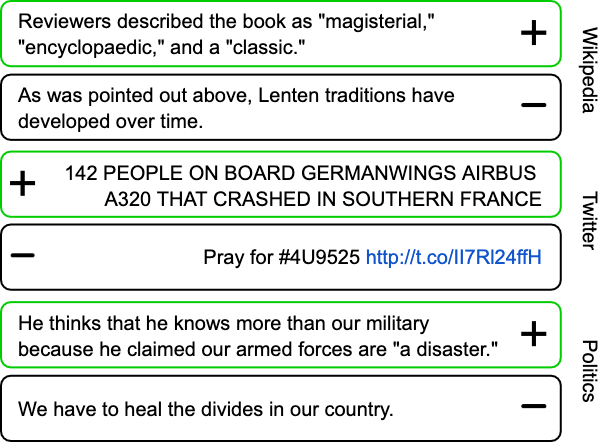}
    \caption{[Paper 1] Examples of check-worthy and non check-worthy statements from three different domains. Check-worthy statements are those which were judged to require evidence or a fact check.}
    \label{fig:paper1}
\end{figure}

This paper addresses the fact checking sub-task of claim check-worthiness detection, a text classification problem where, given a statement, one must predict if the content of that statement makes ``an assertion about the world that is checkable'' (\cite{konstantinovskiy2018towards}).
There are multiple isolated lines of research which have studied variations of this problem: rumour detection on Twitter~(\cite{zubiaga2016analysing,journals/ipm/ZubiagaKLPLBCA18}), check-worthiness ranking in political debates and speeches~(\cite{atanasova2018overview,elsayed2019overview,barron2020checkthat}), and `citation needed' detection on Wikipedia~(\cite{redi2019citation}) (see Figure \ref{fig:paper1}). Each task is concerned with a shared underlying problem: detecting claims which warrant further verification. However, no work has been done to compare all three tasks to understand shared challenges in order to derive shared solutions, which could enable the improvement of claim check-worthiness detection systems across multiple domains. 
Therefore, we ask the following main research question in this work: 
are these (rumour detection on Twitter, check-worthiness ranking in political debates, `citation needed' detection) all variants of the same task (claim check-worthiness detection), and if so, is it possible to have a unified approach to all of them?

In more detail, the contributions of this work are as follows:
\begin{enumerate}[noitemsep]
    \item The first thorough comparison of multiple claim check-worthiness detection tasks.
    \item \textit{Positive Unlabelled Conversion (PUC)}, a novel extension of PU learning to support check-worthiness detection across domains.
    \item Results demonstrating that a unified approach to check-worthiness detection is achievable for 2 out of 3 tasks, improving over the state-of-the-art for those tasks.
\end{enumerate}

\subsubsection{Paper 2: Transformer Based Multi-Source Domain Adaptation}

\begin{figure}[t]
  \centering
    \includegraphics[width=0.7\columnwidth]{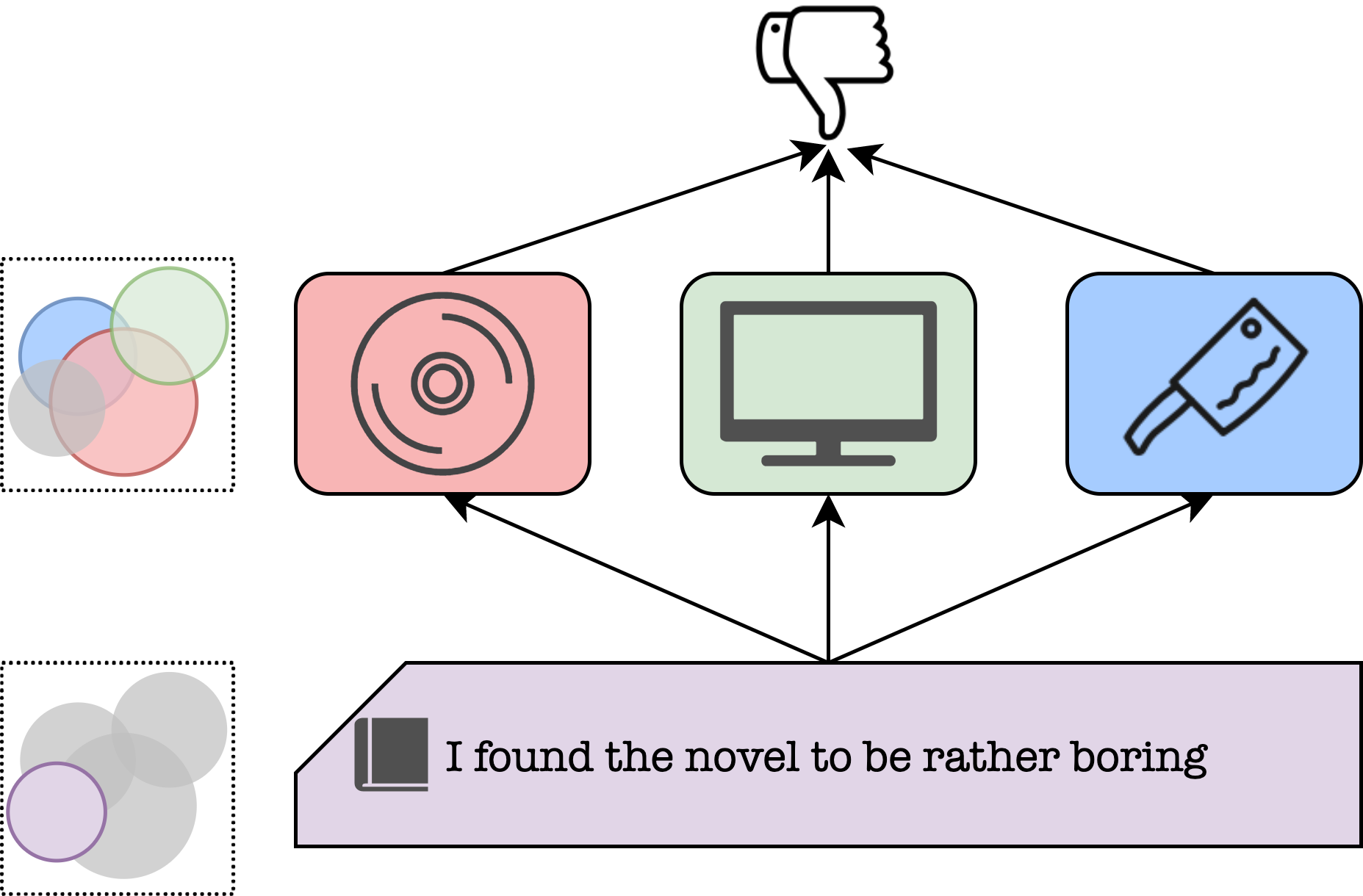}
    \caption{[Paper 2] In multi-source domain adaptation, a model is trained on data drawn from multiple parts of the underlying distribution. At test time, the model must make predictions on data from a potentially non-overlapping part of the distribution.}
    \label{fig:paper2}
\end{figure}

Multi-source domain adaptation is a well studied problem in deep learning for natural language processing. An example of this setting can be found in Figure \ref{fig:paper2}: a model may e.g. be trained to predict the sentiment of product reviews for DVDs, electronics, and kitchen goods, and must utilise this learned knowledge to predict the sentiment of a review about a book. 
Proposed methods have been primarily studied using convolutional nets (CNNs) and recurrent nets (RNNs) trained from scratch, while the NLP community has recently begun to rely more and more on large pretrained transformer (LPX) models e.g. BERT~(\cite{devlin-etal-2019-bert}). 
To date there has been some preliminary investigation of how LPX models perform under domain shift in the single source-single target setting~(\cite{ma2019domain,han2019unsupervised,rietzler2019adapt,gururangan2020don}). What is lacking is a study into the effects of and best ways to apply classic multi-source domain adaptation techniques with LPX models, which can give insight into possible avenues for improved application of these models in settings where there is domain shift.

Given this, Paper 2 presents a study into unsupervised multi-source domain adaptation techniques for large pretrained transformer models, evaluating the proposed models on the tasks of rumour detection on Twitter as well as sentiment analysis of customer reviews. 
Our main research question is: do mixture of experts and domain adversarial training offer any benefit when using LPX models? The answer to this is not immediately obvious, as such models have been shown to generalize quite well across domains and tasks while still learning representations which are not domain invariant. Therefore, we experiment with four mixture of experts models, including one novel technique based on attending to different domain experts; as well as domain adversarial training with gradient reversal. 
Perhaps surprisingly, we find that, while domain adversarial training helps the model learn more domain invariant representations, this does not always result in increased target task performance.
When using mixture-of-experts models, we see significant gains on out-of-domain rumour detection, and some gains on out-of-domain sentiment analysis. Further analysis reveals that the classifiers learned by domain expert models are highly homogeneous, making it challenging to learn a better mixing function than simple averaging.

\subsection{Stance Detection}

The second fact checking area in which contributions are made concerns the classification of stance and textual entailment.

\subsubsection{Paper 3: Bidirectional Conditional Encoding}

\begin{figure*}
\hspace*{-1.5cm}
\begin{tikzpicture}[scale=1.2]
\foreach \i/\l in {1/1,2/2,3/3} {
  \path[draw, thick] (\i-0.2,0) rectangle (\i+0.2,0.75) {};
  \path[draw, thick, fill=nice-red!10] (\i-0.2,1) rectangle (\i+0.2,1.75) {};
  \path[draw, thick, fill=nice-red!20] (\i-0.2,1.75) rectangle (\i+0.2,2.5) {};  
  \path[draw, thick, fill=nice-red!10] (\i-0.2,2.75) rectangle (\i+0.2,3.5) {};
  \path[draw, thick, fill=nice-red!20] (\i-0.2,3.5) rectangle (\i+0.2,4.25) {};
  \draw[->, >=stealth', thick] (\i,0.75) -- (\i,1);
  \draw[->, >=stealth', thick] (\i-0.2,0.4) to[bend left=30] (\i-0.2,2);
  \draw[->, >=stealth', thick] (\i-0.2,2.3) to[bend left=30] (\i-0.2,3.87);
  \draw[->, >=stealth', thick] (\i+0.2,1.55) to[bend right=30] (\i+0.2,3.17);  
  \node[] at (\i,0.4) {\small$\mathbf{x}_\l$};
  \node[] at (\i,1.4) {\small$\mathbf{c}^\rightarrow_\l$};
  \node[] at (\i,2.15) {\small$\mathbf{c}^\leftarrow_\l$};
  \node[] at (\i,3.15) {\small$\mathbf{h}^\rightarrow_\l$};
  \node[] at (\i,3.9) {\small$\mathbf{h}^\leftarrow_\l$};
}
\foreach \i/\l in {5/4,6/5,7/6,8/7,9/8,10/9} {
  \path[draw, thick] (\i-0.2,0) rectangle (\i+0.2,0.75) {};
  \path[draw, thick, fill=nice-blue!10] (\i-0.2,1) rectangle (\i+0.2,1.75) {};
  \path[draw, thick, fill=nice-blue!20] (\i-0.2,1.75) rectangle (\i+0.2,2.5) {};  
  \path[draw, thick, fill=nice-blue!10] (\i-0.2,2.75) rectangle (\i+0.2,3.5) {};
  \path[draw, thick, fill=nice-blue!20] (\i-0.2,3.5) rectangle (\i+0.2,4.25) {};
  \draw[->, >=stealth', thick] (\i,0.75) -- (\i,1);
  \draw[->, >=stealth', thick] (\i-0.2,0.4) to[bend left=30] (\i-0.2,2);
  \draw[->, >=stealth', thick] (\i-0.2,2.3) to[bend left=30] (\i-0.2,3.87);
  \draw[->, >=stealth', thick] (\i+0.2,1.55) to[bend right=30] (\i+0.2,3.17);  
  \node[] at (\i,0.4) {\small$\mathbf{x}_{\l}$};
  \node[] at (\i,1.4) {\small$\mathbf{c}^\rightarrow_{\l}$};
  \node[] at (\i,2.15) {\small$\mathbf{c}^\leftarrow_{\l}$};
  \node[] at (\i,3.15) {\small$\mathbf{h}^\rightarrow_{\l}$};
  \node[] at (\i,3.9) {\small$\mathbf{h}^\leftarrow_{\l}$};
}
\foreach \i in {1,2,5,6,7,8,9} {
  \draw[->, >=stealth', thick] (\i+0.2,1.4) -- (\i+1-0.2,1.4);
  \draw[->, >=stealth', thick] (\i+1-0.2,2.15) -- (\i+0.2,2.15);
}
\foreach \i/\word in {1/Legalization, 2/of, 3/Abortion, 5/A, 6/foetus, 7/has, 8/rights, 9/too, 10/!} {
  \node[anchor=north, text height=1.5ex, text depth=.25ex, yshift=-2em] at (\i, 0.5) {\word};
}

\draw[ultra thick, nice-red] (0.5,-0.75) -- (3.5,-0.75);
\draw[ultra thick, nice-blue] (4.5,-0.75) -- (10.5,-0.75);
\node[anchor=north] at (2,-0.75) {Target};
\node[anchor=north] at (7.5,-0.75) {Tweet};
\draw[->, >=stealth', line width=3pt, color=nice-red, dashed] (3+0.2,1.4) -- (4+1-0.2,1.4);
\draw[->, >=stealth', line width=3pt, color=nice-red, dashed] (1-0.2,2.15) to[bend left=168] (10+0.2,2.15);

\path[draw, ultra thick, color=nice-blue] (5-0.2,3.5) rectangle (5+0.2,4.25) {};
\path[draw, ultra thick, color=nice-blue] (10-0.2,2.75) rectangle (10+0.2,3.5) {};
\path[draw, ultra thick, color=nice-red] (1-0.2,1.75) rectangle (1+0.2,2.5) {};  
\path[draw, ultra thick, color=nice-red] (3-0.2,1) rectangle (3+0.2,1.75) {};
\end{tikzpicture}
\caption{[Paper 3] Bidirectional encoding of tweet conditioned on bidirectional encoding of target ($[\mathbf{c}^\rightarrow_3\;\mathbf{c}^\leftarrow_1]$). The stance is predicted using the last forward and reversed output representations ($[\mathbf{h}^\rightarrow_{9}\;\mathbf{h}^\leftarrow_4]$).}
\label{fig:paper3}
\end{figure*}
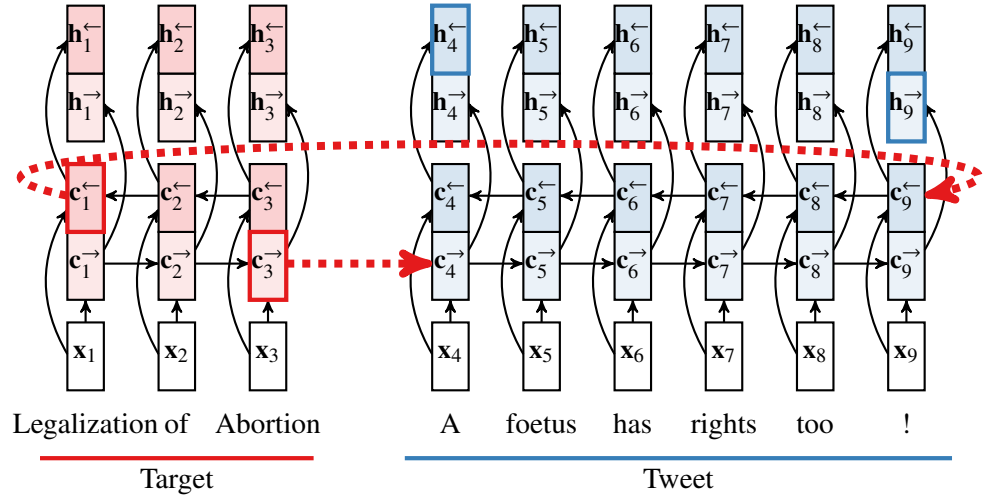

The goal of stance detection is to classify the attitude expressed in a text, towards a given target, here, as ``positive'', ''negative'', or ''neutral''. The focus of this paper is on a novel stance detection task, namely tweet stance detection towards previously unseen target entities (mostly entities such as politicians or issues of public interest), as defined in the SemEval Stance Detection for Twitter task~(\cite{mohammad-etal-2016-semeval}). 
This task is rather difficult, firstly due to not having training data for the targets in the test set, and secondly due to the targets not always being mentioned in the tweet. 
Thus the challenge is twofold. First, we need to learn a model that interprets the tweet stance towards a target that might not be mentioned in the tweet itself. Second, we need to learn such a model without labelled training data for the target with respect to which we are predicting the stance.

To address these challenges, we develop a neural network architecture based on conditional encoding~(\cite{rocktaschel2016reasoning}), visualised in Figure \ref{fig:paper3}. A long-short term memory (LSTM) network~(\cite{hochreiter1997long}) is used to encode the target, followed by a second LSTM that encodes the tweet using the encoding of the target as its initial state. We show that this approach achieves better F1 than standard stance detection baselines, or an independent LSTM encoding of the tweet and the target.
Results improve further with a bidirectional version of our model, which takes into account the context on either side of the word being encoded. In the shared task, this would be the second best result, except for an approach which uses automatically labelled tweets for the test targets. Lastly, when our bidirectional conditional encoding model is trained on such data, it achieves state-of-the-art performance.

\subsubsection{Paper 4: Discourse-Aware Rumour Classification}

\begin{figure*}
\fontsize{10}{10}\selectfont
 \begin{framed}
  \textit{[depth=0]} \noindent \textbf{u1:} These are not timid colours; soldiers back guarding Tomb of Unknown Soldier after today's shooting \#StandforCanada --PICTURE-- \textbf{[support]}
  \begin{addmargin}[2em]{0pt}
   \textit{[depth=1]} \textbf{u2:} @u1 Apparently a hoax. Best to take Tweet down. \textbf{[deny]}
  \end{addmargin}
  \begin{addmargin}[2em]{0pt}
   \textit{[depth=1]} \textbf{u3:} @u1 This photo was taken this morning, before the shooting. \textbf{[deny]}
  \end{addmargin}
  \begin{addmargin}[2em]{0pt}
   \textit{[depth=1]} \textbf{u4:} @u1 I don't believe there are soldiers guarding this area right now. \textbf{[deny]}
  \end{addmargin}
  \begin{addmargin}[4em]{0pt}
   \textit{[depth=2]} \textbf{u5:} @u4 wondered as well. I've reached out to someone who would know just to confirm that. Hopefully get response soon. \textbf{[comment]}
  \end{addmargin}
  \begin{addmargin}[6em]{0pt}
   \textit{[depth=3]} \textbf{u4:} @u5 ok, thanks. \textbf{[comment]}
  \end{addmargin}
 \end{framed}
 \caption{[Paper 4] Example of a tree-structured thread discussing the veracity of a rumour, where the label associated with each tweet is the target of the rumour stance classification task.}
 \label{fig:paper4}
\end{figure*}

In this work we focus on the development of stance classification models for rumour detection.
It has been argued that it could be helpful in determining the likely veracity to aggregate across multiple distinct stances in the multiple tweets discussing a rumour. 
This would provide, for example, the means to flag highly disputed rumours as being potentially false (\cite{malon-2018-team}). This approach has been justified by recent research that has suggested that the aggregation of the different stances expressed by users can be used for determining the veracity of a rumour (\cite{derczynski2015pheme,liu2015realtime}).

In this work, we examine in depth the use of so-called sequential approaches to the rumour stance classification task. Sequential classifiers are able to utilise the discursive nature of social media (\cite{tolmie2017microblog}), learning from how `conversational threads' evolve for a more accurate classification of the stance of each tweet -- see Figure \ref{fig:paper4} for an example of such a conversational thread.

The work presented here advances research in rumour stance classification by performing an exhaustive analysis of different approaches to this task. In particular, we make the following contributions:

\begin{itemize}
\item We perform an analysis of whether -- and the extent to which -- the use of the sequential structure of conversational threads can improve stance classification in comparison to a classifier that determines a tweet's stance from the tweet in isolation. To do so, we evaluate the effectiveness of a range of sequential classifiers: (1) a state-of-the-art classifier that uses Hawkes Processes to model the temporal sequence of tweets (\cite{lukasik-etal-2016-hawkes}); (2) two different variants of Conditional Random Fields (CRF), i.e., a linear-chain CRF and a tree CRF; and (3) a classifier based on Long Short Term Memory (LSTM) networks. We compare the performance of these sequential classifiers with non-sequential baselines, including the non-sequential equivalent of CRF, a Maximum Entropy classifier.

\item We perform a detailed analysis of the results broken down by dataset and by depth of tweet in the thread, as well as an error analysis to further understand the performance of the different classifiers. We complete our analysis of results by delving into the features, and exploring whether and the extent to which they help characterise the different types of stances.
\end{itemize}

Our results show that sequential approaches do perform substantially better in terms of macro-averaged F1 score, proving that exploiting the dialogical structure improves classification performance. Specifically, the LSTM achieves the best performance in terms of macro-averaged F1 scores, with a performance that is largely consistent across different datasets and different types of stances. Our experiments show that LSTMs performs especially well when only local features are used, as compared to the rest of the classifiers, which need to exploit contextual features to achieve comparable -- yet still inferior -- performance scores. Our findings reinforce the importance of leveraging conversational context in stance classification. Our research also sheds light on open research questions that we suggest should be addressed in future work. Our work here complements other components of a rumour classification system that we implemented in the PHEME project, including a rumour detection component (\cite{zubiaga2016learning,zubiaga2017exploiting}), as well as a study into the diffusion of -- and reactions to -- rumour (\cite{zubiaga2016analysing}).

\subsubsection{Paper 5: Multi-Task Learning Over Disparate Label Spaces}

\begin{figure}
	\centering
  	\includegraphics[width=0.7\linewidth]{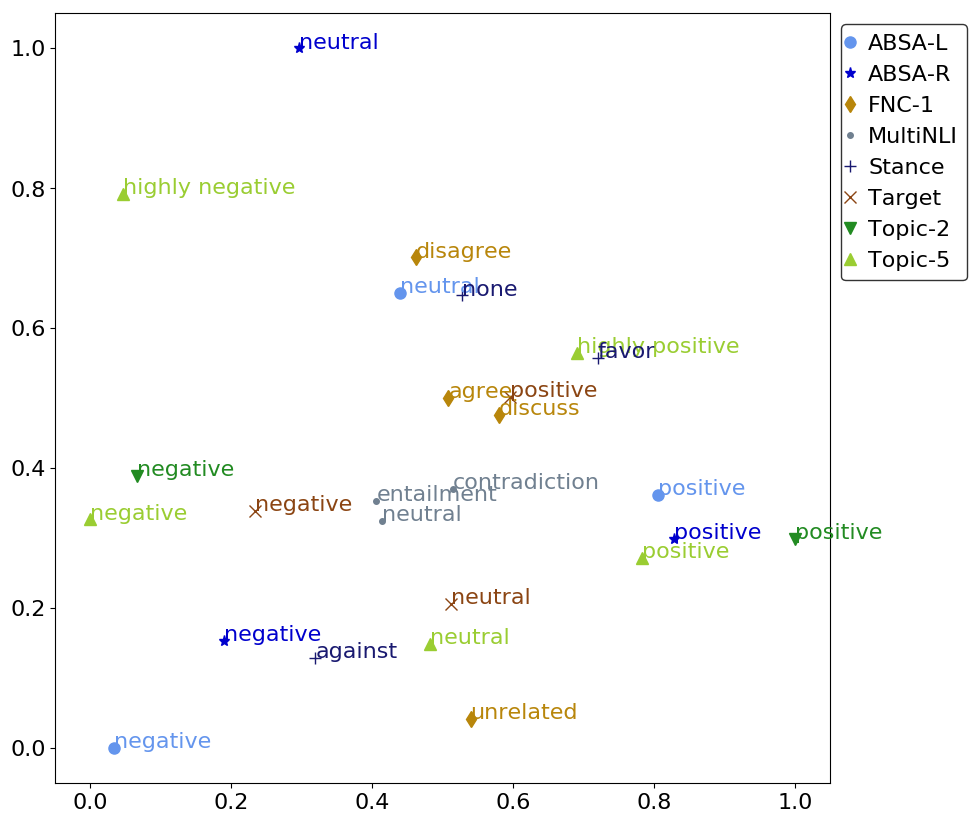}
  	\caption{[Paper 5] Label embeddings of all tasks. Positive, negative, and neutral labels are clustered together.}
  	\label{fig:paper5}
\end{figure}

Contemporary work in multi-task learning for NLP typically focuses on learning representations that are useful across tasks, often through hard parameter sharing of hidden layers of neural networks (\cite{Collobert2011,Soegaard:Goldberg:16}). If tasks share optimal hypothesis classes at the level of these representations, multi-task learning leads to improvements (\cite{Baxter:00}). However, while sharing hidden layers of neural networks is an effective regulariser (\cite{Soegaard:Goldberg:16}), we potentially lose synergies between the classification functions trained to associate these representations with class labels. This paper sets out to build an architecture in which such synergies are exploited, with an application to pairwise sequence classification tasks (topic-based, target-depending and aspect-based sentiment analysis; stance detection; fake news detection; and natural language inference).

For many NLP tasks, disparate label sets are weakly correlated, e.g. part-of-speech tags correlate with dependencies (\cite{Hashimoto2017}), sentiment correlates with emotion (\cite{Felbo2017,eisner-etal-2016-emoji2vec}), etc. We thus propose to induce a joint label embedding space using a Label Embedding Layer that allows us to model these relationships, which we show helps with learning. A visualisation of the learned label embedding space is provided in Figure \ref{fig:paper5}.

In addition, for tasks where labels are closely related, we should be able to not only model their relationship, but also to directly estimate the corresponding label of the target task based on auxiliary predictions. To this end, we propose to train a Label Transfer Network jointly with the model to produce pseudo-labels across tasks. 

In summary, our contributions are as follows. 

\begin{enumerate}[noitemsep]
\item{We model the relationships between labels by inducing a joint label space for multi-task learning.}
\item{We propose a Label Transfer Network that learns to transfer labels between tasks and propose to use semi-supervised learning to leverage them for training.}
\item{We evaluate multi-task learning approaches on a variety of classification tasks and shed new light on settings where multi-task learning works.}
\item{We perform an extensive ablation study of our model.}
\item{We report state-of-the-art performance on topic-based sentiment analysis.}
\end{enumerate}

\subsubsection{Paper 6: Temporal Domain Adaptation with Sequential Alignment}

\begin{figure}
    \centering
    \includegraphics[width=.48\textwidth]{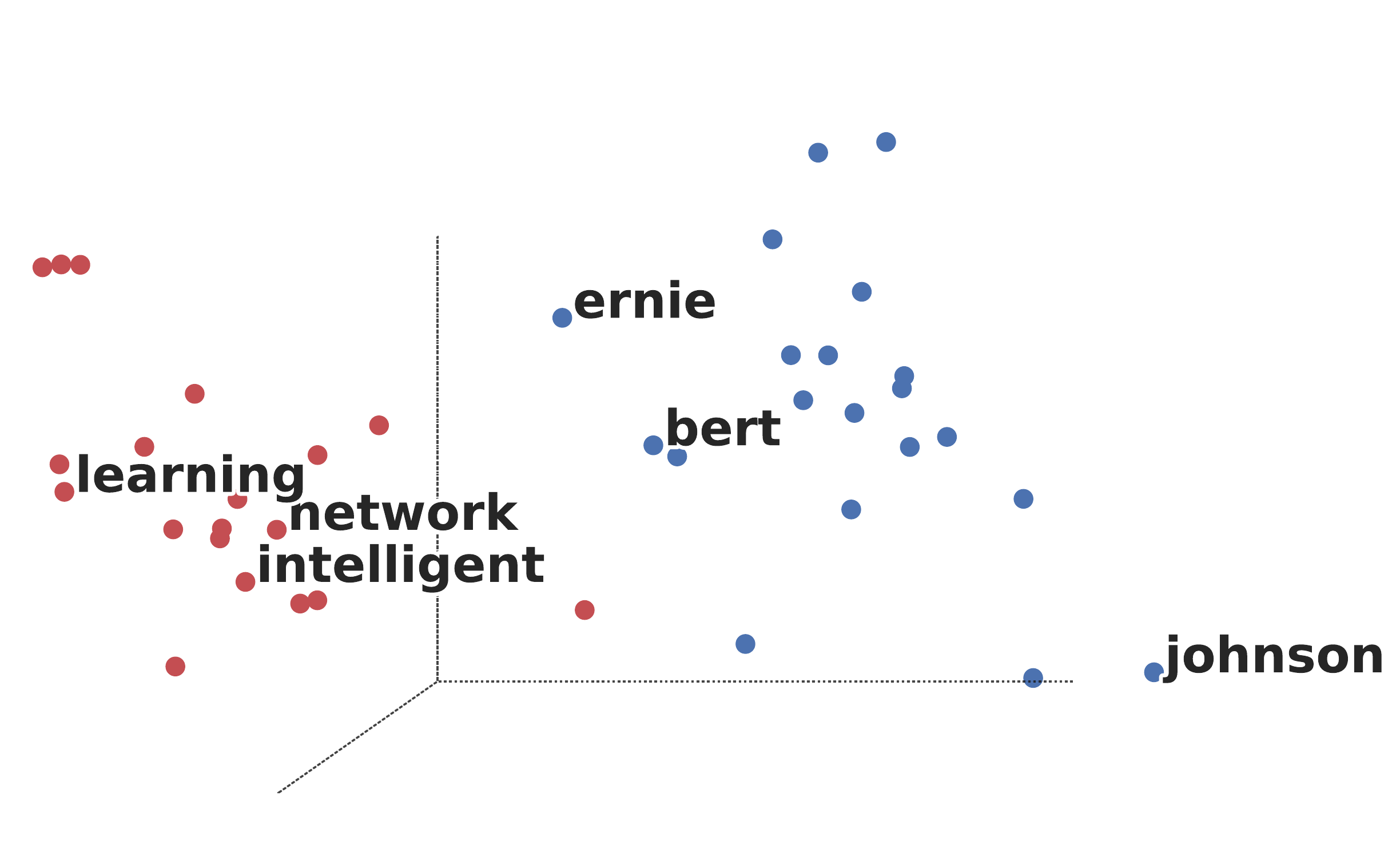}
    \includegraphics[width=.48\textwidth]{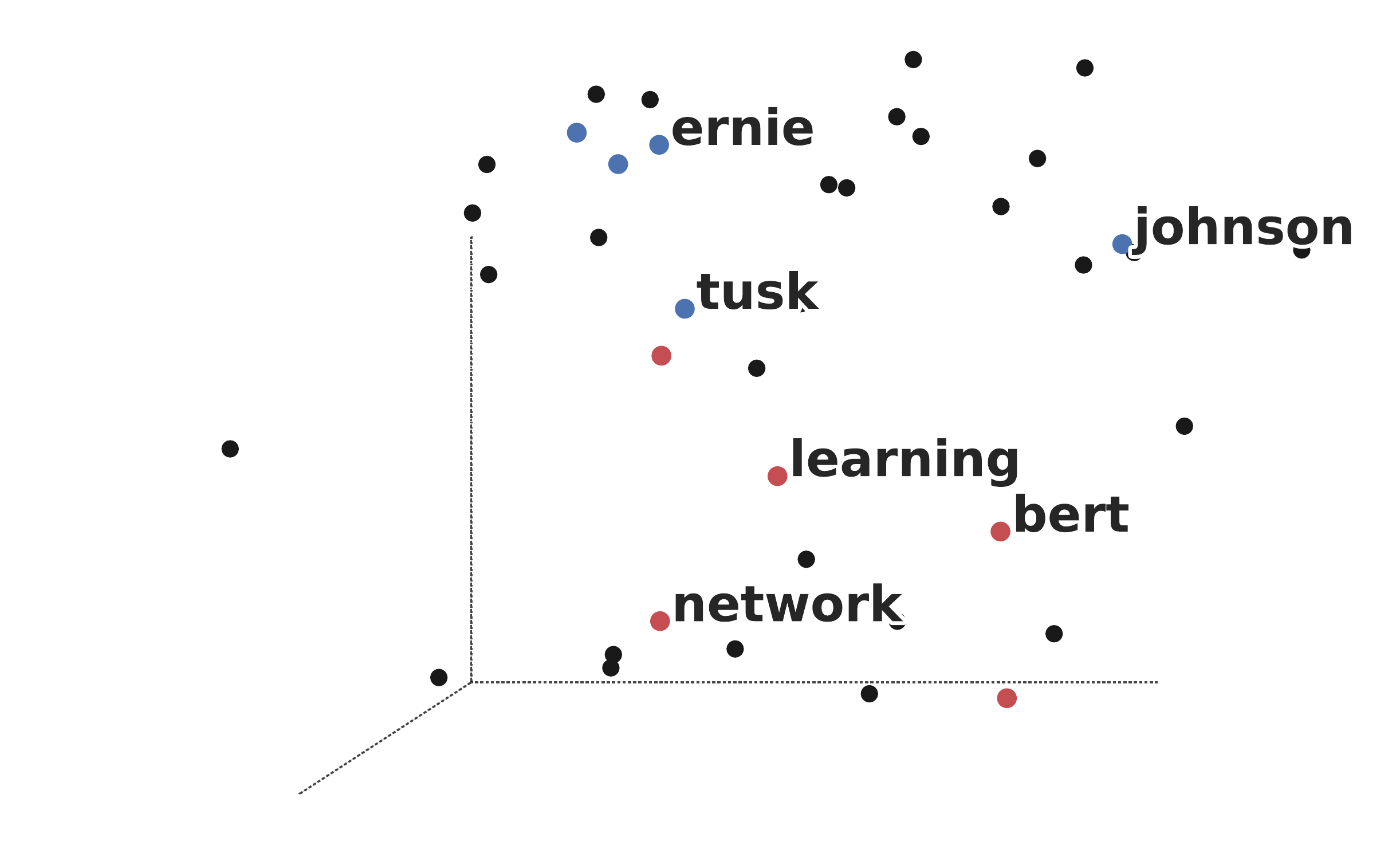}
    \caption{Example of a word embedding at $t_{2017}$ vs $t_{2018}$ (blue={\textsc person}, red={\textsc{artefact}}, black={\textsc{unk}}). Source data (left, $t_{2017}$), target data (right, $t_{2018}$). Note that at $t_{2017}$, 'bert' is a {\textsc{person}}, while at $t_{2018}$, 'bert' is an {\textsc{artefact}}.}
    \label{fig:paper6}
\end{figure}


Evolution of language usage affects natural language processing tasks and, as such, models based on data from one point in time cannot be expected to generalise to the future.
For example, the names `Bert' and `Elmo' would only rarely make an appearance prior to 2018 in the context of scientific writing. After the publication of BERT (\cite{devlin-etal-2019-bert}) and ELMo (\cite{peters-etal-2018-deep}), however, usage has increased in frequency. In the context of named entities on Twitter, it is also likely that these names would be tagged as \textsc{person} prior to 2018, and are now more likely to refer to an \textsc{artefact}. As such, their part-of-speech tags will also differ -- see Figure \ref{fig:paper6}, which aims to illustrate this point.

In order to become more robust to language evolution, data should be collected at multiple points in time. We consider a dynamic learning paradigm where one makes predictions for data points from the current time-step given labelled data points from previous time-steps. As time increments, data points from the current step are labelled and new unlabelled data points are observed. 
Changes in language usage cause a data drift between time-steps and some way of controlling for the shift between time-steps is necessary. Given that linguistic tokens are embedded in some vector space using neural language models, we observe that in time-varying dynamic tasks, the drift causes token embeddings to occupy different parts of embedding space over consecutive time-steps. 

In each time-step we map linguistic tokens using the same pre-trained language model (a ``BERT'' network, \cite{devlin-etal-2019-bert}) and align the resulting embeddings using a second procedure called subspace alignment (\cite{fernando2013unsupervised}). We apply subspace alignment sequentially by finding the principal components in each time-step and transforming linearly the components from the previous step to match the current step. A classifier trained on the aligned embeddings from the previous step will be more suited to classify embeddings in the current step. 

We show that sequential subspace alignment yields substantial improvements in three challenging tasks: paper acceptance prediction on the PeerRead data set (\cite{kang18naacl}); Named Entity Recognition on the Broad Twitter Corpus (\cite{derczynski:2016}); and rumour stance detection on the RumourEval 2019 data set (\cite{rumour:19}).
These tasks are chosen to vary in terms of domains, timescales, and the granularity of the linguistic units. 

In addition to evaluating sequential subspace alignment, we include two technical contributions as we extend the method both to allow for time series of unbounded length and to consider instance similarities between classes.
The best-performing sequential subspace alignment methods proposed here are semi-supervised, but require only between 2 and 10 annotated data points per class from the test year for successful alignment.
Crucially, the best proposed sequential subspace alignment models outperform baselines utilising more data, including the whole data set.

\subsubsection{Paper 7: A Diagnostic Study of Explainability Techniques}

\begin{figure}
\centering
\includegraphics[width=0.7\textwidth]{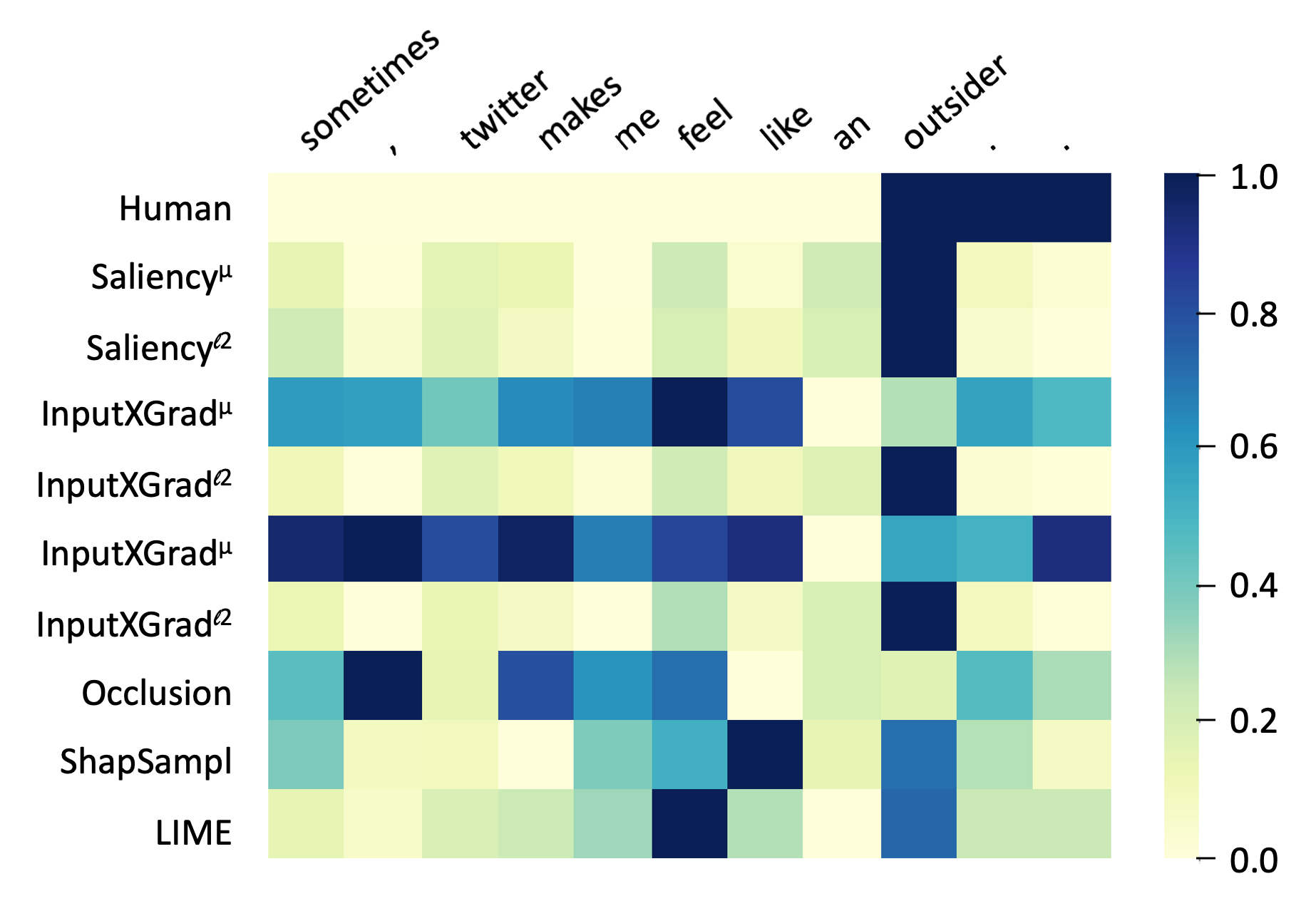}
\caption{[Paper 7] Example of the saliency scores for the words (columns) of an instance from the Twitter Sentiment Extraction dataset. They are produced by the explainability techniques (rows) given a \trans{} model. The first row is the human annotation of the salient words. The scores are normalized in the range $[0, 1]$.}
\label{fig:paper7}
\end{figure}

\textit{Explainability methods} attempt to reveal the reasons behind a model's prediction for a single data point, as shown in Figure~\ref{fig:paper7}. They can be produced post-hoc, i.e., with already trained models. Such post-hoc explanation techniques can be applicable to one specific model~(\cite{barakat2007rule, wagner2019interpretable}) or to a broader range thereof~(\cite{ribeiromodel, lundberg2017unified}). They can further be categorised as: employing model gradients (\cite{sundararajan2017axiomatic, Simonyan2013DeepIC}), being perturbation based (\cite{shapley1953value, zeiler2014visualizing}) or providing explanations through model simplifications (\cite{ribeiromodel, johansson2004truth}). While there is a growing amount of explainability methods, we find that they can produce varying, sometimes contradicting explanations, as illustrated in Figure~\ref{fig:paper7}. 

Hence, it is important to assess existing techniques and to provide a generally applicable and automated methodology for choosing one that is suitable for a particular model architecture and application task~(\cite{jacovi2020towards}). 

In summary, the contributions of this work are:
\begin{itemize}[noitemsep]
\item We compile a comprehensive list of \propertyplural{} for explainability and automatic measurement of them, allowing for their effective assessment in practice.
\item We study and compare the characteristics of different groups of explainability techniques (gradient-based, perturbation-based, simplification-based) in three different application tasks (natural language inference, sentiment analysis of movie reviews, sentiment analysis of tweets) and three different model architectures (CNN, LSTM, and Transformer).
\item We study the attributions of the explainability techniques and human annotations of salient regions to compare and contrast the rationales of humans and machine learning models. 
\end{itemize}

\subsection{Veracity Prediction}

\subsubsection{Paper 8: Multi-Domain Evidence-Based Fact Checking of Claims}

\begin{table}[]
\fontsize{10}{10}\selectfont
\centering
\begin{tabular}{p{2.1cm}l}   
\toprule
Feature & Value \\ \midrule
ClaimID & farg-00004 \\
Claim & Mexico and Canada assemble cars with foreign parts and send them to the U.S. with no tax. \\
Label & distorts \\
Claim URL & \footnotesize{\url{https://www.factcheck.org/2018/10/factchecking-trump-on-trade/}} \\
Reason & None \\
Category & the-factcheck-wire \\
Speaker & Donald Trump \\
Checker & Eugene Kiely \\
Tags & North American Free Trade Agreement \\
Claim Entities & United\_States, Canada, Mexico \\
Article Title & Fact Checking Trump on Trade\\
Publish Date & October 3, 2018 \\
Claim Date & Monday, October 1, 2018 \\ 
\bottomrule
\end{tabular}
\caption{\label{tb:paper8} [Paper 8] An example of a claim instance. Entities are obtained via entity linking. Article and outlink texts, evidence search snippets and pages are not shown.}
\end{table}

Existing efforts for automatic veracity prediction either use small datasets consisting of naturally occurring claims (e.g. \cite{mihalcea2009lie,zubiaga2016analysing}), or datasets consisting of artificially constructed claims such as FEVER (\cite{thorne-etal-2018-fever}). While the latter offer valuable contributions to further automatic claim verification work, they cannot replace real-world datasets.

In summary, Paper 8 makes the following contributions.
\begin{enumerate}[noitemsep]
\item{We introduce the currently largest claim verification dataset of naturally occurring claims.\footnote{The dataset is found here: \url{https://copenlu.github.io/publication/2019_emnlp_augenstein/}} It consists of 34,918 claims, collected from 26 fact checking websites in English; evidence pages to verify the claims; the context in which they occurred; and rich metadata (see Table \ref{tb:paper8} for an example).}
\item{We perform a thorough analysis to identify characteristics of the dataset such as entities mentioned in claims.}
\item{We demonstrate the utility of the dataset by training state of the art veracity prediction models, and find that evidence pages as well as metadata significantly contribute to model performance.}
\item{Finally, we propose a novel model that jointly ranks evidence pages and performs veracity prediction. The best-performing model achieves a Macro F1 of 49.2\%, showing that this is a non-trivial dataset with remaining challenges for future work.}
\end{enumerate}

\subsubsection{Paper 9: Generating Label-Cohesive and Well-Formed Adversarial Claims}

\begin{figure}
\centering
\includegraphics[width=0.6 \textwidth]{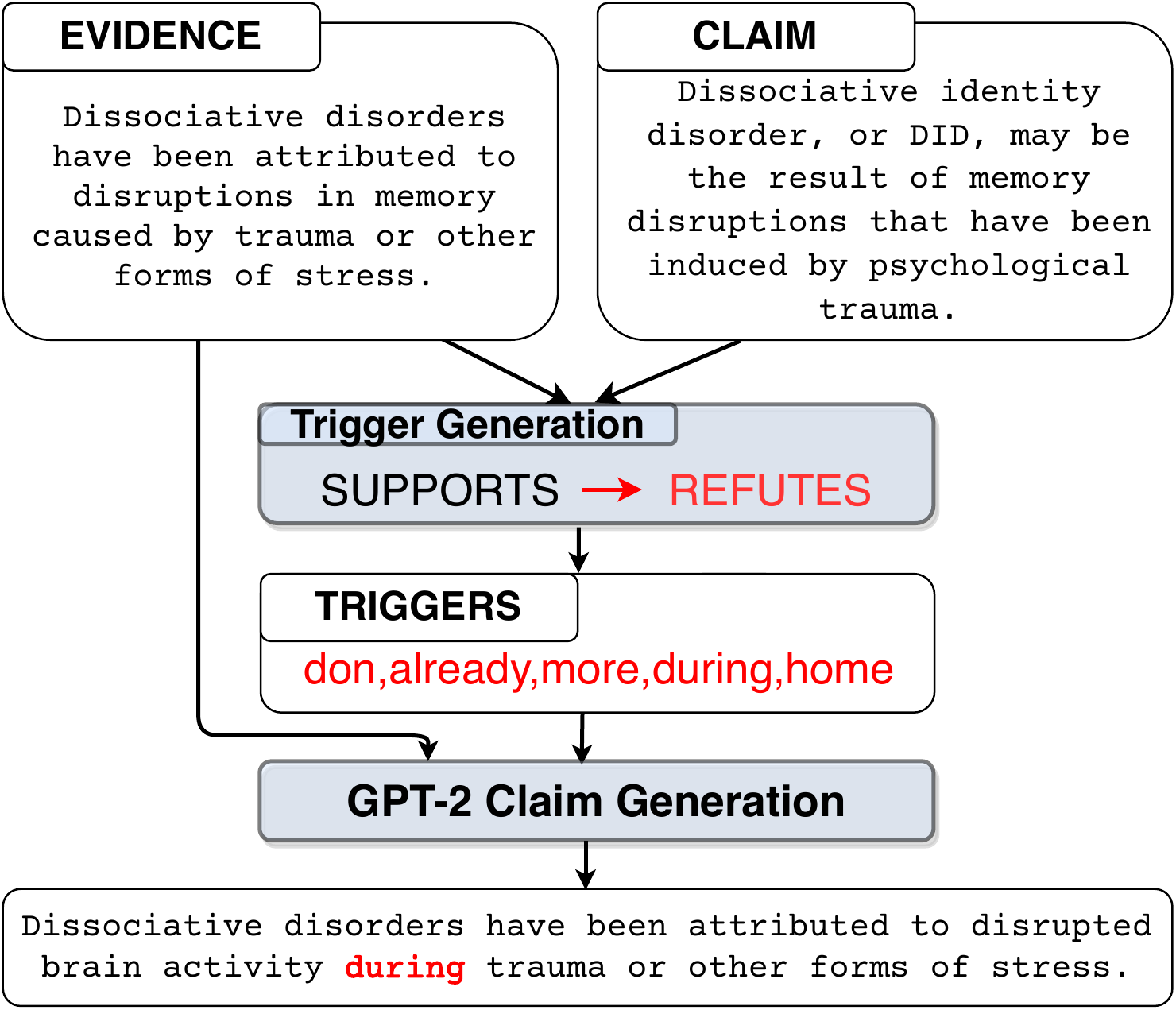}
\caption{[Paper 9] High level overview of our method. First, universal triggers are discovered for flipping a source to a target label (e.g. SUPPORTS $\rightarrow$ REFUTES). These triggers are then used to condition the GPT-2 language model to generate novel claims with the original label, including at least one of the found triggers.}
\label{fig:paper9}
\end{figure}

In this paper, we explore the vulnerabilities of fact checking models trained on the FEVER dataset~(\cite{thorne-etal-2018-fever}), where the inference between a claim and evidence text is predicted. We particularly construct \textit{universal adversarial triggers}~(\cite{wallace2019universal}) -- single n-grams appended to the input text that can shift the prediction of a model from a source class to a target one. Such adversarial examples are of particular concern, as they can apply to a large number of input instances. 

However, we find that the triggers also change the meaning of the claim such that the true label is in fact the target class. For example, when attacking a claim-evidence pair with a `SUPPORTS' label, a common unigram found to be a universal trigger when switching the label to `REFUTES' is `none'. Prepending this token to the claim drastically changes the meaning of the claim such that the new claim is in fact a valid 
`REFUTES' claim as opposed to an adversarial `SUPPORTS' claim.
Furthermore, we find adversarial examples constructed in this way to be nonsensical, as a new token is simply being attached to an existing claim.

In summary, our contributions are as follows. 
\begin{enumerate}[noitemsep]
\item{We \textit{preserve the meaning} of the source text and \textit{improve the semantic validity} of universal adversarial triggers to automatically construct more potent adversarial examples. This is accomplished via: 1) a \textit{novel extension to the HotFlip attack}~(\cite{ebrahimi2018hotflip}), where we jointly minimize the target class loss of a FC model and the attacked class loss of a natural language inference model; 2) a \textit{conditional language model} trained using GPT-2~(\cite{radford2019language}), which takes 
trigger tokens and a piece of evidence, and generates a semantically coherent new claim containing at least one trigger. Our method is illustrated in Figure \ref{fig:paper9}.
}
\item{The resulting triggers maintain potency against a FC model while preserving the original claim label.}
\item{Moreover, the conditional language model produces semantically coherent adversarial examples containing triggers, on which a FC model performs 23.8\% worse than with the original FEVER claims.}
\end{enumerate}

\subsubsection{Paper 10: Generating Fact Checking Explanations}

\setlength{\tabcolsep}{0.2em}

\begin{table}
\fontsize{10}{10}\selectfont
\begin{center}
\begin{tabular}{|L|}
\toprule
\textbf{Claim}: The last major oil spill from a drilling accident in America happened over 40 years ago in 1969.\\ \midrule
\textbf{Ruling Comments}: 
(...) \hlyellow{The last major oil spill from a drilling accident in America happened over 40 years ago in 1969.} \\
\hspace*{3mm}(...) The largest in volume was the Santa Barbara spill of 1969 referenced by Murdock and Johnson, in which an estimated 100,000 barrels of oil spilled into the Pacific Ocean, according to the API. \hlblue{The Santa Barbara spill was so big it ranked seventh among the 10 largest oil spills caused by marine well blowouts in the world, the report states.} Two other U.S. spills, both in 1970, rank eighth and 10th. \hlred{Fourteen marine blowouts have taken place in the U.S. between 1969 and 2007.} Six of them took place after 1990 and spilled a total of nearly 13,700 barrels. \\
\hspace*{3mm}(...) We interviewed three scientists who said that the impact of a spill has little to do with its volume. \hlviolet{Scientists have proven that spills far smaller than Santa Barbara's have been devastating.} \\  \midrule
\textbf{Justification}: While the nation's largest oil well blowout did take place in 1969, it's not factually correct to call it the ``last major oil spill". First of all, two of the largest blowouts in the world took place in the U. S.  the following year. More importantly, experts agree that spills far smaller in volume to the 1969 disaster have been devastating. From a scientific perspective, Johnson's decision to single out the 1969 blowout as the last ``major" one makes no sense. \\ \midrule

\textbf{Ruling}: Half-True \\ \bottomrule
\end{tabular}
\end{center}
\caption{\label{tab:paper10} [Paper 10] Example instance from the LIAR-PLUS dataset, with oracle sentences for generating the justification highlighted.}
\end{table}

A prevalent component of existing fact checking systems is a stance detection or textual entailment model that predicts whether a piece of evidence contradicts or supports a claim (\cite{Ma:2018:DRS:3184558.3188729,mohtarami-etal-2018-automatic,Xu2019AdversarialDA}). Existing research, however, rarely attempts to directly optimise the selection of relevant evidence, i.e., the self-sufficient explanation for predicting the veracity label (\cite{thorne-etal-2018-fever, stammbach-neumann-2019-team}).
On the other hand, \cite{alhindi-etal-2018-evidence} have reported a significant performance improvement of over 10\% macro F1 score when the system is provided with a short human explanation of the veracity label. Still, there are no attempts at automatically producing explanations, and automating the most elaborate part of the process - producing the \emph{justification} for the veracity prediction - is an understudied problem.


In this paper, we research how to generate explanations for veracity prediction. We frame this as a summarisation task, where, provided with elaborate fact checking reports, later referred to as \textit{ruling comments}, the model has to generate \textit{veracity explanations} close to the human justifications as in the example in Table~\ref{tab:paper10}. We then explore the benefits of training a joint model that learns to generate veracity explanations while also predicting the veracity of a claim.

In summary, our contributions are as follows:
\begin{enumerate}[noitemsep]
\item{We present the first study on generating veracity explanations, showing that they can successfully describe the reasons behind a veracity prediction.}
\item{We find that the performance of a veracity classification system can leverage information from the elaborate ruling comments, and can be further improved by training veracity prediction and veracity explanation jointly.}
\item{We show that optimising the joint objective of veracity prediction and veracity explanation produces explanations that achieve better coverage and overall quality and serve better at explaining the correct veracity label than explanations learned solely to mimic human justifications.}
\end{enumerate}

\section{Research Landscape of Content-Based Automatic Fact Checking}

This section contextualises the contributions and findings of this thesis by presenting a broad and up to date literature review of content-based automatic fact checking, highlighting the main research streams and positioning the contributions of the thesis with respect to this. It is not meant to be a comprehensive review of relevant related work, for which the reader is instead referred to the related work sections of the individual papers. 

\subsection{Research Trends over Time}

\begin{figure}
\centering
\resizebox{1.0\linewidth}{!}{
\centering
\includegraphics[]{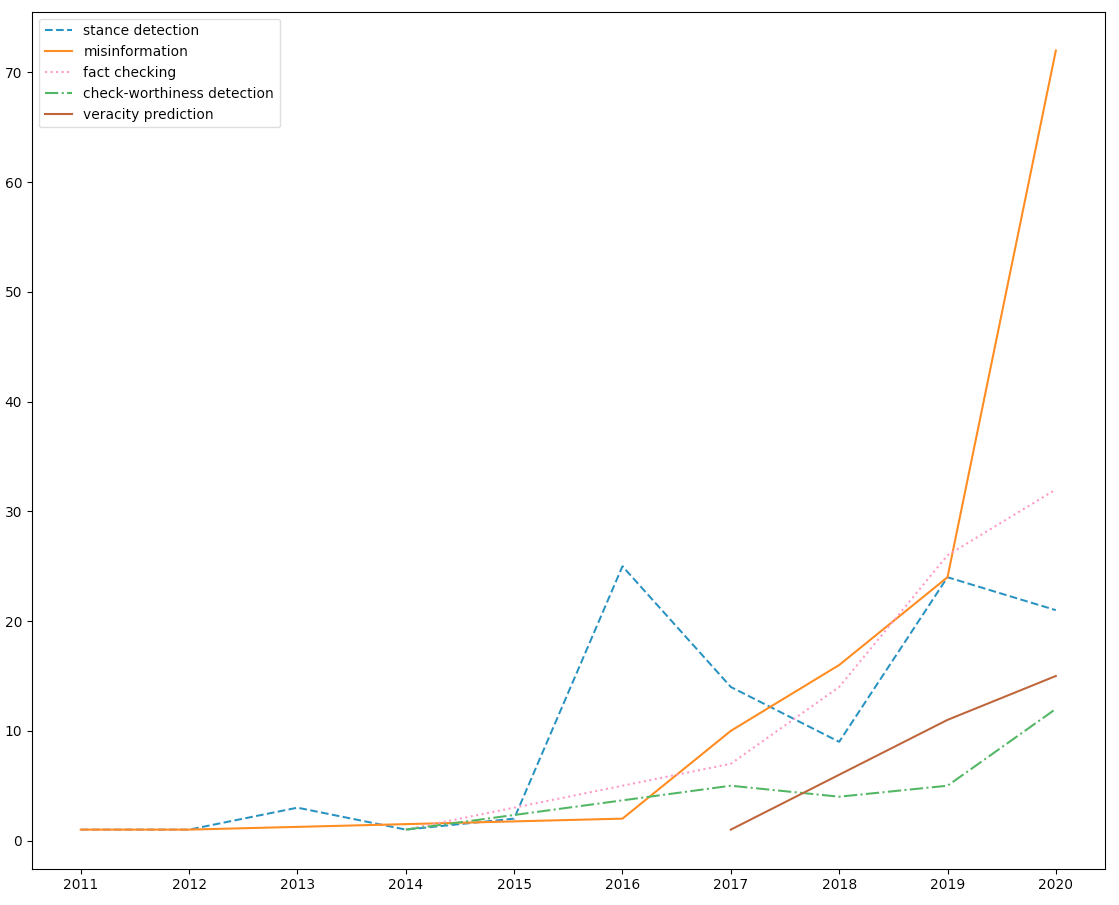} 
}
\caption{This visualises the number of articles on the topics of this thesis published in the ACL Anthology, identified using keyword-based search of these articles' abstracts.}\label{f:fc_pipeline}
\end{figure}

First of all, to get a sense of the popularity of the research area fact checking within natural language processing, Figure \ref{f:fc_pipeline} shows a plot showing the number of paper on the topics over time in the ACL Anthology.\footnote{The ACL Anthology (\url{https://www.aclweb.org/anthology/}) indexes papers for most publication venues within NLP. At the time of writing, the number of papers available there is 62 344.} The x-axis shows the publication year, and the y-axis shows the number of papers on a specific topic. Note that only the year is taken into account to produce this plot, as the exact publication date is not always available.

These papers are identified by using a high-precision keyword-based search of the abstracts of these papers, using the paper's title if the abstract is not readily available.\footnote{It would, of course, be possible to parse the PDF documents to extract those, but this would be much more involved.} More concretely, for each research sub-area addressed by this thesis (misinformation, fact checking, check-worthiness detection, stance detection, veracity prediction), a number of descriptive key phrases\footnote{The full list of key phrases is: \{\textit{misinformation}, fake news, disinformation\}; \{\textit{fact checking}, fact check, fact-check, check fact, checking fact\}; \{\textit{check-worthiness detection}, check-worthy, check-worthiness, check worthiness, claim detection, claim identification, identifying claims, rumour detection, rumor detection\}; \{\textit{stance detection}, stance classification, classifying stance, detecting stance, detect stance\}; \{\textit{veracity prediction}, claim verification, rumour verification, rumor verification\}.} are collected using a top-down process, and if a paper's abstract contains one of these key phrases, it is counted as a match for that research area.\footnote{It might seem odd to the reader that such a simple heuristics-based method is applied here when this thesis otherwise proposes such sophisticated deep learning based methods. As with many real-world problems, no training data was readily available to the author to build a classification model. Hence, this heuristics-based method was chosen instead as a quick and high-precision alternative.}
While this process will not capture all papers on the topic, it should be a reasonably reliable estimate of the development of research trends over time.

As can be seen, there was very little research related to either of the five research topics prior to 2016, which is also when the research documented in this thesis began. Overall, there has been a stark increase in the number of publications on the broad topic of tackling false information over time. One outlier is the sub-area of stance detection; this will be explained further below. Notably, work on veracity prediction itself only began in 2017, and check-worthiness detection is the least researched topic out of the five. 

\subsection{Contextualisation of Contributions}

To understand these research trends in more detail, this section next examines the individual papers on each of the five topics and relates them to the contributions of this thesis.
Research on misinformation as a core NLP problem started around 2011, though only really taking off from 2016 onwards.
There are several papers focusing on misinformation and fact checking in 2016, many of which are inspired by political events, namely the US presidential election.\footnote{The US is one of the two most highly represented countries when it comes to ACL publication (the other one is China) -- see e.g. \url{https://acl2020.org/blog/general-conference-statistics/}. So it is hardly surprising to see NLP research trends inspired by US politics.}
There is another big spike for all fact checking related topics in 2020, when many research papers on tackling misinformation related to COVID-19 were published.

\subsubsection{Misinformation}

Apart from the tasks researched in this thesis -- fact checking as an overall task, check-worthiness detection, stance detection and veracity prediction -- research within NLP has tackled a few other, related problems, which are briefly outlined below for context.


\paragraph{Annotation} \cite{liang-etal-2012-expert} present an annotator matching method for annotating rumourous tweets. \cite{rehm-etal-2018-automatic} present a Web-based annotation tool for misinformation, which offers a combined automatic and manual annotation functionality.  \cite{fan-etal-2020-generating} generate so-called `fact checking briefs', which provide three pieces of information to fact checkers who are given a claim: a passage brief, containing a passage retrieved about the claim; entity briefs, containing more information the entities mentioned in the claim; and question answering briefs, which are questions generated about entities conditioned on the claim with answers. They show that these briefs reduce the time humans needed to complete a fact-check by 20\%. 

\paragraph{Satire Detection}
\cite{W16-0802} focus on satire detection, aiming to automatically distinguish between articles published on two satirical news websites (The Onion and The Beaverton) from two legitimate news sources (The Toronto Star and The New York Times). This turns out to be a relatively easy task, achieving an F1 score of 87\% for the best-performing method, likely in part because the model picked up on the domain discrepancy more than satirical features such as humour.
\cite{de-sarkar-etal-2018-attending} also study satire detection, but using more different sources for real news than \cite{W16-0802}. They combine deep learning with hand-crafted features, namely sentiment look-up, named entity, syntactic and typographic features, and achieve an F1 of 91.59\%.
\cite{rashkin-etal-2017-truth} work on a related problem, namely to analyse stylistic cues of satire, hoaxes and propaganda with that of real news.
\cite{levi-etal-2019-identifying} study how to differentiate fake news from satire, and find that an approach combining state-of-the-art deep learning with hand-crafted stylistic features works well.
\cite{saadany-etal-2020-fake} present a case study of satire detection for Arabic, and \cite{vincze-szabo-2020-automatic} study clickbait detection for Hungarian. \cite{maronikolakis-etal-2020-analyzing} present a large-scale study of parody detection of social media, analysing linguistic markers.

\paragraph{Factuality Detection}
Another task related to satire detection is factuality detection more broadly, namely to automatically distinguish facts from opinions. \cite{dusmanu-etal-2017-argument} study this problem for tweets, and find that it is a reasonably easy task, with their best model achieving an F1 of 78\%.
\cite{alhindi-etal-2020-fact} also study the task of distinguishing fact from opinion articles, and find that argumentation structure features are helpful for this.
\cite{li-etal-2017-nlp} identify to which degree scientific reporting by the news media exaggerates findings in scientific articles they report on.
\cite{baly-etal-2018-predicting} study factuality, not of individual articles, but of media sources as a whole, using features capturing the structure of articles, the sentiment, engagement on social media, topics, linguistic complexity, subjectivity and morality. 

\paragraph{Hyperpartisanship Detection}
\cite{kiesel-etal-2019-semeval} propose a SemEval shared task\footnote{SemEval is a semantic evaluation campaign, which consists of so-called `shared tasks', proposed by groups of researchers, as well as teams of participants, who submit their systems to be evaluated on shared task datasets.} on hyperpartisan news detection. Their large labelled dataset drawn from articles from 383 publishers is annotated using a 5-point Likert scale denoting to what degree an article is hyperpartisan. The best F1 of a submitted system is 82.1\%, achieved by \cite{srivastava-etal-2019-vernon} using sentiment and polarity features combined with sentence embeddings, indicating that the task it not too difficult.

\paragraph{Issue Framing} The task of detecting issue frames is about which facet of an issue is conveyed, for instance, one can discuss COVID-19 with an public health or an economic frame. \cite{field-etal-2018-framing} study political framing for Russian media, \cite{hartmann-etal-2019-issue} study issue framing on Reddit, and \cite{huguet-cabot-etal-2020-pragmatics} study issue frame detection jointly with metaphor and emotion detection in political discourse. A case study of political framing in the Russian news is presented by \cite{field-etal-2018-framing}.

\paragraph{Disinformation Network Detection} As introduced above, disinformation network detection is the task of detecting social sub-networks -- e.g. retweet networks, mention networks, follower networks -- which discuss aspects of a certain topic. In \cite{hartmann-etal-2019-mapping}, we found that that users sharing disinformation about the crash of the of Malaysian Airlines (MH17) flight can be differentiated from those not sharing disinformation using network detection methods.

\paragraph{Relationship to this Thesis} While none of the above-mentioned tasks are core components of fact checking, they are related to it. Finding good annotators for fact checking is important for creating new fact checking datasets. For the one fact checking dataset introduced in this thesis, MultiFC (Paper 8, Section \ref{ch:multifc}), we crawled claims already annotated by journalists. Satire detection is related to fact checking, though it is a slightly different problem -- it is often not categorically false, but rather amusing stories made up around real-word events. Compared to detecting mis- or disinformation, it is a much easier task, as a model can pick up on stylistic cues. 

Factuality detection is one component of check-worthiness detection, and as such also an easier task than what is studied in this thesis. Hyperpartisanship detection can be an important indicator for determining how reliable evidence documents are, though this was out of scope for the study on joint evidence ranking and veracity prediction presented in Paper 8.
Issue framing is a task related to stance detection, in that it determines which sub-aspect of a topic is discussed, which someone might then express a stance on. So far, it has not been directly studied in connection with stance detection. 

Even though I have published on issue framing (\cite{hartmann-etal-2019-issue}), that specific paper has not been included in this thesis as issue framing is currently not part of a standard fact checking pipeline. The same goes for disinformation network detection (\cite{hartmann-etal-2019-mapping}), which I have published on, but is typically viewed as a separate problem from content-based fact checking.

\subsubsection{Fact Checking}\label{ch:Intro:RelWork:FC}

\paragraph{Task definition} Fact checking was first proposed as a task by \cite{vlachos2014fact}, who crawl political statements from Channel 4 and Politifact. They then propose two fact checking methods, though do not test them: comparing claims against previously fact-checked ones, and comparing claims against a large text corpus such as Wikipedia, which is likely to contain the ground truth for some of the claims. They argue that neither method would be able to deal with emerging claims. \cite{thorne-vlachos-2018-automated-fixed} present a survey for fact checking, highlighting multi-modal fact checking, fact checking of longer documents, and fact checking of real-world claims as future directions.

\paragraph{User modelling} Some research further examines users who spread fake news. \cite{del-tredici-fernandez-2020-words} show that tweets by fake news spreaders exhibit certain linguistic trends that hold across domains. \cite{yuan-etal-2020-early} also study users together with source credibility, and find that this can be used for early detection of fake news. 

\paragraph{Generating fake news}
\cite{saldanha-etal-2020-understanding} study how to automatically generate deceptive text, and find that it is important to differentiate topic from stylistic features.
\cite{C18-1287} propose a paired approach, where crowd workers are asked to manipulate credible news to turn them into fake news. \cite{tan-etal-2020-detecting} both generate and explore methods to defend against multi-modal fake news. They find that semantic inconsistencies between images and text serve as useful features.

\paragraph{Adversarial attacks} A significant challenge for fact checking is that systems can easily overfit to spurious patterns. Research has therefore focused on adversarial attacks for fact checking. \cite{thorne2019evaluating} propose hand-crafted adversarial attacks for the FEVER dataset, which they formalise in the FEVER 2.0 shared task~(\cite{thorne-etal-2019-fever2}). The two most highly ranked systems participating in the task~(\cite{niewinski-etal-2019-gem, hidey-etal-2020-deseption}) generate claims that require multi-hop reasoning. 

\paragraph{Generating explanations} \cite{lu-li-2020-gcan} produce explanations for fact checking by highlighting words in tweets using neural attention. However, they do not evaluate the resulting explanations directly; rather, they evaluate how the model the propose as a whole compares to baselines without neural attention. \cite{wu-etal-2020-dtca} create an explainable method by design, namely by proposing to use decision trees for modelling evidence documents, which are inherently interpretable. \cite{kotonya-toni-2020-explainable-automated} propose to generate explanations for fact checking of public health claims, separately from the task of veracity prediction. They model this as a summarisation task, similarly to our work on explanation generation (Paper 9, Section \ref{ch:adversarial}). Another similar approach is \cite{mishra-etal-2020-generating}, who generate summaries of evidence documents from the Web using an attention-based mechanism. They then use these summaries as evidence documents and find that using them performs better than using the original evidence documents directly. 

\paragraph{Relationship to this Thesis} This thesis presents the first work to generate natural language explanations for fact checking (Paper 10, Section \ref{ch:fc_explanations}). While concurrent and subsequent work has investigated the generation of explanations for fact checking, our work is unique in that it generates natural language explanations, and does so jointly with veracity prediction. Unlike many articles on explanation generation, it also includes a thorough manual evaluation of the resulting explanations.

Our work on adversarial claims for fact checking (Paper 9, Section \ref{ch:adversarial}) is unique in that it does not simply propose local perturbations, but identifies universal adversarial triggers for fact checking, which can serve as explanations of the weak points of a fact checking model by highlighting the spurious correlations it has learned. We then show how they can be used to produce diverse semantically coherent and well-formed adversarial claims in a fully automated fashion, whereas prior work mostly focused on narrow categories of claim patterns.

The work on generating fake news is related to that of adversarial attacks for fact checking. The difference is that adversarial attacks are meant to trick (often specific) fact checking models, by generating claims which are confusing for fact checking models. On the other hand, generated fake news claims are not always fact checking model specific, and are mainly aimed at deceiving humans. There are unexploited synergies between the two streams of research though, e.g. in terms of generating claims that seem as natural as possible.

Lastly, user modelling is an orthogonal research stream to what is presented in this thesis. The main barrier to more research in this area is the lack of datasets for which user metadata is available. 

\subsubsection{Claim Check-Worthiness Detection}
Claim check-worthiness detection is the least well-studied sub-task of fact checking, and has mainly taken place in the form of three disconnected research streams described below. In addition to claim check-worthiness detection, there is also the task of claim detection (detecting if a statement is a claim or not, without a regard for check-worthiness). This is omitted here for brevity.

\paragraph{Political debates} Political debates are the primary current source for the task of claim check-worthiness detection. \cite{gencheva-etal-2017-context-fixed} propose a dataset constructed from US presidential debates, which they annotated for check-worthiness by comparing it against nine fact checking sources. Most other datasets of political speeches annotated for check-worthiness were created in the context of the Clef CheckThat! shared tasks~(\cite{atanasova2018overview, elsayed2019overview,barron2020checkthat}) and ClaimRank~(\cite{jaradat2018claimrank}). Each sentence is annotated by an independent news or fact-checking organisation as to whether or not the statement should be checked for veracity.
\cite{vasileva-etal-2019-takes} show that it is helpful to frame check-worthiness detection as a multi-task learning approach, considering multiple sources at once.

\paragraph{Rumour Detection}
Detecting rumours on Twitter is mainly studied using the PHEME dataset~(\cite{zubiaga2016analysing}), which consists of a set of tweets and associated threads from breaking news events which are either rumourous or not. 
There are various papers showing that rumours can be detected based on their propagation structure on social media, i.e. based on how the rumour spreads. \cite{ma-etal-2017-detect} propose a kernel learning method for this, \cite{ma-etal-2018-rumor} a tree RNN; other solutions involve conditional random fields~(\cite{zubiaga2017exploiting,journals/ipm/ZubiagaKLPLBCA18}), or tree Transformers (\cite{ma-gao-2020-debunking}). Other streams of solutions identify salient rumour-related words~(\cite{abulaish2019graph}), use a GAN to generate misinformation in order to improve a downstream discriminator~(\cite{ma2019detect}), or try to identify the minimum amount of posts needed to determine with certainty if a tweet constitutes a rumour or not (\cite{zhou-etal-2019-early,xia-etal-2020-state}).

\paragraph{Citation Needed Detection}
A last task related to claim check-worthiness detection is citation needed detection for Wikpedia articles (\cite{redi2019citation}).
The authors present a dataset of sentences from Wikipedia labelled for whether or not they have a citation attached to them. In addition to this, they also release a set of sentences which have been flagged as not having a citation but needing one (i.e. \textit{unverified}). In contrast to other check-worthiness detection domains, there is much more training data available on Wikipedia. However, the rules for what requires a citation do not necessarily capture all ``checkable'' statements, as ``all material in Wikipedia articles must be verifiable''~(\cite{redi2019citation}). 

\paragraph{Relationship to this Thesis} Paper 1 (Section \ref{ch:pu_learning}) in this thesis presents a holistic approach to claim check-worthiness detection, which revisits different variants of this task previously researched in isolation, namely detecting check-worthy sentences in political debates, detecting rumours on Twitter, and detecting sentences requiring citations on Wikipedia. We suggest a transfer learning approach, pre-training on the large citation needed detection dataset, and fine-tuning on the smaller datasets for political debates and Twitter rumours. By combining this with positive unlabelled learning, we outperform the state of the art in two of the three tasks studied.

Paper 2 (Section \ref{ch:domain}) examines rumour detection as one of the tasks for studying domain adaptation. Among others, we find that using mixture-of-experts methods to adapt Transformer representations is an effective way of generalising across domains for rumour detection, which we find outperforms both internal and external methods on the PHEME rumour detection dataset. Ours it the only paper to use rumour detection for domain adaptation research. We find that the dataset is well-suited to it, due to it being a relatively challenging task, thus offering room for performance gains.

\subsubsection{Stance Detection}\label{ch:Intro:RelWork:Stance}
Stance detection is most popular fact checking sub-task, and the one with the longest history. There have been many different ways of defining the stance detection task. An attempt to group them is presented below.

\paragraph{Stance detection in debates}
Early work on stance detection focused on stance in the context of debates, and mostly considers binary stance -- for or against an issue. One such genre is online debates, studied by \cite{anand-etal-2011-cats,walker-etal-2012-stance,hasan-ng-2013-stance,hasan-ng-2013-frame,ranade-etal-2013-stance}. An interesting observation in that context is that people rarely change their stance on an issue, which can be exploited by incorporating this into a model in the form of soft or hard constraints (\cite{sridhar-etal-2014-collective}). 
\cite{orbach-etal-2020-echo} study stance in speeches, more concretely, trying to find speeches that directly counter one another, i.e. have the opposing stance on a topic.
\cite{sirrianni-etal-2020-agreement} propose to not only model stance polarity, but also stance intensity, and present a new dataset for this task.

\paragraph{Stance detection in tweets}
Stance detection in tweets was first introduced as a SemEval shared task in 2016 (\cite{mohammad-etal-2016-semeval}), consisting of both `seen target' subtask and an `unseen target' subtask (see next paragraph). Some of the interesting approaches on this dataset include \cite{sobhani-etal-2016-detecting,ebrahimi-etal-2016-joint,sun-etal-2018-stance,li-caragea-2019-multi}, who exploit the connection between plain sentiment analysis (not targeted) and stance detection (targeted) by predicting both in a joint model, and show that this improves stance performance. \cite{zhang-etal-2020-enhancing-cross} build on this idea and show that external emotion knowledge can be used to generalise across domains more effectively. \cite{conforti-etal-2020-will} present the largest stance detection dataset to date, consisting of just over 50k tweets. The dataset consists of tweets discussing mergers and acquisitions between companies, and thus presents and interesting alternative to other datasets mainly consisting of political tweets.

\paragraph{Unseen target stance detection}
Unseen target stance detection was formalised as a subtask in the same above-mentioned SemEval 2016 shared task (\cite{mohammad-etal-2016-semeval}). The focus of this task is tweet stance detection towards previously unseen target entities -- mostly entities such as politicians or issues of public interest. This dataset presents a much more difficult task than previous stance detection settings, not just due to the lack of training data for the test target, but also because the targets are often only implicitly mentioned in the tweets.
\cite{ebrahimi-etal-2016-weakly} show that a weak supervision approach, automatically annotating instances originally annotated towards other targets with labels indicating their stance towards the test target, yields a dataset that leads to improved performance on the unseen target test set compared to not using weakly annotated tweets. \cite{allaway-mckeown-2020-zero} present a dataset for unseen target stance detection from news data, which contains annotations for topics in addition to claim targets.

\paragraph{Multi-target stance detection}
A follow-up dataset for unseen target stance detection is presented by \cite{sobhani-etal-2017-dataset}, who annotate the stance of each tweet towards multiple targets. \cite{xu-etal-2018-cross} present a solution to this cross-target stance detection task, consisting of self-attention networks. \cite{ferreira-vlachos-2019-incorporating} present a solution to multi-target stance detection on three datasets, which models the cross-label dependency between softmax probabilities using a cross-label dependency loss.

\paragraph{Rumour stance detection}
Stance detection towards rumours was first introduced as a task by \cite{qazvinian2011rumor}, who propose it as a binary classification task (support vs deny). They train a model on past tweets about a rumour, and at test time, apply the trained model to new tweets about the same rumour. A shared task on rumour stance detection was then proposed at SemEval 2017, and extended in SemEval 2019 (\cite{derczynski2017semeval,rumour:19}). Subtask A is concerned with classifying individual tweets discussing a rumour within a conversational thread as `support', `deny', `query' or `comment'. \cite{scarton-etal-2020-measuring} revisit the evaluation of rumour stance detection systems, proposing a new metric that better captures performance differences for this class-imbalanced task.

\paragraph{Topic stance detection}
\cite{levy-etal-2014-context} present a dataset and an approach to stance detection of topics in Wikipedia. \cite{sasaki-etal-2017-topics,sasaki-etal-2018-predicting} study stance detection of Twitter users towards multiple targets, and find that this can be predicted effectively using a matrix factorisation approach, as common in recommender systems, that predicts the stances of all users towards all topics jointly.

\paragraph{Headline stance detection}
\cite{DBLP:conf/naacl/FerreiraV16} present a dataset and approach for headline stance detection, where the stance of a news article towards a headline is determined. This is also the setting explored in the 2017 Fake News Challenge Stage 1 (FNC-1, \cite{PomerleauRao}), which was later criticised for presenting an class-imbalanced setting (\cite{hanselowski-etal-2018-retrospective}).

\paragraph{Claim perspectives detection}
There is further work on claim perspectives detection (\cite{chenseeing}), meaning a claim is paired with several sentences and evidence documents. The stance of each of these towards the claim is then annotated. \cite{popat-etal-2019-stancy} also address this task, and propose a consistency constraint for the loss function, which enforces that representations of claims and perspectives are similar if the perspective supports the claim, and dissimilar if it opposes the claim.

\paragraph{Detecting previously fact-checked claims}
\cite{shaar-etal-2020-known} study the detection of previously fact-checked claims. They frame this as a semantic matching task similar to stance detection. \cite{vo-lee-2020-facts} present a similar study, but consider images in addition to text to identify previously fact-checked articles.

\paragraph{User Features} Some few works incorporate user features. \cite{lynn-etal-2017-human} incorporate the predicted age, gender and `Big Five' personality traits (\cite{Goldberg_90}) -- training classifiers for these features on external datasets and applying them to the target tweets. They observe a noticeable increase in performance from this. \cite{joseph-etal-2017-constance} use user information not for stance prediction, but for annotation -- they find that annotators have an easier task annotating stance correctly if they are provided profile information, previous tweets and the political party affiliation of the user whose tweet they are annotating.

\paragraph{Relationship to this Thesis} This thesis studies the settings of unseen target stance detection and rumour stance detection.

Our work on unseen target stance detection (Paper 3, Section \ref{ch:conditional}) proposes a method to generalise to any unseen target, using bidirectional conditional encoding. The trick employed in that paper is to make what would ordinarily be part of the output (a stance towards a concrete stance target) part of the input instead, and to model the dependencies between tweets and targets. This, combined with a weak supervision approach, achieved state of the art performance on the SemEval 2016 Unseen Target Stance Detection dataset. This has inspired many follow-up papers, including other papers presented in this thesis (Papers 5 and 8), applying the architecture to not only stance detection, but also other pairwise sequence classification tasks. 

More concretely, Paper 5 (Section \ref{ch:disparate}) uses Paper 3 as a base architecture to study multi-task learning of pairwise sequence classification tasks. The key innovation of that work is to model outputs with label embeddings, which is shown to improve performance. This idea as such is neural architecture agnostic, and has subsequently been used in other settings, including in our own work (e.g. \cite{bjerva-etal-2019-probabilistic}).

Our work on rumour stance detection (Paper 4, Section \ref{ch:discourse}) shows that modelling the tree structure of tweet threads using nested LSTMs significantly improves performance, achieving the best performance on the SemEval 2017 Rumour Stance Detection dataset. While this was originally proposed for an LSTM architecture, the general idea is neural architecture independent. Subsequently, there have been many approaches to modelling the tree structure of tweets for rumour stance detection, specifically for the joint rumour stance and veracity prediction setting, further described in the next section.

Paper 6 (Section \ref{ch:sequential}) is the only paper, to the best of our knowledge, which studies temporal domain adaptation for stance detection, but clearly demonstrates the benefits of doing so. It is also one of the few papers which study temporal domain adaptation for NLP as a whole, because most datasets do not contain time stamps for instances -- they are either removed during dataset pre-processing, or are never available in the first place.

Lastly, Paper 7 (Section \ref{ch:Diagnostic}) studies explainability for sequence classification tasks, including pairwise sequence classification. There is currently no paper on explainable stance detection and in this work too, we study explainable textual entailment recognition, not stance detection. This is because there is currently no stance detection dataset annotated for explanations. As the task is very similar semantically, in form and in terms of the label scheme, the findings can be expected to be transferable. Paper 7 presents a holistic examination of rationale-based explainability techniques and proposes automatic evaluation metrics for them. It finds that gradient-based techniques perform best across different tasks, datasets, and model architectures.

\subsubsection{Veracity Prediction}
Veracity prediction has been studied for claims in isolation, either with or without evidence pages; and for claims appearing in social media contexts.

\paragraph{Claim Only} Early approaches to veracity prediction do not take evidence documents into account, but merely consider claims. \cite{P17-2067} crawl claims from Politifact and propose a CNN-based approach to verify them. \cite{long-etal-2017-fake} extend this by also encoding meta-data about speakers, including speaker name, title, party affiliation, current job, location of speech, and credit history. \cite{alhindi-etal-2018-evidence} propose an extension using not only the claim, but also the gold justification written by journalists and, unsurprisingly, find that this improves results.
\cite{naderi-hirst-2018-automated} explore claim-only prediction for parliamentary debates extracted from the Toronto Star newspaper. They further cross-reference the extracted claims against Politifact. 

\paragraph{Evidence-Based} \cite{karadzhov-etal-2017-fully} propose an evidence-based framework for claim verification, which retrieves Web pages as evidence documents. They first convert the query to query terms, then retrieve Web pages via two search engines, then automatically extract sentences from the Web page to use as evidence documents, in addition to the snippet identified by the search engine. \cite{popat-etal-2018-declare} also propose a Web-based framework, and an attention-based method for veracity prediction. \cite{baly-etal-2018-integrating} present a small dataset for fact checking of claims in Arabic, which, unlike prior work, has annotations not just for veracity, but also for stance towards evidence documents. A similar effort is made by \cite{hanselowski-etal-2019-richly}, who annotate the stance of evidence documents towards claims from the Snopes fact checking portal.

FEVER \cite{thorne-etal-2018-fever} is the first large-scale dataset for fact checking, which, unlike others, consists not only of a couple of thousand claims, but close to 200k claims. It is artificially constructed from Wikipedia to ensure easier benchmarking. Each claim is annotated with a label of `supports', `refutes' or `not enough info'. Evidence sentences are also to be retrieved from Wikipedia in this setting.
There have since been many papers benchmarking their approaches on the FEVER dataset. Some notable ones include: \cite{lee2018improving}, proposing the use of decomposable attention and semantic tagging; \cite{yin-schutze-2018-attentive}, proposing an attention mechanism over CNNs, and applying this to modelling the relationships between evdience sentences; \cite{suntwal-etal-2019-importance}, who show that entity masking as well as paraphrasing improves out-of-domain performance; and \cite{zhong-etal-2020-reasoning}, who improve the modelling of the relationship between evidence documents and claims using semantic role labelling.

The outbreak of COVID-19 further inspired veracity prediction of scientific claims. \cite{wadden-etal-2020-fact} present SciFact, a small dataset consisting of scientific claims, with evidence documents being the abstracts of scientific articles. Scientific claims are obtained from scientific articles themselves. \cite{liu-etal-2020-adapting-open} experiment on this dataset and find that a domain adaptation approach works best to tackle the low-data problem. In addition to this, there are many preliminary studies on veracity prediction of COVID-related public health claims, which are left out here for brevity.

\paragraph{Knowledge Graphs}
\cite{thorne-vlachos-2017-extensible} propose to tackle the verification of numerical claims using a semantic parsing approach, which compares the extracted statements against those found in a knowledge base.
\cite{zhong-etal-2020-logicalfactchecker} also use a semantic parsing approach for the verification of tabular claims against a knowledge base, using a neural module network. \cite{kim-choi-2020-unsupervised} study fact checking in knowledge graphs and propose a new method to find positive and negative evidence paths in them. \cite{zhang-etal-2020-said} propose to track the provenance of claims using a knowledge graph construction approach, and show that this can aid claim verification.

\paragraph{Multi-Modal Approaches} Approaches not only involving text, but also images or other modalities have also been investigated. \cite{zlatkova-etal-2019-fact} study the verification of claims about images. \cite{nakamura-etal-2020-fakeddit} propose a large multi-modal dataset of images paired with captions. The task is to determine if the caption and image together constitute true content, or if the image is manipulated, the connection between the two is not genuine, etc. \cite{medina-serrano-etal-2020-nlp} detect COVID-10 related misinformation in YouTube videos from user comments. \cite{wen-etal-2018-cross} present a cross-lingual cross-platform approach to -- and dataset for rumour veracity prediction for -- 17 events.

\paragraph{Rumour Veracity} A shared task for rumour veracity prediction was SemEval 2017 Task 7 Subtask B (\cite{derczynski2017semeval}). There, participants should predict the veracity of a rumour based on the rumorous tweet alone. Note that this is a slighty different task from rumour detection, a subtask of check-worthiness detection, in that here, the dataset only consists of rumours (not non-rumours) and each rumour's veracity has to be predicted. \cite{kochkina-etal-2018-one,wei-etal-2019-modeling,kumar-carley-2019-tree,li-etal-2019-rumor} later deviate from this setting and demonstrate the benefits of approaching rumour stance and veracity in a joint multi-task learning setting. A follow-up RumourEval shared task was held in 2019, which formalised this joint setting as Subtask B (\cite{rumour:19}). \cite{yu-etal-2020-coupled} propose a Transformer-based coupled rumour stance and veracity prediction model, which models the interactions between the two tasks in a more involved way than previous multi-task learning approaches.
\cite{li-etal-2019-rumor} shows the benefit of incorporating user credibility features into the rumour detection layer, e.g.: if the account is verified, if a location is provided in the profile, and if the profile includes a description. 
\cite{kochkina-liakata-2020-estimating} propose an active learning setting, i.e. to use uncertainty estimates to for rumour veracity prediction as a rumour unfolds, as a way of deciding when to solicit input from human fact checkers.
\cite{hossain-etal-2020-covidlies} present a small social media data of COVID-19 related misconceptions/rumours, with tweets discussing those annotated with their stance.  

\paragraph{Relationship to this Thesis} This thesis presents the first large dataset for evidence-based fact checking of real-world claims (MultiFC, Paper 8, Section \ref{ch:multifc}). It is an order of magnitude larger than previous real-world datasets for fact checking. Unlike the popular FEVER dataset, it does not consist of artificial claims, but of naturally occurring claims.

Other contributions related to veracity prediction are the generation of adversarial claims and the generation of fact checking explanations, which were already discussed in Section \ref{ch:Intro:RelWork:FC}.

\section{Conclusions and Future Work}

This section offers suggestions for future work on explainable fact checking. 

\subsection{Dataset Development}\label{sec:conclusions:dataset}
Research in natural language processing, and by extension also on fact checking, can have different types of core contributions. The most common such  are: \textit{methodology}-centric contributions, which are new methods published for existing tasks or datasets; and \textit{dataset}-centric contributions, i.e. new datasets, potentially for entirely new tasks. Even though most research is on new methodologies, this research cannot exist without the introduction of datasets to benchmark these methods on.
This also explains much of the research progress on, and related to, fact checking (visualised in Figure \ref{f:fc_pipeline}). The first very popular task that constitutes a fact checking component is stance detection, which spurred new research with the introduction of a dataset (SemEval 2016 Task 6, \cite{mohammad-etal-2016-semeval}) in 2016. Work on veracity prediction as a complete task consisting of several components only saw a large increase in popularity in 2018, when the large FEVER dataset (\cite{thorne-etal-2018-fever}) was introduced.

For explainable fact checking, the research presented in this thesis presents some of the very first findings on this topic, published only in 2020. There is no commonly-agreed upon large dataset to benchmark methods for this task yet. The dataset we used for Paper 10 on generating instance-level fact checking explanations in the PolitiFact-based dataset LIAR-PLUS dataset (\cite{alhindi-etal-2018-evidence}), consists of only 10k training instances. As the results presented in the paper show, it is very difficult to beat simple baselines with such small amounts of training data. In Paper 10, on generating adversarial claims to reveal model-level vulnerabilities, the method presented to generate such claims is unsupervised, and thus requires no annotated training data for adversarial claims. Paper 8, which studies post-hoc explainability techniques for several text classification tasks, only studies a sub-task of fact checking, namely natural language inference / stance detection (on the e-SNLI dataset by \cite{NIPS2018_8163}). This is because, to date, there is no dataset for the task of fact checking as a whole annotated with so-called `human rationales', i.e. portions of the input annotated for to what degree they are indicative of the instance's label.

As such, significant progress on explainable fact checking would require the publication of \textit{large, annotated datasets for different explanation types}, which future work should focus on.



\subsection{Synergies between Explainable Fact Checking Methods}

Next, as Table \ref{tab:overview_contributions} providing an overview of the contributions of this thesis illustrates, research on fact checking has mostly taken place in the form of isolated case studies. 

When it comes to fact-checking tasks, early research mostly studied stance detection as a core component of automated fact checking. Even though more recent research considers several components jointly, it is usually the components of stance detection and veracity prediction which are considered jointly, whereas evidence retrieval and especially the identification of check-worthy claims are often assumed to be solved. To the best of my knowledge, there is not a single paper which addresses the identification of check-worthy claims as part of a \textit{larger, joint learning approach}, which I would propose for future work.

Moving on to methods for dealing with limited labelled data, there have been significantly more synergies in that area than on fact checking sub-tasks, especially on combining transfer learning with other types of approaches. There is, however, very little research on temporal domain adaptation, and Paper 7 is the only one I am aware of which tackles this within the broad area of fact checking. Hence, progress could certainly achieved by research on \textit{temporal domain adaptation for more sub-areas of fact checking}.
As outlined in Section \ref{sec:conclusions:dataset} above, a perhaps surprising barrier preventing more research on this is that time steps are not always readily available for benchmark datasets. In cases where an effort would have to be to collect them, they are very rarely available, and in cases where they would be easy to obtain (e.g. for tweets), they are often discarded during dataset pre-processing, and can then be hard to obtain afterwards (because the content has been deleted by then).

Lastly, for explainability methods, it would be very interesting to study how to take advantage of \textit{different types of explainability methods in a joint framework}. I am certain that jointly studying rationale-based, natural language based, and model-level explanations would generate more stable and informative explanations for users. The likely reason this has not been studied yet is simply the state of research as a whole -- the unavailability of datasets, combined with so far relatively little methodology research, and, as is discussed next, the lack of user studies.

\subsection{Explanations to Support Human Decision Making}\label{sec:conclusions:human}

The purpose of generating explanations for fact checking is to provide humans with more in-depth guidance about how a model has arrived at a prediction and whether the prediction can be trusted. Current research on explainable fact checking is still in its infancy, so real user studies, or even deploying explainable fact checking models in real-world settings may seem far off. 

However, very recent EU legislation passed in December 2020, namely the Digital Services Act (\cite{digitalservicesact2020}) will soon require all digital service providers operating in the EU to provide transparent reporting on their services, which includes user-facing information, and, for larger provides, flag harmful content. While there is already legislation requiring decision makers operating in the EU to provide explanations, namely the regulation on a `Right to Explanation' (\cite{goodman2017european}), this is much less far-reaching than this new legislation, as the former only applies to situations where individuals are strongly impacted by an algorithmic decision.
The Digital Services Act is much more broad-reaching in that it affects all digital platform providers and, at its very core, has the goal of providing a better online experience, which protects users and society as a large from disinformation and other types of harm. As false information online is one type of harmful content, digital platform providers will likely soon look for ways to deploy explainable fact checking methods.

This development, inspired by new legislation, will in turn provide new opportunities for \textit{industry-driven research, involving user studies}, which, can be expected to create new opportunities for basic research on explainable fact checking.

\part{Detecting Check-Worthy Claims}\label{part:II}
\chapter{Claim Check-Worthiness Detection as Positive Unlabelled Learning}\label{ch:pu_learning}

\boxabstract{

As the first step of automatic fact checking, claim check-worthiness detection is a critical component of fact checking systems. There are multiple lines of research which study this problem: check-worthiness ranking from political speeches and debates, rumour detection on Twitter, and citation needed detection from Wikipedia. 
To date, there has been no structured comparison of these various tasks to understand their relatedness, and no investigation into whether or not a unified approach to all of them is achievable.
In this work, we illuminate a central challenge in claim check-worthiness detection underlying all of these tasks, being that they hinge upon detecting both how factual a sentence is, as well as how likely a sentence is to be believed without verification. As such, annotators only mark those instances they judge to be clear-cut check-worthy. Our best performing method is a unified approach which automatically corrects for this using a variant of positive unlabelled learning that finds instances which were incorrectly labelled as not check-worthy.
In applying this, we outperform the state of the art in two of the three tasks studied for claim check-worthiness detection in English. 
}\blfootnote{\fullcite{wright-augenstein-2020-claim}}

\section{Introduction}
Misinformation is being spread online at ever increasing rates (\cite{del2016spreading}) and has been identified as one of society's most pressing issues by the World Economic Forum~(\cite{howell2013digital}). In response, there has been a large increase in the number of organizations performing fact checking (\cite{graves2016rise}). However, the rate at which misinformation is introduced and spread vastly outpaces the ability of any organization to perform fact checking, so only the most salient claims are checked. This obviates the need for being able to automatically find check-worthy content online and verify it.

\begin{figure}[t]
  \centering
    \includegraphics[width=0.6\textwidth]{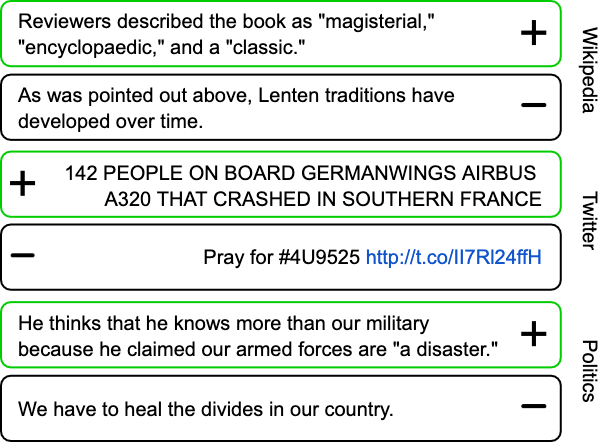}
    \caption{Examples of check-worthy and non check-worthy statements from three different domains. Check-worthy statements are those which were judged to require evidence or a fact check.}
    \label{fig:check-worthy-examples}
\end{figure}

The natural language processing and machine learning communities have recently begun to address the problem of automatic fact checking~(\cite{vlachos2014fact,hassan2017claimbuster, thorne-vlachos-2018-automated-fixed,augenstein-etal-2019-multifc,atanasova-etal-2020-generating-fact,atanasova-etal-2020-generating, journals/corr/abs-2009-06401,journals/corr/abs-2009-06402}). The first step of automatic fact checking is claim check-worthiness detection, a text classification problem where, given a statement, one must predict if the content of that statement makes ``an assertion about the world that is checkable'' (\cite{konstantinovskiy2018towards}).
There are multiple isolated lines of research which have studied variations of this problem. \autoref{fig:check-worthy-examples} provides examples from three tasks which are studied in this work: rumour detection on Twitter~(\cite{zubiaga2016analysing,journals/ipm/ZubiagaKLPLBCA18}), check-worthiness ranking in political debates and speeches~(\cite{atanasova2018overview,elsayed2019overview,barron2020checkthat}), and citation needed detection on Wikipedia~(\cite{redi2019citation}). Each task is concerned with a shared underlying problem: detecting claims which warrant further verification. However, no work has been done to compare all three tasks to understand shared challenges in order to derive shared solutions, which could enable improving claim check-worthiness detection systems across multiple domains. 



Therefore, we ask the following main research question in this work: are these all variants of the same task, and if so, is it possible to have a unified approach to all of them? We answer this question by investigating the problem of annotator subjectivity, where annotator background and expertise causes their judgement of what is check-worthy to differ, leading to false negatives in the data~(\cite{konstantinovskiy2018towards}). Our proposed solution is \textit{Positive Unlabelled Conversion (PUC)}, an extension of Positive Unlabelled (PU) learning, which converts negative instances into positive ones based on the estimated prior probability of an example being positive. We demonstrate that a model trained using \textit{PUC} improves performance on English \textit{citation needed detection} and 
\textit{Twitter rumour detection}. We also show that by pretraining a model on citation needed detection, one can further improve results on Twitter rumour detection over a model trained solely on rumours, highlighting that a unified approach to these problems is achievable. Additionally, we show that one attains better results on 
\textit{political speeches}  check-worthiness ranking without using any form of PU learning, arguing through a dataset analysis that the labels are much more subjective than the other two tasks.


The \textbf{contributions} of this work are as follows:
\begin{enumerate}[noitemsep]
    \item The first thorough comparison of multiple claim check-worthiness detection tasks.
    \item \textit{Positive Unlabelled Conversion (PUC)}, a novel extension of PU learning to support check-worthiness detection across domains.
    \item Results demonstrating that a unified approach to check-worthiness detection is achievable for 2 out of 3 tasks, improving over the state-of-the-art for those tasks.
\end{enumerate}

\section{Related Work}

\subsection{Claim Check-Worthiness Detection}
As the first step in automatic fact checking, claim check-worthiness detection is a binary classification problem which involves determining if a piece of text makes ``an assertion about the world which can be checked''~(\cite{konstantinovskiy2018towards}). We adopt this broad definition as it allows us to perform a structured comparison of many publicly available datasets. The wide applicability of the definition also allows us to study if and how a unified cross-domain approach could be developed. 

Claim check-worthiness detection can be subdivided into three distinct domains: rumour detection on Twitter, check-worthiness ranking in political speeches and debates, and citation needed detection on Wikipedia. A few studies have been done which attempt to create full systems for mining check-worthy statements, including the works of \cite{konstantinovskiy2018towards}, ClaimRank (\cite{jaradat2018claimrank}), and ClaimBuster (\cite{hassan2017claimbuster}). They develop full software systems consisting of relevant source material retrieval, check-worthiness classification, and dissemination to the public via end-user applications. These works are focused solely on the political domain, using data from political TV shows, speeches, and debates. In contrast, in this work we study the claim check-worthiness detection problem across three domains which have publicly available data: Twitter~(\cite{zubiaga2017exploiting}), political speeches~(\cite{atanasova2018overview}), and Wikipedia~(\cite{redi2019citation}).


\paragraph{Rumour Detection on Twitter}
Rumour detection on Twitter is primarily studied using the PHEME dataset~(\cite{zubiaga2016analysing}), a set of tweets and associated threads from breaking news events which are either rumourous or not. Published systems which perform well on this task include contextual models (e.g. conditional random fields) acting on a tweet's thread~(\cite{zubiaga2017exploiting,journals/ipm/ZubiagaKLPLBCA18}), identifying salient rumour-related words~(\cite{abulaish2019graph}), and using a GAN to generate misinformation in order to improve a downstream discriminator~(\cite{ma2019detect}).

\paragraph{Political Speeches}
For political speeches, the most studied datasets come from the Clef CheckThat! shared tasks~(\cite{atanasova2018overview, elsayed2019overview,barron2020checkthat}) and ClaimRank~(\cite{jaradat2018claimrank}). The data consist of transcripts of political debates and speeches where each sentence has been annotated by an independent news or fact-checking organization for whether or not the statement should be checked for veracity. The most recent and best performing system on the data considered in this paper consists of a two-layer bidirectional GRU network which acts on both word embeddings and syntactic parse tags~(\cite{hansen2019neural}). In addition, they augment the native dataset with weak supervision from unlabelled political speeches.

\paragraph{Citation Needed Detection}
Wikipedia citation needed detection has been investigated recently in~\cite{redi2019citation}. The authors present a dataset of sentences from Wikipedia labelled for whether or not they have a citation attached to them. They also released a set of sentences which have been flagged as not having a citation but needing one (i.e. \textit{unverified}). In contrast to other check-worthiness detection domains, there are much more training data available on Wikipedia. However, the rules for what requires a citation do not necessarily capture all ``checkable'' statements, as ``all material in Wikipedia articles must be verifiable''~(\cite{redi2019citation}). 
Given this, we view Wikipedia citation data as a set of positive and unlabelled data: statements which have attached citations are positive samples of check-worthy statements, and within the set of statements without citations there exist some positive samples (those needing a citation) and some negative samples. 
Based on this, this domain constitutes the most general formulation of check-worthiness among the domains we consider. Therefore, we experiment with using data from this domain as a source for transfer learning, training variants of PU learning models on it, then applying them to target data from other domains.

\subsection{Positive Unlabelled Learning}
PU learning methods attempt to learn good binary classifiers given only positive labelled and unlabelled data. Recent applications where PU learning has been shown to be beneficial include detecting deceptive reviews online~(\cite{li2014spotting,ren2014positive}), keyphrase extraction~(\cite{sterckx2016supervised}) and named entity recognition~(\cite{peng2019distantly}). For a survey on PU learning, see~(\cite{bekker2018learning}), and for a formal definition of PU learning, see \S\ref{sec:pu_learning}.

Methods for learning positive-negative (PN) classifiers from PU data have a long history~(\cite{denis1998pac,de1999positive,letouzey2000learning}), with one of the most seminal papers being from~\cite{elkan2008learning}. In this work, the authors show that by assuming the labelled samples are a random subset of all positive samples, one can utilize a classifier trained on PU data in order to train a different classifier to predict if a sample is positive or negative. The process involves training a PN classifier with positive samples being shown to the classifier once and \textit{unlabelled} samples shown as \textit{both} a positive sample and a negative sample. The loss for the duplicated samples is weighted by the confidence of a PU classifier that the sample is positive.

Building on this, \cite{du2014analysis} propose an unbiased estimator which improves the estimator introduced in~\cite{elkan2008learning} by balancing the loss for positive and negative classes. The work of~\cite{kiryo2017positive} extends this method to improve the performance of deep networks on PU learning. Our work builds on the method of~\cite{elkan2008learning} by relabelling samples which are highly confidently positive.

\section{Methods}
\begin{figure}[t]
  \centering
    \includegraphics[width=1.0\textwidth]{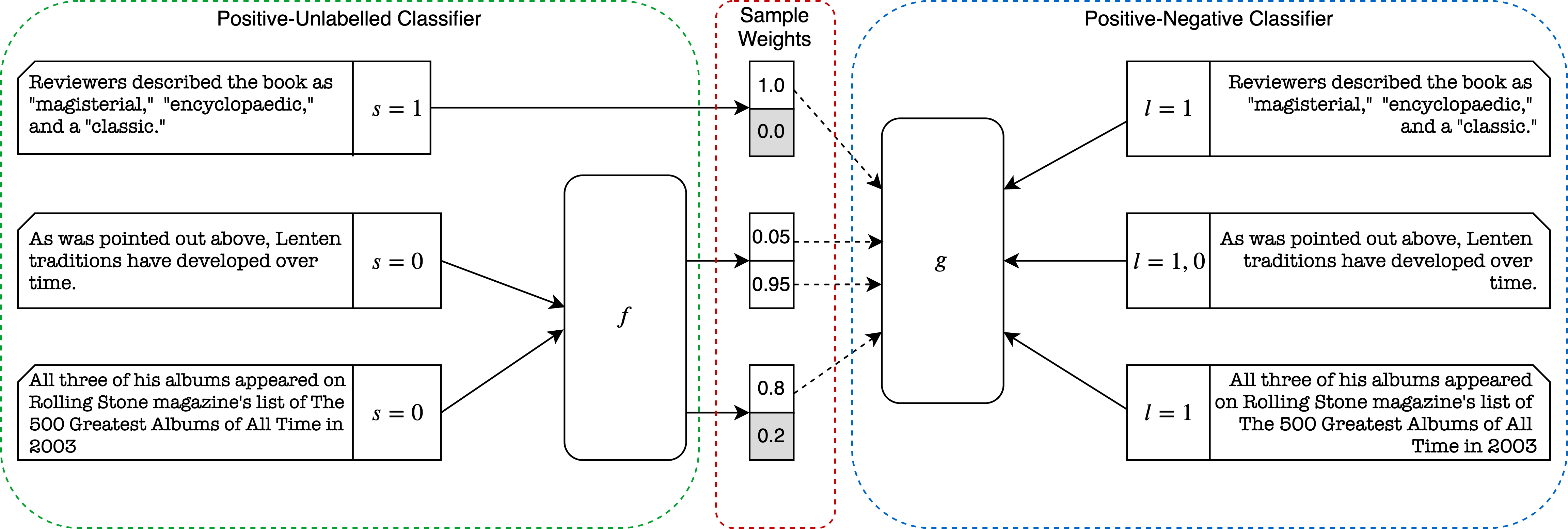}
    \caption{High level view of \textit{PUC}. A PU classifier ($f$, green box) is first learned using PU data (with $s$ indicating if the sample is positive or unlabelled). From this the prior probability of a sample being positive is estimated. Unlabelled samples are then ranked by $f$ (red box) and the most positive samples are converted into positives until the dataset is balanced according to the estimated prior. The model $g$ is then trained using the duplication and weighting method of \cite{elkan2008learning} as described in \S\ref{sec:pu_learning} with labels $l$ (blue box). Greyed out boxes are negative weights which are ignored when training the classifier $g$, as those examples are only trained as positives.}
      \label{fig:puc}
\end{figure}

The task considered in this paper is to predict if a statement makes ``an assertion about the world that is checkable'' (\cite{konstantinovskiy2018towards}). 
As the subjectivity of annotations for existing data on claim check-worthiness detection is a known problem~(\cite{konstantinovskiy2018towards}), we view the data as a set of positive and unlabelled (PU) data. In addition, we unify our approach to each of them by viewing Wikipedia data as an abundant source corpus.
Models are then trained on this source corpus using variants of PU learning and transferred via fine-tuning to the other claim check-worthiness detection datasets, which are subsequently trained on as PU data. On top of vanilla PU learning, we introduce \textit{Positive Unlabelled Conversion (PUC)} which relabels examples that are most confidently positive in the unlabelled data. A formal task definition, description of PU learning, and explanation of the \textit{PUC} extension are given in the following sections.

\subsection{Task Definition}
The fundamental task is binary text classification. In the case of positive-negative (PN) data, we have a labelled dataset $\mathcal{D}: \{(x, y)\}$ with input features $x \in \mathbb{R}^{d}$ and labels $y \in \{0, 1\}$. The goal is to learn a classifier $g: x \rightarrow (0,1)$ indicating the probability that the input belongs to the positive class. With PU data, the dataset $\mathcal{D}$ instead consists of samples $\{(x, s)\}$, where the value $s \in \{0,1\}$ indicates if a sample is labelled or not. The primary difference from the PN case is that, unlike for the labels $y$, a value of $s = 0$ does not denote the sample is negative, but that the label is unknown. The goal is then to learn a PN classifier $g$ using a PU classifier $f: x \rightarrow (0,1)$ which predicts whether or not a sample is labelled~(\cite{elkan2008learning}). 

\subsection{PU Learning}
\label{sec:pu_learning}
Our overall approach is depicted in~\autoref{fig:puc}. We begin with an explanation of the PU learning algorithm described in~\cite{elkan2008learning}. Assume that we have a dataset randomly drawn from some probability distribution $p(x,y,s)$, where samples are of the form $(x,s), ~s \in \{0,1\}$ and $s=1$ indicates that the sample is labelled. The variable $y$ is unknown, but we make two assumptions which allow us to derive an estimator for probabilities involving $y$. The first is that:
\begin{equation}
    p(y = 0 | s = 1) = 0
\end{equation}
In other words, if we know that a sample is labelled, then that label cannot be 0. The second assumption is that labelled samples are Selected Completely At Random from the underlying distribution (also known as the SCAR assumption). Check-worthiness data can be seen as an instance of SCAR PU data; annotators tend to only label those instances which are very clearly check-worthy in \textit{their} opinion~(\cite{konstantinovskiy2018towards}). When combined across several annotators, we assume this leads to a random sample from the total set of check-worthy statements.

Given this, a classifier $f : x \rightarrow (0,1)$ is trained to predict $p(s=1|x)$ from the PU data. It is then employed to train a classifier $g$ to predict $p(y=1|x)$ by first estimating $c = p(s=1|y=1)$ on a set of validation data. Considering a validation set $V$ where $P \subset V$ is the set of positive samples in $V$, $c$ is estimated as:
\begin{equation}
\label{eq:c_estimate}
    c \approx \frac{1}{|P|}\sum_{x \in P}f(x)
\end{equation}
This says our estimate of $p(s=1|y=1)$ is the average confidence of our classifier on known positive samples. Next, we can estimate $E_{p(x,y,s)}[h(x,y)]$ for any arbitrary function $h$ empirically from a dataset of $k$ samples as follows:
\begin{dmath}
    E[h] = \frac{1}{k}(\sum_{(x,s=1)}h(x,1) + \sum_{(x,s=0)}w(x)h(x,1) + (1-w(x))h(x,0))
\end{dmath}
\begin{align}
    w(x) &= p(y=1|x,s=0) = \frac{1-c}{c}\frac{p(s=1|x)}{1-p(s=1|x)}
\end{align}
In this case, $c$ is estimated using~\autoref{eq:c_estimate} and $p(s=1|x)$ is estimated using the classifier $f$. The derivations for these equations can be found in~\cite{elkan2008learning}.

To estimate $p(y=1|x)$ empirically, the unlabelled samples in the training data are duplicated, with one copy negatively labelled and one copy positively labelled. Each copy is trained on with a weighted loss $w(x)$ when the label is positive and $1 - w(x)$ when the label is negative. Labelled samples are trained on normally (i.e. a single copy with unit weight).

\subsection{Positive Unlabelled Conversion}
For \textit{PUC}, the motivation is to relabel those samples from the unlabelled data which are very clear cut positive. To accomplish this, we start with the fact that one can also estimate the prior probability of a sample having a positive label using $f$. If instead of $h$ we want to estimate $E[y] = p(y=1)$, the following is obtained:
\begin{equation}
    p(y=1) \approx \frac{1}{k}(\sum_{x,s=1}1 + \sum_{x,s=0}w(x))
\end{equation}
This estimate is then utilized to convert the most confident unlabelled samples into positives. First, all of the unlabelled samples are ranked according to their calculated weight $w(x)$. The ranked samples are then iterated through and converted into positive-only samples until the distribution of positive samples is greater than or equal to the estimate of $p(y=1)$. Unlike in vanilla PU learning, these samples are discretized to have a positive weight of 1, and trained on by the classifier $g$ once per epoch as positive samples along with the labelled samples. The remaining unlabelled data are trained on in the same way as in vanilla PU learning. 

\subsection{Implementation}
In order to create a unified approach to check-worthiness detection, transfer learning from Wikipedia citation needed detection is employed. 
To accomplish this, we start with a training dataset $\mathcal{D}^{s}$ of statements from Wikipedia featured articles that are either labelled as containing a citation (positive) or unlabelled. We train a classifier $f^{s}$ on this dataset and obtain a classifier $g^{s}$ via \textit{PUC}. For comparison, we also train models with vanilla PU learning and PN learning as baselines. The network architecture for both $f^{s}$ and $g^{s}$ is BERT~(\cite{devlin-etal-2019-bert}), a large pretrained transformer-based~(\cite{vaswani2017}) language model. We use the HuggingFace transformers implementation of the 12-layer 768 dimensional variation of BERT~(\cite{Wolf2019HuggingFacesTS}). The classifier in this implementation is a two layer neural network acting on the \texttt{[CLS]} token.

From $g^{s}$, we train a classifier $g^{t}$ using downstream check-worthiness detection dataset $D^{t}$ by initializing $g^{t}$ with the base BERT network from $g^{s}$ and using a new randomly initialized final layer. In addition, we train a model $f^{t}$ on the target dataset, and train $g^{t}$ with \textit{PUC} from this model to obtain the final classifier. As a baseline, we also experiment with training on just the dataset $D^{t}$ without any pretraining. In the case of citation needed detection, since the data comes from the same domain we simply test on the test split of statements labelled as ``citation needed'' using the classifier $g^{s}$. We compare our models to the published state of the art baselines on each dataset.

For all of our models ($f^s$, $g^s$, $f^t$, $g^t$) we train for two epochs, saving the weights with the best F1 score on validation data as the final model. Training is performed with a max learning rate of 3e-5 and a triangular learning rate schedule~(\cite{howard-ruder-2018-universal}) that linearly warms up for 200 training steps, then linearly decays to 0 for the rest of training. For regularization we add L2 loss with a coefficient of 0.01, and dropout with a rate of 0.1. Finally, we split the training sets into 80\% train and 20\% validation, and train with a batch size of 8. 
The code to reproduce our experiments can be found here.\footnote{\url{https://github.com/copenlu/check-worthiness-pu-learning}} 

\section{Experimental Results}

\begin{table}
    \centering
    \fontsize{10}{10}\selectfont
    \begin{tabular}{l c c c | c c c} 
    \toprule 
    Method & P & R & \multicolumn{1}{c}{F1} & eP & eR & eF1\\
    \midrule 
       \cite{redi2019citation}  & 75.3& 70.9& 73.0 [76.0]*& - & - & -\\
    \midrule 
       \rule{0pt}{2ex}BERT  & \underline{78.8 $\pm$ 1.3}& 83.7 $\pm$ 4.5& 81.0 $\pm$ 1.5 & 79.0 & 85.3 & 82.0 \\
       BERT + PU  & \textbf{78.8 $\pm$ 0.9}& \underline{84.3 $\pm$ 3.0}& \underline{81.4 $\pm$ 1.0} & 79.0 & \underline{85.6} & \underline{82.2}\\
       BERT + \textit{PUC}  & 78.4 $\pm$ 0.9& \textbf{85.6 $\pm$ 3.2}& \textbf{81.8 $\pm$ 1.0} & 78.6 & \textbf{87.1} & \textbf{82.6}\\
    \bottomrule 

    \end{tabular}
    \caption{\label{tab:citation_detection_results}F1 and ensembled F1 score for citation needed detection training on the FA split and testing on the LQN split of \cite{redi2019citation}. The FA split contains statements with citations from featured articles and the LQN split consists of statements which were flagged as not having a citation but needing one. Listed are the mean, standard deviation, and ensembled results across 15 seeds (eP, eR, and eF1). \textbf{Bold} indicates best performance, \underline{underline} indicates second best. *The reported value is from rerunning their released model on the test dataset. The value in brackets is the value reported in the original paper.}
\end{table}

To what degree is claim check-worthiness detection a PU learning problem, and does this enable a unified approach to check-worthiness detection? In our experiments, we progressively answer this question by answering the following: 1) is PU learning beneficial for the tasks considered? 2) Does PU citation needed detection transfer to rumour detection? 3) Does PU citation needed detection transfer to political speeches? To investigate how well the data in each domain reflects the definition of a check-worthy statement as one which ``makes an assertion about the world which is checkable'' and thus understand subjectivity in the annotations, we perform a dataset analysis comparing the provided labels of the top ranked check-worthy claims from the \textit{PUC} model with the labels given by two human annotators. In all experiments, we report the mean performance of our models and standard deviation across 15 different random seeds. Additionally, we report the performance of each model ensembled across the 15 runs through majority vote on each sample.

\subsection{Datasets}

See supplemental material for links to datasets.

\paragraph{Wikipedia Citations}
We use the dataset from \cite{redi2019citation} for citation needed detection. The dataset is split into three sets: one coming from featured articles (deemed `high quality', 10k positive and 10k negative statments), one of statements which have no citation but have been flagged as needing one (10k positive, 10k negative), and one of statements from random articles which have citations (50k positive, 50k negative). In our experiments the models were trained on the high quality statements from featured articles and tested on the statements which were flagged as `citation needed'. The key differentiating features of this dataset from the other two datasets are: 1) the domain of text is Wikipedia and 2) annotations are based on the decisions of Wikipedia editors following Wikipedia guidelines for citing sources\footnote{\url{https://en.wikipedia.org/wiki/Wikipedia:Citing_sources}}.

\paragraph{Twitter Rumours}
The PHEME dataset of rumours is employed for Twitter claim check-worthiness detection~(\cite{zubiaga2016analysing}). The data consists of 5,802 annotated tweets from 5 different events, where each tweet is labelled as rumourous or non-rumourous (1,972 rumours, 3,830 non-rumours). We followed the leave-one-out evaluation scheme of~(\cite{zubiaga2017exploiting}), namely, we performed a 5-fold cross-validation for all methods, training on 4 events and testing on 1. The key differentiating features of this dataset from the other two datasets are: 1) the domain of data is tweets and 2) annotations are collected from professional journalists specifically for building a dataset to train machine learning models.

\paragraph{Political Speeches}
The dataset we adopted in the political speeches domain is the same as in~\cite{hansen2019neural}, consisting of 4 political speeches from the 2018 Clef CheckThat! competition~(\cite{atanasova2018overview}) and 3 political speeches from ClaimRank~(\cite{jaradat2018claimrank}) (2,602 statements total). We performed a 7-fold cross-validation, using 6 splits as training data and 1 as test in our experimental setup. The data from ClaimRank is annotated using the judgements from 9 fact checking organizations, and the data from Clef 2018 is annotated by factcheck.org. The key differentiating features of this dataset from the other two datasets are: 1) the domain of data is transcribed spoken utterances from political speeches and 2) annotations are taken from 9 fact checking organizations gathered independently.

\subsection{Is PU Learning Beneficial for Citation Needed Detection?}
\begin{table}
    \centering
    \fontsize{10}{10}\selectfont
    \begin{tabular}{l c c c | c c c} 
    \toprule
    Method & $\mu$P & $\mu$R & \multicolumn{1}{c}{$\mu$F1}& eP & eR & eF1\\
    \midrule
       \cite{zubiaga2017exploiting}  & 66.7& 55.6& 60.7 & - & - & -\\
       BiLSTM & 62.3 &	56.4 &	59.0 & - & - & -\\
    \midrule 
       \rule{0pt}{2ex}BERT  & \underline{69.9 $\pm$ 1.7}& 60.8 $\pm$ 2.6& 65.0 $\pm$ 1.3 & 71.3 & 61.9	& 66.3\\
       BERT + Wiki  & 69.3 $\pm$ 1.6& 61.4 $\pm$ 2.6& 65.1 $\pm$ 1.2 & 70.7 & 62.2 & 66.2\\
       BERT + WikiPU  & \underline{69.9 $\pm$ 1.3}& 62.5 $\pm$ 1.6& 66.0 $\pm$ 1.1 & \textbf{72.2} & 64.6 & 68.2\\
       BERT + Wiki\textit{PUC}  & \textbf{70.1 $\pm$ 1.1}& 61.8 $\pm$ 1.8& 65.7 $\pm$ 1.0 & \underline{71.5} & 62.7 & 66.8\\
       BERT + PU  & 68.7 $\pm$ 1.2& 64.7 $\pm$ 1.8& 66.6 $\pm$ 0.9 & 69.9 & 65.2 & 67.5\\
       BERT + \textit{PUC}  & 68.1 $\pm$ 1.5& 65.3 $\pm$ 1.6& 66.6 $\pm$ 0.9 & 69.1 & 66.3 & 67.7\\
       BERT + PU + WikiPU  & 68.4 $\pm$ 1.2& \textbf{66.1 $\pm$ 1.2}& \textbf{67.2 $\pm$ 0.6} & 69.3 & \underline{67.2} & \underline{68.3}\\
       BERT + \textit{PUC} + WikiPUC  & 68.0 $\pm$ 1.4& \underline{66.0 $\pm$ 2.0}& \underline{67.0 $\pm$ 1.3} & 69.4 & \textbf{67.5} & \textbf{68.5}\\
    \bottomrule

    \end{tabular}
    \caption{micro-F1 ($\mu$F1) and ensembled F1 (eF1) performance of each system on the PHEME dataset. Performance is averaged across the five splits of~\cite{zubiaga2017exploiting}. Results show the mean, standard deviation, and ensembled score across 15 seeds. \textbf{Bold} indicates best performance, \underline{underline} indicates second best.}
    \label{tab:pheme_results}
\end{table}


Our results for citation needed detection are given in \autoref{tab:citation_detection_results}. The vanilla BERT model already significantly outperforms the state of the art model from~\cite{redi2019citation} (a GRU network with global attention) by 6 F1 points. We see further gains in performance with PU learning, as well as when using \textit{PUC}. Additionally, the models using PU learning have lower variance, indicating more consistent performance across runs. The best performing model we see is the one trained using \textit{PUC} with an F1 score of 82.6. We find that this confirms our hypothesis that citation data is better seen as a set of positive and unlabelled data when used for check-worthiness detection. In addition, it gives some indication that PU learning improves the generalization power of the model, which could make it better suited for downstream tasks.


\subsection{Does PU Citation Needed Detection Transfer to Rumour Detection?}
\subsubsection{Baselines}
The best published method that we compare to is the CRF from~\cite{zubiaga2017exploiting}. which utilizes a combination of content and social features. Content features include word vectors, part-of-speech tags, and various lexical features, and social features include tweet count, listed count, follow ratio, age, and whether or not a user is verified. The CRF acts on a timeline of tweets, making it contextual. In addition, we include results from a 2-layer BiLSTM with FastText embeddings~(\cite{bojanowski2017enriching}). There exist other deep learning models which have been developed for this task, including \cite{ma2019detect} and \cite{abulaish2019graph}, but they do not publish results on the standard splits of the data and we were unable to recreate their results, and thus are omitted.

\subsubsection{Results}
The results for the tested systems are given in \autoref{tab:pheme_results}. Again we see large gains from BERT based models over the baseline from \cite{zubiaga2017exploiting} and the 2-layer BiLSTM. Compared to training solely on PHEME, fine tuning from basic citation needed detection sees little improvement (0.1 F1 points). However, fine tuning a model trained using PU learning leads to an increase of 1 F1 point over the non-PU learning model, indicating that PU learning enables the Wikipedia data to be useful for transferring to rumour detection i.e. the improvement is not only from a better semantic representation learned from Wikipedia data. For \textit{PUC}, we see an improvement of 0.7 F1 points over the baseline and lower overall variance than vanilla PU learning, meaning that the results with \textit{PUC} are more consistent across runs. The best performing models also use PU learning on in-domain data, with the best average performance being from the models trained using PU/\textit{PUC} on in domain data and initialized with weights from a Wikipedia model trained using PU/\textit{PUC}. When models are ensembled, pretraining with vanilla PU learning improves over no pretraining by almost 2 F1 points, and the best performing models which are also trained using PU learning on in domain data improve over the baseline by over 2 F1 points. We conclude that framing rumour detection on Twitter as a PU learning problem leads to improved performance.




Based on these results, we are able to confirm two of our hypotheses. The first is that Wikipedia citation needed detection and rumour detection on Twitter are indeed similar tasks, and a unified approach for both of them is possible. Pretraining a model on Wikipedia provides a clear downstream benefit when fine-tuning on Twitter data, \textit{precisely when PU/PUC is used}. Additionally, training using \textit{PUC} on in domain Twitter data provides further benefit. This shows that \textit{PUC} constitutes a unified approach to these two tasks.

The second hypothesis we confirm is that both Twitter and Wikipedia data are better seen as positive and unlabelled for claim check-worthiness detection. When pretraining with the data as a traditional PN dataset there is no performance gain and in fact a performance loss when the models are ensembled. PU learning allows the model to learn better representations for general claim check-worthiness detection.

To explain why this method performs better, \autoref{tab:citation_detection_results} and \autoref{tab:pheme_results} show that \textit{PUC} improves model recall at very little cost to precision. The aim of this is to mitigate the issue of subjectivity in the annotations of check-worthiness detection datasets noted in previous work \cite{konstantinovskiy2018towards}. Some of the effects of this are illustrated in \autoref{tab:pheme_pos_better} and \autoref{tab:pheme_neg_better} in \autoref{sec:puc_examples}. The \textit{PUC} models are better at distinguishing rumours which involve claims of fact about people i.e. things that people said or did, or qualities about people. For non-rumours, the \textit{PUC} pretrained model is better at recognizing statements which describe qualitative information surrounding the events and information that is self-evident e.g. a tweet showing the map where the Charlie Hebdo attack took place.

\subsection{Does PU Citation Needed Detection Transfer to Political Speeches?}
\subsubsection{Baselines}
The baselines we compare to are the state of the art models from ~\cite{hansen2019neural} and \cite{konstantinovskiy2018towards}. The model from \cite{konstantinovskiy2018towards} consists of InferSent embeddings~(\cite{conneau2017supervised}) concatenated with POS tag and NER features passed through a logistic regression classifier. The model from \cite{hansen2019neural} is a bidirectional GRU network acting on syntatic parse features concatenated with word embeddings as the input representation.

\subsubsection{Results}
The results for political speech check-worthiness detection are given in \autoref{tab:clef_results}. We find that the BERT model initialized with weights from a model trained on plain Wikipedia citation needed statements performs the best of all models. As we add transfer learning and PU learning, the performance steadily drops. We perform a dataset analysis to gain some insight into this effect in \S\ref{sec:dataset_analysis}.
\begin{table}
    \centering
    \fontsize{10}{10}\selectfont
    \begin{tabular}{l c}
    \toprule
    Method & MAP\\
    \midrule
    \cite{konstantinovskiy2018towards} & 26.7\\
    \cite{hansen2019neural}  & 30.2\\
    \midrule
       \rule{0pt}{2ex}BERT  &33.0 $\pm$ 1.8\\
       BERT + Wiki  & \textbf{34.4 $\pm$ 2.7}\\
       BERT + WikiPU  & \underline{33.2 $\pm$ 1.7}\\
       BERT + Wiki\textit{PUC}  & 31.7 $\pm$ 1.8\\
       BERT + PU  & 18.8 $\pm$ 3.7\\
       BERT + \textit{PUC}  & 26.7 $\pm$ 2.8\\
       BERT + PU + WikiPU  & 16.8 $\pm$ 3.5\\
       BERT + \textit{PUC} + Wiki\textit{PUC}  & 27.8 $\pm$ 2.7\\
    \bottomrule

    \end{tabular}
    \caption{Mean average precision (MAP) of models on political speeches. \textbf{Bold} indicates best performance, \underline{underline} indicates second best.}
    \label{tab:clef_results}
\end{table}

\subsection{Dataset Analysis}
\label{sec:dataset_analysis}
In order to understand our results in the context of the selected datasets, we perform an analysis to learn to what extent the positive samples in each dataset reflect the definition of a check-worthy claim as ``an assertion about the world that is checkable''. We ranked all of the statements based on the predictions of 15 \textit{PUC} models trained with different seeds, where more positive class predictions means a higher rank (thus more check-worthy), and had two experts manually relabel the top 100 statements. The experts were informed to label the statements based on the definition of check-worthy given above. We then compared the manual annotation to the original labels using F1 score. Higher F1 score indicates the dataset better reflects the definition of check-worthy we adopt in this work. Our results are given in \autoref{tab:relabel_results}.
\begin{table}
    \centering
    \fontsize{10}{10}\selectfont
    \begin{tabular}{l c c c}
    \toprule
    Dataset & P & R & F1\\
    \midrule
                  &  81.7 &87.0 &84.3\\
       Wikipedia  & 84.8 & 87.0 & 85.9\\
                  & \textit{83.3}& \textit{87.0}& \textit{85.1}\\
       \midrule
        \rule{0pt}{2ex}  & 87.5& 82.4& 84.8 \\
       Twitter  & 86.3 & 81.2 & 83.6\\
                & \textit{86.9}& \textit{81.8}& \textit{84.2} \\
       \midrule
        \rule{0pt}{2ex} &33.8& 89.3& 49.0\\
       Politics  &31.1&  100.0&  47.5\\
                &\textit{32.5} &\textit{94.7}& \textit{48.3}\\
    \bottomrule

    \end{tabular}
    \caption{F1 score comparing manual relabelling of the top 100 predictions by \textit{PUC} model with the original labels in each dataset by two different annotators. \textit{Italics} are average value between the two annotators.}
    \label{tab:relabel_results}
\end{table}

We find that the Wikipedia and Twitter datasets contain labels which are more general, evidenced by similar high F1 scores from both annotators ($>$ 80.0). For political speeches, we observe that the human annotators both found many more examples to be check-worthy than were labelled in the dataset. This is evidenced by examples such as \textit{It's why our unemployment rate is the lowest it's been in so many decades} being labelled as not check-worthy and \textit{New unemployment claims are near the lowest we've seen in almost half a century} being labelled as check-worthy in the same document in the dataset's original annotations. This characteristic has been noted for political debates data previously~(\cite{konstantinovskiy2018towards}), which was also collected using the judgements of independent fact checking organizations~(\cite{gencheva-etal-2017-context-fixed}). Labels for this dataset were collected from various news outlets and fact checking organizations, which may only be interested in certain types of claims such as those most likely to be false. This makes it difficult to train supervised machine learning models for general check-worthiness detection based solely on text content and document context due to labelling inconsistencies. 

\section{Discussion and Conclusion}
In this work, we approached claim check-worthiness detection by examining how to unify three distinct lines of work. We found that check-worthiness detection is challenging in any domain as there exist stark differences in how annotators judge what is check-worthy. We showed that one can correct for this and improve check-worthiness detection across multiple domains by using positive unlabelled learning. Our method enabled us to perform a structured comparison of datasets in different domains, developing a unified approach which outperforms state of the art in 2 of 3 domains and illuminating to what extent these datasets reflect a general definition of check-worthy. 

Future work could explore different neural base architectures. Further, 
it could potentially benefit all tasks to consider the greater context in which statements are made. We would also like to acknowledge again that all experiments have only focused on English language datasets; developing models for other, especially low-resource languages, would likely result in additional challenges. We hope that this work will inspire future research on check-worthiness detection, which we see as an under-studied problem, with a focus on developing resources and models across many domains such as Twitter, news media, and spoken rhetoric.

\section*{Acknowledgements}
$\begin{array}{l}\includegraphics[width=1cm]{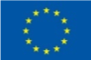} \end{array}$ This project has received funding from the European Union's Horizon 2020 research and innovation programme under the Marie Sk\l{}odowska-Curie grant agreement No 801199.

\newpage

\section{Appendix}

\subsection{Examples of PUC Improvements for Rumour Detection}
\label{sec:puc_examples}

Examples of improvements for rumour detection using \textit{PUC} can be found in \autoref{tab:pheme_pos_better}.
\begin{table}
    \begin{tabular}{p{12cm} c c}
    \toprule
    \multicolumn{1}{c}{Rumour text} & nPUC & \multicolumn{1}{c}{nBaseline} \\
    \midrule
       Germanwings co-pilot had serious depressive episode: Bild newspaper http://t.co/RgSTrehD21 & 13 & 5\\
       \hline
       Now hearing 148 passengers + crew on board the \#A320 that has crashed in southern French Alps. \#GermanWings flight. @BBCWorld & 10 & 2 \\
       \hline
       It appears that \#Ferguson PD are trying to assassinate Mike Brown's character after literally assassinating Mike Brown. & 13 & 5 \\
       \hline
       \#Ferguson cops beat innocent man then charged him for bleeding on them: http://t.co/u1ot9Eh5Cq via @MichaelDalynyc http://t.co/AGJW2Pid1r & 9 & 2\\
    \bottomrule
    \end{tabular}
    \caption{\label{tab:pheme_pos_better}Examples of rumours which the \textit{PUC} model judges correctly vs the baseline model with no pretraining on citation needed detection. n* is the number of models among the 15 seeds which predicted the correct label (rumour).}
\end{table}

\begin{table}
    \begin{tabular}{p{12cm} c c}
    \toprule
    \multicolumn{1}{c}{Non-Rumour text} & nPUC & \multicolumn{1}{c}{nBaseline} \\
    \midrule
       A female hostage stands by the front entrance of the cafe as she turns the lights off in Sydney. \#sydneysiege http://t.co/qNfCMv9yZt & 11 & 5\\
       \hline
       Map shows where gun attack on satirical magazine \#CharlieHebdo took place in central Paris http://t.co/5AZAKumpNd http://t.co/ECFYztMVk9 & 10 & 4 \\
       \hline
       "Hands up! Don't shoot!" \#ferguson https://t.co/svCE1S0Zek & 12 & 7 \\
       \hline
       Australian PM Abbott: Motivation of perpetrator in Sydney hostage situation is not yet known - @9NewsAUS http://t.co/SI01B997xf & 10 & 6\\
    \bottomrule
    \end{tabular}
    \caption{\label{tab:pheme_neg_better}Examples of non-rumours which the \textit{PUC} model judges correctly vs the baseline model with no pretraining on citation needed detection. n* is the number of models among the 15 seeds which predicted the correct label (non-rumour).}
\end{table}

\subsection{Reproducibility}
\subsubsection{Computing Infrastructure}
All experiments were run on a shared cluster. Requested jobs consisted of 16GB of RAM and 4 Intel Xeon Silver 4110 CPUs. We used a single NVIDIA Titan X GPU with 12GB of RAM.

\subsubsection{Average Runtimes}
See \autoref{tab:runtimes} for model runtimes.
\begin{table}
    \centering
    \fontsize{10}{10}\selectfont
    \begin{tabular}{l c c c}
    \toprule
    Method & Wikipedia & PHEME & Political Speeches\\
    \midrule
       \rule{0pt}{2ex}BERT&  34m30s& 14m25s& 8m11s\\
       BERT + PU& 40m7s& 20m40s&  15m38s\\
       BERT + \textit{PUC}& 40m8s& 21m20s& 15m32s\\
       BERT + Wiki& - & 14m28s& 8m50s\\
       BERT + WikiPU& -& 14m25s& 8m41s\\
       BERT + Wiki\textit{PUC}& -& 14m28s& 8m38s\\
       BERT + PU + WikiPU& -& 20m41s& 15m32s\\
       BERT + \textit{PUC} + WikiPUC& -& 21m52s& 15m40s\\
    \bottomrule

    \end{tabular}
    \caption{Average runtime of each tested system for each split of the data}
    \label{tab:runtimes}
\end{table}

\subsubsection{Number of Parameters per Model}
We used BERT with a classifier on top for each model which consists of 109,483,778 parameters.

\subsubsection{Validation Performance}
Validation performances for the tested models are given in \autoref{tab:validation}.
\begin{table}
    \centering
    \fontsize{10}{10}\selectfont
    \begin{tabular}{l c c c}
    \toprule
    Method & Wikipedia & PHEME & Political Speeches\\
    \midrule
       \rule{0pt}{2ex}BERT&  88.9& 81.6& 31.3\\
       BERT + PU& 89.0& 83.7& 18.2\\
       BERT + \textit{PUC}& 89.2& 82.8& 32.0\\
       BERT + Wiki& - & 80.8& 32.3\\
       BERT + WikiPU& -& 82.0& 35.7\\
       BERT + Wiki\textit{PUC}& -& 80.4& 34.3\\
       BERT + PU + WikiPU& -& 82.9& 33.3\\
       BERT + \textit{PUC} + WikiPUC& -& 84.1& 34.0\\
    \bottomrule

    \end{tabular}
    \caption{Validation F1 performances for each tested model.}
    \label{tab:validation}
\end{table}

\subsubsection{Evaluation Metrics}
The primary evaluation metric used was F1 score. We used the sklearn implementation of \texttt{precision\_recall\_fscore\_support}, which can be found here: \url{https://scikit-learn.org/stable/modules/generated/sklearn.metrics.precision_recall_fscore_support.html}. Briefly:
\begin{equation*}
   p = \frac{tp}{tp + fp} 
\end{equation*}
\begin{equation*}
   r = \frac{tp}{tp + fn} 
\end{equation*}
\begin{equation*}
   F1 = \frac{2*p*r}{p+r} 
\end{equation*}
where $tp$ are true positives, $fp$ are false positives, and $fn$ are false negatives.

Additionally, we used the mean average precision calculation from the Clef19 Check That! challenge for political speech data, which can be found here: \url{https://github.com/apepa/clef2019-factchecking-task1/tree/master/scorer}. Briefly:
\begin{equation*}
    \text{AP} = \frac{1}{|P|}\sum_{i}\frac{tp(i)}{i}
\end{equation*}
\begin{equation*}
    \text{mAP} = \frac{1}{|Q|}\sum_{q\in Q}\text{AP}(q)
\end{equation*}
where $P$ are the set of positive instances, $tp(i)$ is an indicator function which equals one when the $i$th ranked sample is a true positive, and $Q$ is the set of queries. In this work $Q$ consists of the ranking of statements from each split of the political speech data.

\subsubsection{Links to Data}
\begin{itemize}
    \item Citation Needed Detection~(\cite{redi2019citation}):  \url{https://drive.google.com/drive/folders/1zG6orf0_h2jYBvGvso1pSy3ikbNiW0xJ}
    
    \item PHEME~(\cite{zubiaga2016analysing}): \url{https://figshare.com/articles/PHEME_dataset_for_Rumour_Detection_and_Veracity_Classification/6392078}.
    
    \item Political Speeches: We use the same 7 splits as used in~\cite{hansen2019neural}. The first 5 can be found here: \url{http://alt.qcri.org/clef2018-factcheck/data/uploads/clef18_fact_checking_lab_submissions_and_scores_and_combinations.zip}. The files can be found under ``task1\_test\_set/English/task1-en-file(3,4,5,6,7)''. The last two files can be found here: \url{https://github.com/apepa/claim-rank/tree/master/data/transcripts_all_sources}. The files are ``clinton\_acceptance\_speech\_ann.tsv'' and ``trump\_inauguration
    \_ann.tsv''.
\end{itemize}

\subsubsection{Hyperparameters}
We found that good defaults worked well, and thus did not perform hyperparameter search. The hyperparameters we used are given in \autoref{tab:hyperparams}.

\begin{table}
    \centering
    \fontsize{10}{10}\selectfont
    \begin{tabular}{l c}
    \toprule
    Hyperparameter & Value \\
    \midrule
      Learning Rate& 3e-5 \\
      Weight Decay& 0.01 \\
      Batch Size& 8 \\
      Dropout& 0.1 \\
      Warmup Steps& 200 \\
      Epochs& 2 \\
    \bottomrule

    \end{tabular}
    \caption{Validation F1 performances used for each tested model.}
    \label{tab:hyperparams}
\end{table}
\chapter{Transformer Based Multi-Source Domain Adaptation}\label{ch:domain}

\boxabstract{
In practical machine learning settings, the data on which a model must make predictions often come from a different distribution than the data it was trained on. 
Here, we investigate the problem of \textit{unsupervised multi-source domain adaptation}, where a model is trained on labelled data from multiple source domains and must make predictions on a domain for which no labelled data has been seen. 
Prior work with CNNs and RNNs has demonstrated the benefit of mixture of experts, where the predictions of multiple domain expert classifiers are combined; as well as domain adversarial training, to induce a domain agnostic representation space. Inspired by this, we investigate how such methods can be effectively applied to large pretrained transformer models. 
We find that domain adversarial training has an effect on the learned representations of these models while having little effect on their performance, suggesting that large transformer-based models are already relatively robust across domains.
Additionally, we show that mixture of experts leads to significant performance improvements by comparing several variants of mixing functions, including one novel mixture based on attention. 
Finally, we demonstrate that the predictions of large pretrained transformer based domain experts are highly homogenous, making it challenging to learn effective functions for mixing their predictions.
}\blfootnote{\fullcite{wright-augenstein-2020-transformer}}

\section{Introduction}
\begin{figure}[t]
  
  \centering
    \includegraphics[width=0.7\columnwidth]{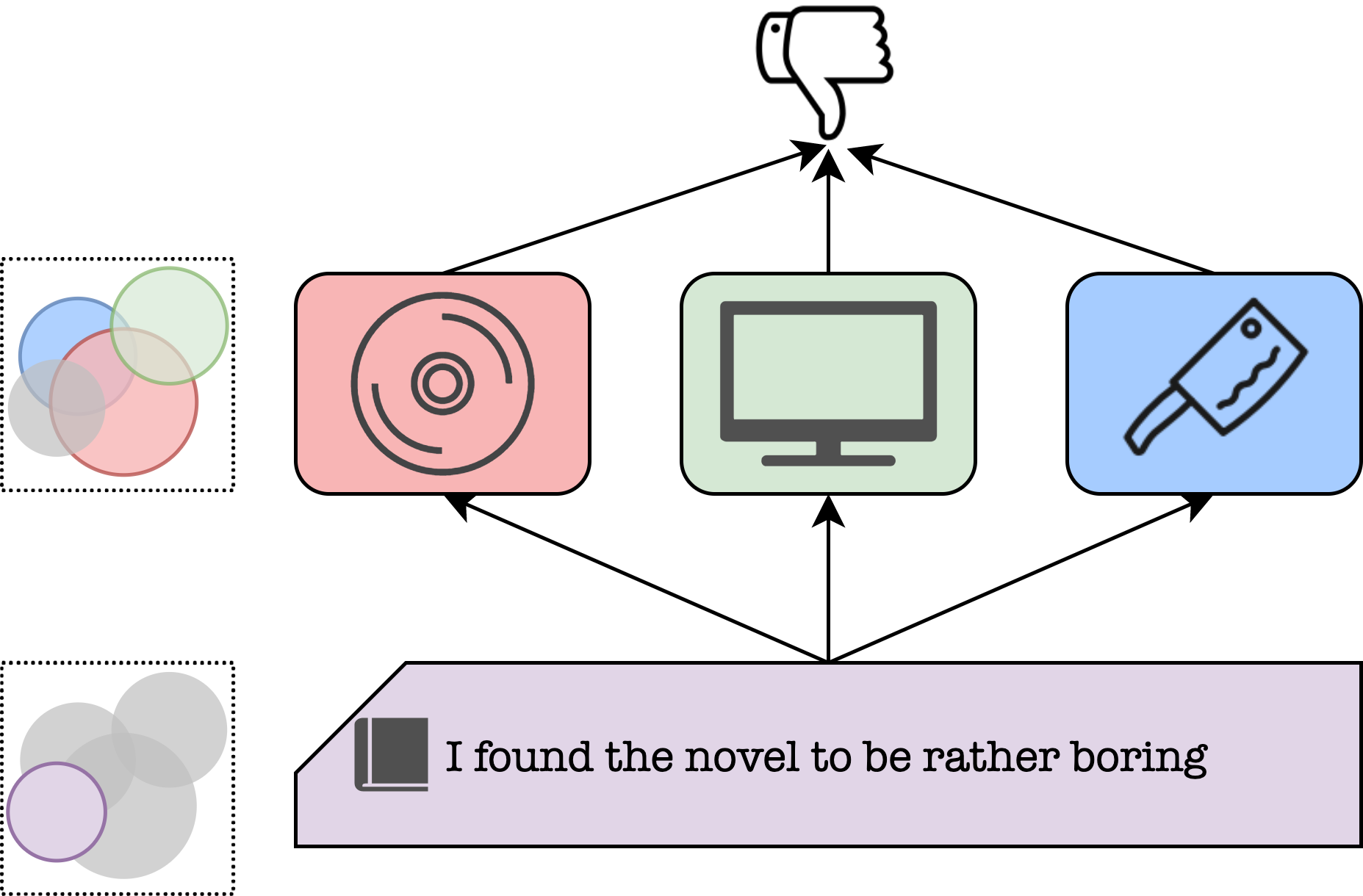}
    \caption{In multi-source domain adaptation, a model is trained on data drawn from multiple parts of the underlying distribution. At test time, the model must make predictions on data from a potentially non-overlapping part of the distribution.}
    \label{fig:msda}
\end{figure}
Machine learning practitioners are often faced with the problem of evolving test data, leading to mismatches in training and test set distributions.
As such, the problem of \textit{domain adaptation} is of particular interest to the natural language processing community in order to build models which are robust this shift in distribution. For example, a model may be trained to predict the sentiment of product reviews for DVDs, electronics, and kitchen goods, and must utilize this learned knowledge to predict the sentiment of a review about a book (\autoref{fig:msda}). This paper is concerned with this setting, namely \textit{unsupervised multi-source domain adaptation}.

Multi-source domain adaptation is a well studied problem in deep learning for natural language processing. Prominent techniques are generally based on data selection strategies and representation learning. For example, a popular representation learning method is to induce domain invariant representations using unsupervised target data and domain adversarial learning~(\cite{ganin2015unsupervised}). Adding to this, mixture of experts techniques attempt to learn both domain specific and global shared representations and combine their predictions~(\cite{guo2018multi,li2018s,ma2019domain}). These methods have been primarily studied using convolutional nets (CNNs) and recurrent nets (RNNs) trained from scratch, while the NLP community has recently begun to rely more and more on large pretrained transformer (LPX) models e.g. BERT~(\cite{devlin-etal-2019-bert}). 
To date there has been some preliminary investigation of how LPX models perform under domain shift in the single source-single target setting~(\cite{ma2019domain,han2019unsupervised,rietzler2019adapt,gururangan2020don}). What is lacking is a study into the effects of and best ways to apply classic multi-source domain adaptation techniques with LPX models, which can give insight into possible avenues for improved application of these models in settings where there is domain shift.


Given this, we present a study into unsupervised multi-source domain adaptation techniques for large pretrained transformer models. 
Our main research question is: do mixture of experts and domain adversarial training offer any benefit when using LPX models? The answer to this is not immediately obvious, as such models have been shown to generalize quite well across domains and tasks while still learning representations which are not domain invariant. Therefore, we experiment with four mixture of experts models, including one novel technique based on attending to different domain experts; as well as domain adversarial training with gradient reversal. 
Surprisingly, we find that, while domain adversarial training helps the model learn more domain invariant representations, this does not always result in increased target task performance.
When using mixture of experts, we see significant gains on out of domain rumour detection, and some gains on out of domain sentiment analysis. Further analysis reveals that the classifiers learned by domain expert models are highly homogeneous, making it challenging to learn a better mixing function than simple averaging.



\section{Related Work}
Our primary focus is multi-source domain adaptation with LPX models. We first review domain adaptation in general, followed by studies into domain adaptation with LPX models.

\subsection{Domain Adaptation}
Domain adaptation approaches generally fall into three categories: \textit{supervised} approaches (e.g. \cite{daumeiii:2007:ACLMain,finkel-manning-2009-hierarchical-fixed,conf/cvpr/KulisSD11}), where both labels for the source and the target domain are available; \textit{semi-supervised} approaches (e.g. \cite{conf/cvpr/DonahueHRSD13,conf/cvpr/YaoPNLM15}), where labels for the source and a small set of labels for the target domain are provided; and lastly \textit{unsupervised} approaches (e.g. \cite{blitzer-etal-2006-domain-fixed,ganin2015unsupervised,conf/aaai/SunFS16,conf/icml/LiptonWS18}), where only labels for the source domain are given. Since the focus of this paper is the latter, we restrict our discussion to unsupervised approaches. A more complete recent review of unsupervised domain adaptation approaches is given in \cite{kouw2019review}.

A popular approach to unsupervised domain adaptation is to induce representations which are invariant to the shift in distribution between source and target data. For deep networks, this can be accomplished via domain adversarial training using a simple gradient reversal trick~(\cite{ganin2015unsupervised}). This has been shown to work in the multi-source domain adaptation setting too~(\cite{li2018s}). Other popular representation learning methods include minimizing the covariance between source and target features~(\cite{conf/aaai/SunFS16}) and using maximum-mean discrepancy between the marginal distribution of source and target features as an adversarial objective~(\cite{guo2018multi}).

Mixture of experts has also been shown to be effective for multi-source domain adaptation. \cite{kim2017domain} use 
attention to combine the predictions of domain experts. \cite{guo2018multi} propose learning a mixture of experts using a point to set metric, which combines the posteriors of models trained on individual domains. Our work attempts to build on this to study how multi-source domain adaptation can be improved with LPX models.

\subsection{Transformer Based Domain Adaptation}
There are a handful of studies which investigate how LPX models can be improved in the presence of domain shift. These methods tend to focus on the data and training objectives for single-source single-target unsupervised domain adaptation. The work of \cite{ma2019domain} shows that curriculum learning based on the similarity of target data to source data improves the performance of BERT on out of domain natural language inference. Additionally, \cite{han2019unsupervised} demonstrate that domain adaptive fine-tuning with the masked language modeling objective of BERT leads to improved performance on domain adaptation for sequence labelling. \cite{rietzler2019adapt} offer similar evidence for task adaptive fine-tuning on aspect based sentiment analysis. \cite{gururangan2020don} take this further, showing that significant gains in performance are yielded when progressively fine-tuning on in domain data, followed by task data, using the masked language modeling objective of RobERTa. Finally, \cite{lin2020does} explore whether domain adversarial training with BERT would improve performance for clinical negation detection, finding that the best performing method is a plain BERT model, giving some evidence that perhaps well-studied domain adaptation methods may not be applicable to LPX models.

What has not been studied, to the best of our knowledge, is the impact of domain adversarial training via gradient reversal on LPX models on natural language processing tasks, as well as if mixture of experts techniques can be beneficial. As these methods have historically benefited deep 
models for domain adaptation, we explore their effect when applied to LPX models in this work. 

\section{Methods}
\begin{figure}[t]
  
  \centering
    \includegraphics[width=0.7\textwidth]{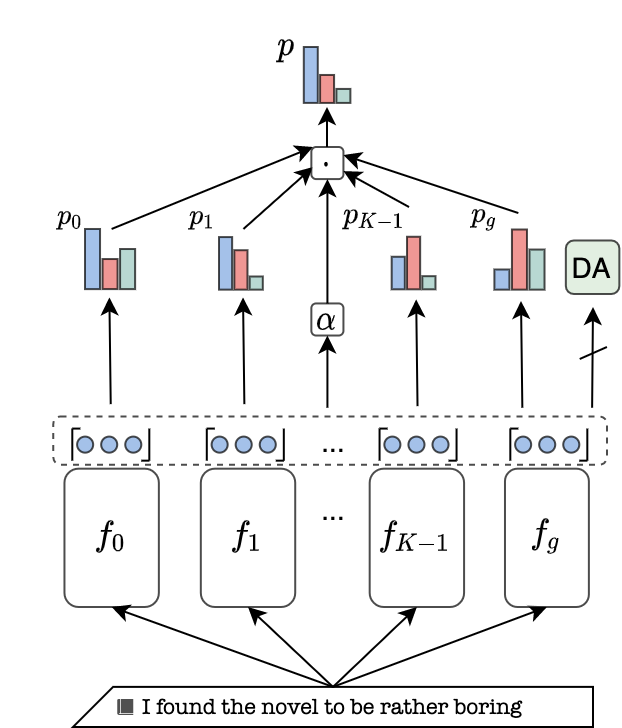}
    \caption{The overall approach tested in this work. A sample is input to a set of expert and one shared LPX model as described in \S\ref{sec:modeling}. The output probabilities of these models are then combined using an attention parameter alpha (\S\ref{sec:avg}, \S\ref{sec:fta}, \S\ref{sec:dc}, \S\ref{sec:attention}). In addition, a global model $f_g$ learns domain invariant representations via a classifier \texttt{DA} with gradient reversal (indicated by the slash, see \S\ref{sec:da_method}).}
    \label{fig:check-worthy-examples}
\end{figure}
This work is motivated by previous research on domain adversarial training and mixture of domain experts for domain adaptation. In this, the data consists of 
$K$ source domains $\mathcal{S}$ and a target domain $\mathcal{T}$. The source domains consist of labelled datasets $D_{s}, s \in \{1,...,K\}$ and the target domain consists only of unlabelled data $U_{t}$. The goal is to learn a classifier $f$, which generalizes well to $\mathcal{T}$ using only the labelled data from $\mathcal{S}$ and optionally unlabelled data from $\mathcal{T}$. We consider a base network $f_{z}, z \in \mathcal{S} \cup \{g\}$ corresponding to either a domain specific network or a global shared network. These $f_{z}$ networks are initialized using LPX models, in particular DistilBert~(\cite{sanh2019distilbert}). 

\subsection{Mixture of Experts Techniques}\label{sec:modeling}
We study four different mixture of expert techniques: simple averaging, fine-tuned averaging, attention with a domain classifier, and a novel sample-wise attention mechanism based on transformer attention~(\cite{vaswani2017attention}). Prior work reports that utilizing mixtures of domain experts and shared classifiers leads to improved performance when having access to multiple source domains~(\cite{guo2018multi,li2018s}). Given this, we investigate if mixture of experts can have any benefit when using LPX models. 

Formally, for a setting with $K$ domains, we have set of $K$ different LPX models $f_{k}, k \in \{0...K-1\}$ corresponding to each domain. There is also an additional LPX model $f_{g}$ corresponding to a global shared model. The output predictions of these models are $p_{k}, k \in \{0...K-1\}$ and $p_{g}$, respectively. Since the problems we are concerned with are binary classification, these are single values in the range $(0,1)$. The final output probability is calculated as a weighted combination of a set of domain expert probabilities $\bar{\mathcal{K}} \subseteq \mathcal{S}$ and the probability from the global shared model. Four methods are used for calculating the weighting. 

\subsubsection{Averaging}
\label{sec:avg}
The first method is a simple averaging of the predictions of domain specific and shared classifiers. The final output of the model is
\begin{equation}
    p_A(x,\bar{\mathcal{K}}) = \frac{1}{|\bar{\mathcal{K}}|+1}\sum_{k \in \bar{\mathcal{K}}}p_{k}(x) + p_{g}(x)
\end{equation}

\subsubsection{Fine Tuned Averaging}
\label{sec:fta}
As an extension to simple averaging, we fine tune the weight given to each of the domain experts and global shared model. This is performed via randomized grid search evaluated on validation data, after the models have been trained. A random integer between zero and ten is generated for each of the models, which is then normalized to a set of probabilities $\alpha_{F}$. The final output probability is then given as follows.

\begin{equation}
    p_F(x) = \sum_{k \in \bar{\mathcal{K}}}p_k(x) * \alpha_{F}^{(k)}(x) + p_g(x) * \alpha_{F}^{(g)}(x)
\end{equation}


\subsubsection{Domain Classifier}
\label{sec:dc}
It was recently shown that curriculum learning using a domain classifier can lead to improved performance for single-source domain adaptation~(\cite{ma2019domain}) when using LPX models. Inspired by this, we experiment with using a domain classifier as a way to attend to the predictions of domain expert models. First, a domain classifier $f_C$ is trained to predict the domain of an input sample $x$ given $\mathbf{r}_{g} \in \mathbb{R}^{d}$, the representation of the \texttt{[CLS]} token at the output of a LPX model. From the classifier, a vector $\alpha_{C}$ is produced with the probabilities that a sample belongs to each source domain. 
\begin{equation}
    \alpha_{C} = f_{C}(x) = \text{softmax}(\mathbf{W}_C\mathbf{r}_{g} + b_{C})
\end{equation}
where $\mathbf{W}_{C} \in \mathbb{R}^{d \times K}$ and $b_{C} \in \mathbb{R}^{K}$. The domain classifier is trained before the end-task network and is held static throughout training on the end-task. For this, a set of domain experts $f_{k}$ are trained and their predictions combined through a weighted sum of the attention vector $\alpha_{C}$.
\begin{equation}
    p_C(x) = \sum_{k \in S}p_k(x) * \alpha_C^{(k)}(x)
\end{equation}
where the superscript $(k)$ indexes into the $\alpha_C$ vector. Note that in this case we only use domain experts and not a global shared model. In addition, the probability is always calculated with respect to each source domain.

\subsubsection{Attention Model}
\label{sec:attention}
Finally, a novel parameterized attention model is learned which attends to different domains based on the input sample. The attention method is based on the scaled dot product attention applied in transformer models~(\cite{vaswani2017attention}), where a global shared model acts as a query network attending to each of the expert and shared models. As such, a shared model $f_{g}$ produces a vector $\mathbf{r}_{g} \in \mathbb{R}^{d}$, and each domain expert produces a vector $\mathbf{r}_{k} \in \mathbb{R}^{d}$. First, for an input sample $x$, a probability for the end task is obtained from the classifier of each model yielding probabilities $p_{g}$ and $p_{k}, k \in {0...K-1}$.
An attention vector $\alpha_{X}$ is then obtained via the following transformations.
\begin{equation}
    \mathbf{q} = \mathbf{g}\mathbf{Q}^{T}
\end{equation}
\begin{equation}
    \mathbf{k} = \begin{bmatrix}
           \mathbf{r}_{1} \\
           \vdots \\
           \mathbf{r}_{K} \\
           \mathbf{r}_{g}
         \end{bmatrix} \mathbf{K}^{T}
\end{equation}
\begin{equation}
    \alpha_{X} = \text{softmax}(\mathbf{q}\mathbf{k}^{T})
\end{equation}
where $\mathbf{Q} \in \mathbb{R}^{d \times d}$ and $\mathbf{K} \in \mathbb{R}^{d \times d}$. The attention vector $\alpha_{X}$ then attends to the individual predictions of each domain expert and the global shared model. 
\begin{equation}
    p_X(x,\bar{\mathcal{K}}) = \sum_{k \in \bar{\mathcal{K}}}p_k(x) * \alpha_{X}^{(k)}(x) + p_g(x) * \alpha_{X}^{(g)}(x)
\end{equation}

To ensure that each model is trained as a domain specific expert, a similar training procedure to that of \cite{guo2018multi} is utilized, described in \S\ref{sec:training}.

\subsection{Domain Adversarial Training}
\label{sec:da_method}
The method of domain adversarial adaptation we investigate here is the well-studied technique described in~\cite{ganin2015unsupervised}. It has been shown to benefit both convolutional nets and recurrent nets on NLP problems~(\cite{li2018s,gui2017part}), so is a prime candidate to study in the context of LPX models. Additionally, some preliminary evidence indicates that adversarial training might improve LPX generalizability for single-source domain adaptation~(\cite{ma2019domain}). 

To learn domain invariant representations, we train a model such that the learned representations maximally confuse a domain classifier $f_d$. This is accomplished through a min-max objective between the domain classifier parameters $\theta_{D}$ and the parameters $\theta_{G}$ of an encoder $f_g$. The objective can then be described as follows.
\begin{equation}
    \mathcal{L}_D = \max_{\theta_{D}}\min_{\theta_{G}}-d\log  f_{d}(f_{g}(x))
\end{equation}
where $d$ is the domain of input sample $x$. The effect of this is to improve the ability of the classifier to determine the domain of an instance, while encouraging the model to generate maximally confusing representations via minimizing the negative loss. In practice, this is accomplished by training the model using standard cross entropy loss, but reversing the gradients of the loss with respect to the model parameters $\theta_{G}$.

\subsection{Training}
\label{sec:training}
Our training procedure follows a multi-task learning setup in which the data from a single batch comes from a single domain. Domains are thus shuffled on each round of training and the model is optimized for a particular domain on each batch. 

For the attention based (\S\ref{sec:attention}) and averaging (\S\ref{sec:avg}) models we adopt a similar training algorithm to~\cite{guo2018multi}. For each batch of training, a meta-target $t$ is selected from among the source domains, with the rest of the domains treated as meta-sources $\mathcal{S}' \in \mathcal{S} \setminus \{t\}$. Two losses are then calculated. The first is with respect to all of the meta-sources, where the attention vector is calculated for only those domains. For target labels $y_{i}$ and a batch of size $N$ with samples from a single domain, this is given as follows.
\begin{equation}
    \mathcal{L}_{s} = -\frac{1}{N}\sum_{i}y_{i}\log p_{X}(x, \mathcal{S}')
\end{equation}
The same procedure is followed for the averaging model $p_{A}$. The purpose is to encourage the model to learn attention vectors for out of domain data, thus why the meta-target is excluded from the calculation. 

The second loss is with respect to the meta-target, where the cross-entropy loss is calculated directly for the domain expert network of the meta-target.
\begin{equation}
    \mathcal{L}_{t} = -\frac{1}{N}\sum_{i}y_{i}\log p_t(x)
\end{equation}
This allows each model to become a domain expert through strong supervision. The final loss of the network is a combination of the three losses described previously, with $\lambda$ and $\gamma$  hyperparameters controlling the weight of each loss.
\begin{equation}
    \mathcal{L} = \lambda \mathcal{L}_s + (1 - \lambda) \mathcal{L}_t + \gamma \mathcal{L}_D
\end{equation}

For the domain classifier (\S\ref{sec:dc}) and fine-tuned averaging (\S\ref{sec:fta}), the individual LPX models are optimized directly with no auxiliary mixture of experts objective. In addition, we experiment with training the simple averaging model directly.

\section{Experiments and Results}

\begin{table}[t!]
    \centering
    \fontsize{10}{10}\selectfont
    \begin{tabular}{l c c c c c | c c c c c c c }
    \toprule 
     Method &\multicolumn{5}{c}{Sentiment Analysis (Accuracy)}&\multicolumn{6}{c}{Rumour Detection (F1)}\\
    \cmidrule(lr){2-6}
    \cmidrule(lr){7-12}
     & D & E & K & B & macroA & CH & F & GW & OS & S & $\mu$F1\\
    \midrule 
    \cite{li2018s} &77.9 &80.9 &80.9 &77.1 & 79.2 & - & - & - & - & - & -\\
      \cite{guo2018multi} &87.7 &89.5 & 90.5& 87.9 &88.9 & - & - & - & - & - & -\\
      \cite{zubiaga2017exploiting}&- &- &- &- & - & 63.6& \textbf{46.5}& 70.4& 69.0& 61.2& 60.7\\
    \midrule
        Basic & 89.1& 89.8& 90.1& 89.3& 89.5& 66.1& 44.7& 71.9& 61.0& 63.3& 62.3\\
        \midrule
        Adv-6 & 88.3& 89.7& 90.0& 89.0& 89.3 & 65.8& 42.0& 66.6& 61.7& 63.2& 61.4\\
        Adv-3 & 89.0& 89.9& 90.3& 89.0& 89.6& 65.7& 43.2& 72.3& 60.4& 62.1& 61.7\\
        \midrule
        Independent-Avg & 88.9& \textbf{90.6}& 90.4& \textbf{90.0}& \textbf{90.0}& 66.1& 45.6& 71.7& 59.4& 63.5& 62.2\\
        Independent-Ft & 88.9& 90.3& \textbf{90.8}& \textbf{90.0}& \textbf{90.0}& 65.9& 45.7& 72.2& 59.3& 62.4& 61.9\\
        MoE-Avg & \textbf{89.3}& 89.9& 90.5& 89.9& 89.9& \textbf{67.9}& 45.4& \textbf{74.5}& 62.6& \textbf{64.7}& \textbf{64.1}\\
        MoE-Att & 88.6& 90.0& 90.4& 89.6& 89.6& 65.9& 42.3& 72.5& 61.2& 63.3& 62.2\\
        MoE-Att-Adv-6 & 87.8& 89.0 & 90.5& 88.3& 88.9& 66.0& 40.7& 69.0& 63.8& 63.7& 61.8\\
        MoE-Att-Adv-3 & 88.6& 89.1& 90.4& 88.9& 89.2& 65.6& 42.7& 73.4& 60.9& 61.0& 61.8\\
        MoE-DC & 87.8& 89.2& 90.2& 87.9& 88.8& 66.5& 40.6& 70.5& \textbf{70.8}& 62.8& 63.8\\
    \bottomrule 

    \end{tabular}
    \caption{Experiments for sentiment analysis in (D)VD, (E)lectronics, (K)itchen and housewares, and (B)ooks domains and rumour detection for different events ((C)harlie(H)ebdo, (F)erguson, (G)erman(W)ings, (O)ttawa(S)hooting, and (S)ydneySiege) using leave-one-out cross validation. Results are averaged across 5 random seeds. The results for sentiments analysis are in terms of accuracy and the results for rumour detection are in terms of F1.}
    \label{tab:sentiment_results}
\end{table}
We focus our experiments on text classification problems with data from multiple domains. To this end, we experiment with sentiment analysis from Amazon product reviews and rumour detection from tweets. For both tasks, we perform cross-validation on each domain, holding out a single domain for testing and training on the remaining domains, allowing a comparison of each method on how well they perform under domain shift. The code to reproduce all of the experiments in this paper can be found here.\footnote{\url{https://github.com/copenlu/xformer-multi-source-domain-adaptation}}

\paragraph{Sentiment Analysis Data} The data used for sentiment analysis come from the legacy Amazon Product Review dataset~(\cite{blitzer2007biographies}). This dataset consists of 8,000 total tweets from four product categories: books, DVDs, electronics, and kitchen and housewares. Each domain contains 1,000 positive and 1,000 negative reviews. In addition, each domain has associated unlabelled data. Following previous work we focus on the transductive setting~(\cite{guo2018multi,ziser2017neural}) where we use the same 2,000 out of domain tweets as unlabelled data for training the domain adversarial models.  This data has been well studied in the context of domain adaptation, making for easy comparison with previous work.

\paragraph{Rumour Detection Data} The data used for rumour detection come from the PHEME dataset of rumourous tweets~(\cite{zubiaga2016analysing}). There are a total of 5,802 annotated tweets from 5 different events labelled as rumourous or non-rumourous (1,972 rumours, 3,830 non-rumours). Methods which have been shown to work well on this data include context-aware classifiers (\cite{zubiaga2017exploiting}) and positive-unlabelled learning (\cite{wright-augenstein-2020-claim}). Again, we use this data in the transductive setting when testing domain adversarial training. 

\subsection{Baselines}

\paragraph{What's in a Domain?}
We use the model from~\cite{li2018s} as a baseline for sentiment analysis. This model consists of a set of domain experts and one general CNN, and is trained with a domain adversarial auxiliary objective.

\paragraph{Mixture of Experts}
Additionally, we present the results from~\cite{guo2018multi} representing the most recent state of the art on the Amazon reviews dataset. Their method consists of domain expert classifiers trained on top of a shared encoder, with predictions being combined via a novel metric which considers the distance between the mean representations of target data and source data. 

\paragraph{\cite{zubiaga2017exploiting}} Though not a domain adaptation technique, we include the results from \cite{zubiaga2017exploiting} on rumour detection to show the current state of the art performance on this task. The model is a CRF, which utilizes a combination of content and social features acting on a timeline of tweets.
\begin{figure}[t]
  
  \centering
    \includegraphics[width=1.0\textwidth]{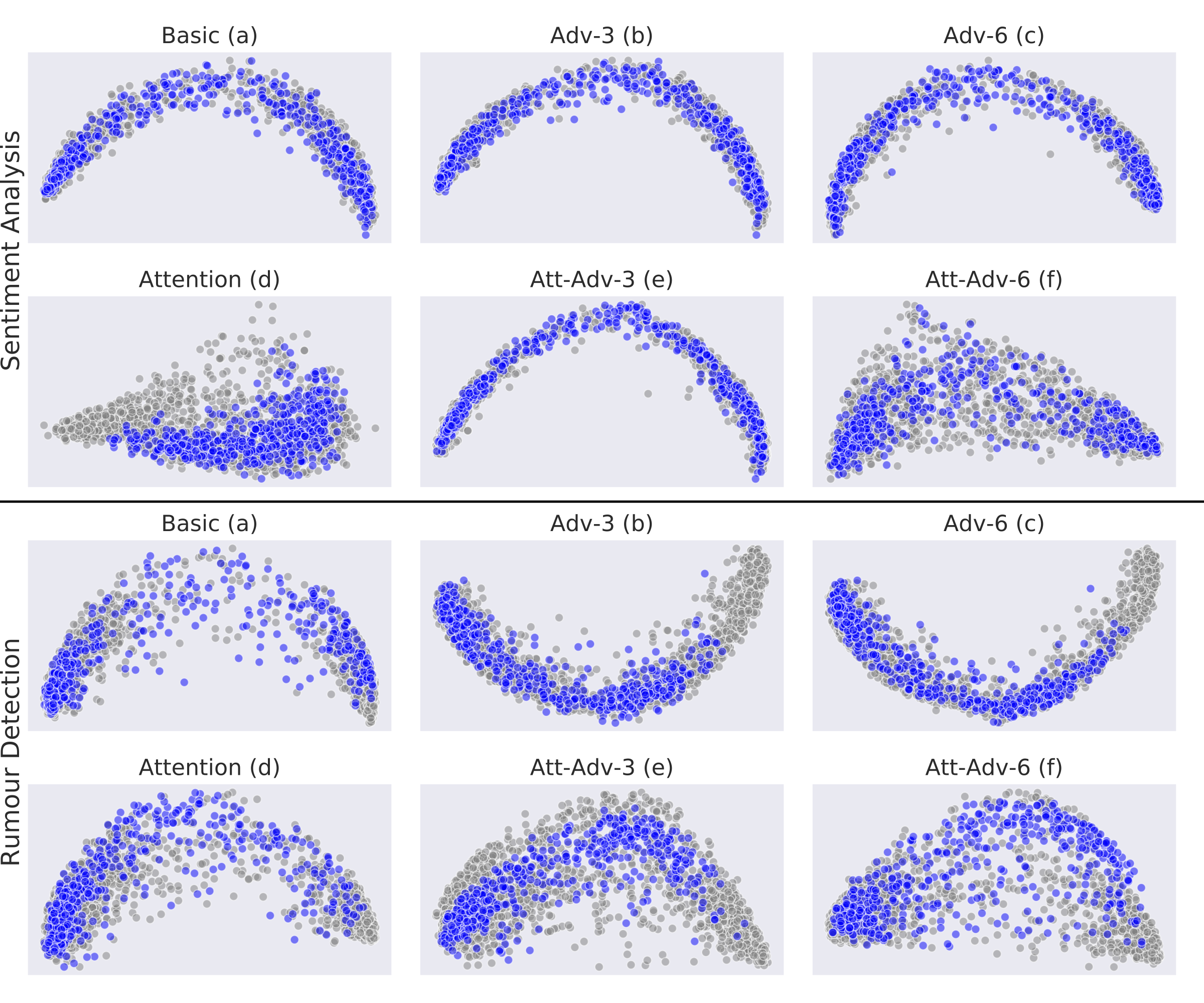}
    \caption{Final layer DistilBert embeddings for 500 randomly selected examples from each split for each tested model for sentiment data (top two rows) and rumour detection (bottom two rows). The blue points are out of domain data (in this case from Kitchen and Housewares for sentiment analysis and Sydney Siege for rumour detection) and the gray points are in domain data. }
    \label{fig:all_reps}
\end{figure}

\subsection{Model Variants}
A variety of models are tested in this work. Therefore, each model is referred to by the following.

\paragraph{Basic} Basic DistilBert with a single classification layer at the output. 

\paragraph{Adv-$X$} DistilBert with domain adversarial supervision applied at the $X$'th layer (\S\ref{sec:da_method}).

\paragraph{Independent-Avg} DistilBert mixture of experts averaged but trained individually (not with the algorithm described in \S\ref{sec:training}).

\paragraph{Independent-FT} DistilBert mixture of experts averaged with mixing attention fine tuned after training (\S\ref{sec:fta}), trained individually.

\paragraph{MoE-Avg} DistilBert mixture of experts using averaging (\S\ref{sec:avg}).

\paragraph{MoE-Att} DistilBert mixture of experts using our novel attention based technique (\S\ref{sec:attention}).

\paragraph{MoE-Att-Adv-$X$} DistilBert mixture of experts using attention and domain adversarial supervision applied at the $X$'th layer.

\paragraph{MoE-DC} DistilBert mixture of experts using a domain classifier for attention (\S\ref{sec:dc}).

\subsection{Results}

  

Our results are given in \autoref{tab:sentiment_results}. Similar to the findings of \cite{lin2020does} on clinical negation, we see little overall difference in performance from both the individual model and the mixture of experts model when using domain adversarial training on sentiment analysis. For the base model, there is a slight improvement when domain adversarial supervision is applied at a lower layer of the model, but a drop when applied at a higher level. Additionally, mixture of experts provides some benefit, especially using the simpler methods such as averaging.


For rumour detection, again we see little performance change from using domain adversarial training, with a slight drop when supervision is applied at either layer. The mixture of experts methods overall perform better than single model methods, suggesting that mixing domain experts is still effective when using large pretrained transformer models. In this case, the best mixture of experts methods are simple averaging and static grid search for mixing weights, indicating the difficulty in learning an effective way to mix the predictions of domain experts. We elaborate on our findings further in \S\ref{sec:discussion}. Additional experiments on domain adversarial training using Bert can be found in \autoref{tab:bert_appendix_results} in \S\ref{sec:bert_appendix}, where we similarly find that domain adversarial training leads to a drop in performance on both datasets.

\section{Discussion}
\label{sec:discussion}
We now discuss our initial research questions in light of the results we obtained, and provide explanations for the observed behavior.
  
\subsection{What is the Effect of Domain Adversarial Training?}
We present PCA plots of the representations learned by different models in \autoref{fig:all_reps}. These are the final layer representations of 500 randomly sampled points for each split of the data. In the ideal case, the representations for out of domain samples would be indistinguishable from the representations for in domain data.

In the case of basic DistilBert, we see a slight change in the learned representations of the domain adversarial models versus the basic model (\autoref{fig:all_reps} top half, a-c) for sentiment analysis. When the attention based mixture of experts model is used, the representations of out of domain data cluster in one region of the representation space (d). With the application of adversarial supervision, the model learns representations which are more domain agnostic. Supervision applied at layer 6 of DistilBert (plot f) yields a representation space similar to the version without domain adversarial supervision. Interestingly, the representation space of the model with supervision at layer 3 (plot e) yields representations similar to the basic classifier. This gives some potential explanation as to the similar performance of this model to the basic classifier on this split (kitchen and housewares). Overall, domain adversarial supervision has some effect on performance, leading to gains in both the basic classifier and the mixture of experts model for this split. Additionally, there are minor improvements overall for the basic case, and a minor drop in performance with the mixture of experts model.



The effect of domain adversarial training is more pronounced on the rumour detection data for the basic model (\autoref{fig:all_reps} bottom half, a), where the representations exhibit somewhat less variance when domain adversarial supervision is applied. Surprisingly, this leads to a slight drop in performance for the split of the data depicted here (Sydney Siege). For the attention based model, the variant without domain adversarial supervision (d) already learns a somewhat domain agnostic representation. The model with domain adversarial supervision at layer 6 (f) furthers this, and the classifier learned from these representations perform better on this split of the data. Ultimately, the best performing models for rumour detection do not use domain supervision, and the effect on performance on the individual splits are mixed, suggesting that domain adversarial supervision can potentially help, but not in all cases.


\subsection{Is Mixture of Experts Useful with LPX Models?}
\label{sec:moe_discussion}
\begin{figure}
  
  \centering
    \includegraphics[width=0.7\textwidth]{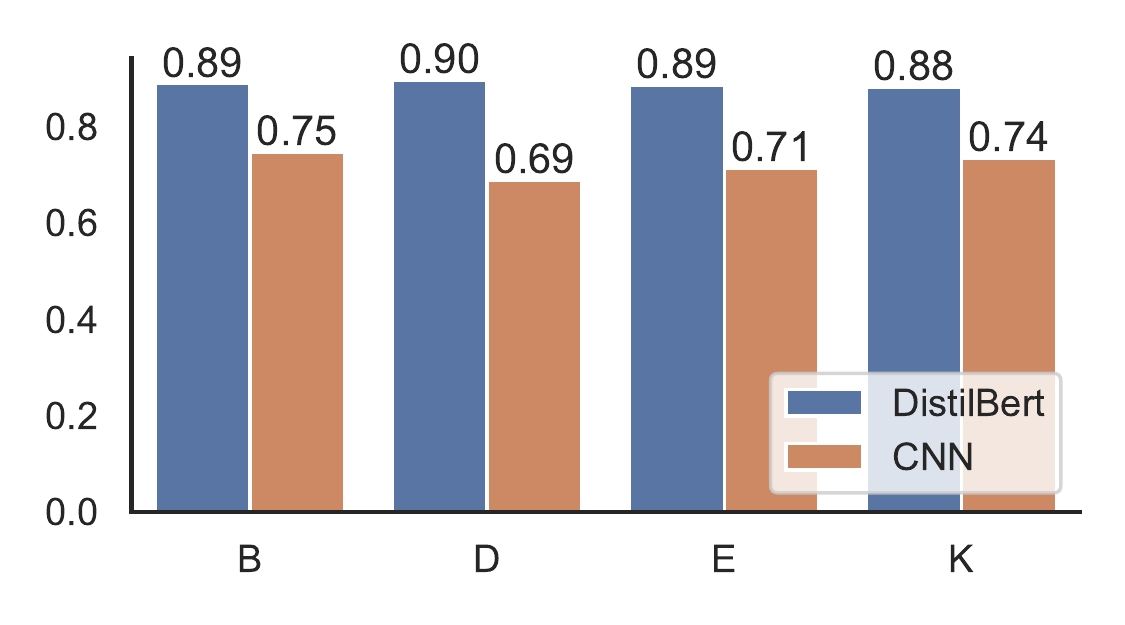}
    \caption{Comparison of agreement (Krippendorff's alpha) between domain expert models when the models are either DistilBert or a CNN. Predictions are made on unseen test data by each domain expert, and agreement is measured between their predictions ((B)ooks, (D)VD, (E)lectronics, and (K)itchen). The overall agreement between the DistilBert experts is greater than the CNNs, suggesting that the learned classifiers are much more homogenous.}
    \label{fig:krippendorff}
\end{figure}
We performed experiments with several variants of mixture of experts, finding that overall, it can help, but determining the optimal way to mix LPX domain experts remains challenging. Simple averaging of domain experts (\S\ref{sec:avg}) gives better performance on both sentiment analysis and rumour detection over the single model baseline. Learned attention (\S\ref{sec:attention}) has a net positive effect on performance for sentiment analysis and a negative effect for rumour detection compared to the single model baseline. Additionally, simple averaging of domain experts consistently outperforms a learned sample by sample attention. This highlights the difficulty in utilizing large pretrained transformer models to learn to attend to the predictions of domain experts.

\paragraph{Comparing agreement} To provide some potential explanation for why it is difficult to learn to attend to domain experts, we compare the agreement on the predictions of domain experts of one of our models based on DistilBert, versus a model based on CNNs (\autoref{fig:krippendorff}). CNN models are chosen in order to compare the agreement using our approach with an approach which has been shown to work well with mixture of experts on this data~(\cite{guo2018multi}). Each CNN consists of an embedding layer initialized with 300 dimensional FastText embeddings~(\cite{bojanowski2017enriching}), a series of 100 dimensional convolutional layers with widths 2, 4, and 5, and a classifier. The end performance is on par with previous work using CNNs~(\cite{li2018s}) (78.8 macro averaged accuracy, validation accuracies of the individual models are between 80.0 and 87.0). Agreement is measured using Krippendorff's alpha~(\cite{krippendorff2011computing}) between the predictions of domain experts on test data. 

We observe that the agreement between DistilBert domain experts on test data is significantly higher than that of CNN domain experts, indicating that the learned classifiers of each expert are much more similar in the case of DistilBert. Therefore, it will potentially be more difficult for a mixing function on top of DistilBert domain experts to gain much beyond simple averaging, while with CNN domain experts, there is more to be gained from mixing their predictions. This effect may arise because each DistilBert model is highly pre-trained already, hence there is little change in the final representations, and therefore similar classifiers are learned between each domain expert.
\section{Conclusion}
In this work, we investigated the problem of multi-source domain adaptation with large pretrained transformer models. Both domain adversarial training and mixture of experts techniques were explored. While domain adversarial training could effectively induce more domain agnostic representations, it had a mixed effect on model performance. Additionally, we demonstrated that while techniques for mixing domain experts can lead to improved performance for both sentiment analysis and rumour detection, determining a beneficial mixing of such experts is challenging. The best method we tested was a simple averaging of the domain experts, and we provided some evidence as to why this effect was observed. We find that LPX models may be better suited for data-driven techniques such as that of~\cite{gururangan2020don}, which focus on inducing a better prior into the model through pretraining, as opposed to techniques which focus on learning a better posterior with architectural enhancements. We hope that this work can help inform researchers of considerations to make when using LPX models in the presence of domain shift.

\section*{Acknowledgements}
$\begin{array}{l}\includegraphics[width=1cm]{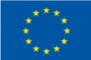} \end{array}$ This project has received funding from the European Union's Horizon 2020 research and innovation programme under the Marie Sk\l{}odowska-Curie grant agreement No 801199. 




\clearpage

\section{Appendix}

\subsection{BERT Domain Adversarial Training Results}
\label{sec:bert_appendix}
Additional results on domain adversarial training with Bert can be found in \autoref{tab:bert_appendix_results}.

\begin{table}[t!]
    \centering
    \fontsize{10}{10}\selectfont
    \begin{tabular}{l c c c c c | c c c c c c c }
    \toprule 
     Method &\multicolumn{5}{c}{Sentiment Analysis (Accuracy)}&\multicolumn{6}{c}{Rumour Detection (F1)}\\
    \cmidrule(lr){2-6}
    \cmidrule(lr){7-12}
     & D & E & K & B & macroA & CH & F & GW & OS & S & $\mu$F1\\
    \midrule
        Bert & 90.3& 91.6& 91.7& 90.4& 91.0& 66.4& 46.2& 68.3& 67.3& 62.3& 63.3\\
        \midrule
        Bert-Adv-12 & 89.8& 91.4& 91.2& 90.1& 90.6 & 66.6& 47.8& 62.5& 65.3& 62.8& 62.5\\
        Bert-Adv-4 & 89.9& 91.1& 91.7& 90.4& 90.8& 65.6& 43.6& 71.0& 68.1& 60.8& 62.8\\
    \bottomrule 

    \end{tabular}
    \caption{Experiments for sentiment analysis in (D)VD, (E)lectronics, (K)itchen and housewares, and (B)ooks domains and rumour detection for different events ((C)harlie(H)ebdo, (F)erguson, (G)erman(W)ings, (O)ttawa(S)hooting, and (S)ydneySiege) using leave-one-out cross validation for BERT. Results are averaged across 3 random seeds. The results for sentiments analysis are in terms of accuracy and the results for rumour detection are in terms of F1.}
    \label{tab:bert_appendix_results}
\end{table}

\subsection{Reproducibility}

\subsubsection{Computing Infrastructure}
All experiments were run on a shared cluster. Requested jobs consisted of 16GB of RAM and 4 Intel Xeon Silver 4110 CPUs. We used a single NVIDIA Titan X GPU with 12GB of RAM.

\subsubsection{Average Runtimes}
The average runtime performance of each model is given in \autoref{tab:runtimes}. Note that different runs may have been placed on different nodes within a shared cluster, thus why large time differences occurred. 
\begin{table}
    \centering
    \fontsize{10}{10}\selectfont
    \begin{tabular}{l c c}
    \toprule 
     Method &Sentiment Analysis &Rumour Detection\\
    \midrule
        Basic & 0h44m37s& 0h23m52s\\
        \midrule
        Adv-6 & 0h54m53s& 0h59m31s\\
        Adv-3 & 0h53m43s& 0h57m29s\\
        \midrule
        Independent-Avg & 1h39m13s& 1h19m27\\
        Independent-Ft & 1h58m55s& 1h43m13\\
        MoE-Avg & 2h48m23s& 4h03m46s\\
        MoE-Att & 2h49m44s& 4h07m3s\\
        MoE-Att-Adv-6 & 4h51m38s& 4h58m33s\\
        MoE-Att-Adv-3 & 4h50m13s& 4h54m56s\\
        MoE-DC & 3h23m46s& 4h09m51s\\
    \bottomrule 

    \end{tabular}
    \caption{Average runtimes for each model on each dataset (runtimes are taken for the entire run of an experiment).}
    \label{tab:runtimes}
\end{table}

\subsubsection{Number of Parameters per Model}
The number of parameters in each model is given in \autoref{tab:num_params}.
\begin{table}
    \centering
    \fontsize{10}{10}\selectfont
    \begin{tabular}{l c c}
    \toprule 
     Method &Sentiment Analysis &Rumour Detection\\
    \midrule
        Basic & 66,955,010& 66,955,010\\
        \midrule
        Adv-6 & 66,958,082& 66,958,850\\
        Adv-3 & 66,958,082& 66,958,850\\
        \midrule
        Independent-Avg & 267,820,040& 334,775,050\\
        Independent-Ft & 267,820,040& 334,775,050\\
        MoE-Avg & 267,820,040& 334,775,050\\
        MoE-Att & 268,999,688& 335,954,698\\
        MoE-Att-Adv-6 & 269,002,760& 335,958,538\\
        MoE-Att-Adv-3 & 269,002,760& 335,958,538\\
        MoE-DC & 267,821,576& 334,777,354\\
    \bottomrule 

    \end{tabular}
    \caption{Number of parameters in each model}
    \label{tab:num_params}
\end{table}

\subsubsection{Validation Performance}
The validation performance of each tested model is given in \autoref{tab:val_performance}.
\begin{table}
    \centering
    \fontsize{10}{10}\selectfont
    \begin{tabular}{l c c}
    \toprule 
     Method &Sentiment Analysis (Acc) &Rumour Detection (F1)\\
    \midrule
        Basic & 91.7& 82.4\\
        \midrule
        Adv-6 & 91.5& 83.3\\
        Adv-3 & 91.2& 83.4\\
        \midrule
        Independent-Avg & 92.7& 82.8\\
        Independent-Ft & 92.6& 82.5\\
        MoE-Avg & 92.2& 83.5\\
        MoE-Att & 92.0& 83.3\\
        MoE-Att-Adv-6 & 91.2& 83.3\\
        MoE-Att-Adv-3 & 91.4& 82.8\\
        MoE-DC & 89.8& 84.6\\
    \bottomrule 

    \end{tabular}
    \caption{Average validation performance for each of the models on both datasets.}
    \label{tab:val_performance}
\end{table}

\subsubsection{Evaluation Metrics}
The primary evaluation metrics used were accuracy and F1 score. For accuracy, we used our implementation provided with the code. The basic implementation is as follows.
\begin{equation*}
    \text{accuracy} = \frac{tp + tn}{tp+fp+tn+fn}
\end{equation*}
We used the sklearn implementation of \texttt{precision\_recall\_fscore\_support} for F1 score, which can be found here: \url{https://scikit-learn.org/stable/modules/generated/sklearn.metrics.precision_recall_fscore_support.html}. Briefly:
\begin{equation*}
   p = \frac{tp}{tp + fp} 
\end{equation*}
\begin{equation*}
   r = \frac{tp}{tp + fn} 
\end{equation*}
\begin{equation*}
   F1 = \frac{2*p*r}{p+r} 
\end{equation*}
where $tp$ are true positives, $fp$ are false positives, and $fn$ are false negatives.

\subsubsection{Hyperparameters}
We performed and initial hyperparameter search to obtain good hyperparameters that we used across models. The bounds for each hyperparameter was as follows:
\begin{itemize}[noitemsep]
    \item Learning rate: [0.00003, 0.00004, 0.00002, 0.00001, 0.00005, 0.0001, 0.001].
    \item Weight decay: [0.0, 0.1, 0.01, 0.005, 0.001, 0.0005, 0.0001].
    \item Epochs: [2, 3, 4, 5, 7, 10].
    \item Warmup steps: [0, 100, 200, 500, 1000, 5000, 10000].
    \item Gradient accumulation: [1,2]
\end{itemize}
We kept the batch size at 8 due to GPU memory constraints and used gradient accumulation instead. We performed a randomized hyperparameter search for 70 trials. Best hyperparameters are chosen based on validation set performance (accuracy for sentiment data, F1 for rumour detection data). The final hyperparameters selected are as follows:
\begin{itemize}[noitemsep]
    \item Learning rate: 3e-5.
    \item Weight decay: 0.01.
    \item Epochs: 5.
    \item Warmup steps: 200.
    \item Batch Size: 8
    \item Gradient accumulation: 1
\end{itemize}
Additionally, we set the objective weighting parameters to $\lambda = 0.5$ for the mixture of experts models and $\gamma = 0.003$ for the adversarial models, in line with previous work~(\cite{guo2018multi,li2018s}).

\subsubsection{Links to data}
\begin{itemize}[noitemsep]
    \item Amazon Product Reviews~(\cite{blitzer2007biographies}):  \url{https://www.cs.jhu.edu/~mdredze/datasets/sentiment/}
    
    \item PHEME~(\cite{zubiaga2016analysing}): \url{https://figshare.com/articles/PHEME_dataset_for_Rumour_Detection_and_Veracity_Classification/6392078}.
    
\end{itemize}

\part{Stance Detection}\label{part:III}
\pdfbookmark[1]{Stance Detection}{stance detection}
\chapter{Stance Detection with Bidirectional Conditional Encoding}\label{ch:conditional}

\boxabstract{Stance detection is the task of classifying the attitude
expressed in a text towards a target such as 
``Climate Change is a Real Concern'' to be ``positive'', ``negative'' or ``neutral''.
Previous work has assumed that either the target is mentioned in the text or that training data for every target is given. This paper considers the more challenging version of this task, where targets are not always mentioned and no training data is available for the test targets. 
We experiment with conditional LSTM encoding, which builds a representation of the tweet that is dependent on the target, and demonstrate that it outperforms the independent encoding of tweet and target. Performance improves even further when the conditional model is augmented with bidirectional encoding. The method is evaluated on the SemEval 2016 Task 6 Twitter Stance Detection corpus and achieves performance second best only to a system trained on semi-automatically labelled tweets for the test target. When such weak supervision is added, our approach achieves state--of-the-art results.}\blfootnote{\fullcite{augenstein-etal-2016-stance}}

\section{Introduction}

The goal of stance detection is to classify the attitude expressed in a text, towards a given target, as ``positive'', ''negative'', or ''neutral''. 
Such information can be useful for a variety of tasks, e.g.\ \cite{Mendoza2010} showed that tweets stating actual facts were affirmed by 90\% of the tweets related to them, while tweets conveying false information were predominantly questioned or denied.
The focus of this paper is on a novel stance detection task, namely tweet stance detection towards previously unseen target entities (mostly entities such as politicians or issues of public interest), as defined in the SemEval Stance Detection for Twitter task~(\cite{mohammad-etal-2016-semeval}). 
This task is rather difficult, firstly due to not having training data for the targets in the test set, and secondly, due to the targets not always being mentioned in the tweet. 
For example, the tweet
``@realDonaldTrump is the only honest voice of the @GOP'' expresses a positive stance towards the target {\em Donald Trump}. However, when stance is predicted with respect to {\em Hillary Clinton} as the target, this tweet expresses a negative stance, since supporting candidates from one party implies negative stance towards candidates from other parties. 

Thus the challenge is twofold. First, we need to learn a model that interprets the tweet stance towards a target that might not be mentioned in the tweet itself. Second, we need to learn such a model without labelled training data for the target with respect to which we are predicting the stance. In the example above, we need to learn a model for {\em Hillary Clinton} by only using training data for other targets. While this renders the task more challenging, it is a more realistic scenario, as it is unlikely that labelled training data for each target of interest will be available.

To address these challenges we develop a neural network architecture based on conditional encoding~(\cite{rocktaschel2016reasoning}). A long-short term memory (LSTM) network~(\cite{hochreiter1997long}) is used to encode the target, followed by a second LSTM that encodes the tweet using the encoding of the target as its initial state. We show that this approach achieves better F1 than standard stance detection baselines, or an independent LSTM encoding of the tweet and the target. The latter achieves an F1 of 0.4169 on the test set.
Results improve further (F1 of 0.4901) with a bidirectional version of our model, which takes into account the context on either side of the word being encoded. In the context of the shared task, this would be the second best result, except for an approach which uses automatically labelled tweets for the test targets (F1 of 0.5628). Lastly, when our bidirectional conditional encoding model is trained on such data, it achieves state-of-the-art performance (F1 of 0.5803).
 
\section{Task Setup}\label{sec:TaskSetup}

The SemEval 2016 Stance Detection for Twitter task~(\cite{mohammad-etal-2016-semeval}) consists of two subtasks, Task A and Task B. In Task A the goal is to detect the stance of tweets towards targets given labelled training data for all test targets ({\em Climate Change is a Real Concern}, {\em Feminist Movement}, {\em Atheism}, {\em Legalization of Abortion} and {\em Hillary Clinton}). 
In Task B, which is the focus of this paper, the goal is to detect stance with respect to an \emph{unseen target} different from the ones considered in Task A, namely {\em Donald Trump}, for which labeled training/development data is not provided.

Systems need to classify the stance of each tweet as ``positive'' (FAVOR), ``negative'' (AGAINST) or ``neutral'' (NONE) towards the target.
The official metric reported is F1 macro-averaged over FAVOR and AGAINST. Although the F1 of NONE is not considered, systems still need to predict it to avoid precision errors for the other two classes.

Although participants were not allowed to manually label data for the test target {\em Donald Trump}, they were allowed to label data automatically. The two best performing systems submitted to Task B, pkudblab~(\cite{StanceSemEval2016pkudblab}) and LitisMind~(\cite{StanceSemEval2016MITRE}), both made use of this. Making use of such techniques renders the task intp \emph{weakly supervised seen target stance detection}, instead of an unseen target task.
Although the goal of this paper is to present stance detection methods for targets for which no training data is available, we show that they can also be used in a weakly supervised framework and outperform the state-of-the-art on the SemEval 2016 Stance Detection for Twitter dataset.

\section{Methods}\label{sec:Methods}

A common stance detection approach is to treat it as a sentence-level classification task similar to sentiment analysis~(\cite{pang2008opinion,socher-EtAl:2013:EMNLP}). However, such an approach cannot capture the stance of a tweet with respect to a particular target, unless training data is available for each of the test targets. In such cases, we could learn that a tweet mentioning {\em Donald Trump} in a positive manner expresses a negative stance towards {\em Hillary Clinton}.
Despite this limitation, we use two such baselines, one implemented with a Support Vector Machine (SVM) classifier and one with an LSTM, in order to assess whether we are successful in incorporating the target in stance prediction.

A naive approach to incorporate the target in stance prediction would be to  generate features concatenating the target with words from the tweet.
In principle, this could allow the classifier to learn that some words in the tweets have target-dependent stance weights, but it still assumes that training data is available for each target.

In order to learn how to combine 
the target with the tweet in a way that generalises to unseen targets, we focus on learning distributed representations and ways to combine them.
The following sections develop progressively the proposed bidirectional conditional LSTM encoding  model, starting from the independent LSTM encoding.

\begin{figure*}
\hspace*{-1.5cm}
\begin{tikzpicture}[scale=1.2]
\foreach \i/\l in {1/1,2/2,3/3} {
  \path[draw, thick] (\i-0.2,0) rectangle (\i+0.2,0.75) {};
  \path[draw, thick, fill=nice-red!10] (\i-0.2,1) rectangle (\i+0.2,1.75) {};
  \path[draw, thick, fill=nice-red!20] (\i-0.2,1.75) rectangle (\i+0.2,2.5) {};  
  \path[draw, thick, fill=nice-red!10] (\i-0.2,2.75) rectangle (\i+0.2,3.5) {};
  \path[draw, thick, fill=nice-red!20] (\i-0.2,3.5) rectangle (\i+0.2,4.25) {};
  \draw[->, >=stealth', thick] (\i,0.75) -- (\i,1);
  \draw[->, >=stealth', thick] (\i-0.2,0.4) to[bend left=30] (\i-0.2,2);
  \draw[->, >=stealth', thick] (\i-0.2,2.3) to[bend left=30] (\i-0.2,3.87);
  \draw[->, >=stealth', thick] (\i+0.2,1.55) to[bend right=30] (\i+0.2,3.17);  
  \node[] at (\i,0.4) {\small$\mathbf{x}_\l$};
  \node[] at (\i,1.4) {\small$\mathbf{c}^\rightarrow_\l$};
  \node[] at (\i,2.15) {\small$\mathbf{c}^\leftarrow_\l$};
  \node[] at (\i,3.15) {\small$\mathbf{h}^\rightarrow_\l$};
  \node[] at (\i,3.9) {\small$\mathbf{h}^\leftarrow_\l$};
}
\foreach \i/\l in {5/4,6/5,7/6,8/7,9/8,10/9} {
  \path[draw, thick] (\i-0.2,0) rectangle (\i+0.2,0.75) {};
  \path[draw, thick, fill=nice-blue!10] (\i-0.2,1) rectangle (\i+0.2,1.75) {};
  \path[draw, thick, fill=nice-blue!20] (\i-0.2,1.75) rectangle (\i+0.2,2.5) {};  
  \path[draw, thick, fill=nice-blue!10] (\i-0.2,2.75) rectangle (\i+0.2,3.5) {};
  \path[draw, thick, fill=nice-blue!20] (\i-0.2,3.5) rectangle (\i+0.2,4.25) {};
  \draw[->, >=stealth', thick] (\i,0.75) -- (\i,1);
  \draw[->, >=stealth', thick] (\i-0.2,0.4) to[bend left=30] (\i-0.2,2);
  \draw[->, >=stealth', thick] (\i-0.2,2.3) to[bend left=30] (\i-0.2,3.87);
  \draw[->, >=stealth', thick] (\i+0.2,1.55) to[bend right=30] (\i+0.2,3.17);  
  \node[] at (\i,0.4) {\small$\mathbf{x}_{\l}$};
  \node[] at (\i,1.4) {\small$\mathbf{c}^\rightarrow_{\l}$};
  \node[] at (\i,2.15) {\small$\mathbf{c}^\leftarrow_{\l}$};
  \node[] at (\i,3.15) {\small$\mathbf{h}^\rightarrow_{\l}$};
  \node[] at (\i,3.9) {\small$\mathbf{h}^\leftarrow_{\l}$};
}
\foreach \i in {1,2,5,6,7,8,9} {
  \draw[->, >=stealth', thick] (\i+0.2,1.4) -- (\i+1-0.2,1.4);
  \draw[->, >=stealth', thick] (\i+1-0.2,2.15) -- (\i+0.2,2.15);
}
\foreach \i/\word in {1/Legalization, 2/of, 3/Abortion, 5/A, 6/foetus, 7/has, 8/rights, 9/too, 10/!} {
  \node[anchor=north, text height=1.5ex, text depth=.25ex, yshift=-2em] at (\i, 0.5) {\word};
}

\draw[ultra thick, nice-red] (0.5,-0.75) -- (3.5,-0.75);
\draw[ultra thick, nice-blue] (4.5,-0.75) -- (10.5,-0.75);
\node[anchor=north] at (2,-0.75) {Target};
\node[anchor=north] at (7.5,-0.75) {Tweet};
\draw[->, >=stealth', line width=3pt, color=nice-red, dashed] (3+0.2,1.4) -- (4+1-0.2,1.4);
\draw[->, >=stealth', line width=3pt, color=nice-red, dashed] (1-0.2,2.15) to[bend left=168] (10+0.2,2.15);

\path[draw, ultra thick, color=nice-blue] (5-0.2,3.5) rectangle (5+0.2,4.25) {};
\path[draw, ultra thick, color=nice-blue] (10-0.2,2.75) rectangle (10+0.2,3.5) {};
\path[draw, ultra thick, color=nice-red] (1-0.2,1.75) rectangle (1+0.2,2.5) {};  
\path[draw, ultra thick, color=nice-red] (3-0.2,1) rectangle (3+0.2,1.75) {};
\end{tikzpicture}
\caption{Bidirectional encoding of tweet conditioned on bidirectional encoding of target ($[\mathbf{c}^\rightarrow_3\;\mathbf{c}^\leftarrow_1]$). The stance is predicted using the last forward and reversed output representations ($[\mathbf{h}^\rightarrow_{9}\;\mathbf{h}^\leftarrow_4]$).}
\label{fig:cond}
\end{figure*}
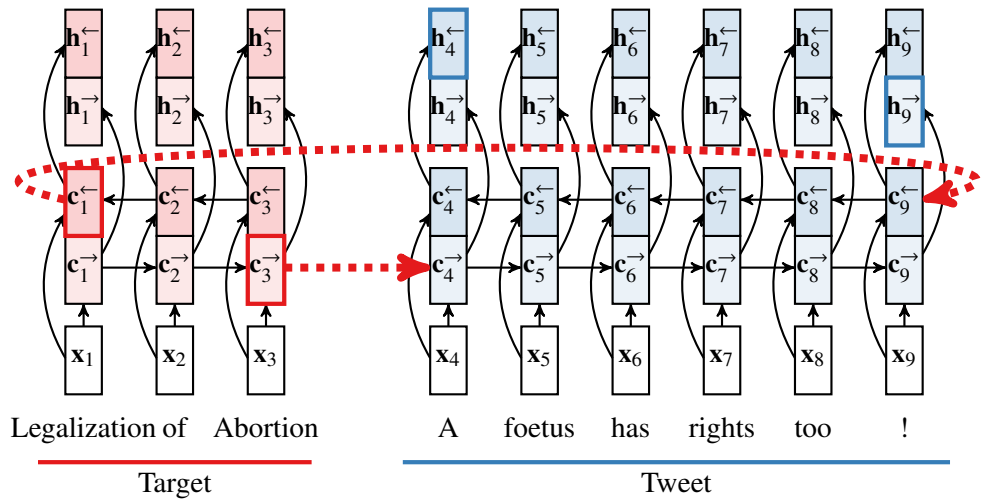

\subsection{Independent Encoding}
\label{sec:concat}

Our initial attempt to learn distributed representations for the tweets and the targets is to encode the target and tweet independently  as $k$-dimensional dense vectors using two LSTMs~(\cite{hochreiter1997long}).

  \begin{align*}
    \mathbf{H} &= \left[{
      \begin{array}{*{20}c}
        \mathbf{x}_t \\
        \mathbf{h}_{t-1}
      \end{array} }
    \right]\\
    \mathbf{i}_t &= \sigma(\mathbf{W}^i\mathbf{H}+\mathbf{b}^i)\\
    \mathbf{f}_t &= \sigma(\mathbf{W}^f\mathbf{H}+\mathbf{b}^f)\\
    \mathbf{o}_t &= \sigma(\mathbf{W}^o\mathbf{H}+\mathbf{b}^o)\\
    \mathbf{c}_t &= \mathbf{f}_t \odot \mathbf{c}_{t-1} + \mathbf{i}_t \odot
    \tanh(\mathbf{W}^c\mathbf{H}+\mathbf{b}^c)\\
    \mathbf{h}_t &= \mathbf{o}_t \odot \tanh(\mathbf{c}_t)
  \end{align*}
Here, $\mathbf{x}_t$ is an input vector at time step $t$, $\mathbf{c}_t$ denotes the LSTM memory, $\mathbf{h}_t \in \mathbb{R}^k$ is an output vector and the remaining weight matrices and biases are trainable parameters.
We concatenate the two output vector representations and classify the stance using the softmax over a non-linear projection 
  \begin{align*}
{\text{softmax}}(\tanh(\mathbf{W}^{\text{ta}}\mathbf{h}_{\text{target}}+\mathbf{W}^{\text{tw}}\mathbf{h}_{\text{tweet}} + \mathbf{b}))
\end{align*}
into the space of the three classes for stance detection where $\mathbf{W}^{\text{ta}}, \mathbf{W}^{\text{tw}} \in \mathbb{R}^{3\times k}$ are trainable weight matrices and $\mathbf{b}\in\mathbb{R}^3$ is a trainable class bias. 
This model learns target-independent distributed representations for the tweets and relies on the non-linear projection layer to incorporate the target in the stance prediction.

\subsection{Conditional Encoding}\label{sec:cond_enc_LSTM}
In order to learn target-dependent tweet representations, we use conditional encoding as previously applied to the task of recognizing textual entailment (\cite{rocktaschel2016reasoning}). We use one LSTM to encode the target as a fixed-length vector. Then, we encode the tweet with another LSTM, whose state is initialised with the representation of the target. Finally, we use the last output vector of the tweet LSTM to predict the stance of the target-tweet pair. 

This effectively allows the second LSTM to read the tweet in a target-specific manner, which is crucial since the stance of the tweet depends on the target (recall the Donald Trump example above).

\subsection{Bidirectional Conditional Encoding}
Bidirectional LSTMs (\cite{graves2005framewise}) have been shown to learn improved representations of sequences by encoding a sequence from left to right and from right to left. Therefore, we adapt the conditional encoding model from Section~\ref{sec:cond_enc_LSTM} to use bidirectional LSTMs, which represent the target and the tweet using two vectors for each of them, one obtained by reading the target and then the tweet left-to-right (as in the conditional LSTM encoding) and one obtained by reading them right-to-left. To achieve this, we initialise the state of the bidirectional LSTM that reads the tweet by the last state of the forward and reversed encoding of the target  (see Figure~\ref{fig:cond}). 
The bidirectional encoding allows the model to construct target-dependent representations of the tweet such that when each word is considered, they take into account both the left- and the right-hand side context.

\subsection{Unsupervised Pretraining}\label{sec:Pretraining}

In order to counter-balance the relatively small training data available (5628 instances in total), unsupervised pre-training is employed. It initialises the word embeddings used in the LSTMs with an appropriately trained word2vec model (\cite{mikolov2013efficient}).  Note that these embeddings are used only for initialisation, as we allow them to be optimised further during training. 

In more detail, we train a word2vec model on a corpus of 395,212 unlabelled tweets, collected with the Twitter Keyword Search API\footnote{\url{https://dev.twitter.com/rest/public/search}} between November 2015 and January 2016, plus all the tweets contained in the official SemEval 2016 Stance Detection datasets~(\cite{mohammad-etal-2016-semeval}). 
 The unlabelled tweets are collected so they contain the training, dev and test targets, using up to two keywords per target, namely ``hillary'', ``clinton'', ``trump'', ``climate'', ``femini'', ``aborti''.
Note that Twitter does not allow for regular expression search, so this is a free text search disregarding possible word boundaries.
We combine this large unlabelled corpus with the official training data and train a skip-gram word2vec model (dimensionality 100, 5 min words, context window of 5).
Tweets and targets are tokenised with the Twitter-adapted tokeniser twokenize\footnote{\url{https://github.com/leondz/twokenize}}. Subsequently, all tokens are normalised to lower case, URLs are removed, and stopword tokens are filtered (i.e. punctuation characters, Twitter-specific stopwords (``rt'', ``\#semst'', ``via'').

As demonstrated in our experiments, unsupervised pre-training is quite helpful, since it is difficult to learn representations for all the words using only the relatively small training datasets available.
Finally, to ensure that the proposed neural network architectures contribute to the performance, we also use the word vectors from word2vec in a Bag-of-Word-Vectors baseline (\textsf{BOWV}), in which the tweet and target representations are fed into a logistic regression classifier with L2 regularization~(\cite{scikit-learn}).

\setlength{\tabcolsep}{0.3em}
\begin{table}[t]
\fontsize{10}{10}\selectfont
\begin{center}
\begin{tabular}{l c c c c}
\toprule
\bf Corpus & \bf Favor & \bf Against & \bf None & \bf All  \\ 
\midrule
TaskA\_Tr+Dv & 1462 & 2684 & 1482 & 5628 \\
TaskA\_Tr+Dv\_HC & 224 & 722 & 332 & 1278 \\
TaskB\_Unlab & - & - & - & 278,013 \\
TaskB\_Auto-lab* & 4681 & 5095 & 4026 & 13,802 \\
TaskB\_Test & 148 & 299 & 260 & 707 \\
Crawled\_Unlab* & - & - & - & 395,212 \\
\bottomrule
\end{tabular}
\end{center}
\caption{\label{tab:DataStats} Data sizes of available corpora. \textsf{TaskA\_Tr+Dv\_HC} is the part of \textsf{TaskA\_Tr+Dv} with tweets for the target Hillary Clinton only, which we use for development. \textsf{TaskB\_Auto-lab} is an automatically labelled version of \textsf{TaskB\_Unlab}. Crawled\_Unlab is an unlabelled tweet corpus collected by us.}

\end{table}

\section{Experiments}

Experiments are performed on the SemEval 2016 Task 6 corpus for Stance Detection on Twitter~(\cite{mohammad-etal-2016-semeval}). 
We report experiments for two different experimental setups: one is the {\em unseen target} setup (Section~\ref{sec:UnseenTarget}), which is the main focus of this paper, i.e. detecting the stance of tweets towards previously unseen targets. 
We show that conditional encoding, by reading the tweets in a target-specific way, generalises to unseen targets better than baselines which ignore the target.
Next, we compare our approach to previous work in a {\em weakly supervised framework} (Section~\ref{sec:WeaklySup}) and show that our approach outperforms the state-of-the-art on the SemEval 2016 Stance Detection Subtask B corpus.

Table~\ref{tab:DataStats} lists the various corpora used in the experiments and their sizes. 
\textsf{TaskA\_Tr+Dv} is the official SemEval 2016 Twitter Stance Detection TaskA training and development corpus, which contain instances for the targets {\em Legalization of Abortion}, {\em Atheism}, {\em Feminist Movement}, {\em Climate Change is a Real Concern} and {\em Hillary Clinton}.
\textsf{TaskA\_Tr+Dv\_HC} is the part of the corpus which contains only the {\em Hillary Clinton} tweets, which we use for development purposes. \textsf{TaskB\_Test} is the TaskB test corpus on which we report results containing {\em Donald Trump} testing instances. \textsf{TaskB\_Unlab} is an unlabelled corpus containing {\em Donald Trump} tweets supplied by the task organisers, and \textsf{TaskB\_Auto-lab*} is an automatically labelled version of a small portion of the corpus for the weakly supervised stance detection experiments reported in Section~\ref{sec:WeaklySup}. Finally, \textsf{Crawled\_Unlab*} is a corpus we collected for unsupervised pre-training (see Section~\ref{sec:Pretraining}).

For all experiments, the official task evaluation script is used. 
Predictions are postprocessed so that if the target is contained in a tweet, the highest-scoring non-neutral stance is chosen. 
This was motivated by the observation that in the training data most target-containing tweets express a stance, with only 16\% of them 
being neutral.

\subsection{Methods}

We compare the following baseline methods: 
\begin{itemize}[noitemsep]
\item{SVM trained with word and character tweet n-grams features (\textsf{SVM-ngrams-comb})~(\cite{mohammad-etal-2016-semeval})}
\item{a majority class baseline (\textsf{Majority baseline}), reported in~\cite{mohammad-etal-2016-semeval}}
\item{bag of word vectors (\textsf{BoWV}) (see Section~\ref{sec:Pretraining})}
\item{independent encoding of tweet and the target with two LSTMs (\textsf{Concat}) (see Section~\ref{sec:concat})}
\item{encoding of the tweet only with an LSTM (\textsf{TweetOnly}) (see Section~\ref{sec:concat}}
\end{itemize}
to three versions of conditional encoding:
\begin{itemize}[noitemsep]
\item{target conditioned on tweet (\textsf{TarCondTweet})}
\item{tweet conditioned on target (\textsf{TweetCondTar})}
\item{a bidirectional encoding model (\textsf{BiCond})}
\end{itemize}

\begin{table}[t]
\fontsize{10}{10}\selectfont
\begin{center}
\begin{tabular}{l c c c c}
\toprule
\bf Method & \bf Stance & \bf P & \bf R & \bf F1 \\
\midrule
\multirow{3}{*}{\textsf{BoWV}} & FAVOR & 0.2444 & 0.0940 & 0.1358 \\
 & AGAINST & 0.5916 & 0.8626 & 0.7019 \\ \cline{2-5}
 & Macro & & & 0.4188 \\
 \midrule
\multirow{3}{*}{\textsf{TweetOnly}} & FAVOR & 0.2127 & 0.5726 & 0.3102 \\
 & AGAINST & 0.6529 & 0.4020 & 0.4976 \\ \cline{2-5}
 & Macro & & & 0.4039 \\
 \midrule
 \multirow{3}{*}{\textsf{Concat}} & FAVOR & 0.1811 & 0.6239 & 0.2808 \\
 & AGAINST & 0.6299 & 0.4504 & 0.5252 \\ \cline{2-5}
 & Macro & & & 0.4030 \\
 \midrule \midrule
 \multirow{3}{*}{\textsf{TarCondTweet}} & FAVOR & 0.3293 & 0.3649 & 0.3462 \\
 & AGAINST & 0.4304 & 0.5686 & 0.4899 \\ \cline{2-5}
 & Macro & & & 0.4180 \\
 \midrule
 \multirow{3}{*}{\textsf{TweetCondTar}} & FAVOR & 0.1985 & 0.2308 & 0.2134 \\
 & AGAINST & 0.6332 & 0.7379 & 0.6816 \\ \cline{2-5} 
 & Macro & & & 0.4475 \\
 \midrule

  \multirow{3}{*}{\textsf{BiCond}} & FAVOR & 0.2588 & 0.3761 & 0.3066 \\
 & AGAINST & 0.7081 & 0.5802 & 0.6378 \\ \cline{2-5}
& Macro & & & \bf{0.4722} \\
\bottomrule
\end{tabular}
\end{center}
\caption{\label{tab:ResultsDev} Results for the {\em unseen target} stance detection  development setup.}
\end{table}

\begin{table}[t]
\fontsize{10}{10}\selectfont
\begin{center}
\begin{tabular}{l c c c c}
\toprule
\bf Method & \bf Stance & \bf P & \bf R & \bf F1 \\
\midrule
\multirow{3}{*}{\textsf{BoWV}} & FAVOR & 0.3158 & 0.0405 & 0.0719 \\
 & AGAINST & 0.4316 & 0.8963 & 0.5826 \\ \cline{2-5}
 & Macro & & & 0.3272 \\
 \midrule
 \multirow{3}{*}{\textsf{TweetOnly}} & FAVOR & 0.2767 & 0.3851 & 0.3220 \\
 & AGAINST & 0.4225 & 0.5284 & 0.4695 \\ \cline{2-5}
 & Macro & & & 0.3958 \\
  \midrule
 \multirow{3}{*}{\textsf{Concat}} & FAVOR & 0.3145 & 0.5270 & 0.3939 \\
 & AGAINST & 0.4452 & 0.4348 & 0.4399 \\ \cline{2-5}
 & Macro & & & 0.4169 \\
 \midrule \midrule
\multirow{3}{*}{\textsf{TarCondTweet}} & FAVOR & 0.2322 & 0.4188 & 0.2988 \\
 & AGAINST & 0.6712 & 0.6234 & 0.6464 \\ \cline{2-5}
 & Macro & & & 0.4726 \\
\midrule
\multirow{3}{*}{\textsf{TweetCondTar}} & FAVOR & 0.3710 & 0.5541 & 0.4444 \\
 & AGAINST & 0.4633 & 0.5485 & 0.5023 \\ \cline{2-5}
& Macro & & & 0.4734 \\
\midrule
\multirow{3}{*}{\textsf{BiCond}} & FAVOR & 0.3033 & 0.5470 & 0.3902 \\
 & AGAINST & 0.6788 & 0.5216 & 0.5899 \\ \cline{2-5}
 & Macro & & & \bf{0.4901} \\
\bottomrule
\end{tabular}
\end{center}
\caption{\label{tab:ResultsTest} Results for the {\em unseen target} stance detection  test setup.}
\end{table}

\section{Unseen Target Stance Detection}\label{sec:UnseenTarget}

As explained, the challenge is to learn a model without any manually labelled training data for the test target, but only using the data from the Task A targets. In order  to avoid using any labelled data for {\em Donald Trump}, while still having a (labelled) development set to tune and evaluate our models, we used the tweets labelled for {\em Hillary Clinton} as a development set and the tweets for the remaining four targets as training. We refer to this as the \emph{development setup}, and all models are tuned using this setup.
The labelled {\em Donald Trump} tweets were only used in reporting our final results.
For the final results we train on all the data from the development setup and evaluate on the official Task B test set, i.e. the {\em Donald Trump} tweets. We refer to this as our \emph{test setup}. 

Based on a small grid search using the development setup, the following settings for LSTM-based models were chosen: 
input layer size 100 (equal to word embedding dimensions), hidden layer size 60, training for max 50 epochs with initial learning rate 1e-3 using ADAM~(\cite{kingma2014adam}) for optimisation,
 dropout 0.1.
Using one, relatively small hidden layer and dropout help avoid overfitting.

\subsection{Results and Discussion}

Results for the unseen target setting show how well conditional encoding is suited for learning target-dependent representations of tweets, and crucially, how well such representations generalise to unseen targets. 
The best performing method on both development (Table~\ref{tab:ResultsDev}) and test setups (Table~\ref{tab:ResultsTest}) is \textsf{BiCond}, which achieves an F1 of 0.4722 and 0.4901 respectively. Notably, \textsf{Concat}, which learns an indepedent encoding of the target and the tweets, does not achieve big F1 improvements over \textsf{TweetOnly}, which learns a representation of the tweets only. This shows that it is not only important to learn target-depedent encodings, but also the way in which they are learnt matters.
Models that learn to condition the encoding of tweets on targets outperform all baselines on the test set. 

It is further worth noting that the Bag-of-Word-Vectors baseline achieves results comparable with \textsf{TweetOnly}, \textsf{Concat} and one of the conditional encoding models, \textsf{TarCondTweet}, on the dev set, even though it achieves significantly lower performance on the test set. This indicates that the pre-trained word embeddings on their own are already very useful for stance detection.

Our best result in the test setup with \textsf{BiCond} is currently the second highest reported result on the Stance Detection corpus, however the first, third and fourth best approaches achieved their results by automatically labelling {\em Donald Trump} training data.  \textsf{BiCond} for the unseen target setting outperforms the third and fourth best approaches by a large margin (5 and 7 points in Macro F1, respectively), as can be seen in Table~\ref{tab:ResultsTestSoA}. Results for weakly supervised stance detection are discussed in the next section.

\paragraph{Unsupervised Pre-Training}

Table~\ref{tab:ResultsEmbIni} shows the effect of unsupervised pre-training of word embeddings, and furthermore, the results of sharing these representations between the tweets and targets, on the development set.
The first set of results is with a uniformly \textsf{Random} embeddings initialisation in $[-0.1, 0.1]$. \textsf{PreFixed} uses the pre-trained word embeddings, whereas \textsf{PreCont} uses the pre-trained word embeddings and continues training them during LSTM training. 

Our results show that, in the absence of a large labelled training dataset, unsupervised pre-training of word embeddings 
is more helpful than random initialisation of embeddings.
\textsf{Sing} vs \textsf{Sep} shows the difference between using shared vs two separate embeddings matrices for looking up the word embeddings. \textsf{Sing} means the word representations for tweet and target vocabularies are shared, whereas \textsf{Sep} means they are different.
Using shared embeddings performs better, which we hypothesise is because the tweets contain some mentions of targets that are tested.

\paragraph{Target in Tweet vs Not in Tweet}

Table~\ref{tab:ErrorAna} shows results on the development set for \textsf{BiCond}, compared to the best unidirectional encoding model, \textsf{TweetCondTar} and the baseline \textsf{Concat}, split by tweets that contain the target and those that do not.
All three models perform well when the target is mentioned in the tweet, but less so when the targets are not mentioned explicitly. In the case where the target is mentioned in the tweet, biconditional encoding outperforms unidirectional encoding and unidirectional encoding outperforms \textsf{Concat}. This shows that conditional encoding is able to learn useful dependencies between the tweets and the targets.

\begin{table}[t]
\fontsize{10}{10}\selectfont
\begin{center}
\begin{tabular}{l l c c c c}
\toprule
\bf EmbIni & \bf NumMatr & \bf Stance & \bf P & \bf R & \bf F1 \\
\midrule
\multirow{6}{*}{\textsf{Random}} & \multirow{3}{*}{\textsf{Sing}} & FAVOR & 0.1982 & 0.3846 & 0.2616 \\
 & & AGAINST & 0.6263 & 0.5929 & 0.6092 \\ \cline{3-6}
 & & Macro & & & 0.4354 \\
\cmidrule{2-6}
 & \multirow{3}{*}{\textsf{Sep}} & FAVOR & 0.2278 & 0.5043 & 0.3138 \\
& & AGAINST & 0.6706 & 0.4300 & 0.5240 \\ \cline{3-6}
& & Macro & & & 0.4189 \\
\midrule
\multirow{6}{*}{\textsf{PreFixed}} & \multirow{3}{*}{\textsf{Sing}} & FAVOR & 0.6000 & 0.0513 & 0.0945 \\
& & AGAINST & 0.5761 & 0.9440 & 0.7155 \\ \cline{3-6}
& & Macro & & & 0.4050 \\
\cmidrule{2-6}
& \multirow{3}{*}{\textsf{Sep}} & FAVOR & 0.1429 & 0.0342 & 0.0552 \\
& & AGAINST & 0.5707 & 0.9033 & 0.6995 \\ \cline{3-6}
& & Macro & & & 0.3773 \\
\midrule
\multirow{6}{*}{\textsf{PreCont}} & \multirow{3}{*}{\textsf{Sing}} & FAVOR & 0.2588 & 0.3761 & 0.3066 \\
& & AGAINST & 0.7081 & 0.5802 & 0.6378 \\ \cline{3-6}
& & Macro & & & \bf{0.4722} \\
\cmidrule{2-6}
& \multirow{3}{*}{\textsf{Sep}} & FAVOR & 0.2243 & 0.4103 & 0.2900 \\
& & AGAINST & 0.6185 & 0.5445 & 0.5792 \\ \cline{3-6}
& & Macro & & & 0.4346 \\
\bottomrule
\end{tabular}
\end{center}
\caption{\label{tab:ResultsEmbIni} Results for the {\em unseen target} stance detection development setup using \textsf{BiCond}, with single vs separate embeddings matrices for tweet and target and different initialisations} 
\end{table}

\begin{table}[t]
\fontsize{10}{10}\selectfont
\begin{center}
\begin{tabular}{l l c c c c}
\toprule
\bf Method & \bf inTwe & \bf Stance & \bf P & \bf R & \bf F1 \\
\midrule
 \multirow{6}{*}{\textsf{Concat}} & \multirow{3}{*}{\textsf{Yes}} & FAVOR     & 0.3153 & 0.6214 & 0.4183 \\
& & AGAINST & 0.7438 & 0.4630 & 0.5707 \\ \cline{3-6} 
 & & Macro & & & 0.4945 \\
\cmidrule{2-6}
 & \multirow{3}{*}{\textsf{No}} & FAVOR & 0.0450 & 0.6429 & 0.0841 \\
 & & AGAINST & 0.4793 & 0.4265 & 0.4514 \\ \cline{3-6}
& & Macro & & & 0.2677 \\
 \midrule
 \multirow{6}{*}{\textsf{TweetCondTar}} & \multirow{3}{*}{\textsf{Yes}} & FAVOR & 0.3529 & 0.2330 & 0.2807 \\
& & AGAINST & 0.7254 & 0.8327 & 0.7754 \\ \cline{3-6}
& & Macro & & & 0.5280 \\
\cmidrule{2-6}
 & \multirow{3}{*}{\textsf{No}} & FAVOR & 0.0441 & 0.2143 & 0.0732 \\
 & & AGAINST & 0.4663 & 0.5588 & 0.5084 \\ \cline{3-6}
 & & Macro & & & 0.2908 \\
 \midrule
 \multirow{6}{*}{\textsf{BiCond}} & \multirow{3}{*}{\textsf{Yes}} & FAVOR & 0.3585 & 0.3689 & 0.3636
\\
& & AGAINST & 0.7393 & 0.7393 & 0.7393 \\ \cline{3-6}
&  & Macro & & & \bf{0.5515} \\
\cmidrule{2-6}
 & \multirow{3}{*}{\textsf{No}} & FAVOR & 0.0938 & 0.4286 & 0.1538 \\
 & & AGAINST & 0.5846 & 0.2794 & 0.3781 \\ \cline{3-6}
 & & Macro & & & 0.2660 \\
 
\bottomrule
\end{tabular}
\end{center}
\caption{\label{tab:ErrorAna} Results for the {\em unseen target} stance detection development setup for tweets containing the target vs tweets not containing the target.}
\end{table}

\begin{table}[t]
\fontsize{10}{10}\selectfont
\begin{center}
\begin{tabular}{l c c c c}
\toprule
\bf Method & \bf Stance & \bf P & \bf R & \bf F1 \\
\midrule
\multirow{3}{*}{\textsf{BoWV}} & FAVOR & 0.5156 & 0.6689 & 0.5824 \\
 & AGAINST & 0.6266 & 0.3311 & 0.4333 \\ \cline{2-5}
 & Macro & & & 0.5078 \\
\midrule
\multirow{3}{*}{\textsf{TweetOnly}} & FAVOR & 0.5284 & 0.6284 & 0.5741 \\
& AGAINST & 0.5774 & 0.4615 & 0.5130 \\ \cline{2-5}
& Macro & & & 0.5435 \\
\midrule
\multirow{3}{*}{\textsf{Concat}} & FAVOR & 0.5506 & 0.5878 & 0.5686 \\
 & AGAINST & 0.5794 & 0.4883 & 0.5299 \\ \cline{2-5}
 & Macro & & & 0.5493 \\
 \midrule \midrule
\multirow{3}{*}{\textsf{TarCondTweet}} & FAVOR & 0.5636 & 0.6284 & 0.5942 \\
 & AGAINST & 0.5947 & 0.4515 & 0.5133 \\ \cline{2-5}
& Macro & & & 0.5538 \\
\midrule
\multirow{3}{*}{\textsf{TweetCondTar}} & FAVOR & 0.5868 & 0.6622 & 0.6222 \\
 & AGAINST & 0.5915 & 0.4649 & 0.5206 \\ \cline{2-5}
 & Macro & & & 0.5714 \\
 \midrule
\multirow{3}{*}{\textsf{BiCond}}
& FAVOR & 0.6268 & 0.6014 & 0.6138 \\
& AGAINST & 0.6057 & 0.4983 & 0.5468 \\ \cline{2-5}
& Macro & & & \bf{0.5803} \\
\bottomrule
\end{tabular}
\end{center}
\caption{\label{tab:ResultsTest2} Stance Detection test results for weakly supervised setup, trained on automatically labelled pos+neg+neutral Trump data, and reported on the official test set.}
\end{table}

\begin{table}[t]
\fontsize{10}{10}\selectfont
\begin{center}
\begin{tabular}{l c c}
\toprule
\bf Method & \bf Stance & \bf F1 \\
\midrule
\multirow{3}{*}{\textsf{SVM-ngrams-comb} ({\em Unseen Target})} 
 & FAVOR &  0.1842 \\
 & AGAINST & 0.3845 \\  \cline{2-3}
& Macro & 0.2843 \\ 
\midrule
\multirow{3}{*}{\textsf{Majority baseline} ({\em Unseen Target})} 
 & FAVOR &  0.0 \\
 & AGAINST & 0.5944 \\  \cline{2-3}
& Macro & 0.2972 \\ 
\midrule
\multirow{3}{*}{\textsf{BiCond} ({\em{Unseen Target}})} 
& FAVOR & 0.3902 \\
 & AGAINST  & 0.5899 \\  \cline{2-3}
 & Macro & \bf{0.4901} \\ 
 \midrule \midrule
  \multirow{3}{*}{\textsf{INF-UFRGS} ({\em Weakly Supervised*})}
& FAVOR & 0.3256 \\
 & AGAINST  & 0.5209 \\ \cline{2-3}
 & Macro & 0.4232 \\
 \midrule
 \multirow{3}{*}{\textsf{LitisMind} ({\em Weakly Supervised*})}
& FAVOR & 0.3004 \\
 & AGAINST  & 0.5928 \\ \cline{2-3}
 & Macro & 0.4466 \\
 \midrule
 \multirow{3}{*}{\textsf{pkudblab} ({\em Weakly Supervised*})}
& FAVOR & 0.5739 \\
& AGAINST & 0.5517 \\  \cline{2-3}
& Macro & 0.5628 \\ 
\midrule
\multirow{3}{*}{\textsf{BiCond} ({\em Weakly Supervised})}
& FAVOR & 0.6138 \\
& AGAINST & 0.5468 \\  \cline{2-3}
& Macro & \bf{0.5803} \\
\bottomrule
\end{tabular}
\end{center}
\caption{\label{tab:ResultsTestSoA} Stance Detection test results, compared against the state of the art. \textsf{SVM-ngrams-comb} and \textsf{Majority baseline} are reported in~\protect\cite{mohammad-etal-2016-semeval}, pkudblab in  \protect\cite{StanceSemEval2016pkudblab}, LitisMind in  \protect\cite{StanceSemEval2016MITRE}, INF-UFRGS in  \protect\cite{StanceSemEval2016inf}}
\end{table}

\section{Weakly Supervised Stance Detection}\label{sec:WeaklySup}

The previous section showed the usefulness of conditional encoding for unseen target stance detection and compared results against internal baselines. The goal of experiments reported in this section is to compare against participants in the SemEval 2016 Stance Detection Task B. While we consider an {\em unseen target} setup, most submissions, including the three highest ranking ones for Task B, pkudblab~(\cite{StanceSemEval2016pkudblab}), LitisMind~(\cite{StanceSemEval2016MITRE}) and INF-UFRGS~(\cite{StanceSemEval2016inf}) considered a different experimental setup. They automatically annotated training data for the test target {\em Donald Trump}, thus rendering the task as a weakly supervised seen target stance detection. The pkudblab system uses a deep convolutional neural network that learns to make 2-way predictions on automatically labelled positive and negative training data for {\em Donald Trump}. The neutral class is predicted according to rules which are applied at test time.

Since the best performing systems which participated in the shared task consider a weakly supervised setup, we further compare our proposed approach to the state-of-the-art using such a weakly supervised setup. Note that, even though pkudblab, LitisMind and INF-UFRGS also use regular expressions to label training data automatically, the resulting datasets were not made available to us. Therefore, we had to develop our own automatic labelling method and dataset, which will be made publicly available on publication.

\paragraph{Weakly Supervised Test Setup}
For this setup, the unlabelled {\em Donald Trump} corpus \textsf{TaskB\_Unlab} is annotated automatically. For this purpose we created a small set of regular expressions\footnote{Note that ``$|$'' indiates ``or'', ( ?) indicates optional space}, based on inspection of the \textsf{TaskB\_Unlab} corpus, expressing positive and negative stance towards the target. The regular expressions for the positive stance were:
\begin{itemize}[noitemsep,nolistsep]
\item{make( ?)america( ?)great( ?)again}
\item{trump( ?)(for$|$4)( ?)president}
\item{votetrump}
\item{trumpisright}
\item{the truth}
\item{\#trumprules}
\end{itemize}
The keyphrases for negative stance were: \#dumptrump, \#notrump, \#trumpwatch, racist, idiot, fired\\

A tweet is labelled as positive if one of the positive expressions is detected, else negative if a negative expressions is detected. If neither are detected, the tweet is annotated as neutral randomly with 2\% chance. The resulting corpus size per stance is shown in Table~\ref{tab:DataStats}.
The same hyperparameters for the LSTM-based models are used as for the {\em unseen target} setup described in the previous section.

\subsection{Results and Discussion}

Table~\ref{tab:ResultsTest2} lists our results in the weakly supervised setting.
Table~\ref{tab:ResultsTestSoA} shows all our results, including those using the unseen target setup, compared against the state-of-the-art on the stance detection corpus. It further lists baselines reported by~\cite{mohammad-etal-2016-semeval}, namely a majority class baseline (\textsf{Majority baseline}), and a method using 1 to 3-gram bag-of-word and character n-gram features (\textsf{SVM-ngrams-comb}), which are extracted from the tweets and used to train a 3-way SVM classifier.
Bag-of-word baselines (\textsf{BoWV}, \textsf{SVM-ngrams-comb}) achieve results comparable to the majority baseline (F1 of 0.2972), which shows how difficult the task is.
The baselines which only extract features from the tweets, \textsf{SVM-ngrams-comb} and \textsf{TweetOnly} perform worse than the baselines which also learn representations for the targets (\textsf{BoWV}, \textsf{Concat}). 
By training conditional encoding models on automatically labelled stance detection data we achieve state-of-the-art results.
The best result (F1 of 0.5803) is achieved with the bi-directional conditional encoding model (\textsf{BiCond}). This shows that such models are suitable for unseen, as well as seen target stance detection.

\section{Related Work}

{\bf Stance Detection}: 
Previous work mostly considered target-specific stance prediction in debates (\cite{hasan-ng-2013-stance,walker-etal-2012-stance}) or student essays~(\cite{Faulkner14}). Recent work studied Twitter-based stance detection~(\cite{RajadesinganL14}), which is also a task at SemEval 2016~(\cite{mohammad-etal-2016-semeval}). The latter is more challenging than stance detection in debates because, in addition to irregular language, the~(\cite{mohammad-etal-2016-semeval}) dataset is offered without any context, e.g., conversational structure or tweet metadata. The targets are also not always mentioned in the tweets, which makes the task very challenging~(\cite{augenstein-etal-2016-usfd}) and distinguishes it from target-dependent (\cite{conf/ijcai/VoZ15,Zhang2016Gated,alghunaim-EtAl:2015:VSM-NLP}) and open-domain target-dependent sentiment analysis (\cite{mitchell-EtAl:2013:EMNLP,zhang-zhang-vo:2015:EMNLP}).

{\bf Conditional Encoding}:
Conditional encoding has been applied in the related task of recognising textual entailment~(\cite{rocktaschel2016reasoning}), using a dataset of half a million training examples~(\cite{bowman-etal-2015-large}) and numerous different hypotheses. Our experiments show that conditional encoding is also successful on a relatively small training set and when applied to an unseen testing target. Moreover, we augment conditional encoding with bidirectional encoding and demonstrate the added benefit of unsupervised pre-training of word embeddings on unlabelled domain data.

\section{Conclusions and Future Work}
This paper showed that conditional LSTM encoding is a successful approach to stance detection for unseen targets. Our unseen target bidirectional conditional encoding approach achieves the second best results reported to date on the SemEval 2016 Twitter Stance Detection corpus. In a seen target minimally supervised scenario, as considered by prior work, our approach achieves the best results to date on the SemEval Task B dataset. 
We further show that in the absence of large labelled corpora, unsupervised pre-training can be used to learn target representations for stance detection and improves results on the SemEval corpus. Future work will investigate further the challenge of stance detection for tweets which do not contain explicit mentions of the target.

\chapter{Discourse-Aware Rumour Stance Classification}\label{ch:discourse}

\boxabstract{Rumour stance classification, defined as classifying the stance of specific social media posts into one of supporting, denying, querying or commenting on an earlier post, is becoming of increasing interest to researchers. While most previous work has focused on using individual tweets as classifier inputs, here we report on the performance of sequential classifiers that exploit the discourse features inherent in social media interactions or `conversational threads'. Testing the effectiveness of four sequential classifiers -- Hawkes Processes, Linear-Chain Conditional Random Fields (Linear CRF), Tree-Structured Conditional Random Fields (Tree CRF) and Long Short Term Memory networks (LSTM) -- on eight datasets associated with breaking news stories, and looking at different types of local and contextual features, our work sheds new light on the development of accurate stance classifiers. We show that sequential classifiers that exploit the use of discourse properties in social media conversations while using only local features, outperform non-sequential classifiers. Furthermore, we show that LSTM using a reduced set of features can outperform the other sequential classifiers; this performance is consistent across datasets and across types of stances. To conclude, our work also analyses the different features under study, identifying those that best help characterise and distinguish between stances, such as supporting tweets being more likely to be accompanied by evidence than denying tweets. We also set forth a number of directions for future research.}\blfootnote{\fullcite{journals/ipm/ZubiagaKLPLBCA18}}

\section{Introduction}

Social media platforms have established themselves as important sources for learning about the latest developments in breaking news. People increasingly use social media for news consumption (\cite{hermida2012share,mitchell2015millennials,zubiaga2015real}), while media professionals, such as journalists, increasingly turn to social media for news gathering (\cite{zubiaga2013curating}) and for gathering potentially exclusive updates from eyewitnesses (\cite{diakopoulos2012finding,tolmie2017microblog}). Social media platforms such as Twitter are a fertile and prolific source of breaking news, occasionally even outpacing traditional news media organisations (\cite{kwak2010twitter}). This has led to the development of multiple data mining applications for mining and discovering events and news from social media (\cite{dong2015multiscale,stilo2016efficient}). However, the use of social media also comes with the caveat that some of the reports are necessarily rumours at the time of posting, as they have yet to be corroborated and verified (\cite{malon-2018-team,procter2013readingb,procter2013readinga}). The presence of rumours in social media has hence provoked a growing interest among researchers for devising ways to determine veracity in order to avoid the diffusion of misinformation (\cite{derczynski2015pheme}).

Resolving the veracity of social rumours requires the development of a rumour classification system and we described in (\cite{zubiaga2017exploiting}), a candidate architecture for such a system consisting of the following four components: (1) detection, where emerging rumours are identified, (2) tracking, where those rumours are monitored to collect new related tweets, (3) stance classification, where the views expressed by different tweet authors are classified, and (4) veracity classification, where knowledge garnered from the stance classifier is put together to determine the likely veracity of a rumour.

In this work we focus on the development of the third component, a stance classification system, which is crucial to subsequently determining the veracity of the underlying rumour. The stance classification task consists in determining how individual posts in social media observably orientate to the postings of others (\cite{walker-etal-2012-stance,qazvinian2011rumor}). For instance, a post replying with ``no, that's definitely false'' is \textit{denying} the preceding claim, whereas ``yes, you're right'' is \textit{supporting} it. It has been argued that aggregation of the distinct stances evident in the multiple tweets discussing a rumour could help in determining its likely veracity, providing, for example, the means to flag highly disputed rumours as being potentially false (\cite{malon-2018-team}). This approach has been justified by recent research that has suggested that the aggregation of the different stances expressed by users can be used for determining the veracity of a rumour (\cite{derczynski2015pheme,liu2015realtime}).

In this work we examine in depth the use of so-called sequential approaches to the rumour stance classification task. Sequential classifiers are able to utilise the discursive nature of social media (\cite{tolmie2017microblog}), learning from how `conversational threads' evolve for a more accurate classification of the stance of each tweet. The use of sequential classifiers to model the conversational properties inherent in social media threads is still in its infancy. For example, in preliminary work we showed that a sequential classifier modelling the temporal sequence of tweets outperforms standard classifiers (\cite{lukasik-etal-2016-hawkes,zubiaga2016coling}). Here we extend this preliminary experimentation in four different directions that enable exploring further the stance classification task using sequential classifiers: (1) we perform a comparison of a range of sequential classifiers, including a Hawkes Process classifier, a Linear CRF, a Tree CRF and an LSTM; (2) departing from the use of only local features in our previous work, we also test the utility of contextual features to model the conversational structure of Twitter threads; (3) we perform a more exhaustive analysis of the results looking into the impact of different datasets and the depth of the replies in the conversations on the classifiers' performance, as well as performing an error analysis; and (4) we perform an analysis of features that gives insight into what characterises the different kinds of stances observed around rumours in social media. To the best of our knowledge, dialogical structures in Twitter have not been studied in detail before for classifying each of the underlying tweets and our work is the first to evaluate it exhaustively for stance classification. Twitter conversational threads are identifiable by the relational features that emerge as users respond to each others' postings, leading to tree-structured interactions. The motivation behind the use of these dialogical structures for determining stance is that users' opinions are expressed and evolve in a discursive manner, and that they are shaped by the interactions with other users.

The work presented here advances research in rumour stance classification by performing an exhaustive analysis of different approaches to this task. In particular, we make the following contributions:

\begin{itemize}
\item We perform an analysis of whether and the extent to which use of the sequential structure of conversational threads can improve stance classification in comparison to a classifier that determines a tweet's stance from the tweet in isolation. To do so, we evaluate the effectiveness of a range of sequential classifiers: (1) a state-of-the-art classifier that uses Hawkes Processes to model the temporal sequence of tweets (\cite{lukasik-etal-2016-hawkes}); (2) two different variants of Conditional Random Fields (CRF), i.e., a linear-chain CRF and a tree CRF; and (3) a classifier based on Long Short Term Memory (LSTM) networks. We compare the performance of these sequential classifiers with non-sequential baselines, including the non-sequential equivalent of CRF, a Maximum Entropy classifier.

\item We perform a detailed analysis of the results broken down by dataset and by depth of tweet in the thread, as well as an error analysis to further understand the performance of the different classifiers. We complete our analysis of results by delving into the features, and exploring whether and the extent to which they help characterise the different types of stances.
\end{itemize}

Our results show that sequential approaches do perform substantially better in terms of macro-averaged F1 score, proving that exploiting the dialogical structure improves classification performance. Specifically, the LSTM achieves the best performance in terms of macro-averaged F1 scores, with a performance that is largely consistent across different datasets and different types of stances. Our experiments show that LSTM performs especially well when only local features are used, as compared to the rest of the classifiers, which need to exploit contextual features to achieve comparable -- yet still inferior -- performance scores. Our findings reinforce the importance of leveraging conversational context in stance classification. Our research also sheds light on open research questions that we suggest should be addressed in future work. Our work here complements other components of a rumour classification system that we implemented in the PHEME project, including a rumour detection component (\cite{zubiaga2016learning,zubiaga2017exploiting}), as well as a study into the diffusion of and reactions to rumour (\cite{zubiaga2016analysing}).

\section{Related Work}

Stance classification is applied in a number of different scenarios and domains, usually aiming to classify stances as one of ``in favour'' or ``against''. This task has been studied in political debates (\cite{hasan2013extra,walker2012your}), in arguments in online fora (\cite{hasan-ng-2013-stance,sridhar2014collective}) and in attitudes towards topics of political significance (\cite{augenstein-etal-2016-stance,mohammad-etal-2016-semeval,augenstein-etal-2016-usfd}). In work that is closer to our objectives, stance classification has also been used to help determine the veracity of information in micro-posts (\cite{qazvinian2011rumor}), often referred to as \textit{rumour stance classification} (\cite{lukasik2015classifying,lukasik-etal-2016-hawkes,procter2013readinga,zubiaga2016coling}). The idea behind this task is that the aggregation of distinct stances expressed by users in social media can be used to assist in deciding if a report is actually true or false (\cite{derczynski2015pheme}). This may be particularly useful in the context of rumours emerging during breaking news stories, where reports are released piecemeal and which may be lacking authoritative review; in consequence, using the `wisdom of the crowd' may provide a viable, alternative approach. The types of stances observed while rumours circulate, however, tend to differ from the original ``in favour/against'', and different types of stances have been discussed in the literature, as we review next.

Rumour stance classification of tweets was introduced in early work by \cite{qazvinian2011rumor}. The line of research initiated by \cite{qazvinian2011rumor} has progressed substantially with revised definitions of the task and hence the task tackled in this paper differs from this early work in a number of aspects. \cite{qazvinian2011rumor} performed 2-way classification of each tweet as \textit{supporting} or \textit{denying} a long-standing rumour such as disputed beliefs that \textit{Barack Obama is reportedly Muslim}. The authors used tweets observed in the past to train a classifier, which was then applied to new tweets discussing the same rumour. In recent work, rule-based methods have been proposed as a way of improving on \cite{qazvinian2011rumor}'s baseline method; however, rule-based methods are likely to be difficult -- if not impossible -- to generalise to new, unseen rumours. \cite{hamidian2016rumor} extended that work to analyse the extent to which a model trained from historical tweets could be used for classifying new tweets discussing the same rumour.

The work we present here has three different objectives towards improving stance classification. First, we aim to classify the stance of tweets towards rumours that emerge while breaking news stories unfold; these rumours are unlikely to have been observed before and hence rumours from previously observed events, which are likely to diverge, need to be used for training. As far as we know, only work by \cite{lukasik2015classifying,lukasik2016using,lukasik-etal-2016-hawkes} has tackled stance classification in the context of breaking news stories applied to new rumours. \cite{zeng2016unconfirmed} have also performed stance classification for rumours around breaking news stories, but overlapping rumours were used for training and testing. \cite{augenstein-etal-2016-stance,augenstein-etal-2016-usfd} studied stance classification of unseen events in tweets, but ignored the conversational structure. Second, recent research has proposed that a 4-way classification is needed to encompass responses seen in breaking news stories (\cite{procter2013readinga,zubiaga2016analysing}). Moving away from the 2-way classification above, which \cite{procter2013readinga} found to be limited in the context of rumours during breaking news, we adopt this expanded scheme to include tweets that are \textit{supporting}, \textit{denying}, \textit{querying} or \textit{commenting} rumours. This adds more categories to the scheme used in early work, where tweets would only support or deny a rumour, or where a distinction between querying and commenting is not made (\cite{augenstein-etal-2016-stance,mohammad-etal-2016-semeval,augenstein-etal-2016-usfd}). Moreover, our approach takes into account the interaction between users on social media, whether it is about appealing for more information in order to corroborate a rumourous statement (\textit{querying}) or to post a response that does not contribute to the resolution of the rumour's veracity (\textit{commenting}). Finally -- and importantly -- instead of dealing with tweets as single units in isolation, we exploit the emergent structure of interactions between users on Twitter, building a classifier that learns the dynamics of stance in tree-structured conversational threads by exploiting its underlying interactional features. While these interactional features do not, in the final analysis, map directly onto those of conversation as revealed by Conversation Analysis (\cite{sacks1974simplest}), we argue that there are sufficient relational similarities to justify this approach (\cite{tolmie2017ugc}). The closest work is by \cite{ritter2010unsupervised} who modelled linear sequences of replies in Twitter conversational threads with Hidden Markov Models for dialogue act tagging, but the tree structure of the thread as a whole was not exploited.

As we were writing this article, we also organised, in parallel, a shared task on rumour stance classification, RumourEval (\cite{derczynski2017semeval}), at the well-known natural language processing competition SemEval 2017. The subtask A consisted in stance classification of individual tweets discussing a rumour within a conversational thread as one of \textit{support}, \textit{deny}, \textit{query}, or \textit{comment}, which specifically addressed the task presented in this paper. Eight participants submitted results to this task, including work by \cite{kochkina-etal-2017-turing} using an LSTM classifier which is being also analysed in this paper. In this shared task, most of the systems viewed this task as a 4-way single tweet classification task, with the exception of the best performing system by \cite{kochkina-etal-2017-turing}, as well as the systems by \cite{wang2017ecnu} and \cite{singh2017iitp}. The winning system addressed the task as a sequential classification problem, where the stance of each tweet takes into consideration the features and labels of the previous tweets. The system by \cite{singh2017iitp} takes as input pairs of source and reply tweets, whereas \cite{wang2017ecnu} addressed class imbalance by decomposing the problem into a two step classification task, first distinguishing between comments and non-comments, to then classify non-comment tweets as one of support, deny or query. Half of the systems employed ensemble classifiers, where classification was obtained through majority voting (\cite{wang2017ecnu,garcialozano2017mama,bahuleyan2017uwaterloo,srivastava2017dfki}). In some cases the ensembles were hybrid, consisting both of machine learning classifiers and manually created rules with differential weighting of classifiers for different class labels (\cite{wang2017ecnu,garcialozano2017mama,srivastava2017dfki}). Three systems used deep learning, with \cite{kochkina-etal-2017-turing} employing LSTMs for sequential classification, \cite{chen2017ikm} used convolutional neural networks (CNN) for obtaining the representation of each tweet, assigned a probability for a class by a softmax classifier and \cite{garcialozano2017mama} used CNN as one of the classifiers in their hybrid conglomeration. The remaining two systems by \cite{enayet2017niletmrg} and \cite{singh2017iitp} used support vector machines with a linear and polynomial kernel respectively. Half of the systems invested in elaborate feature engineering, including cue words and expressions denoting Belief, Knowledge, Doubt and Denial (\cite{bahuleyan2017uwaterloo}) as well as Tweet domain features, including meta-data about users, hashtags and event specific keywords (\cite{wang2017ecnu,bahuleyan2017uwaterloo,singh2017iitp,enayet2017niletmrg}). The systems with the least elaborate features were \cite{chen2017ikm} and \cite{garcialozano2017mama} for CNNs (word embeddings), \cite{srivastava2017dfki} (sparse word vectors as input to logistic regression) and \cite{kochkina-etal-2017-turing} (average word vectors, punctuation, similarity between word vectors in current tweet, source tweet and previous tweet, presence of negation, picture, URL). Five out of the eight systems used pre-trained word embeddings, mostly Google News word2vec embeddings\footnote{\url{https://github.com/mmihaltz/word2vec-GoogleNews-vectors}}, whereas \cite{wang2017ecnu} used four different types of embeddings. The winning system used a sequential classifier, however the rest of the participants opted for other alternatives.

To the best of our knowledge Twitter conversational thread structure has not been explored in detail in the stance classification problem. Here we extend the experimentation presented in our previous work using Conditional Random Fields for rumour stance classification (\cite{zubiaga2016coling}) in a number of directions: (1) we perform a comparison of a broader range of classifiers, including state-of-the-art rumour stance classifiers such as Hawkes Processes introduced by \cite{lukasik-etal-2016-hawkes}, as well as a new LSTM classifier, (2) we analyse the utility of a larger set of features, including not only local features as in our previous work, but also contextual features that further model the conversational structure of Twitter threads, (3) we perform a more exhaustive analysis of the results, and (4) we perform an analysis of features that gives insight into what characterises the different kinds of stances observed around rumours in social media.

\section{Research Objectives}

The main objective of our research is to analyse whether, the extent to which and how the sequential structure of social media conversations can be exploited to improve the classification of the stance expressed by different posts towards the topic under discussion. Each post in a conversation makes its own contribution to the discussion and hence has to be assigned its own stance value. However, posts in a conversation contribute to previous posts, adding up to a discussion attempting to reach a consensus. Our work looks into the exploitation of this evolving nature of social media discussions with the aim of improving the performance of a stance classifier that has to determine the stance of each tweet. We set forth the following six research objectives:

\textbf{RO 1.} \textit{Quantify performance gains of using sequential classifiers compared with the use of non-sequential classifiers.}

Our first research objective aims to analyse how the use of a sequential classifier that models the evolving nature of social media conversations can perform better than standard classifiers that treat each post in isolation. We do this by solely using local features to represent each post, so that the analysis focuses on the benefits of the sequential classifiers.

\textbf{RO 2.} \textit{Quantify the performance gains using contextual features extracted from the conversation.}

With our second research objective we are interested in analysing whether the use of contextual features (i.e. using other tweets surrounding in a conversation to extract the features of a given tweet) are helpful to boost the classification performance. This is particularly interesting in the case of tweets as they are very short, and inclusion of features extracted from surrounding tweets would be especially helpful. The use of contextual features is motivated by the fact that tweets in a discussion are adding to each other, and hence they cannot be treated alone.

\textbf{RO 3.} \textit{Evaluate the consistency of classifiers across different datasets.}

Our aim is to build a stance classifier that will generalise to multiple different datasets comprising data belonging to different events. To achieve this, we evaluate our classifiers on eight different events.

\textbf{RO 4.} \textit{Assess the effect of the depth of a post in its classification performance.}

We want to build a classifier that will be able to classify stances of different posts occurring at different levels of depth in a conversation. A post can be from a source tweet that initiates a conversation, to a nested reply that occurs later in the sequence formed by a conversational thread. The difficulty increases as replies are deeper as there is more preceding conversation to be aggregated for the classification task. We assess the performance over different depths to evaluate this.

\textbf{RO 5.} \textit{Perform an error analysis to assess when and why each classifier performs best.}

We want to look at the errors made by each of the classifiers. This will help us understand when we are doing well and why, as well as in what cases and with which types of labels we need to keep improving.

\textbf{RO 6.} \textit{Perform an analysis of features to understand and characterise stances in social media discussions.}

In our final objective we are interested in performing an exploration of different features under study, which is informative in two different ways. On the one hand, to find out which features are best for a stance classifier and hence improve performance; on the other hand, to help characterise the different types of stances and hence further understand how people respond in social media discussions.

\section{Rumour Stance Classification}

In what follows we formally define the rumour stance classification task, as well as the datasets we use for our experiments.

\subsection{Task Definition}

The rumour stance classification task consists in determining the type of orientation that each individual post expresses towards the disputed veracity of a rumour. We define the rumour stance classification task as follows: we have a set of conversational threads, each discussing a rumour, $D = \{C_1, ..., C_n\}$. Each conversational thread $C_j$ has a variably sized set of tweets $|C_j|$ discussing it, with a source tweet (the root of the tree), $t_{j,1}$, that initiates it. The source tweet $t_{j,1}$ can receive replies by a varying number $k$ of tweets $Replies_{t_{j,1}} = \{t_{j,1,1}, ..., t_{j,1,k}\}$, which can in turn receive replies by a varying number $k$ of tweets, e.g., $Replies_{t_{j,1,1}} = \{t_{j,1,1,1}, ..., t_{j,1,1,k}\}$, and so on. An example of a conversational thread is shown in Figure \ref{fig:example}.
The task consists in determining the stance of each of the tweets $t_j$ as one of $Y = \{supporting, denying, querying, commenting\}$.

\subsection{Dataset}

As part of the PHEME project (\cite{derczynski2015pheme}), we collected a rumour dataset associated with eight events corresponding to breaking news events (\cite{zubiaga2016analysing}).\footnote{The entire dataset included nine events, but here we describe the eight events with tweets in English, which we use for our classification experiments. The ninth dataset with tweets in German was not considered for this work.} Tweets in this dataset include tree-structured conversations, which are initiated by a tweet about a rumour (source tweet) and nested replies that further discuss the rumour circulated by the source tweet (replying tweets). The process of collecting the tree-structured conversations initiated by rumours, i.e. having a rumour discussed in the source tweet, and associated with the breaking news events under study was conducted with the assistance of journalist members of the Pheme project team. Tweets comprising the rumourous tree-structured conversations were then annotated for stance using CrowdFlower\footnote{\url{https://www.crowdflower.com/}} as a crowdsourcing platform. The annotation process is further detailed in \cite{zubiaga2015crowdsourcing}.


The resulting dataset includes 4,519 tweets and the transformations of annotations described above only affect 24 tweets (0.53\%), i.e., those where the source tweet denies a rumour, which is rare. The example in Figure \ref{fig:example} shows a rumour thread taken from the dataset along with our inferred annotations, as well as how we establish the depth value of each tweet in the thread.

\begin{figure*}
\fontsize{10}{10}\selectfont
 \begin{framed}
  \textit{[depth=0]} \noindent \textbf{u1:} These are not timid colours; soldiers back guarding Tomb of Unknown Soldier after today's shooting \#StandforCanada --PICTURE-- \textbf{[support]}
  \begin{addmargin}[2em]{0pt}
   \textit{[depth=1]} \textbf{u2:} @u1 Apparently a hoax. Best to take Tweet down. \textbf{[deny]}
  \end{addmargin}
  \begin{addmargin}[2em]{0pt}
   \textit{[depth=1]} \textbf{u3:} @u1 This photo was taken this morning, before the shooting. \textbf{[deny]}
  \end{addmargin}
  \begin{addmargin}[2em]{0pt}
   \textit{[depth=1]} \textbf{u4:} @u1 I don't believe there are soldiers guarding this area right now. \textbf{[deny]}
  \end{addmargin}
  \begin{addmargin}[4em]{0pt}
   \textit{[depth=2]} \textbf{u5:} @u4 wondered as well. I've reached out to someone who would know just to confirm that. Hopefully get response soon. \textbf{[comment]}
  \end{addmargin}
  \begin{addmargin}[6em]{0pt}
   \textit{[depth=3]} \textbf{u4:} @u5 ok, thanks. \textbf{[comment]}
  \end{addmargin}
 \end{framed}
 \caption{Example of a tree-structured thread discussing the veracity of a rumour, where the label associated with each tweet is the target of the rumour stance classification task.}
 \label{fig:example}
\end{figure*}

One important characteristic of the dataset, which affects the rumour stance classification task, is that the distribution of categories is clearly skewed towards \textit{commenting} tweets, which account for over 64\% of the tweets. This imbalance varies slightly across the eight events in the dataset (see Table \ref{tab:dataset-stats}). Given that we consider each event as a separate fold that is left out for testing, this varying imbalance makes the task more realistic and challenging. The striking imbalance towards \textit{commenting} tweets is also indicative of the increased difficulty with respect to previous work on stance classification, most of which performed binary classification of tweets as supporting or denying, which account for less than 28\% of the tweets in our case representing a real world scenario.

\begin{table}[htb]
\fontsize{10}{10}\selectfont
 \centering
 \begin{tabular}{ l c c c c c }
  \toprule
  Event & Supporting & Denying & Querying & Commenting & Total \\
  \midrule
  charliehebdo & 239 (22.0\%) & 58 (5.0\%) & 53 (4.0\%) & 721 (67.0\%) & 1,071 \\
  ebola-essien & 6 (17.0\%) & 6 (17.0\%) & 1 (2.0\%) & 21 (61.0\%) & 34 \\
  ferguson & 176 (16.0\%) & 91 (8.0\%) & 99 (9.0\%) & 718 (66.0\%) & 1,084 \\
  germanwings-crash & 69 (24.0\%) & 11 (3.0\%) & 28 (9.0\%) & 173 (61.0\%) & 281 \\
  ottawashooting & 161 (20.0\%) & 76 (9.0\%) & 63 (8.0\%) & 477 (61.0\%) & 777 \\
  prince-toronto & 21 (20.0\%) & 7 (6.0\%) & 11 (10.0\%) & 64 (62.0\%) & 103 \\
  putinmissing & 18 (29.0\%) & 6 (9.0\%) & 5 (8.0\%) & 33 (53.0\%) & 62 \\
  sydneysiege & 220 (19.0\%) & 89 (8.0\%) & 98 (8.0\%) & 700 (63.0\%) & 1,107 \\
  \midrule
  Total & 910 (20.1\%) & 344 (7.6\%) & 358 (7.9\%) & 2,907 (64.3\%) & 4,519 \\
  \bottomrule
 \end{tabular}
 \caption{Distribution of categories for the eight events in the dataset.}
 \label{tab:dataset-stats}
\end{table}

\section{Classifiers}

In this section we describe the different classifiers that we used for our experiments. Our focus is on sequential classifiers, especially looking at classifiers that exploit the discursive nature of social media, which is the case for Conditional Random Fields in two different settings -- i.e. Linear CRF and tree CRF -- as well as that of a Long Short-Term Memory (LSTM) in a linear setting -- Branch LSTM. We also experiment with a sequential classifier based on Hawkes Processes that instead exploits the temporal sequence of tweets and has been shown to achieve state-of-the-art performance (\cite{lukasik-etal-2016-hawkes}). After describing these three types of classifiers, we outline a set of baseline classifiers.

\subsection{Hawkes Processes}

One approach for modelling arrival of tweets around rumours is based on point processes, a probabilistic framework where tweet occurrence likelihood is modelled using an intensity function over time. Intuitively, higher values of intensity function denote higher likelihood of tweet occurrence. For example, \cite{lukasik-etal-2016-hawkes} modelled tweet occurrences over time with a log-Gaussian Cox Process, a point process which models its intensity function as an exponentiated sample of a Gaussian Process (\cite{lukasik15_dynamics,lukasik15_tweetarrival,lukasik16_conv}). In related work, tweet arrivals were modelled with a Hawkes Process and a resulting model was applied for stance classification of tweets around rumours (\cite{lukasik-etal-2016-hawkes}). In this subsection we describe the sequence classification algorithm based on Hawkes Processes.

\paragraph{Intensity Function}
The intensity function in a Hawkes Process is expressed as a summation of base intensity and the intensities corresponding to influences of previous tweets,
\begin{flalign}
\label{eq:HawkesCI2}
\lambda_{y,m}(t)\!\! =\!\! \displaystyle \mu_y \! +\!\! \sum_{t_\ell < t} \mathbb{I}(m_\ell = m)
\alpha_{y_\ell,y} \kappa(t - t_\ell),
\end{flalign}
where the first term  represents the constant base intensity of generating label $y$. The second term represents the influence from the previous tweets. The influence from each tweet is modelled with an exponential kernel function $\kappa(t - t_\ell) = \omega \exp(-\omega (t - t_\ell))$. 
The matrix $\alpha$ of size $|Y| \times |Y|$ encodes how pairs of labels corresponding to tweets influence one another, e.g. how a \emph{querying} label influences a \emph{rejecting} label.

\paragraph{Likelihood function}
The parameters governing the intensity function are learnt by maximising the likelihood of generating the tweets:

\begin{flalign}
\label{eq:factorizedCL}
L(\bm{t}, \bm{y}, \bm{m}, \bm{W})  = \prod_{n=1}^N  p(\bm{W}_n | y_n)  \times \Big [ \prod_{n=1}^N \lambda_{y_n,m_n}(t_n) \Big ] \! \times \! p(E_T),
\end{flalign}

where the likelihood of generating text given the label is modelled as a multinomial distribution conditioned on the label (parametrised by matrix $\beta$). The second term provides the likelihood of occurrence of tweets at times $t_1, \ldots , t_n$ and the third term provides the likelihood that no tweets happen in the interval $[0,T]$ except at times $t_1, \ldots, t_n$. We estimate the parameters of the model by maximising the log-likelihood. As in \cite{lukasik-etal-2016-hawkes}, Laplacian smoothing is applied to the estimated language parameter $\beta$.

In one approach to $\mu$ and $\alpha$ optimisation (\textit{Hawkes Process with Approximated Likelihood}, or \emph{HP Approx.} \cite{lukasik-etal-2016-hawkes}) a closed form updates for $\mu$ and $\alpha$ are obtained using an approximation of the log-likelihood of the data. In a different approach (\textit{Hawkes Process with Exact Likelihood}, or \emph{HP Grad.} \cite{lukasik-etal-2016-hawkes}) parameters are found using joint gradient based optimisation over $\mu$ and $\alpha$, using derivatives of log-likelihood\footnote{For both implementations we used the `seqhawkes' Python package: \url{https://github.com/mlukasik/seqhawkes}}. L-BFGS approach is employed for gradient search. Parameters are initialised with those found by the \emph{HP Approx.} method. Moreover, following previous work we fix the decay parameter $\omega$ to $0.1$.

We predict the most likely sequence of labels, thus maximising the likelihood of occurrence of the tweets from Equation (\ref{eq:factorizedCL}), or the approximated likelihood in case of \emph{HP Approx.} Similarly as in \cite{lukasik-etal-2016-hawkes}, we follow a greedy approach, where we choose the most likely label for each consecutive tweet. 

\subsection{Conditional Random Fields (CRF): Linear CRF and Tree CRF}

We use CRF as a structured classifier to model sequences observed in Twitter conversations. With CRF, we can model the conversation as a graph that will be treated as a sequence of stances, which also enables us to assess the utility of harnessing the conversational structure for stance classification. Different to traditionally used classifiers for this task, which choose a label for each input unit (e.g. a tweet), CRF also consider the neighbours of each unit, learning the probabilities of transitions of label pairs to be followed by each other. The input for CRF is a graph $G = (V, E)$, where in our case each of the vertices $V$ is a tweet, and the edges $E$ are relations of tweets replying to each other. Hence, having a data sequence $X$ as input, CRF outputs a sequence of labels $Y$ (\cite{lafferty2001conditional}), where the output of each element $y_i$ will not only depend on its features, but also on the probabilities of other labels surrounding it. The generalisable conditional distribution of CRF is shown in Equation \ref{eq:crf} (\cite{sutton2011introduction}).

\begin{equation}
 p(y|x) = \frac{1}{Z(x)} \prod_{a = 1}^{A} \Psi_a (y_a, x_a)
 \label{eq:crf}
\end{equation}

where Z(x) is the normalisation constant, and $\Psi_a$ is the set of factors in the graph $G$.

\begin{figure}
 \centering
 \includegraphics[width=0.8\textwidth]{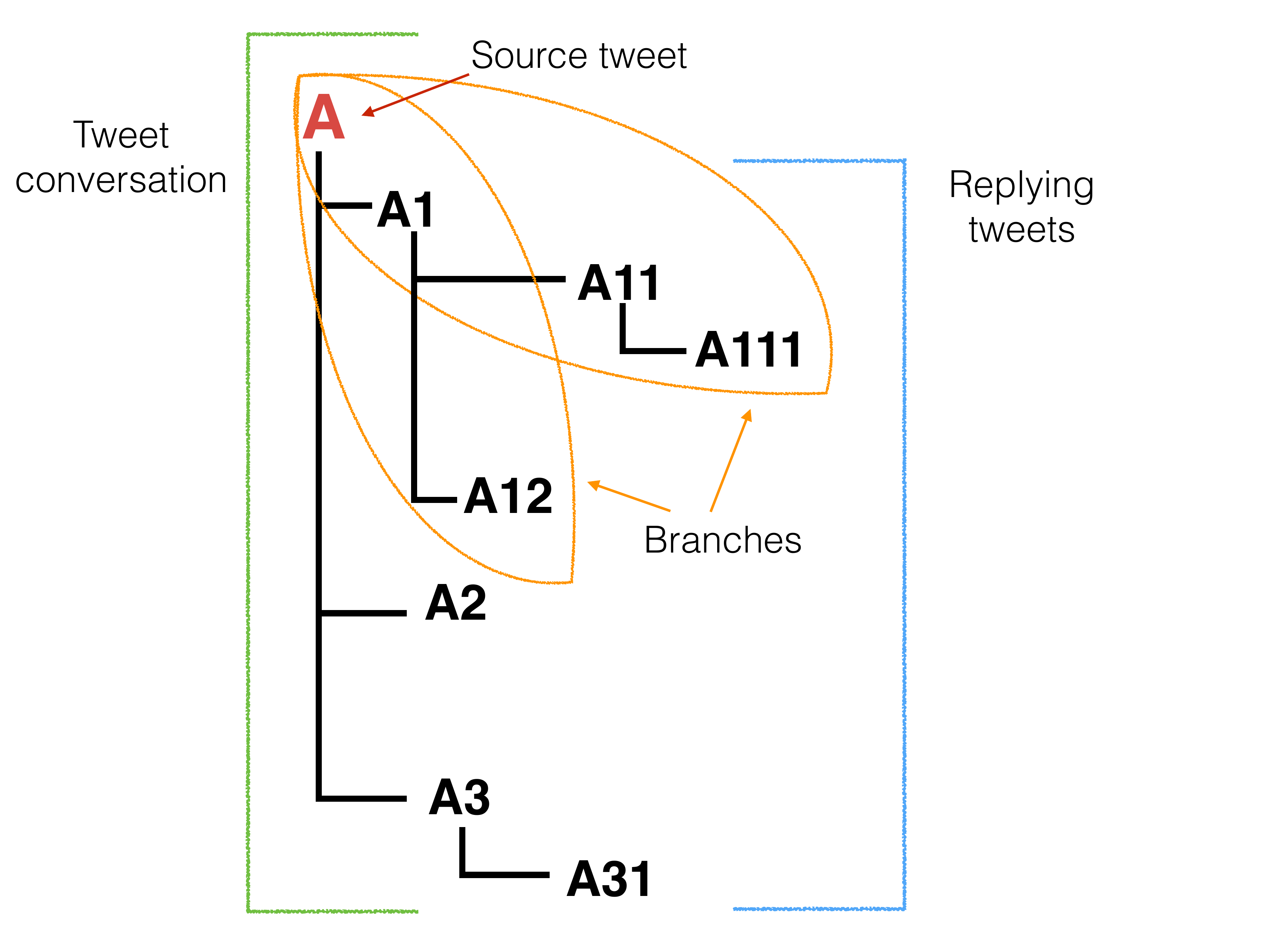}
 \caption{Example of a tree-structured conversation, with two overlapping branches highlighted.}
 \label{fig:tree-and-branches}
\end{figure}

We use CRFs in two different settings.\footnote{We use the PyStruct to implement both variants of CRF \cite{muller2014pystruct}.} First, we use a linear-chain CRF (Linear CRF) to model each branch as a sequence to be input to the classifier. We also use Tree-Structured CRFs (Tree CRF) or General CRFs to model the whole, tree-structured conversation as the sequence input to the classifier. So in the first case the sequence unit is a branch and our input is a collection of branches and in the second case our sequence unit is an entire conversation, and our input is a collection of trees. An example of the distinction of dealing with branches or trees is shown in Figure \ref{fig:tree-and-branches}. With this distinction we also want to experiment whether it is worthwhile building the whole tree as a more complex graph, given that users replying in one branch might not have necessarily seen and be conditioned by tweets in other branches. However, we believe that the tendency of types of replies observed in a branch might also be indicative of the distribution of types of replies in other branches, and hence useful to boost the performance of the classifier when using the tree as a whole. An important caveat of modelling a tree in branches is also that there is a need to repeat parts of the tree across branches, e.g., the source tweet will repeatedly occur as the first tweet in every branch extracted from a tree.\footnote{Despite this also leading to having tweets repeated across branches in the test set and hence producing an output repeatedly for the same tweet with Linear CRF, this output does is consistent and there is no need to aggregate different outputs.}

To account for the imbalance of classes in our datasets, we perform cost-sensitive learning by assigning weighted probabilities to each of the classes, these probabilities being the inverse of the number of occurrences observed in the training data for a class.

\subsection{Branch LSTM}

\begin{figure}
 \centering
 \includegraphics[width=0.8\textwidth]{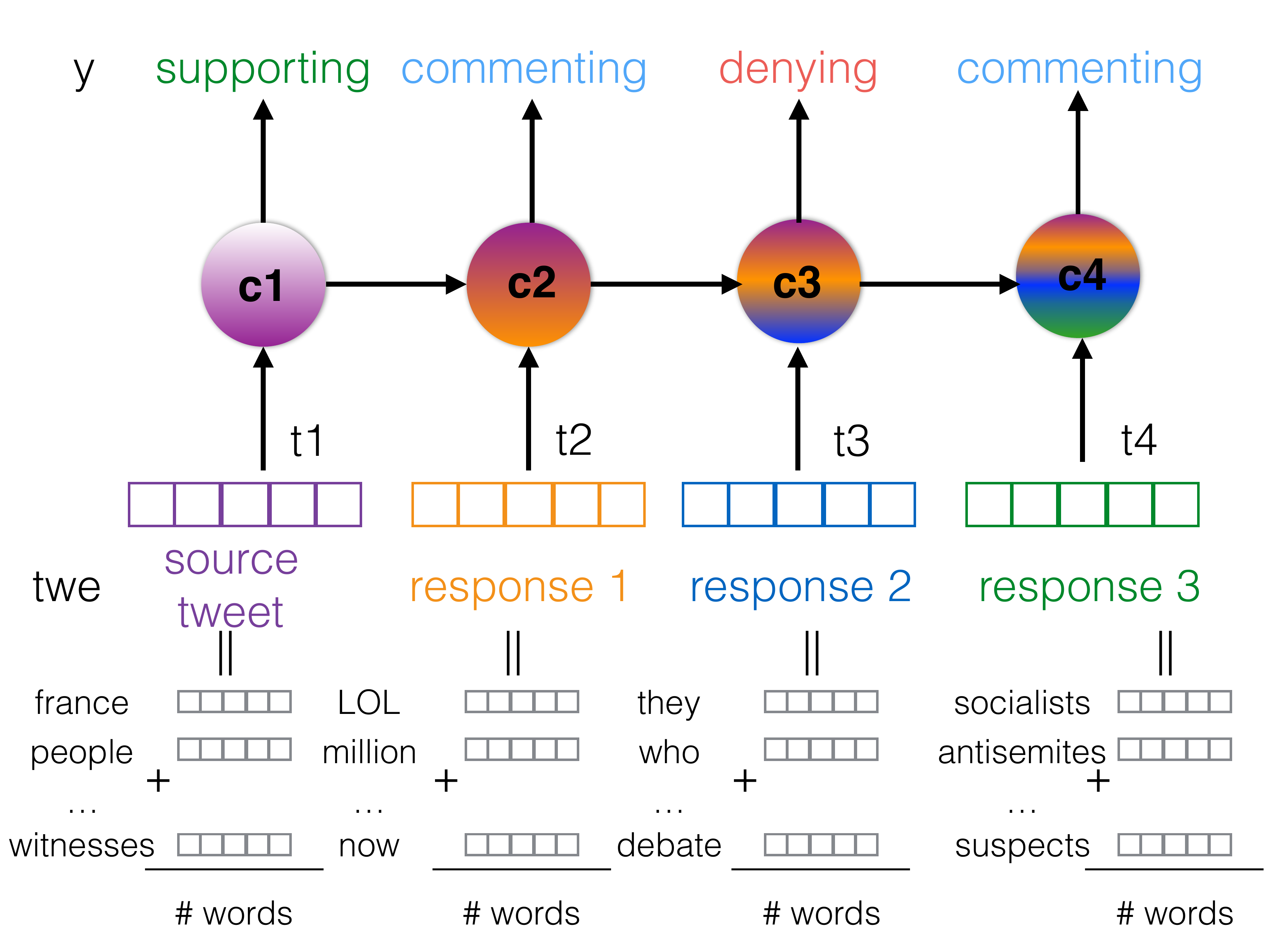}
 \caption{Illustration of the input/output structure of the LSTM-branch model}
 \label{fig:LSTMbranchio}
\end{figure}

Another model that works with structured input is a neural network with Long Short-Term Memory (LSTM) units (\cite{hochreiter1997long}). LSTMs are able to model discrete time series and possess a `memory' property of the previous time steps, therefore we propose a \textit{branch-LSTM} model that utilises them to process branches of tweets.

Figure \ref{fig:LSTMbranchio} illustrates how the input of the time step of the LSTM layer is a vector that is an average of word vectors from each tweet and how the information propagates between time steps. 

The full model consists of several LSTM layers that are connected to several feed-forward ReLU layers and a softmax layer to obtain predicted probabilities of a tweet belonging to certain class. As a means for weight regularisation we utilise \textit{dropout} and \textit{l2-norm}. We use categorical cross-entropy as the loss function. The model is trained using mini-batches and the Adam optimisation algorithm (\cite{kingma2014adam}).\footnote{For implementation of all models we used Python libraries Theano (\cite{bastien2012theano}) and Lasagne (\cite{lasagne}).}

The number of layers, number of units in each layer, regularisation strength, mini-batch size and learning rate are determined using the Tree of Parzen Estimators (TPE) algorithm (\cite{bergstra2011algorithms})\footnote{We use the implementation in the hyperopt package (\cite{bergstra2013making}).} on the development set.\footnote{For this setting, we use the `Ottawa shooting' event for development.} 

The \textit{branch-LSTM} takes as input tweets represented as the average of its word vectors. We also experimented with obtaining tweet representations through per-word nested LSTM layers, however, this approach did not result in significantly better results than the average of word vectors.

Extracting branches from a tree-structured conversation presents the caveat that some tweets are repeated across branches after this conversion. We solve this issue by applying a mask to the loss function to not take repeated tweets into account.

\subsection{Summary of Sequential Classifiers}

All of the classifiers described above make use of the sequential nature of Twitter conversational threads. These classifiers take a sequence of tweets as input, where the relations between tweets are formed by replies. If C replies to B, and B replies to A, it will lead to a sequence ``A $\rightarrow$ B $\rightarrow$ C''. Sequential classifiers will use the predictions on preceding tweets to determine the possible label for each tweet. For instance, the classification for B will depend on the prediction that has been previously made for A, and the probabilities of different labels for B will vary for the classifier depending on what has been predicted for A.

Among the four classifiers described above, the one that differs in how the sequence is treated is the Tree CRF. This classifier builds a tree-structured graph with the sequential relationships composed by replying tweets. The rest of the classifiers, Hawkes Processes, Linear CRF and LSTM, will break the entire conversational tree into linear branches, and the input to the classifiers will be linear sequences. The use of a graph with the Tree CRF has the advantage of building a single structure, while the rest of the classifiers building linear sequences inevitably need to repeat tweets across different linear sequences. All the linear sequences will repeatedly start with the source tweet, while some of the subsequent tweets may also be repeated. The use of linear sequences has however the advantages of simplifying the model being used, and one may also hypothesise that inclusion of the entire tree made of different branches into the same graph may not be suitable when they may all be discussing issues that differ to some extent from one another. Figure \ref{fig:tree-and-branches} shows an example of a conversation tree, how the entire tree would make a graph, as well as how we break it down into smaller branches or linear sequences.

\subsection{Baseline Classifiers}

\noindent \textbf{Maximum Entropy classifier (MaxEnt).} As the non-sequential counterpart of CRF, we use a Maximum Entropy (or logistic regression) classifier, which is also a conditional classifier but which will operate at the tweet level, ignoring the conversational structure. This enables us to directly compare the extent to which treating conversations as sequences instead of having each tweet as a separate unit can boost the performance of the CRF classifiers. We perform cost-sensitive learning by assigning weighted probabilities to each class as the inverse of the number of occurrences in the training data.

\noindent \textbf{Additional baselines.} We also compare two more non-sequential classifiers\footnote{We use their implementation in the scikit-learn Python package, using the \textit{class\_weight=``balanced''} parameter to perform cost-sensitive learning.}: Support Vector Machines (SVM), and Random Forests (RF).

\subsection{Experiment Settings and Evaluation Measures}

We experiment in an 8-fold cross-validation setting. Given that we have 8 different events in our dataset, we create 8 different folds, each having the data linked to an event. In our cross-validation setting, we run the classifier 8 times, on each occasion having a different fold for testing, with the other 7 for training. In this way, each fold is tested once, and the aggregation of all folds enables experimentation on all events. For each of the events in the test set, the experiments consist in classifying the stance of each individual tweet. With this, we simulate a realistic scenario where we need to use knowledge from past events to train a model that will be used to classify tweets in new events.

Given that the classes are clearly imbalanced in our case, evaluation based on accuracy arguably cannot suffice to capture competitive performance beyond the majority class. To account for the imbalance of the categories, we report the macro-averaged F1 scores, which measures the overall performance assigning the same weight to each category. We aggregate the macro-averaged F1 scores to get the final performance score of a classifier. We also use the McNemar test (\cite{mcnemar1947note}) throughout the analysis of results to further compare the performance of some classifiers.

It is also worth noting that all the sequential classifiers only make use of preceding tweets in the conversation to classify a tweet, and hence no later tweets are used. That is the case of a sequence \textit{$t_1$, $t_2$, $t_3$} of tweets, each responding to the preceding tweet. The sequential classifier attempting to classify $t_2$ would incorporate $t_1$ in the sequence, but $t_3$ would not be considered.

\section{Features}

While focusing on the study of sequential classifiers for discursive stance classification, we perform our experiments with three different types of features: local features, contextual features and Hawkes features. First, local features enable us to evaluate the performance of sequential classifiers in a comparable setting to non-sequential classifiers where features are extracted solely from the current tweet; this makes it a fairer comparison where we can quantify the extent to which mining sequences can boost performance. In a subsequent step, we also incorporate contextual features, i.e. features from other tweets in a conversation, which enables us to further boost performance of the sequential classifiers. Finally, and to enable comparison with the Hawkes process classifier, we describe the Hawkes features.

Table \ref{tab:features} shows the list of features used, both local and contextual, each of which can be categorised into several subtypes of features, as well as the Hawkes features. For more details on these features, please see \ref{ap:features}.


\begin{table}
\fontsize{10}{10}\selectfont
 \centering
 \begin{tabular*}{0.66\textwidth}{l | l}
  \toprule
  \multicolumn{2}{c}{\textbf{Local features}} \\
  \toprule
  \multirow{4}{*}{\textbf{Lexicon}} & Word embeddings \\
   & POS tags \\
   & Negation \\
   & Swear words \\
  \midrule
  \multirow{2}{*}{\textbf{Content formatting}} & Tweet length \\
   & Word count \\
  \midrule
  \multirow{2}{*}{\textbf{Punctuation}} & Question mark \\
   & Exclamation mark \\
  \midrule
  \multirow{1}{*}{\textbf{Tweet formatting}} & URL attached \\
  \toprule
  \multicolumn{2}{c}{\textbf{Contextual features}} \\
  \toprule
  \multirow{3}{*}{\textbf{Relational}} & Word2Vec similarity wrt source tweet \\
   & Word2Vec similarity wrt preceding tweet \\
   & Word2Vec similarity wrt thread \\
  \midrule
  \multirow{4}{*}{\textbf{Structural}} & Is leaf \\
   & Is source tweet \\
   & Is source user \\
  \midrule
  \multirow{3}{*}{\textbf{Social}} & Has favourites \\
   & Has retweets \\
   & Persistence \\
   & Time difference \\
  \toprule
  \multicolumn{2}{c}{\textbf{Hawkes features}} \\
  \toprule
  \toprule
  \multirow{2}{*}{\textbf{Hawkes features}} & Bag of words \\
   & Timestamp \\
  \bottomrule
 \end{tabular*}
 \caption{List of features.}
 \label{tab:features}
\end{table}


\section{Experimental Results}
\label{sec:experimental-results}

\subsection{Evaluating Sequential Classifiers (RO 1)}
\label{sec:eval-sequential}

First, we evaluate the performance of the classifiers by using only local features. As noted above, this enables us to perform a fairer comparison of the different classifiers by using features that can be obtained solely from each tweet in isolation; likewise, it enables us to assess whether and the extent to which the use of a sequential classifier to exploit the discursive structure of conversational threads can be of help to boost performance of the stance classifier while using the same set of features as non-sequential classifiers.

Therefore, in this section we make use of the local features described in Section \ref{ssec:local-features}. Additionally, we also use the Hawkes features described in Section \ref{ssec:hawkes-features} for comparison with the Hawkes processes. For the set of local features, we show the results for three different scenarios: (1) using each subgroup of features alone, (2) in a leave-one-out setting where one of the subgroups is not used, and (3) using all of the subgroups combined.

Table \ref{tab:results-sequence} shows the results for the different classifiers using the combinations of local features as well as Hawkes features. We make the following observations from these results:

\begin{itemize}
 \item LSTM consistently performs very well with different features.
 \item Confirming our main hypothesis and objective, sequential classifiers do show an overall superior performance to the non-sequential classifiers. While the two CRF alternatives perform very well, the LSTM classifier is slightly superior (the differences between CRF and LSTM results are statistically significant at $p < 0.05$, except for the LF1 features). Moreover, the CRF classifiers outperform their non-sequential counterpart MaxEnt, which achieves an overall lower performance (all the differences between CRF and MaxEnt results being statistically significant at $p < 0.05$).
 \item The LSTM classifier is, in fact, superior to the Tree CRF classifier (all statistically significant except LF1). While the Tree CRF needs to make use of the entire tree for the classification, the LSTM classifier only uses branches, reducing the amount of data and complexity that needs to be processed in each sequence.
 \item Among the local features, combinations of subgroups of features lead to clear improvements with respect to single subgroups without combinations.
 \item Even though the combination of all local features achieves good performance, there are alternative leave-one-out combinations that perform better. The feature combination leading to the best macro-F1 score is that combining lexicon, content formatting and punctuation (i.e. LF123, achieving a score of 0.449).
\end{itemize}

Summarising, our initial results show that exploiting the sequential properties of conversational threads, while still using only local features to enable comparison, leads to superior performance with respect to the classification of each tweet in isolation by non-sequential classifiers. Moreover, we observe that the local features combining lexicon, content formatting and punctuation lead to the most accurate results. In the next section we further explore the use of contextual features in combination with local features to boost performance of sequential classifiers; to represent the local features, we rely on the best approach from this section (i.e. LF123).


\begin{table}
\fontsize{10}{10}\selectfont
  \centering
  \begin{tabular}{l || l l l l l l l l l l }
   \toprule
   \multicolumn{11}{c}{\textbf{Macro-F1}} \\
   \toprule
    & HF & LF1 & LF2 & LF3 & LF4 & LF123 & LF124 & LF134 & LF234 & LF1234 \\
   \midrule
   SVM & 0.336 & 0.356 & 0.231 & 0.258 & 0.313 & 0.403 & 0.365 & 0.403 & 0.420 & 0.408 \\
   Random Forest & 0.325 & 0.308 & 0.276 & 0.267 & \textbf{0.437*} & 0.322 & 0.310 & 0.351 & 0.357 & 0.329 \\
   MaxEnt & 0.338 & 0.363 & 0.272 & 0.263 & 0.428 & 0.415 & 0.363 & 0.421 & 0.427 & 0.422 \\
   \midrule
   Hawkes-approx & 0.309 & -- & -- & -- & -- & -- & -- & -- & -- & -- \\
   Hawkes-grad & 0.307 & -- & -- & -- & -- & -- & -- & -- & -- & -- \\
   Linear CRF & \textbf{0.362*} & 0.357 & 0.268 & 0.318 & 0.317 & 0.413 & 0.365 & 0.403 & 0.425 & 0.412 \\
   Tree CRF & 0.350 & \textbf{0.375*} & 0.285 & 0.221 & 0.217 & 0.433 & 0.385 & \textbf{0.413} & \textbf{0.436*} & 0.433 \\
   LSTM & 0.318 & 0.362 & \textbf{0.318*} & \textbf{0.407*} & 0.419 & \textbf{0.449*} & \textbf{0.395*} & 0.412 & 0.429 & \textbf{0.437*} \\
   \bottomrule
  \end{tabular}
  \caption{Macro-F1 performance results using local features. HF: Hawkes features. LF: local features, where numbers indicate subgroups of features as follows, 1: Lexicon, 2: Content formatting, 3: Punctuation, 4: Tweet formatting. An '*' indicates that the differences between the best performing classifier and the second best classifier for that feature set are statistically significant at $p < 0.05$.}
  \label{tab:results-sequence}
\end{table}


\subsection{Exploring Contextual Features (RO 2)}
\label{sec:eval-contextual}

The experiments in the previous section show that sequential classifiers that model discourse, especially the LSTM classifier, can provide substantial improvements over non-sequential classifiers that classify each tweet in isolation, in both cases using only local features to represent each tweet. To complement this, we now explore the inclusion of contextual features described in Section \ref{ssec:contextual-features} for the stance classification. We perform experiments with four different groups of features in this case, including local features and the three subgroups of contextual features, namely relational features, structural features and social features. As in the previous section, we show results for the use of each subgroup of features alone, in a leave-one-out setting, and using all subgroups of features together.

Table \ref{tab:results-contextual} shows the results for the classifiers incorporating contextual features along with local features. We make the following observations from these results:

\begin{itemize}
 \item The use of contextual features leads to substantial improvements for non-sequential classifiers, getting closer to and even in some cases outperforming some of the sequential classifiers.
 \item Sequential classifiers, however, do not benefit much from using contextual features. It is important to note that sequential classifiers are taking the surrounding context into consideration when they aggregate sequences in the classification process. This shows that the inclusion of contextual features is not needed for sequential classifiers, given that they are implicitly including context through the use of sequences.
 \item In fact, for the LSTM, which is still the best-performing classifier, it is better to only rely on local features, as the rest of the features do not lead to any improvements. Again, the LSTM is able to handle context on its own, and therefore inclusion of contextual features is redundant and may be harmful.
 \item Addition of contextual features leads to substantial improvements for the non-sequential classifiers, achieving similar macro-averaged scores in some cases (e.g. MaxEnt / All vs LSTM / LF). This reinforces the importance of incorporating context in the classification process, which leads to improvements for the non-sequential classifier when contextual features are added, but especially in the case of sequential classifiers that can natively handle context.
\end{itemize}


\begin{table}
\fontsize{10}{10}\selectfont
  \centering
  \begin{tabular}{l || l l l l l l l l l }
   \toprule
   \multicolumn{10}{c}{\textbf{Macro-F1}} \\
   \toprule
    & LF & R & ST & SO & LF+R+ST & LF+R+SO & LF+ST+SO & R+ST+SO & All \\
   \midrule
   SVM & 0.403 & \textbf{0.335*} & \textbf{0.318} & 0.260 & 0.429 & 0.347 & 0.388 & 0.295 & 0.375 \\
   Random Forest & 0.322 & 0.325 & 0.269 & 0.328 & 0.356 & 0.358 & 0.376 & \textbf{0.343*} & 0.364 \\
   MaxEnt & 0.415 & 0.333 & \textbf{0.318} & 0.310 & 0.434 & \textbf{0.447} & 0.447 & 0.318 & \textbf{0.449} \\
   \midrule
   Linear CRF & 0.413 & 0.318 & \textbf{0.318} & \textbf{0.334*} & 0.424 & 0.431 & 0.431 & 0.342 & 0.437 \\
   Tree CRF & 0.433 & 0.322 & 0.317 & 0.312 & 0.425 & 0.429 & 0.430 & 0.232 & 0.433 \\
   LSTM & \textbf{0.449*} & 0.318 & \textbf{0.318} & 0.315 & \textbf{0.445*} & 0.436 & \textbf{0.448} & 0.314 & 0.437 \\
   \bottomrule
  \end{tabular}
  \caption{Macro-F1 performance results incorporating contextual features. LF: local features, R: relational features, ST: structural features, SO: social features. An '*' indicates that the differences between the best performing classifier and the second best classifier for that feature set are statistically significant.}
  \label{tab:results-contextual}
\end{table}


Summarising, we observe that the addition of contextual features is clearly useful for non-sequential classifiers, which do not consider context natively. For the sequential classifiers, which natively consider context in the classification process, the inclusion of contextual features is not helpful and is even harmful in most cases, potentially owing to the contextual information being used twice. Still, sequential classifiers, and especially LSTM, are the best classifiers to achieve optimal results, which also avoid the need for computing contextual features.

\subsection{Analysis of the Best-Performing Classifiers}

Despite the clear superiority of LSTM with the sole use of local features, we now further examine the results of the best-performing classifiers to understand when they perform well. We compare the performance of the following five classifiers in this section: (1) LSTM with only local features, (2) Tree CRF with all the features, (3) Linear CRF with all the features, (4) MaxEnt with all the features, and (5) SVM using local features, relational and structural features. Note that while for LSTM we only need local features, for the rest of the classifiers we need to rely on all or almost all of the features. For these best-performing combinations of classifiers and features, we perform additional analyses by event and by tweet depth, and perform an analysis of errors.

\subsubsection{Evaluation by Event (RO 3)}

The analysis of the best-performing classifiers, broken down by event, is shown in Table \ref{tab:results-events}. These results suggest that there is not a single classifier that performs best in all cases; this is most likely due to the diversity of events. However, we see that the LSTM is the classifier that outperforms the rest in the greater number of cases; this is true for three out of the eight cases (the difference with respect to the second best classifier being always statistically significant). Moreover, sequential classifiers perform best in the majority of the cases, with only three cases where a non-sequential classifier performs best. Most importantly, these results suggest that sequential classifiers outperform non-sequential classifiers across the different events under study, with LSTM standing out as a classifier that performs best in numerous cases using only local features.


\begin{table}
\fontsize{10}{10}\selectfont
  \centering
  \begin{tabular}{l || l l l l l l l l }
   \toprule
   \multicolumn{9}{c}{\textbf{Macro-F1}} \\
   \toprule
   & CH & Ebola & Ferg. & GW crash & Ottawa & Prince & Putin & Sydney \\
   \midrule
   SVM & 0.399 & 0.380 & 0.382 & 0.427 & \textbf{0.492} & 0.491 & 0.509 & 0.427 \\
   MaxEnt & 0.446 & 0.425 & \textbf{0.418} & 0.475 & 0.468 & \textbf{0.514} & 0.381 & 0.443 \\
   \midrule
   Linear CRF & 0.443 & 0.619 & 0.380 & 0.470 & 0.412 & 0.512 & \textbf{0.528} & \textbf{0.454} \\
   Tree CRF & 0.457 & 0.557 & 0.356 & 0.523 & 0.441 & 0.505 & 0.491 & 0.426 \\
   LSTM & \textbf{0.465} & \textbf{0.657} & 0.373 & \textbf{0.543} & 0.475 & 0.379 & 0.457 & 0.446 \\
   \bottomrule
  \end{tabular}
  \caption{Macro-F1 results for the best-performing classifiers, broken down by event.}
  \label{tab:results-events}
\end{table}


\subsubsection{Evaluation by Tweet Depth (RO 4)}

The analysis of the best-performing classifiers, broken down by depth of tweets, is shown in Table \ref{tab:results-depth}. Note that the depth of the tweet reflects, as shown in Figure \ref{fig:example}, the number of steps from the source tweet to the current tweet. We show results for all the depths from 0 to 4, as well as for the subsequent depths aggregated as 5+.

Again, we see that there is not a single classifier that performs best for all depths. We see, however, that sequential classifiers (Linear CRF, Tree CRF and LSTM) outperform non-sequential classifiers (SVM and MaxEnt) consistently. However, the best sequential classifier varies. While LSTM is the best-performing classifier overall when we look at macro-averaged F1 scores, as shown in Section \ref{sec:eval-contextual}, surprisingly it does not achieve the highest macro-averaged F1 scores at any depth. It does, however, perform well for each depth compared to the rest of the classifiers, generally being close to the best classifier in that case. Its consistently good performance across different depths makes it the best overall classifier, despite only using local features.


\begin{table}
\fontsize{10}{10}\selectfont
  \centering
  \begin{tabular}{l || l l l l l l }
   \toprule
   \multicolumn{7}{c}{\textbf{Tweets by depth}} \\
   \toprule
   & 0 & 1 & 2 & 3 & 4 & 5+ \\
   \midrule
   Counts & 297 & 2,602 & 553 & 313 & 195 & 595 \\
   \toprule
   \multicolumn{7}{c}{\textbf{Macro-F1}} \\
   \toprule
   & 0 & 1 & 2 & 3 & 4 & 5+ \\
   \midrule
   SVM & 0.272 & 0.368 & 0.298 & 0.314 & 0.331 & 0.274 \\
   MaxEnt & 0.238 & 0.385 & 0.286 & 0.279 & \textbf{0.369} & \textbf{0.290} \\
   \midrule
   Linear CRF & \textbf{0.286} & 0.394 & \textbf{0.306} & 0.282 & 0.271 & 0.266 \\
   Tree CRF & 0.278 & \textbf{0.404} & 0.280 & \textbf{0.331} & 0.230 & 0.237 \\
   LSTM & 0.271 & 0.381 & 0.298 & 0.274 & 0.307 & 0.286 \\
   \bottomrule
  \end{tabular}
  \caption{Macro-F1 results for the best-performing classifiers, broken down by tweet depth.}
  \label{tab:results-depth}
\end{table}


\subsubsection{Error Analysis (RO 5)}

To analyse the errors that the different classifiers are making, we look at the confusion matrices in Table \ref{tab:confusion}. If we look at the correct guesses, highlighted in bold in the diagonals, we see that the LSTM clearly performs best for three of the categories, namely \textit{support}, \textit{deny} and \textit{query}, and it is just slightly behind the other classifiers for the majority class, \textit{comment}. Besides LSTM's overall superior performance as we observed above, this also confirms that the LSTM is doing better than the rest of the classifiers in dealing with the imbalance inherent in our datasets. For instance, the \textit{Deny} category proves especially challenging for being less common than the rest (only 7.6\% of instances in our datasets); the LSTM still achieves the highest performance for this category, which, however, only achieves 0.212 in accuracy and may benefit from having more training instances.

We also notice that a large number of instances are misclassified as \textit{comments}, due to this being the prevailing category and hence having a much larger number of training instances. One could think of balancing the training instances to reduce the prevalence of \textit{comments} in the training set, however, this is not straightforward for sequential classifiers as one needs to then break sequences, losing not only some instances of \textit{comments}, but also connections between instances of other categories that belong to those sequences. Other solutions, such as labelling more data or using more sophisticated features to distinguish different categories, might be needed to deal with this issue; given that the scope of this paper is to assess whether and the extent to which sequential classifiers can be of help in stance classification, further tackling this imbalance is left for future work.


\begin{table}
\fontsize{10}{10}\selectfont
  \centering
  \begin{tabular}{l || l l l l }
   \toprule
   \multicolumn{5}{c}{\textbf{SVM}} \\
   \toprule
   & Support & Deny & Query & Comment \\
   \midrule
   Support & \textbf{0.657} & 0.041 & 0.018 & 0.283 \\
   Deny & 0.185 & \textbf{0.129} & 0.107 & 0.579 \\
   Query & 0.083 & 0.081 & \textbf{0.343} & 0.494 \\
   Comment & 0.150 & 0.075 & 0.053 & \textbf{0.723} \\
   \toprule
   \multicolumn{5}{c}{\textbf{MaxEnt}} \\
   \toprule
   & Support & Deny & Query & Comment \\
   \midrule
   Support & \textbf{0.794} & 0.044 & 0.003 & 0.159 \\
   Deny & 0.156 & \textbf{0.130} & 0.079 & 0.634 \\
   Query & 0.088 & 0.066 & \textbf{0.366} & 0.480 \\
   Comment & 0.152 & 0.074 & 0.048 & \textbf{0.726} \\
   \toprule
   \multicolumn{5}{c}{\textbf{Linear CRF}} \\
   \toprule
   & Support & Deny & Query & Comment \\
   \midrule
   Support & \textbf{0.603} & 0.048 & 0.013 & 0.335 \\
   Deny & 0.219 & \textbf{0.140} & 0.050 & 0.591 \\
   Query & 0.071 & 0.095 & \textbf{0.357} & 0.476 \\
   Comment & 0.139 & 0.072 & 0.062 & \textbf{0.726} \\
   \toprule
   \multicolumn{5}{c}{\textbf{Tree CRF}} \\
   \toprule
   & Support & Deny & Query & Comment \\
   \midrule
   Support & \textbf{0.552} & 0.066 & 0.019 & 0.363 \\
   Deny & 0.145 & \textbf{0.169} & 0.081 & 0.605 \\
   Query & 0.077 & 0.081 & \textbf{0.401} & 0.441 \\
   Comment & 0.128 & 0.074 & 0.068 & \textbf{0.730} \\
   \toprule
   \multicolumn{5}{c}{\textbf{LSTM}} \\
   \toprule
   & Support & Deny & Query & Comment \\
   \midrule
   Support & \textbf{0.825} & 0.046 & 0.003 & 0.127 \\
   Deny & 0.225 & \textbf{0.212} & 0.125 & 0.438 \\
   Query & 0.090 & 0.087 & \textbf{0.432} & 0.390 \\
   Comment & 0.144 & 0.076 & 0.057 & \textbf{0.723} \\
   \bottomrule
  \end{tabular}
  \caption{Confusion matrices for the best-performing classifiers.}
  \label{tab:confusion}
\end{table}


\subsection{Feature Analysis (RO 6)}

\begin{figure}
 \centering
 \includegraphics[width=\textwidth]{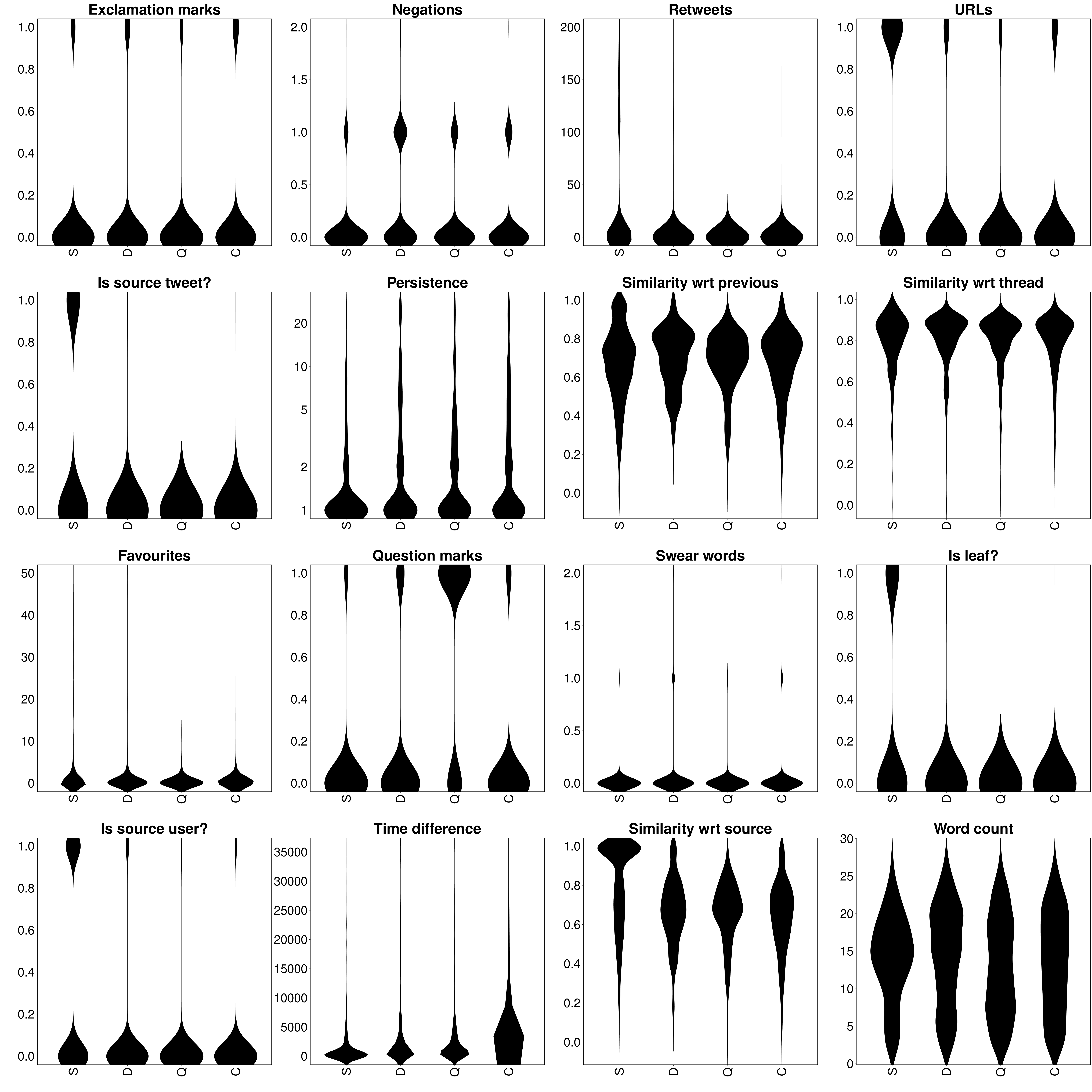}
 \caption{Distributions of feature values across the four categories: Support, Deny, Query and Comment.}
 \label{fig:features}
\end{figure}

To complete the analysis of our experiments, we now look at the different features we used in our study and perform an analysis to understand how distinctive the different features are for the four categories in the stance classification problem. We visualise the different distributions of features for the four categories in beanplots (\cite{kampstra2008beanplot}). We show the visualisations pertaining to 16 of the features under study in Figure \ref{fig:features}. This analysis leads us to some interesting observations towards characterising the different types of stances:

\begin{itemize}
 \item As one might expect, \textit{querying tweets} are more likely to have question marks.
 \item Interestingly, \textit{supporting tweets} tend to have a higher similarity with respect to the source tweet, indicating that the similarity based on word embeddings can be a good feature to identify those tweets.
 \item \textit{Supporting tweets} are more likely to come from the user who posted the source tweet.
 \item \textit{Supporting tweets} are more likely to include links, which is likely indicative of tweets pointing to evidence that supports their position.
 \item Looking at the delay in time of different types of tweets (i.e., the \textit{time difference} feature), we see that \textit{supporting}, \textit{denying} and \textit{querying tweets} are more likely to be observed only in the early stages of a rumour, while later tweets tend to be mostly comments. In fact, these suggests that discussion around the veracity of a rumour occurs especially in the period just after it is posted, whereas the conversation then evolves towards comments that do not discuss the veracity of the rumour in question.
 \item \textit{Denying tweets} are more likely to use negating words. However, negations are also used in other kinds of tweets to a lesser extent, which also makes it more complicated for the classifiers to identify denying tweets. In addition to the low presence of denying tweets in the datasets, the use of negations also in other kinds of responses makes it more challenging to classify them. A way to overcome this may be to use more sophisticated approaches to identify negations that are rebutting the rumour initiated in the source tweet, while getting rid of the rest of the negations.
 \item When we look at the extent to which users persist in their participation in a conversational thread (i.e., the \textit{persistence} feature), we see that users tend to participate more when they are posting \textit{supporting tweets}, showing that users especially insistent when they support a rumour. However, we observe a difference that is not highly remarkable in this particular case.
\end{itemize}

The rest of the features do not show a clear tendency that helps visually distinguish characteristics of the different types of responses. While some features like swear words or exclamation marks may seem indicative of how they orient to somebody else's earlier post, there is no clear difference in reality in our datasets. The same is true for social features like retweets or favourites, where one may expect, for instance, that denying tweets may attract more retweets than comments, as people may want to let others know about rebuttals; the distributions of retweets and favourites are, however, very similar for the different categories.

One possible concern from this analysis is that there are very few features that characterise \textit{commenting tweets}. In fact, the only feature that we have identified as being clearly distinct for \textit{comments} is the \textit{time difference}, given that they are more likely to appear later in the conversations. This may well help classify those late \textit{comments}, however, early comments will be more difficult to be classified based on that feature. Finding additional features to distinguish \textit{comments} from the rest of the tweets may be of help for improving the overall classification.

\section{Conclusions and Future Work}

While discourse and sequential structure of social media conversations have been barely explored in previous work, our work has performed an analysis on the use of different sequential classifiers for the rumour stance classification task. Our work makes three core contributions to existing work on rumour stance classification: (1) we focus on the stance of tweets towards rumours that emerge while breaking news stories unfold; (2) we broaden the stance types considered in previous work to encompass all types of responses observed during breaking news, performing a 4-way classification task; and (3) instead of dealing with tweets as single units in isolation, we exploit the emergent structure of interactions between users on Twitter. In this task, a classifier has to determine if each tweet is supporting, denying, querying or commenting on a rumour's truth value. We mine the sequential structure of Twitter conversational threads in the form of users' replies to one another, extending existing approaches that treat each tweet as a separate unit. We have used four different sequential classifiers: (1) a Hawkes Process classifier that exploits temporal sequences, which showed state-of-the-art performance (\cite{lukasik-etal-2016-hawkes}); (2) a linear-chain CRF modelling tree-structured conversations broken down into branches; (3) a tree CRF modelling them as a graph that includes the whole tree; and (4) an LSTM classifier that also models the conversational threads as branches. These classifiers have been compared with a range of baseline classifiers, including the non-sequential equivalent Maximum Entropy classifier, on eight Twitter datasets associated with breaking news.

While previous stance detection work had mostly limited classifiers to looking at tweets as single units, we have shown that exploiting the discursive characteristics of interactions on Twitter, by considering probabilities of transitions within tree-structured conversational threads, can lead to substantial improvements. Among the sequential classifiers, our results show that the LSTM classifier using a more limited set of features performs the best, thanks to its ability to natively handle context, as well as only relying on branches instead of the whole tree, which reduces the amount of data and complexity that needs to be processed in each sequence. The LSTM has been shown to perform consistently well across datasets, as well as across different types of stances. Besides the comparison of classifiers, our analysis also looks at the distributions of the different features under study as well as how well they characterise the different types of stances. This enables us both to find out which features are the most useful, as well as to suggest improvements needed in future work for improving stance classifiers.

To the best of our knowledge, this is the first attempt at aggregating the conversational structure of Twitter threads to produce classifications at the tweet level. Besides the utility of mining sequences from conversational threads for stance classification, we believe that our results will, in turn, encourage the study of sequential classifiers applied to other natural language processing and data mining tasks where the output for each tweet can benefit from the structure of the entire conversation, e.g., sentiment analysis (\cite{kouloumpis2011twitter,tsytsarau2012survey,saif2016contextual,liu2016identifying,vilares2017supervised,pandey2017twitter}), tweet geolocation (\cite{han2014text,zubiaga2017towards}), language identification (\cite{bergsma2012language,zubiaga2016tweetlid}), event detection (\cite{srijith2017sub}) and analysis of public perceptions on news (\cite{reis2015breaking,an2011media}) and other issues (\cite{pak2010twitter,bian2016mining}).

Our plans for future work include further developing the set of features that characterise the most challenging and least-frequent stances, i.e., denying tweets and querying tweets. These need to be investigated as part of a more detailed and interdisciplinary, thematic analysis of threads (\cite{tolmie2017microblog,housley2017digitizing,housley2017membership}). We also plan to develop an LSTM classifier that mines the entire conversation as a single tree. Our approach assumes that rumours have been already identified or input by a human, hence a final and ambitious aim for future work is the integration with our rumour detection system (\cite{zubiaga2016learning}), whose output would be fed to the stance classification system. The output of our stance classification will also be integrated with a veracity classification system, where the aggregation of stances observed around a rumour will be exploited to determine the likely veracity of the rumour.

\section*{Acknowledgments}

This work has been supported by the PHEME FP7 project (grant No. 611233), the EPSRC Career Acceleration Fellowship EP/I004327/1, Elsevier through the UCL Big Data Institute, and The Alan Turing Institute under the EPSRC grant EP/N510129/1. 


\section{Appendix}\label{ap:features}


\subsection{Local Features}
\label{ssec:local-features}

Local features are extracted from each of the tweets in isolation, and therefore it is not necessary to look at other features in a thread to generate them. We use four types of features to represent the tweets locally.

\noindent \textbf{Local feature type \#1: Lexicon.}
\begin{itemize}
 \item \textit{Word Embeddings:} we use Word2Vec \cite{mikolov2013distributed} to represent the textual content of each tweet. First, we trained a separate Word2Vec model for each of the eight folds, each having the seven events in the training set as input data, so that the event (and the vocabulary) in the test set is unknown. We use large datasets associated with the seven events in the training set, including all the tweets we collected for those events. Finally, we represent each tweet as a vector with 300 dimensions averaging vector representations of the words in the tweet using Word2Vec.
 \item \textit{Part of speech (POS) tags:} we parse the tweets to extract the part-of-speech (POS) tags using Twitie (\cite{bontcheva2013twitie}). Once the tweets are parsed, we represent each tweet with a vector that counts the number of occurrences of each type of POS tag. The final vector therefore has as many features as different types of POS tags we observe in the dataset.
 \item \textit{Use of negation:} this is a feature determining the number of negation words found in a tweet. The existence of negation words in a tweet is determined by looking at the presence of the following words: not, no, nobody, nothing, none, never, neither, nor, nowhere, hardly, scarcely, barely, don't, isn't, wasn't, shouldn't, wouldn't, couldn't, doesn't.
 \item \textit{Use of swear words:} this is a feature determining the number of `bad' words present in a tweet. We use a list of 458 bad words\footnote{\url{http://urbanoalvarez.es/blog/2008/04/04/bad-words-list/}}.
\end{itemize}

\noindent \textbf{Local feature type \#2: Content formatting.}
\begin{itemize}
 \item \textit{Tweet length:} the length of the tweet in number of characters.
 \item \textit{Word count:} the number of words in the tweet, counted as the number of space-separated tokens.
\end{itemize}

\noindent \textbf{Local feature type \#3: Punctuation.}
\begin{itemize}
 \item \textit{Use of question mark:} binary feature indicating the presence or not of at least one question mark in the tweet.
 \item \textit{Use of exclamation mark:} binary feature indicating the presence or not of at least one exclamation mark in the tweet.
\end{itemize}

\noindent \textbf{Local feature type \#4: Tweet formatting.}
\begin{itemize}
 \item \textit{Attachment of URL:} binary feature, capturing the presence or not of at least one URL in the tweet.
\end{itemize}

\subsection{Contextual Features}
\label{ssec:contextual-features}

\noindent \textbf{Contextual feature type \#1: Relational features.}

\begin{itemize}
 \item \textit{Word2Vec similarity wrt source tweet:} we compute the cosine similarity between the word vector representation of the current tweet and the word vector representation of the source tweet. This feature intends to capture the semantic relationship between the current tweet and the source tweet and therefore help inferring the type of response.
 \item \textit{Word2Vec similarity wrt preceding tweet:} likewise, we compute the similarity between the current tweet and the preceding tweet, the one that is directly responding to.
 \item \textit{Word2Vec similarity wrt thread:} we compute another similarity score between the current tweet and the rest of the tweets in the thread excluding the tweets from the same author as that in the current tweet.
\end{itemize}

\noindent \textbf{Contextual feature type \#2: Structural features.}

\begin{itemize}
 \item \textit{Is leaf:} binary feature indicating if the current tweet is a leaf, i.e. the last tweet in a branch of the tree, with no more replies following.
 \item \textit{Is source tweet:} binary feature determining if the tweet is a source tweet or is instead replying to someone else. Note that this feature can also be extracted from the tweet itself, checking if the tweet content begins with a Twitter user handle or not.
 \item \textit{Is source user:} binary feature indicating if the current tweet is posted by the same author as that in the source tweet.
\end{itemize}

\noindent \textbf{Contextual feature type \#3: Social features.}

\begin{itemize}
 \item \textit{Has favourites:} feature indicating the number of times a tweet has been favourited.
 \item \textit{Has retweets:} feature indicating the number of times a tweet has been retweeted.
 \item \textit{Persistence:} this feature is the count of the total number of tweets posted in the thread by the author in the current tweet. High numbers of tweets in a thread indicate that the author participates more.
 \item \textit{Time difference:} this is the time elapsed, in seconds, from when the source tweet was posted to the time the current tweet was posted.
\end{itemize}

\subsection{Hawkes Features}
\label{ssec:hawkes-features}

\begin{itemize}
  \item \textit{Bag of words:} a vector where each token in the dataset represents a feature, where each feature is assigned a number pertaining its count of occurrences in the tweet.
  \item \textit{Timestamp:} The UNIX time in which the tweet was posted.
\end{itemize}

\chapter{Multi-task Learning Over Disparate Label Spaces}\label{ch:disparate}

\boxabstract{
We combine multi-task learning and semi-supervised learning by inducing a joint embedding space between disparate label spaces and learning transfer functions between label embeddings, enabling us to jointly leverage unlabelled data and auxiliary, annotated datasets. We evaluate our approach on a variety of sequence classification tasks with disparate label spaces. We outperform strong single and multi-task baselines and achieve a new state-of-the-art for topic-based sentiment analysis.}\blfootnote{\fullcite{augenstein-etal-2018-multi}}

\newenvironment{starfootnotes}
  {\par\edef\savedfootnotenumber{\number\value{footnote}}
   \renewcommand{\thefootnote}{$\star$} 
   \setcounter{footnote}{0}}
  {\par\setcounter{footnote}{\savedfootnotenumber}}


\section{Introduction}

Multi-task learning (MTL) and semi-supervised learning are both successful paradigms for learning in scenarios with limited labelled data and have in recent years been applied to almost all areas of NLP. Applications of MTL in NLP, for example, include partial parsing (\cite{Soegaard:Goldberg:16}), text normalisation (\cite{Bollman:ea:17}), neural machine translation (\cite{Luong:ea:16}), and keyphrase boundary classification (\cite{augenstein-sogaard-2017-multi}).

Contemporary work in MTL for NLP typically focuses on learning representations that are useful across tasks, often through hard parameter sharing of hidden layers of neural networks (\cite{Collobert2011,Soegaard:Goldberg:16}). If tasks share optimal hypothesis classes at the level of these representations, MTL leads to improvements (\cite{Baxter:00}). However, while sharing hidden layers of neural networks is an effective regulariser (\cite{Soegaard:Goldberg:16}), we potentially {\em lose synergies between the classification functions} trained to associate these representations with class labels. This paper sets out to build an architecture in which such synergies are exploited, with an application to pairwise sequence classification tasks. Doing so, we achieve a new state of the art on topic-based sentiment analysis.

For many NLP tasks, disparate label sets are weakly correlated, e.g. part-of-speech tags correlate with dependencies (\cite{Hashimoto2017}), sentiment correlates with emotion (\cite{Felbo2017,eisner-etal-2016-emoji2vec}), etc. We thus propose to induce a joint label embedding space (visualised in Figure \ref{fig:label_embeddings}) using a Label Embedding Layer that allows us to model these relationships, which we show helps with learning.

In addition, for tasks where labels are closely related, we should be able to not only model their relationship, but also to directly estimate the corresponding label of the target task based on auxiliary predictions. To this end, we propose to train a Label Transfer Network (LTN) jointly with the model to produce pseudo-labels across tasks. 

The LTN can be used to label unlabelled and auxiliary task data by utilising the `dark knowledge' (\cite{Hinton2015}) contained in auxiliary model predictions. This pseudo-labelled data is then incorporated into the model via semi-supervised learning, leading to a natural combination of multi-task learning and semi-supervised learning. We additionally augment the LTN with data-specific diversity features (\cite{ruder2017emnlp}) that aid in learning. 

\paragraph{Contributions} Our contributions are: a) We model the relationships between labels by inducing a joint label space for multi-task learning. b) We propose a Label Transfer Network that learns to transfer labels between tasks and propose to use semi-supervised learning to leverage them for training. c) We evaluate MTL approaches on a variety of classification tasks and shed new light on settings where multi-task learning works. d) We perform an extensive ablation study of our model. e) We report state-of-the-art performance on topic-based sentiment analysis.

\section{Related work}

\paragraph{Learning task similarities} Existing approaches for learning similarities between tasks enforce a clustering of tasks (\cite{Evgeniou2005,Jacob2009}), induce a shared prior (\cite{Yu2005,Xue2007,DaumeIII2009}), or learn a grouping (\cite{Kang2011,Kumar2012}). These approaches focus on homogeneous tasks and employ linear or Bayesian models. They can thus not be directly applied to our setting with tasks using disparate label sets.

\paragraph{Multi-task learning with neural networks} Recent work in multi-task learning goes beyond hard parameter sharing~(\cite{Caruana:93}) and considers different sharing structures, e.g. only sharing at lower layers (\cite{Soegaard:Goldberg:16}) and induces private and shared subspaces (\cite{Liu2017,ruder2017sluice}). These approaches, however, are not able to take into account relationships between labels that may aid in learning. Another related direction is to train on disparate annotations of the same task (\cite{chen-zhang-liu:2016:EMNLP2016,Peng2017}). In contrast, the different nature of our tasks requires a modelling of their label spaces.

\paragraph{Semi-supervised learning} There exists a wide range of semi-supervised learning algorithms, e.g., self-training, co-training, tri-training, EM, and combinations thereof, several of which have also been used in NLP. Our approach is probably most closely related to an algorithm called {\em co-forest} (\cite{Li:Zhou:07}). In co-forest, like here, each learner is improved with unlabeled instances labeled by the ensemble consisting of all the other learners. 
Note also that several researchers have proposed using auxiliary tasks that are unsupervised (\cite{Plank2016a,Rei2017}), which also leads to a form of semi-supervised models. 

\paragraph{Label transformations} The idea of manually mapping between label sets or learning such a mapping to facilitate transfer is not new. \cite{Zhang:ea:12} use distributional information to map from a language-specific tagset to a tagset used for other languages, in order to facilitate cross-lingual transfer. More related to this work, \cite{Kim:ea:15} use canonical correlation analysis to transfer between tasks with disparate label spaces. There has also been work on label transformations in the context of multi-label classification problems (\cite{Yeh:ea:17}).


\section{Multi-task learning with disparate label spaces}

\subsection{Problem definition}

In our multi-task learning scenario, we have access to labelled datasets for $T$ tasks $\mathcal{T}_1, \ldots, \mathcal{T}_T$ at training time with a target task $\mathcal{T}_T$ that we particularly care about. The training dataset for task $\mathcal{T}_i$ consists of $N_k$ examples $X_{\mathcal{T}_i} = \{x_1^{\mathcal{T}_i}, \ldots, x_{N_k}^{\mathcal{T}_i}\}$ and their labels $Y_{\mathcal{T}_i} = \{\mathbf{y}_1^{\mathcal{T}_i}, \ldots, \mathbf{y}_{N_k}^{\mathcal{T}_i}\}$.
Our base model is a deep neural network that performs classic hard parameter sharing (\cite{Caruana:93}): It shares its parameters across tasks and has task-specific softmax output layers, which output a probability distribution $\mathbf{p}^{\mathcal{T}_i}$ for task $\mathcal{T}_i$ according to the following equation:

\begin{equation}
\mathbf{p}^{\mathcal{T}_i} = \mathrm{softmax}(\mathbf{W}^{\mathcal{T}_i}\mathbf{h} + \mathbf{b}^{\mathcal{T}_i})
\end{equation}

where $\mathrm{softmax}(\mathbf{x}) = e^{\mathbf{x}} / \sum^{ \|\mathbf{x}\| }_{i=1} e^{\mathbf{x}_i}$, $\mathbf{W}^{\mathcal{T}_i} \in \mathbb{R}^{L_i \times h}$, $\mathbf{b}^{\mathcal{T}_i} \in \mathbb{R}^{L_i}$
is the weight matrix and bias term of the output layer of task $\mathcal{T}_i$ respectively, $\mathbf{h} \in \mathbb{R}^h$ is the jointly learned hidden representation, $L_i$ is the number of labels for task $\mathcal{T}_i$, and $h$ is the dimensionality of $\mathbf{h}$.

\begin{figure}
    \begin{subfigure}{.32\linewidth}
      \centering
         \includegraphics[height=2.2in]{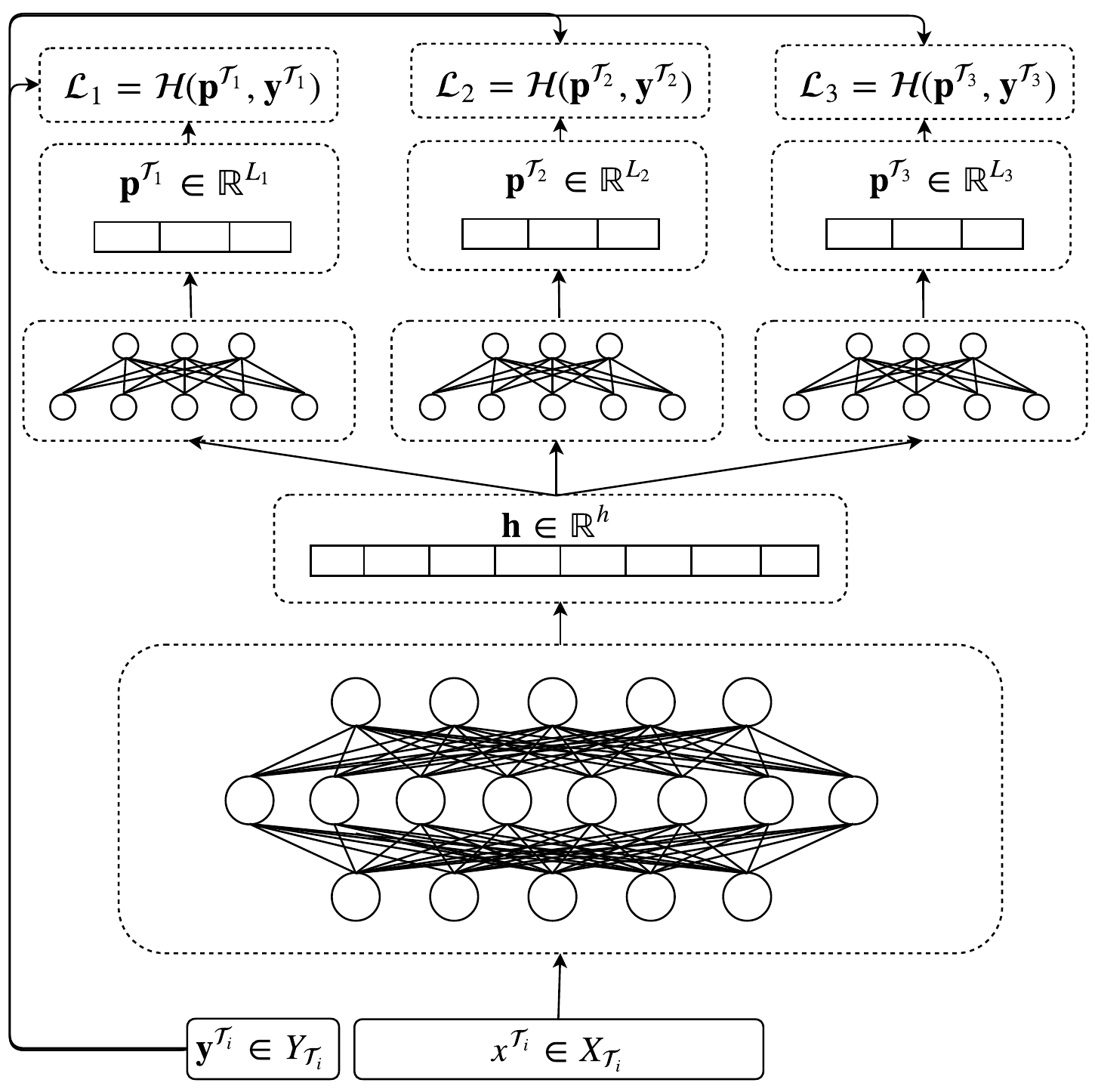}
    \caption{Multi-task learning} \label{fig:mtl}
    \end{subfigure}%
    \hspace*{0.4cm}
    \begin{subfigure}{.21\linewidth}
      \centering
         \includegraphics[height=2.2in]{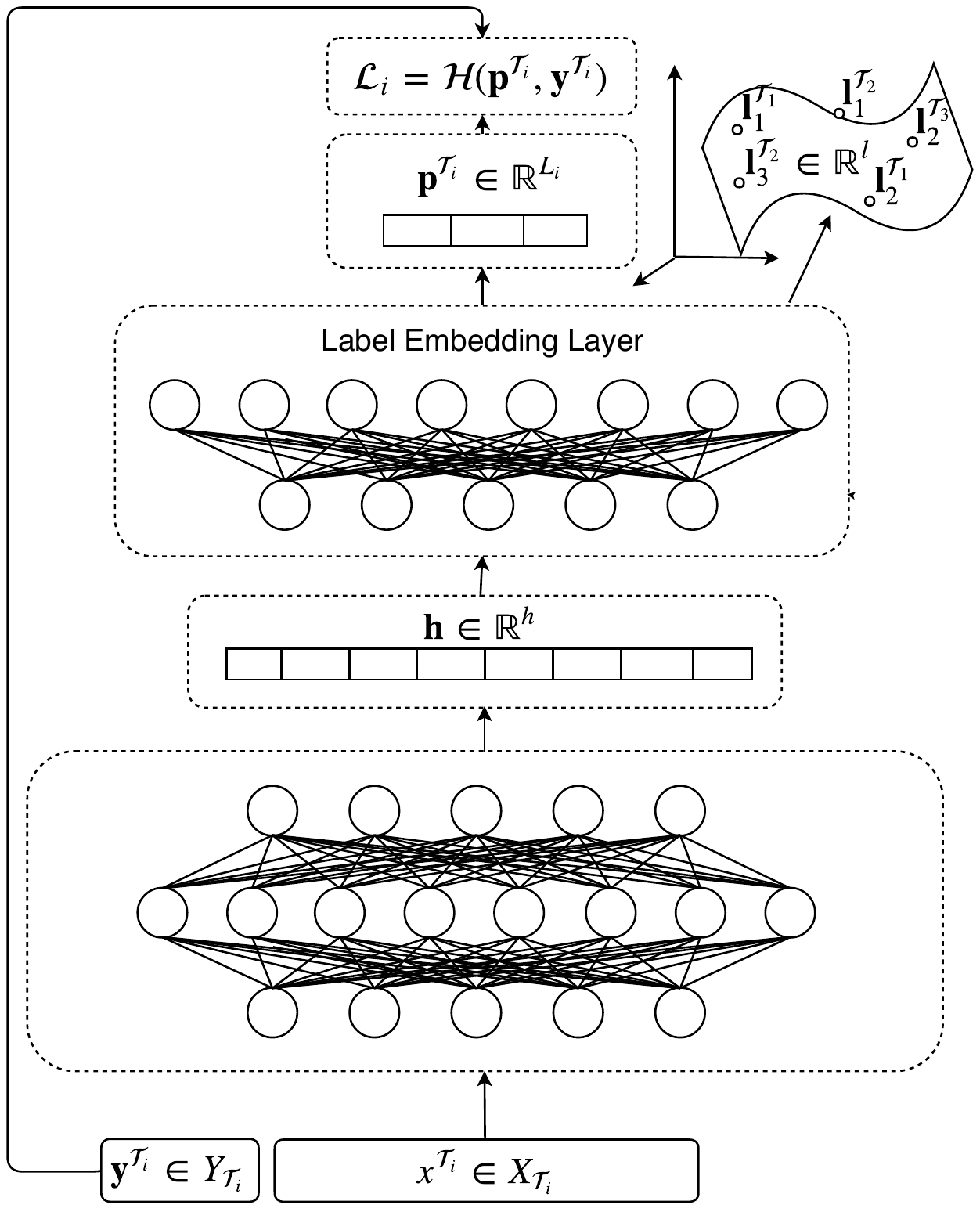}
    \caption{MTL with LEL} \label{fig:lel}
    \end{subfigure}
    \hspace*{0.4cm}
    \begin{subfigure}{.44\linewidth}
      \centering
         \includegraphics[height=2.2in]{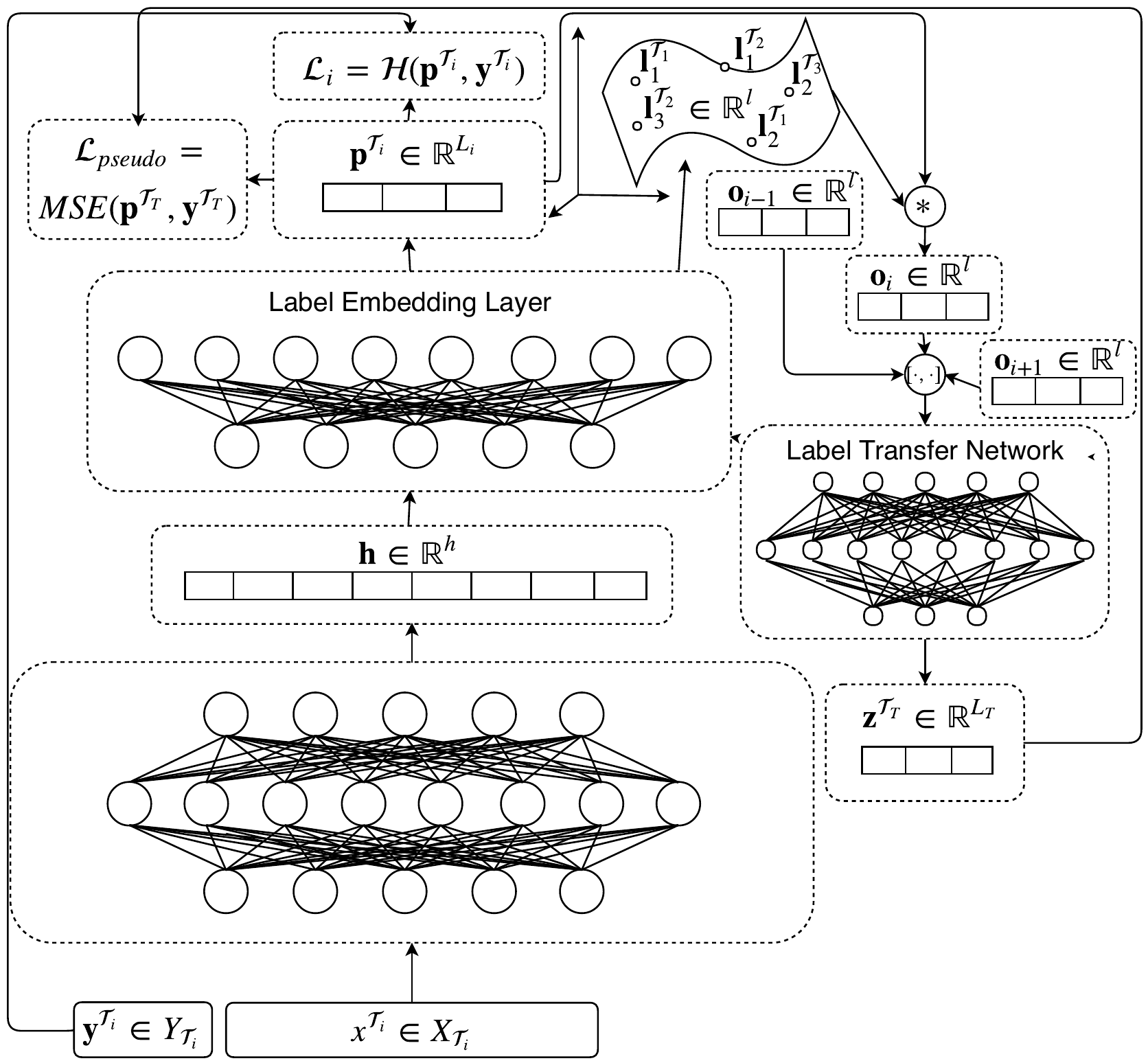}
    \caption{Semi-supervised MTL with LTN} \label{fig:semi-supervised_mtl}
    \end{subfigure}
    \caption{a) Multi-task learning (MTL) with hard parameter sharing and 3 tasks $\mathcal{T}_{1-3}$ and $L_{1-3}$ labels per task. A shared representation $\mathbf{h}$ is used as input to task-specific softmax layers, which optimise cross-entropy losses $\mathcal{L}_{1-3}$. b) MTL with the Label Embedding Layer (LEL) embeds task labels $\mathbf{l}_{1-L_i}^{\mathcal{T}_{1-3}}$ in a joint embedding space and uses these for prediction with a label compatibility function. c) Semi-supervised MTL with the Label Transfer Network (LTN) in addition optimises an unsupervised loss $\mathcal{L}_{pseudo}$ over pseudo-labels $\mathbf{z}^{\mathcal{T}_T}$ on auxiliary/unlabelled data. The pseudo-labels $\mathbf{z}^{\mathcal{T}_T}$ are produced by the LTN for the main task $\mathcal{T}_T$ using the concatenation of auxiliary task label output embeddings $[\mathbf{o}_{i-1},\mathbf{o}_i, \mathbf{o}_{i+1}]$ as input.}
\label{fig:training-procedures}
\end{figure}

The MTL model is then trained to minimise the sum of the individual task losses:

\begin{equation} \label{eq:mtl_loss}
\mathcal{L} = \lambda_1 \mathcal{L}_1 + \ldots + \lambda_T \mathcal{L}_T
\end{equation}

where $\mathcal{L}_i$ is the negative log-likelihood objective $\mathcal{L}_i = \mathcal{H}(\mathbf{p}^{\mathcal{T}_i},\mathbf{y}^{\mathcal{T}_i}) = - \frac{1}{N} \sum_n \sum_j \log \mathbf{p}_j^{\mathcal{T}_i} \mathbf{y}_j^{\mathcal{T}_i} $ and $\lambda_i$ is a parameter that determines the weight of task $\mathcal{T}_i$. In practice, we apply the same weight to all tasks. We show the full set-up in Figure \ref{fig:mtl}.

\subsection{Label Embedding Layer}

In order to learn the relationships between labels, we propose a Label Embedding Layer (LEL) that embeds the labels of all tasks in a joint space. Instead of training separate softmax output layers as above, we introduce a label compatibility function $c(\cdot, \cdot)$ that measures how similar a label with embedding $\mathbf{l}$ is to the hidden representation $\mathbf{h}$:

\begin{equation}
c(\mathbf{l},\mathbf{h}) = \mathbf{l} \cdot \mathbf{h}
\end{equation}

where $\cdot$ is the dot product. This is similar to the Universal Schema Latent Feature Model introduced by \cite{Riedel2013}. In contrast to other models that use the dot product in the objective function, we do not have to rely on negative sampling and a hinge loss (\cite{Collobert2008}) as negative instances (labels) are known. For efficiency purposes, we use matrix multiplication instead of a single dot product and softmax instead of sigmoid activations:

\begin{equation}
\mathbf{p} = \mathrm{softmax}(\mathbf{L} \mathbf{h})
\end{equation}

where $\mathbf{L} \in \mathbb{R}^{(\sum_i L_i) \times l}$ is the label embedding matrix for all tasks and $l$ is the dimensionality of the label embeddings. In practice, we set $l$ to the hidden dimensionality $h$. We use padding if $l < h$. We apply a task-specific mask to $\mathbf{L}$ in order to obtain a task-specific probability distribution $\mathbf{p}^{\mathcal{T}_i}$. The LEL is shared across all tasks, which allows us to learn the relationships between the labels in the joint embedding space. We show MTL with the LEL in Figure \ref{fig:lel}.

\subsection{Label Transfer Network}

The LEL allows us to learn the relationships between labels. In order to make use of these relationships, we would like to leverage the predictions of our auxiliary tasks to estimate a label for the target task. To this end, we introduce the Label Transfer Network (LTN). This network takes the auxiliary task outputs as input. In particular, we define the output label embedding $\mathbf{o}_i$ of task $\mathcal{T}_i$ as the sum of the task's label embeddings $\mathbf{l}_j$ weighted with their probability $\mathbf{p}^{\mathcal{T}_i}_j$:

\begin{equation}
\mathbf{o}_i = \sum^{L_i}_{j=1} \mathbf{p}^{\mathcal{T}_i}_j \mathbf{l}_j 
\end{equation}

The label embeddings $\mathbf{l}$ encode general relationship between labels, while the model's probability distribution $\mathbf{p}^{\mathcal{T}_i}$ over its predictions encodes fine-grained information useful for learning (\cite{Hinton2015}). The LTN is trained on labelled target task data. For each example, the corresponding label output embeddings of the auxiliary tasks are fed into a  multi-layer perceptron (MLP), which is trained with a negative log-likelihood objective $\mathcal{L}_{\mathrm{LTN}}$ to produce a pseudo-label ${\mathbf{z}}^{\mathcal{T}_T}$ for the target task ${\mathcal{T}_T}$:

\begin{equation}
\mathrm{LTN}_T = \mathrm{MLP}([\mathbf{o}_1, \ldots, \mathbf{o}_{T-1}]) 
\end{equation}

where $[\cdot, \cdot]$ designates concatenation. The mapping of the tasks in the LTN yields another signal that can be useful for optimisation and act as a regulariser. The LTN can also be seen as a mixture-of-experts layer (\cite{Jacobs1991}) where the experts are the auxiliary task models. As the label embeddings are learned jointly with the main model, the LTN is more sensitive to the relationships between labels than a separately learned mixture-of-experts model that only relies on the experts' output distributions. As such, the LTN can be directly used to produce predictions on unseen data.

\subsection{Semi-supervised MTL}

The downside of the LTN is that it requires additional parameters and relies on the predictions of the auxiliary models, which impacts the runtime during testing. Instead, of using the LTN for prediction directly, we can use it to provide pseudo-labels for unlabelled or auxiliary task data by utilising auxiliary predictions for semi-supervised learning.

We train the target task model on the pseudo-labelled data to minimise the squared error between the model predictions $\mathbf{p}^{\mathcal{T}_i}$ and the pseudo labels $\mathbf{z}^{\mathcal{T}_i}$ produced by the LTN:

\begin{equation}
\mathcal{L}_{pseudo} = MSE(\mathbf{p}^{\mathcal{T}_T}, \mathbf{z}^{\mathcal{T}_T}) = ||\mathbf{p}^{\mathcal{T}_T} - \mathbf{z}^{\mathcal{T}_T}|| ^ 2
\end{equation}

We add this loss term to the MTL loss in Equation \ref{eq:mtl_loss}.
As the LTN is learned together with the MTL model, pseudo-labels produced early during training will likely not be helpful as they are based on unreliable auxiliary predictions. For this reason, we first train the base MTL model until convergence and then augment it with the LTN.
We show the full semi-supervised learning procedure in Figure \ref{fig:semi-supervised_mtl}.

\subsection{Data-specific features}

When there is a domain shift between the datasets of different tasks as is common for instance when learning NER models with different label sets, the output label embeddings might not contain sufficient information to bridge the domain gap.

To mitigate this discrepancy, we augment the LTN's input with features that have been found useful for transfer learning (\cite{ruder2017emnlp}). In particular, we use the number of word types, type-token ratio, entropy, Simpson's index, and R{\'e}nyi entropy as diversity features. We calculate each feature for each example.
\footnote{For more information regarding the feature calculation, refer to \cite{ruder2017emnlp}.} The features are then concatenated with the input of the LTN. 

\subsection{Other multi-task improvements}

Hard parameter sharing can be overly restrictive and provide a regularisation that is too heavy when jointly learning many tasks. For this reason, we propose several additional improvements that seek to alleviate this burden: We use skip-connections, which have been shown to be useful for multi-task learning in recent work (\cite{ruder2017sluice}). Furthermore, we add a task-specific layer before the output layer, which is useful for learning task-specific transformations of the shared representations (\cite{Soegaard:Goldberg:16,ruder2017sluice}).

\section{Experiments}

For our experiments, we evaluate on a wide range of text classification tasks. In particular, we choose pairwise classification tasks---i.e. those that condition the reading of one sequence on another sequence---as we are interested in understanding if knowledge can be transferred even for these more complex interactions. To the best of our knowledge, this is the first work on transfer learning between such pairwise sequence classification tasks. We implement all our models in Tensorflow (\cite{abadi2016tensorflow}) and release the code at \url{https://github.com/coastalcph/mtl-disparate}. 

\begin{table}
\centering
\fontsize{10}{10}\selectfont
\begin{tabular}{l l c c c}
\toprule
Task & Domain & $N$ & $L$ & Metric\\
\midrule
{\texttt Topic-2} & Twitter & 4,346 & 2 & $\rho^{PN}$ \\
{\texttt Topic-5} & Twitter & 6,000 & 5 & $MAE^M$\\
{\texttt Target} & Twitter & 6,248 & 3 & $F_1^M$ \\
{\texttt Stance} & Twitter & 2,914 & 3 & $F_1^{FA}$\\
{\texttt ABSA-L} & Reviews & 2,909 & 3 & $Acc$\\
{\texttt ABSA-R} & Reviews & 2,507 & 3 & $Acc$\\
{\texttt FNC-1} & News & 39,741 & 4 & $Acc$\\
{\texttt MultiNLI} & Diverse & 392,702 & 3 & $Acc$\\
\bottomrule
\end{tabular}%
\caption{Training set statistics and evaluation metrics of every task. $N$: \# of examples. $L$: \# of labels.}
\label{tab:dataset-stats}
\end{table}

\subsection{Tasks and datasets}\label{sec:datasets}

\setlength{\tabcolsep}{0.3em}
\begin{table}[h]
\fontsize{10}{10}\selectfont
\begin{center}
\begin{tabular}{|L|}
\toprule
\textbf{Topic-based sentiment analysis}: \\
\textit{Tweet}: No power at home, sat in the dark listening to AC/DC in the hope it'll make the electricity come back again \\
\textit{Topic}: AC/DC \\
\textit{Label}: positive \\
\midrule
\textbf{Target-dependent sentiment analysis}: \\
\textit{Text}: how do you like settlers of catan for the wii? \\
\textit{Target}: wii  \\
\textit{Label}: neutral \\
\midrule
\textbf{Aspect-based sentiment analysis}: \\
\textit{Text}: For the price, you cannot eat this well in Manhattan\\
\textit{Aspects}: restaurant prices, food quality \\
\textit{Label}: positive\\
\midrule
\textbf{Stance detection}: \\
\textit{Tweet}: Be prepared - if we continue the policies of the liberal left, we will be \#Greece\\
\textit{Target}: Donald Trump\\
\textit{Label}: favor\\
\midrule
\textbf{Fake news detection}: \\
\textit{Document}: Dino Ferrari hooked the whopper wels catfish, (...), which could be the biggest in the world.\\
\textit{Headline}: Fisherman lands 19 STONE catfish which could be the biggest in the world to be hooked\\
\textit{Label}: agree\\
\midrule
\textbf{Natural language inference}: \\
\textit{Premise}: Fun for only children\\
\textit{Hypothesis}: Fun for adults and children\\
\textit{Label}: contradiction\\
\bottomrule
\end{tabular}
\end{center}
\caption{\label{tab:dataset-examples} Example instances from the datasets described in Section \ref{sec:datasets}.}
\end{table}

We use the following tasks and datasets for our experiments, show task statistics in Table \ref{tab:dataset-stats}, and summarise examples in Table \ref{tab:dataset-examples}:

\paragraph{Topic-based sentiment analysis} Topic-based sentiment analysis aims to estimate the sentiment of a tweet known to be about a given topic. We use the data from SemEval-2016 Task 4 Subtask B and C (\cite{SemEval:2016:task4}) for predicting on a two-point scale of positive and negative ({\texttt Topic-2}) and five-point scale ranging from highly negative to highly positive ({\texttt Topic-5}) respectively. An example from this dataset would be to classify the tweet ``No power at home, sat in the dark listening to AC/DC in the hope it'll make the electricity come back again'' known to be about the topic ``AC/DC'', which is labelled as a positive sentiment. The evaluation metrics for {\texttt Topic-2} and {\texttt Topic-5} are macro-averaged recall ($\rho^{PN}$) and macro-averaged mean absolute error ($MAE^M$) respectively, which are both averaged across topics.

\paragraph{Target-dependent sentiment analysis} Target-dependent sentiment analysis ({\texttt Target}) seeks to classify the sentiment of a text's author towards an entity that occurs in the text as positive, negative, or neutral. We use the data from \cite{Dong2014}. An example instance is the expression ``how do you like settlers of catan for the wii?'' which is labelled as neutral towards the target ``wii'.' The evaluation metric is macro-averaged $F_1$ ($F_1^M$).

\paragraph{Aspect-based sentiment analysis} Aspect-based sentiment analysis is the task of identifying whether an aspect, i.e. a particular property of an item is associated with a positive, negative, or neutral sentiment (\cite{Ruder2016a}). We use the data of SemEval-2016 Task 5 Subtask 1 Slot 3 (\cite{Pontiki2016Aspect}) for the laptops ({\texttt ABSA-L}) and restaurants ({\texttt ABSA-R}) domains. An example is the sentence ``For the price, you cannot eat this well in Manhattan'', labelled as positive towards both the aspects ``restaurant prices'' and ``food quality''. The evaluation metric for both domains is accuracy ($Acc$).

\paragraph{Stance detection} Stance detection ({\texttt Stance}) requires a model, given a text and a target entity, which might not appear in the text, to predict whether the author of the text is in favour or against the target or whether neither inference is likely (\cite{augenstein-etal-2016-stance}). We use the data of SemEval-2016 Task 6 Subtask B (\cite{mohammad-etal-2016-semeval}). An example from this dataset would be to predict the stance of the tweet ``Be prepared - if we continue the policies of the liberal left, we will be \#Greece'' towards the topic ``Donald Trump'', labelled as ``favor''. The evaluation metric is the macro-averaged $F_1$ score of the ``favour'' and ``against'' classes ($F_1^{FA}$).

\begin{table}
\centering
\resizebox{\textwidth}{!}{%
\begin{tabular}{l c c c c c c c c}
\toprule
 & {\texttt Stance} & {\texttt FNC} & {\texttt MultiNLI} & {\texttt Topic-2} & {\texttt Topic-5}* & {\texttt ABSA-L} & {\texttt ABSA-R} & {\texttt Target}\\
\midrule
\cite{augenstein-etal-2016-stance} & \textbf{49.01} & - & - & - & - & - & - & -\\
\cite{journals/corr/RiedelASR17} & - & \textbf{88.46} & - & - & - & - & - & -\\
\cite{chen2017recurrent} & - & - & \textbf{74.90} & - & - & - & - & -\\
\cite{palogiannidi2016tweester}  & - & - & - & \underline{79.90} & - & - & - & -\\
\cite{balikas2016twise}  & - & - & - & - & \textbf{0.719} & - & - & - \\
\cite{Brun2016} & - & - & - & - & - & - & \textbf{88.13} & -\\
\cite{Kumar2016} & - & - & - & - & - & \textbf{82.77} & \underline{86.73} & -\\
\cite{conf/ijcai/VoZ15}  & - & - & - & - & - & - &- & \textbf{69.90} \\
\midrule
STL & 41.1 & 72.72 & 49.25 & 63.92 & 0.919 & \underline{76.74} & 67.47 & 64.01 \\
\midrule
MTL + LEL & \underline{46.26} & 72.71 & \underline{49.94}  & \textbf{80.52} & 0.814 & 74.94 & 79.90 & \underline{66.42} \\
MTL + LEL + LTN, main model & 43.16  & \underline{72.73} &  48.75 & 73.90 & \underline{0.810} & 75.06 & 83.71 & 66.10 \\
MTL + LEL + LTN + semi, main model & 43.56 & 72.72 & 48.00 & 72.35  & 0.821 & 75.42 & 83.26 & 63.00 \\
\bottomrule
\end{tabular}%
}
\caption{Comparison of our best performing models on the test set against a single task baseline and the state of the art, with task specific metrics. *: lower is better. Bold: best. Underlined: second-best.}
\label{tab:results-sota}
\end{table}

\paragraph{Fake news detection} The goal of fake news detection in the context of the Fake News Challenge\footnote{\url{http://www.fakenewschallenge.org/}} is to estimate whether the body of a news article agrees, disagrees, discusses, or is unrelated towards a headline. We use the data from the first stage of the Fake News Challenge ({\texttt FNC-1}). An example for this dataset is the document ``Dino Ferrari hooked the whopper wels catfish, (...), which could be the biggest in the world.'' with the headline ``Fisherman lands 19 STONE catfish which could be the biggest in the world to be hooked'' labelled as ``agree''. The evaluation metric is accuracy ($Acc$)\footnote{We use the same metric as \cite{journals/corr/RiedelASR17}.}.

\paragraph{Natural language inference} Natural language inference is the task of predicting whether one sentences entails, contradicts, or is neutral towards another one. We use the Multi-Genre NLI corpus ({\texttt MultiNLI}) from the RepEval 2017 shared task (\cite{Nangia2017}). An example for an instance would be the sentence pair ``Fun for only children'', ``Fun for adults and children'', which are in a ``contradiction'' relationship. The evaluation metric is accuracy ($Acc$).

\subsection{Base model}

Our base model is the Bidirectional Encoding model (\cite{augenstein-etal-2016-stance}), a state-of-the-art model for stance detection that conditions a bidirectional LSTM (BiLSTM) encoding of a text on the BiLSTM encoding of the target. Unlike \cite{augenstein-etal-2016-stance}, we do not pre-train word embeddings on a larger set of unlabelled in-domain text for each task as we are mainly interested in exploring the benefit of multi-task learning for generalisation.

\subsection{Training settings}

We use BiLSTMs with one hidden layer of $100$ dimensions, $100$-dimensional randomly initialised word embeddings, a label embedding size of $100$. We train our models with RMSProp, a learning rate of $0.001$, a batch size of $128$, and early stopping on the validation set of the main task with a patience of $3$.

\section{Results}

Our main results are shown in Table \ref{tab:results-sota}, with a comparison against the state of the art. We present the results of our multi-task learning network with label embeddings (MTL + LEL), multi-task learning with label transfer (MTL + LEL + LTN), and the semi-supervised extension of this model. On 7/8 tasks, at least one of our architectures is better than single-task learning; and in 4/8, all our architectures are much better than single-task learning. 

The state-of-the-art systems we compare against are often highly specialised, task-dependent architectures. Our architectures, in contrast, have not been optimised to compare favourably against the state of the art, as our main objective is to develop a novel approach to multi-task learning leveraging synergies between label sets and knowledge of marginal distributions from unlabeled data. For example, we do not use pre-trained word embeddings (\cite{augenstein-etal-2016-stance,palogiannidi2016tweester,conf/ijcai/VoZ15}), class weighting to deal with label imbalance (\cite{balikas2016twise}), or domain-specific sentiment lexicons (\cite{Brun2016,Kumar2016}). Nevertheless, our approach outperforms the state-of-the-art on two-way topic-based sentiment analysis ({\texttt Topic-2}).

The poor performance compared to the state-of-the-art on {\texttt FNC} and {\texttt MultiNLI} is expected; as we alternate among the tasks during training, our model only sees a comparatively small number of examples of both corpora, which are one and two orders of magnitude larger than the other datasets. For this reason, we do not achieve good performance on these tasks as main tasks, but they are still useful as auxiliary tasks as seen in Table \ref{tab:auxiliary-tasks}.

\section{Analysis}
\subsection{Label Embeddings}

\begin{figure}
	\centering
  	\includegraphics[width=0.7\linewidth]{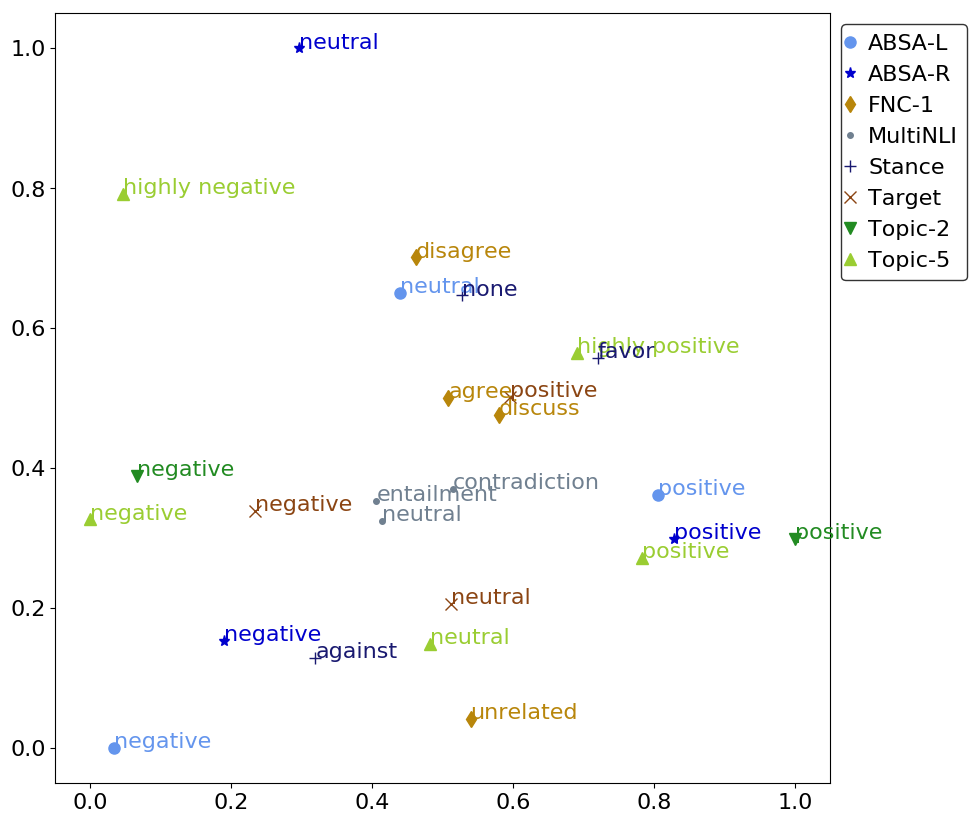}
  	\caption{Label embeddings of all tasks. Positive, negative, and neutral labels are clustered together.}
  	\label{fig:label_embeddings}
\end{figure}

Our results above show that, indeed, modelling the similarity between tasks using label embeddings sometimes leads to much better performance. Figure \ref{fig:label_embeddings} shows why. In Figure~\ref{fig:label_embeddings}, we visualise the label embeddings of an MTL+LEL model trained on all tasks, using PCA. As we can see, similar labels are clustered together across tasks, e.g. there are two positive clusters (middle-right and top-right), two negative clusters (middle-left and bottom-left), and two neutral clusters (middle-top and middle-bottom).


Our visualisation also provides us with a picture of what auxilary tasks are beneficial, and to what extent we can expect synergies from multi-task learning. For instance, the notion of positive sentiment appears to be very similar across the topic-based and aspect-based tasks, while the conceptions of negative and neutral sentiment differ. In addition, we can see that the model has failed to learn a relationship between {\texttt MultiNLI} labels and those of other tasks, possibly accounting for its poor performance on the inference task. We did not evaluate the correlation between label embeddings and task performance, but \cite{Bjerva2017} recently suggested that mutual information of target and auxiliary task label sets is a good predictor of gains from multi-task learning.  

\subsection{Auxilary Tasks}


\begin{table}
\centering
\begin{tabular}{l l}
\toprule
Main task & Auxiliary tasks\\
\midrule
{\texttt Topic-2} & {\texttt FNC-1}, {\texttt MultiNLI}, {\texttt Target} \\
\multirow{2}{*}{{\texttt Topic-5}} & {\texttt FNC-1}, {\texttt MultiNLI}, {\texttt ABSA-L}, \\
 & {\texttt Target}\\
{\texttt Target} & {\texttt FNC-1}, {\texttt MultiNLI}, {\texttt Topic-5} \\
{\texttt Stance} & {\texttt FNC-1}, {\texttt MultiNLI}, {\texttt Target} \\
{\texttt ABSA-L} & {\texttt Topic-5} \\
{\texttt ABSA-R} & {\texttt Topic-5}, {\texttt ABSA-L}, {\texttt Target}\\
\multirow{2}{*}{{\texttt FNC-1}} & {\texttt Stance}, {\texttt MultiNLI}, {\texttt Topic-5},\\
& {\texttt ABSA-R}, {\texttt Target} \\
{\texttt MultiNLI} & {\texttt Topic-5} \\
\bottomrule
\end{tabular}%
\caption{Best-performing auxiliary tasks for different main tasks.}
\label{tab:auxiliary-tasks}
\end{table}


For each task, we show the auxiliary tasks that achieved the best performance on the development data in Table \ref{tab:auxiliary-tasks}. In contrast to most existing work, we did not restrict ourselves to performing multi-task learning with only one auxiliary task (\cite{Soegaard:Goldberg:16,Bingel:ea:17}). Indeed we find that most often a combination of auxiliary tasks achieves the best performance. In-domain tasks are less used than we assumed; only {\texttt Target} is consistently used by all Twitter main tasks. In addition, tasks with a higher number of labels, e.g. {\texttt Topic-5} are used more often. Such tasks provide a more fine-grained reward signal, which may help in learning representations that generalise better. Finally, tasks with large amounts of training data such as {\texttt FNC-1} and {\texttt MultiNLI} are also used more often. Even if not directly related, the larger amount of training data that can be indirectly leveraged via multi-task learning may help the model focus on relevant parts of the representation space (\cite{Caruana:93}). These observations shed additional light on when multi-task learning may be useful that go beyond existing studies (\cite{Bingel:ea:17}).

\begin{table*}[h]
\centering
\resizebox{\textwidth}{!}{%
\begin{tabular}{l c c c c c c c c}
\toprule
 & {\texttt Stance} & {\texttt FNC} & {\texttt MultiNLI} & {\texttt Topic-2} & {\texttt Topic-5}* & {\texttt ABSA-L} & {\texttt ABSA-R} & {\texttt Target}\\
\midrule
MTL & 44.12 & \underline{72.75} & \underline{49.39} & \textbf{80.74} & 0.859 & 74.94 & 82.25 & 65.73 \\
\midrule
MTL + LEL & \textbf{46.26} & 72.71 & \textbf{49.94}  & \underline{80.52} & 0.814 & 74.94 & 79.90 & \textbf{66.42} \\
MTL + LTN & 40.95 & 72.72 & 44.14 & 78.31  & 0.851 & 73.98  & 82.37 & 63.71 \\
MTL + LTN, main model & 41.60 & 72.72 & 47.62 & 79.98  & 0.814 & \underline{75.54} & 81.70 & 65.61  \\
MTL + LEL + LTN & \underline{44.48} & \textbf{72.76} & 43.72 & 74.07 & 0.821 & \textbf{75.66} & 81.92 & 65.00 \\
MTL + LEL + LTN, main model & 43.16  & 72.73 &  48.75 & 73.90 & 0.810 & 75.06 & \textbf{83.71} & \underline{66.10} \\
\midrule
MTL + LEL + LTN + main preds feats & 42.78 & 72.72 & 45.41 & 66.30  & 0.835 & 73.86 & 81.81 & 65.08 \\
MTL + LEL + LTN + main preds feats, main model &  42.65 & 72.73 & 48.81 & 67.53  & \textbf{0.803} & 75.18 & 82.59 & 63.95 \\
\midrule
MTL + LEL + LTN + main preds feats -- diversity feats & 42.78 & 72.72 & 43.13 & 66.3 & 0.835 & 73.5 & 81.7 & 63.95 \\
MTL + LEL + LTN + main preds feats -- diversity feats, main model & 42.47 & 72.74 & 47.84 & 67.53 & \underline{0.807} & 74.82 & 82.14 & 65.11 \\
\midrule
MTL + LEL + LTN + semi & 42.65 & \underline{72.75} & 44.28 & 77.81  & 0.841 & 74.10 & 81.36 & 64.45 \\
MTL + LEL + LTN + semi, main model & 43.56 & 72.72 & 48.00 & 72.35  & 0.821 & 75.42 & \underline{83.26} & 63.00 \\

\bottomrule
\end{tabular}%
}
\caption{Ablation results with task-specific evaluation metrics on test set with early stopping on dev set. \textit{LTN} means the output of the relabelling function is shown, which does not use the task predictions, only predictions from other tasks. \textit{LTN + main preds feats} means main model predictions are used as features for the relabelling function. \textit{LTN, main model} means that the main model predictions of the model that trains a relabelling function are used. Note that for {\texttt MultiNLI}, we down-sample the training data. *: lower is better. Bold: best. Underlined: second-best.} 
\label{tab:ablation}
\end{table*}

\subsection{Ablation analysis}

We now perform a detailed ablation analysis of our model, the results of which are shown in Table \ref{tab:ablation}. We ablate whether to use the LEL (\textit{+ LEL}), whether to use the LTN (\textit{+ LTN}), whether to use the LEL output or the main model output for prediction (main model output is indicated by \textit{, main model}), and whether to use the LTN as a regulariser or for semi-supervised learning (semi-supervised learning is indicated by \textit{+ semi}). We further test whether to use diversity features (\textit{-- diversity feats}) and whether to use main model predictions for the LTN (\textit{+ main model feats}).

Overall, the addition of the Label Embedding Layer improves the performance over regular MTL in almost all cases.

\begin{table}
\centering
\resizebox{0.48\textwidth}{!}{%
\begin{tabular}{l c c c c }
\toprule
Task & Main & LTN & Main (Semi) & LTN (Semi) \\
\midrule
{\texttt Stance} & 2.12 & 2.62 & 1.94 & 1.28 \\
{\texttt FNC} & 4.28 & 2.49 & 6.92 & 4.84 \\
{\texttt MultiNLI}  & 1.5 & 1.95 & 1.94 & 1.28 \\
{\texttt Topic-2} & 6.45 & 4.44 & 5.87 & 5.59 \\
{\texttt Topic-5}* & 9.22 & 9.71 & 11.3 & 5.90 \\
{\texttt ABSA-L} & 3.79 & 2.52 & 9.06 & 6.63 \\
{\texttt ABSA-R} & 10.6 & 6.70 & 9.06 & 6.63 \\
{\texttt Target} & 26.3 & 14.6 & 20.1 & 15.7 \\
\bottomrule
\end{tabular}
}
\caption{Error analysis of LTN with and without semi-supervised learning for all tasks. Metric shown: percentage of correct predictions only made by either the relabelling function or the main model, respectively, relative to the the number of all correct predictions.}
\label{tab:ltn-errorana}
\end{table}

\begin{figure}
	\centering
  	\includegraphics[width=1.0\linewidth]{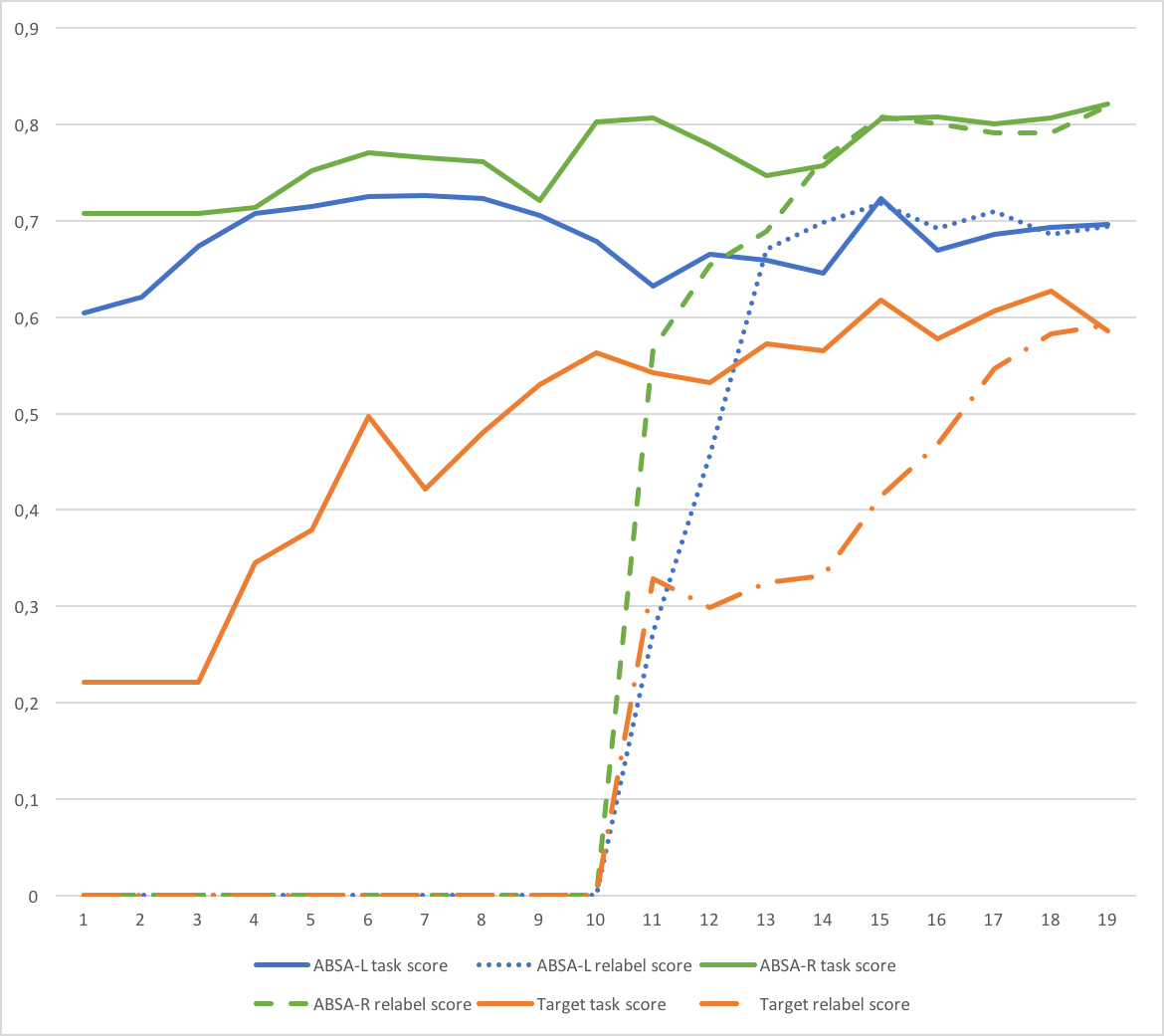}
  	\caption{Learning curves with LTN for selected tasks, dev performances shown. The main model is pre-trained for 10 epochs, after which the relabelling function is trained.}
  	\label{fig:ltn_learning_curve}
\end{figure}

\subsection{Label transfer network}

To understand the performance of the LTN, we analyse learning curves of the relabelling function vs. the main model. Examples for all tasks without semi-supervised learning are shown in Figure \ref{fig:ltn_learning_curve}.
One can observe that the relabelling model does not take long to converge as it has fewer parameters than the main model. Once the relabelling model is learned alongside the main model, the main model performance first stagnates, then starts to increase again. For some of the tasks, the main model ends up with a higher task score than the relabelling model.
We hypothesise that the softmax predictions of other, even highly related tasks are less helpful for predicting main labels than the output layer of the main task model. At best, learning the relabelling model alongside the main model might act as a regulariser to the main model and thus improve the main model's performance over a baseline MTL model, as it is the case for {\texttt TOPIC-5} (see Table \ref{tab:ablation}).

To further analyse the performance of the LTN, we look into to what degree predictions of the main model and the relabelling model for individual instances are complementary to one another. Or, said differently, we measure the percentage of correct predictions made only by the relabelling model or made only by the main model, relative to the number of correct predictions overall. Results of this for each task are shown in Table \ref{tab:ltn-errorana} for the LTN with and without semi-supervised learning. One can observe that, even though the relabelling function overall contributes to the score to a lesser degree than the main model, a substantial number of correct predictions are made by the relabelling function that are missed by the main model. This is most prominently pronounced for {\texttt ABSA-R}, where the proportion is 14.6.

\section{Conclusion}

We have presented a multi-task learning architecture that (i) leverages potential synergies between classifier functions relating shared representations with disparate label spaces and (ii) enables learning from mixtures of labeled and unlabeled data. We have presented experiments with combinations of eight pairwise sequence classification tasks. Our results show that leveraging synergies between label spaces sometimes leads to big improvements, and we have presented a new state of the art for topic-based sentiment analysis. Our analysis further showed that (a) the learned label embeddings were indicative of gains from multi-task learning, (b) auxiliary tasks were often beneficial across domains,  and (c) label embeddings almost always led to better performance. We also investigated the dynamics of the label transfer network we use for exploiting the synergies between disparate label spaces. 

\section*{Acknowledgments}

Sebastian Ruder is supported by the Irish Research Council Grant Number EBPPG/2014/30 and Science Foundation Ireland Grant Number SFI/12/RC/2289. Anders Søgaard is supported by the ERC Starting Grant Number 313695. Isabelle Augenstein is supported by Eurostars grant Number E10138. We further gratefully acknowledge the support of NVIDIA Corporation with the donation of the Titan Xp GPU used for this research.

\chapter{Sequential Alignment of Text Representations}\label{ch:sequential}

\boxabstract{
  Language evolves over time in many ways relevant to natural language processing tasks. For example, recent occurrences of tokens 'BERT' and 'ELMO' in publications refer to neural network architectures rather than persons. This type of temporal signal is typically overlooked, but is important if one aims to deploy a machine learning model over an extended period of time. In particular, language evolution causes data drift between time-steps in sequential decision-making tasks. Examples of such tasks include prediction of paper acceptance for yearly conferences (regular intervals) or author stance prediction for rumours on Twitter (irregular intervals). Inspired by successes in computer vision, we tackle data drift by sequentially aligning learned representations. 
  We evaluate on three challenging tasks varying in terms of time-scales, linguistic units, and domains. These tasks show our method outperforming several strong baselines, including using all available data. We argue that, due to its low computational expense, sequential alignment is a practical solution to dealing with language evolution.
}\blfootnote{\fullcite{Bjerva_Kouw_Augenstein_2020}}

\section{Introduction}
\noindent As time passes, language usage changes.
For example, the names `Bert' and `Elmo' would only rarely make an appearance prior to 2018 in the context of scientific writing. After the publication of BERT (\cite{devlin-etal-2019-bert}) and ELMo (\cite{peters-etal-2018-deep}), however, usage has increased in frequency. In the context of named entities on Twitter, it is also likely that these names would be tagged as \textsc{person} prior to 2018, and are now more likely to refer to an \textsc{artefact}. As such, their part-of-speech tags will also differ. Evidently, evolution of language usage affects natural language processing (NLP) tasks, and as such, models based on data from one point in time cannot be expected to generalise to the future.

In order to become more robust to language evolution, data should be collected at multiple points in time. We consider a dynamic learning paradigm where one makes predictions for data points from the current time-step given labelled data points from previous time-steps. As time increments, data points from the current step are labelled and new unlabelled data points are observed. This setting occurs in NLP in, for instance, the prediction of paper acceptance to conferences (\cite{kang18naacl}) or named entity recognition from yearly data dumps of Twitter (\cite{derczynski:2016}). Changes in language usage cause a data drift between time-steps and some way of controlling for the shift between time-steps is necessary. 


In this paper, we apply a domain adaptation technique to correct for shifts. Domain adaptation is a furtive area of research within machine learning that deals with learning from training data drawn from one data-generating distribution (source domain) and generalising to test data drawn from another, different data-generating distribution (target domain) (\cite{kouw2019review}). 
We are interested in whether a sequence of adaptations can compensate for the data drift caused by shifts in the meaning of words or features across time. Given that linguistic tokens are embedded in some vector space using neural language models, we observe that in time-varying dynamic tasks, the drift causes token embeddings to occupy different parts of embedding space over consecutive time-steps. We want to avoid the computational expense of re-training a neural network every time-step. Instead, in each time-step, we map linguistic tokens using the same pre-trained language model (a "BERT" network, \cite{devlin-etal-2019-bert}) and align the resulting embeddings using a second procedure called subspace alignment (\cite{fernando2013unsupervised}). We apply subspace alignment sequentially: find the principal components in each time-step and linearly transform the components from the previous step to match the current step. A classifier trained on the aligned embeddings from the previous step will be more suited to classify embeddings in the current step. We show that sequential subspace alignment (SSA) yields substantial improvements in three challenging tasks: paper acceptance prediction on the PeerRead data set (\cite{kang18naacl}); Named Entity Recognition on the Broad Twitter Corpus (\cite{derczynski:2016}); and rumour stance detection on the RumourEval 2019 data set (\cite{rumour:19}). 
These tasks are chosen to vary in terms of domains, timescales, and the granularity of the linguistic units. 
In addition to evaluating SSA, we include two technical contributions as we extend the method both to allow for time series of unbounded length and to consider instance similarities between classes.
The best-performing SSA methods proposed here are semi-supervised, but require only between 2 and 10 annotated data points per class from the test year for successful alignment.
Crucially, the best proposed SSA models outperform baselines utilising more data, including the whole data set. 


\section{Subspace Alignment}
\begin{figure}
    \centering
    \includegraphics[width=.48\textwidth]{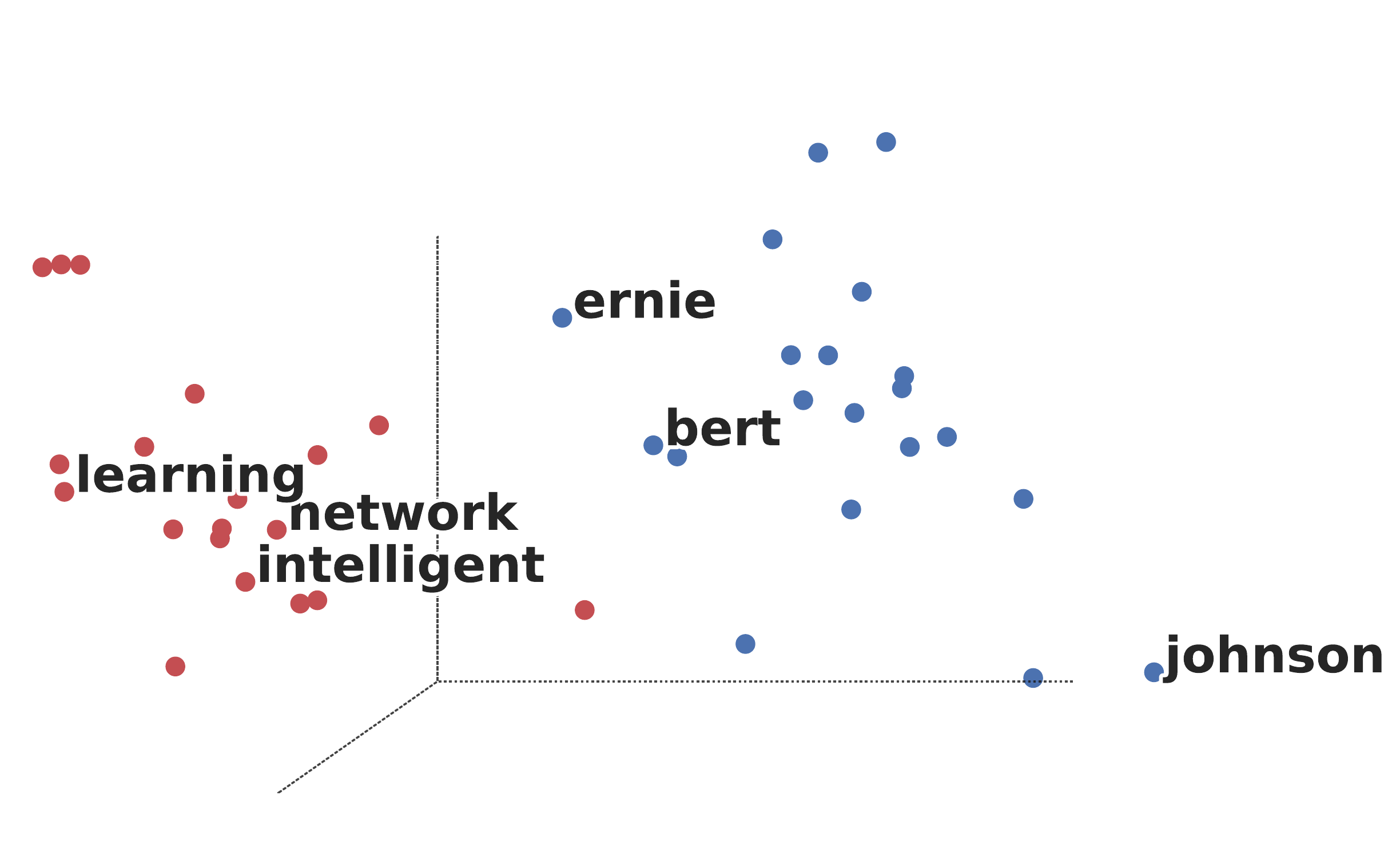}
    \includegraphics[width=.48\textwidth]{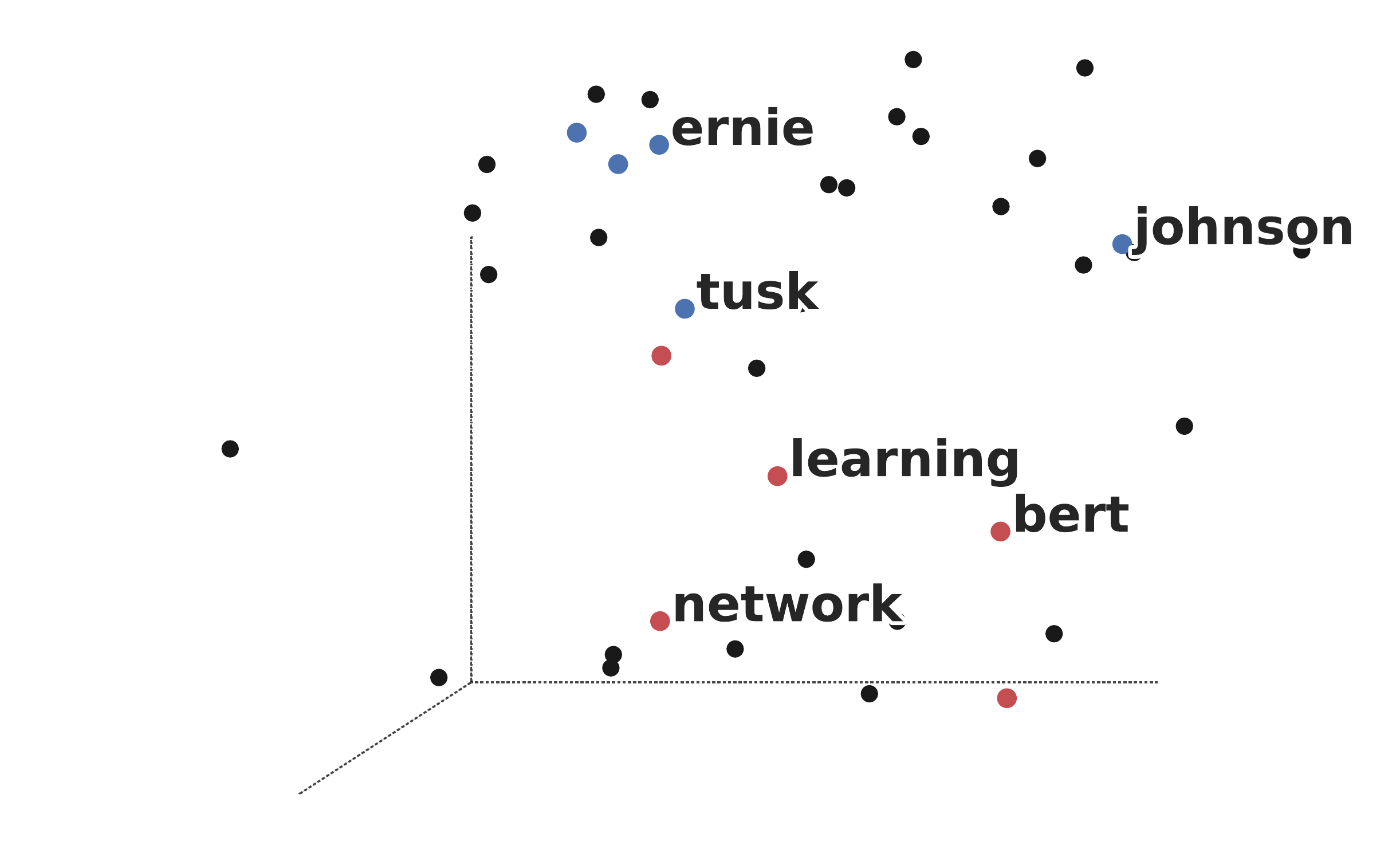}
    \caption{Example of a word embedding at $t_{2017}$ vs $t_{2018}$ (blue={\textsc{ person}}, red={\textsc{artefact}}, black={\textsc{unk}}). Source data (left, $t_{2017}$), target data (right, $t_{2018}$). Note that at $t_{2017}$, 'bert' is a {\textsc{person}}, while at $t_{2018}$, 'bert' is an {\textsc{artefact}}.}
    \label{fig:2DG}
\end{figure}

Suppose we embed words from a named entity recognition task, where {\textsc{artefact}}s should be distinguished from {\textsc{person}}s. Figure \ref{fig:2DG} shows scatterplots with data collected at two different time-points, say 2017 (top; source domain) and 2018 (bottom; target domain). Red points are examples of {\textsc{ artefact}}s  embedded in this space and blue points are examples of {\textsc{person}}s. We wish to classify the unknown points (black) from 2018 using the labeled points (red/blue bottom) from 2018 and the labeled points from 2017 (red/blue top).

As can be seen, the data from 2017 is not particularly relevant to classification of data from 2018, because the red and blue point clouds do not match. In other words, a classifier trained to discriminate red from blue in 2017 would make a lot of mistakes when applied directly to the data from 2018, partly because words such as 'bert' have changed from being {\textsc{person}}s to being {\textsc{artefact}}s. 
To make the source data from 2017 relevant -- and reap the benefits of having more data -- we wish to \emph{align} source and target data points.

\begin{figure}
    \centering
    \includegraphics[width=1.0\textwidth]{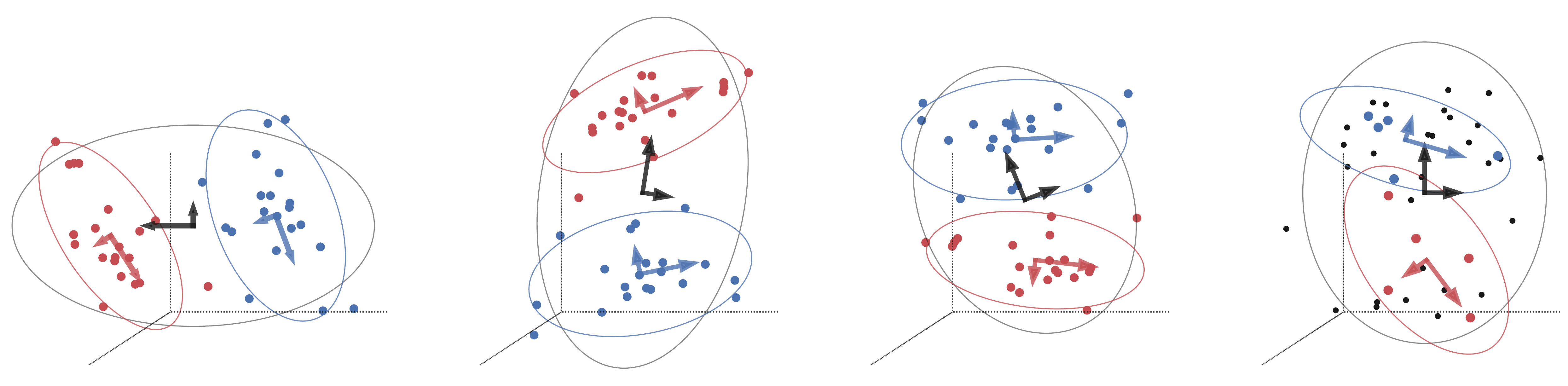}
    \caption{Illustration of subspace alignment procedures. Red vs blue dots indicate samples from different classes, arrows (black for total data and red vs blue for each class) indicate scaled eigenvectors of the covariance matrix (error ellipses indicate regions within 2 standard deviations). (Leftmost) Source data, fully labeled. (Left middle). Unsupervised subspace alignment: the total principal components from the source data (black arrows in leftmost figure) have been aligned to the total principal components of the target data (black arrows in rightmost figure). (Right middle) Semi-supervised subspace alignment: the class-specific principal components of the source data (red/blue arrows from leftmost figure) have been aligned to the class-specific components of the target data (red/blue arrows from the rightmost figure). Note that unsupervised alignment fails to match the red and blue classes across domains, while semi-supervised alignment succeeds. (Rightmost) Target data, with few labeled samples per class (black dots are unlabeled samples).}
    \label{fig:2DG_subalign}
\end{figure}

\subsection{Unsupervised Subspace Alignment}
Unsupervised alignment extracts a set of bases from each data set and transforms the source components such that they match the target components (\cite{fernando2013unsupervised}). Let $C_{\cal S}$ be the principal components of the source data $X_{t-1}$ and $C_{\cal T}$ be the components of the target data set $X_t$. The optimal linear transformation matrix is found by minimising the difference between the transformed source components and the target components:
\begin{align}
    M^{*} =& \ \underset{M}{\arg \min} \ \| C_{\cal S}M - C_{\cal T}\|^{2}_{F} \nonumber \\
    =& \ \underset{M}{\arg \min} \ \| C_{\cal S}^{\top} C_{\cal S}M - C_{\cal S}^{\top}C_{\cal T}\|^{2}_{F} \nonumber \\
    =& \ \underset{M}{\arg \min} \ \| M - C_{\cal S}^{\top}C_{\cal T}\|^{2}_{F} \ = \ C_{\cal S}^{\top} C_{\cal T} \label{eq:unsup_align} \, ,
\end{align}
where $\| \cdot \|_{F}$ denotes the Frobenius norm. Note that we left-multiplied both terms in the norm with the same matrix $C_{\cal S}^{\top}$ and that due to orthonormality of the principal components, $C_{\cal S}^{\top} C_{\cal S}$ is the identity and drops out. 
Source data $X_{t-1}$ is aligned to target data by first mapping it onto its own principal components and then applying the transformation matrix, $X_{t-1} C_{\cal S} M^{*}$. Target data $X_t$ is also projected onto its target components, $X_t C_{\cal T}$. 
The alignment is performed on the $d$ largest principal components, i.e.~a \emph{subspace} of the embedding. Keeping $d$ small avoids the high computational expense of eigendecomposition in high-dimensional data. 

Unsupervised alignment will only match the total structure of both data sets. Therefore, global shifts between domains can be accounted for, but not local shifts. Figure \ref{fig:2DG} is an example of a setting with local shifts, i.e. red and blue classes are shifted differently. Performing unsupervised alignment on this setting would fail. Figure  \ref{fig:2DG_subalign} (left middle) shows the source data (leftmost) aligned to the target data (rightmost) in an unsupervised fashion. Note that although the total data sets roughly match, the classes (red and blue ellipses) are not matched.

\subsection{Semi-Supervised Subspace Alignment}
In semi-supervised alignment, one performs subspace alignment \textit{per class}. As such, at least $1$ target label per class needs to be available. However, even then, with only $1$ target label per class, we would only be able to find $1$ principal component. To allow for the estimation of more components, we provisionally label all target samples using a $1$-nearest-neighbour classifier, starting from the given target labels. Using pseudo-labelled target samples, we estimate $d$ components.

Now, the optimal linear transformation matrix for each class can be found with an equivalent procedure as in Equation \ref{eq:unsup_align}:
\begin{align}
    M_{k}^{*} = \ \underset{M}{\arg \min} \ \| C_{{\cal S}, k} M - C_{{\cal T},_k} \|^{2}_{F} \ = \ C_{{\cal S},k}^{\top} C_{{\cal T},k} \label{eq:semisup_align} \, .
\end{align}
Afterwards, we transform the source samples of each class $X_{t-1}^k$ through the projection onto class-specific components $C_{{\cal S},k}$ and the optimal transformation: $X_{t-1}^{k} C_{{\cal S}, k} M_{k}^{*}$. Additionally, we centre each transformed source class on the corresponding target class. Figure \ref{fig:2DG_subalign} (right middle) shows the source documents transformed through semi-supervised alignment. Now, the classes match the target data classes.


\subsection{Extending SSA to Unbounded Time}
Semi-supervised alignment allows for aligning \emph{two} time steps, $t_1$ and $t_2$, to a joint space $t'_{1,2}$.
However, when considering a further alignment to another time step $t_3$, this can not trivially be mapped, since the joint space $t'_{1,2}$ necessarily has a lower dimensionality.
Observing that two independently aligned spaces, $t'_{1,2}$ and $t'_{2,3}$, \emph{do} have the same dimensionality, we further learn a new alignment between the two, resulting in the joint space of $t'_{1,2}$ and $t'_{2,3}$, namely $t''_{1,2,3}$.
While there are many ways of joining individual time steps to a single joint space, we approach this by building a binary branching tree, first joining adjacent timesteps with each other, and then joining the new adjacent subspaces with each other.

Although this is seemingly straight-forward, there is no guarantee that $t'_{1,2}$ and $t'_{2,3}$ will be coherent with one another, in the same way that two word embedding spaces trained with different algorithms might also differ in spite of having the same dimensionality.
This issue is partially taken care of by using semi-supervised alignment which takes class labels into account when learning the 'deeper' alignment $t''$.
We further find that it is beneficial to also take the similarities between samples into account when aligning.

\subsection{Considering Sample Similarities between Classes}

Since intermediary spaces, such as  $t'_{1,2}$ and $t'_{2,3}$, do not necessarily share the same semantic properties, we add a step to the semi-supervised alignment procedure.
Given that the initial unaligned spaces do encode similarities between instances, we run the $k$-means clustering algorithm ($k=5$) to give us some course-grained indication of instance similarities in the original embedding space.
This cluster ID is passed to SSA, resulting in an alignment which both attempts to match classes across time steps, in addition to instance similarities.
Hence, even though $t'_{1,2}$ and $t'_{2,3}$ are not necessarily semantically coherent, an alignment to $t''_{1,2,3}$ is made possible.

\section{Experimental Setup} \label{sec:experiments}
In the past year, several approaches to pre-training representations on language modelling based on transformer architectures (\cite{vaswani2017attention}) have been proposed.
These models essentially use a multi-head self-attention mechanism in order to learn representations which are able to attend directly to any part of a sequence. 
Recent work has shown that such contextualised representations pre-trained on language modelling tasks offer highly versatile representations which can be fine-tuned on seemingly any given task (\cite{peters-etal-2018-deep,devlin-etal-2019-bert,gpt,gpt2}).
In line with the recommendations from experiments on fine-tuning representations (\cite{ruder:freeze}), we use a frozen BERT to extract a consistent task-agnostic representation. Using a frozen BERT with subsequent subspace alignment allows us to avoid re-training a neural network each time-step while still working in an embedding learned by a neural language model. It also allows us to test the effectiveness of SSA without the confounding influence of representation updates.

\paragraph{Three Tasks.}
We consider three tasks representing a broad selection of natural language understanding scenarios: paper acceptance prediction based on the PeerRead data set (\cite{kang18naacl}), Named Entity Recognition (NER) based on the Broad Twitter Corpus (\cite{derczynski:2016}), and author stance prediction based on the RumourEval-2019 data set (\cite{rumour:19}).
These tasks were chosen so as to represent i) different textual domains, across ii) differing time scales, and iii) operating at varying levels of linguistic granularity.
As we are dealing with dynamical learning, the vast majority of NLP data sets can unfortunately not be used since they do not include time stamps.

\section{Paper Acceptance Prediction}
\label{sec:peerread}

\begin{table}
    \centering
\fontsize{10}{10}\selectfont
    \begin{tabular}{lrrr|rr|rrr}
    \toprule
    \textbf{Test year} & \textbf{All} & \textbf{Same} & \textbf{Prev} & \textbf{Unsup.} & \textbf{Semi-sup.} & \textbf{Unsup. Unb.} & \textbf{S. Unb.} & \textbf{S. Unb. w/Clst} \\
    \midrule
    2010      & 61.77   & 67.64 & 35.29 & \textbf{70.59} & \textbf{70.59} & 70.58 &  \textbf{70.59}  & \textbf{70.59} \\
    2011      & 61.77   & 58.82 & 55.88 & 14.71          & \textbf{72.35} & 24.71 &  \textbf{72.35}  & \textbf{72.35} \\
    2012      & 56.25   & 56.25 & 58.75 & 50.00          & 72.50 & 45.00 &  \textbf{72.80}  & 72.30 \\
    2013      & 67.54   & 56.14 & 58.78 & 76.31          & 78.07 & 72.31 &  78.97  & \textbf{79.03} \\
    2014      & 50.53   & 51.64 & 51.64 & 36.88          & 68.03 & 31.88 &  69.03  & \textbf{69.45}      \\
    2015      & 57.83   & 54.05 & 54.05 & 49.19          & 58.37 & 41.19 &  \textbf{59.97}  & 59.93      \\
    2016      & 58.89   & 57.36 & 57.36 & 50.61          & 61.04 & 38.61 &  \textbf{63.04}  & \textbf{63.04}      \\
    2017      & 56.04   & 58.24 & 58.24 & 68.13 & 63.73          & 58.13 &  68.73  & \textbf{69.80}      \\
    \midrule
    avg       & 58.82   & 57.52 & 53.75 & 52.05          & 68.09 & 47.80 & 69.44 & \textbf{69.56} \\
    \bottomrule
    \end{tabular}
    \caption{Paper acceptance prediction (acc.) on the PeerRead data set (\cite{kang18naacl}). Abbreviations represent Unsupervised, Semi-supervised, Unsupervised Unbounded, Semi-supervised Unbounded, and Semi-supervised Unbounded with Clustering.}
    \label{tab:acceptance_bert}
\end{table}

The PeerRead data set contains papers from ten years of arXiv history, as well as papers and reviews from major AI and NLP conferences (\cite{kang18naacl}).\footnote{\url{https://github.com/allenai/PeerRead}} 
From the perspective of evaluating our method, the arXiv sub-set of this data set offers the possibility of evaluating our method while adapting to ten years of history.
This is furthermore the only subset of the data annotated with both timestamps and with a relatively balanced accept/reject annotation.\footnote{The NIPS selection, ranging from 2013-2017, only contains accepted papers. The other conferences contain accept/reject annotation, but only represent single years.}
As arXiv naturally contains both accepted and rejected papers, this acceptance status has been assigned based on \cite{sutton:2017} who match arXiv submissions to bibliographic entries in DBLP, and additionally defining acceptance as having been accepted to major conferences, and not to workshops.
This results in a data set of nearly 12,000 papers, from which we use the raw abstract text as input to our system. The first three years were filtered out due to containing very few papers.
We use the standard train/test splits supplied with the data set.

\cite{kang18naacl} show that it is possible to predict paper acceptance status at major conferences at above baseline levels.
Our intuition in applying SSA to this problem, is that the topic of a paper is likely to bias acceptance to certain conferences \textit{across time}.
For instance, it is plausible that the likelihood of a neural paper being accepted to an NLP conference before and after 2013 differs wildly.
Hence, we expect that our model will, to some extent, represent the topic of an article, and that this will lend itself nicely to SSA.

\subsection{Model}
\begin{figure}
    \centering
    \includegraphics[width=0.75\columnwidth]{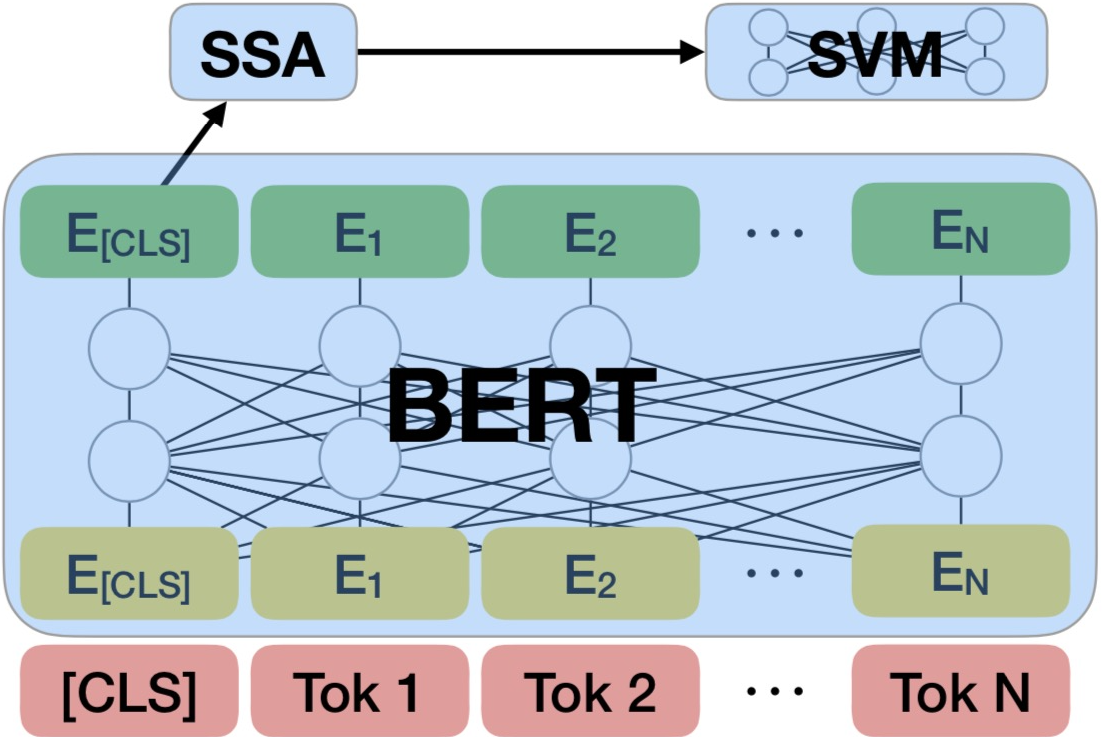}
    \caption{Paper acceptance model (BERT and SSA).}
    \label{fig:bertseq}
\end{figure}

We use the pre-trained \textsc{bert-base-uncased} model as the base for our paper acceptance prediction model.
Following the approach of \cite{devlin-etal-2019-bert}, we take the final hidden state (i.e., the output of the transformer) corresponding to the special \texttt{[CLS]} token of an input sequence to be our representation of a paper, as this has aggregated information through the sequence (Figure~\ref{fig:bertseq}).
This gives us a $d$-dimensional representation of each document, where $d=786$.
In all of the experiments for this task, we train an SVM with an RBF kernel on these representations, either with or without SSA depending on the setting.

\subsection{Experiments \& Results}
We set up a series of experiments where we observe past data, and evaluate on present data. 
We compare both unsupervised and semi-supervised subspace alignment, with several strong baselines.
The baselines represent cases in which we have access to more data, and consist of training our model on either \textbf{all} data (i.e.~both past and future data), on the \textbf{same} year as the evaluation year, and on the \textbf{previous} year.
In our alignment settings, we only observe data from the previous year, and apply subspace alignment.
This is a different task than presented by \citeauthor{kang18naacl}, as we evaluate paper acceptance for papers in the present. Hence, our scores are not directly comparable to theirs.

\begin{figure}
    \centering
    \includegraphics[width=0.8\columnwidth]{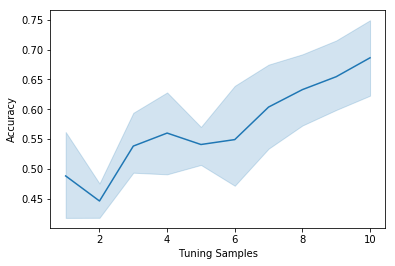}
    \caption{Tuning semi-supervised subspace alignment on PeerRead development data (95\% CI shaded).}
    \label{fig:hyperparam}
\end{figure}

One parameter which significantly influences performance, is the number of labelled data points we use for learning the semi-supervised subspace alignment.
We tuned this hyperparameter on the development set, finding an increasing trend. Using as few as 2 tuning points per class yielded an increase in performance in some cases (Figure~\ref{fig:hyperparam}).

Our results are shown in Table~\ref{tab:acceptance_bert}, using 10 tuning samples per class.
With unsupervised subspace alignment, we observe relatively unstable results -- in one exceptional case, namely testing on 2010, unsupervised alignment is as helpful as semi-supervised alignment.
Semi-supervised alignment, however, yields consistent improvements in performance across the board.
It is especially promising that adapting from past data outperforms training on all available data, as well as training on the actual in-domain data.
This highlights the importance of controlling for data drift due to language evolution. It shows that this signal can be taken advantage of to increase performance on present data with only a small amount of annotated data.
We further find that using several past time steps in the Unbounded condition is generally helpful, as is using instance similarities in the alignment.

\section{Named Entity Recognition}

\begin{table}
    \centering
    \fontsize{10}{10}\selectfont
    \begin{tabular}{lrrr|rr|rrr}
    \toprule
    \textbf{Test year} & \textbf{All} & \textbf{Same} & \textbf{Prev} & \textbf{Unsup.} & \textbf{Semi-sup.} & \textbf{Unsup. Unb.} & \textbf{S. Unb.} & \textbf{S. Unb. w/Clst} \\
    \midrule
    2013      & 62.95  & 42.24 & 54.16 & 42.25  & 63.82 & 42.25  & 63.82 & \textbf{63.95} \\
    2014      & 72.77  & 77.76 & 59.53 & 50.43  & 73.67 & 50.43  & 73.67 & \textbf{78.75} \\
    \midrule
    avg       & 67.86  & 60.00 & 56.85 & 46.34 & 68.75  & 46.34 & 68.75 &  \textbf{71.35} \\
    \bottomrule
    \end{tabular}
    \caption{NER (F1 score) on the Broad Twitter Corpus (\cite{derczynski:2016}).}
    \label{tab:NER}
\end{table}

The Broad Twitter Corpus contains tweets annotated with named entities, collected between the years 2009 and 2014 (\cite{derczynski:2016}).
However, as only a handful of tweets are collected before 2012, we focus our analysis on the final three years of this period (i.e.~two test years).
The corpus includes diverse data, annotated in part via crowdsourcing and in part by experts.
The inventory of tags in their tag scheme is relatively small, including Person, Location, and Organisation.
To the best of our knowledge no one has evaluated on this corpus either in general or per year, and so we cannot compare with previous work.

In the case of NER, we expect the adaptation step of our model to capture the fact that named entities may change their meaning across time (e.g. the example with ``Bert'' and ``BERT'' in Figure \ref{fig:2DG}).
This is related to work showing temporal drift of topics (\cite{wang2006topics}).

\subsection{Model}
Since casing is typically an important feature in NER, we use the pre-trained \textsc{bert-base-cased} model as our base for NER.
For each token, we extract its contextualised representation from BERT, before applying SSA.
As \cite{devlin-etal-2019-bert} achieve state-of-the-art results without conditioning the predicted tag sequence on surrounding tags (as would be the case with a CRF, for example), we also opt for this simpler architecture.
The resulting contextualised representations are therefore passed to an MLP with a single hidden layer (200 hidden units, ReLU activation), before predicting NER tags.
We train the MLP over 5 epochs using the Adam optimiser (\cite{kingma2014adam}).

\subsection{Experiments \& Results}
As with previous experiments, we compare unsupervised and semi-supervised subspace alignment with baselines corresponding to using all data, data from the same year as the evaluation year, and data from the previous year.
For each year, we divide the data into 80/10/10 splits for training, development, and test.
Results on the two test years 2013 and 2014 are shown in Table~\ref{tab:NER}.
In the case of NER, we do not observe any positive results for unsupervised subspace alignment.
In the case of semi-supervised alignment, however, we find increased performance as compared to training on the previous year, and compared to training on all data.
This shows that learning an alignment from just a few data points can help the model to generalise from past data.
However, unlike our previous experiments, results are somewhat better when given access to the entire set of training data from the test year itself in the case of NER.
The fact that training on only 2013 and evaluating on the same year does not work well can be explained by the fact that the amount of data available for 2013 is only 10\% of that for 2012.
The identical results for the unbounded extension is because aligning from a single time step renders this irrelevant.

\section{SDQC Stance Classification}
The RumourEval-2019 data set consists of roughly 5500 tweets collected for 8 events surrounding well-known incidents, such as the Charlie Hebdo shooting in Paris (\cite{rumour:19}).\footnote{\url{http://alt.qcri.org/semeval2019/index.php?id=tasks}} 
Since the shared task test set is not available, we split the training set into a training, dev and test part based on rumours (one rumour will be training data with a 90/10 split for development and another rumour will be the test data, with a few samples labelled). For Subtask A, tweets are annotated with stances, denoting whether it is in the category \texttt{Support}, \texttt{Deny}, \texttt{Query}, or \texttt{Comment} (SDQC). 

Each rumour only lasts a couple of days, but the total data set spans years, from August 2014 to November 2016.  We regard each rumour as a time-step and adapt from the rumour at time $t$-$1$ to the rumour at time $t$. We note that this setting is more difficult than the previous two due to the irregular time intervals. We disregard the rumour \texttt{ebola-essien} as it has too few samples per class.

\subsection{Model}
For this task, we use the same modelling approach as described for paper acceptance prediction. 
This method is also suitable here, since we simply require a condensed representation of a few sentences on which to base our temporal adaptation and predictions.
In the last iteration of the task, the winning system used hand-crafted features to achieve a high performance (\cite{kochkina-etal-2017-turing}). Including these would complicate SSA, so we opt for this simpler architecture instead.
We use the shorter time-scale of approximately weeks rather than years as rumours can change rapidly (\cite{kwon2017rumor}).

\subsection{Experiments \& Results}
In this experiment, we start with the earliest rumour and adapt to the next rumour in time. As before, we run the following baselines: training on all available labelled data (i.e.~all previous rumours and the labelled data for the current rumour), training on the labelled data from the current rumour (designated as `same') and training on the labelled data from the previous rumour. We perform both unsupervised and semi-supervised alignment using data from the previous rumour. We label 5 samples per class for each rumour.

In this data set, there is a large class imbalance, with a large majority of \texttt{comment} tweets and few \texttt{support} or \texttt{deny} tweets. To address this, we over-sample the minority classes. Afterwards, a SVM with RBF is trained and we test on unlabelled tweets for the current rumour. 
Table~\ref{tab:SDQC} shows the performance of the baselines and the two alignment procedures. As with the previous tasks, semi-supervised alignment generally helps, except for in the \texttt{charliehebdo} rumour.

\begin{table}
    \centering
    \fontsize{10}{10}\selectfont
    \begin{tabular}{lrrr|rr|rrr}
    \toprule
       \textbf{Test year} & \textbf{All} & \textbf{Same} & \textbf{Prev} & \textbf{Unsup.} & \textbf{Semi-sup.} & \textbf{Unsup. Unb.} & \textbf{S. Unb.} & \textbf{S. Unb. w/Clst} \\
    \midrule
    \texttt{ottawashooting}     & 31.51  & 23.67  & 30.77  & 30.77  & \textbf{31.88}  & 28.37       & 30.68 & 30.88 \\
    \texttt{prince-toronto}     & 36.27  & 23.37  & 34.46  & 34.46  & \textbf{40.32}  & 31.36       & 39.12 & 39.52 \\
    \texttt{sydney-siege}       & 32.34  & 27.17  & 41.23  & 41.23  & \textbf{43.60}  & 33.23       & 43.50 & 43.54 \\
\texttt{charliehebdo}       & 38.51  & 31.67  & \textbf{35.73}  & \textbf{35.73}  & 33.76  & 33.71  & 32.70 & 32.61 \\
    \texttt{putinmissing}       & 28.33  & 22.38  & 34.53  & 34.53  & \textbf{36.11}  & 31.95       & 35.10 & 35.81 \\
    \texttt{germanwings-crash}  & 29.38  & 22.01  & 44.79  & 44.79  & \textbf{44.84}  & 40.30       & 44.88 & 44.80 \\
    \texttt{illary}             & 29.24  & 25.81  & 37.53  & 37.53 & \textbf{40.08}   & 34.10       & 39.30 & 38.95 \\
    \midrule
    avg & 31.13 & 25.16 & 37.00 & 37.00 & \textbf{38.65} & 33.29 & 37.90 & 38.02 \\
    \bottomrule
    \end{tabular}
    \caption{F1 score in SDQC task of RumourEval-2019 (\cite{rumour:19})\label{tab:SDQC}}
\end{table}

\section{Analysis and Discussion} \label{sec:discussion}
We have shown that sequential subspace alignment is useful across natural language processing tasks. 
For the PeerRead data set we were particularly successful. This might be explained by the fact that the topic of a paper is a simple feature for SSA to pick up on, while being predictive of a paper's acceptance chances.
For NER, on the other hand, named entities can change in less predictable ways across time, proving a larger challenge for our approach.
For SDQC, we were successful in cases where the tweets are nicely clustered by class. For instance, where both rumours are about terrorist attacks, many of the \texttt{support} tweets were headlines from reputable newspaper agencies. These agencies structure tweets in a way that is consistently dissimilar from \texttt{comments} and \texttt{queries}.

The effect of our unbounded time extension boosts results on the PeerRead data set, as the data stretches across a range of years.
In the case of NER, however, this extension is excessive as only two time steps are available.
In the case of SDQC, the lack of improvement could be due to the irregular time intervals, making it hard to learn consistent mappings from rumour to rumour.
Adding instance similarity clustering aids alignment, since considering sample similarities across classes is important over longer time scales.

\subsection{Example of Aligning Tweets}
Finally, we set up the following simplified experiment to investigate the effect of alignment on SDQC data. First, we consider the rumour \texttt{charliehebdo}, where we picked the following tweet:

\vspace{5px}
\noindent \fbox{
	\parbox{1.0\textwidth}{%
		\small
		\textbf{Support:}\\ 
		\texttt{France: 10 people dead after shooting at HQ of satirical weekly newspaper \#CharlieHebdo, according to witnesses <URL>}
	}
}
\vspace{5px}

\noindent It has been labeled to be in support of the veracity of the rumour. We will consider the scenario where we use this tweet and others involving the \texttt{charliehebdo} incident to predict author stance in the rumour \texttt{germanwings-crash}. Before alignment, the following 2 \texttt{germanwings-crash} tweets are among the nearest neighbours in the embedding space:

\vspace{5px}
\noindent \fbox{
	\parbox{1.0\textwidth}{%
    \small
   \textbf{Query:} \\
   \texttt{@USER @USER if they had, it’s likely the descent rate would’ve been steeper and the speed not reduce, no ?}
	}
}
\vspace{5px}
\noindent \fbox{
	\parbox{1.0\textwidth}{%
    \small
    \textbf{Comment:} \\
   \texttt{@USER Praying for the families and friends of those involved in crash. I'm so sorry for your loss.}
}}

%

\noindent The second tweet is semantically similar (both are on the topic of tragedy), but the other is unrelated. Note that the news agency tweet differs from the comment and query tweets in that it stems from a reputable source, mentions details and includes a reference. After alignment, the \texttt{charliehebdo} tweet has the following 2 nearest neighbours:

\vspace{5px}
\noindent \fbox{
	\parbox{1.0\textwidth}{%
    \small
    \textbf{Support:} \\
    \texttt{“@USER: 148 passengers were on board \#GermanWings Airbus A320 which has crashed in the southern French Alps <URL>”}
}}
\vspace{5px}
\noindent \fbox{
	\parbox{1.0\textwidth}{%
    \small
    \textbf{Support:} \\
    \texttt{Report: Co-Pilot Locked Out Of Cockpit Before Fatal Plane Crash <URL> \#Germanwings <URL>}
}}
Now, both neighbours are of the \texttt{support} class. This example shows that semi-supervised alignment maps source tweets from one class close to target tweets of the same class. 

\subsection{Limitations}
A necessary assumption in subspace alignment is that classes are clustered in the embedding space: most embedded tokens should lie closer to \textit{other} embedded tokens of the \textit{same} class than to embedded tokens of another class. If this is not the case, then aligning based on a few labelled samples of class $k$ does not imply that the embedded source tokens are aligned to other target points of class $k$. This assumption is violated if, for instance, people only discuss one aspect of a rumour on day one and discuss several aspects of a rumour simultaneously on day two. One would observe a single cluster of token embeddings for supporters of the rumour initially and several clusters at a later time-step. 
Note that there is no unique solution for aligning a single cluster to multiple clusters.

Additionally, if those few samples labeled in the current time-step (for semi-supervised alignment) are falsely labeled or their label is ambiguous (e.g. a tweet that could equally be labeled as {\textsc query} or {\textsc deny}), then the source data could be aligned to the wrong point cloud. It is important that the few labeled tokens actually represent their classes. This is a common requirement in semi-supervised learning and is not specific to sequential alignment of text representations.

\subsection{Related Work}

The temporal nature of data can have a significant impact in natural language processing tasks. 
For instance, \cite{kutuzov2018diachronic} compare a number of approaches to diachronic word embeddings, and detection of semantic shifts across time.
For instance, such representations can be used to uncover changes of word meanings, or senses of new words altogether (\cite{gulordava2011distributional,heyer2009change,michel2011quantitative,mitra2014senses,wijaya2011understanding}).
Other work has investigated changes in the usage of parts of speech across time (\cite{mihalcea2012word}).
\cite{yao2018dynamic} investigate the changing meanings and associations of words across time, in the perspective of language change. By learning time-aware embeddings, they are able to outperform standard word representation learning algorithms, and can discover, e.g., equivalent technologies through time.
\cite{lukes:2018} show that lexical features can change their polarity across time, which can have a significant impact in sentiment analysis.
\cite{wang2006topics} show that associating topics with continuous distributions of timestamps yields substantial improvements in terms of topic prediction and interpretation of trends.
Temporal effects in NLP have also been studied in the context of scientific journals, for instance in the context of emerging themes and viewpoints (\cite{blei2006dynamic,sipos2012temporal}), and in terms of topic modelling on news corpora across time (\cite{allan2001temporal}).
Finally, in the context of rumour stance classification, \cite{lukasik-etal-2016-hawkes} show that temporal information as a feature in addition to textual content offers an improvement in results.
While this previous work has highlighted the extent to which language change across time is relevant for NLP, we present a concrete approach to taking advantage of this change. Nonetheless, these results could inspire more specialised forms of sequential adaptation for specific tasks.

Unsupervised subspace alignment has been used in computer vision to adapt between various types of representations of objects, such as high-definition photos, online retail images and illustrations (\cite{fernando2013unsupervised}). 
Alignment is not restricted to linear transformations, but can be made non-linear through kernelisation (\cite{aljundi2015landmarks}). 
An extension to semi-supervised alignment has been done for images (\cite{conf/cvpr/YaoPNLM15}), but not in the context of classification of text embeddings or domain adaptation on a sequential basis. 

\section{Conclusions}

In this paper, we introduced sequential subspace alignment (SSA) for natural language processing (NLP), which allows for improved generalisation from past to present data. Experimental evidence shows that this method is useful across diverse NLP tasks, in various temporal settings ranging from weeks to years, and for word-level and document-level representations. The best-performing SSA method, aligning sub-spaces in a semi-supervised way, outperforms simply training on all data with no alignment.

\section*{Acknowledgements}
WMK was supported by the Niels Stensen Fellowship. 


\chapter{A Diagnostic Study of Explainability Techniques}\label{ch:Diagnostic}


\boxabstract{
Recent developments in machine learning have introduced models that approach human performance at the cost of increased architectural complexity. Efforts to make the rationales behind the models' predictions transparent have inspired an abundance of new explainability techniques. Provided with an already trained model, they compute saliency scores for the words of an input instance. However, there exists no definitive guide on (i) how to choose such a technique given a particular application task and model architecture, and (ii) the benefits and drawbacks of using each such technique. In this paper, we develop a comprehensive list of \propertyplural{} for evaluating existing explainability techniques. We then employ the proposed list to compare a set of diverse explainability techniques on downstream text classification tasks and neural network architectures. We also compare the saliency scores assigned by the explainability techniques with human annotations of salient input regions to find relations between a model's performance and the agreement of its rationales with human ones. Overall, we find that the gradient-based explanations perform best across tasks and model architectures, and we present further insights into the properties of the reviewed explainability techniques.
}\blfootnote{\fullcite{atanasova-etal-2020-diagnostic}}

\section{Introduction}
\begin{figure}
\centering
\includegraphics[width=0.7\textwidth]{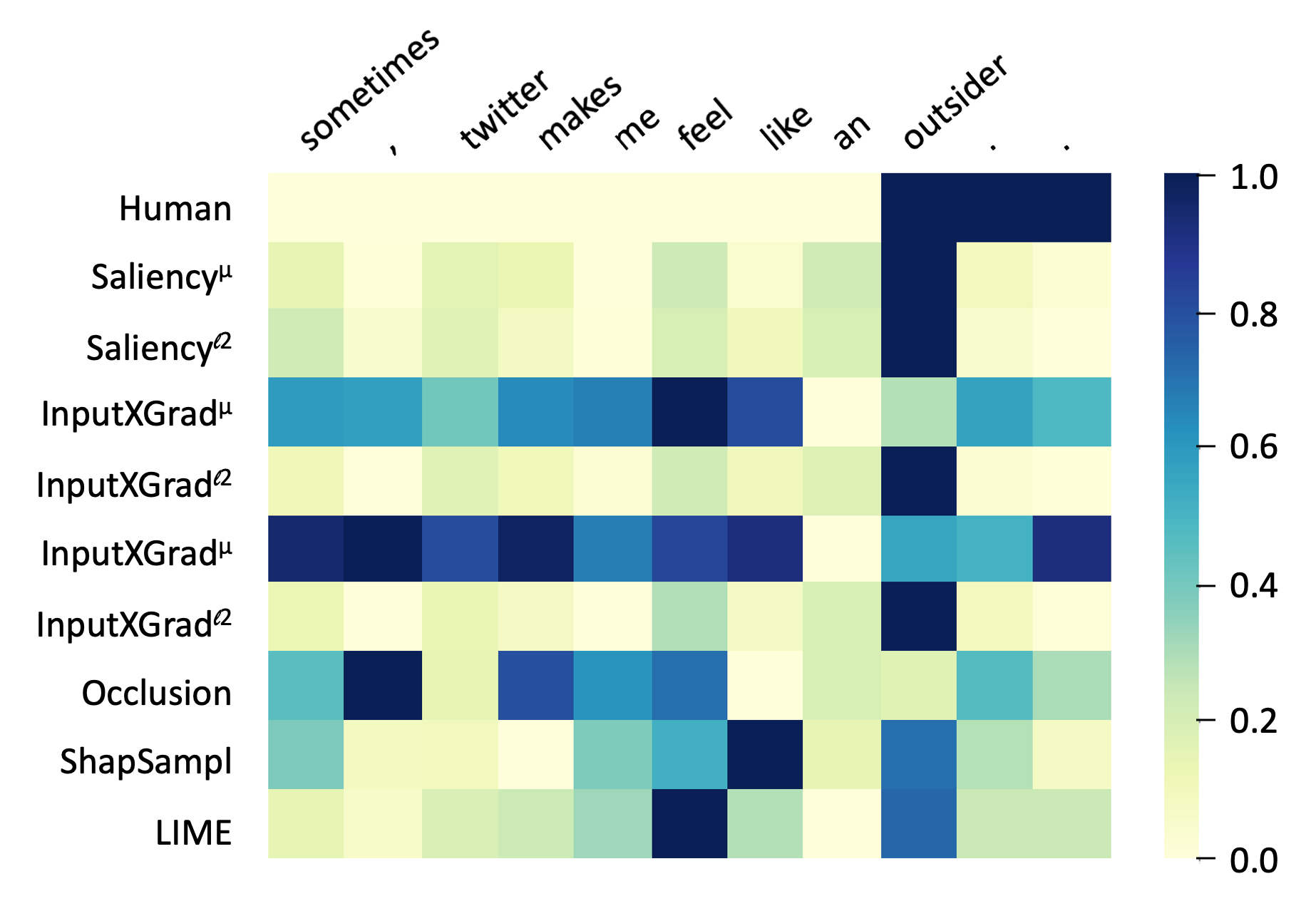}
\caption{Example of the saliency scores for the words (columns) of an instance from the Twitter Sentiment Extraction dataset. They are produced by the explainability techniques (rows) given a \trans{} model. The first row is the human annotation of the salient words. The scores are normalized in the range $[0, 1]$.}
\label{fig:diagnostic_example}
\end{figure}
Understanding the rationales behind models' decisions is becoming a topic of pivotal importance, as both the architectural complexity of machine learning models and the number of their application domains increases. Having greater insight into the models' reasons for making a particular prediction has already proven to be essential for discovering potential flaws or biases in medical diagnosis (\cite{caruana2015intelligible}) and judicial sentencing (\cite{rich2016machine}). In addition, European law has mandated ``the right $\dots$ to obtain an explanation of the decision reached'' ~(\cite{regulation2016regulation}).

\textit{Explainability methods} attempt to reveal the reasons behind a model's prediction for a single data point, as shown in Figure~\ref{fig:diagnostic_example}. They can be produced post-hoc, i.e., with already trained models. Such post-hoc explanation techniques can be applicable to one specific model~(\cite{barakat2007rule, wagner2019interpretable}) or to a broader range thereof~(\cite{ribeiromodel, lundberg2017unified}). 
They can further be categorised as: employing model gradients (\cite{sundararajan2017axiomatic, Simonyan2013DeepIC}), being perturbation based (\cite{shapley1953value, zeiler2014visualizing}) or providing explanations through model simplifications (\cite{ribeiromodel, johansson2004truth}). There also exist explainability methods that generate textual explanations~(\cite{NIPS2018_8163}) and are trained post-hoc or jointly with the model at hand. 

While there is a growing amount of explainability methods, we find that they can produce varying, sometimes contradicting explanations, as illustrated in Figure~\ref{fig:diagnostic_example}.
Hence, it is important to \emph{assess existing techniques} and to \emph{provide a generally applicable and automated methodology} for choosing one that is suitable for a particular model architecture and application task~(\cite{jacovi2020towards}). 
\cite{robnik2018perturbation} compiles a list of property definitions for explainability techniques, but it remains a challenge to evaluate them in practice. Several other studies have independently proposed different setups for probing varied aspects of explainability techniques~(\cite{deyoung2019eraser, sundararajan2017axiomatic}).
However, existing studies evaluating explainability methods are discordant and do not compare to properties from previous studies. In our work, we consider properties from related work and extend them to be applicable to a broader range of downstream tasks.

Furthermore, to create a thorough setup for evaluating explainability methods, one should include at least: (i) different groups of explainability methods (explanation by simplification, gradient-based, etc.), (ii) different downstream tasks, and (iii) different model architectures. However, existing studies usually consider at most two of these aspects, thus providing insights tied to a specific setup.  

We propose a number of \propertyplural{} for explainability methods and evaluate them in a comparative study. We consider explainability methods from different groups, all widely applicable to most ML models and application tasks. We conduct an evaluation on three text classification tasks, which contain human annotations of salient tokens. Such annotations are available for Natural Language Processing (NLP) tasks, as they are relatively easy to obtain. This is in contrast to ML sub-fields such as image analysis, for which we only found one relevant dataset -- 536 manually annotated object bounding boxes for Visual Question Answering~(\cite{subramanian2020obtaining}). 

We further compare explainability methods across three of the most widely used model architectures -- \cnn, \lstm, and \trans{}. The \trans{} model achieves state-of-the-art performance on many text classification tasks but has a complex architecture, hence methods to explain its predictions are strongly desirable. The proposed properties can also be directly applied to Machine Learning (ML) subfields other than NLP. The code for the paper is publicly available.\footnote{https://github.com/copenlu/xai-benchmark} \\In summary, the \textbf{contributions} of this work are:
\begin{itemize}[noitemsep]
\item We compile a comprehensive list of \propertyplural{} for explainability and automatic measurement of them, allowing for their effective assessment in practice.
\item We study and compare the characteristics of different groups of explainability techniques in three different application tasks and three different model architectures.
\item We study the attributions of the explainability techniques and human annotations of salient regions to compare and contrast the rationales of humans and machine learning models. 
\end{itemize}

\section{Related Work}
Explainability methods can be divided into explanations by simplification, e.g., LIME (\cite{ribeiromodel}); gradient-based explanations (\cite{sundararajan2017axiomatic}); perturbation-based explanations (\cite{shapley1953value, zeiler2014visualizing}). Some studies propose the generation of text serving as an explanation, e.g., \cite{NIPS2018_8163,lei2016rationalizing,atanasova-etal-2020-generating-fact}. For extensive overviews of existing explainability approaches, see \cite{BARREDOARRIETA202082}.

Explainability methods provide explanations of different qualities, so assessing them systematically is pivotal. A common attempt to reveal shortcomings in explainability techniques is to reveal a model's reasoning process with counter-examples (\cite{alvarez2018robustness, kindermans2019reliability,atanasova-etal-2020-generating}), finding different explanations for the same output. However, single counter-examples do not provide a measure to evaluate explainability techniques~(\cite{jacovi2020towards}).

Another group of studies performs human evaluation of the outputs of explainability methods (\cite{Lertvittayakumjorn2019HumangroundedEO, narayanan2018humans}). Such studies exhibit low inter-annotator agreement and reflect mostly what appears to be reasonable and appealing to the annotators, not the actual properties of the method.

The most related studies to our work design measures and properties of explainability techniques. \cite{robnik2018perturbation} propose an extensive list of properties. The \textit{Consistency} property captures the difference between explanations of different models that produce the same prediction; and the \textit{Stability} property measures the difference between the explanations of similar instances given a single model. We note that similar predictions can still stem from different reasoning paths. Instead, we propose to explore instance activations, which reveal more of the model's reasoning process than just the final prediction. The authors propose other properties as well, which we find challenging to apply in practice. We construct a comprehensive list of \propertyplural{} tied with measures that assess the degree of each characteristic.

Another common approach to evaluate explainability methods is to measure the sufficiency of the most salient tokens for predicting the target label~(\cite{deyoung2019eraser}). We also include a sufficiency estimate, but instead of fixing a threshold for the tokens to be removed, we measure the decrease of a model's performance, varying the proportion of excluded tokens. Other perturbation-based evaluation studies and measures exist~(\cite{sundararajan2017axiomatic, Adebayo:2018:SCS:3327546.3327621}), but we consider the above, as it is the most widely applied.

Another direction of explainability evaluation is to compare the agreement of salient words annotated by humans to the saliency scores assigned by explanation techniques (\cite{deyoung2019eraser}). We also consider the latter and further study the agreement across model architectures, downstream tasks, and explainability methods. 
While we consider human annotations at the word level (\cite{NIPS2018_8163, lei2016rationalizing}), there are also datasets (\cite{clark2019boolq,khashabi-etal-2018-looking}) with annotations at the sentence-level, which would require other model architectures, so we leave this for future work.

Existing studies for evaluating explainability heavily differ in their scope. Some concentrate on a \textbf{single model architecture} - BERT-LSTM (\cite{deyoung2019eraser}), RNN (\cite{arras-etal-2019-evaluating}), CNN (\cite{Lertvittayakumjorn2019HumangroundedEO}), whereas a few consider \textbf{more than one} model (\cite{guan2019towards, poerner-etal-2018-evaluating}). Some studies concentrate on one \textbf{particular dataset} (\cite{guan2019towards,arras-etal-2019-evaluating}), while only a few generalize their findings over \textbf{downstream tasks} (\cite{deyoung2019eraser, vashishth2019attention}). Finally, existing studies focus on one (\cite{vashishth2019attention}) or a single group of explainability methods (\cite{deyoung2019eraser, Adebayo:2018:SCS:3327546.3327621}). Our study is the first to propose a unified comparison of different groups of explainability techniques across three text classification tasks and three model architectures.


\section{Evaluating Attribution Maps}
We now define a set of \propertyplural{} of explainability techniques, and propose how to quantify them. Similar notions can be found in related work~(\cite{robnik2018perturbation, deyoung2019eraser}), and we extend them to be generally applicable to downstream tasks. We first introduce the prerequisite notation. 
Let $X = \{(x_i, y_i, w_i)|i \in [1,N]\}$ be the test dataset, where each instance consists of a 
list of \emph{tokens} $x_i$ = $\{x_{i,j}| j \in [1, |x_i|]\}$, a \emph{gold label} $y_i$, and a 
\emph{gold saliency score} for each of the tokens in $x_i$: $w_i = \{w_{i,j} | j \in [1, |x_i|]\}$ 
with each $w_{i,j} \in \{0, 1\}$. Let $\omega$ be an explanation technique that, given a model $M$,  a class $c$, and a single instance $x_i$, computes saliency scores for each token in the input: 
\salscores $ = \{\omega_{(i,j),c}^{M} |j \in [1, |x_i|]\}$. Finally, let $M = M_1, \dots M_K$ be models with the same architecture, each trained from a randomly chosen seed, and let $M' = M_1', \dots M_K'$ be models of the same architecture, but with randomly initialized weights.

\textbf{Agreement with human rationales (HA)}. This \property{} measures the degree of overlap between saliency scores provided by human annotators, specific to the particular task, and the word saliency scores computed by an explainability technique on each instance. The property is a simple way of approximating the quality of the produced feature attributions. While it does not necessarily mean that the saliency scores explain the predictions of a model, we assume that explanations with high agreement scores would be more comprehensible for the end-user as they would adhere more to human reasoning. With this \property{}, we can also compare how the type and the performance of a model and/or dataset affect the agreement with human rationales when observing one type of explainability technique. 

During evaluation, we provide an estimate of the average agreement of the explainability technique across the dataset. To this end, we start at the instance level and compute the Average Precision (AP) of produced saliency scores \salscores{} by comparing them to the gold saliency annotations $w_i$. Here, the label for computing the saliency scores is the gold label: $c=y_i$.
Then, we compute the average across all instances, arriving at Mean AP (MAP):
\begin{equation}
\textrm{MAP}(\omega, M, X) = \frac{1}{N}\sum \limits_{i \in [1, N]} AP(w_{i}, \omega_{x_i, y_i}^M)   
\end{equation}
\textbf{Confidence Indication (CI)}. A token from a single instance can receive several saliency scores, indicating its contribution to the prediction of each of the classes. Thus, when a model recognizes a highly indicative pattern of the predicted class $k$, the tokens involved in the pattern would have highly positive saliency scores for this class and highly negative saliency scores for the remaining classes. On the other hand, when the model is not highly confident, we can assume that it is unable to recognize a strong indication of any class, and the tokens accordingly do not have high saliency scores for any class. Thus, the computed explanation of an instance $i$ should indicate the confidence $p_{i,k}$ of the model in its prediction.

We propose to measure the predictive power of the produced explanations for the confidence of the model. We start by computing the Saliency Distance (SD) between the saliency scores for the predicted class $k$ to the saliency scores of the other classes $K/k$ (Eq.~\ref{eq:conf1}). Given the distance between the saliency scores, we predict the confidence of the class with logistic regression (LR) and finally compute the Mean Absolute Error -- MAE (Eq.~\ref{eq:conf2}), of the predicted confidence to the actual one.
\begin{gather}
\textrm{SD} = \sum \limits_{j \in [0,|x|]}D(\omega_{x_{i,j}, k}^{M}, \omega_{x_{i,j}, K/k}^{M}) \label{eq:conf1}\\
\textrm{MAE}(\omega, M, X) = \sum \limits_{\substack{i \in [1, N]}} |p_{i,k} - \textrm{LR}(\salmap)|~\label{eq:conf2}
\end{gather}
For tasks with two classes, D is the subtraction of the saliency value for class k and the other class. For more than two classes, D is the concatenation of the max, min, and average across the differences of the saliency value for class k and the other classes. Low MAE indicates that model's confidence can be easily identified by looking at the produced explanations.

\textbf{Faithfulness (F)}. Since explanation techniques are employed to explain model predictions for a single instance, an essential property is that they are faithful to the model's inner workings and not based on arbitrary choices. A well-established way of measuring this property is by replacing a number of the most-salient words with a mask token~(\cite{deyoung2019eraser}) and observing the drop in the model's performance. To avoid choosing an unjustified percentage of words to be perturbed, we produce several dataset perturbations by masking 0, 10, 20, \dots, 100\% of the tokens in order of decreasing saliency, thus arriving at $X^{\omega^0}$, $X^{\omega^{10}}$, \dots, $X^{\omega^{100}}$. Finally, to produce a single number to measure faithfulness, we compute the area under the threshold-performance curve (AUC-TP):
\begin{equation}
\begin{aligned}
\textrm{AUC-TP}(\omega, M, X) = \\ 
\textrm{AUC}([(i, P(M(X^{\omega^0}))-M(X^{\omega^i}))]) 
\end{aligned}
\end{equation}
where P is a task specific performance measure and $i \in [0, 10, \dots, 100]$. 
We also compare the AUC-TP of the saliency methods to a random saliency map to find whether there are explanation techniques producing saliency scores without any contribution over a random score. 

Using AUC-TP, we perform an ablation analysis
which is a good approximation of whether the most salient words are also the most important ones for a model's prediction. 
However, some prior studies~(\cite{feng2018pathologies}) find that models remain confident about their prediction even after stripping most input tokens, leaving a few that might appear nonsensical to humans. The \propertyplural\ that follow aim to facilitate a more in-depth analysis of the alignment between the inner workings of a model and produced saliency maps.

\textbf{Rationale Consistency (RC)}.
A desirable property of an explainability technique is to be consistent with the similarities in the reasoning paths of several models on a single instance. Thus, when two reasoning paths are similar, the scores provided by an explainability technique $\omega$ should also be similar, and vice versa. Note that we are interested in similar reasoning paths as opposed to similar predictions, as the latter does not guarantee analogous model rationales. For models with diverse architectures, we expect rationales to be diverse as well and to cause low consistency. Therefore, we focus on a set of models with the same architecture, trained from different random seeds as well as the same architecture, but with randomly initialized weights. The latter would ensure that we can have model pairs $(M_s, M_p)$ with similar and distant rationales. We further claim that the similarity in the reasoning paths could be measured effectively with the distance between the activation maps (averaged across layers and neural nodes) produced by two distinct models (Eq.~\ref{eq:consist1}). The distance between the explanation scores is computed simply by subtracting the two (Eq.~\ref{eq:consist2}). Finally, we compute  Spearman's $\rho$ between the similarity of the explanation scores and the similarity of the attribution maps (Eq.~\ref{eq:consist3}).

\begin{gather}
D(M_s, M_p, x_i) = D(M_s(x_i), M_p(x_i)) \label{eq:consist1}\\
D(M_s, M_p, x_i, \omega) = D(\omega_{x_i, y_i}^{M_s}, \omega_{x_i, y_i}^{M_p}) \label{eq:consist2} \\
\begin{split}
\rho(M_s, M_p, X, \omega) = \rho(D(M_s, M_p, x_i), \\
D(M_s, M_p, x_i, \omega)| i \in [1, N] ) \label{eq:consist3}
\end{split}
\end{gather}
The higher the positive correlation is, the more consistent the attribution method would be. We choose Spearman's $\rho$ as it measures the monotonic correlation between the two variables. On the other hand, Pearson's $\rho$ measures only the linear correlation, and we can have a non-linear correlation between the activation difference and the saliency score differences.
When subtracting saliency scores and layer activations, we also take the absolute value of the vector difference as the property should be invariant to order of subtraction.
An additional benefit of the property is that low correlation scores would also help to identify explainability techniques that are not faithful to a model's rationales.

\textbf{Dataset Consistency (DC)}. 
The next \property{} is similar to the above notion of rationale consistency but focuses on consistency across instances of a dataset as opposed to consistency across different models of the same architecture. In this case, we test whether instances with similar rationales also receive similar explanations. While Rationale Consistency compares instance explanations of the same instance for different model rationales, Dataset Consistency compares explanations for pairs of instances on the same model. We again measure the similarity between instances $x_i$ and $x_j$ by comparing their activation maps, as in Eq.~\ref{eq:consistdata1}. The next step is to measure the similarity of the explanations produced by an explainability technique $\omega$, which is the difference between the saliency scores as in Eq.~\ref{eq:consistdata2}. Finally, we measure Spearman's $\rho$  between the similarity in the activations and the saliency scores as in Eq.~\ref{eq:consistdata3}. We again take the absolute value of the difference.
\begin{gather}
D(M, x_i, x_j) = D(M(x_i), M(x_j)) \label{eq:consistdata1} \\
D(M, x_i, x_j, \omega) = D(\omega_{x_i, y_i}^{M}, \omega_{x_j, y_i}^{M})
\label{eq:consistdata2} \\
\begin{split}
\rho(M, X, \omega) = \rho(D(M, x_i, x_j), \\ D(M, x_i, x_j, \omega)| i, j \in [1, N])
\end{split} 
\label{eq:consistdata3}
\end{gather}

\section{Experiments}
\subsection{Datasets}
\begin{table}
\centering
\fontsize{10}{10}\selectfont
\begin{tabular}{p{20mm}p{70mm}p{28mm}p{22mm}}
\toprule
\textbf{Dataset} & \textbf{Example} & \textbf{Size} & \textbf{Length} \\ \midrule
e-SNLI \newline \cite{NIPS2018_8163} & \textit{Premise:} An adult dressed in black \textbf{holds a stick.} \newline \textit{Hypothesis:} An adult is walking away, \textbf{empty-handed}. \newline \textit{Label}: contradiction & 549 367 Train \newline 9 842 Dev \newline 9 824 Test & 27.4 inst. \newline 5.3 expl. \\ \midrule
Movie \newline Reviews \newline \cite{zaidan-etal-2007-using} \newline & \textit{Review:} he is one of \textbf{the most exciting martial artists on the big screen}, continuing to perform his own stunts and \textbf{dazzling audiences} with his flashy kicks and punches.\newline \textit{Class:} Positive & 1 399 Train \newline 199 Dev \newline 199 Test & 834.9 inst. \newline 56.18 expl. \\ \midrule
Tweet \newline Sentiment \newline Extraction \newline (TSE)~\footnotemark & 
\textit{Tweet:} im soo \textbf{bored}...im deffo missing my music channels \newline
\textit{Class:} Negative & 
21 983 Train \newline 2 747 Dev \newline 2 748 Test & 20.5 inst. \newline 9.99 expl. \\
\bottomrule
\end{tabular}
\caption{Datasets with human-annotated saliency explanations. The \textit{Size} column presents the dataset split sizes we use in our experiments. The \textit{Length} column presents the average number of instance tokens in the test set \textit{(inst.)} and the average number of human annotated explanation tokens \textit{(expl.)}.}
\label{tab:datasets}
\end{table}
\footnotetext{\urlx{https://www.kaggle.com/c/tweet-sentiment-extraction}}

Table~\ref{tab:datasets} provides an overview of the used datasets.
For e-SNLI, models predict inference -- contradiction, neutral, or entailment -- between sentence tuples. For the Movie Reviews dataset, models predict the sentiment -- positive, negative, or neutral -- of reviews with multiple sentences. Finally, for the TSE dataset, models predict tweets' sentiment -- positive, negative, or neutral.
The e-SNLI dataset provides three dataset splits with human-annotated rationales, which we use as training, dev, and test sets, respectively. The Movie Reviews dataset provides rationale annotations for nine out of ten splits. Hence, we use the ninth split as a test and the eighth split as a dev set, while the rest are used for training. Finally, the TSE dataset only provides rationale annotations for the training dataset, and we therefore  randomly split it into 80/10/10\% chunks for training, development and testing.

\subsection{Models}  
\begin{table}
\centering
\fontsize{10}{10}\selectfont
\begin{tabular}{lrr}
\toprule
\textbf{Model} & \textbf{Val} & \textbf{Test} \\ \midrule
\multicolumn{3}{c}{\textbf{e-SNLI}} \\
\trans & 0.897 ($\pm$0.002) & 0.892 ($\pm 0.002$) \\
\transrand & 0.167 ($\pm$0.003) & 0.167 ($\pm 0.003$)\\
\cnn & 0.773 ($\pm$0.003) & 0.768 ($\pm 0.002$)\\
\cnnrand & 0.195 ($\pm 0.038$) & 0.194 ($\pm 0.037$) \\
\lstm & 0.794 ($\pm$0.005) & 0.793 ($\pm 0.009$)\\
\lstmrand & 0.176 ($\pm 0.013$) & 0.176 ($\pm 0.000$) \\ \midrule

\multicolumn{3}{c}{\textbf{Movie Reviews}}\\
\trans & 0.859 ($\pm$0.044) & 0.856 ($\pm$0.018) \\
\transrand & 0.335 ($\pm$0.003)& 0.333 ($\pm 0.000$)\\
\cnn & 0.831 ($\pm$0.014) & 0.773 ($\pm$0.005)\\
\cnnrand & 0.343 ($\pm$0.020) & 0.333 ($\pm 0.001$) \\
\lstm & 0.614 ($\pm$0.017) & 0.567 ($\pm 0.019$)\\
\lstmrand & 0.362 ($\pm$0.030) & 0.363 ($\pm 0.041$) \\ \midrule

\multicolumn{3}{c}{\textbf{TSE}} \\
\trans & 0.772 ($\pm$0.005) & 0.781 ($\pm 0.009$) \\
\transrand &0.165 ($\pm$0.025) & 0.171 ($\pm 0.022$)\\
\cnn & 0.708 ($\pm$0.007) &  0.730 ($\pm 0.007$)\\
\cnnrand & 0.221 ($\pm$0.060) & 0.226 ($\pm 0.055$) \\
\lstm & 0.701 ($\pm$0.005) & 0.727 ($\pm 0.004$)\\
\lstmrand & 0.196 ($\pm$0.070) & 0.204 ($\pm 0.070$) \\
\bottomrule
\end{tabular}
\caption{Models' F1 score on the test and the validation datasets. The results present the average and the standard deviation of the Performance measure over five models trained from different seeds. The random versions of the models are again five models, but only randomly initialized, without training.}
\label{tab:modeleval}
\end{table}

We experiment with different commonly used base models, namely \cnn{}~(\cite{fukushima1980neocognitron}), \lstm{} ~(\cite{hochreiter1997long}), and the \trans{} ~(\cite{vaswani2017attention}) architecture BERT (\cite{devlin-etal-2019-bert}). The selected models allow for a comparison of the explainability techniques on diverse model architectures.  Table~\ref{tab:modeleval} presents the performance of the separate models on the datasets.

For the \cnn{} model, we use an embedding, a convolutional, a max-pooling, and a linear layer. The embedding layer is initialized with GloVe~(\cite{pennington2014glove}) embeddings and is followed by a dropout layer. The convolutional layer computes convolutions with several window sizes and multiple-output channels with ReLU~(\cite{hahnloser2000digital}) as an activation function. The result is compressed down with a max-pooling layer, passed through a dropout layer, and into a fine linear layer responsible for the prediction. The final layer has a size equal to the number of classes in the dataset.

The \lstm{} model again contains an embedding layer initialized with the GloVe embeddings. The embeddings are passed through several bidirectional LSTM layers. The final output of the recurrent layers is passed through three linear layers and a final dropout layer.

For the \trans{} model, we fine-tune the pre-trained basic, uncased language model (LM)~(\cite{Wolf2019HuggingFacesTS}). The fine-tuning is performed with a linear layer on top of the LM with a size equal to the number of classes in the corresponding task. Further implementation details for all of the models, as well as their F1 scores, are presented in~\ref{appendix:A}.

\subsection{Explainability Techniques}
We select the explainability techniques to be representative of different groups -- gradient~(\cite{sundararajan2017axiomatic, Simonyan2013DeepIC}), perturbation ~(\cite{shapley1953value, zeiler2014visualizing}) and simplification based~(\cite{ribeiromodel, johansson2004truth}). 

Starting with the \textbf{gradient-based} approaches, we select \textit{Saliency}~(\cite{Simonyan2013DeepIC}) as many other gradient-based explainability methods build on it. It computes the gradient of the output w.r.t. the input. We also select two widely used improvements of the \textit{Saliency} technique, namely \textit{InputXGradient}~(\cite{Kindermans2016InvestigatingTI}), and \textit{Guided Backpropagation}~(\cite{springenberg2014striving}). InputXGradient additionally multiplies the gradient with the input and \textit{Guided Backpropagation} overwrites the gradients of ReLU functions so that only non-negative gradients are backpropagated.

From the \textbf{perturbation-based} approaches, we employ \textit{Occlusion}~(\cite{zeiler2014visualizing}), which replaces each token with a baseline token (as per standard, we use the value zero) and measures the change in the output. Another popular perturbation-based technique is the \textit{Shapley Value Sampling}~(\cite{castro2009polynomial}). It is based on the Shapley Values approach that computes the average marginal contribution of each word across all possible word perturbations. The Sampling variant allows for a faster approximation of Shapley Values by considering only a fixed number of random perturbations as opposed to all possible perturbations.

Finally, we select the \textbf{simplification-based} explanation technique LIME~(\cite{ribeiromodel}). For each instance in the dataset, LIME trains a linear model to approximate the local decision boundary for that instance.

\textbf{Generating explanations.} 
The saliency scores from each of the explainability methods are generated for each of the classes in the dataset. As all of the gradient approaches provide saliency scores for the embedding layer (the last layer that we can compute the gradient for), we have to aggregate them to arrive at one saliency score per input token. As we found different aggregation approaches in related studies~(\cite{bansal2016ask, deyoung2019eraser}), we employ the two most common methods -- 
L2 norm and averaging (denoted as $\mu$ and $\ell2$ in the explainability method names). 
\section{Results and Discussion}
We report the measures of each \property{} as well as FLOPs as a measure of the computing time used by the particular method. For all \propertyplural, we also include the randomly assigned saliency as a baseline.

\subsection{Results}
\begin{table}
\centering
\fontsize{10}{10}\selectfont
\begin{tabular}{lrrr}
\toprule
\textbf{Saliency} & \textbf{e-SNLI} & \textbf{IMDB} & \textbf{TSE} \\
\midrule
\multicolumn{4}{c}{\trans}\\
\rand & 0.201 & 0.517 & 0.185  \\
\shapsamp & 0.479 & 0.481 & 0.667  \\
\lime & \textbf{0.809} & 0.604 & 0.553  \\
\occlusion & 0.523 & 0.323 & 0.556  \\
\salmean & 0.772 & 0.671 & \underline{0.707}  \\
\salnorm & 0.781 & \textbf{0.687} & 0.696   \\
\inputxmean & 0.364 & 0.432 & 0.307  \\
\inputxnorm & \underline{0.796} & \underline{0.676} & \textbf{0.754}  \\
\guidedmean & 0.468 & 0.236 & 0.287  \\
\guidednorm & 0.782 & \underline{0.676} & 0.685  \\
\midrule
\multicolumn{4}{c}{\cnn}\\
\rand & 0.209 & 0.468 & 0.384  \\
\shapsamp & 0.460 & 0.648 & 0.630  \\
\lime & 0.571 & 0.572 & \textbf{0.681} \\
\occlusion & 0.554 & 0.411 & 0.594  \\
\salmean & 0.853 & 0.712 & 0.595  \\
\salnorm & \underline{0.875} & \textbf{0.796} & 0.631  \\
\inputxmean & 0.576 & 0.662 & 0.613  \\
\inputxnorm & \textbf{0.881} & 0.759 & \underline{0.636}  \\
\guidedmean & 0.403 & 0.346 & 0.438  \\
\guidednorm & \underline{0.875} & \underline{0.788} & 0.628  \\
\midrule
\multicolumn{4}{c}{\lstm}\\
\rand & 0.166 & 0.343 & 0.225  \\
\shapsamp & 0.606 & 0.605 & 0.526  \\
\lime & 0.759 & 0.233 & 0.630 \\
\occlusion & 0.609 & 0.589 & 0.681  \\
\salmean & 0.795 & 0.568 & 0.702  \\
\salnorm & 0.800 & 0.583 & \textbf{0.704}  \\
\inputxmean & 0.432 & 0.481 & 0.441  \\
\inputxnorm & \textbf{0.820} & \textbf{0.685} & 0.693  \\
\guidedmean & 0.492 & 0.553 & 0.410  \\
\guidednorm & \underline{0.805} & \underline{0.660} & \textbf{0.720} \\
\bottomrule
\end{tabular}
\caption{Mean of the \property{} measures for all tasks and models. The best result for the particular model architecture and downstream task is in bold and the second-best is underlined.}
\label{tab:meanprop}
\end{table}
Of the three model architectures, unsurprisingly, the \trans\ model performs best, while the \cnn\ and the \lstm\ models are close in performance. It is only for the IMDB dataset that the \lstm\ model performs considerably worse than the \cnn, which we attribute to the fact that the instances contain a large number of tokens, as shown in Table~\ref{tab:datasets}. As this is not the core focus of this paper, detailed results can be found in the supplementary material.

\textbf{Overall results.} Table~\ref{tab:meanprop} presents the mean of all properties across tasks and models with all property measures normalized to be in the range [0,1]. 
We see that gradient-based explainability techniques always have the best or the second-best performance for the \propertyplural{} across all three model architectures and all three downstream tasks. Note that, \inputxmean{} and \guidedmean{}, which are computed with a mean aggregation of the scores, have some of the worst results. We conjecture that this is due to the large number of values that are averaged -- the mean smooths out any differences in the values. In contrast, the L2 norm aggregation amplifies the presence of large and small values in the vector. From the non-gradient based explainability methods, \lime{} has the best performance, where in two out of nine cases it has the best performance. It is followed by \shapsamp{} and \occlusion{}. We can conclude that the occlusion based methods overall have the worst performance according to the \propertyplural{}.

Furthermore, we see that the explainability methods achieve better performance for the e-SNLI and the TSE datasets with the \trans{} and \lstm{} architectures, whereas the results for the IMDB dataset are the worst. We hypothesize that this is due to the longer text of the input instances in the IMDB dataset. The scores also indicate that the explainability techniques have the highest \property{} measures for the \cnn{} model with the e-SNLI and the IMDB datasets, followed by the \lstm{}, and the \trans{} model. We suggest that the performance of the explainability tools can be worse for large complex architectures with a huge number of neural nodes, like the \trans{} one, and perform better for small, linear architectures like the \cnn{}.
\begin{figure}
\centering
\begin{subfigure}{.5\textwidth}
  \centering
  \includegraphics[width=215pt]{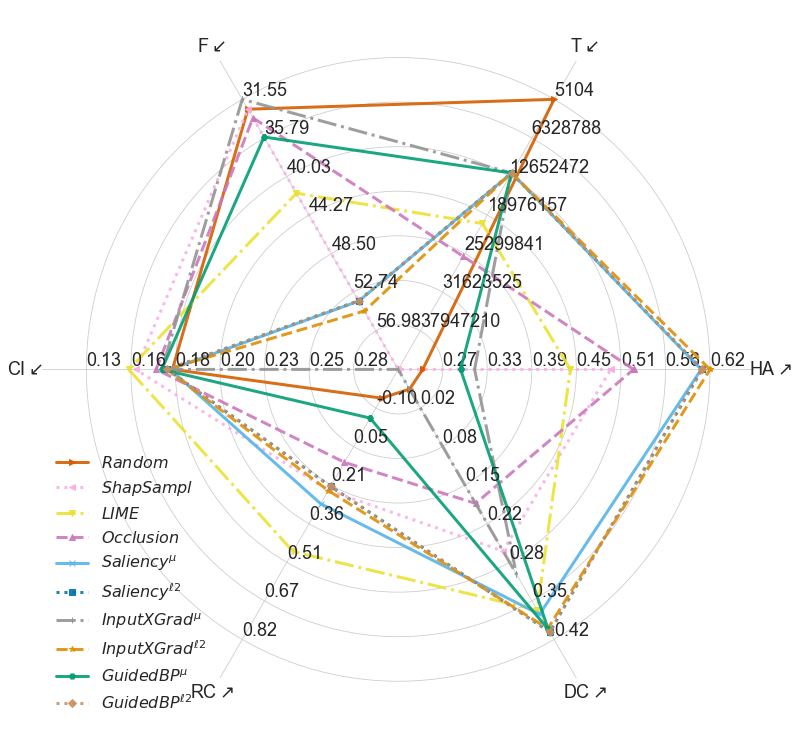}
  \caption{\trans}
  \label{fig:sub11}
\end{subfigure} 
\begin{subfigure}{.5\textwidth}
  \centering
  \includegraphics[width=215pt]{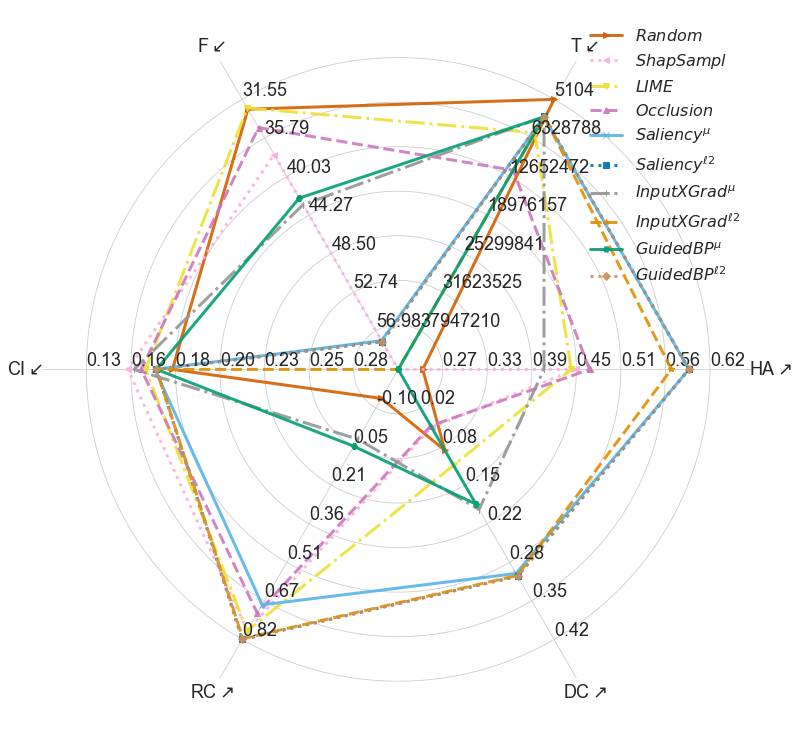}
  \caption{\cnn}
  \label{fig:sub12}
\end{subfigure}
\begin{subfigure}{.5\textwidth}
  \centering
  \includegraphics[width=215pt]{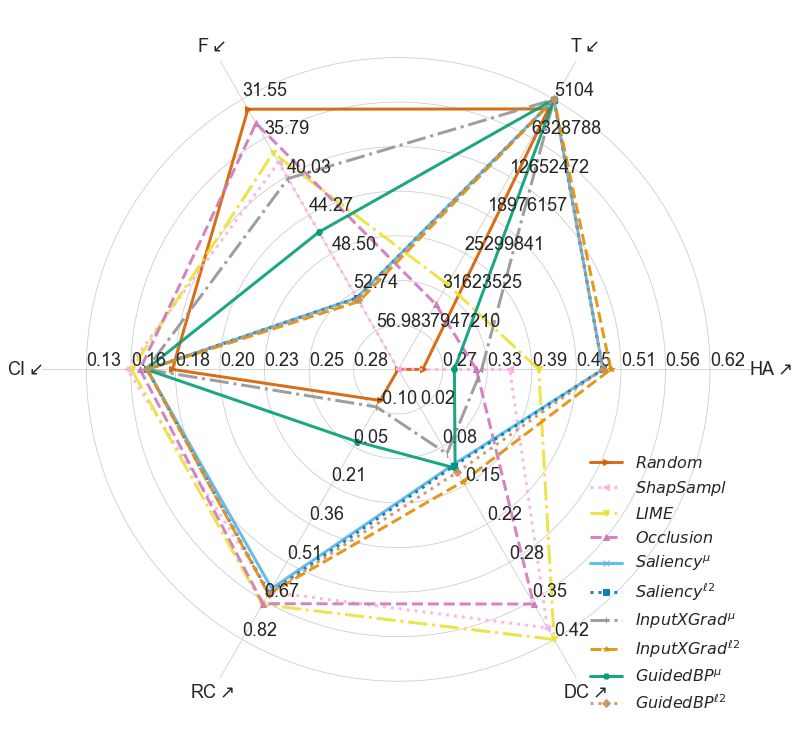}
  \caption{\lstm}
  \label{fig:sub13}
\end{subfigure}
\caption{Diagnostic property evaluation for 
all explainability techniques, on the e-SNLI dataset. 
The $\nearrow$ and $\swarrow$ signs indicate that higher, correpspondingly lower, values of the property measure are better.}
\label{fig:spider1}
\end{figure}

\begin{figure}
\centering
\begin{subfigure}{.5\textwidth}
  \centering
  \includegraphics[width=215pt]{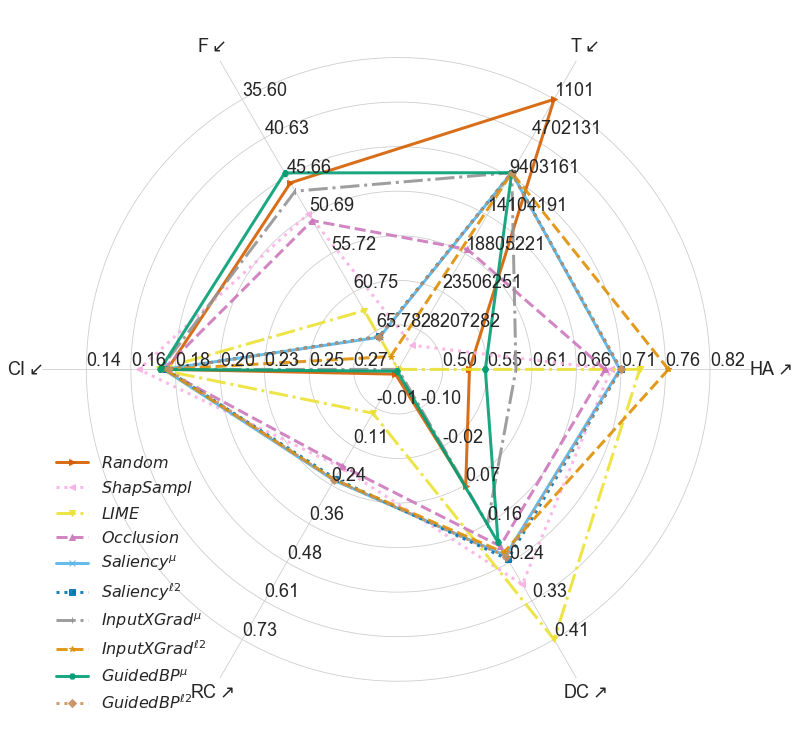}
  \caption{\trans}
  \label{fig:sub31}
\end{subfigure} 
\begin{subfigure}{.5\textwidth}
  \centering
  \includegraphics[width=215pt]{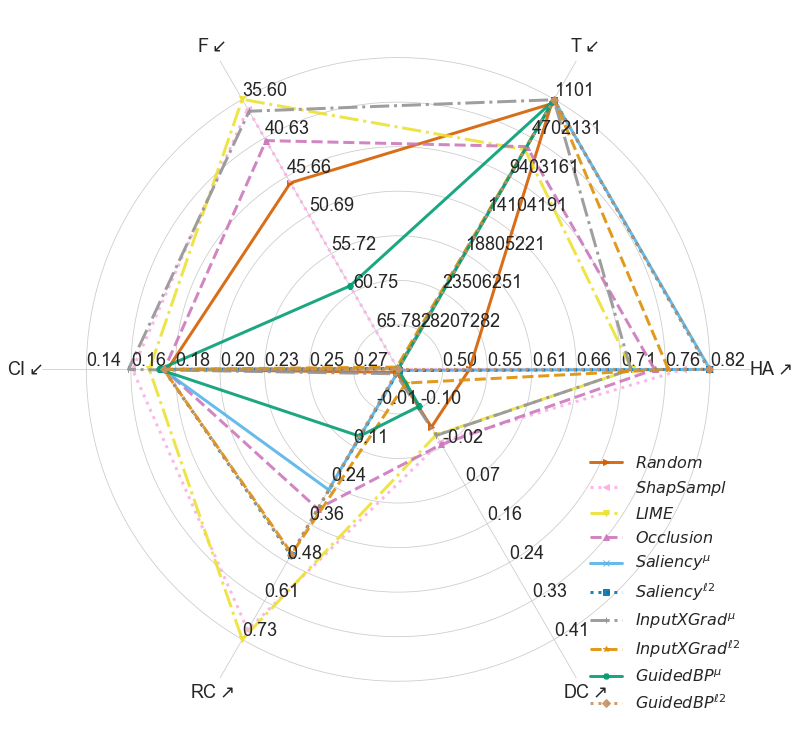}
  \caption{\cnn}
  \label{fig:sub32}
\end{subfigure}
\begin{subfigure}{.5\textwidth}
  \centering
  \includegraphics[width=215pt]{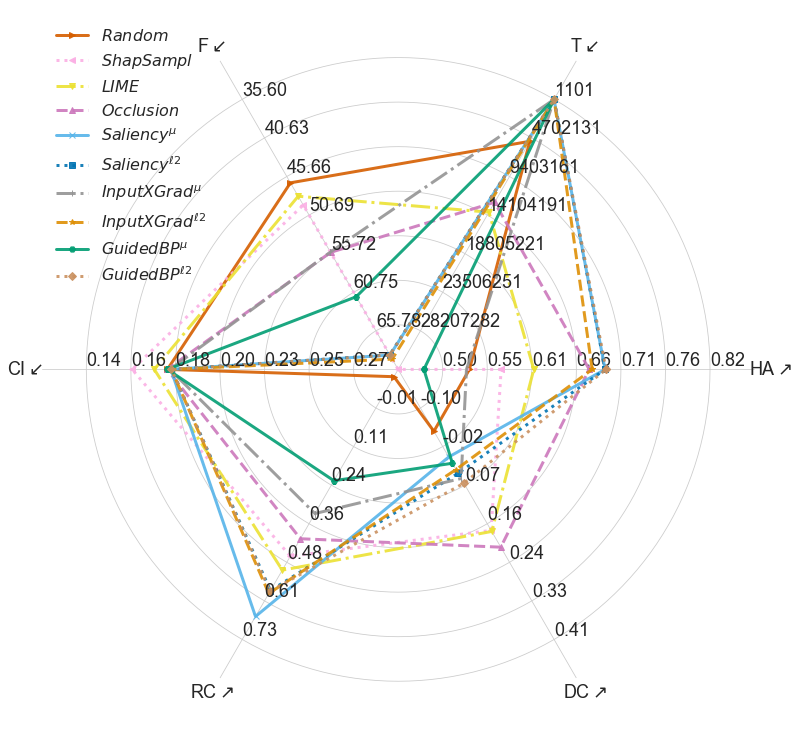}
  \caption{\lstm}
  \label{fig:sub33}
\end{subfigure}
 \caption{Diagnostic property evaluation for 
all explainability techniques, on the TSE dataset. 
The $\nearrow$ and $\swarrow$ signs indicate that higher, correspondingly lower, values of the property measure are better.}
\label{fig:spider3}
\end{figure}

\textbf{Diagnostic property performance.} Figure~\ref{fig:spider1} shows the performance of each explainability technique for all \propertyplural\ on the e-SNLI dataset, and Figure~\ref{fig:spider3} -- for the TSE dataset, which are considerably bigger than IMDB.  The IMDB dataset shows similar tendencies and a corresponding figure can be found in the supplementary material.

\textbf{Agreement with human rationales.}
We observe that the best performing explainability technique for the \trans{} model is \inputxnorm{} followed by the gradient-based ones with L2 norm aggregation.
While for the \cnn{} and the \lstm{} models, we observe similar trends, their MAP scores are always lower than for the \trans, which indicates a correlation between the performance of a model and its agreement with human rationales. 
Furthermore, the MAP scores of the \cnn{} model are higher than for the \lstm{} model, even though the latter achieves higher F1 scores on the e-SNLI dataset.
This might indicate that the representations of the \lstm{} model are less in line with human rationales.
Finally, we note that the mean aggregations of the gradient-based explainability techniques have MAP scores close to or even worse than those from the randomly initialized models.

\textbf{Faithfulness.} 
We find that gradient-based techniques have the best performance for the Faithfulness \property. On the e-SNLI dataset, it is particularly \inputxnorm{}, which performs well across all model architectures. We further find that the \cnn{} exhibits the highest Faithfulness scores for seven out of nine explainability methods. We hypothesize that this is due to the simple architecture with relatively few neural nodes compared to the recurrent nature of the \lstm{} model and the large number of neural nodes in the \trans{} architecture. Finally, models with high Faithfulness scores do not necessarily have high Human agreement scores and vice versa. This suggests that these two are indeed separate \propertyplural{}, and the first should not be confused with estimating the faithfulness of the techniques.

\textbf{Confidence Indication.} 
We find that the Confidence Indication of all models is predicted most accurately by the \shapsamp{}, \lime{}, and \occlusion{} explainability methods. This result is expected, as they compute the saliency of words based on differences in the model's confidence using different instance perturbations. We further find that the \cnn{} model's confidence 
is better predicted with \inputxmean{}. The lowest MAE with the balanced dataset is for the \cnn{} and \lstm{} models. We hypothesize that this could be due to these models' overconfidence, which makes it challenging to detect when the model is not confident of its prediction.


\textbf{Rationale Consistency.} 
There is no single universal explainability technique that achieves the highest score for Rationale Consistency property. 
We see that \lime{} can be good at achieving a high performance, which is expected, as it is trained to approximate the model's performance. The latter is beneficial, especially for models with complex architectures like the \trans. The gradient-based approaches also have high Rationale Consistency scores. We find that the \occlusion{} technique is the best performing for the \lstm{} across all tasks, as it is the simplest of the explored explainability techniques, and does not inspect the model's internals or try to approximate them. This might serve as an indication that \lstm{} models, due to their recurrent nature, can be best explained with simple perturbation based methods that do not examine a model's reasoning process. 

\textbf{Dataset Consistency.} Finally, the results for the Dataset Consistency property show low to moderate correlations of the explainability techniques with similarities across instances in the dataset. The correlation is present for LIME and the gradient-based techniques, again with higher scores for the L2 aggregated gradient-based methods. 

\textbf{Overall.} To summarise, the proposed list of \propertyplural{} allows for assessing existing explainability techniques from different perspectives and supports the choice of the best performing one. Individual property results indicate that gradient-based methods have the best performance.
The only strong exception to the above is the better performance of \shapsamp{} and \lime{} for the Confidence Indication \property. However, \shapsamp{}, \lime{} and \occlusion{} take considerably more time to compute and have worse performance for all other \propertyplural. 

\section{Conclusion}
We proposed a comprehensive list of \propertyplural{} for the evaluation of explainability techniques from different perspectives. We further used them to compare and contrast different groups of explainability techniques on three downstream tasks and three diverse architectures. We found that gradient-based explanations are the best for all of the three models and all of the three downstream text classification tasks that we consider in this work. Other explainability techniques, such as \shapsamp{}, \lime{} and \occlusion{} take more time to compute, and are in addition considerably less faithful to the models and less consistent with the rationales of the models and similarities in the datasets. 

\section*{Acknowledgements}
$\begin{array}{l}\includegraphics[width=1cm]{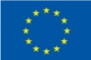} \end{array}$ This project has received funding from the European Union's Horizon 2020 research and innovation programme under the Marie Sk\l{}odowska-Curie grant agreement No 801199.






\clearpage
\section{Appendix}
\subsection{Experimental Setup}~\label{appendix:A}
\begin{table}[h!]
\centering
\small
\begin{tabular}{lrr}
\toprule
\textbf{Model} & \textbf{Time} & \textbf{Score} \\ \midrule
\multicolumn{3}{c}{\textbf{e-SNLI}} \\
\trans & 244.763 ($\pm$62.022) & 0.523 ($\pm$0.356) \\
\cnn & 195.041 ($\pm$53.994) & 0.756 ($\pm$0.028) \\
\lstm & 377.180 ($\pm$232.918) & 0.708 ($\pm$0.205)\\

\multicolumn{3}{c}{\textbf{Movie Reviews}}\\
\trans & 3.603 ($\pm$0.031) & 0.785 ($\pm$0.226) \\
\cnn & 4.777 ($\pm$1.953) & 0.756 ($\pm$0.058)\\
\lstm & 5.344 ($\pm$1.593) & 0.584 ($\pm$0.061) \\

\multicolumn{3}{c}{\textbf{TSE}} \\
\trans & 9.393 ($\pm$1.841) & 0.783 ($\pm$0.006) \\
\cnn & 2.240 ($\pm$0.544) & 0.730 ($\pm$0.035) \\
\lstm & 3.781 ($\pm$1.196) & 0.713 ($\pm$0.076) \\
\bottomrule
\end{tabular}
\caption{Hyper-parameter tuning details. \textit{Time} is the average time (mean and standard deviation in brackets) measured in minutes required for a particular model with all hyper-parameter combinations. \textit{Score} is the mean and standard deviation of the performance on the validation set as a function of the number of the different hyper-parameter searches.}
\label{tab:modeleval}
\end{table}

\paragraph{Machine Learning Models}. The models used in our experiments are trained on the training splits, and the parameters are selected according to the development split. We conducted fine-tuning in a grid-search manner with the ranges and parameters we describe next. We use superscripts to indicate when a parameter value was selected for one of the datasets e-SNLI -- 1, Movie Review -- 2, and TSE -- 3. For the \cnn{} model, we experimented with the following parameters: embedding dimension $\in \{50, 100, 200, 300^{1, 2, 3}\}$, batch size $\in \{16^{2}, 32, 64^{3}, 128, 256^{1}\}$, dropout rate $\in \{0.05^{1,2,3}, 0.1, 0.15, 0.2\}$, learning rate for an Adam optimizer $\in \{0.01, 0.03, 0.001^{2, 3}, 0.003, 0.0001^{1}, 0.0003\}$, window sizes $\in \{[2, 3, 4]^{2}, [2, 3, 4, 5], [3, 4, 5]^{3}, [3, 4, 5, 6],$ $[4, 5, 6], [4, 5, 6, 7]^{1}\}$, and number of output channels $\in \{50^{2, 3}, 100, 200, 300^{1}\}$. We leave the stride and the padding parameters to their default values -- one and zero. 

For the \lstm{} model we fine-tuned over the following grid of parameters: embedding dimension $\in \{50, 100^{1, 2}, 200^{3}, 300\}$, batch size $\in \{16^{2,3}, 32, 64, 128, 256^{1}\}$, dropout rate $\in \{0.05^{3}, 0.1^{1, 2}, 0.15, 0.2\}$, learning rate for an Adam optimizer $\in \{0.01^{1}, 0.03^{2}, 0.001^{2, 3}, 0.003, 0.0001, 0.0003\}$, number of LSTM layers $\in \{1^{2, 3}, 2, 3, 4^{1}\}$, LSTM hidden layer size $\in \{50, 100^{1, 2, 3}, 200, 300\}$, and size of the two linear layers $\in \{[50, 25]^{2}, [100, 50]^{1}, [200, 100]^{3}\}$. We also experimented with other numbers of linear layers after the recurrent ones, but having three of them, where the final was the prediction layer, yielded the best results. 

The \cnn{} and \lstm{} models are trained with an early stopping over the validation accuracy with a patience of five and a maximum number of training epochs of 100. We also experimented with other optimizers, but none yielded improvements.

Finally, for the \trans{} model we fine-tuned the pre-trained basic, uncased LM~(\cite{Wolf2019HuggingFacesTS})(110M parameters) where the maximum input size is 512, and the hidden size of each layer of the 12 layers is 768. We performed a grid-search over learning rate of $\in \{1e-5, 2e-5^{1, 2}, 3e-5^{3}, 4e-5, 5e-5\}$. The models were trained with a warm-up period where the learning rate increases linearly between 0 and 1 for 0.05\% of the steps found with a grid-search. We train the models for five epochs with an early stopping with patience of one as the Transformer models are easily fine-tuned for a small number of epochs.

All experiments were run on a single NVIDIA TitanX GPU with 8GB, and 4GB of RAM and 4 Intel Xeon Silver 4110 CPUs.

The models were evaluated with macro F1 score, which can be found here \url{https://scikit-learn.org/stable/modules/generated/sklearn.metrics.precision_recall_fscore_support.html} and is defined as follows:
\begin{equation*}
   Precision (P) = \frac{\mathrm{TP}}{\mathrm{TP} + \mathrm{FP}} 
\end{equation*}
\begin{equation*}
   Recall (R) = \frac{\mathrm{TP}}{\mathrm{TP} + \mathrm{FN}} 
\end{equation*}
\begin{equation*}
   F1 = \frac{2*\mathrm{P}*\mathrm{R}}{\mathrm{P}+\mathrm{R}} 
\end{equation*}
where TP is the number of true positives, FP is the number of false positives, and FN is the number of false negatives.

\textbf{Explainability generation}. When evaluating the Confidence Indication property of the explainability measures, we train a logistic regression for 5 splits and provide the MAE over the five test splits. As for some of the models, e.g. \trans{}, the confidence is always very high, the LR starts to predict only the average confidence. To avoid this, we additionally randomly up-sample the training instances with a smaller confidence, making the number of instances in each confidence interval [0.0-0.1],\dots[0.9-1.0]) to be the same as the maximum number of instances found in one of the separate intervals.

For both Rationale and Dataset Consistency properties, we consider Spearman's $\rho$. While 
Pearson's $\rho$ measures only the linear correlation between two variables (a change in one variable should be proportional to the change in the other variable), Spearman's $\rho$ measures the monotonic correlation (when one variable increases, the other increases, too). In our experiments, we are interested in the monotonic correlation as all activation differences don't have to be linearly proportional to the differences of the explanations and therefore measure Spearman's $\rho$. 

The Dataset Consistency property is estimated over instance pairs from the test dataset. As computing it for all possible pairs in the dataset is computationally expensive, we select 2 000 pairs from each dataset in order of their decreasing word overlap and sample 2 000 from the remaining instance pairs. This ensures that we compute the \property\ on a set containing tuples of similar and different instances. 

Both the Dataset Consistency property and the Rationale Consistency property estimate the difference between the instances based on their activations. For the \lstm\ model, the activations of the LSTM layers are limited to the output activation also used for prediction as it isn't possible to compare activations with different lengths due to the different token lengths of the different instances. We also use min-max scaling of the differences in the activations and the saliencies as the saliency scores assigned by some explainability techniques are very small. 


\subsection{Spider Figures for the IMDB Dataset}
~\label{appendix:C}
\begin{figure}
\centering
\begin{subfigure}{.5\textwidth}
  \centering
  \includegraphics[width=215pt]{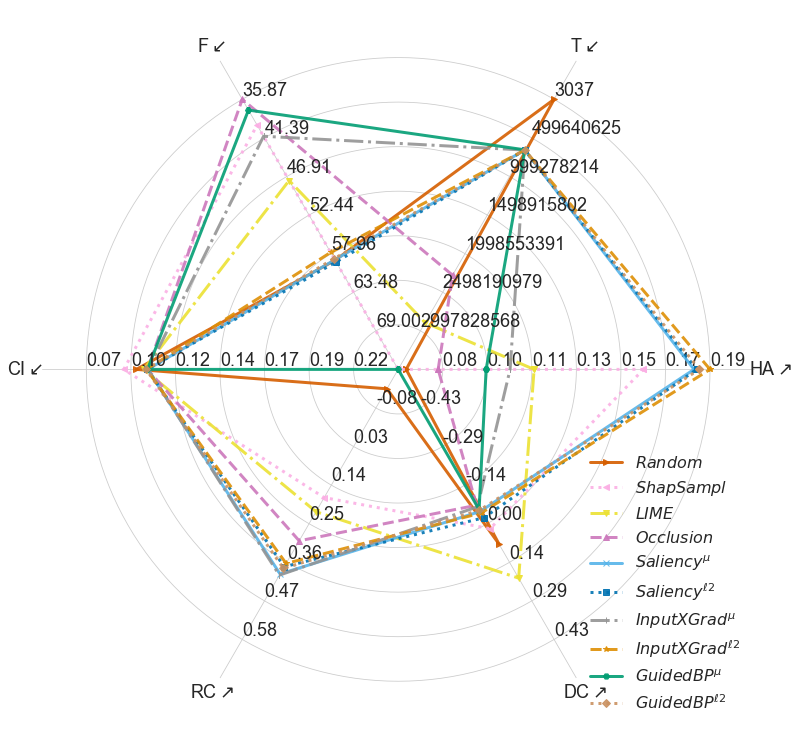}
  \caption{\trans}
  \label{fig:sub21}
  
\end{subfigure} 
\begin{subfigure}{.5\textwidth}
  \centering
  \includegraphics[width=215pt]{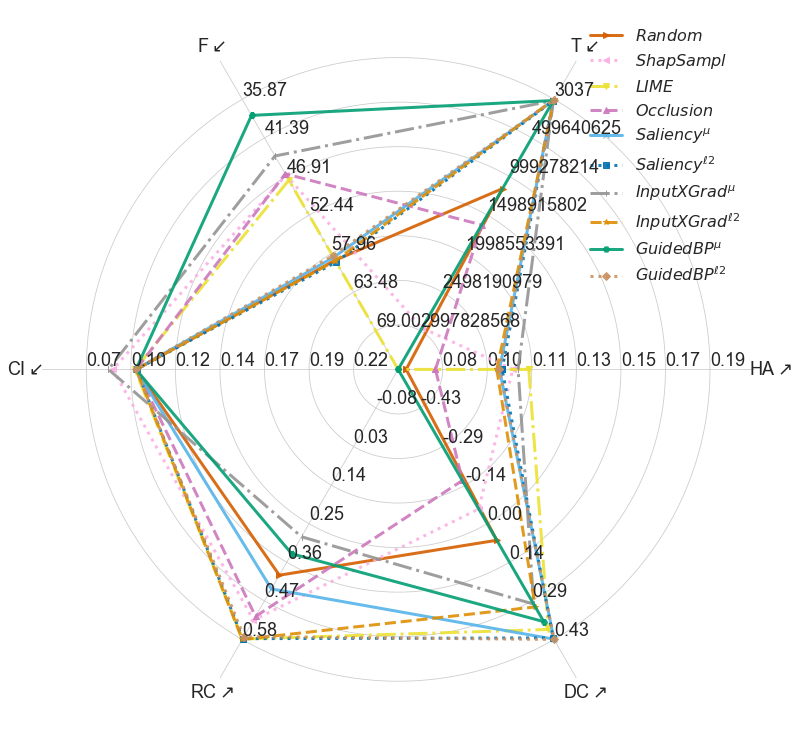}
  \caption{\cnn}
  \label{fig:sub22}
\end{subfigure}
\begin{subfigure}{.5\textwidth}
  \centering
  \includegraphics[width=215pt]{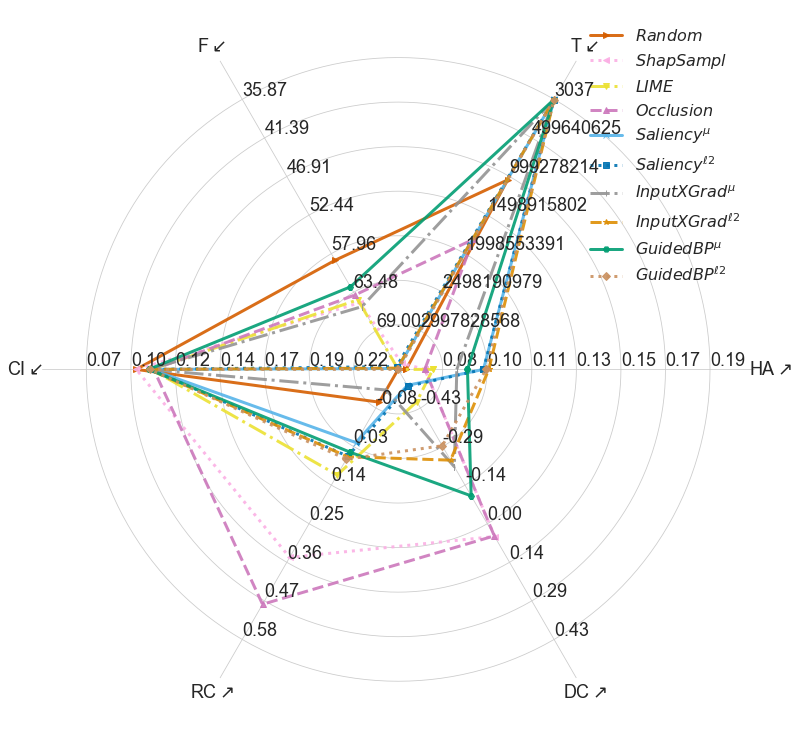}
  \caption{\lstm}
  \label{fig:sub23}
\end{subfigure}
\caption{Diagnostic property evaluation for 
all explainability techniques, on the IMDB dataset. 
The $\nearrow$ and $\swarrow$ signs following the names of each explainability method indicate that higher, correspondingly lower, values of the property measure are better.}
\label{fig:spider2}
\end{figure}

\subsection{Detailed Evaluation Results for the Explainability Techniques}
~\label{appendix:B}
\begin{landscape}
\begin{table}
\centering
\fontsize{9}{9}\selectfont
\begin{tabular}{l@{\hspace{0.7\tabcolsep}}|r@{\hspace{0.7\tabcolsep}}r@{\hspace{0.7\tabcolsep}}l@{\hspace{0.5\tabcolsep}}|r@{\hspace{0.7\tabcolsep}}r@{\hspace{0.7\tabcolsep}}l@{\hspace{0.5\tabcolsep}}|r@{\hspace{0.7\tabcolsep}}r@{\hspace{0.7\tabcolsep}}l@{\hspace{0.5\tabcolsep}}}
\toprule
\textbf{Explain.}&\multicolumn{3}{c}{\textbf{e-SNLI}}&\multicolumn{3}{c}{\textbf{IMDB}}&\multicolumn{3}{c}{\textbf{TSE}}\\
&\textbf{MAP}&\textbf{MAP RI}&\textbf{FLOPs}&\textbf{MAP}&\textbf{MAP RI}&\textbf{FLOPs}&\textbf{MAP}&\textbf{MAP RI}&\textbf{FLOPs}\\
\midrule
\rand&.297 ($\pm$.001)&--&6.12e+3 ($\pm$4.6e+1)&.079 ($\pm$.001)&--&9.41e+4 ($\pm$1.8e+2)&.573 ($\pm$.001)&--&4.62e+3 ($\pm$2.2e+1)\\
\midrule
\multicolumn{10}{c}{\trans}\\
\shapsamp&.511 ($\pm$.004)&.292 ($\pm$.011)&1.78e+7 ($\pm$5.5e+5)&.168 ($\pm$.003)&.084 ($\pm$.001)&3.00e+9 ($\pm$1.3e+8)&.716 ($\pm$.003)&.575 ($\pm$.027)&1.29e+7 ($\pm$2.0e+6)\\
\lime&.465 ($\pm$.008)&.264 ($\pm$.004)&2.39e+5 ($\pm$1.5e+4)&.127 ($\pm$.004)&.075 ($\pm$.004)&4.98e+8 ($\pm$1.4e+8)&.745 ($\pm$.003)&.570 ($\pm$.028)&2.82e+7 ($\pm$1.6e+6)\\
\occlusion&.537 ($\pm$.014)&.292 ($\pm$.009)&6.33e+5 ($\pm$1.0e+3)&.091 ($\pm$.001)&.084 ($\pm$.001)&8.05e+7 ($\pm$4.5e+5)&.710 ($\pm$.008)&.577 ($\pm$.012)&5.86e+5 ($\pm$1.6e+2)\\
\salmean&.614 ($\pm$.003)&.255 ($\pm$.008)&5.38e+4 ($\pm$1.8e+2)&.187 ($\pm$.005)&.079 ($\pm$.001)&6.59e+5 ($\pm$1.8e+3)&.725 ($\pm$.011)&.499 ($\pm$.002)&4.93e+4 ($\pm$2.1e+2)\\
\salnorm&.615 ($\pm$.003)&.255 ($\pm$.009)&5.39e+4 ($\pm$1.3e+2)&.188 ($\pm$.006)&.078 ($\pm$.001)&6.62e+5 ($\pm$8.4e+2)&.726 ($\pm$.014)&.498 ($\pm$.001)&4.93e+4 ($\pm$1.4e+2)\\
\inputxmean&.356 ($\pm$.005)&.280 ($\pm$.016)&5.38e+4 ($\pm$1.8e+2)&.118 ($\pm$.003)&.083 ($\pm$.001)&6.60e+5 ($\pm$4.5e+3)&.620 ($\pm$.008)&.558 ($\pm$.011)&4.92e+4 ($\pm$1.4e+2)\\
\inputxnorm&\underline{\textbf{.624 ($\boldsymbol \pm$.004)}}&.254 ($\pm$.013)&5.39e+4 ($\pm$1.5e+2)&\underline{\textbf{.193 ($\boldsymbol \pm$.005)}}&.079 ($\pm$.001)&6.62e+5 ($\pm$2.1e+3)&\textbf{.774 ($\boldsymbol \pm$.009)}&.499 ($\pm$.005)&4.92e+4 ($\pm$8.0e+1)\\
\guidedmean&.340 ($\pm$.012)&.281 ($\pm$.025)&5.39e+4 ($\pm$1.8e+2)&.109 ($\pm$.003)&.086 ($\pm$.005)&6.54e+5 ($\pm$7.5e+3)&.589 ($\pm$.006)&.567 ($\pm$.008)&4.94e+4 ($\pm$4.1e+2)\\
\guidednorm&.615 ($\pm$.003)&.255 ($\pm$.009)&5.38e+4 ($\pm$1.1e+2)&.189 ($\pm$.005)&.079 ($\pm$.001)&6.59e+5 ($\pm$2.8e+3)&.726 ($\pm$.012)&.498 ($\pm$.001)&4.97e+4 ($\pm$4.2e+2)\\
\midrule
\multicolumn{10}{c}{\cnn}\\
\shapsamp&.471 ($\pm$.003)&.298 ($\pm$.008)&3.79e+7 ($\pm$3.1e+3)&.119 ($\pm$.004)&.084 ($\pm$.001)&1.26e+7 ($\pm$1.6e+5)&.789 ($\pm$.004)&.586 ($\pm$.017)&4.53e+6 ($\pm$2.1e+4)\\
\lime&.466 ($\pm$.002)&.300 ($\pm$.017)&1.81e+4 ($\pm$1.2e+3)&\textbf{.125 ($\boldsymbol \pm$.005)}&.079 ($\pm$.004)&5.39e+7 ($\pm$1.9e+4)&.737 ($\pm$.002)&.581 ($\pm$.021)&1.52e+4 ($\pm$7.1e+1)\\
\occlusion&.487 ($\pm$.003)&.298 ($\pm$.006)&6.06e+4 ($\pm$2.9e+2)&.090 ($\pm$.001)&.084 ($\pm$.001)&3.36e+5 ($\pm$2.6e+3)&.760 ($\pm$.004)&.580 ($\pm$.006)&1.40e+4 ($\pm$3.6e+1)\\
\salmean&\textbf{.600 ($\boldsymbol \pm$.002)}&.339 ($\pm$.007)&1.08e+4 ($\pm$5.6e+1)&.114 ($\pm$.005)&.091 ($\pm$.001)&4.28e+3 ($\pm$2.3e+2)&\underline{\textbf{.816 ($\pm$.003)}}&.593 ($\pm$.008)&4.16e+3 ($\pm$1.9e+1)\\
\salnorm&\textbf{.600 ($\boldsymbol \pm$.002)}&.339 ($\pm$.007)&1.06e+4 ($\pm$5.6e+1)&.115 ($\pm$.005)&.090 ($\pm$.001)&4.29e+3 ($\pm$9.9e+1)&.815 ($\pm$.003)&.596 ($\pm$.009)&4.16e+3 ($\pm$1.2e+1)\\
\inputxmean&.435 ($\pm$.001)&.294 ($\pm$.014)&1.07e+4 ($\pm$2.3e+1)&.121 ($\pm$.003)&.086 ($\pm$.002)&4.27e+3 ($\pm$1.8e+2)&.736 ($\pm$.002)&.572 ($\pm$.011)&4.16e+3 ($\pm$1.2e+1)\\
\inputxnorm&.580 ($\pm$.001)&.280 ($\pm$.003)&1.06e+4 ($\pm$6.5e+1)&.113 ($\pm$.004)&.093 ($\pm$.002)&4.09e+3 ($\pm$1.8e+2)&.774 ($\pm$.003)&.501 ($\pm$.006)&4.12e+3 ($\pm$2.7e+1)\\
\guidedmean&\textcolor{bad_res}{.269 ($\pm$.001)}&.299 ($\pm$.017)&1.08e+4 ($\pm$1.7e+2)&\textcolor{bad_res}{.076 ($\pm$.002)}&.086 ($\pm$.002)&4.27e+3 ($\pm$2.2e+2)&\textcolor{bad_res}{.501 ($\pm$.006)}&.573 ($\pm$.013)&4.32e+3 ($\pm$4.0e+2)\\
\guidednorm&\textbf{.600 ($\boldsymbol \pm$.002)}&.339 ($\pm$.007)&1.07e+4 ($\pm$3.4e+1)&.114 ($\pm$.005)&.091 ($\pm$.002)&4.21e+3 ($\pm$2.2e+2)&.815 ($\pm$.003)&.594 ($\pm$.009)&4.14e+3 ($\pm$1.7e+1)\\
\midrule
\multicolumn{10}{c}{\lstm}\\
\shapsamp&.396 ($\pm$.012)&.291 ($\pm$.008)&8.42e+5 ($\pm$1.2e+4)&.086 ($\pm$.001)&.084 ($\pm$.000)&2.30e+8 ($\pm$2.5e+5)&.605 ($\pm$.034)&.588 ($\pm$.020)&1.12e+7 ($\pm$2.1e+6)\\
\lime&.429 ($\pm$.012)&.309 ($\pm$.018)&1.68e+5 ($\pm$2.1e+5)&.089 ($\pm$.001)&.081 ($\pm$.002)&3.00e+8 ($\pm$1.8e+5)&.638 ($\pm$.025)&.588 ($\pm$.021)&5.20e+4 ($\pm$4.1e+3)\\
\occlusion&.358 ($\pm$.003)&.281 ($\pm$.007)&2.46e+5 ($\pm$5.7e+0)&.086 ($\pm$.002)&.083 ($\pm$.002)&1.18e+6 ($\pm$1.1e+3)&.694 ($\pm$.011)&.578 ($\pm$.016)&3.71e+4 ($\pm$2.7e+0)\\
\salmean&.502 ($\pm$.008)&.411 ($\pm$.011)&5.11e+3 ($\pm$6.8e+0)&.108 ($\pm$.001)&.106 ($\pm$.000)&3.04e+3 ($\pm$7.7e+1)&.710 ($\pm$.009)&.546 ($\pm$.000)&1.11e+3 ($\pm$2.8e+0)\\
\salnorm&.502 ($\pm$.008)&.410 ($\pm$.010)&5.12e+3 ($\pm$4.6e+0)&.108 ($\pm$.002)&.106 ($\pm$.002)&3.07e+3 ($\pm$3.9e+1)&.710 ($\pm$.010)&.546 ($\pm$.001)&1.10e+3 ($\pm$1.4e+0)\\
\inputxmean&.364 ($\pm$.004)&.349 ($\pm$.027)&5.12e+3 ($\pm$7.2e+0)&.098 ($\pm$.002)&.096 ($\pm$.002)&3.06e+3 ($\pm$7.0e+1)&\textcolor{bad_res}{.570 ($\pm$.010)}&.601 ($\pm$.017)&1.11e+3 ($\pm$2.2e+0)\\
\inputxnorm&\textbf{.511 ($\pm$.007)}&.389 ($\pm$.004)&5.12e+3 ($\pm$4.2e+0)&\textbf{.110 ($\boldsymbol \pm$.001)}&.107 ($\pm$.000)&3.05e+3 ($\pm$9.9e+1)&.697 ($\pm$.007)&.544 ($\pm$.001)&1.10e+3 ($\pm$1.6e+0)\\
\guidedmean&.333 ($\pm$.009)&.382 ($\pm$.033)&5.11e+3 ($\pm$4.4e+0)&.102 ($\pm$.005)&.098 ($\pm$.003)&3.06e+3 ($\pm$1.0e+2)&\textcolor{bad_res}{.527 ($\pm$.005)}&.570 ($\pm$.031)&1.10e+3 ($\pm$2.2e+0)\\
\guidednorm&.502 ($\pm$.009)&.410 ($\pm$.009)&5.10e+3 ($\pm$2.5e+1)&.109 ($\pm$.001)&.107 ($\pm$.001)&3.08e+3 ($\pm$9.2e+1)&\textbf{.711 ($\boldsymbol \pm$.009)}&.547 ($\pm$.001)&1.10e+3 ($\pm$2.4e+0)\\
\bottomrule
\end{tabular}
\caption{Evaluation of the explainability techniques with Human Agreement (HA) and time for computation. HA is measured with Mean Average Precision (MAP) with the gold human annotations, MAP of a Randomly initialized model (MAP RI). The time is computed with FLOPs. The presented numbers are averaged over five different models and the standard deviation of the scores is presented in brackets. Explainability methods with the best MAP for a particular dataset and model are in bold, while the best MAP across all models for a dataset is underlined as well. Methods that have MAP worse than the randomly generated saliency are in \textcolor{bad_res}{red}.}
\label{tab:human}
\end{table}
\end{landscape}

\begin{table*}[h!]
\centering
\begin{tabular}{lrrr}
\toprule
\textbf{Explain.}&\textbf{e-SNLI}&\textbf{IMDB}&\textbf{TSE}\\
\midrule
\rand&56.05 ($\pm$0.71)&49.26 ($\pm$1.94)&56.45 ($\pm$2.37)\\
\midrule
\multicolumn{4}{c}{\textbf{\trans}}\\
\shapsamp&56.05 ($\pm$0.71)&\textcolor{bad_res}{65.84 ($\pm$11.8)}&52.99 ($\pm$4.24)\\
\lime&48.14 ($\pm$10.8)&\textcolor{bad_res}{59.04 ($\pm$13.7)}& 42.17 ($\pm$7.89)\\
\occlusion&55.24 ($\pm$3.77)&\textcolor{bad_res}{69.00 ($\pm$6.22)}&52.23 ($\pm$4.29)\\
\salmean&37.98 ($\pm$2.18)&\textcolor{bad_res}{49.32 ($\pm$9.01)}&\textbf{39.20 ($\boldsymbol \pm$3.06)}\\
\salnorm&38.01 ($\pm$2.19)&\textbf{49.05 ($\boldsymbol \pm$9.16)}&39.29 ($\pm$3.14)\\
\inputxmean&\textcolor{bad_res}{56.98 ($\pm$1.89)}&\textcolor{bad_res}{64.47 ($\pm$8.70)}&55.52 ($\pm$2.59)\\
\inputxnorm&\textbf{37.05 ($\boldsymbol \pm$2.29)}&\textcolor{bad_res}{50.22 ($\pm$8.85)}&37.04 ($\pm$2.69)\\
\guidedmean&53.43 ($\pm$1.00)&\textcolor{bad_res}{67.68 ($\pm$6.94)}&\textcolor{bad_res}{57.56 ($\pm$2.60)}\\
\guidednorm&38.01 ($\pm$2.19)&\textcolor{bad_res}{49.47 ($\pm$8.89)}&39.26 ($\pm$3.18)\\
\midrule
\multicolumn{4}{c}{\cnn}\\
\shapsamp&51.78 ($\pm$2.24)&\textcolor{bad_res}{59.69 ($\pm$8.37)}&\textcolor{bad_res}{64.72 ($\pm$1.75)}\\
\lime&\textcolor{bad_res}{56.16 ($\pm$1.67)}&\textcolor{bad_res}{59.09 ($\pm$8.48)}&\textcolor{bad_res}{65.78 ($\pm$1.59)}\\
\occlusion&54.32 ($\pm$0.94)&\textcolor{bad_res}{59.86 ($\pm$7.78)}&\textcolor{bad_res}{61.17 ($\pm$1.48)}\\
\salmean&34.26 ($\pm$1.78)&\textcolor{bad_res}{49.61 ($\pm$5.26)}&35.70 ($\pm$2.94)\\
\salnorm&34.16 ($\pm$1.81)&\textbf{49.04 ($\boldsymbol \pm$5.60)}&35.67 ($\pm$2.91)\\
\inputxmean&47.06 ($\pm$3.82)&\textcolor{bad_res}{62.05 ($\pm$7.54)}&\textcolor{bad_res}{64.45 ($\pm$2.99)}\\
\inputxnorm&\underline{\textbf{31.55 ($\boldsymbol \pm$2.83)}}&49.20 ($\pm$5.96)&35.86 ($\pm$3.22)\\
\guidedmean&47.68 ($\pm$2.65)&\textcolor{bad_res}{67.03 ($\pm$4.36)}&44.93 ($\pm$1.57)\\
\guidednorm&34.16 ($\pm$1.81)&\textcolor{bad_res}{49.80 ($\pm$5.99)}&\underline{\textbf{35.60 ($\boldsymbol \pm$2.91)}}\\
\midrule
\multicolumn{4}{c}{\lstm}\\
\shapsamp&51.05 ($\pm$4.47)&44.05 ($\pm$3.06)&53.97 ($\pm$6.00)\\
\lime&51.93 ($\pm$7.73)&\textcolor{bad_res}{44.41 ($\pm$3.04)}&54.95 ($\pm$3.19)\\
\occlusion&54.73 ($\pm$3.12)&45.01 ($\pm$3.84)&48.68 ($\pm$2.28)\\
\salmean&38.29 ($\pm$1.77)&35.98 ($\pm$2.11)&\textbf{37.20 ($\boldsymbol \pm$3.48)}\\
\salnorm&38.26 ($\pm$1.84)&36.22 ($\pm$2.04)&37.23 ($\pm$3.50)\\
\inputxmean&49.52 ($\pm$1.81)&43.57 ($\pm$4.98)&48.71 ($\pm$3.23)\\
\inputxnorm&\textbf{37.95 ($\boldsymbol \pm$2.06)}&36.03 ($\pm$1.97)&36.75 ($\pm$3.35)\\
\guidedmean&44.48 ($\pm$2.12)&46.00 ($\pm$3.20)&43.72 ($\pm$5.69)\\
\guidednorm&38.17 ($\pm$1.80)&\underline{\textbf{35.87 ($\boldsymbol \pm$1.99)}}&37.21 ($\pm$3.48)\\
\bottomrule
\end{tabular}

\caption{Faithfulness-AUC for thresholds $\in$ [0, 10, 20, \dots, 100]. \textit{Lower scores} indicate the ability of the saliency approach to assign higher scores to words more responsible for the final prediction. The presented scores are averaged over the different random initializations and the standard deviation is shown in brackets. Explainability methods with the smallest AUC for a particular dataset and model are in bold, while the smallest AUC across all models for a dataset is underlined as well. Methods that have AUC worse than the randomly generated saliency are in \textcolor{bad_res}{red}.}
\label{tab:faith}
\end{table*}

\begin{landscape}
\begin{table}[p]
\centering
\fontsize{9}{9}\selectfont
\begin{tabular}{l@{\hspace{0.2\tabcolsep}}|r@{\hspace{0.7\tabcolsep}}r@{\hspace{0.7\tabcolsep}}r@{\hspace{0.7\tabcolsep}}r@{\hspace{0.5\tabcolsep}}|r@{\hspace{0.7\tabcolsep}}r@{\hspace{0.7\tabcolsep}}r@{\hspace{0.7\tabcolsep}}r@{\hspace{0.5\tabcolsep}}|r@{\hspace{0.7\tabcolsep}}r@{\hspace{0.7\tabcolsep}}r@{\hspace{0.7\tabcolsep}}r}
\toprule
& \multicolumn{4}{c}{\textbf{e-SNLI}}&\multicolumn{4}{c}{\textbf{IMDB}}&\multicolumn{4}{c}{\textbf{TSE}} \\
\textbf{Explain.} & \textbf{MAE} & \textbf{MAX} &\textbf{MAE-up} & \textbf{MAX-up} & \textbf{MAE} & \textbf{MAX} & \textbf{MAE-up} & \textbf{MAX-up} & \textbf{MAE} & \textbf{MAX} & \textbf{MAE-up} & \textbf{MAX-up} \\ \midrule
\rand&.087 ($\pm$.004)&.527 ($\pm$.007)&.276 ($\pm$.005)&.377 ($\pm$.002)&.130 ($\pm$.007)&.286 ($\pm$.014)&.160 ($\pm$.003)&.251 ($\pm$.008)&.092 ($\pm$.009)&.466 ($\pm$.021)&.260 ($\pm$.017)&.428 ($\pm$.064) \\
\midrule
\multicolumn{13}{c}{\textbf{\trans}} \\
\shapsamp&.071 ($\pm$.005)&.456 ($\pm$.037)&.158 ($\pm$.029)&.437 ($\pm$.046)&\textbf{.071 ($\boldsymbol \pm$.008)}&\textbf{.238 ($\boldsymbol \pm$.036)}&\textbf{.120 ($\boldsymbol \pm$.033)}&\textbf{.213 ($\boldsymbol \pm$.035)}&\underline{\textbf{.073 ($\boldsymbol \pm$.012)}}&\textbf{.408 ($\boldsymbol \pm$.043)}&\textbf{.169 ($\boldsymbol \pm$.052)}&\textbf{.415 ($\boldsymbol \pm$.030)} \\
\lime&\underline{\textbf{.068 ($\boldsymbol \pm$.002)}}& \underline{\textbf{.368 ($\boldsymbol \pm$.151)}}&\textbf{.136 ($\boldsymbol \pm$.028)}&\textbf{.395 ($\boldsymbol \pm$.128)}&.077 ($\pm$.008)&.288 ($\pm$.024)&.184 ($\pm$.018)&.260 ($\pm$.021)&.084 ($\pm$.009)&.521 ($\pm$.072)&.232 ($\pm$.013)&.661 ($\pm$.225) \\
\occlusion&.074 ($\pm$.004)&.499 ($\pm$.020)&.224 ($\pm$.006)&.518 ($\pm$.048)&.085 ($\pm$.011)&.306 ($\pm$.015)&.196 ($\pm$.015)&.252 ($\pm$.011)&.085 ($\pm$.011)&.463 ($\pm$.035)&.247 ($\pm$.015)&.482 ($\pm$.091) \\
\salmean&.078 ($\pm$.005)&.544 ($\pm$.014)&.269 ($\pm$.004)&.416 ($\pm$.043)&.083 ($\pm$.009)&.303 ($\pm$.008)&.197 ($\pm$.017)&.269 ($\pm$.023)&.085 ($\pm$.012)&.474 ($\pm$.021)&.248 ($\pm$.017)&.467 ($\pm$.091) \\ 
\salnorm&.078 ($\pm$.005)&.565 ($\pm$.051)&.259 ($\pm$.007)&.571 ($\pm$.095)&.083 ($\pm$.009)&.306 ($\pm$.017)&.195 ($\pm$.021)&.245($\pm$.004)&.085 ($\pm$.012)&.465 ($\pm$.021)&.255 ($\pm$.012)&.479 ($\pm$.074) \\ 
\inputxmean&.079 ($\pm$.005)&.502 ($\pm$.015)&.242 ($\pm$.006)&.518 ($\pm$.031)&.084 ($\pm$.011)&.310 ($\pm$.011)&.198 ($\pm$.013)&.246 ($\pm$.008)&.085 ($\pm$.011)&.463 ($\pm$.015)&.237 ($\pm$.010)&.480 ($\pm$.071) \\
\inputxnorm&.078 ($\pm$.005)&.568 ($\pm$.057)&.258 ($\pm$.007)&.581 ($\pm$.096)&.083 ($\pm$.011)&.301 ($\pm$.014)&.193 ($\pm$.023)&.249 ($\pm$.016)&.086 ($\pm$.013)&.469 ($\pm$.022)&.252 ($\pm$.016)&.480 ($\pm$.087) \\
\guidedmean&.080 ($\pm$.005)&.505 ($\pm$.016)&.242 ($\pm$.008)&.519 ($\pm$.037)&.084 ($\pm$.011)&.308 ($\pm$.009)&.196 ($\pm$.014)&.245 ($\pm$.014)&.085 ($\pm$.011)&.456 ($\pm$.014)&.237 ($\pm$.013)&.494 ($\pm$.069) \\
\guidednorm&.078 ($\pm$.005)&.565 ($\pm$.051)&.258 ($\pm$.007)&.573 ($\pm$.095)&.080 ($\pm$.012)&.306 ($\pm$.009)&.192 ($\pm$.018)&.244 ($\pm$.008)&.086 ($\pm$.012)&.503 ($\pm$.053)&.261 ($\pm$.017)&.450 ($\pm$.081) \\
\midrule
\multicolumn{13}{c}{\textbf{\cnn}} \\
\shapsamp&\textbf{.103 ($\boldsymbol \pm$.001)}&.439 ($\pm$.020)&\textbf{.133 ($\boldsymbol \pm$.003)}&.643 ($\pm$.032)&.077 ($\pm$.018)&.210 ($\pm$.041)&.085 ($\pm$.023)&.196 ($\pm$.026)&.093 ($\pm$.002)&\underline{\textbf{.372 ($\boldsymbol \pm$.011)}}&.148 ($\pm$.004)&.479 ($\pm$.030) \\
\lime&.125 ($\pm$.003)&.498 ($\pm$.018)&.190 ($\pm$.006)&.494 ($\pm$.028)&.128 ($\pm$.006)&.289 ($\pm$.019)&.156 ($\pm$.003)&.260 ($\pm$.011)&.103 ($\pm$.001)&.469 ($\pm$.027)&.202 ($\pm$.014)&.633 ($\pm$.090) \\
\occlusion&.119 ($\pm$.004)&.492 ($\pm$.018)&.176 ($\pm$.007)&.507 ($\pm$.037)&.130 ($\pm$.007)&.289 ($\pm$.018)&.160 ($\pm$.006)&.254 ($\pm$.005)&.114 ($\pm$.002)&.463 ($\pm$.018)&.250 ($\pm$.007)&.418 ($\pm$.035) \\
\salmean&.137 ($\pm$.002)&.496 ($\pm$.011)&.220 ($\pm$.006)&.399 ($\pm$.010)&.129 ($\pm$.007)&.288 ($\pm$.021)&.159 ($\pm$.003)&.253 ($\pm$.013)&.115 ($\pm$.002)&.467 ($\pm$.014)&.245 ($\pm$.007)&.425 ($\pm$.028) \\
\salnorm&.140 ($\pm$.003)&.492 ($\pm$.009)&.225 ($\pm$.005)&.354 ($\boldsymbol \pm$.009)&.130 ($\pm$.006)&.286 ($\pm$.019)&.161 ($\pm$.004)&.250 ($\pm$.005)&.114 ($\pm$.002)&.475 ($\pm$.016)&.248 ($\pm$.006)&.405 ($\pm$.031) \\
\inputxmean&.110 ($\pm$.001)&\textbf{.436 ($\boldsymbol \pm$.014)}&.153 ($\pm$.007)&.460 ($\pm$.009)&\textbf{.071 ($\boldsymbol \pm$.004)}&\underline{\textbf{.191 ($\boldsymbol \pm$.010)}}&\underline{\textbf{.071 ($\boldsymbol \pm$.005)}}&\underline{\textbf{.190 ($\boldsymbol \pm$.010)}}&\textbf{.090 ($\boldsymbol \pm$.002)}&.379 ($\pm$.012)&\underline{\textbf{.135 ($\boldsymbol \pm$.004)}}&.477 ($\boldsymbol \pm$.025) \\ 
\inputxnorm&.140 ($\pm$.003)&.492 ($\pm$.009)&.225 ($\pm$.005)&.355 ($\pm$.007)&.130 ($\pm$.007)&.285 ($\pm$.019)&.160 ($\pm$.004)&.251 ($\pm$.011)&.114 ($\pm$.002)&.475 ($\pm$.014)&.248 ($\pm$.006)&.416 ($\pm$.033) \\
\guidedmean&.140 ($\pm$.003)&.485 ($\pm$.011)&.225 ($\pm$.005)&.367 ($\pm$.023)&.129 ($\pm$.006)&.286 ($\pm$.019)&.159 ($\pm$.003)&.253 ($\pm$.011)&.114 ($\pm$.002)&.462 ($\pm$.013)&.234 ($\pm$.011)&.441 ($\pm$.036) \\
\guidednorm&.140 ($\pm$.003)&.492 ($\pm$.009)&.225 ($\pm$.005)&\underline{\textbf{.353 ($\boldsymbol \pm$.008)}}&.130 ($\pm$.007)&.289 ($\pm$.018)&.159 ($\pm$.004)&.252 ($\pm$.011)&.114 ($\pm$.002)&.473 ($\pm$.015)&.249 ($\pm$.006)&\textbf{.404 ($\boldsymbol \pm$.029)} \\
\midrule
\multicolumn{13}{c}{\textbf{\lstm}} \\
\shapsamp&\textbf{.118 ($\boldsymbol \pm$.003)}&.622 ($\boldsymbol \pm$.035)&\underline{\textbf{.131 ($\boldsymbol \pm$.005)}}&.648 ($\pm$.054)&\underline{\textbf{.060 ($\boldsymbol \pm$.018)}}&\textbf{.279 ($\boldsymbol \pm$.065)}&\textbf{.160 ($\boldsymbol \pm$.014)}&.277 ($\pm$.038)&\textbf{.087 ($\boldsymbol \pm$.007)}&\textbf{.433 ($\boldsymbol \pm$.053)}&\textbf{.147 ($\boldsymbol \pm$.015)}&\underline{\textbf{.393 ($\boldsymbol \pm$.029)}}\\
\lime&.127 ($\pm$.004)&.512 ($\pm$.052)&.145 ($\pm$.009)&.490 ($\pm$.040)&.069 ($\pm$.018)&.300 ($\pm$.051)&.209 ($\pm$.024)&.267 ($\pm$.031)&.090 ($\pm$.007)&.667 ($\pm$.150)&.218 ($\pm$.010)&.864 ($\pm$.362) \\
\occlusion&.147 ($\pm$.003)&.579 ($\pm$.065)&.172 ($\pm$.007)&.593 ($\pm$.083)&.069 ($\pm$.017)&.304 ($\pm$.055)&.216 ($\pm$.014)&.324 ($\pm$.032)&.099 ($\pm$.006)&.509 ($\pm$.015)&.259 ($\pm$.012)&.723 ($\pm$.063) \\
\salmean&.163 ($\pm$.002)&.450 ($\pm$.008)&.195 ($\pm$.008)&.398 ($\pm$.031)&.069 ($\pm$.018)&.301 ($\pm$.051)&.208 ($\pm$.026)&\textbf{.259 ($\boldsymbol \pm$.022)}&.101 ($\pm$.007)&.518 ($\pm$.013)&.271 ($\pm$.008)&.469 ($\pm$.071) \\
\salnorm&.163 ($\pm$.002)&.448 ($\pm$.011)&.195 ($\pm$.008)&.399 ($\pm$.034)&.070 ($\pm$.018)&.299 ($\pm$.051)&.206 ($\pm$.024)&.263 ($\pm$.027)&.101 ($\pm$.007)&.523 ($\pm$.011)&.273 ($\pm$.008)&.441 ($\pm$.051) \\
\inputxmean&.161 ($\pm$.002)&.454 ($\pm$.018)&.193 ($\pm$.007)&.502 ($\pm$.033)&.066 ($\pm$.018)&.295 ($\pm$.059)&.201 ($\pm$.033)&\textbf{.262 ($\boldsymbol \pm$.014)}&.098 ($\pm$.007)&.527 ($\pm$.005)&.268 ($\pm$.008)&.425 ($\pm$.035) \\
\inputxnorm&.163 ($\pm$.002)&\textbf{.445 ($\boldsymbol \pm$.011)}&.195 ($\pm$.007)&\textbf{.394 ($\boldsymbol \pm$.029)}&.068 ($\pm$.018)&.303 ($\pm$.050)&.201 ($\pm$.031)&.277 ($\pm$.024)&.101 ($\pm$.007)&.523 ($\pm$.008)&.273 ($\pm$.007)&.445 ($\pm$.038) \\
\guidedmean&.161 ($\pm$.001)&.453 ($\pm$.014)&.192 ($\pm$.007)&.516 ($\pm$.058)&.068 ($\pm$.019)&.298 ($\pm$.055)&.200 ($\pm$.024)&.287 ($\pm$.045)&.097 ($\pm$.006)&.523 ($\pm$.017)&.260 ($\pm$.016)&.460 ($\pm$.045) \\
\guidednorm&.163 ($\pm$.002)&.446 ($\pm$.010)&.195 ($\pm$.007)&.396 ($\pm$.042)&.069 ($\pm$.017)&.300 ($\pm$.050)&.204 ($\pm$.024)&.279 ($\pm$.025)&.101 ($\pm$.007)&.525 ($\pm$.010)&.273 ($\pm$.007)&.474 ($\pm$.051) \\
\bottomrule
\end{tabular}
\caption{Confidence Indication experiments are measured with the Mean Absolute Error (MAE) of the generated saliency scores when used to predict the confidence of the class predicted by the model and the Maximum Error (MAX). We present the result with and without up-sampling(MAE-up, MAX-up) of the model confidence.
The presented measures are an average over the set of models trained from from different random seeds. The standard deviation of the scores is presented in brackets.  AVG Conf. is the average confidence of the model for the predicted class. The best results for a particular dataset and model are in bold and the best results across a dataset are also underlined. Lower results are better.}
~\label{tab:confidence}
\end{table}
\end{landscape}

\begin{table*}[h!]
\centering
\begin{tabular}{lrrr}
\toprule
\textbf{Explain.} & \textbf{e-SNLI} & \textbf{IMDB} & \textbf{TSE} \\
\midrule
\multicolumn{4}{c}{\textbf{\trans}} \\
\rand & -0.004 (2.6e-01)    & -0.035 (1.4e-01)  & 0.003 (6.1e-01) \\
\shapsamp & 0.310 (0.0e+00) & 0.234 (3.6e-12)   & 0.259 (0.0e+00) \\
\lime & \textbf{0.519 (0.0e+00)} & 0.269 (3.0e-31) & 0.110 (2.0e-29) \\
\occlusion & 0.215 (0.0e+00) & 0.341 (2.6e-50) & 0.255 (0.0e+00) \\
\salmean & 0.356 (0.0e+00) & 0.423 (3.9e-79) & \textbf{0.294 (0.0e+00)} \\
\salnorm & 0.297 (0.0e+00) & 0.405 (6.9e-72) & 0.289 (0.0e+00) \\
\inputxmean & \textcolor{bad_res}{-0.102 (2.0e-202)} & \textbf{0.426 (2.5e-80)} & \textcolor{bad_res}{-0.010 (1.3e-01)} \\
\inputxnorm & 0.311 (0.0e+00) & 0.397 (3.8e-69) & 0.292 (0.0e+00) \\
\guidedmean & 0.064 (1.0e-79) & \textcolor{bad_res}{-0.083 (4.2e-04)} & \textcolor{bad_res}{-0.005 (4.9e-01)} \\
\guidednorm & 0.297 (0.0e+00) & 0.409 (1.2e-73) & 0.293 (0.0e+00) \\
\midrule
\multicolumn{4}{c}{\textbf{\cnn}} \\
\rand & -0.003 (4.0e-01) & 0.426 (2.6e-106) & -0.002 (7.4e-01) \\
\shapsamp & 0.789 (0.0e+00) & 0.537 (1.4e-179) & 0.704 (0.0e+00) \\
\lime & 0.790 (0.0e+00) & 0.584 (1.9e-219) & \underline{\textbf{0.730 (0.0e+00)}} \\
\occlusion & 0.730 (0.0e+00) & 0.528 (2.4e-172) & 0.372 (0.0e+00) \\
\salmean & 0.701 (0.0e+00) & 0.460 (4.5e-126) & 0.320 (0.0e+00) \\
\salnorm & \underline{\textbf{0.819 (0.0e+00)}} & 0.583 (4.0e-218) & 0.499 (0.0e+00) \\
\inputxmean & 0.136 (0.0e+00) & \textcolor{bad_res}{0.331 (1.2e-62)} & 0.002 (7.5e-01) \\
\inputxnorm & 0.816 (0.0e+00) & \underline{\textbf{0.585 (8.6e-221)}} & 0.495 (0.0e+00) \\
\guidedmean & 0.160 (0.0e+00) &\textcolor{bad_res}{ 0.373 (5.5e-80)} & 0.173 (6.3e-121) \\
\guidednorm & \underline{\textbf{0.819 (0.0e+00)}} & 0.578 (2.4e-214) & 0.498 (0.0e+00) \\
\midrule
\multicolumn{4}{c}{\textbf{\lstm}} \\ 
\rand & 0.004 (1.8e-01) & 0.002 (9.2e-01) & 0.010 (1.8e-01) \\
\shapsamp & 0.657 (0.0e+00) & 0.382 (1.7e-63) & 0.502 (0.0e-00) \\
\lime & \textbf{0.700 (0.0e+00)} & 0.178 (3.3e-14) & 0.540 (0.0e-00) \\
\occlusion & 0.697 (0.0e+00) & \textbf{0.498 (1.7e-113)} & 0.454 (0.0e-00) \\
\salmean & 0.645 (0.0e+00) & 0.098 (3.1e-05) & \textbf{0.667 (0.0e-00)} \\
\salnorm & 0.662 (0.0e+00) & 0.132 (1.8e-08) & 0.596 (0.0e-00) \\
\inputxmean & 0.026 (1.9e-14) & \textcolor{bad_res}{-0.032 (1.7e-01)} & 0.385 (0.0e-00) \\
\inputxnorm & 0.664 (0.0e+00) & 0.133 (1.5e-08) & 0.604 (0.0e-00) \\
\guidedmean & 0.144 (0.0e+00) & 0.122 (2.0e-07) & 0.295 (0.0e-00) \\
\guidednorm & 0.663 (0.0e+00) & 0.139 (3.1e-09) & 0.598 (0.0e-00) \\
\bottomrule
\end{tabular}
\caption{Rationale Consistency Spearman's $\rho$ correlation. The estimated p-value for the correlation is provided in the brackets. The best results for a particular dataset and model are in bold and the best results across a dataset are also underlined. Correlation lower that the one of the randomly sampled saliency scores are colored in \textcolor{bad_res}{red}.}
\label{tab:consistency:rat}
\end{table*}

\begin{table*}[h!]
\centering
\begin{tabular}{llll}
\toprule
\textbf{Explain.} & \textbf{e-SNLI} & \textbf{IMDB} & \textbf{TSE} \\
\midrule
\multicolumn{4}{c}{\textbf{\trans}}  \\
\rand & 0.047 (2.7e-04) & 0.127 (6.6e-07)/ & 0.121 (2.5e-01) \\
\shapsamp & 0.285 (1.8e-02) & \textcolor{bad_res}{0.078 (5.8e-04)} & 0.308 (3.4e-36) \\
\lime & 0.372 (3.1e-90) & \textbf{0.236 (4.6e-07)} & \underline{\textbf{0.413 (3.4e-120)}} \\
\occlusion & 0.215 (9.6e-02) & \textcolor{bad_res}{0.003 (2.0e-04)} & 0.235 (7.3e-05) \\
\salmean & 0.378 (4.3e-57) & \textcolor{bad_res}{0.023 (4.3e-02)} & 0.253 (1.4e-20) \\
\salnorm & 0.027 (3.0e-05) & \textcolor{bad_res}{-0.043 (5.6e-02)} & 0.260 (6.8e-21) \\
\inputxmean & 0.319 (3.0e-03) & \textcolor{bad_res}{0.008 (1.2e-01)} & 0.193 (7.5e-05) \\
\inputxnorm & 0.399 (1.9e-78) & \textcolor{bad_res}{0.028 (2.3e-03)} & 0.247 (4.9e-17) \\
\guidedmean & 0.400 (6.7e-31) & \textcolor{bad_res}{0.017 (1.9e-01)} & 0.228 (5.2e-09) \\
\guidednorm & \underline{\textbf{0.404 (1.4e-84)}} & \textcolor{bad_res}{0.019 (4.3e-04)} & 0.255 (3.1e-20) \\
\midrule
\multicolumn{4}{c}{\textbf{\cnn}} \\
\rand & 0.018 (2.4e-01) & 0.115 (1.8e-04) & 0.008 (2.0e-01) \\
\shapsamp & \textcolor{bad_res}{0.015 (1.8e-01)} & \textcolor{bad_res}{-0.428 (5.3e-153)} & 0.037 (1.4e-01) \\
\lime & \textcolor{bad_res}{0.000 (4.4e-02)} & 0.400 (1.4e-126) & 0.023 (4.0e-01) \\
\occlusion & \textcolor{bad_res}{-0.076 (6.5e-02)} & \textcolor{bad_res}{-0.357 (1.9e-85)} & \textbf{0.041 (1.7e-01)} \\
\salmean & 0.381 (6.9e-91) & 0.431 (1.1e-146) & \textcolor{bad_res}{-0.100 (3.9e-06)} \\
\salnorm & 0.391 (1.7e-98) & 0.427 (3.5e-135) & \textcolor{bad_res}{-0.100 (3.7e-06)} \\
\inputxmean & 0.171 (5.1e-04) & 0.319 (1.4e-69) & 0.024 (3.5e-01) \\
\inputxnorm & \textbf{0.399 (1.0e-93)} & 0.428 (1.4e-132) & \textcolor{bad_res}{-0.076 (1.2e-03)} \\
\guidedmean & 0.091 (7.9e-02) & 0.375 (5.7e-109) & \textcolor{bad_res}{-0.032 (1.1e-01)} \\
\guidednorm & \textbf{0.391 (1.7e-98)} & \underline{\textbf{0.432 (3.5e-140)}} & \textcolor{bad_res}{-0.102 (1.7e-06)} \\
\midrule
\multicolumn{4}{c}{\textbf{\lstm}} \\
\rand & 0.018 (3.9e-01) & 0.037 (1.8e-01) & 0.016 (9.2e-03) \\
\shapsamp & 0.398 (3.5e-81) & 0.230 (8.9e-03) & 0.205 (2.1e-16) \\
\lime & \underline{\textbf{0.415 (1.2e-80)}} & 0.079 (8.6e-04) & 0.207 (4.3e-16) \\
\occlusion & 0.363 (1.1e-37) & \textbf{0.429 (7.5e-137)} & \textbf{0.237 (2.9e-29)} \\
\salmean & 0.158 (1.7e-17) & \textcolor{bad_res}{-0.177 (1.6e-10)} & 0.065 (5.8e-03) \\
\salnorm & 0.160 (7.5e-19) & \textcolor{bad_res}{-0.168 (2.0e-15)} & 0.096 (8.2e-03) \\
\inputxmean & 0.142 (3.3e-06) & \textcolor{bad_res}{-0.152 (1.2e-14)} & 0.106 (2.8e-02) \\
\inputxnorm & 0.183 (7.0e-24) & \textcolor{bad_res}{-0.175 (4.7e-17)} & 0.089 (8.4e-03) \\
\guidedmean & 0.163 (1.9e-12) & \textcolor{bad_res}{-0.060 (4.7e-02)} & 0.077 (1.2e-02) \\
\guidednorm & 0.169 (1.8e-12) & \textcolor{bad_res}{-0.214 (5.8e-16)} & 0.115 (4.3e-02) \\
\bottomrule
\end{tabular}
\caption{Dataset Consistency results with Spearman $\rho$. The estimated p-value for the correlation is provided in the brackets. The best results for a particular dataset and model are in bold and the best results across a dataset are also underlined. Correlation lower that the one of the randomly samples saliency scores are colored in \textcolor{bad_res}{red}.}
\label{tab:consistency:data}
\end{table*}

\part{Veracity Prediction}\label{part:IV}
\chapter{Multi-Domain Evidence-Based Fact Checking of Claims}\label{ch:multifc}


\boxabstract{We contribute the largest publicly available dataset of naturally occurring factual claims for the purpose of automatic claim verification. It is collected from 26 fact checking websites in English, paired with textual sources and rich metadata, and labelled for veracity by human expert journalists.
We present an in-depth analysis of the dataset, highlighting characteristics and challenges. 
Further, we present results for automatic veracity prediction, both with established baselines and with a novel method for joint ranking of evidence pages and predicting veracity that outperforms all baselines. Significant performance increases are achieved by encoding evidence, and by modelling metadata.
Our best-performing model achieves a Macro F1 of 49.2\%, showing that this is a challenging testbed for claim veracity prediction.}\blfootnote{\fullcite{augenstein-etal-2019-multifc}}

\section{Introduction}
\label{s:intro}

Misinformation and disinformation are two of the most pertinent and difficult challenges of the information age, exacerbated by the popularity of social media. In an effort to counter this, a significant amount of manual labour has been invested in fact checking claims, often collecting the results of these manual checks on fact checking portals or websites such as politifact.com or snopes.com. In a parallel development, researchers have recently started to view fact checking as a task that can be partially automated, using machine learning and NLP to automatically predict the \textit{veracity} of claims. 
However, existing efforts either use small datasets consisting of naturally occurring claims (e.g. \cite{mihalcea2009lie,zubiaga2016analysing}), or datasets consisting of artificially constructed claims such as FEVER (\cite{thorne-etal-2018-fever}). While the latter offer valuable contributions to further automatic claim verification work, they cannot replace real-world datasets.


\begin{table}[]
\fontsize{10}{10}\selectfont
\centering
\begin{tabular}{p{2.1cm}l}   
\toprule
Feature & Value \\ \midrule
ClaimID & farg-00004 \\
Claim & Mexico and Canada assemble cars with foreign parts and send them to the U.S. with no tax. \\
Label & distorts \\
Claim URL & \footnotesize{\url{https://www.factcheck.org/2018/10/factchecking-trump-on-trade/}} \\
Reason & None \\
Category & the-factcheck-wire \\
Speaker & Donald Trump \\
Checker & Eugene Kiely \\
Tags & North American Free Trade Agreement \\
Claim Entities & United\_States, Canada, Mexico \\
Article Title & Fact Checking Trump on Trade\\
Publish Date & October 3, 2018 \\
Claim Date & Monday, October 1, 2018 \\ 
\bottomrule
\end{tabular}
\caption{\label{tb:examp_claim} An example of a claim instance. Entities are obtained via entity linking. Article and outlink texts, evidence search snippets and pages are not shown.}
\end{table}


\paragraph{Contributions.} We introduce the currently largest claim verification dataset of naturally occurring claims.\footnote{The dataset is found here: \url{https://copenlu.github.io/publication/2019_emnlp_augenstein/}} It consists of 34,918 claims, collected from 26 fact checking websites in English; evidence pages to verify the claims; the context in which they occurred; and rich metadata (see Table \ref{tb:examp_claim} for an example).
We perform a thorough analysis to identify characteristics of the dataset such as entities mentioned in claims. We demonstrate the utility of the dataset by training state of the art veracity prediction models, and find that evidence pages as well as metadata significantly contribute to model performance. Finally, we propose a novel model that jointly ranks evidence pages and performs veracity prediction. The best-performing model achieves a Macro F1 of 49.2\%, showing that this is a non-trivial dataset with remaining challenges for future work.



\section{Related Work}
\label{s:rw}

\begin{table}[!htbp]
\fontsize{10}{10}\selectfont
    \centering
   \begin{tabular}{l c c c c c c c}
\toprule
\bf Dataset & \bf \# Claims & \bf Labels & \bf metadata & \bf Claim Sources\\
\midrule
\multicolumn{3}{l}{\bf I: Veracity prediction w/o evidence} & & & \\
\cite{wang2017liar} & 12,836 & 6  & Yes & Politifact\\
\cite{C18-1287}& 980 & 2  & No & News Websites \\
\midrule
\multicolumn{3}{l}{\bf II: Veracity} & &  & \\
\cite{bachenko2008verification} & 275 & 2 & No & Criminal Reports \\
\cite{mihalcea2009lie}& 600 & 2 & No & Crowd Authors \\
\cite{mitra2015credbank}$\dagger$ & 1,049 & 5 & No & Twitter \\
\cite{ciampaglia2015computational}$\dagger$ & 10,000 & 2 &  No & Google, Wikipedia \\
\cite{PopatMSW16} & 5,013 & 2  & Yes & Wikipedia, Snopes \\
\cite{2018arXiv180901286S}$\dagger$ &  23,921 & 2 & Yes & Politifact, gossipcop.com \\
Datacommons Fact Check\footnote{https://datacommons.org/factcheck/download} & 10,564 & 2-6 &  Yes & Fact Checking Websites \\
 \midrule
\multicolumn{3}{l}{\bf III: Veracity (evidence encouraged, but not provided)} & & \\
\cite{thorne-etal-2018-fever}$\dagger$& 185,445 & 3 & No & Wikipedia \\
\cite{barron2018overview} & 150 & 3 &  No & factcheck.org, Snopes \\
\midrule
 \multicolumn{3}{l}{\bf IV: Veracity + stance} & & \\
\cite{vlachos2014fact}& 106 & 5 & Yes & Politifact, Channel 4 News \\
\cite{zubiaga2016analysing}& 330 & 3  & Yes & Twitter \\
\cite{derczynski2017semeval}& 325 & 3  & Yes & Twitter \\
\cite{baly-etal-2018-integrating} & 422 & 2  & No & ara.reuters.com, verify-sy.com\\
\midrule \midrule
  \multicolumn{3}{l}{\bf V: Veracity + evidence relevancy} & & \\
MultiFC & 36,534 & 2-40 & Yes & Fact Checking Websites & \\
\bottomrule
\end{tabular}
    \caption{\label{tab:Datasets} Comparison of fact checking datasets. $\dagger$ indicates claims are not ``naturally occuring'': \cite{mitra2015credbank} use events as claims; \cite{ciampaglia2015computational} use DBPedia tiples as claims; \cite{2018arXiv180901286S} use tweets as claims; and \cite{thorne-etal-2018-fever} rewrite sentences in Wikipedia as claims.}
\end{table}

\subsection{Datasets}

Over the past few years, a variety of mostly small datasets related to fact checking have been released. An overview over core datasets is given in Table \ref{tab:Datasets}, and a version of this table extended with the number of documents, source of annotations and SoA performances can be found in the appendix (Table \ref{tab:Datasets2}). The datasets can be grouped into four categories (I--IV). Category I contains datasets aimed at testing how well the veracity\footnote{We use \emph{veracity}, \emph{claim credibility}, and \emph{fake news} prediction interchangeably here -- these terms are often conflated in the literature and meant to have the same meaning.} of a claim can be predicted using the claim alone, without context or evidence documents.
Category II contains datasets bundled with documents related to each claim --  either topically related to provide context, or serving as evidence. Those documents are, however, not annotated.
Category III is for predicting veracity; they encourage retrieving evidence documents as part of their task description, but do not distribute them. 
Finally, category IV comprises datasets annotated for both veracity and stance. Thus, every document is annotated with a label indicating whether the document supports or denies the claim, or is unrelated to it. Additional labels can then be added to the datasets to better predict veracity, for instance by jointly training stance and veracity prediction models. 

Methods not shown in the table, but related to fact checking, are stance detection for claims (\cite{DBLP:conf/naacl/FerreiraV16,PomerleauRao,augenstein-etal-2016-stance,kochkina-etal-2017-turing,augenstein-etal-2016-usfd,journals/ipm/ZubiagaKLPLBCA18,journals/corr/RiedelASR17}), satire detection (\cite{W16-0802}), clickbait detection (\cite{KGNK2017}), conspiracy news detection (\cite{DBLP:journals/corr/TacchiniBVMA17}), rumour cascade detection (\cite{Vosoughi1146}) and claim perspectives detection (\cite{chenseeing}).

Claims are obtained from a variety of sources, including Wikipedia, Twitter, criminal reports and fact checking websites such as politifact.com and snopes.com. The same goes for documents -- these are often websites obtained through Web search queries, or Wikipedia documents, tweets or Facebook posts.
Most datasets contain a fairly small number of claims, and those that do not, often lack evidence documents. An exception is \cite{thorne-etal-2018-fever}, who create a Wikipedia-based fact checking dataset. While a good testbed for developing deep neural architectures, their dataset is artificially constructed and can thus not take metadata about claims into account.

{\bf Contributions:} We provide a dataset that, uniquely among extant datasets, contains a large number of \emph{naturally occurring} claims 
and rich additional meta-information. 

\subsection{Methods}

Fact checking methods partly depend on the type of dataset used. Methods only taking into account claims typically encode those with CNNs or RNNs (\cite{P17-2067,C18-1287}), and potentially encode metadata (\cite{P17-2067}) in a similar way. Methods for small datasets often use hand-crafted features that are a mix of bag of word and other lexical features, e.g.\ LIWC, and then use those as input to a SVM or MLP (\cite{mihalcea2009lie,C18-1287,baly-etal-2018-integrating}). Some use additional Twitter-specific features (\cite{enayet2017niletmrg}). 
More involved methods taking into account evidence documents, often trained on larger datasets, consist of evidence identification and ranking following a neural model that measures the compatibility between claim and evidence (\cite{thorne-etal-2018-fever,DBLP:conf/aaai/MihaylovaNMBMKG18,yin-roth-2018-twowingos}). 

{\bf Contributions:} The latter category above is the most related to our paper as we consider evidence documents. However, existing models are not trained jointly for evidence identification, or for stance and veracity prediction, but rather employ a pipeline approach. 
Here, we show that a joint approach that learns to weigh evidence pages by their importance for veracity prediction can improve downstream veracity prediction performance.




\section{Dataset Construction}
\label{s:crawl}



We crawled a total of 43,837 claims with their metadata (see details in Table \ref{tb:stats4all}). 
We present the data collection in terms of selecting sources, crawling claims and associated metadata (Section \ref{ss:sources}); retrieving evidence pages; 
and linking entities in the crawled claims (Section \ref{ss:ent}).


\subsection{Selection of sources}
\label{ss:sources}
We crawled all active fact checking websites in English listed by Duke Reporters' Lab\footnote{\url{https://reporterslab.org/fact-checking/}} and on the Fact Checking Wikipedia page.\footnote{\url{https://en.wikipedia.org/wiki/Fact_checking}} This resulted in 38 websites in total (shown in Table \ref{tb:stats4all}). Ten websites could not be crawled, as further detailed in Table \ref{tb:not_cralwed_list}.
In the later experimental descriptions, we refer to the part of the dataset crawled from a specific fact checking website as a \textit{domain}, and we refer to each website as \textit{source}.

From each source, we crawled the ID, claim, label, URL, reason for label, categories, person making the claim (speaker), person fact checking the claim (checker), tags, article title, publication date, claim date, as well as the full text that appears when the claim is clicked. Lastly, the above full text contains hyperlinks, so we further crawled the full text that appears when each of those hyperlinks are clicked (outlinks).

There were a number of crawling issues, e.g. security protection of websites with SSL/TLS protocols, time out, URLs that pointed to pdf files instead of HTML content, or unresolvable encoding. In all of these cases, the content could not be retrieved.
For some websites, no veracity labels were available, in which case, they were not selected as domains for training a veracity prediction model. Moreover, not all types of metadata (category, speaker, checker, tags, claim date, publish date) were available for all websites; and availability of articles and full texts differs as well.

We performed semi-automatic cleansing of the dataset as follows. First, we double-checked that the veracity labels would not appear in claims. For some domains, the first or last sentence of the claim would sometimes contain the veracity label, in which case we would discard either the full sentence or part of the sentence. Next, we checked the dataset for duplicate claims. We found 202 such instances, 69 of them with different labels. Upon manual inspection, this was mainly due to them appearing on different websites, with labels not differing much in practice (e.g. `Not true', vs. `Mostly False'). We made sure that all such duplicate claims would be in the training split of the dataset, so that the models would not have an unfair advantage. Finally, we performed some minor manual merging of label types for the same domain where it was clear that they were supposed to denote the same level of veracity (e.g. `distorts', `distorts the facts').


This resulted in a total of 36,534 claims with their metadata.
For the purposes of fact verification, we discarded instances with labels that occur fewer than 5 times, resulting in 34,918 claims. The number of instances, as well as labels per domain, are shown in Table \ref{tab:results_per-domain} and label names in Table \ref{tb:dataset_labels_full} in the appendix. The dataset is split into a training part (80\%) and a development and testing part (10\% each) in a label-stratified manner.
Note that the domains vary in the number of labels, ranging from 2 to 27. Labels include both  straight-forward ratings of veracity (`correct', `incorrect'), but also labels that would be more difficult to map onto a veracity scale (e.g. `grass roots movement!', `misattributed', `not the whole story'). We therefore do not postprocess label types across domains to map them onto the same scale, 
and rather treat them as is. In the methodology section (Section \ref{s:exp}), we show how a model can be trained on this dataset regardless by framing this multi-domain veracity prediction task as a multi-task learning (MTL) one.

\subsection{Retrieving Evidence Pages}\label{ss:evidence}
\label{ss:retrieval}

The text of each claim is submitted verbatim as a query to the Google Search API (without quotes). The 10 most highly ranked search results are retrieved, for each of which we save the title; Google search rank; URL; time stamp of last update; search snippet; as well as the full Web page. We acknowledge that search results change over time, which might have an effect on veracity prediction. However, studying such temporal effects is outside the scope of this paper. Similar to Web crawling claims, as described in Section \ref{ss:sources}, the corresponding Web pages can in some cases not be retrieved, in which case fewer than 10 evidence pages are available.
The resulting evidence pages are from a wide variety of URL domains, though with a predictable skew towards popular websites, such as Wikipedia or The Guardian (see Table \ref{tb:domain_percent} in the appendix for detailed statistics).

\subsection{Entity Detection and Linking}
\label{ss:ent}

\begin{figure}
\centering
\includegraphics[width=0.7\textwidth]{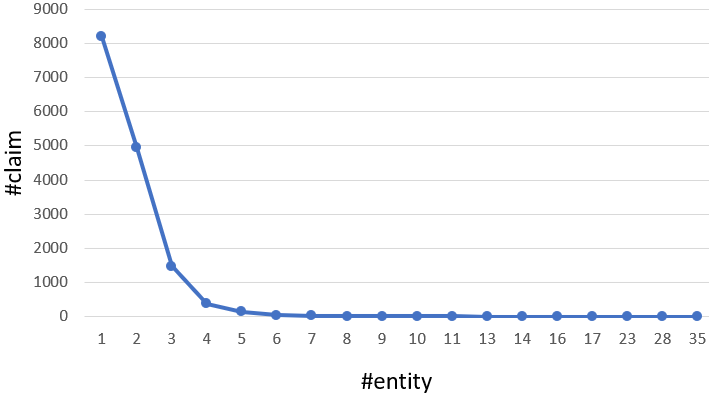}
\caption{\label{fig:claim_distru4entity} Distribution of entities in claims. }
\end{figure}

To better understand what claims are about, we conduct entity linking 
for all claims. Specifically, mentions of people, places, organisations, and other named entities within a claim are recognised and linked to their respective Wikipedia pages, if available. Where there are different entities with the same name, they are disambiguated. For this, we apply the state-of-the-art neural entity linking model by \cite{kolitsas2018end}. 
This results
in a total of 25,763 entities detected and linked to Wikipedia, with a total of 15,351 claims involved, 
meaning that 42\% of all claims contain entities that can be linked to Wikipedia. Later on, we use entities as additional metadata (see Section \ref{ss:metadata}). 
The distribution of claim numbers according to the number of entities they contain is shown in Figure \ref{fig:claim_distru4entity}. We observe that the majority of claims have one to four entities, and the maximum number of 35 entities occurs in one claim only. Out of the 25,763 entities, 2,767 are unique entities. The top 30 most frequent entities are listed in Table \ref{tab:entiyFreq}. This clearly shows that most of the claims involve entities related to the United States, which is to be expected, as most of the fact checking websites are US-based.

\begin{table}
\centering
\fontsize{10}{10}\selectfont
\begin{tabular}{lr}
\toprule
\textbf{Entity} & \textbf{Frequency} \\ 
\midrule
United\_States & 2810 \\
Barack\_Obama & 1598 \\
Republican\_Party\_(United\_States) & 783 \\
Texas & 665 \\
Democratic\_Party\_(United\_States) & 560 \\
Donald\_Trump & 556 \\
Wisconsin & 471 \\
United\_States\_Congress & 354 \\
Hillary\_Rodham\_Clinton & 306 \\
Bill\_Clinton & 292 \\
California & 285 \\
Russia & 275 \\
Ohio & 239 \\
China & 229 \\
George\_W.\_Bush & 208 \\
Medicare\_(United\_States) & 206 \\
Australia & 186 \\
Iran & 183 \\
Brad\_Pitt & 180 \\
Islam & 178 \\
Iraq & 176 \\
Canada & 174 \\
White\_House & 166 \\
New\_York\_City & 164 \\
Washington,\_D.C. & 164 \\
Jennifer\_Aniston & 163 \\
Mexico & 158 \\
Ted\_Cruz & 152 \\
Federal\_Bureau\_of\_Investigation & 146 \\
Syria & 130 \\ 
\bottomrule
\end{tabular}
\caption{\label{tab:entiyFreq} Top 30 most frequent entities listed by their Wikipedia URL with prefix omitted}
\end{table}

\section{Claim Veracity Prediction}
\label{s:exp}

We train several models to predict the veracity of claims. Those fall into two categories: those that only consider the claims themselves, and those that encode evidence pages as well. In addition, claim metadata (speaker, checker, linked entities) is optionally encoded for both categories of models, and ablation studies with and without that metadata are shown.
We first describe the base model used in Section \ref{ss:mtl}, followed by introducing our novel evidence ranking and veracity prediction model in Section \ref{ss:ranking}, and lastly the metadata encoding model in Section \ref{ss:metadata}.

\subsection{Multi-Domain Claim Veracity Prediction with Disparate Label Spaces}\label{ss:mtl}
Since not all fact checking websites use the same claim labels (see Table \ref{tab:results_per-domain}, and Table \ref{tb:dataset_labels_full} in the appendix), training a claim veracity prediction model is not entirely straight-forward. One option would be to manually map those labels onto one another. However, since the sheer number of labels is rather large (165), and it is not always clear from the guidelines on fact checking websites how they can be mapped onto one another, we opt to learn how these labels relate to one another as part of the veracity prediction model. To do so, we employ the multi-task learning (MTL) approach inspired by collaborative filtering presented in \cite{augenstein-etal-2018-multi} (\textit{MTL with LEL}--multitask learning with label embedding layer) that excels on pairwise sequence classification tasks with disparate label spaces. 
More concretely, each domain is modelled as its own task in a MTL architecture, and labels are projected into a fixed-length label embedding space. Predictions are then made by taking the dot product between the claim-evidence embeddings and the label embeddings. By doing so, the model implicitly learns how semantically close the labels are to one another, and can benefit from this knowledge when making predictions for individual tasks, which on their own might only have a small number of instances. When making predictions for individual domains/tasks, both at training and at test time, as well as when calculating the loss, a mask is applied such that the valid and invalid labels for that task are restricted to the set of known task labels.

Note that the setting here slightly differs from \cite{augenstein-etal-2018-multi}. There, tasks are less strongly related to one another; for example, they consider stance detection, aspect-based sentiment analysis and natural language inference. Here, we have different domains, as opposed to conceptually different tasks, but use their framework, as we have the same underlying problem of disparate label spaces.
A more formal problem definition follows next, as our evidence ranking and veracity prediction model in Section \ref{ss:ranking} then builds on it.

\subsubsection{Problem Definition}

We 
frame our problem as a multi-task learning one, where access to labelled datasets for $T$ tasks $\mathcal{T}_1, \ldots, \mathcal{T}_T$ is given at training time with a target task $\mathcal{T}_T$ that is of particular interest. The training dataset for task $\mathcal{T}_i$ consists of $N$ examples $X_{\mathcal{T}_i} = \{x_1^{\mathcal{T}_i}, \ldots, x_{N}^{\mathcal{T}_i}\}$ and their labels $Y_{\mathcal{T}_i} = \{\mathbf{y}_1^{\mathcal{T}_i}, \ldots, \mathbf{y}_{N}^{\mathcal{T}_i}\}$.
The base model is a classic deep neural network MTL model (\cite{Caruana:93}) that shares its parameters across tasks and has task-specific softmax output layers that output a probability distribution $\mathbf{p}^{\mathcal{T}_i}$ for task $\mathcal{T}_i$:

\begin{equation}
\mathbf{p}^{\mathcal{T}_i} = \mathrm{softmax}(\mathbf{W}^{\mathcal{T}_i}\mathbf{h} + \mathbf{b}^{\mathcal{T}_i})
\end{equation}

\noindent where $\mathrm{softmax}(\mathbf{x}) = e^{\mathbf{x}} / \sum^{ \|\mathbf{x}\| }_{i=1} e^{\mathbf{x}_i}$, $\mathbf{W}^{\mathcal{T}_i} \in \mathbb{R}^{L_i \times h}$, $\mathbf{b}^{\mathcal{T}_i} \in \mathbb{R}^{L_i}$ is the weight matrix and bias term of the output layer of task $\mathcal{T}_i$ respectively, $\mathbf{h} \in \mathbb{R}^h$ is the jointly learned hidden representation, $L_i$ is the number of labels for task $\mathcal{T}_i$, and $h$ is the dimensionality of $\mathbf{h}$.
The MTL model is trained to minimise the sum of individual task losses $\mathcal{L}_1 + \ldots + \mathcal{L}_T$ using a negative log-likelihood objective.






\paragraph{Label Embedding Layer.}

To learn the relationships between labels, a Label Embedding Layer (LEL) embeds labels of all tasks in a joint Euclidian space. Instead of training separate softmax output layers as above, a label compatibility function $c(\cdot, \cdot)$ measures how similar a label with embedding $\mathbf{l}$ is to the hidden representation $\mathbf{h}$:

\begin{equation}\label{eq:labelemb1}
c(\mathbf{l},\mathbf{h}) = \mathbf{l} \cdot \mathbf{h}
\end{equation}

\noindent where $\cdot$ is the dot product. Padding is applied such that $l$ and $h$ have the same dimensionality. 
Matrix multiplication and softmax are used for making predictions:
\begin{equation}
\mathbf{p} = \mathrm{softmax}(\mathbf{L} \mathbf{h})
\end{equation}

\noindent where $\mathbf{L} \in \mathbb{R}^{(\sum_i L_i) \times l}$ is the label embedding matrix for all tasks and $l$ is the dimensionality of the label embeddings. 
We apply a task-specific mask to $\mathbf{L}$ in order to obtain a task-specific probability distribution $\mathbf{p}^{\mathcal{T}_i}$. The LEL is shared across all tasks, which allows the model to learn the relationships between labels in the joint embedding space. 




\subsection{Joint Evidence Ranking and Claim Veracity Prediction}\label{ss:ranking}

\begin{figure}[!t]
      \centering
         \includegraphics[width=0.6\linewidth]{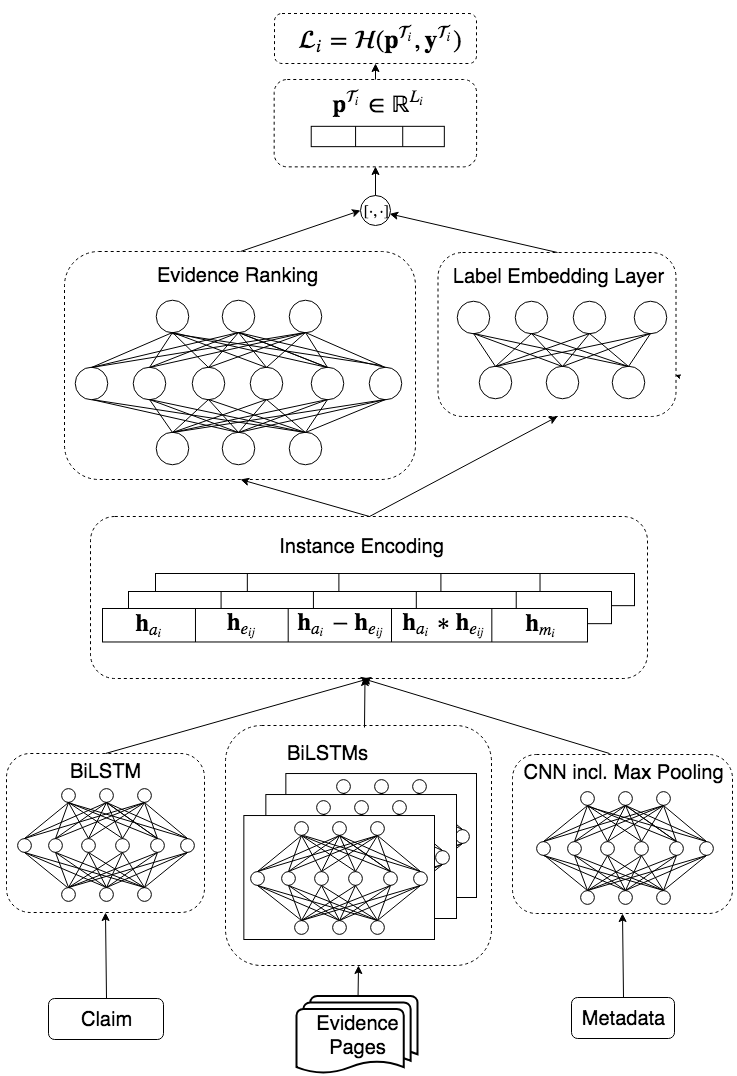}
    \caption{The Joint Veracity Prediction and Evidence Ranking model, shown for one task.}
\label{fig:training-procedures}
\end{figure}

So far, we have ignored the issue of how to obtain claim representation, as the base model described in the previous section is agnostic to how instances are encoded.
A very simple approach, which we report as a baseline, is to encode claim texts only. Such a model ignores evidence for and against a claim, and ends up guessing the veracity based on surface patterns observed in the claim texts.

We next introduce two variants of evidence-based veracity prediction models that encode 10 pieces of evidence in addition to the claim. Here, we opt to encode search snippets as opposed to whole retrieved pages. While the latter would also be possible, it comes with a number of additional challenges, such as encoding large documents, parsing tables or PDF files, and encoding images or videos on these pages, which we leave to future work. Search snippets also have the benefit that they already contain summaries of the part of the page content that is most related to the claim.

\subsubsection{Problem Definition}

Our problem is to obtain encodings for $N$ examples $X_{\mathcal{T}_i} = \{x_1^{\mathcal{T}_i}, \ldots, x_{N}^{\mathcal{T}_i}\}$. For simplicity, we will henceforth drop the task superscript and refer to instances as $X = \{x_1, \ldots, x_N\}$, as instance encodings are learned in a task-agnostic fashion.
Each example further consists of a claim $a_i$ and $k=10$ evidence pages $E_k = \{e_{1_{0}}, \ldots, e_{N_{10}}\}$.

Each claim and evidence page is encoded with a BiLSTM to obtain a sentence embedding, which is the concatenation of the last state of the forward and backward reading of the sentence, i.e. $\mathbf{h} = BiLSTM(\cdot)$, where $\mathbf{h}$ is the sentence embedding.

Next, we want to combine claims and evidence sentence embeddings into joint instance representations.
In the simplest case, referred to as model variant \textit{crawled\_avg}, we mean average the BiLSTM sentence embeddings of all evidence pages (signified by the overline) and concatenate those with the claim embeddings, i.e. 

\begin{equation}
\mathbf{s}_{{g}_i} = [ \mathbf{h}_{a_i};\overline{\mathbf{h}_{E_i}} ]
\end{equation}

\noindent where $s_{{g}_i}$ is the resulting encoding for training example $i$ and $[\cdot ; \cdot]$ denotes vector concatenation.
However, this has the disadvantage that all evidence pages are considered equal.

\paragraph{Evidence Ranking}
The here proposed alternative instance encoding model, \textit{crawled\_ranked}, which achieves the highest overall performance as discussed in Section \ref{s:results}, learns the compatibility between an instance's claim and each evidence page. It ranks evidence pages by their utility for the veracity prediction task, and then uses the resulting ranking to obtain a weighted combination of all claim-evidence pairs. No direct labels are available to learn the ranking of individual documents, only for the veracity of the associated claim, so the model has to learn evidence ranks implicitly.

To combine claim and evidence representations, we use the matching model proposed for the task of natural language inference by \cite{conf/acl/MouMLX0YJ16} and adapt it to combine an instance's claim representation with each evidence representation, i.e.

\begin{equation}
s_{{r}_{i_j}} = [ \mathbf{h}_{a_i};\mathbf{h}_{e_{i_j}};\mathbf{h}_{a_i} - \mathbf{h}_{e_{i_j}};\mathbf{h}_{a_i} \cdot \mathbf{h}_{e_{i_j}} ]
\end{equation}

\noindent where $s_{{r}_{i_j}}$ is the resulting encoding for training example $i$ and evidence page $j$ , $[\cdot ; \cdot]$ denotes vector concatenation, and $\cdot$ denotes the dot product.

All joint claim-evidence representations $\mathbf{s}_{{r}_{i_0}} , \ldots, \mathbf{s}_{{r}_{i_{10}}}$ are then projected into the binary space via a fully connected layer $\mathrm{FC}$, followed by a non-linear activation function $f$, to obtain a soft ranking of claim-evidence pairs, in practice a 10-dimensional vector,

\begin{equation}\label{eq:labelemb2}
\mathbf{o}_i = [ f(\mathrm{FC}( s_{{r}_{i_0}} )) ; \ldots ; f(\mathrm{FC}( s_{{r}_{i_{10}}} ))] 
\end{equation}

\noindent where $[\cdot ; \cdot]$ denotes concatenation.

Scores for all labels are obtained as per (\ref{eq:labelemb2}) above, with the same input instance embeddings as for the evidence ranker, i.e. $s_{{r}_{i_j}}$.
Final predictions for all claim-evidence pairs are then obtained by taking the dot product between the label scores and binary evidence ranking scores, i.e.

\begin{equation}\label{eq:labelemb}
\mathbf{p}_i = \mathrm{softmax}( c(\mathbf{l},\mathbf{h}) \cdot \mathbf{o}_i )
\end{equation}

\noindent Note that the novelty here is that, unlike for the model described in \cite{conf/acl/MouMLX0YJ16}, we have no direct labels for learning weights for this matching model. Rather, our model has to implicitly learn these weights for each claim-evidence pair in an end-to-end fashion given the veracity labels.

\begin{table}
\centering
\fontsize{10}{10}\selectfont
\begin{tabular}{@{}lcc@{}}
\toprule
\textbf{Model} & \textbf{Micro F1} & \textbf{Macro F1} \\ \midrule
claim-only & 0.469 & 0.253 \\
claim-only\_embavg & 0.384 & 0.302 \\
crawled-docavg & 0.438 & 0.248 \\
crawled\_ranked & 0.613 & 0.441 \\
\midrule
claim-only + meta & 0.494 & 0.324 \\
claim-only\_embavg + meta & 0.418 & 0.333 \\
crawled-docavg + meta & 0.483 & 0.286 \\
crawled\_ranked + meta & \bf 0.611 & \bf 0.459 \\
\bottomrule
\end{tabular}
\caption{\label{tab:results_main} Results with different model variants on the test set, ``meta'' means all metadata is used.}\end{table}

\subsection{Metadata}\label{ss:metadata}

We experiment with how useful claim metadata is, and encode the following as one-hot vectors: speaker, category, tags and linked entities. We do not encode `Reason' as it gives away the label, and do not include `Checker' as there are too many unique checkers for this information to be relevant. The claim publication date is potentially relevant, but it does not make sense to merely model this as a one-hot feature, so we leave incorporating temporal information to future work.
Since all metadata consists of individual words and phrases, a sequence encoder is not necessary, and we opt for a CNN followed by a max pooling operation as used in \cite{P17-2067} to encode metadata for fact checking.
The max-pooled metadata representations, denoted $h_m$, are then concatenated with the instance representations, e.g. for the most elaborate model, \textit{crawled\_ranked}, these would be concatenated with $s_{{cr}_{i_j}}$.

\section{Experiments}\label{s:results}

\subsection{Experimental Setup}

The base sentence embedding model is a BiLSTM over all words in the respective sequences with randomly initialised word embeddings, following \cite{augenstein-etal-2018-multi}. 
We opt for this strong baseline sentence encoding model, as opposed to engineering sentence embeddings that work particularly well for this dataset, to showcase the dataset. We would expect pre-trained contextual encoding models, e.g. ELMO (\cite{peters-etal-2018-deep}), ULMFit (\cite{howard-ruder-2018-universal}), BERT (\cite{devlin-etal-2019-bert}), to offer complementary performance gains, as has been shown for a few recent papers (\cite{conf/emnlp/WangSMHLB18,conf/acl/RajpurkarJL18}).

For claim veracity prediction without evidence documents with the MTL with LEL model, we use the following sentence encoding variants: \textit{claim-only}, which uses a BiLSTM-based sentence embedding as input, and \textit{claim-only\_embavg}, which uses a sentence embedding based on mean averaged word embeddings as input.

We train one multi-task model per task (i.e., one model per domain).
We perform a grid search over the following hyperparameters, tuned on the respective dev set, and evaluate on the correspoding test set (final settings are underlined): 
word embedding size [64, \underline{128}, 256], BiLSTM hidden layer size [64, \underline{128}, 256], number of BiLSTM hidden layers [1, \underline{2}, 3], BiLSTM dropout on input and output layers [0.0, \underline{0.1}, 0.2, 0.5], word-by-word-attention for BiLSTM with window size 10 (\cite{bahdanau2014neural}) [True, \underline{False}], skip-connections for the BiLSTM [\underline{True}, False], batch size [\underline{32}, 64, 128], label embedding size [\underline{16}, 32, 64]. We use ReLU as an activation function for both the BiLSTM and the CNN. 
For the CNN, the following hyperparameters are used: number filters [\underline{32}], kernel size [\underline{32}]. 
We train using cross-entropy loss and the RMSProp optimiser with initial learning rate of $0.001$ and perform early stopping on the dev set with a patience of $3$.

\subsection{Results}

\begin{table}
\centering
\fontsize{10}{10}\selectfont
\begin{tabular}{@{}lrccc@{}}
\toprule
\textbf{Domain} & \bf \# Insts &\bf  \# Labs & \textbf{Micro F1} & \textbf{Macro F1} \\ \midrule
ranz & 21 & 2 & 1.000 & 1.000 \\
bove & 295 & 2 & 1.000 & 1.000 \\
abbc & 436 & 3 & 0.463 & 0.453 \\
huca & 34 & 3 & 1.000 & 1.000 \\
mpws & 47 & 3 & 0.667 & 0.583 \\ 
peck & 65 & 3 & 0.667 & 0.472 \\
faan & 111 & 3 & 0.682 & 0.679 \\
clck & 38 & 3 & 0.833 & 0.619 \\ 
fani & 20 & 3 & 1.000 & 1.000 \\ 
chct & 355 & 4 & 0.550 & 0.513 \\
obry & 59 & 4 & 0.417 & 0.268 \\
vees & 504 & 4 & 0.721 & 0.425 \\
faly & 111 & 5 & 0.278 & 0.5 \\ 
goop & 2943 & 6 & 0.822 & 0.387 \\ 
pose & 1361 & 6 & 0.438 & 0.328 \\
thet & 79 & 6 & 0.55 & 0.37 \\
thal & 163 & 7 & 1.000 & 1.000 \\ 
afck & 433 & 7 & 0.357 & 0.259 \\
hoer & 1310 & 7 & 0.694 & 0.549 \\
para & 222 & 7 & 0.375 & 0.311 \\
wast & 201 & 7 & 0.344 & 0.214 \\
vogo & 654 & 8 & 0.594 & 0.297 \\ 
pomt & 15390 & 9 & 0.321 & 0.276 \\ 
snes & 6455 & 12 & 0.551 & 0.097 \\
farg & 485 & 11 & 0.500 & 0.140 \\ 
tron & 3423 & 27 & 0.429 & 0.046 \\ 
\midrule
avg &  & 7.17 & 0.625 & 0.492 \\ 
\bottomrule
\end{tabular}
\caption{\label{tab:results_per-domain} Total number of instances and unique labels per domain, as well as per-domain results with model \textit{crawled\_ranked + meta}, sorted by label size}
\end{table}

\begin{table}
\centering
\fontsize{10}{10}\selectfont
\begin{tabular}{@{}lcc@{}}
\toprule
\bf Metadata & \bf Micro F1 & \bf Macro F1 \\
\midrule
None & \bf 0.627 & 0.441 \\
\midrule
Speaker & 0.602 & 0.435 \\
  + Tags & 0.608 & 0.460 \\
\midrule
Tags & 0.585 & 0.461 \\
\midrule
Entity & 0.569 & 0.427 \\
  + Speaker & 0.607 & 0.477 \\
  + Tags & 0.625 & \bf 0.492 \\
\bottomrule
\end{tabular}
\caption{\label{tab:results_meta-ablation} Ablation results with base model \textit{crawled\_ranked} for different types of metadata}
\end{table}

\begin{table}
\centering
\fontsize{10}{10}\selectfont
\begin{tabular}{@{}lcc@{}}
\toprule
\bf Model & \bf Micro F1 & \bf Macro F1 \\
\midrule
STL & 0.527 & 0.388 \\
MTL & 0.556 & 0.448 \\
MTL + LEL & \bf 0.625 & \bf 0.492 \\
\bottomrule
\end{tabular}
\caption{\label{tab:results_training-ablation} Ablation results with \textit{crawled\_ranked + meta} encoding for STL vs. MTL vs. MTL + LEL training}
\end{table}

For each domain, we compute the Micro as well as Macro F1, then mean average results over all domains. Core results with all vs. no metadata are shown in Table \ref{tab:results_main}.
We first experiment with different base model variants and find that label embeddings improve results, and that the best proposed models utilising multiple domains outperform single-task models (see Table \ref{tab:results_training-ablation}). This corroborates the findings of \cite{augenstein-etal-2018-multi}.
Per-domain results with the best model are shown in Table \ref{tab:results_per-domain}. Domain names are from hereon after abbreviated for brevity, see Table \ref{tb:stats4all} in the appendix for correspondences to full website names. Unsurprisingly, it is hard to achieve a high Macro F1 for domains with many labels, e.g. tron and snes. Further, some domains, surprisingly mostly with small numbers of instances, seem to be very easy -- a perfect Micro and Macro F1 score of 1.0 is achieved on ranz, bove, buca, fani and thal. 
We find that for those domains, the verdict is often already revealed as part of the claim using explicit wording.

\paragraph{Claim-Only vs. Evidence-Based Veracity Prediction.}
Our evidence-based claim veracity prediction models outperform claim-only veracity prediction models by a large margin. Unsurprisingly, \textit{claim-only\_embavg} is outperformed by \textit{claim-only}. Further, \textit{crawled\_ranked} is our best-performing model in terms of Micro F1 and Macro F1, meaning that our model captures that not every piece of evidence is equally important, and can utilise this for veracity prediction.

\paragraph{Metadata.}

We perform an ablation analysis of how metadata impacts results, shown in Table \ref{tab:results_meta-ablation}. 
Out of the different types of metadata, topic tags on their own contribute the most. This is likely because they offer highly complementary information to the claim text of evidence pages. Only using all metadata together achieves a higher Macro F1 at similar Micro F1 than using no metadata at all. 
To further investigate this, we split the test set into those instances for which no metadata is available vs. those for which metadata is available. We find that encoding metadata within the model hurts performance for domains where no metadata is available, but improves performance where it is. In practice, an ensemble of both types of models would be sensible, as well as exploring more involved methods of encoding metadata.  



\begin{figure}
\centering
\includegraphics[width=0.7\linewidth]{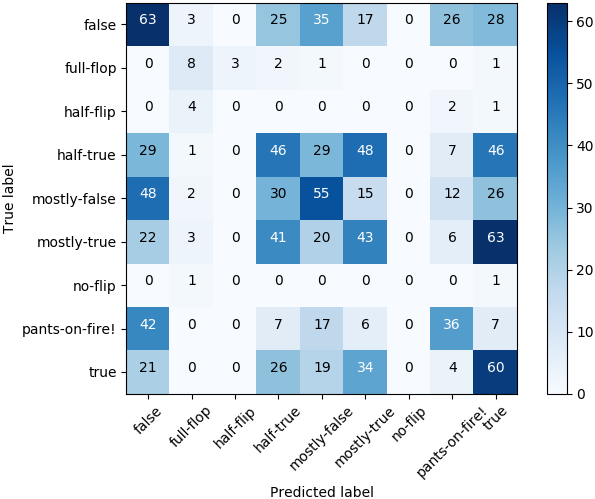}
\caption{\label{fig:confusion_matrix} Confusion matrix of predicted labels with best-performing model, \textit{crawled\_ranked + meta}, on the `pomt' domain}
\end{figure}

\section{Analysis and Discussion}

An analysis of labels frequently confused with one another, for the largest domain `pomt' and best-performing model \textit{crawled\_ranked + meta} is shown in Figure \ref{fig:confusion_matrix}. The diagonal represents when gold and predicted labels match, and the numbers signify the number of test instances. One can observe that the model struggles more to detect claims with labels `true' than those with label `false'. Generally, many confusions occur over close labels, e.g. `half-true' vs. `mostly true'. 

We further analyse what properties instances that are predicted correctly vs. incorrectly have, using the model \textit{crawled\_ranked meta}.
We find that, unsurprisingly, longer claims are harder to classify correctly, and that claims with a high direct token overlap with evidence pages lead to a high evidence ranking.
When it comes to frequently occurring tags and entities, 
very general tags such as `government-and-politics' or `tax' that do not give away much, frequently co-occur with incorrect predictions, whereas more specific tags such as `brisbane-4000' or `hong-kong' tend to co-occur with correct predictions. 
Similar trends are observed for bigrams.
This means that the model has an easy time succeeding for instances where the claims are short, where specific topics tend to co-occur with certain veracities, and where evidence documents are highly informative. Instances with longer, more complex claims where evidence is ambiguous remain challenging.

\section{Conclusions}

We present a new, real-world fact checking dataset, currently the largest of its kind. It consists of 34,918 claims collected from 26 fact checking websites, rich metadata and 10 retrieved evidence pages per claim.
We find that encoding the metadata as well evidence pages helps, and introduce a new joint model for ranking evidence pages and predicting veracity. 

\section*{Acknowledgments}
This research is partially supported by QUARTZ (721321, EU H2020 MSCA-ITN) and DABAI (5153-00004A, Innovation Fund Denmark).
\section{Appendix}

\setlength{\tabcolsep}{0.2em}
\begin{table}[]
\fontsize{10}{12}\selectfont
\begin{tabular}{@{}lll@{}}
\toprule
 & \bf Websites (Sources)                                                           & \bf Reason                                                                                              \\ \midrule
 & Mediabiasfactcheck                                                           & Website that checks other news websites                                                             \\
 & CBC                                                                          & No pattern to crawl                                                                                 \\
 & apnews.com/APFactCheck                                                       & No categorical label and no structured claim                                                        \\
 & weeklystandard.com/tag/fact-check                                            & Mostly no label, and they are placed anywhere \\
 & ballotpedia.org                                                              & No categorical label and no structured claim                                                        \\
 & channel3000.com/news/politics/reality-check                                  & No categorical label, lack of structure, and no clear claim                                         \\
 & npr.org/sections/politics-fact-check & No label and no clear claim (only some titles are claims) \\
 & dailycaller.com/buzz/check-your-fact                                         & Is a subset of checkyourfact which has already been crawled                                         \\
 & sacbee.com\footnote{sacbee.com/news/politics-government/election/california-elections/poligraph/} & Contains very few labelled articles, and without clear claims                      \\
 & TheGuardian & Only a few websites have a pattern for labels. \\ \bottomrule
\end{tabular}
\caption{\label{tb:not_cralwed_list} The list of websites that we did not crawl and reasons for not crawling them.}
\end{table}

\begin{table}[]
\centering
\fontsize{10}{10}\selectfont
\begin{tabular}{@{}lllll@{}}
\toprule
\bf Domain                          & \bf \%       &  &  &  \\ \midrule
https://en.wikipedia.org/       & 4.425 \\
https://www.snopes.com/         & 3.992 \\
https://www.washingtonpost.com/ & 3.025 \\
https://www.nytimes.com/        & 2.478 \\
https://www.theguardian.com/    & 1.807 \\
https://www.youtube.com/        & 1.712 \\
https://www.dailymail.co.uk/    & 1.558 \\
https://www.usatoday.com/       & 1.279 \\
https://www.politico.com/       & 1.241 \\
http://www.politifact.com/      & 1.231 \\
https://www.pinterest.com/      & 1.169 \\
https://www.factcheck.org/      & 1.09  \\
https://www.gossipcop.com/      & 1.073 \\
https://www.cnn.com/            & 1.065 \\
https://www.npr.org/            & 0.957 \\
https://www.forbes.com/         & 0.911 \\
https://www.vox.com/            & 0.89  \\
https://www.theatlantic.com/    & 0.88  \\
https://twitter.com/            & 0.767 \\
https://www.hoax-slayer.net/    & 0.655 \\
http://time.com/                & 0.554 \\
https://www.bbc.com/            & 0.551 \\
https://www.nbcnews.com/        & 0.515 \\
https://www.cnbc.com/           & 0.514 \\
https://www.cbsnews.com/        & 0.503 \\
https://www.facebook.com/       & 0.5   \\
https://www.newyorker.com/      & 0.495 \\
https://www.foxnews.com/        & 0.468 \\
https://people.com/             & 0.439 \\
http://www.cnn.com/             & 0.419                 \\ \bottomrule
\end{tabular}
\caption{\label{tb:domain_percent}The top 30 most frequently occurring URL domains.}
\end{table}

\begin{table}[]
\fontsize{10}{10}\selectfont
\begin{tabular}{p{1.2cm}p{1.2cm}p{1.3cm}p{11cm}}   
\toprule
\bf Domain & \bf \# Insts & \bf \# Labels & \bf Labels \\
\midrule
abbc & 436 & 3 & in-between, in-the-red, in-the-green \\
afck & 433 & 7 & correct, incorrect, mostly-correct, unproven, misleading, understated, exaggerated \\
bove & 295 & 2 & none, rating: false \\
chct & 355 & 4 & verdict: true, verdict: false, verdict: unsubstantiated, none \\
clck & 38 & 3 & incorrect, unsupported, misleading \\
faan & 111 & 3 & factscan score: false, factscan score: true, factscan score: misleading \\
faly & 71 & 5 & true, none, partly true, unverified, false \\
fani & 20 & 3 & conclusion: accurate, conclusion: false, conclusion: unclear \\
farg & 485 & 11 & false, none, distorts the facts, misleading, spins the facts, no evidence, not the whole story, unsupported, cherry picks, exaggerates, out of context \\
goop & 2943 & 6 & 0, 1, 2, 3, 4, 10 \\
hoer & 1310 & 7 & facebook scams, true messages, bogus warning, statirical reports, fake news, unsubstantiated messages, misleading recommendations \\
huca & 34 & 3 & a lot of baloney, a little baloney, some baloney \\
mpws & 47 & 3 & accurate, false, misleading \\
obry & 59 & 4 & mostly\_true, verified, unobservable, mostly\_false \\
para & 222 & 7 & mostly false, mostly true, half-true, false, true, pants on fire!, half flip \\
peck & 65 & 3 & false, true, partially true \\
pomt & 15390 & 9 & half-true, false, mostly true, mostly false, true, pants on fire!, full flop, half flip, no flip \\
pose & 1361 & 6 & promise kept, promise broken, compromise, in the works, not yet rated, stalled \\
ranz & 21 & 2 & fact, fiction \\
snes & 6455 & 12 & false, true, mixture, unproven, mostly false, mostly true, miscaptioned, legend, outdated, misattributed, scam, correct attribution \\
thet & 79 & 6 & none, mostly false, mostly true, half true, false, true \\
thal & 74 & 2 & none, we rate this claim false \\
tron & 3423 & 27 & fiction!, truth!, unproven!, truth! \& fiction!, mostly fiction!, none, disputed!, truth! \& misleading!, authorship confirmed!, mostly truth!, incorrect attribution!, scam!, investigation pending!, confirmed authorship!, commentary!, previously truth! now resolved!, outdated!, truth! \& outdated!, virus!, fiction! \& satire!, truth! \& unproven!, misleading!, grass roots movement!, opinion!, correct attribution!, truth! \& disputed!, inaccurate attribution! \\
vees & 504 & 4 & none, fake, misleading, false \\
vogo & 653 & 8 & none, determination: false, determination: true, determination: mostly true, determination: misleading, determination: barely true, determination: huckster propaganda, determination: false, determination: a stretch \\
wast & 201 & 7 & 4 pinnochios, 3 pinnochios, 2 pinnochios, false, not the whole story, needs context, none \\
\bottomrule
\end{tabular}
\caption{\label{tb:dataset_labels_full} Number of instances, and labels per domain sorted by number of occurrences}
\end{table}

\begin{sidewaystable}
\fontsize{8}{8}\selectfont
\begin{tabular}{llllllllllllll}
\toprule
\bf Website & \bf Domain & \bf Claims & \bf Labels & \bf Category & \bf Speaker & \bf Checker & \bf Tags & \bf Article & \bf Claim date & \bf Publish date & \bf Full text & \bf Outlinks \\ 
\midrule
abc & abbc & 436 & 436 & 436 & - & - & 436 & 436 & - & 436 & 436 & 7676 \\
africacheck & afck & 436 & 436 & - & - & - & - & 436 & - & 436 & 436 & 2325 \\
altnews & - & 496 & - & - & - & 496 & - & 496 & - & 496 & 496 & 6389 \\
boomlive & - & 302 & 302 & - & - & - & - & 302 & - & 302 & 302 & 6054 \\
checkyourfact & chht & 358 & 358 & - & - & 358 & - & - & - & 358 & 358 & 5271 \\
climatefeedback & clck & 45 & 45 & - & - & - & - & 45 & - & 45 & 45 & 489 \\
crikey & - & 18 & 18 & 18 & - & 18 & 18 & 18 & - & 18 & 18 & 212 \\
factcheckni & - & 36 & 36 & 36 & - & - & - & 36 & - & - & 36 & 151 \\
factcheckorg & farg & 512 & 512 & 512 & 512 & 512 & 512 & 512 & 512 & 512 & 512 & 8282 \\
factly & - & 77 & 77 & - & - & - & - & 77 & - & - & 77 & 658 \\
factscan & - & 115 & 115 & - & 115 & - & - & - & 115 & 115 & 115 & 1138 \\
fullfact & - & 336 & 336 & 336 & - & 336 & - & 336 & - & 336 & 336 & 3838 \\
gossipcop & goop & 2947 & 2947 & - & - & 2947 & - & 2947 & - & 2947 & 2947 & 12583 \\
hoaxslayer & hoer & 1310 & 1310 & - & - & 1310 & - & 1310 & - & 1310 & 1310 & 14499 \\
huffingtonpostca & huca & 38 & 38 & - & 38 & 38 & - & 38 & 38 & 38 & 38 & 78 \\
leadstories & - & 1547 & 1547 & - & - & 1547 & - & 1547 & - & 1547 & 1547 & 12015 \\
mprnews & mpws & 49 & 49 & - & - & 49 & - & 49 & - & 49 & 49 & 319 \\
nytimes & - & 17 &  17 & - & - & 17 & - & 17 & - & 17 & 17 & 271 \\
observatory & obry & 60 & 60 & - & - & 60 & - & 60 & - & 60 & 60 & 592 \\
pandora & para & 225 & 225 & 225 & 225 & 225 & - & 225 & - & 225 & 225 & 114 \\
pesacheck & peck & 67 & 67 & - & - & 67 & - & 67 & - & 67 & 67 & 521 \\
politico & - & 102 & 102 & - & - & 102 & - & 102 & - & 102 & 102 & 150 \\
politifact\_promise & pose & 1361 & 1361 & 1361 & 1361 & - & - & 1361 & - & 1361 & 1361 & 6279 \\
politifact\_stmt & pomt & 15390 & 15390 & - & 15390 & - & - & - & 15390 & 15390 & 15390 & 78543 \\
politifact\_story & - & 5460 & - & - & - & 5460 & - & - & - & 5460 & 5460 & 24836 \\
radionz & ranz & 32 & 32 & 32 & 32 & - & - & 32 & 32 & 32 & 32 & 44 \\
snopes & snes & 6457 & 6457 & 6457 & - & 6457 & - & 6457 & - & 6457 & 6457 & 46735 \\
swissinfo & - & 20 & 20 & 20 & 20 & 20 & - & 20 & - & 20 & 20 & 40 \\
theconversation & - & 62 & 62 & 62 & 62 & 62 & 62 & 62 & - & 62 & 62 & 723 \\
theferret & thet & 81 & 81 & 81 & 81) & - & - & 81 & - & 81(81) & 81 & 885 \\
theguardian & - & 155 & 155 & 155 & - & 155 & - & 155 & - & 155 & 155 & 2600 \\
thejournal & thal & 179 & 179 & - & - & - & - & 179 & - & 179 & 179 & 2375 \\
truthorfiction & tron & 3674 & 3674 & 3674 & - & - & 3674 & 3674 & - & 3674 & 3674 & 8268 \\
verafiles & vees & 509 & 509 & - & - & - & 509 & 509 & - & 509 & 509 & 23 \\
voiceofsandiego & vogo & 660 & 660 & - & - & - & - & 660 & - & 660 & 660 & 2352 \\
washingtonpost & wast & 227 & 227 & - & 227 & 227 & - & 227 & - & 227 & 227 & 2470 \\
wral & - & 20 & 20 & - & - & 20 & 20 & 20 & - & 20 & 20 & 355 \\
zimfact & - & 21 & 21 & 21 & 21 & 21 & - & 21 & - & 21 & 21 & 179 \\
\midrule
Total & & 43837 & 43837 & 43837 & 43837 & 43837 & 43837 & 43837 & 43837 & 43837 & 43837 & 260330 \\
\bottomrule
\end{tabular}
\caption{\label{tb:stats4all} Summary statistics for claim collection. ``Domain'' indicates the domain name used for the veracity prediction experiments, ``--'' indicates that the website was not used due to missing or insufficient claim labels, see Section \ref{ss:evidence}.}
\end{sidewaystable}

\begin{sidewaystable}
\fontsize{10}{12}\selectfont
    \centering
   \begin{tabular}{l c c c c c c c}
\toprule
\bf Dataset & \bf \# Claims & \bf Labels & \bf \# Doc & \bf Source of Annos & \bf metadata & \bf Claim Sources & \bf SoA performance \\
\midrule
\multicolumn{3}{l}{\bf Veracity prediction w/o evidence} & & & & & \\
\cite{P17-2067} & 12836 & 6 & no info & Journalists & Yes & Politifact & 38.81 (Acc)$\diamond$ \\
\cite{C18-1287}& 980 & 2 & no info & Crowd annotators & No & News Websites & 76 (Acc) \\
\midrule
\multicolumn{3}{l}{\bf Veracity} & & & & & \\
\cite{bachenko2008verification} & 275 & 2& no info & Linguists & No & Criminal Reports & 0.749 (Acc) \\
\cite{mihalcea2009lie}& 600 & 2& no info & Crowd annotators & No & Crowd Authors & 0.708 (Acc) \\
\cite{mitra2015credbank}$\dagger$ & 1049 & 5 & 60 m & Crowd annotators & No & Twitter &  \\
\cite{ciampaglia2015computational}$\dagger$ & 10000 & 2& no info & Crowd annotation &  No & Google+Wikipedia &  \\
\cite{PopatMSW16} & 5013 & 2 & 133272 & Community/scholars & Yes & Wikipedia, Snopes & 0.7196 (Acc) \\
\cite{DBLP:conf/aaai/MihaylovaNMBMKG18} & 249 & 6 & variable & Paper authors & No & qatarliving.com/forum & 0.7454 (F1), 0.7229 (Acc) \\
\cite{2018arXiv180901286S}$\dagger$ &  23,921 & 2 & 602,659 & Journalists & Yes & Politifact, gossipcop.com & 0.691(Pol),0.822(Gos) (Acc) \\
\cite{santia2018buzzface} & 2263 & 4 & $>$ 1.6 m & Journalists & Yes & News Websites & no info given \\
Datacommons Fact Check\footnote{https://datacommons.org/factcheck/download} & 10564 & 2-6 & no info &  Journalists & Yes & Fact Checking Websites & no info given \\
 \midrule
\multicolumn{3}{l}{\bf Veracity (evidence encouraged, but not provided)} & & & & & \\
\cite{thorne-etal-2018-fever}& 185445 & 3 & no info & Crowd Annotators & No & wikipedia & 54.33(SCOREEV, FC);47.15(F1) $\diamond$ \\
\cite{barron2018overview} & 150 & 3& no info & Journalists & No & factcheck.org, snopes.com & MAE0.705(En);0.658(Arabic) $\diamond$ \\
\midrule
 \multicolumn{3}{l}{\bf Veracity + stance} & & & & & \\
\cite{vlachos2014fact}& 106 & 5 & no info &  Journalists & Yes & Politifact, Channel 4 News & no info given \\
\cite{zubiaga2016analysing}& 330 & 3 & 4,842 & Crowd annotators & Yes & Twitter & no info \\
\cite{derczynski2017semeval}& 325 & 3 & 5,274 & Crowd annotators & Yes & Twitter & 0.536(macro accu+RMSE) $\diamond$ \\
\cite{baly-etal-2018-integrating} & 422 & 2 & 3,042 & Journalists & No & ara.reuters.com, verify-sy.com & 55.8 (Macro F1) $\diamond$\\
\midrule \midrule
  \multicolumn{3}{l}{\bf Veracity + evidence relevancy} & & & & & \\
MultiFC & \claimNum & 2-40 & \docNum & Journalist & Yes & Fact Checking Websites & 45.9 (Macro F1) \\
\bottomrule
\end{tabular}
    \caption{\label{tab:Datasets2} Comparison of fact checking datasets. Doc = all doc types (including tweets, replies, etc.). SoA perform indicates state-of-the-art performance. 
    $\dagger$ indicates that claims are not naturally occuring: \cite{mitra2015credbank} use events as claims; \cite{ciampaglia2015computational} use DBPedia tiples as claims; \cite{2018arXiv180901286S} use tweets as claims; and \cite{thorne-etal-2018-fever} rewrite sentences in Wikipedia as claims.
$\diamond$ denotes that the SoA performance is from other papers. Best performance for \cite{P17-2067} is from \cite{karimi-etal-2018-multi}; \cite{thorne-etal-2018-fever} from \cite{yin-roth-2018-twowingos}; \cite{barron2018overview} from \cite{wang2018copenhagen} in English, \cite{derczynski2017semeval} from \cite{enayet2017niletmrg}; and \cite{baly-etal-2018-integrating} from \cite{hanselowski2017team}. }
\end{sidewaystable}

\chapter{Generating Label Cohesive and Well-Formed Adversarial Claims}\label{ch:adversarial}


\boxabstract{Adversarial attacks reveal important vulnerabilities and flaws of trained models. One potent type of attack are \textit{universal adversarial triggers}, which are individual n-grams that, when appended to instances of a class under attack, can trick a model into predicting a target class. However, for inference tasks such as fact checking, these triggers often inadvertently invert the meaning of instances they are inserted in. In addition, such attacks produce semantically nonsensical inputs, as they simply concatenate triggers to existing samples. Here, we investigate how to generate adversarial attacks against fact checking systems that preserve the ground truth meaning and are semantically valid. We extend the HotFlip attack algorithm used for universal trigger generation by jointly minimizing the target class loss of a fact checking model and the entailment class loss of an auxiliary natural language inference model. We then train a conditional language model to generate semantically valid statements, which include the found universal triggers. We find that the generated attacks maintain the directionality and semantic validity of the claim better than previous work.}\blfootnote{\fullcite{atanasova-etal-2020-generating}}

\section{Introduction}
Adversarial examples~(\cite{goodfellow2015explaining, szegedy2013intriguing}) are deceptive model inputs designed to mislead an ML system into making the wrong prediction. They expose regions of the input space that are outside the training data distribution where the model is unstable. 
It is important to reveal such vulnerabilities and correct for them, especially for tasks such as fact checking (FC). 

\begin{figure}
\centering
\includegraphics[width=0.6 \textwidth]{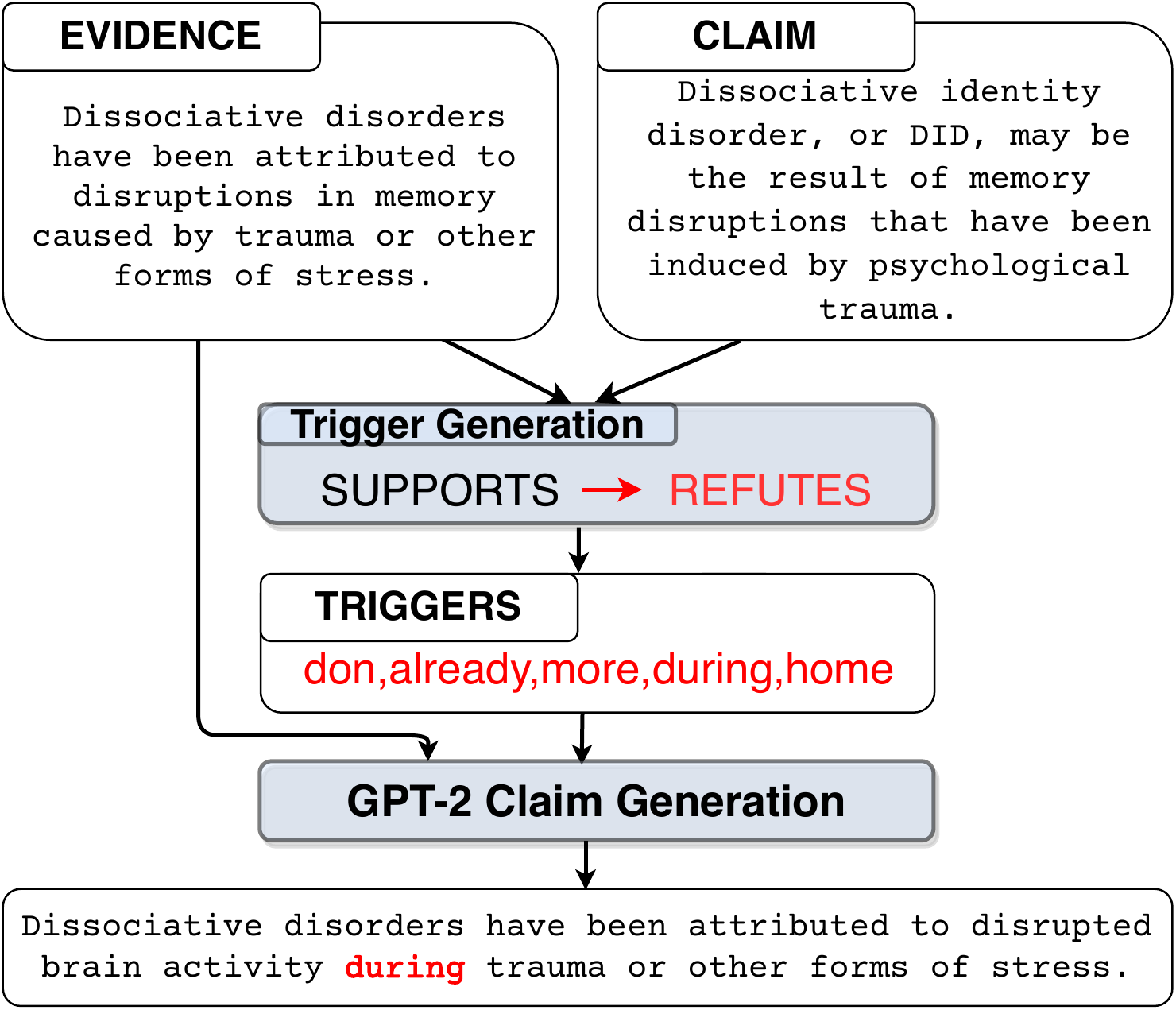}
\caption{High level overview of our method. First, universal triggers are discovered for flipping a source to a target label (e.g. SUPPORTS $\rightarrow$ REFUTES). These triggers are then used to condition the GPT-2 language model to generate novel claims with the original label, including at least one of the found triggers.}
\label{fig:puc}
\end{figure}

In this paper, we explore the vulnerabilities of FC models trained on the FEVER dataset~(\cite{thorne-etal-2018-fever}), where the inference between a claim and evidence text is predicted. We particularly construct \textit{universal adversarial triggers}~(\cite{wallace2019universal}) -- single n-grams appended to the input text that can shift the prediction of a model from a source class to a target one. Such adversarial examples are of particular concern, as they can apply to a large number of input instances. 

However, we find that the triggers also change the meaning of the claim such that the true label is in fact the target class. For example, when attacking a claim-evidence pair with a `SUPPORTS' label, a common unigram found to be a universal trigger when switching the label to `REFUTES' is `none'. Prepending this token to the claim drastically changes the meaning of the claim such that the new claim is in fact a valid 
`REFUTES' claim as opposed to an adversarial `SUPPORTS' claim. 
Furthermore, we find adversarial examples constructed in this way to be nonsensical, as a new token is simply being attached to an existing claim. 
%

Our \textbf{contributions} are as follows. We \textit{preserve the meaning} of the source text and \textit{improve the semantic validity} of universal adversarial triggers to automatically construct more potent adversarial examples. This is accomplished via: 1) a \textit{novel extension to the HotFlip attack}~(\cite{ebrahimi2018hotflip}), where we jointly minimize the target class loss of a FC model and the attacked class loss of a natural language inference model; 2) a \textit{conditional language model} trained using GPT-2~(\cite{radford2019language}), which takes 
trigger tokens and a piece of evidence, and generates a semantically coherent new claim containing at least one trigger. 
The resulting triggers maintain potency against a FC model while preserving the original claim label. Moreover, the conditional language model produces semantically coherent adversarial examples containing triggers, on which a FC model performs 23.8\% worse than with the original FEVER claims. The code for the paper is publicly available.\footnote{https://github.com/copenlu/fever-adversarial-attacks}


\section{Related Work}
\subsection{Adversarial Examples}
Adversarial examples for 
NLP systems can be constructed as automatically generated text~(\cite{ren2019generating}) or perturbations of existing input instances~(\cite{jintextfool,ebrahimi2018hotflip}). For a 
detailed literature overview, see~\cite{zhang2019adversarial}.

One potent type of adversarial techniques are universal adversarial attacks~(\cite{gao2019universal, wallace2019universal}) -- single perturbation changes that can be applied to a large number of input instances and that cause significant performance decreases of the model under attack. 
~(\cite{wallace2019universal}) find universal adversarial triggers that can change the prediction of the model using the HotFlip algorithm~(\cite{ebrahimi2018hotflip}). 

However, for NLI tasks, they also change the meaning of the instance they are appended to, and the prediction of the model remains correct. ~\cite{michel2019evaluation} 
address this by exploring only perturbed instances in the neighborhood of the original one.
Their approach is for instance-dependent attacks, whereas we suggest finding \textit{universal} adversarial triggers that also preserve the original meaning of input instances. 
Another approach to this 
are rule-based perturbations of the input~(\cite{ribeiro2018semantically}) or imposing adversarial constraints on the produced perturbations~(\cite{dia2019semantics}). 
By contrast, we extend the HotFlip method by including an auxiliary Semantic Textual Similarity (STS) objective. We additionally use the extracted universal adversarial triggers to generate adversarial examples with low perplexity.

\subsection{Fact Checking}

Fact checking systems consist of components to identify check-worthy claims (\cite{atanasova2018overview,hansen2019neural,wright-augenstein-2020-claim}), retrieve and rank evidence documents (\cite{yin-roth-2018-twowingos,journals/corr/abs-2009-06402}), determine the relationship between claims and evidence documents (\cite{bowman-etal-2015-large,augenstein-etal-2016-stance,baly-etal-2018-integrating}), and finally predict the claims' veracity (\cite{thorne-etal-2018-fever,augenstein-etal-2019-multifc}).
As this is a relatively involved task, models easily overfit to shallow textual patterns, necessitating the need for adversarial examples to evaluate the limits of their performance.

\cite{thorne2019evaluating} are the first to propose hand-crafted adversarial attacks. 
They follow up on this with the FEVER 2.0 
task~(\cite{thorne-etal-2019-fever2}), where participants design adversarial attacks for existing FC systems. The first two winning systems~(\cite{niewinski-etal-2019-gem, hidey-etal-2020-deseption}) produce claims requiring multi-hop reasoning, which has been shown to be challenging for fact checking models (\cite{ostrowski2020multihop}). The other remaining system~(\cite{kim-allan-2019-fever}) generates adversarial attacks manually. We instead find universal adversarial attacks that can be applied to most existing inputs while markedly decreasing fact checking performance.
\cite{niewinski-etal-2019-gem} additionally feed a pre-trained GPT-2 model with the target label of the instance along with the text for conditional adversarial claim generation. Conditional language generation has also been employed by \cite{keskar2019ctrl} to control the style, content, and the task-specific behavior of a Transformer.

\section{Methods}

\subsection{Models}
We take a RoBERTa~(\cite{liu2019roberta}) model pretrained with a LM objective and fine-tune it to classify claim-evidence pairs from the FEVER dataset as SUPPORTS, REFUTES, and NOT ENOUGH INFO (NEI). The evidence used is the gold evidence, available for the SUPPORTS and REFUTES classes. For NEI claims, we use the system of \cite{malon-2018-team} to retrieve evidence sentences. 
To measure the semantic similarity between the claim before and after prepending a trigger, we use a large RoBERTa model fine-tuned on the Semantic Textual Similarity Task.\footnote{https://huggingface.co/SparkBeyond/roberta-large-sts-b} For further details, we refer the reader to \S\ref{sec:appendixA}.

\subsection{Universal Adversarial Triggers Method}
The Universal Adversarial Triggers method is developed to find n-gram trigger tokens $\mathbf{t_{\alpha}}$, which, appended to the original input $x$,  $f(x) = y$, cause the model to predict a target class $\widetilde{y}$ : $f(t_{\alpha}, x) = \widetilde{y}$. In our work, we generate unigram triggers, as generating longer triggers would require additional objectives to later produce well-formed adversarial claims. We start by initializing the triggers with the token `a'. Then, we update the embeddings of the initial trigger tokens $\mathbf{e}_{\alpha}$ with embeddings $\mathbf{e}_{w_i}$ of candidate adversarial trigger tokens $w_i$ that minimize the loss $\mathcal{L}$ for the target class $\widetilde{y}$. Following the HotFlip algorithm, we reduce the brute-force optimization problem using a first-order Taylor approximation around the initial trigger embeddings:
\begin{equation}
\underset{\mathbf{w}_{i} \in \mathcal{V}}{\arg \min }\left[\mathbf{e}_{w_i}-\mathbf{e}_{\alpha}\right]^{\top} \nabla_{\mathbf{e}_{\alpha}} \mathcal{L}
\end{equation}
where $\mathcal{V}$ is the vocabulary of the RoBERTa model and $\nabla_{\mathbf{e}_{\alpha}} \mathcal{L}$ is the average gradient of the task loss accumulated for all batches. This approximation allows for a $\mathcal{O}(|\mathcal{V}|)$ space complexity of the brute-force candidate trigger search.

While HotFlip 
finds universal adversarial triggers that successfully fool the model for many instances, we find that the most potent triggers are often negation words, e.g., `not', `neither', `nowhere'. Such triggers change the meaning of the text, making the prediction of the target class correct. Ideally, adversarial triggers would preserve the original label of the claim. To this end, we propose to include an auxiliary STS model objective when searching for candidate triggers. The additional objective is used to minimize the loss $\mathcal{L'}$ for the maximum similarity score (5 out of 0) between the original claim and the claim with the prepended trigger. Thus, we arrive at the combined optimization problem:
\begin{equation}
\small
\underset{\mathbf{w}_{i} \in \mathcal{V}}{\arg \min }([\mathbf{e}_{w_i}-\mathbf{e}_{\alpha}]^{\top} \nabla_{\mathbf{e}_{\alpha}} \mathcal{L} + [\mathbf{o}_{w_i}-\mathbf{o}_{\alpha}]^{\top} \nabla_{\mathbf{o}_{\alpha}} \mathcal{L'})
\end{equation}
where $\mathbf{o}_w$ is the STS model embedding of word $w$. For the initial trigger token, we use ``[MASK]'' as STS selects candidates from the neighborhood of the initial token.

\subsection{Claim Generation}
\label{sec:claim_generation}
In addition to finding highly potent adversarial triggers, it is also of interest to generate coherent statements containing the triggers. To accomplish this, we use the HuggingFace implementation of the GPT-2 language model~(\cite{radford2019language,Wolf2019HuggingFacesTS}), a large transformer-based language model trained on 40GB of text. 
The objective is to generate a coherent claim, which either entails, refutes, or is unrelated a given piece of evidence, while also including trigger words.

The language model is first fine tuned on the FEVER FC corpus with a specific input format. FEVER consists of claims and evidence with the labels \texttt{SUPPORTS}, \texttt{REFUTES}, or \texttt{NOT ENOUGH INFO} (NEI). We first concatenate evidence and claims with a special token. 
Next, to encourage generation of claims with certain tokens, a sequence of tokens separated by commas is prepended to the input. For training, the sequence consists of a single token randomly selected from the original claim, and four random tokens from the vocabulary. 
This encourages the model to only select the one token most likely to form a coherent and correct claim. The final input format is \texttt{[trigger tokens]}\textbar\textbar\texttt{[evidence]}\textbar\textbar\texttt{[claim]}.
Adversarial claims are then generated by providing an initial input of a series of five comma-separated trigger tokens plus evidence, and progressively generating the rest of the sequence. Subsequently, the set of generated claims is pruned to include only those which contain a trigger token, 
and constitute the desired label. The latter is ensured by passing both evidence and claim through an external NLI model trained on SNLI (\cite{bowman-etal-2015-large}). 

\section{Results}
We present results for universal adversarial trigger generation and coherent claim generation. 
Results are measured using the original FC model on claims with added triggers and generated claims (macro F1). We also measure how well the added triggers maintain the claim's original label (semantic similarity score), the perplexity (PPL) of the claims with prepended triggers, and the semantic quality of generated claims (manual annotation). PPL is measured with a pretrained RoBERTa LM.

\subsection{Adversarial Triggers}
Table~\ref{tab:eval} presents the results of applying universal adversarial triggers to claims from the source class.
The top-performing triggers for each direction are found in \S\ref{sec:appendixC}. 
The adversarial method with a single FC objective successfully deteriorates model performance by a margin of 0.264 F1 score overall. The biggest performance decrease is when the adversarial triggers are constructed to flip the predicted class from SUPPORTS to REFUTES. We also find that 8 out of 18 triggers from the top-3 triggers for each direction, are negation words such as  `nothing', `nobody', `neither', `nowhere' (see Table~\ref{tab:evalonetrig} in the appendix). The first of these triggers decreases the performance of the model to 0.014 in F1. While this is a significant performance drop, these triggers also flip the meaning of the text. The latter is again indicated by the decrease of the semantic similarity between the claim before and after prepending a trigger token, which is the largest for the SUPPORTS to REFUTES direction. We hypothesise that the success of the best performing triggers is partly due to the meaning of the text being flipped.

Including the auxiliary STS objective increases the similarity between the claim before and after prepending the trigger for five out of six directions. Moreover, we find that now only one out of the 18 top-3 triggers for each direction are negation words. Intuitively, these adversarial triggers are worse at fooling the FC model as they also have to preserve the label of the original claim. Notably, for the SUPPORTS to REFUTES direction the trigger performance is decreased with a margin of 0.642 compared to the single FC objective.
We conclude that including the STS objective for generating Universal Adversarial triggers helps to preserve semantic similarity with the original claim, but also makes it harder to both find triggers preserving the label of the claim while substantially decreasing the performance of the model.

\begin{table}
\small
\centering
\begin{tabular}{l@{\hspace{1.2\tabcolsep}}l@{\hspace{1.2\tabcolsep}}l@{\hspace{1.2\tabcolsep}}l}
\toprule
\textbf{Class} & \textbf{F1} & \textbf{STS} & \textbf{PPL}\\ \midrule
\multicolumn{4}{c}{\bf No Triggers} \\
All & .866 & 5.139 & 11.92 ($\pm$45.92) \\
S & .938 & 5.130 & 12.22 ($\pm$40.34) \\
R & .846 & 5.139 &  12.14 ($\pm$37.70) \\
NEI & .817 & 5.147 & 14.29 ($\pm$84.45) \\
\midrule
\multicolumn{4}{c}{\bf FC Objective} \\
All & .602 ($\pm$.289) & 4.586 ($\pm$.328) & 12.96 ($\pm$55.37) \\
S$\rightarrow$R & .060 ($\pm$.034) & 4.270 ($\pm$.295) & 12.44 ($\pm$41.74) \\
S$\rightarrow$NEI & .611 ($\pm$.360) & 4.502 ($\pm$.473) & 12.75 ($\pm$40.50) \\
R$\rightarrow$S & .749 ($\pm$.027) & 4.738 ($\pm$.052) & 11.91 ($\pm$36.53) \\
R$\rightarrow$NEI & .715 ($\pm$.026) & 4.795 ($\pm$.094) & 11.77 ($\pm$36.98) \\
NEI$\rightarrow$R & .685 ($\pm$.030) & 4.378 ($\pm$.232) & 14.20 ($\pm$83.32) \\
NEI$\rightarrow$S & .793 ($\pm$.054) & 4.832 ($\pm$.146) & 14.72 ($\pm$93.15) \\
\midrule
\multicolumn{4}{c}{\bf FC+STS Objectives} \\
All & .763 ($\pm$.123) & 4.786 ($\pm$.156) & 12.97 ($\pm$58.30) \\
S$\rightarrow$R & .702 ($\pm$.237) & 4.629 ($\pm$.186) & 12.62 ($\pm$41.91) \\
S$\rightarrow$NEI & .717 ($\pm$.161) & 4.722 ($\pm$.152) & 12.41 ($\pm$39.66) \\
R$\rightarrow$S & .778 ($\pm$.010) & 4.814 ($\pm$.141) & 11.93 ($\pm$37.04) \\
R$\rightarrow$NEI & .779 ($\pm$.009) & 4.855 ($\pm$.098) & 12.20 ($\pm$37.67) \\
NEI$\rightarrow$R & .780 ($\pm$.078) & 4.894 ($\pm$.115) & 15.27 ($\pm$111.2) \\
NEI$\rightarrow$S & .821 ($\pm$.008) & 4.800 ($\pm$.085) & 13.42 ($\pm$82.30) \\
\bottomrule
\end{tabular}
\caption{Universal Adversarial Trigger method performance. Triggers are generated given claims from a source class to fool the classifier to predict a target class (column \textit{Class}, with SUPPORTS (S), REFUTES (R), NEI). 
The results are averaged over the top 10 triggers.}
\label{tab:eval}
\end{table}

\begin{table}
\fontsize{10}{10}\selectfont
\renewcommand{\arraystretch}{1.4}
\centering
\begin{tabular}{p{6cm} P{4.2cm} p{4.6cm}}
\toprule
\textbf{Evidence} & \textbf{Triggers} & \textbf{Generated Claim}\\ 
\midrule

\multicolumn{3}{c}{\bf SUPPORTS Claims} \\
Since the 19th century, some Romani have also migrated to the Americas. & don,already,more,\textcolor{red}{during},home & Romani have moved to the Americas \textcolor{red}{during} the 19th century. \\

Cyprus is a major tourist destination in the Mediterranean. & foreign,biggest,\textcolor{red}{major},every, friends & Cyprus is a \textcolor{red}{major} tourist destination. \\

The first Nobel Prize in Chemistry was awarded in 1901 to Jacobus Henricus van't Hoff, of the Netherlands, ``for his discovery of the laws of chemical dynamics and osmotic pressure in solutions.'' & later,\textcolor{red}{already},quite,altern, whereas & Henricus Van't Hoff was \textcolor{red}{already} awarded the Nobel Prize. \\
\midrule
\multicolumn{3}{c}{\bf REFUTES Claims} \\

California Attorney General Kamala Harris defeated Sanchez , 61.6\% to 38.4\%. & phys,incarn,\textcolor{red}{not},occasionally,
something & Kamala Harris did \textcolor{red}{not} defeat Sanchez, 61.6\% to 38.4\%. \\

Uganda is in the African Great Lakes region. & unless,endorsed,picks,pref, \textcolor{red}{against} & Uganda is \textcolor{red}{against} the African Great Lakes region. \\

Times Higher Education World University Rankings is an annual publication of university rankings by Times Higher Education (THE) magazine. & interested,reward,visit, \textcolor{red}{consumer},conclusion & Times Higher Education World University Rankings is a \textcolor{red}{consumer} magazine. \\

\midrule
\multicolumn{3}{c}{\bf NOT ENOUGH INFO Claims} \\
The KGB was a military service and was governed by army laws and regulations, similar to the Soviet Army or MVD Internal Troops. & nowhere,\textcolor{red}{only},none,no,nothing & The KGB was \textcolor{red}{only} controlled by a military service. \\

The series revolves around Frank Castle, who uses lethal methods to fight crime as the vigilante ``the Punisher'', with Jon Bernthal reprising the role from Daredevil. & says,said,\textcolor{red}{take},say,is & \textcolor{red}{Take} Me High is about Frank Castle's use of lethal techniques to fight crime. \\

The Suite Life of Zack \& Cody is an American sitcom created by Danny Kallis and Jim Geoghan. & whilst,interest,applic,\textcolor{red}{someone}, nevertheless & The Suite Life of Zack \& Cody was created by \textcolor{red}{someone} who never had the chance to work in television. \\
\bottomrule
\end{tabular}
\caption{Examples of generated adversarial claims. These are all claims which the FC model incorrectly classified.}
\label{tab:generation_examples}
\end{table}

\subsection{Generation}
We use the method described in \S\ref{sec:claim_generation} to generate 156 claims using triggers found with the additional STS objective, and 156 claims without. 52 claims are generated for each class (26 flipping to one class, 26 flipping to the other). A different GPT-2 model is trained to generate claims for each specific class, with triggers specific to attacking that class used as input. The generated claims are annotated manually (see \S\ref{app:B3} for the procedure). The overall average claim quality is 4.48, indicating that most generated statements are highly semantically coherent. The macro F1 of the generative model w.r.t. the intended label is 58.9 overall. For the model without the STS objective, the macro F1 is 56.6, and for the model with the STS objective, it is 60.7, meaning that using triggers found with the STS objective helps the generated claims to retain their intended label.

We measure the performance of the original FC model on generated claims (\autoref{tab:generation_eval}). We compare between using triggers that are generated with the STS objective (Ex2) and without (Ex1). In both cases, the adversarial claims effectively fool the FC model, which performs 38.4\% worse and 23.8\% worse on Ex1 and Ex2, respectively.  Additionally, the overall sentence quality increases when the triggers are found with the STS objective (Ex2). The FC model's performance is higher on claims using triggers generated with the STS objective but still significantly worse than on the original claims. We provide examples of generated claims with their evidence in \autoref{tab:generation_examples}.
\begin{table}
\fontsize{10}{10}\selectfont
\centering
\begin{tabular}{lccc}
\toprule
\textbf{Target} & \textbf{F1} & \textbf{Avg Quality} & \textbf{\# Examples}\\ \midrule
\multicolumn{4}{c}{\bf FC Objective} \\
Overall& 0.534& 4.33&156\\
SUPPORTS& 0.486& 4.79& 39\\
REFUTES& 0.494& 4.70&32\\
NEI& 0.621& 3.98 &85\\
\midrule
\multicolumn{4}{c}{\bf FC+STS Objectives} \\
Overall& 0.635& 4.63&156\\
SUPPORTS& 0.617& 4.77&67\\
REFUTES& 0.642& 4.68&28\\
NEI& 0.647& 4.44&61\\
\bottomrule
\end{tabular}
\caption{FC performance for generated claims.}
\label{tab:generation_eval}
\end{table}

Comparing FC performance with our generated claims vs. those from the development set of adversarial claims from the FEVER shared task 
, we see similar drops in performance (0.600 and 0.644 macro F1, respectively). While the adversarial triggers from FEVER cause a larger performance drop, they were manually selected to meet the label coherence and grammatical correctness requirements. Conversely, we automatically generate claims that meet these requirements.

\section{Conclusion}
We present a method for automatically generating highly potent, well-formed, label cohesive claims for FC. 
We improve upon previous work on universal adversarial triggers by determining how to construct valid claims containing a trigger word. 
Our method is fully automatic, whereas previous work on generating claims for fact checking is generally rule-based or requires manual intervention. As FC is only one test bed for adversarial attacks, it would be interesting to test this method on other NLP tasks requiring semantic understanding such as question answering 
to better understand shortcomings of models. 

\section*{Acknowledgements}
$\begin{array}{l}\includegraphics[width=1cm]{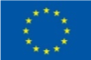} \end{array}$ This project has received funding from the European Union's Horizon 2020 research and innovation programme under the Marie Sk\l{}odowska-Curie grant agreement No 801199.

\clearpage

\clearpage
\section{Appendix}
\subsection{Implementation Details}
\label{sec:appendixA}
\textbf{Models}. The RoBERTa FC model (125M parameters) is fine-tuned with a batch size of 8, learning rate of 2e-5 and for a total of 4 epochs, where the epoch with the best performance is saved. We used the implementation provided by HuggingFace library. We performed a grid hyper-parameter search for the learning rate between the values 1e-5, 2e-5, and 3e-5. The average time for training a model with one set of hyperparameters is 155 minutes ($\pm3$). The average accuracy over the different hyperparameter runs is 0.862($\pm$ 0.005) F1 score on the validation set.

For the models that measure the perplexity and the semantical similarity we use the pretrained models provided by HuggingFace-- RoBERTa large model (125M parameters) fine tuned on the STS-b task and RoBERTa base model (355M parameters) pretrained on a LM objective.

We used the HuggingFace implementation of the small GPT-2 model, which consists of 124,439,808 parameters and is fine-tuned with a batch size of 4, learning rate of 3e-5, and for a total of 20 epochs. We perform early stopping on the loss of the model on a set of validation data. The average validation loss is 0.910. The average runtime for training one of the models is 31 hours and 28 minutes.

We note that, the intermediate models used in this work and described in this section, are trained on large relatively general-purpose datasets. While, they can make some mistakes, they work well enough and using them, we don't have to rely on additional human annotations for the intermediate task.

\textbf{Adversarial Triggers.} The adversarial triggers are generated based on instances from the validation set. We run the algorithm for three epochs to allow for the adversarial triggers to converge. At each epoch the initial trigger is updated with the best performing trigger for the epoch (according to the loss of the FC or FC+STS objective). At the last step, we select only the top 10 triggers and remove any that have a negative loss. We choose the top 10 triggers as those are the most potent ones, adding more than top ten of the triggers preserves the same tendencies in the results, but smooths them as further down the list of adversarial attacks, the triggers do not decrease the performance of the model substantially. This is also supported by related literature~(\cite{wallace2019universal}), where only the top few triggers are selected.

The adversarial triggers method is run for 28.75 ($\pm$ 1.47) minutes for with the FC objective and 168.6($\pm$ 28.44) minutes for the FC+STS objective. We perform the trigger generation with a batch size of four. We additionally normalize the loss for each objective to be in the range [0,1] and also re-weight the losses with a wieht of 0.6 for the FC loss and a weight of 0.4 for the SST loss as when generated with an equal weight, the SST loss tends to preserve the same initial token in all epochs.

\textbf{Datasets.} 
The datasets used for training the FC model consist of 161,249 SUPPORTS, 60,227 REFUTES, and 69,885 NEI claims for the training split; 6,207 SUPPORTS, 6,235 REFUTES, and 6,554 NEI claims for the dev set; 6,291 SUPPORTS, 5,992 REFUTES, and 6522 NEI claims. The evidence for each claim is the gold evidence provided from the FEVER dataset, which is available for REFUTES and SUPPORTS claims. When there is more than one annotation of different evidence sentences for an instance, we include them as separate instances in the datasets. For NEI claims, we use the system of \cite{malon-2018-team} to retrieve evidence sentences. 

\subsection{Top Adversarial Triggers}
Table~\ref{tab:evalonetrig} presents the top adversarial triggers for each direction found with the Universal Adversarial Triggers method. It offers an additional way of estimating the effectiveness of the STS objective by comparing the number of negation words generated by the basic model (8) and the STS objective (2) in the top-3 triggers for each direction.
\label{sec:appendixC}
\begin{table}
\centering
\begin{tabular}{l@{\hspace{1.2\tabcolsep}}l@{\hspace{1.2\tabcolsep}}l@{\hspace{1.2\tabcolsep}}l@{\hspace{1.2\tabcolsep}}l}
\toprule
\textbf{Class} & \textbf{Trigger} & \textbf{F1} & \textbf{STS} & \textbf{PPL}\\ \midrule
\multicolumn{5}{c}{\bf FC Objective} \\
S$\rightarrow$R & only &  0.014 &  4.628 &  11.660 (36.191) \\
S$\rightarrow$R & nothing &  0.017 &  4.286 &  13.109 (56.882) \\
S$\rightarrow$R & nobody &  0.036 &  4.167 &  12.784 (37.390) \\
S$\rightarrow$NEI & neither &   0.047 &  3.901 &  11.509 (31.413) \\
S$\rightarrow$NEI & none &  0.071 &  4.016 &  13.136 (39.894) \\
S$\rightarrow$NEI & Neither &  0.155 &  3.641 &  11.957 (44.274) \\
R$\rightarrow$S & some &  0.687 &  4.694 &  11.902 (33.348) \\
R$\rightarrow$S & Sometimes &  0.724 &  4.785 &  10.813 (32.058) \\
R$\rightarrow$S & Some &  0.743 &  4.713 &  11.477 (37.243) \\
R$\rightarrow$NEI & recommended &  0.658 &  4.944 &  12.658 (36.658) \\
R$\rightarrow$NEI & Recommend &  0.686 &  4.789 &  10.854 (32.432) \\
R$\rightarrow$NEI & Supported &  0.710 &  4.739 &  11.972 (40.267) \\
NEI$\rightarrow$R & Only &  0.624 &   4.668 &  12.939 (57.666) \\
NEI$\rightarrow$R & nothing &  0.638 &  4.476 &   11.481 (48.781) \\
NEI$\rightarrow$R & nobody & 0.678 &  4.361 &  16.345 (111.60) \\
NEI$\rightarrow$S & nothing &  0.638 &  4.476 &  18.070 (181.85) \\
NEI$\rightarrow$S & existed &  0.800 &  4.950  &  15.552 (79.823) \\
NEI$\rightarrow$S & area &  0.808 &  4.834  &  13.857 (93.295) \\

\midrule
\multicolumn{5}{c}{\bf FC+STS Objectives} \\
S$\rightarrow$R & never & 0.048 & 4.267 & 12.745 (50.272) \\
S$\rightarrow$R & every & 0.637 & 4.612 & 13.714 (51.244) \\
S$\rightarrow$R & didn & 0.719 & 4.986 & 12.416 (41.080) \\
S$\rightarrow$NEI & always  & 0.299 &  4.774 &  11.906 (35.686) \\
S$\rightarrow$NEI & every & 0.637 & 4.612 & 12.222 (38.440) \\
S$\rightarrow$NEI & investors & 0.696 & 4.920 & 12.920 (42.567) \\
R$\rightarrow$S & over &  0.761 &  4.741 &  12.139 (33.611) \\
R$\rightarrow$S & about &  0.765 &   4.826 &  12.052 (37.677) \\
R$\rightarrow$S & her &   0.774 &   4.513 &   12.624 (41.350) \\
R$\rightarrow$NEI & top &  0.757 &  4.762 &  12.787 (39.418) \\
R$\rightarrow$NEI & also &   0.770 &   5.034 &   11.751 (35.670) \\
R$\rightarrow$NEI & when &   0.776 &   4.843 &   12.444 (37.658) \\
NEI$\rightarrow$R & only &  0.562 &  4.677 &  14.372 (83.059) \\
NEI$\rightarrow$R & there &   0.764 & 4.846 &    11.574 (42.949) \\
NEI$\rightarrow$R & just &   0.786 & 4.916 &   16.879 (135.73) \\
NEI$\rightarrow$S & of&   0.802 & 4.917 &  11.844 (55.871) \\
NEI$\rightarrow$S & is &   0.815 & 4.931 & 17.507 (178.55) \\
NEI$\rightarrow$S & A &   0.818 & 4.897 & 12.526 (67.880) \\

\bottomrule
\end{tabular}
\caption{Top-3 triggers found with the Universal Adversarial Triggers methods. The triggers are generated given claims from a source class (column \textit{Class}), so that the classifier is fooled to predict a different target class. The classes are SUPPORTS (S), REFUTES (R), NOT ENOUGH INFO (NEI).}
\label{tab:evalonetrig}
\end{table}

\subsection{Computing Infrastructure}
All experiments were run on a shared cluster. Requested jobs consisted of 16GB of RAM and 4 Intel Xeon Silver 4110 CPUs. We used two NVIDIA Titan RTX GPUs with 12GB of RAM for training GPT-2 and one NVIDIA Titan X GPU with 8GB of RAM for training the FC models and finding the universal adversarial triggers.

\subsection{Evaluation Metrics}
The primary evaluation metric used was macro-F1 score. We used the sklearn implementation of \texttt{precision\_recall\_fscore\_support}, which can be found here: \url{https://scikit-learn.org/stable/modules/generated/sklearn.metrics.precision_recall_fscore_support.html}. Briefly:
\begin{equation*}
   p = \frac{tp}{tp + fp} 
\end{equation*}
\begin{equation*}
   r = \frac{tp}{tp + fn} 
\end{equation*}
\begin{equation*}
   F1 = \frac{2*p*r}{p+r} 
\end{equation*}
where $tp$ are true positives, $fp$ are false positives, and $fn$ are false negatives.

\subsection{Manual Evaluation}
\label{app:B3}
After generating the claims, two independent annotators label the overall claim quality (score of 1-5) and the true label for the claim. The inter-annotator agreement for the quality label using Krippendorff's alpha is 0.54 for the quality score and 0.38 for the claim label. Given this, we take the average of the two annotator's scores for the final quality score and have a third expert annotator examine and select the best label for each contested claim label.

\chapter{Generating Fact Checking Explanations}\label{ch:fc_explanations}

\boxabstract{
Most existing work on automated fact checking is concerned with predicting the veracity of claims based on metadata, social network spread, language used in claims, and, more recently, evidence supporting or denying claims. A crucial piece of the puzzle that is still missing is to understand how to automate the most elaborate part of the process -- generating justifications for verdicts on claims.
This paper provides the first study of how these explanations can be generated automatically based on available claim context, and how this task can be modelled jointly with veracity prediction. Our results indicate that optimising both objectives at the same time, rather than training them separately, improves the performance of a fact checking system. The results of a manual evaluation further suggest that the informativeness, coverage and overall quality of the generated explanations are also improved in the multi-task model.

}\blfootnote{\fullcite{atanasova-etal-2020-generating-fact}}

\section{Introduction}\label{sec:intro}
\noindent When a potentially viral news item is rapidly or indiscriminately published by a news outlet, the responsibility of verifying the truthfulness of the item is often passed on to the audience. To alleviate this problem, independent teams of professional fact checkers manually verify the veracity and credibility of common or particularly check-worthy statements circulating the web. However, these teams have limited resources to perform manual fact checks, thus creating a need for automating the fact checking process.
\setlength{\tabcolsep}{0.2em}

\begin{table}
\fontsize{10}{10}\selectfont
\begin{center}
\begin{tabular}{|L|}
\toprule
\textbf{Claim}: The last major oil spill from a drilling accident in America happened over 40 years ago in 1969.\\ \midrule
\textbf{Ruling Comments}: 
(...) \hlyellow{The last major oil spill from a drilling accident in America happened over 40 years ago in 1969.} \\
\hspace*{3mm}(...) The largest in volume was the Santa Barbara spill of 1969 referenced by Murdock and Johnson, in which an estimated 100,000 barrels of oil spilled into the Pacific Ocean, according to the API. \hlblue{The Santa Barbara spill was so big it ranked seventh among the 10 largest oil spills caused by marine well blowouts in the world, the report states.} Two other U.S. spills, both in 1970, rank eighth and 10th. \hlred{Fourteen marine blowouts have taken place in the U.S. between 1969 and 2007.} Six of them took place after 1990 and spilled a total of nearly 13,700 barrels. \\
\hspace*{3mm}(...) We interviewed three scientists who said that the impact of a spill has little to do with its volume. \hlviolet{Scientists have proven that spills far smaller than Santa Barbara's have been devastating.} \\  \midrule
\textbf{Justification}: While the nation's largest oil well blowout did take place in 1969, it's not factually correct to call it the ``last major oil spill". First of all, two of the largest blowouts in the world took place in the U. S.  the following year. More importantly, experts agree that spills far smaller in volume to the 1969 disaster have been devastating. From a scientific perspective, Johnson's decision to single out the 1969 blowout as the last ``major" one makes no sense. \\ \midrule

\textbf{Ruling}: Half-True \\ \bottomrule
\end{tabular}
\end{center}
\caption{\label{tab:Example} Example instance from the LIAR-PLUS dataset, with oracle sentences for generating the justification highlighted.}
\end{table}

The current research landscape in automated fact checking is comprised of systems that estimate the veracity of claims based on available metadata and evidence pages. Datasets like LIAR (\cite{P17-2067}) and the multi-domain dataset MultiFC (\cite{augenstein-etal-2019-multifc}) provide real-world benchmarks for evaluation. There are also artificial datasets of a larger scale, e.g., the FEVER (\cite{thorne-etal-2018-fever}) dataset based on Wikipedia articles. As evident from the effectiveness of state-of-the-art methods for both real-world -- 0.492 macro F1 score (\cite{augenstein-etal-2019-multifc}), and artificial data -- 68.46 FEVER score (label accuracy conditioned on evidence provided for `supported' and `refuted' claims) (\cite{stammbach-neumann-2019-team}),  the task of automating fact checking remains a significant and poignant research challenge.

A prevalent component of existing fact checking systems is a stance detection or textual entailment model that predicts whether a piece of evidence contradicts or supports a claim (\cite{Ma:2018:DRS:3184558.3188729,mohtarami-etal-2018-automatic,Xu2019AdversarialDA}). Existing research, however, rarely attempts to directly optimise the selection of relevant evidence, i.e., the self-sufficient explanation for predicting the veracity label (\cite{thorne-etal-2018-fever, stammbach-neumann-2019-team}).
On the other hand, \cite{alhindi-etal-2018-evidence} have reported a significant performance improvement of over 10\% macro F1 score when the system is provided with a short human explanation of the veracity label. Still, there are no attempts at automatically producing explanations, and automating the most elaborate part of the process - producing the \emph{justification} for the veracity prediction - is an understudied problem.

In the field of NLP as a whole, both explainability and interpretability methods have gained importance recently, because most state-of-the-art models are large, neural black-box models. Interpretability, on one hand, provides an overview of the inner workings of a trained model such that a user could, in principle, follow the same reasoning to come up with predictions for new instances. However, with the increasing number of neural units in published state-of-the-art models, it becomes infeasible for users to track all decisions being made by the models.
Explainability, on the other hand, deals with providing local explanations about single data points that suggest the most salient areas from the input or are generated textual explanations for a particular prediction.

Saliency explanations have been studied extensively (\cite{Adebayo:2018:SCS:3327546.3327621, arras-etal-2019-evaluating, poerner-etal-2018-evaluating}), however, they only uncover regions with high contributions for the final prediction, while the reasoning process still remains behind the scenes. An alternative method explored in this paper is to generate textual explanations. In one of the few prior studies on this, the authors find that feeding generated explanations about multiple choice question answers to the answer predicting system improved QA performance (\cite{rajani-etal-2019-explain}).

Inspired by this, we research how to generate explanations for veracity prediction. We frame this as a summarisation task, where, provided with elaborate fact checking reports, later referred to as \textit{ruling comments}, the model has to generate \textit{veracity explanations} close to the human justifications as in the example in Table~\ref{tab:Example}. We then explore the benefits of training a joint model that learns to generate veracity explanations while also predicting the veracity of a claim.\\
In summary, our \textbf{contributions} are as follows:
\begin{enumerate}[noitemsep]
\item{We present the first study on generating veracity explanations, showing that they can successfully describe the reasons behind a veracity prediction.}
\item{We find that the performance of a veracity classification system can leverage information from the elaborate ruling comments, and can be further improved by training veracity prediction and veracity explanation jointly.}
\item{We show that optimising the joint objective of veracity prediction and veracity explanation produces explanations that achieve better coverage and overall quality and serve better at explaining the correct veracity label than explanations learned solely to mimic human justifications.}
\end{enumerate}

\section{Dataset}\label{sec:dataset}

Existing fact checking websites publish claim veracity verdicts along with ruling comments to support the verdicts. Most ruling comments span over long pages and contain redundancies, making them hard to follow. Textual explanations, by contrast, are succinct and provide the main arguments behind the decision. PolitiFact\footnote{https://www.politifact.com/} provides a summary of a claim's ruling comments that summarises the whole explanation in just a few sentences. 

We use the PolitiFact-based dataset LIAR-PLUS (\cite{alhindi-etal-2018-evidence}), which contains 12,836 statements with their veracity justifications. The justifications are automatically extracted from the long ruling comments, as their location is clearly indicated at the end of the ruling comments. Any sentences with words indicating the label, which \cite{alhindi-etal-2018-evidence} select to be identical or similar to the label, are removed. We follow the same procedure to also extract the ruling comments without the summary at hand.

We remove instances that contain fewer than three sentences in the ruling comments as they indicate short veracity reports, where no summary is present. The final dataset consists of 10,146 training, 1,278 validation, and 1,255 test data points. A claim's ruling comments in the dataset span over 39 sentences or 904 words on average, while the justification fits in four sentences or 89 words on average. 

\section{Method}\label{sec:method}
We now describe the models we employ for training separately (1) an explanation extraction and (2) veracity prediction, as well as (3) the joint model trained to optimise both.

The models are based on DistilBERT (\cite{sanh2019distilbert}), which is a reduced version of BERT (\cite{devlin-etal-2019-bert}) performing on par with it as reported by the authors. For each of the models described below, we take the version of DistilBERT that is pre-trained with a language-modelling objective and further fine-tune its embeddings for the specific task at hand. 

\subsection{Generating Explanations}\label{sec:explanationGen}

\begin{figure*}
\centering
\includegraphics[width=\linewidth]{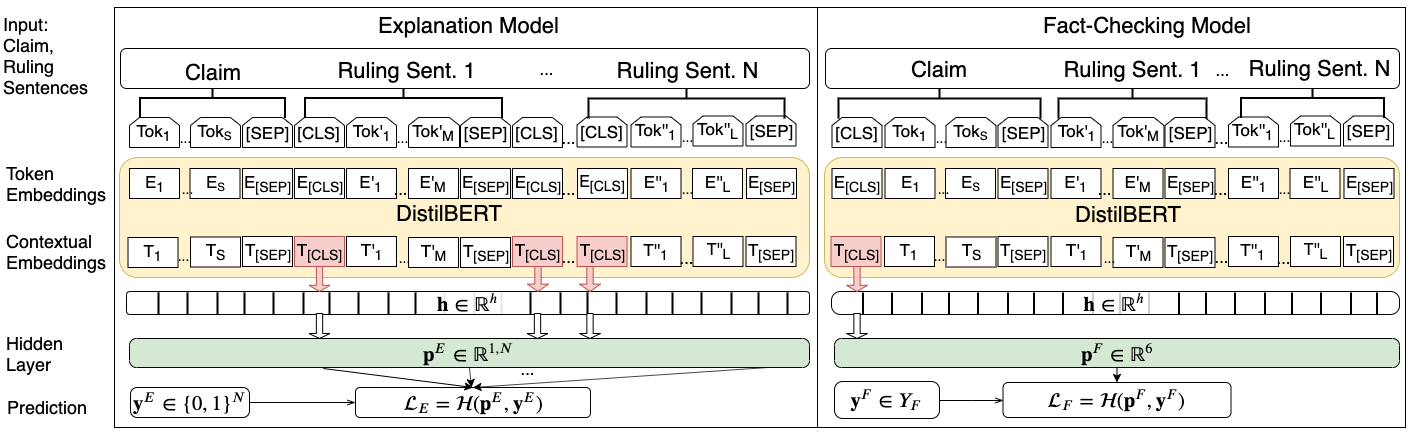}
\caption{Architecture of the \textit{Explanation} (left) and \textit{Fact-Checking} (right) models that optimise separate objectives.} 
\label{figure:separateModels}
\end{figure*}

Our explanation model, shown in Figure~\ref{figure:separateModels} (left) is inspired by the recent success of utilising the transformer model architecture for extractive summarisation (\cite{liu-lapata-2019-text}). It learns to maximize the similarity of the extracted explanation with the human justification.  

We start by greedily selecting the top $k$ sentences from each claim's ruling comments that achieve the highest ROUGE-2 F1 score when compared to the gold justification. We choose $k = 4$, as that is the average number of sentences in veracity justifications. The selected sentences, referred to as oracles, serve as positive gold labels - $\mathbf{y}^E \in \{0,1\}^N $, where $N$ is the total number of sentences present in the ruling comments. Appendix~\ref{appendix:a} provides an overview of the coverage that the extracted oracles achieve compared to the gold justification. Appendix~\ref{appendix:o} further presents examples of the selected oracles, compared to the gold justification. 

At training time, we learn a function $f(X) = \mathbf{p}^E$, $\mathbf{p}^E \in \mathbb{R}^{1, N}$ that, based on the input $X$, the text of the claim and the ruling comments, predicts which sentence should be selected - \{0,1\}, to constitute the explanation. At inference time, we select the top $n = 4$ sentences with the highest confidence scores.

Our extraction model, represented by function $f(X)$, takes the contextual representations produced by the last layer of DistilBERT and feeds them into a feed-forward task-specific layer - $\mathbf{h} \in \mathbb{R}^{h}$. It is followed by the prediction layer $\mathbf{p}^{E} \in \mathbb{R}^{1,N}$ with sigmoid activation. The prediction is used  to optimise the cross-entropy loss function $\mathcal{L}_{E}=\mathcal{H}(\mathbf{p}^{E}, \mathbf{y}^{E})$.

\subsection{Veracity Prediction}\label{sec:veracityPred}
For the veracity prediction model, shown in Figure~\ref{figure:separateModels} (right), we learn a function $g(X) = \mathbf{p}^F$ that, based on the input X, predicts the veracity of the claim $\mathbf{y}^{F} \in Y_{F}$, $Y_F =$ \textit{\{true, false, half-true, barely-true, mostly-true, pants-on-fire\}}. 

The function $g(X)$ takes the contextual token representations from the last layer of DistilBERT and feeds them to a task-specific feed-forward layer $\mathbf{h} \in \mathbb{R}^{h}$. It is followed by the prediction layer with a softmax activation $\mathbf{p}^{F} \in \mathbb{R}^{6}$. We use the prediction to optimise a cross-entropy loss function $\mathcal{L}_{F}= \mathcal{H}(\mathbf{p}^{F}, \mathbf{y}^{F})$.

\subsection{Joint Training}\label{sec:jointTraining}
\begin{figure}
\centering
\includegraphics[width=0.6\linewidth]{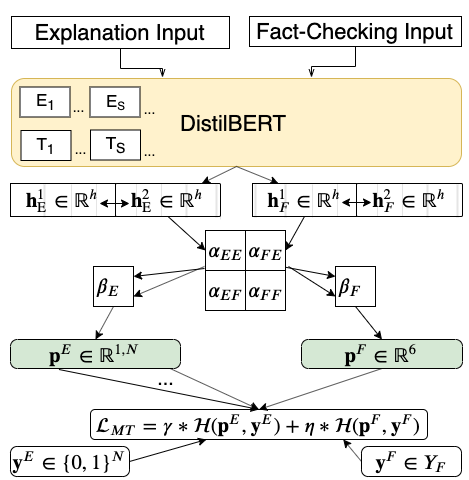}
\caption{Architecture of the \textit{Joint} model learning Explanation (E) and Fact-Checking (F) at the same time.}
\label{figure:jointmodel} 
\end{figure}

Finally, we learn a function $h(X) = (\mathbf{p}^E, \mathbf{p}^F)$ that, given the input X - the text of the claim and the ruling comments, predicts both the veracity explanation $\mathbf{p}^E$ and the veracity label $\mathbf{p}^F$ of a claim. The model is shown Figure~\ref{figure:jointmodel}. The function $h(X)$ takes the contextual embeddings $\mathbf{c}^E$ and $\mathbf{c}^F$ produced by the last layer of DistilBERT and feeds them into a cross-stitch layer (\cite{misra2016cross,conf/aaai/RuderBAS19}), which consists of two layers with two shared subspaces each - $\mathbf{h}_{E}^1$ and $\mathbf{h}_{E}^2$ for the explanation task and $\mathbf{h}_F^1$ and $\mathbf{h}_F^2$ for the veracity prediction task. In each of the two layers, there is one subspace for task-specific representations and one that learns cross-task representations. The subspaces and layers interact trough $\alpha$ values, creating the linear combinations $\widetilde{h}^i_E$ and $\widetilde{h}^j_F$, where i,j$\in \{1,2\}$:
%
\begin{equation}
\centering
\begin{bmatrix}
\widetilde{h}^i_E\\ 
\widetilde{h}^j_F
\end{bmatrix}
=
\begin{bmatrix}
\alpha_{EE} & \alpha_{EF}\\ 
\alpha_{FE} & \alpha_{FF}
\end{bmatrix}
\begin{bmatrix}
{h^i_E}^T & {h^j_F}^T\\ 
\end{bmatrix}  
\end{equation}

We further combine the resulting two subspaces for each task - $\widetilde{h}^i_E$ and $\widetilde{h}^j_F$ with parameters $\beta$ to produce one representation per task:
\begin{equation}
\centering
\widetilde{h}^T_P
=
\begin{bmatrix}
\beta_P^1\\ 
\beta_P^2
\end{bmatrix}^T
\begin{bmatrix}
\widetilde{h}^1_P & \widetilde{h}^2_P\\ 
\end{bmatrix}^T
\end{equation}
where P $\in \{E, F\}$ is the corresponding task.

Finally, we use the produced representation to predict $\mathbf{p}^{E}$ and $\mathbf{p}^{F}$, with feed-forward layers followed by sigmoid and softmax activations accordingly. We use the prediction to optimise the joint loss function $\mathcal{L}_{MT}= \gamma*\mathcal{H}(\mathbf{p}^{E}, \mathbf{y}^{E}) + \eta * \mathcal{H}(\mathbf{p}^{F}, \mathbf{y}^{F})$, where $\gamma$ and $\eta$ are used for weighted combination of the individual loss functions.

\section{Automatic Evaluation}\label{sec:automaticEval}
We first conduct an automatic evaluation of both the veracity prediction and veracity explanation models. 

\subsection{Experiments}\label{subsec:automaticExperiments}
In Table~\ref{tab:results:explanation}, we compare the performance of the two proposed models for generating extractive explanations. \textit{Explain-MT} is trained jointly with a veracity prediction model, and \textit{Explain-Extractive} is trained separately. We include the \textit{Lead-4} system (\cite{Nallapati:2017:SRN:3298483.3298681}) as a baseline, which selects as a summary the first four sentences from the ruling comments. The \textit{Oracle} system presents the best greedy approximation of the justification with sentences extracted from the ruling comments. It indicates the upper bound that could be achieved by extracting sentences from the ruling comments as an explanation. The performance of the models is measured using ROUGE-1, ROUGE-2, and ROUGE-L F1 scores.

In Table~\ref{tab:results:fact-checking}, we again compare two models - one trained jointly - \textit{MT-Veracity@Rul}, with the explanation generation task and one trained separately - \textit{Veracity@Rul}. As a baseline, we report the work of \cite{P17-2067}, who train a model based on the metadata available about the claim. It is the best known model that uses only the information available from the LIAR dataset and not the gold justification, which we aim at generating. 

We also provide two upper bounds serving as an indication of the approximate best performance that can be achieved given the gold justification. The first is the reported system performance from \cite{alhindi-etal-2018-evidence}, and the second - \textit{Veracity@Just}, is our veracity prediction model but trained on gold justifications. The \cite{alhindi-etal-2018-evidence} system is trained using a BiLSTM, while we train the \textit{Veracity@Just} model using the same model architecture as for predicting the veracity from the ruling comments with \textit{Veracity@Rul}. 

Lastly, \textit{Veracity@RulOracles} is the veracity model trained on the gold oracle sentences from the ruling comments. It provides a rough estimate of how much of the important information from the ruling comments is preserved in the oracles. The models are evaluated with a macro F1 score.

\subsection{Experimental Setup}\label{sec:experiments}

Our models employ the base, uncased version of the pre-trained DistilBERT model. The models are fed with text depending on the task set-up - claim and ruling sentences for the explanation and joint models; claim and ruling sentences, claim and oracle sentences or claim and justification for the fact-checking model. We insert a `[CLS]' token before the start of each ruling sentence (explanation model), before the claim (fact-checking model), or at the combination of both for the joint model. The text sequence is passed through a number of Transformer layers from DistilBERT. We use the `[CLS]' embeddings from the final contextual layer of DistilBERT and feed that in task-specific feed-forward layers $\mathbf{h} \in \mathbb{R}^{h}$, where h is 100 for the explanation task, 150 for the veracity prediction one and 100 for each of the joint cross-stitch subspaces. Following are the task-specific prediction layers ${p}^E$. 

The size of $h$ is picked with grid-search over \{50, 100, 150, 200, 300\}. We also experimented with replacing the feed-forward task-specific layers with an RNN or Transformer layer or including an activation function, which did not improve task performance.

The models are trained for up to 3 epochs, and, following \cite{liu-lapata-2019-text}, we evaluate the performance of the fine-tuned model on the validation set at every 50 steps, after the first epoch. We then select the model with the best ROUGE-2 F1 score on the validation set, thus, performing a potential early stopping. The learning rate used is 3e-5, which is chosen with a grid search over \{3e-5, 4e-5, 5e-5\}. We perform 175 warm-up steps (5\% of the total number of steps), after also experimenting with 0, 100, and 1000 warm-up steps. Optimisation is performed with AdamW (\cite{loshchilov2017fixing}), and the learning rate is scheduled with a warm-up linear schedule (\cite{goyal2017accurate}). The batch size during training and evaluation is 8.

The maximum input words to DistilBERT are 512, while the average length of the ruling comments is 904 words. To prevent the loss of any sentences from the ruling comments, we apply a sliding window over the input of the text and then merge the contextual representations of the separate sliding windows, mean averaging the representations in the overlap of the windows. The size of the sliding window is 300, with a stride of 60 tokens, which is the number of overlapping tokens between two successive windows. The maximum length of the encoded sequence is 1200. We find that these hyper-parameters have the best performance after experimenting with different values in a grid search.

We also include a dropout layer (with 0.1 rate for the separate and 0.15 for the joint model) after the contextual embedding provided by the transformer models and after the first linear layer as well.

The models optimise cross-entropy loss, and the joint model optimises a weighted combination of both losses. Weights are selected with a grid search - 0.9 for the task of explanation generation and 0.1 for veracity prediction. The best performance is reached with weights that bring the losses of the individual models to roughly the same scale.

\subsection{Results and Discussion}\label{subsec:automaticResults}

\begin{table}
\fontsize{10}{10}\selectfont
\centering
\begin{tabular}{lll}
\toprule
\textbf{Model} & \textbf{Val} & \textbf{Test}  \\ 
\midrule
\cite{P17-2067}, all metadata & 0.247 & 0.274 \\
\midrule
Veracity@RulOracles & 0.308 & 0.300 \\ 
Veracity@Rul & 0.313 & 0.313 \\ 
MT-Veracity@Rul & \textbf{0.321} & \textbf{0.323}  \\ 
\midrule
\cite{alhindi-etal-2018-evidence}@Just & 0.37 & 0.37 \\ 
Veracity@Just & \textbf{0.443}& \textbf{0.443} \\ 
\bottomrule
\end{tabular}
\caption{Results (Macro F1 scores) of the veracity prediction task on all of the six classes. The models are trained using the text from the ruling oracles (@RulOracles), ruling comment (@Rul), or the gold justification (@Just).}
\label{tab:results:fact-checking}
\end{table}

\begin{table}
\fontsize{10}{10}\selectfont
\centering
\begin{tabular}{l|ccc|ccc}
\toprule
\multirow{2}{*}{\textbf{Model}} & \multicolumn{3}{c|}{\textbf{Validation}}&  \multicolumn{3}{c}{\textbf{Test}} \\ 
& \textbf{ROUGE-1} & \textbf{ROUGE-2} & \textbf{ROUGE-L} & \textbf{ROUGE-1} & \textbf{ROUGE-2} & \textbf{ROUGE-L} \\ \midrule
Lead-4 & 27.92 & 6.94 & 24.26 & 28.11 & 6.96 & 24.38 \\
Oracle & 43.27 & 22.01 & 38.89 & 43.57 & 22.23 & 39.26 \\
\midrule
Explain-Extractive & \textbf{35.64} & \textbf{13.50} & \textbf{31.44} & \textbf{35.70} & \textbf{13.51} & \textbf{31.58} \\
Explain-MT & 35.18 & 12.94 & 30.95 & 35.13 & 12.90 & 30.93 \\
\bottomrule 		
\end{tabular}
\caption{Results of the veracity explanation generation task. The results are ROUGE-N F1 scores of the gene- \newline rated explanation w.r.t. the gold justification.} 
\label{tab:results:explanation}
\end{table} 

For each claim, our proposed joint model (see \ref{sec:jointTraining}) provides both (i) a veracity explanation and (ii) a veracity prediction. We compare our model's performance with models that learn to optimise these objectives \emph{separately}, as no other joint models have been proposed.
Table~\ref{tab:results:fact-checking} shows the results of veracity prediction, measured in terms of macro F1. 

Judging from the performance of both \textit{Veracity@Rul} and \textit{MT-Veracity@Rul}, we can assume that the task is very challenging. Even given a gold explanation (\cite{alhindi-etal-2018-evidence} and \textit{Veracity@Just}), the macro F1 remains below 0.5. This can be due to the small size of the dataset and/or the difficulty of the task even for human annotators. We further investigate the difficulty of the task in a human evaluation, presented in Section~\ref{sec:manualEval}. 


Comparing \textit{Veracity@RulOracles} and \textit{Veracity@Rul}, the latter achieves a slightly higher macro F1 score, indicating that the extracted ruling oracles, while approximating the gold justification, omit information that is important for veracity prediction. Finally, when the fact checking system is learned jointly with the veracity explanation system - \textit{MT-Veracity@Rul}, it achieves the best macro F1 score of the three systems. The objective to extract explanations provides information about regions in the ruling comments that are close to the gold explanation, which helps the veracity prediction model to choose the correct piece of evidence.

In Table~\ref{tab:results:explanation}, we present an evaluation of the generated explanations, computing ROUGE F1 score w.r.t. gold justification. 
Our first model, the \textit{Explain-Extractive} system, optimises the single objective of selecting explanation sentences. It outperforms the baseline, indicating that generating veracity explanations is possible.

\textit{Explain-Extractive} also outperforms the \textit{Explain-MT} system. While we would expect that training jointly with a veracity prediction objective would improve the performance of the explanation model, as it does for the veracity prediction model, we observe the opposite. This indicates a potential mismatch between the ruling oracles and the salient regions for the fact checking model. We also find a potential indication of that in the observed performance decrease when the veracity model is trained solely on the ruling oracles compared to the one trained on all of the ruling comments. We hypothesise that, when trained jointly with the veracity extraction component, the explanation model starts to also take into account the actual knowledge needed to perform the fact check, which might not match the exact wording present in the oracles, thus decreasing the overall performance of the explanation system. We further investigate this in a manual evaluation of which of the systems - Explain-MT and Explain-Extractive, generates explanations with better qualities and with more information about the veracity label.

Finally, comparing the performance of the extractive models and the \textit{Oracle}, we can conclude that there is still room for improvement of explanation systems when only considering extractive summarisation.

\subsection{A Case Study}\label{section:case}
Table~\ref{tab:example} presents two example explanations generated by the extractive vs. the multi-task model.
In the first example, the multi-task explanation achieves higher ROUGE scores than the extractive one. The corresponding extractive summary contains information that is not important for the final veracity label, which also appears to affect the ROUGE scores of the explanation. On the other hand, the multi-task model, trained jointly with a veracity prediction component, selects sentences that are more important for the fact check, which in this case is also beneficial for the final ROUGE score of the explanation.

In the second example, the multi-task explanation has lower ROUGE scores than the extractive one. We observe that the gold justification contains some sentences that are not relevant to the fact check, and the extractive summary is fooled to select explanation sentences that are close to the gold summary. As a result, the explanation does not provide enough information about the chosen veracity label. The multi-task model, on the other hand, selects sentences that are also contributing to the prediction of the veracity labels. Thus, its explanation turns out to be more beneficial for the final fact check even though it has a lower ROUGE score compared to the gold justification.

\begin{table}
\centering
\fontsize{10}{10}\selectfont
\begin{tabular}{p{1.0\textwidth}}
\toprule

\textbf{Label: }Mostly-true \textbf{Claim:} Says opponent U.S. Rep. Gwen Moore is Wisconsin's most absent member of Congress, missing nearly 17 percent of the House votes in the second quarter of 2012. \\
\textbf{Just:} Sebring says Moore is the state's "most absent" member of Congress and that she missed 17 percent of the House votes in the second quarter of 2012. Indeed, Moore missed nearly 17 percent of the votes in the second quarter. That is Moore's worst quarter, but it also amounts to Sebring doing some cherry-picking of data. A longer period shows a lower percentage, but even that places Moore at the top of the state's eight-member delegation. \\
\textbf{Explain-Extr:} Then he turns to the ad's main point: "Gwen Moore is Wisconsin's most absent member of Congress, missing nearly 17 percent of the House votes in the second quarter of 2012." That, again, compares to the median of 2.5 percent, according to GovTrack. Overall, the state delegation missed an average of 2.53 percent of the votes this session. But what about the other part of Sebring's claim, that Moore is the "most absent" member of the state delegation? \\
\textbf{Explain-MT:} Then he turns to the ad's main point: "Gwen Moore is Wisconsin's most absent member of Congress, missing nearly 17 percent of the House votes in the second quarter of 2012." If you take out the first two quarters of 2012, Moore missed an average of 3.4 percent of the votes in the other five quarters of this session. In the second quarter, the one cited by Sebring, the missed votes easily put Moore at the top of the state's eight-member House delegation. Looking at the full session, Moore missed 112 votes, or 6.98 percent. \\
\midrule
\textbf{Label:} Half-true \textbf{Claim:} Of the more than 1.3 million temporary mortgage modifications, over half have defaulted. \\
\textbf{Just:} In the final full week of the U.S. Senate race, how did Rubio fare on his numbers about the mortgage modification program? Rubio said "over 1.3 million temporary work-outs, over half have now defaulted,'' referring to a temporary mortgage modification program. \\
\textbf{Explain-Extr:} Over 1.3 million temporary work-outs, over half have now defaulted,'' he said. There have been 500,000 permanent (modifications). Rubio also said that more than half of those 1.3 million had defaulted." Rubio: "The temporary modifications." \\
\textbf{Explain-MT:} Rubio also said that more than half of those 1.3 million had "defaulted." "Over 1.3 million temporary work-outs, over half have now defaulted,'' he said. Of those permanent modifications, the majority survived while almost 29,000 were cancelled. The chart states that 699,924 trials had been cancelled -- so note that is slightly more than half.\\
\bottomrule
\end{tabular}
\caption{Examples of the generated explanation of the extractive (Explain-Extr) and the multi-task model (Explain-MT) compared to the gold justification (Just).}
\label{tab:example}
\end{table}

\section{Manual Evaluation}\label{sec:manualEval}
As the ROUGE score only accounts for word-level similarity between gold and predicted justifications, we also conduct a manual evaluation of the quality of the produced veracity explanations.

\subsection{Experiments}\label{subsec:manualExperiments}
\textbf{Explanation Quality}. We first provide a manual evaluation of the properties of three different types of explanations - gold justification, veracity explanation generated by the  \textit{Explain-MT}, and the ones generated by \textit{Explain-Extractive}. We ask three annotators to rank these explanations with the ranks 1, 2, 3, (first, second, and third place) according to four different criteria:

\begin{enumerate}[noitemsep]
    \item \textbf{Coverage.} The explanation contains important, salient information and does not miss any important points that contribute to the fact check.
    \item \textbf{Non-redundancy.} The summary does not contain any information that is redundant/repeated/not relevant to the claim and the fact check.
    \item \textbf{Non-contradiction.} The summary does not contain any pieces of information that are contradictory to the claim and the fact check. 
    \item \textbf{Overall.} Rank the explanations by their overall quality.
\end{enumerate}

We also allow ties, meaning that two veracity explanations can receive the same rank if they appear the same. 

For the annotation task set-up, we randomly select a small set of 40 instances from the test set and collect the three different veracity explanations for each of them. We did not provide the participants with information of the three different explanations and shuffled them randomly to prevent easily creating a position bias for the explanations. The annotators worked separately without discussing any details about the annotation task.

\textbf{Explanation Informativeness}. In the second manual evaluation task, we study how well the veracity explanations manage to address the information need of the user and if they sufficiently describe the veracity label. We, therefore, design the annotation task asking annotators to provide a veracity label for a claim based on a veracity explanation coming from the justification, the \textit{Explain-MT}, or the \textit{Explain-Extractive} system. The annotators have to provide a veracity label on two levels - binary classification - true or false, and six-class classification - true, false, half-true, barely-true, mostly-true, pants-on-fire. Each of them has to provide the label for 80 explanations, and there are two annotators per explanation. 

\subsection{Results and Discussion}\label{sec:manualResults}
\textbf{Explanation Quality}. Table~\ref{tab:results:man1} presents the results from the manual evaluation in the first set-up, described in Section~\ref{sec:manualEval}, where annotators ranked the explanations according to four different criteria. 

We compute Krippendorff's $\alpha$ inter-annotator agreement (IAA, \cite{hayes2007answering}) as it is suited for ordinal values. The corresponding alpha values are 0.26 for \textit{Coverage}, 0.18 for \textit{Non-redundancy}, -0.1 for \textit{Non-contradiction}, and 0.32 for \textit{Overall}, where $0.67<\alpha <0.8$ is regarded as significant, but vary a lot for different domains. 

We assume that the low IAA can be attributed to the fact that in ranking/comparison tasks for manual evaluation, the agreement between annotators might be affected by small differences in one rank position in one of the annotators as well as by the annotator bias towards ranking explanations as ties. Taking this into account, we choose to present the mean average recall for each of the annotators instead. Still, we find that their preferences are not in a perfect agreement and report only what the majority agrees upon. We also consider that the low IAA reveals that the task might be ``already too difficult for humans''. This insight proves to be important on its own as existing machine summarisation/question answering studies involving human evaluation do not report IAA scores (\cite{liu-lapata-2019-text}), thus, leaving essential details about the nature of the evaluation tasks ambiguous. 


We find that the gold explanation is ranked the best for all criteria except for \textit{Non-contradiction}, where one of the annotators found that it contained more contradictory information than the automatically generated explanations, but Krippendorff's $\alpha$ indicates that there is no agreement between the annotations for this criterion. 

Out of the two extractive explanation systems, \textit{Explain-MT} ranks best in Coverage and Overall criteria, with 0.21 and 0.13 corresponding improvements in the ranking position. These results contradict the automatic evaluation in Section~\ref{subsec:automaticResults}, where the explanation of \textit{Explain-MT} had lower ROUGE F1 scores. This indicates that an automatic evaluation might be insufficient in estimating the information conveyed by the particular explanation.

On the other hand, \textit{Explain-Extr} is ranked higher than \textit{Explain-MT} in terms of Non-redundancy and Non-contradiction, where the last criterion was disagreed upon, and the rank improvement for the first one is only marginal at 0.04. 

This implies that a veracity prediction objective is not necessary to produce natural-sounding explanations (\textit{Explain-Extr}), but that the latter is useful for generating better explanations overall and with higher coverage \textit{Explain-MT}. 

\begin{table}
\fontsize{10}{10}\selectfont
\centering
\begin{tabular}{lccc}
\toprule
\textbf{Annotators} & \textbf{Just} & \textbf{Explain-Extr} & \textbf{Explain-MT}\\ \midrule
\multicolumn{4}{c}{Coverage} \\ \midrule
All & \textbf{1.48} & 1.89 & \cellcolor{myblue}1.68 \\
1st & \textbf{1.50} & 2.08 & \cellcolor{myblue}1.87 \\
2nd & \textbf{1.74} & 2.16 & \cellcolor{myblue}1.84 \\
3rd & \textbf{1.21} & 1.42 & \cellcolor{myblue}1.34 \\ \midrule
\multicolumn{4}{c}{Non-redundancy} \\ \midrule
All & \textbf{1.48} & \cellcolor{myblue}1.75 & 1.79 \\
1st & \textbf{1.34} & 1.84 & \cellcolor{myblue}1.76 \\
2nd & \textbf{1.71} & \cellcolor{myblue}1.97 & 2.08 \\
3rd & \textbf{1.40} & \cellcolor{myblue}1.42 & 1.53 \\ \midrule
\multicolumn{4}{c}{Non-contradiction} \\ \midrule
All & 1.45 & \cellcolor{myblue}\textbf{1.40} & 1.48 \\
1st & \textbf{1.13} & 1.45 & \cellcolor{myblue}1.34 \\
2nd & 2.18  & \cellcolor{myblue}\textbf{1.63} & 1.92 \\
3rd & \textbf{1.03} & \cellcolor{myblue}1.13  & 1.18 \\ \midrule
\multicolumn{4}{c}{Overall} \\ \midrule
All & \textbf{1.58} & 2.03 & \cellcolor{myblue}1.90 \\
1st & \textbf{1.58} & 2.18 & \cellcolor{myblue}1.95 \\
2nd & \textbf{1.74} & 2.13 & \cellcolor{myblue}1.92 \\
3rd & \textbf{1.42} & \cellcolor{myblue}1.76  & 1.82 \\
\bottomrule 		
\end{tabular}

\caption{Mean Average Ranks (MAR) of the explanations for each of the four evaluation criteria. The explanations come from the gold justification (Just), the generated explanation (Explain-Extr), and the explanation learned jointly (Explain-MT) with the veracity prediction model. The lower MAR indicates a higher ranking, i.e., a better quality of an explanation. For each row, the best results are in bold, and the best results with automatically generated explanations are in blue.} 
\label{tab:results:man1}
\end{table} 

\textbf{Explanation Informativeness}. Table~\ref{tab:results:man2} presents the results from the second manual evaluation task, where annotators provided the veracity of a claim based on an explanation from one of the systems. We here show the results for binary labels, as annotators struggled to distinguish between 6 labels. 
The latter follows the same trends and are shown in Appendix~\ref{appendix:q}. 

The Fleiss' $\kappa$ IAA for binary prediction is: \textit{Just} -- 0.269,  \textit{Explain-MT} -- 0.345, \textit{Explain-Extr} -- 0.399. The highest agreement is achieved for \textit{Explain-Extr}, which is supported by the highest proportion of agreeing annotations from Table~\ref{tab:results:man2}. Surprisingly, the gold explanations from \textit{Just} were most disagreed upon. 
Apart from that, looking at the agreeing annotations, gold explanations were found most sufficient in providing information about the veracity label and also were found to explain the correct label most of the time. They are followed by the explanations produced by \textit{Explain-MT}. This supports the findings of the first manual evaluation, where the \textit{Explain-MT} ranked better in coverage and overall quality than \textit{Explain-Extr}.

\begin{table}
\fontsize{10}{10}\selectfont
\centering
\begin{tabular}{clrrr}
\toprule
 & & \textbf{Just} & \textbf{Explain-Extr} & \textbf{Explain-MT} \\ \midrule
$\nwarrow$ & Agree-C & \textbf{0.403} & 0.237 & \cellcolor{myblue}0.300 \\
$\searrow$ & Agree-NS & \textbf{0.065} &  0.250 & \cellcolor{myblue}0.188 \\
$\searrow$ & Agree-NC & \textbf{0.064} & 0.113 & \cellcolor{myblue}0.088 \\
$\searrow$ & Disagree & 0.468 & \cellcolor{myblue}\textbf{0.400} & 0.425\\
\bottomrule 		
\end{tabular}
\caption{Manual veracity labelling, given a particular explanation from the gold justification (Just), the generated explanation (Explain-Extr), and the explanation learned jointly (Explain-MT) with the veracity prediction model. Percentages of the dis/agreeing annotator predictions are shown, with agreement percentages split into: \emph{correct} according to the gold label (Agree-C), \emph{incorrect} (Agree-NC) or \emph{insufficient information} (Agree-NS). The first column indicates whether higher ($\nwarrow$) or lower ($\searrow$) values are better. For each row, the best results are in bold, and the best results with automatically generated explanations are in blue.}
\label{tab:results:man2}
\end{table} 

\section{Related Work}
 
\textbf{Generating Explanations.}
Generating textual explanations for model predictions is an understudied problem. The first study was \cite{NIPS2018_8163}, who generate explanations for the task of natural language inference. The authors explore three different set-ups: prediction pipelines with explanation followed by prediction, and prediction followed by explanation, and a joint multi-task learning setting. They find that first generating the explanation produces better results for the explanation task, but harms classification accuracy. 

We are the first to provide a study on generating veracity explanations. We show that the generated explanations improve veracity prediction performance, and find that jointly optimising the veracity explanation and veracity prediction objectives improves the coverage and the overall quality of the explanations.

\textbf{Fact Checking Interpretability.} Interpreting fact checking systems has been explored in a few studies. \cite{shu2019defend} study the interpretability of a system that fact checks full-length news pages by leveraging user comments from social platforms. They propose a co-attention framework, which selects both salient user comments and salient sentences from news articles. \cite{yang2019xfake} build an interpretable fact-checking system XFake, where shallow student and self-attention, among others, are used to highlight parts of the input. This is done solely based on the statement without considering any supporting facts. 
In our work, we research models that generate human-readable explanations, and directly optimise the quality of the produced explanations instead of using attention weights as a proxy. We use the LIAR dataset to train such models, which contains fact checked single-sentence claims that already contain professional justifications.
As a result, we make an initial step towards automating the generation of professional fact checking justifications.
 
\textbf{Veracity Prediction.}
Several studies have built fact checking systems for the LIAR dataset (\cite{P17-2067}). The model proposed by \cite{karimi-etal-2018-multi} reaches 0.39 accuracy by using metadata, ruling comments, and justifications. \cite{alhindi-etal-2018-evidence} also trains a classifier, that, based on the statement and the justification, achieves 0.37 accuracy. To the best of our knowledge, \cite{long2017fake} is the only system that, without using justifications, achieves a performance above the baseline of \cite{P17-2067}, an accuracy of 0.415---the current state-of-the-art performance on the LIAR dataset. Their model learns a veracity classifier with speaker profiles. 
While using metadata and external speaker profiles might provide substantial information for fact checking, they also have the potential to introduce biases towards a certain party or a speaker. 

In this study, we propose a method to generate veracity explanations that would explain the reasons behind a certain veracity label independently of the speaker profile. 
Once trained, such methods could then be applied to other fact checking instances without human-provided explanations or even to perform end-to-end veracity prediction and veracity explanation generation given a claim.

Substantial research on fact checking methods exists for the FEVER dataset~(\cite{thorne-etal-2018-fever}), which comprises rewritten claims from Wikipedia.
Systems typically perform document retrieval, evidence selection, and veracity prediction. Evidence selection is performed using keyword matching (\cite{malon-2018-team,yoneda-etal-2018-ucl}), supervised learning (\cite{hanselowski-etal-2018-ukp, chakrabarty-etal-2018-robust}) or sentence similarity scoring (\cite{Ma:2018:DRS:3184558.3188729, mohtarami-etal-2018-automatic, Xu2019AdversarialDA}). More recently, the multi-domain dataset MultiFC (\cite{augenstein-etal-2019-multifc}) has been proposed, which is also distributed with evidence pages. Unlike FEVER, it contains real-world claims, crawled from different fact checking portals.

While FEVER and MultiFC are larger datasets for fact checking than LIAR-PLUS, they do not contain veracity explanations and can thus not easily be used to train joint veracity prediction and explanation generation models, hence we did not use them in this study.

\section{Conclusions}

We presented the first study on generating veracity explanations, and we showed that veracity prediction can be combined with veracity explanation generation and that the multi-task set-up improves the performance of the veracity system. A manual evaluation shows that the coverage and the overall quality of the explanation system is also improved in the multi-task set-up.

For future work, an obvious next step is to investigate the possibility of generating veracity explanations from evidence pages crawled from the Web. Furthermore, other approaches of generating veracity explanations should be investigated, especially as they could improve fluency or decrease the redundancy of the generated text.

\section{Acknowledgments}
$\begin{array}{l}\includegraphics[width=1cm]{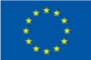} \end{array}$ 
This project has received funding from the European Union’s Horizon 2020 research and innovation programme under the Marie Skłodowska-Curie grant agreement No 801199.

\clearpage



\section{Appendix}\label{appendices}

\subsection{Comparison of Different Sources of Evidence}\label{appendix:a}
Table \ref{tab:evidence} provides an overview of the ruling comments and the ruling oracles compared to the justification. The high recall in both ROUGE-1 and ROUGE-F achieved by the ruling comments indicates that there is a substantial coverage, i.e. over 70\% of the words and long sequences in the justification can be found in the ruling comments. On the other hand, there is a small coverage for the bi-grams. Selecting the oracles from all of the ruling sentences increases ROUGE-F1 scores mainly by improving the precision. 

\begin{table}[h!] 
\centering
\begin{tabular}{l|rrr|rrr|rrr}
\toprule
\multirow{2}{*}{\textbf{Evidence Source}}& \multicolumn{3}{c|}{\textbf{ROUGE-1}}& \multicolumn{3}{c|}{\textbf{ROUGE-2}} & \multicolumn{3}{c}{\textbf{ROUGE-L}} \\ 
& P & R & F1 & P & R & F1 & P & R & F1 \\ \midrule
Ruling & 8.65 & 78.65 & 14.84 & 3.53 & 33.76 & 6.16 & 8.10 & 74.14 & 13.92 \\
Ruling Oracle & 43.97 & 49.24 & 43.79 & 22.45 & 24.50 & 22.03 & 39.70 & 44.10 & 39.37 \\ \bottomrule
\end{tabular}
\caption{Comparison of sources of evidence - Ruling Comments and Ruling Oracles comapred to the target \newline justification summary.}
\label{tab:evidence}
\end{table}

\subsection{Extractive Gold Oracle Examples}
\label{appendix:o}
Table~\ref{tab:oracle-example} presents examples of selected oracles that serve as gold labels during training the extractive summarization model. The three examples represent oracles with different degrees of matching the gold summary. The first row presents an oracle that matches the gold summary with a ROUGE-L F1 score of 60.40 compared to the gold summary. It contains all of the important information from the gold summary and even points precise, not rounded, numbers. The next example has a ROUGE-L F1 score of 43.33, which is close to the average ROUGE-L F1 score for the oracles. The oracle again conveys the main points from the gold justification, thus, being sufficient for the claim's explanation. Finally, the third example is of an oracle with a ROUGE-L F1 score of 25.59. The selected oracle sentences still succeed in presenting the main points from the gold justification, which is at a more detailed level presenting specific findings. The latter might be found as a positive consequence as it presents the particular findings of the journalist that led to selecting the veracity label.

\begin{table}
\centering
\scriptsize
\begin{tabular}{p{0.98\textwidth}}
\toprule
\textbf{Claim: }``The president promised that if he spent money on a stimulus program that unemployment would go to 5.7 percent or 6 percent. Those were his words.'' \\
\textbf{Label: }Mostly-False  \\ 
\textbf{Just:} Bramnick said ``the president promised that if he spent money on a stimulus program that unemployment would go to 5.7 percent or 6 percent.
Those were his words.''
Two economic advisers estimated in a 2009 report that with the stimulus plan, the unemployment rate would peak near 8 percent before dropping to less than 6 percent by now.
Those are critical details Bramnick’s statement ignores.
To comment on this ruling, go to NJ.com. \\
\textbf{Oracle:}  ``The president promised that if he spent money on a stimulus program that unemployment would go to 5.7 percent or 6 percent.
Those were his words,'' Bramnick said in a Sept. 7 interview on NJToday.
But with the stimulus plan, the report projected the nation’s jobless rate would peak near 8 percent in 2009 before falling to about 5.5 percent by now.
So the estimates in the report were wrong.\\

\midrule
\textbf{Claim: }The Milwaukee County bus system has ``among the highest fares in the nation.''  \\
\textbf{Label:} False \\
\textbf{Just:} Larson said the Milwaukee County bus system has ``among the highest fares in the nation.''
But the system’s’ \$2.25 cash fare wasn’t at the top of a national comparison, with fares reaching as high as \$4 per trip.
And regular patrons who use a Smart Card are charged just \$1.75 a ride, making the Milwaukee County bus system about on par with average costs. \\
\textbf{Oracle:} Larson said the Milwaukee County bus system has ``among the highest fares in the nation.''
Patrons who get a Smart Card pay \$1.75 per ride.
At the time, nine cities on that list charged more than Milwaukee’s \$2.25 cash fare.
The highest fare -- in Nashville -- was \$4 per ride.\\
\midrule
 \textbf{Claim: }``The Republican who was just elected governor of the great state of Florida paid his campaign staffers, not with money, but with American Express gift cards.''  \\
 \textbf{Label: }Half-True \\
\textbf{Just:} First, we think many people might think Maddow was referring to all campaign workers, but traditional campaign staffers -- the people working day in and day out on the campaign -- were paid by check, like any normal job.
A Republican Party official said it was simply an easier, more efficient and quicker way to pay people.
And second, it's not that unusual.
In 2008, Obama did the same thing. \\
\textbf{Oracle:}
``It's a simpler and quicker way of compensating short-term help.''
Neither Conston nor Burgess said how many temporary campaign workers were paid in gift cards.
When asked how he was paid, Palecheck said: ``Paid by check, like any normal employee there.''
In fact, President Barack Obama's campaign did the same thing in 2008. \\
\bottomrule
\end{tabular}
\caption{Examples of the extracted oracle summaries (Oracle) compared to the gold justification (Just).}
\label{tab:oracle-example}
\end{table}

\subsection{Manual 6-Way Veracity Prediction from Explanations}\label{appendix:q}
The Fleiss' $\kappa$ agreement for the 6-label manual annotations is: 0.20 on the \textit{Just} explanations, 0.230 on the \textit{Explain-MT} explanations, and 0.333 on the \textit{Explain-Extr} system. Table~\ref{tab:results:man2:6-way} represent the results of the manual veracity prediction with six classes.

\begin{table}[h!]
\centering
\begin{tabular}{clrrr}
\toprule
 & & Just & Explain-Extr & Explain-MT \\ \midrule
$\nwarrow$ & Agree-C & \textbf{0.208} & 0.138 & \cellcolor{myblue}0.163 \\
$\searrow$ & Agree-NS & \textbf{0.065} &  0.250 & \cellcolor{myblue}0.188 \\
$\searrow$ & Agree-NC & \textbf{0.052} & 0.100 & \cellcolor{myblue}0.075 \\
$\searrow$ & Disagree & 0.675 & \cellcolor{myblue}\textbf{0.513} & 0.575\\
\bottomrule 		
\end{tabular}
\caption{Manual classification of veracity label - true, false, half-true, barely-true, mostly-true, pants-on-fire, given a particular explanations from the gold justification (Just), the generated explanation (Explain-Extr) and the explanation learned jointly with the veracity prediction model (Explain-MT). Presented are percentages of the dis/agreeing annotator predictions, where the agreement percentages are split to: correct according to the gold label (Agree-C) , incorrect (Agree-NC) or with not sufficient information (Agree-NS). The first column indicates whether higher ($\nwarrow$) or lower ($\searrow$) values are better. At each row, the best set of explanations is in bold and the best automatic explanations are in blue.}
\label{tab:results:man2:6-way}
\end{table}

\medskip

\printbibliography[
heading=bibintoc,
title={Bibliography}
]

@inproceedings{atanasova-etal-2020-generating,
    title = "Generating Label Cohesive and Well-Formed Adversarial Claims",
    author = "Atanasova, Pepa  and
      Wright, Dustin  and
      Augenstein, Isabelle",
    booktitle = "Proceedings of the 2020 Conference on Empirical Methods in Natural Language Processing (EMNLP)",
    month = nov,
    year = "2020",
    address = "Online",
    publisher = "Association for Computational Linguistics",
    url = "https://www.aclweb.org/anthology/2020.emnlp-main.256",
    doi = "10.18653/v1/2020.emnlp-main.256",
    pages = "3168--3177",
}

@inproceedings{atanasova-etal-2020-diagnostic,
    title = "A Diagnostic Study of Explainability Techniques for Text Classification",
    author = "Atanasova, Pepa  and
      Simonsen, Jakob Grue  and
      Lioma, Christina  and
      Augenstein, Isabelle",
    booktitle = "Proceedings of the 2020 Conference on Empirical Methods in Natural Language Processing (EMNLP)",
    month = nov,
    year = "2020",
    address = "Online",
    publisher = "Association for Computational Linguistics",
    url = "https://www.aclweb.org/anthology/2020.emnlp-main.263",
    doi = "10.18653/v1/2020.emnlp-main.263",
    pages = "3256--3274",
}

@inproceedings{nooralahzadeh-etal-2020-zero,
    title = "Zero-Shot Cross-Lingual Transfer with Meta Learning",
    author = "Nooralahzadeh, Farhad  and
      Bekoulis, Giannis  and
      Bjerva, Johannes  and
      Augenstein, Isabelle",
    booktitle = "Proceedings of the 2020 Conference on Empirical Methods in Natural Language Processing (EMNLP)",
    month = nov,
    year = "2020",
    address = "Online",
    publisher = "Association for Computational Linguistics",
    url = "https://www.aclweb.org/anthology/2020.emnlp-main.368",
    pages = "4547--4562",
}

@inproceedings{bjerva-etal-2020-subjqa,
    title = "{SubjQA}: {A} {D}ataset for {S}ubjectivity and {R}eview {C}omprehension",
    author = "Bjerva, Johannes  and
      Bhutani, Nikita  and
      Golshan, Behzad  and
      Tan, Wang-Chiew  and
      Augenstein, Isabelle",
    booktitle = "Proceedings of the 2020 Conference on Empirical Methods in Natural Language Processing (EMNLP)",
    month = nov,
    year = "2020",
    address = "Online",
    publisher = "Association for Computational Linguistics",
    url = "https://www.aclweb.org/anthology/2020.emnlp-main.442",
    pages = "5480--5494",
}

@inproceedings{wright-augenstein-2020-transformer,
    title = "Transformer Based Multi-Source Domain Adaptation",
    author = "Wright, Dustin  and
      Augenstein, Isabelle",
    booktitle = "Proceedings of the 2020 Conference on Empirical Methods in Natural Language Processing (EMNLP)",
    month = nov,
    year = "2020",
    address = "Online",
    publisher = "Association for Computational Linguistics",
    url = "https://www.aclweb.org/anthology/2020.emnlp-main.639",
    doi = "10.18653/v1/2020.emnlp-main.639",
    pages = "7963--7974",
}

@inproceedings{a-augenstein-2020-2kenize,
    title = "2kenize: Tying Subword Sequences for {C}hinese Script Conversion",
    author = "A, Pranav  and
      Augenstein, Isabelle",
    booktitle = "Proceedings of the 58th Annual Meeting of the Association for Computational Linguistics",
    month = jul,
    year = "2020",
    address = "Online",
    publisher = "Association for Computational Linguistics",
    url = "https://www.aclweb.org/anthology/2020.acl-main.648",
    doi = "10.18653/v1/2020.acl-main.648",
    pages = "7257--7272",
}

@inproceedings{atanasova-etal-2020-generating-fact,
    title = "Generating Fact Checking Explanations",
    author = "Atanasova, Pepa  and
      Simonsen, Jakob Grue  and
      Lioma, Christina  and
      Augenstein, Isabelle",
    booktitle = "Proceedings of the 58th Annual Meeting of the Association for Computational Linguistics",
    month = jul,
    year = "2020",
    address = "Online",
    publisher = "Association for Computational Linguistics",
    url = "https://www.aclweb.org/anthology/2020.acl-main.656",
    doi = "10.18653/v1/2020.acl-main.656",
    pages = "7352--7364",
}

@inproceedings{wright-augenstein-2020-claim,
    title = "Claim Check-Worthiness Detection as Positive Unlabelled Learning",
    author = "Wright, Dustin  and
      Augenstein, Isabelle",
    booktitle = "Findings of the Association for Computational Linguistics: EMNLP 2020",
    month = nov,
    year = "2020",
    address = "Online",
    publisher = "Association for Computational Linguistics",
    url = "https://www.aclweb.org/anthology/2020.findings-emnlp.43",
    doi = "10.18653/v1/2020.findings-emnlp.43",
    pages = "476--488",
}

@inproceedings{rogers-augenstein-2020-improve,
    title = "What Can We Do to Improve Peer Review in {NLP}?",
    author = "Rogers, Anna  and
      Augenstein, Isabelle",
    booktitle = "Findings of the Association for Computational Linguistics: EMNLP 2020",
    month = nov,
    year = "2020",
    address = "Online",
    publisher = "Association for Computational Linguistics",
    url = "https://www.aclweb.org/anthology/2020.findings-emnlp.112",
    pages = "1256--1262",
}

@inproceedings{muttenthaler-etal-2020-unsupervised,
    title = "Unsupervised Evaluation for Question Answering with Transformers",
    author = "Muttenthaler, Lukas  and
      Augenstein, Isabelle  and
      Bjerva, Johannes",
    booktitle = "Proceedings of the Third BlackboxNLP Workshop on Analyzing and Interpreting Neural Networks for NLP",
    month = nov,
    year = "2020",
    address = "Online",
    publisher = "Association for Computational Linguistics",
    url = "https://www.aclweb.org/anthology/2020.blackboxnlp-1.8",
    pages = "83--90",
}

@inproceedings{bjerva-etal-2020-sigtyp,
    title = "{SIGTYP} 2020 Shared Task: Prediction of Typological Features",
    author = "Bjerva, Johannes  and
      Salesky, Elizabeth  and
      Mielke, Sabrina J.  and
      Chaudhary, Aditi  and
      Giuseppe, Celano  and
      Ponti, Edoardo Maria  and
      Vylomova, Ekaterina  and
      Cotterell, Ryan  and
      Augenstein, Isabelle",
    booktitle = "Proceedings of the Second Workshop on Computational Research in Linguistic Typology",
    month = nov,
    year = "2020",
    address = "Online",
    publisher = "Association for Computational Linguistics",
    url = "https://www.aclweb.org/anthology/2020.sigtyp-1.1",
    pages = "1--11",
}

@inproceedings{augenstein-etal-2019-multifc,
    title = "{M}ulti{FC}: A Real-World Multi-Domain Dataset for Evidence-Based Fact Checking of Claims",
    author = "Augenstein, Isabelle  and
      Lioma, Christina  and
      Wang, Dongsheng  and
      Chaves Lima, Lucas  and
      Hansen, Casper  and
      Hansen, Christian  and
      Simonsen, Jakob Grue",
    booktitle = "Proceedings of the 2019 Conference on Empirical Methods in Natural Language Processing and the 9th International Joint Conference on Natural Language Processing (EMNLP-IJCNLP)",
    month = nov,
    year = "2019",
    address = "Hong Kong, China",
    publisher = "Association for Computational Linguistics",
    url = "https://www.aclweb.org/anthology/D19-1475",
    doi = "10.18653/v1/D19-1475",
    pages = "4685--4697",
}

@inproceedings{hartmann-etal-2019-mapping,
    title = "Mapping (Dis-)Information Flow about the {MH}17 Plane Crash",
    author = "Hartmann, Mareike  and
      Golovchenko, Yevgeniy  and
      Augenstein, Isabelle",
    booktitle = "Proceedings of the Second Workshop on Natural Language Processing for Internet Freedom: Censorship, Disinformation, and Propaganda",
    month = nov,
    year = "2019",
    address = "Hong Kong, China",
    publisher = "Association for Computational Linguistics",
    url = "https://www.aclweb.org/anthology/D19-5006",
    doi = "10.18653/v1/D19-5006",
    pages = "45--55",
}

@inproceedings{bjerva-etal-2019-transductive,
    title = "Transductive Auxiliary Task Self-Training for Neural Multi-Task Models",
    author = "Bjerva, Johannes  and
      Kann, Katharina  and
      Augenstein, Isabelle",
    booktitle = "Proceedings of the 2nd Workshop on Deep Learning Approaches for Low-Resource NLP (DeepLo 2019)",
    month = nov,
    year = "2019",
    address = "Hong Kong, China",
    publisher = "Association for Computational Linguistics",
    url = "https://www.aclweb.org/anthology/D19-6128",
    doi = "10.18653/v1/D19-6128",
    pages = "253--258",
}

@inproceedings{abdou-etal-2019-x,
    title = "{X}-{W}iki{RE}: A Large, Multilingual Resource for Relation Extraction as Machine Comprehension",
    author = "Abdou, Mostafa  and
      Sas, Cezar  and
      Aralikatte, Rahul  and
      Augenstein, Isabelle  and
      S{\o}gaard, Anders",
    booktitle = "Proceedings of the 2nd Workshop on Deep Learning Approaches for Low-Resource NLP (DeepLo 2019)",
    month = nov,
    year = "2019",
    address = "Hong Kong, China",
    publisher = "Association for Computational Linguistics",
    url = "https://www.aclweb.org/anthology/D19-6130",
    doi = "10.18653/v1/D19-6130",
    pages = "265--274",
}

@article{bjerva-etal-2019-language,
    title = "What Do Language Representations Really Represent?",
    author = {Bjerva, Johannes  and
      {\"O}stling, Robert  and
      Veiga, Maria Han  and
      Tiedemann, J{\"o}rg  and
      Augenstein, Isabelle},
    journal = "Computational Linguistics",
    volume = "45",
    number = "2",
    month = jun,
    year = "2019",
    url = "https://www.aclweb.org/anthology/J19-2006",
    doi = "10.1162/coli_a_00351",
    pages = "381--389",
}

@proceedings{ws-2019-representation,
    title = "Proceedings of the 4th Workshop on Representation Learning for NLP (RepL4NLP-2019)",
    editor = "Augenstein, Isabelle  and
      Gella, Spandana  and
      Ruder, Sebastian  and
      Kann, Katharina  and
      Can, Burcu  and
      Welbl, Johannes  and
      Conneau, Alexis  and
      Ren, Xiang  and
      Rei, Marek",
    month = aug,
    year = "2019",
    address = "Florence, Italy",
    publisher = "Association for Computational Linguistics",
    url = "https://www.aclweb.org/anthology/W19-4300",
}

@inproceedings{hoyle-etal-2019-combining,
    title = "{C}ombining {S}entiment {L}exica with a {M}ulti-{V}iew {V}ariational {A}utoencoder",
    author = "Hoyle, Alexander Miserlis  and
      Wolf-Sonkin, Lawrence  and
      Wallach, Hanna  and
      Cotterell, Ryan  and
      Augenstein, Isabelle",
    booktitle = "Proceedings of the 2019 Conference of the North {A}merican Chapter of the Association for Computational Linguistics: Human Language Technologies, Volume 1 (Long and Short Papers)",
    month = jun,
    year = "2019",
    address = "Minneapolis, Minnesota",
    publisher = "Association for Computational Linguistics",
    url = "https://www.aclweb.org/anthology/N19-1065",
    doi = "10.18653/v1/N19-1065",
    pages = "635--640",
}

@inproceedings{hartmann-etal-2019-issue,
    title = "Issue Framing in Online Discussion Fora",
    author = "Hartmann, Mareike  and
      Jansen, Tallulah  and
      Augenstein, Isabelle  and
      S{\o}gaard, Anders",
    booktitle = "Proceedings of the 2019 Conference of the North {A}merican Chapter of the Association for Computational Linguistics: Human Language Technologies, Volume 1 (Long and Short Papers)",
    month = jun,
    year = "2019",
    address = "Minneapolis, Minnesota",
    publisher = "Association for Computational Linguistics",
    url = "https://www.aclweb.org/anthology/N19-1142",
    doi = "10.18653/v1/N19-1142",
    pages = "1401--1407",
}

@inproceedings{bjerva-etal-2019-probabilistic,
    title = "A Probabilistic Generative Model of Linguistic Typology",
    author = "Bjerva, Johannes  and
      Kementchedjhieva, Yova  and
      Cotterell, Ryan  and
      Augenstein, Isabelle",
    booktitle = "Proceedings of the 2019 Conference of the North {A}merican Chapter of the Association for Computational Linguistics: Human Language Technologies, Volume 1 (Long and Short Papers)",
    month = jun,
    year = "2019",
    address = "Minneapolis, Minnesota",
    publisher = "Association for Computational Linguistics",
    url = "https://www.aclweb.org/anthology/N19-1156",
    doi = "10.18653/v1/N19-1156",
    pages = "1529--1540",
}

@inproceedings{hoyle-etal-2019-unsupervised,
    title = "Unsupervised Discovery of Gendered Language through Latent-Variable Modeling",
    author = "Hoyle, Alexander Miserlis  and
      Wolf-Sonkin, Lawrence  and
      Wallach, Hanna  and
      Augenstein, Isabelle  and
      Cotterell, Ryan",
    booktitle = "Proceedings of the 57th Annual Meeting of the Association for Computational Linguistics",
    month = jul,
    year = "2019",
    address = "Florence, Italy",
    publisher = "Association for Computational Linguistics",
    url = "https://www.aclweb.org/anthology/P19-1167",
    doi = "10.18653/v1/P19-1167",
    pages = "1706--1716",
}

@inproceedings{bjerva-etal-2019-uncovering,
    title = "Uncovering Probabilistic Implications in Typological Knowledge Bases",
    author = "Bjerva, Johannes  and
      Kementchedjhieva, Yova  and
      Cotterell, Ryan  and
      Augenstein, Isabelle",
    booktitle = "Proceedings of the 57th Annual Meeting of the Association for Computational Linguistics",
    month = jul,
    year = "2019",
    address = "Florence, Italy",
    publisher = "Association for Computational Linguistics",
    url = "https://www.aclweb.org/anthology/P19-1382",
    doi = "10.18653/v1/P19-1382",
    pages = "3924--3930",
}

@inproceedings{gonzalez-etal-2018-strong,
    title = "A strong baseline for question relevancy ranking",
    author = "Gonzalez, Ana  and
      Augenstein, Isabelle  and
      S{\o}gaard, Anders",
    booktitle = "Proceedings of the 2018 Conference on Empirical Methods in Natural Language Processing",
    month = oct # "-" # nov,
    year = "2018",
    address = "Brussels, Belgium",
    publisher = "Association for Computational Linguistics",
    url = "https://www.aclweb.org/anthology/D18-1515",
    doi = "10.18653/v1/D18-1515",
    pages = "4810--4815",
}

@inproceedings{de-lhoneux-etal-2018-parameter,
    title = "Parameter sharing between dependency parsers for related languages",
    author = "de Lhoneux, Miryam  and
      Bjerva, Johannes  and
      Augenstein, Isabelle  and
      S{\o}gaard, Anders",
    booktitle = "Proceedings of the 2018 Conference on Empirical Methods in Natural Language Processing",
    month = oct # "-" # nov,
    year = "2018",
    address = "Brussels, Belgium",
    publisher = "Association for Computational Linguistics",
    url = "https://www.aclweb.org/anthology/D18-1543",
    doi = "10.18653/v1/D18-1543",
    pages = "4992--4997",
}

@inproceedings{nyegaard-signori-etal-2018-ku,
    title = "{KU}-{MTL} at {S}em{E}val-2018 Task 1: Multi-task Identification of Affect in Tweets",
    author = "Nyegaard-Signori, Thomas  and
      Helms, Casper Veistrup  and
      Bjerva, Johannes  and
      Augenstein, Isabelle",
    booktitle = "Proceedings of The 12th International Workshop on Semantic Evaluation",
    month = jun,
    year = "2018",
    address = "New Orleans, Louisiana",
    publisher = "Association for Computational Linguistics",
    url = "https://www.aclweb.org/anthology/S18-1058",
    doi = "10.18653/v1/S18-1058",
    pages = "385--389",
}

@inproceedings{bjerva-augenstein-2018-tracking,
    title = "Tracking Typological Traits of Uralic Languages in Distributed Language Representations",
    author = "Bjerva, Johannes  and
      Augenstein, Isabelle",
    booktitle = "Proceedings of the Fourth International Workshop on Computational Linguistics of Uralic Languages",
    month = jan,
    year = "2018",
    address = "Helsinki, Finland",
    publisher = "Association for Computational Linguistics",
    url = "https://www.aclweb.org/anthology/W18-0207",
    doi = "10.18653/v1/W18-0207",
    pages = "76--86",
}

@proceedings{ws-2018-representation,
    title = "Proceedings of The Third Workshop on Representation Learning for {NLP}",
    editor = "Augenstein, Isabelle  and
      Cao, Kris  and
      He, He  and
      Hill, Felix  and
      Gella, Spandana  and
      Kiros, Jamie  and
      Mei, Hongyuan  and
      Misra, Dipendra",
    month = jul,
    year = "2018",
    address = "Melbourne, Australia",
    publisher = "Association for Computational Linguistics",
    url = "https://www.aclweb.org/anthology/W18-3000",
}

@inproceedings{kann-etal-2018-character,
    title = "Character-level Supervision for Low-resource {POS} Tagging",
    author = "Kann, Katharina  and
      Bjerva, Johannes  and
      Augenstein, Isabelle  and
      Plank, Barbara  and
      S{\o}gaard, Anders",
    booktitle = "Proceedings of the Workshop on Deep Learning Approaches for Low-Resource {NLP}",
    month = jul,
    year = "2018",
    address = "Melbourne",
    publisher = "Association for Computational Linguistics",
    url = "https://www.aclweb.org/anthology/W18-3401",
    doi = "10.18653/v1/W18-3401",
    pages = "1--11",
}

@inproceedings{sogaard-etal-2018-nightmare,
    title = "Nightmare at test time: How punctuation prevents parsers from generalizing",
    author = "S{\o}gaard, Anders  and
      de Lhoneux, Miryam  and
      Augenstein, Isabelle",
    booktitle = "Proceedings of the 2018 {EMNLP} Workshop {B}lackbox{NLP}: Analyzing and Interpreting Neural Networks for {NLP}",
    month = nov,
    year = "2018",
    address = "Brussels, Belgium",
    publisher = "Association for Computational Linguistics",
    url = "https://www.aclweb.org/anthology/W18-5404",
    doi = "10.18653/v1/W18-5404",
    pages = "25--29",
}

@inproceedings{bjerva-augenstein-2018-phonology,
    title = "From Phonology to Syntax: Unsupervised Linguistic Typology at Different Levels with Language Embeddings",
    author = "Bjerva, Johannes  and
      Augenstein, Isabelle",
    booktitle = "Proceedings of the 2018 Conference of the North {A}merican Chapter of the Association for Computational Linguistics: Human Language Technologies, Volume 1 (Long Papers)",
    month = jun,
    year = "2018",
    address = "New Orleans, Louisiana",
    publisher = "Association for Computational Linguistics",
    url = "https://www.aclweb.org/anthology/N18-1083",
    doi = "10.18653/v1/N18-1083",
    pages = "907--916",
}

@inproceedings{augenstein-etal-2018-multi,
    title = "Multi-Task Learning of Pairwise Sequence Classification Tasks over Disparate Label Spaces",
    author = "Augenstein, Isabelle  and
      Ruder, Sebastian  and
      S{\o}gaard, Anders",
    booktitle = "Proceedings of the 2018 Conference of the North {A}merican Chapter of the Association for Computational Linguistics: Human Language Technologies, Volume 1 (Long Papers)",
    month = jun,
    year = "2018",
    address = "New Orleans, Louisiana",
    publisher = "Association for Computational Linguistics",
    url = "https://www.aclweb.org/anthology/N18-1172",
    doi = "10.18653/v1/N18-1172",
    pages = "1896--1906",
}

@inproceedings{kementchedjhieva-etal-2018-copenhagen,
    title = "Copenhagen at {C}o{NLL}{--}{SIGMORPHON} 2018: Multilingual Inflection in Context with Explicit Morphosyntactic Decoding",
    author = "Kementchedjhieva, Yova  and
      Bjerva, Johannes  and
      Augenstein, Isabelle",
    booktitle = "Proceedings of the {C}o{NLL}{--}{SIGMORPHON} 2018 Shared Task: Universal Morphological Reinflection",
    month = oct,
    year = "2018",
    address = "Brussels",
    publisher = "Association for Computational Linguistics",
    url = "https://www.aclweb.org/anthology/K18-3011",
    doi = "10.18653/v1/K18-3011",
    pages = "93--98",
}

@inproceedings{weissenborn-etal-2018-jack,
    title = "Jack the Reader {--} A Machine Reading Framework",
    author = {Weissenborn, Dirk  and
      Minervini, Pasquale  and
      Augenstein, Isabelle  and
      Welbl, Johannes  and
      Rockt{\"a}schel, Tim  and
      Bo{\v{s}}njak, Matko  and
      Mitchell, Jeff  and
      Demeester, Thomas  and
      Dettmers, Tim  and
      Stenetorp, Pontus  and
      Riedel, Sebastian},
    booktitle = "Proceedings of {ACL} 2018, System Demonstrations",
    month = jul,
    year = "2018",
    address = "Melbourne, Australia",
    publisher = "Association for Computational Linguistics",
    url = "https://www.aclweb.org/anthology/P18-4005",
    doi = "10.18653/v1/P18-4005",
    pages = "25--30",
}

@inproceedings{collins-etal-2017-supervised,
    title = "A Supervised Approach to Extractive Summarisation of Scientific Papers",
    author = "Collins, Ed  and
      Augenstein, Isabelle  and
      Riedel, Sebastian",
    booktitle = "Proceedings of the 21st Conference on Computational Natural Language Learning ({C}o{NLL} 2017)",
    month = aug,
    year = "2017",
    address = "Vancouver, Canada",
    publisher = "Association for Computational Linguistics",
    url = "https://www.aclweb.org/anthology/K17-1021",
    doi = "10.18653/v1/K17-1021",
    pages = "195--205",
}

@inproceedings{augenstein-sogaard-2017-multi,
    title = "Multi-Task Learning of Keyphrase Boundary Classification",
    author = "Augenstein, Isabelle  and
      S{\o}gaard, Anders",
    booktitle = "Proceedings of the 55th Annual Meeting of the Association for Computational Linguistics (Volume 2: Short Papers)",
    month = jul,
    year = "2017",
    address = "Vancouver, Canada",
    publisher = "Association for Computational Linguistics",
    url = "https://www.aclweb.org/anthology/P17-2054",
    doi = "10.18653/v1/P17-2054",
    pages = "341--346",
}

@inproceedings{kochkina-etal-2017-turing,
    title = "{T}uring at {S}em{E}val-2017 Task 8: Sequential Approach to Rumour Stance Classification with Branch-{LSTM}",
    author = "Kochkina, Elena  and
      Liakata, Maria  and
      Augenstein, Isabelle",
    booktitle = "Proceedings of the 11th International Workshop on Semantic Evaluation ({S}em{E}val-2017)",
    month = aug,
    year = "2017",
    address = "Vancouver, Canada",
    publisher = "Association for Computational Linguistics",
    url = "https://www.aclweb.org/anthology/S17-2083",
    doi = "10.18653/v1/S17-2083",
    pages = "475--480",
}

@inproceedings{augenstein-etal-2017-semeval,
    title = "{S}em{E}val 2017 Task 10: {S}cience{IE} - Extracting Keyphrases and Relations from Scientific Publications",
    author = "Augenstein, Isabelle  and
      Das, Mrinal  and
      Riedel, Sebastian  and
      Vikraman, Lakshmi  and
      McCallum, Andrew",
    booktitle = "Proceedings of the 11th International Workshop on Semantic Evaluation ({S}em{E}val-2017)",
    month = aug,
    year = "2017",
    address = "Vancouver, Canada",
    publisher = "Association for Computational Linguistics",
    url = "https://www.aclweb.org/anthology/S17-2091",
    doi = "10.18653/v1/S17-2091",
    pages = "546--555",
}

@inproceedings{eisner-etal-2016-emoji2vec,
    title = "emoji2vec: Learning Emoji Representations from their Description",
    author = {Eisner, Ben  and
      Rockt{\"a}schel, Tim  and
      Augenstein, Isabelle  and
      Bo{\v{s}}njak, Matko  and
      Riedel, Sebastian},
    booktitle = "Proceedings of The Fourth International Workshop on Natural Language Processing for Social Media",
    month = nov,
    year = "2016",
    address = "Austin, TX, USA",
    publisher = "Association for Computational Linguistics",
    url = "https://www.aclweb.org/anthology/W16-6208",
    doi = "10.18653/v1/W16-6208",
    pages = "48--54",
}

@inproceedings{lendvai-etal-2016-monolingual,
    title = "Monolingual Social Media Datasets for Detecting Contradiction and Entailment",
    author = "Lendvai, Piroska  and
      Augenstein, Isabelle  and
      Bontcheva, Kalina  and
      Declerck, Thierry",
    booktitle = "Proceedings of the Tenth International Conference on Language Resources and Evaluation ({LREC}'16)",
    month = may,
    year = "2016",
    address = "Portoro{\v{z}}, Slovenia",
    publisher = "European Language Resources Association (ELRA)",
    url = "https://www.aclweb.org/anthology/L16-1729",
    pages = "4602--4605",
}

@inproceedings{augenstein-etal-2016-usfd,
    title = "{USFD} at {S}em{E}val-2016 Task 6: Any-Target Stance Detection on {T}witter with Autoencoders",
    author = "Augenstein, Isabelle  and
      Vlachos, Andreas  and
      Bontcheva, Kalina",
    booktitle = "Proceedings of the 10th International Workshop on Semantic Evaluation ({S}em{E}val-2016)",
    month = jun,
    year = "2016",
    address = "San Diego, California",
    publisher = "Association for Computational Linguistics",
    url = "https://www.aclweb.org/anthology/S16-1063",
    doi = "10.18653/v1/S16-1063",
    pages = "389--393",
}

@inproceedings{augenstein-etal-2016-stance,
    title = "Stance Detection with Bidirectional Conditional Encoding",
    author = {Augenstein, Isabelle  and
      Rockt{\"a}schel, Tim  and
      Vlachos, Andreas  and
      Bontcheva, Kalina},
    booktitle = "Proceedings of the 2016 Conference on Empirical Methods in Natural Language Processing",
    month = nov,
    year = "2016",
    address = "Austin, Texas",
    publisher = "Association for Computational Linguistics",
    url = "https://www.aclweb.org/anthology/D16-1084",
    doi = "10.18653/v1/D16-1084",
    pages = "876--885",
}

@inproceedings{spithourakis-etal-2016-numerically,
    title = "Numerically Grounded Language Models for Semantic Error Correction",
    author = "Spithourakis, Georgios  and
      Augenstein, Isabelle  and
      Riedel, Sebastian",
    booktitle = "Proceedings of the 2016 Conference on Empirical Methods in Natural Language Processing",
    month = nov,
    year = "2016",
    address = "Austin, Texas",
    publisher = "Association for Computational Linguistics",
    url = "https://www.aclweb.org/anthology/D16-1101",
    doi = "10.18653/v1/D16-1101",
    pages = "987--992",
}

@inproceedings{derczynski-etal-2015-usfd,
    title = "{USFD}: {T}witter {NER} with Drift Compensation and Linked Data",
    author = "Derczynski, Leon  and
      Augenstein, Isabelle  and
      Bontcheva, Kalina",
    booktitle = "Proceedings of the Workshop on Noisy User-generated Text",
    month = jul,
    year = "2015",
    address = "Beijing, China",
    publisher = "Association for Computational Linguistics",
    url = "https://www.aclweb.org/anthology/W15-4306",
    doi = "10.18653/v1/W15-4306",
    pages = "48--53",
}

@article{journals/corr/abs-2010-01061,
  author = {Rethmeier, Nils and Augenstein, Isabelle},
  ee = {https://arxiv.org/abs/2010.01061},
  journal = {CoRR},
  title = {Long-Tail Zero and Few-Shot Learning via Contrastive Pretraining on and for Small Data.},
  url = {http://dblp.uni-trier.de/db/journals/corr/corr2010.html#abs-2010-01061},
  volume = {abs/2010.01061},
  year = 2020
}

@article{journals/corr/abs-2009-06401,
  author = {Ostrowski, Wojciech and Arora, Arnav and Atanasova, Pepa and Augenstein, Isabelle},
  ee = {https://arxiv.org/abs/2009.06401},
  journal = {CoRR},
  title = {Multi-Hop Fact Checking of Political Claims.},
  url = {http://dblp.uni-trier.de/db/journals/corr/corr2009.html#abs-2009-06401},
  volume = {abs/2009.06401},
  year = 2020
}

@article{journals/corr/abs-2009-06402,
  author = {Allein, Liesbeth and Augenstein, Isabelle and Moens, Marie-Francine},
  ee = {https://arxiv.org/abs/2009.06402},
  journal = {CoRR},
  title = {Time-Aware Evidence Ranking for Fact-Checking.},
  url = {http://dblp.uni-trier.de/db/journals/corr/corr2009.html#abs-2009-06402},
  volume = {abs/2009.06402},
  year = 2020
}

@article{journals/corr/abs-2008-09112,
  author = {Zhao, Wei and Eger, Steffen and Bjerva, Johannes and Augenstein, Isabelle},
  ee = {https://arxiv.org/abs/2008.09112},
  journal = {CoRR},
  title = {Inducing Language-Agnostic Multilingual Representations.},
  url = {http://dblp.uni-trier.de/db/journals/corr/corr2008.html#abs-2008-09112},
  volume = {abs/2008.09112},
  year = 2020
}

@article{journals/corr/abs-2012-05742,
  author={Andreas Nugaard Holm and Barbara Plank and Dustin Wright and Isabelle Augenstein},
  journal = {CoRR},
  title = {Longitudinal Citation Prediction using Temporal Graph Neural Networks.},
  volume = {abs/2012.05742},
  year = 2020
}

@article{journals/corr/abs-2012-05776,
  author={Andrea Lekkas and Peter Schneider-Kamp and Isabelle Augenstein},
  journal = {CoRR},
  title = {Multi-Sense Language Modelling},
  volume = {abs/2012.05776},
  year = 2020
}

@article{waseem2020disembodied,
title={Disembodied Machine Learning: On the Illusion of Objectivity in NLP},
author={Waseem, Zeerak and Lulz, Smarika and Bingel, Joachim and Augenstein, Isabelle},
journal={OpenReview Preprint},
year={2020},
%note={preprint under review},
url={https://openreview.net/forum?id=fkAxTMzy3fs},
month = {Jun.}
}

@inproceedings{10.1145/3366424.3383758,
author = {Debnath, Alok and Pinnaparaju, Nikhil and Shrivastava, Manish and Varma, Vasudeva and Augenstein, Isabelle},
title = {Semantic Textual Similarity of Sentences with Emojis},
year = {2020},
isbn = {9781450370240},
publisher = {Association for Computing Machinery},
address = {New York, NY, USA},
url = {https://doi.org/10.1145/3366424.3383758},
doi = {10.1145/3366424.3383758},
booktitle = {Companion Proceedings of the Web Conference 2020},
pages = {426–430},
numpages = {5},
location = {Taipei, Taiwan},
series = {WWW '20}
}

@article{Bjerva_Kouw_Augenstein_2020, 
title={Back to the Future – Temporal Adaptation of Text Representations}, 
volume={34}, url={https://ojs.aaai.org/index.php/AAAI/article/view/6240}, 
DOI={10.1609/aaai.v34i05.6240}, 
abstractNote={&lt;p&gt;Language evolves over time in many ways relevant to natural language processing tasks. For example, recent occurrences of tokens ’BERT’ and ’ELMO’ in publications refer to neural network architectures rather than persons. This type of temporal signal is typically overlooked, but is important if one aims to deploy a machine learning model over an extended period of time. In particular, language evolution causes data drift between time-steps in sequential decision-making tasks. Examples of such tasks include prediction of paper acceptance for yearly conferences (regular intervals) or author stance prediction for rumours on Twitter (irregular intervals). Inspired by successes in computer vision, we tackle data drift by sequentially aligning learned representations. We evaluate on three challenging tasks varying in terms of time-scales, linguistic units, and domains. These tasks show our method outperforming several strong baselines, including using all available data. We argue that, due to its low computational expense, sequential alignment is a practical solution to dealing with language evolution.&lt;/p&gt;}, 
number={05}, 
journal={Proceedings of the AAAI Conference on Artificial Intelligence}, 
author={Bjerva, Johannes and Kouw, Wouter and Augenstein, Isabelle}, 
year={2020}, 
month={Apr.}, 
pages={7440-7447} 
}

@article{journals/corr/abs-1911-08782,
  author = {Bruyne, Luna De and Atanasova, Pepa and Augenstein, Isabelle},
  ee = {http://arxiv.org/abs/1911.08782},
  journal = {CoRR},
  title = {Joint Emotion Label Space Modelling for Affect Lexica.},
  url = {http://dblp.uni-trier.de/db/journals/corr/corr1911.html#abs-1911-08782},
  volume = {abs/1911.08782},
  year = 2019
}

@inproceedings{gonzalez2019retrieval,
  title={Retrieval-based Goal-Oriented Dialogue Generation},
  author={Gonzalez, Ana Valeria and Augenstein, Isabelle and S{\o}gaard, Anders},
  booktitle={The 3rd NeurIPS Workshop on Conversational AI},
  url={http://alborz-geramifard.com/workshops/neurips19-Conversational-AI/Papers/34.pdf},
  month = dec,
  address = "Vancouver, Canada",
  year={2019}
}

@inproceedings{bingel2019domain,
  title={Domain Transfer in Dialogue Systems without Turn-Level Supervision},
  author={Bingel, Joachim and Hansen, Victor Petr{\'e}n Bach and Gonzalez, Ana Valeria and Budzianowski, Pavel and Augenstein, Isabelle and S{\o}gaard, Anders},
  booktitle={The 3rd NeurIPS Workshop on Conversational AI},
  url={http://alborz-geramifard.com/workshops/neurips19-Conversational-AI/Papers/6.pdf},
  month = dec,
  address = "Vancouver, Canada",
  year={2019}
}

@inproceedings{conf/aaai/RuderBAS19,
  author = {Ruder, Sebastian and Bingel, Joachim and Augenstein, Isabelle and S{\o}gaard, Anders},
  booktitle = {AAAI},
  crossref = {conf/aaai/2019},
  ee = {https://doi.org/10.1609/aaai.v33i01.33014822},
  isbn = {978-1-57735-809-1},
  pages = {4822-4829},
  publisher = {AAAI Press},
  title = {Latent Multi-Task Architecture Learning.},
  url = {http://dblp.uni-trier.de/db/conf/aaai/aaai2019.html#RuderBAS19},
  year = 2019
}

@article{journals/ipm/ZubiagaKLPLBCA18,
title = "Discourse-aware rumour stance classification in social media using sequential classifiers",
journal = "Information Processing \& Management",
volume = "54",
number = "2",
pages = "273 - 290",
year = "2018",
issn = "0306-4573",
doi = "https://doi.org/10.1016/j.ipm.2017.11.009",
url = "http://www.sciencedirect.com/science/article/pii/S0306457317303746",
author = "Arkaitz Zubiaga and Elena Kochkina and Maria Liakata and Rob Procter and Michal Lukasik and Kalina Bontcheva and Trevor Cohn and Isabelle Augenstein"
}

@article{journals/corr/RiedelASR17,
  author = {Riedel, Benjamin and Augenstein, Isabelle and Spithourakis, Georgios P. and Riedel, Sebastian},
  ee = {http://arxiv.org/abs/1707.03264},
  journal = {CoRR},
  title = {A simple but tough-to-beat baseline for the Fake News Challenge stance detection task.},
  url = {http://dblp.uni-trier.de/db/journals/corr/corr1707.html#RiedelASR17},
  volume = {abs/1707.03264},
  year = 2017
}

@article{journals/semweb/ZhangGBAC17,
  author = {Zhang, Ziqi and Gentile, Anna Lisa and Blomqvist, Eva and Augenstein, Isabelle and Ciravegna, Fabio},
  ee = {https://doi.org/10.3233/SW-150193},
  journal = {Semantic Web},
  number = 2,
  pages = {197-223},
  title = {An Unsupervised Data-driven Method to Discover Equivalent Relations in Large Linked Datasets},
  url = {http://www.semantic-web-journal.net/content/unsupervised-data-driven-method-discover-equivalent-relations-large-linked-datasets},
  volume = 8,
  year = 2017
}

@book{series/synthesis/2016Maynard,
  author = {Maynard, Diana and Bontcheva, Kalina and Augenstein, Isabelle},
  booktitle = {Natural Language Processing for the Semantic Web},
  ee = {https://www.wikidata.org/entity/Q64392146},
  publisher = {Morgan \& Claypool Publishers},
  series = {Synthesis Lectures on the Semantic Web: Theory and Technology},
  title = {Natural Language Processing for the Semantic Web},
  url={https://www.morganclaypool.com/doi/abs/10.2200/S00741ED1V01Y201611WBE015},
  month={Dec.},
  year = 2016
}

@inproceedings{yu-etal-2020-coupled,
    title = "Coupled Hierarchical Transformer for Stance-Aware Rumor Verification in Social Media Conversations",
    author = "Yu, Jianfei  and
      Jiang, Jing  and
      Khoo, Ling Min Serena  and
      Chieu, Hai Leong  and
      Xia, Rui",
    booktitle = "Proceedings of the 2020 Conference on Empirical Methods in Natural Language Processing (EMNLP)",
    month = nov,
    year = "2020",
    address = "Online",
    publisher = "Association for Computational Linguistics",
    url = "https://www.aclweb.org/anthology/2020.emnlp-main.108",
    doi = "10.18653/v1/2020.emnlp-main.108",
    pages = "1392--1401",
}

@inproceedings{allaway-mckeown-2020-zero,
    title = "{Z}ero-{S}hot {S}tance {D}etection: {A} {D}ataset and {M}odel using {G}eneralized {T}opic {R}epresentations",
    author = "Allaway, Emily  and
      McKeown, Kathleen",
    booktitle = "Proceedings of the 2020 Conference on Empirical Methods in Natural Language Processing (EMNLP)",
    month = nov,
    year = "2020",
    address = "Online",
    publisher = "Association for Computational Linguistics",
    url = "https://www.aclweb.org/anthology/2020.emnlp-main.717",
    doi = "10.18653/v1/2020.emnlp-main.717",
    pages = "8913--8931",
}

@inproceedings{xia-etal-2020-state,
    title = "A State-independent and Time-evolving Network for Early Rumor Detection in Social Media",
    author = "Xia, Rui  and
      Xuan, Kaizhou  and
      Yu, Jianfei",
    booktitle = "Proceedings of the 2020 Conference on Empirical Methods in Natural Language Processing (EMNLP)",
    month = nov,
    year = "2020",
    address = "Online",
    publisher = "Association for Computational Linguistics",
    url = "https://www.aclweb.org/anthology/2020.emnlp-main.727",
    doi = "10.18653/v1/2020.emnlp-main.727",
    pages = "9042--9051",
}

@inproceedings{ma-gao-2020-debunking,
    title = "Debunking Rumors on {T}witter with Tree Transformer",
    author = "Ma, Jing  and
      Gao, Wei",
    booktitle = "Proceedings of the 28th International Conference on Computational Linguistics",
    month = dec,
    year = "2020",
    address = "Barcelona, Spain (Online)",
    publisher = "International Committee on Computational Linguistics",
    url = "https://www.aclweb.org/anthology/2020.coling-main.476",
    pages = "5455--5466",
}

@inproceedings{wu-etal-2020-dtca,
    title = "{DTCA}: Decision Tree-based Co-Attention Networks for Explainable Claim Verification",
    author = "Wu, Lianwei  and
      Rao, Yuan  and
      Zhao, Yongqiang  and
      Liang, Hao  and
      Nazir, Ambreen",
    booktitle = "Proceedings of the 58th Annual Meeting of the Association for Computational Linguistics",
    month = jul,
    year = "2020",
    address = "Online",
    publisher = "Association for Computational Linguistics",
    url = "https://www.aclweb.org/anthology/2020.acl-main.97",
    doi = "10.18653/v1/2020.acl-main.97",
    pages = "1024--1035",
}

@inproceedings{conforti-etal-2020-will,
    title = "Will-They-Won{'}t-They: A Very Large Dataset for Stance Detection on {T}witter",
    author = "Conforti, Costanza  and
      Berndt, Jakob  and
      Pilehvar, Mohammad Taher  and
      Giannitsarou, Chryssi  and
      Toxvaerd, Flavio  and
      Collier, Nigel",
    booktitle = "Proceedings of the 58th Annual Meeting of the Association for Computational Linguistics",
    month = jul,
    year = "2020",
    address = "Online",
    publisher = "Association for Computational Linguistics",
    url = "https://www.aclweb.org/anthology/2020.acl-main.157",
    doi = "10.18653/v1/2020.acl-main.157",
    pages = "1715--1724",
}

@inproceedings{zhang-etal-2020-enhancing-cross,
    title = "Enhancing Cross-target Stance Detection with Transferable Semantic-Emotion Knowledge",
    author = "Zhang, Bowen  and
      Yang, Min  and
      Li, Xutao  and
      Ye, Yunming  and
      Xu, Xiaofei  and
      Dai, Kuai",
    booktitle = "Proceedings of the 58th Annual Meeting of the Association for Computational Linguistics",
    month = jul,
    year = "2020",
    address = "Online",
    publisher = "Association for Computational Linguistics",
    url = "https://www.aclweb.org/anthology/2020.acl-main.291",
    doi = "10.18653/v1/2020.acl-main.291",
    pages = "3188--3197",
}

@inproceedings{zhang-etal-2020-said,
    title = "{``}Who said it, and Why?{''} Provenance for Natural Language Claims",
    author = "Zhang, Yi  and
      Ives, Zachary  and
      Roth, Dan",
    booktitle = "Proceedings of the 58th Annual Meeting of the Association for Computational Linguistics",
    month = jul,
    year = "2020",
    address = "Online",
    publisher = "Association for Computational Linguistics",
    url = "https://www.aclweb.org/anthology/2020.acl-main.406",
    doi = "10.18653/v1/2020.acl-main.406",
    pages = "4416--4426",
}

@inproceedings{sirrianni-etal-2020-agreement,
    title = "Agreement Prediction of Arguments in Cyber Argumentation for Detecting Stance Polarity and Intensity",
    author = "Sirrianni, Joseph  and
      Liu, Xiaoqing  and
      Adams, Douglas",
    booktitle = "Proceedings of the 58th Annual Meeting of the Association for Computational Linguistics",
    month = jul,
    year = "2020",
    address = "Online",
    publisher = "Association for Computational Linguistics",
    url = "https://www.aclweb.org/anthology/2020.acl-main.509",
    doi = "10.18653/v1/2020.acl-main.509",
    pages = "5746--5758",
}

@inproceedings{kochkina-liakata-2020-estimating,
    title = "Estimating predictive uncertainty for rumour verification models",
    author = "Kochkina, Elena  and
      Liakata, Maria",
    booktitle = "Proceedings of the 58th Annual Meeting of the Association for Computational Linguistics",
    month = jul,
    year = "2020",
    address = "Online",
    publisher = "Association for Computational Linguistics",
    url = "https://www.aclweb.org/anthology/2020.acl-main.623",
    doi = "10.18653/v1/2020.acl-main.623",
    pages = "6964--6981",
}

@inproceedings{scarton-etal-2020-measuring,
    title = "Measuring What Counts: The Case of Rumour Stance Classification",
    author = "Scarton, Carolina  and
      Silva, Diego  and
      Bontcheva, Kalina",
    booktitle = "Proceedings of the 1st Conference of the Asia-Pacific Chapter of the Association for Computational Linguistics and the 10th International Joint Conference on Natural Language Processing",
    month = dec,
    year = "2020",
    address = "Suzhou, China",
    publisher = "Association for Computational Linguistics",
    url = "https://www.aclweb.org/anthology/2020.aacl-main.92",
    pages = "925--932",
}

@inproceedings{li-etal-2019-rumor,
    title = "Rumor Detection by Exploiting User Credibility Information, Attention and Multi-task Learning",
    author = "Li, Quanzhi  and
      Zhang, Qiong  and
      Si, Luo",
    booktitle = "Proceedings of the 57th Annual Meeting of the Association for Computational Linguistics",
    month = jul,
    year = "2019",
    address = "Florence, Italy",
    publisher = "Association for Computational Linguistics",
    url = "https://www.aclweb.org/anthology/P19-1113",
    doi = "10.18653/v1/P19-1113",
    pages = "1173--1179",
}

@inproceedings{kumar-carley-2019-tree,
    title = "Tree {LSTM}s with Convolution Units to Predict Stance and Rumor Veracity in Social Media Conversations",
    author = "Kumar, Sumeet  and
      Carley, Kathleen",
    booktitle = "Proceedings of the 57th Annual Meeting of the Association for Computational Linguistics",
    month = jul,
    year = "2019",
    address = "Florence, Italy",
    publisher = "Association for Computational Linguistics",
    url = "https://www.aclweb.org/anthology/P19-1498",
    doi = "10.18653/v1/P19-1498",
    pages = "5047--5058",
}

@inproceedings{zhou-etal-2019-early,
    title = "Early Rumour Detection",
    author = "Zhou, Kaimin  and
      Shu, Chang  and
      Li, Binyang  and
      Lau, Jey Han",
    booktitle = "Proceedings of the 2019 Conference of the North {A}merican Chapter of the Association for Computational Linguistics: Human Language Technologies, Volume 1 (Long and Short Papers)",
    month = jun,
    year = "2019",
    address = "Minneapolis, Minnesota",
    publisher = "Association for Computational Linguistics",
    url = "https://www.aclweb.org/anthology/N19-1163",
    doi = "10.18653/v1/N19-1163",
    pages = "1614--1623",
}

@inproceedings{wei-etal-2019-modeling,
    title = "Modeling Conversation Structure and Temporal Dynamics for Jointly Predicting Rumor Stance and Veracity",
    author = "Wei, Penghui  and
      Xu, Nan  and
      Mao, Wenji",
    booktitle = "Proceedings of the 2019 Conference on Empirical Methods in Natural Language Processing and the 9th International Joint Conference on Natural Language Processing (EMNLP-IJCNLP)",
    month = nov,
    year = "2019",
    address = "Hong Kong, China",
    publisher = "Association for Computational Linguistics",
    url = "https://www.aclweb.org/anthology/D19-1485",
    doi = "10.18653/v1/D19-1485",
    pages = "4787--4798",
}

@inproceedings{ferreira-vlachos-2019-incorporating,
    title = "Incorporating Label Dependencies in Multilabel Stance Detection",
    author = "Ferreira, William  and
      Vlachos, Andreas",
    booktitle = "Proceedings of the 2019 Conference on Empirical Methods in Natural Language Processing and the 9th International Joint Conference on Natural Language Processing (EMNLP-IJCNLP)",
    month = nov,
    year = "2019",
    address = "Hong Kong, China",
    publisher = "Association for Computational Linguistics",
    url = "https://www.aclweb.org/anthology/D19-1665",
    doi = "10.18653/v1/D19-1665",
    pages = "6350--6354",
}

@inproceedings{li-caragea-2019-multi,
    title = "Multi-Task Stance Detection with Sentiment and Stance Lexicons",
    author = "Li, Yingjie  and
      Caragea, Cornelia",
    booktitle = "Proceedings of the 2019 Conference on Empirical Methods in Natural Language Processing and the 9th International Joint Conference on Natural Language Processing (EMNLP-IJCNLP)",
    month = nov,
    year = "2019",
    address = "Hong Kong, China",
    publisher = "Association for Computational Linguistics",
    url = "https://www.aclweb.org/anthology/D19-1657",
    doi = "10.18653/v1/D19-1657",
    pages = "6299--6305",
}

@inproceedings{popat-etal-2019-stancy,
    title = "{STANCY}: Stance Classification Based on Consistency Cues",
    author = "Popat, Kashyap  and
      Mukherjee, Subhabrata  and
      Yates, Andrew  and
      Weikum, Gerhard",
    booktitle = "Proceedings of the 2019 Conference on Empirical Methods in Natural Language Processing and the 9th International Joint Conference on Natural Language Processing (EMNLP-IJCNLP)",
    month = nov,
    year = "2019",
    address = "Hong Kong, China",
    publisher = "Association for Computational Linguistics",
    url = "https://www.aclweb.org/anthology/D19-1675",
    doi = "10.18653/v1/D19-1675",
    pages = "6413--6418",
}

@article{yin-schutze-2018-attentive,
    title = "Attentive Convolution: Equipping {CNN}s with {RNN}-style Attention Mechanisms",
    author = {Yin, Wenpeng  and
      Sch{\"u}tze, Hinrich},
    journal = "Transactions of the Association for Computational Linguistics",
    volume = "6",
    year = "2018",
    url = "https://www.aclweb.org/anthology/Q18-1047",
    doi = "10.1162/tacl_a_00249",
    pages = "687--702",
}

@inproceedings{xu-etal-2018-cross,
    title = "Cross-Target Stance Classification with Self-Attention Networks",
    author = "Xu, Chang  and
      Paris, C{\'e}cile  and
      Nepal, Surya  and
      Sparks, Ross",
    booktitle = "Proceedings of the 56th Annual Meeting of the Association for Computational Linguistics (Volume 2: Short Papers)",
    month = jul,
    year = "2018",
    address = "Melbourne, Australia",
    publisher = "Association for Computational Linguistics",
    url = "https://www.aclweb.org/anthology/P18-2123",
    doi = "10.18653/v1/P18-2123",
    pages = "778--783",
}

@inproceedings{ma-etal-2018-rumor,
    title = "Rumor Detection on {T}witter with Tree-structured Recursive Neural Networks",
    author = "Ma, Jing  and
      Gao, Wei  and
      Wong, Kam-Fai",
    booktitle = "Proceedings of the 56th Annual Meeting of the Association for Computational Linguistics (Volume 1: Long Papers)",
    month = jul,
    year = "2018",
    address = "Melbourne, Australia",
    publisher = "Association for Computational Linguistics",
    url = "https://www.aclweb.org/anthology/P18-1184",
    doi = "10.18653/v1/P18-1184",
    pages = "1980--1989",
}

@inproceedings{wen-etal-2018-cross,
    title = "Cross-Lingual Cross-Platform Rumor Verification Pivoting on Multimedia Content",
    author = "Wen, Weiming  and
      Su, Songwen  and
      Yu, Zhou",
    booktitle = "Proceedings of the 2018 Conference on Empirical Methods in Natural Language Processing",
    month = oct # "-" # nov,
    year = "2018",
    address = "Brussels, Belgium",
    publisher = "Association for Computational Linguistics",
    url = "https://www.aclweb.org/anthology/D18-1385",
    doi = "10.18653/v1/D18-1385",
    pages = "3487--3496",
}

@inproceedings{field-etal-2018-framing,
    title = "Framing and Agenda-setting in {R}ussian News: a Computational Analysis of Intricate Political Strategies",
    author = "Field, Anjalie  and
      Kliger, Doron  and
      Wintner, Shuly  and
      Pan, Jennifer  and
      Jurafsky, Dan  and
      Tsvetkov, Yulia",
    booktitle = "Proceedings of the 2018 Conference on Empirical Methods in Natural Language Processing",
    month = oct # "-" # nov,
    year = "2018",
    address = "Brussels, Belgium",
    publisher = "Association for Computational Linguistics",
    url = "https://www.aclweb.org/anthology/D18-1393",
    doi = "10.18653/v1/D18-1393",
    pages = "3570--3580",
}

@inproceedings{hanselowski-etal-2018-retrospective,
    title = "A Retrospective Analysis of the Fake News Challenge Stance-Detection Task",
    author = "Hanselowski, Andreas  and
      PVS, Avinesh  and
      Schiller, Benjamin  and
      Caspelherr, Felix  and
      Chaudhuri, Debanjan  and
      Meyer, Christian M.  and
      Gurevych, Iryna",
    booktitle = "Proceedings of the 27th International Conference on Computational Linguistics",
    month = aug,
    year = "2018",
    address = "Santa Fe, New Mexico, USA",
    publisher = "Association for Computational Linguistics",
    url = "https://www.aclweb.org/anthology/C18-1158",
    pages = "1859--1874",
}

@inproceedings{sun-etal-2018-stance,
    title = "Stance Detection with Hierarchical Attention Network",
    author = "Sun, Qingying  and
      Wang, Zhongqing  and
      Zhu, Qiaoming  and
      Zhou, Guodong",
    booktitle = "Proceedings of the 27th International Conference on Computational Linguistics",
    month = aug,
    year = "2018",
    address = "Santa Fe, New Mexico, USA",
    publisher = "Association for Computational Linguistics",
    url = "https://www.aclweb.org/anthology/C18-1203",
    pages = "2399--2409",
}

@inproceedings{sasaki-etal-2018-predicting,
    title = "Predicting Stances from Social Media Posts using Factorization Machines",
    author = "Sasaki, Akira  and
      Hanawa, Kazuaki  and
      Okazaki, Naoaki  and
      Inui, Kentaro",
    booktitle = "Proceedings of the 27th International Conference on Computational Linguistics",
    month = aug,
    year = "2018",
    address = "Santa Fe, New Mexico, USA",
    publisher = "Association for Computational Linguistics",
    url = "https://www.aclweb.org/anthology/C18-1286",
    pages = "3381--3390",
}

@inproceedings{kochkina-etal-2018-one,
    title = "All-in-one: Multi-task Learning for Rumour Verification",
    author = "Kochkina, Elena  and
      Liakata, Maria  and
      Zubiaga, Arkaitz",
    booktitle = "Proceedings of the 27th International Conference on Computational Linguistics",
    month = aug,
    year = "2018",
    address = "Santa Fe, New Mexico, USA",
    publisher = "Association for Computational Linguistics",
    url = "https://www.aclweb.org/anthology/C18-1288",
    pages = "3402--3413",
}

@inproceedings{sasaki-etal-2017-topics,
    title = "Other Topics You May Also Agree or Disagree: Modeling Inter-Topic Preferences using Tweets and Matrix Factorization",
    author = "Sasaki, Akira  and
      Hanawa, Kazuaki  and
      Okazaki, Naoaki  and
      Inui, Kentaro",
    booktitle = "Proceedings of the 55th Annual Meeting of the Association for Computational Linguistics (Volume 1: Long Papers)",
    month = jul,
    year = "2017",
    address = "Vancouver, Canada",
    publisher = "Association for Computational Linguistics",
    url = "https://www.aclweb.org/anthology/P17-1037",
    doi = "10.18653/v1/P17-1037",
    pages = "398--408",
}

@inproceedings{ma-etal-2017-detect,
    title = "Detect Rumors in Microblog Posts Using Propagation Structure via Kernel Learning",
    author = "Ma, Jing  and
      Gao, Wei  and
      Wong, Kam-Fai",
    booktitle = "Proceedings of the 55th Annual Meeting of the Association for Computational Linguistics (Volume 1: Long Papers)",
    month = jul,
    year = "2017",
    address = "Vancouver, Canada",
    publisher = "Association for Computational Linguistics",
    url = "https://www.aclweb.org/anthology/P17-1066",
    doi = "10.18653/v1/P17-1066",
    pages = "708--717",
}

@inproceedings{sobhani-etal-2017-dataset,
    title = "A Dataset for Multi-Target Stance Detection",
    author = "Sobhani, Parinaz  and
      Inkpen, Diana  and
      Zhu, Xiaodan",
    booktitle = "Proceedings of the 15th Conference of the {E}uropean Chapter of the Association for Computational Linguistics: Volume 2, Short Papers",
    month = apr,
    year = "2017",
    address = "Valencia, Spain",
    publisher = "Association for Computational Linguistics",
    url = "https://www.aclweb.org/anthology/E17-2088",
    pages = "551--557",
}

@inproceedings{joseph-etal-2017-constance,
    title = "{C}on{S}tance: Modeling Annotation Contexts to Improve Stance Classification",
    author = "Joseph, Kenneth  and
      Friedland, Lisa  and
      Hobbs, William  and
      Lazer, David  and
      Tsur, Oren",
    booktitle = "Proceedings of the 2017 Conference on Empirical Methods in Natural Language Processing",
    month = sep,
    year = "2017",
    address = "Copenhagen, Denmark",
    publisher = "Association for Computational Linguistics",
    url = "https://www.aclweb.org/anthology/D17-1116",
    doi = "10.18653/v1/D17-1116",
    pages = "1115--1124",
}

@article{Goldberg_90,
  author = {Goldberg, Lewis R.},
  data = {available},
  journal = {Journal of Personality and Social Psychologs},
  number = 6,
  online = {Goldberg.Big-Five.pdf},
  pages = {1216--12297},
  title = {An Alternative {"}Description of Personality{"}: {The} Big-Five Factor Structure},
  volume = 59,
  year = 1990
}

@inproceedings{lynn-etal-2017-human,
    title = "Human Centered {NLP} with User-Factor Adaptation",
    author = "Lynn, Veronica  and
      Son, Youngseo  and
      Kulkarni, Vivek  and
      Balasubramanian, Niranjan  and
      Schwartz, H. Andrew",
    booktitle = "Proceedings of the 2017 Conference on Empirical Methods in Natural Language Processing",
    month = sep,
    year = "2017",
    address = "Copenhagen, Denmark",
    publisher = "Association for Computational Linguistics",
    url = "https://www.aclweb.org/anthology/D17-1119",
    doi = "10.18653/v1/D17-1119",
    pages = "1146--1155",
}

@inproceedings{sobhani-etal-2016-detecting,
    title = "Detecting Stance in Tweets And Analyzing its Interaction with Sentiment",
    author = "Sobhani, Parinaz  and
      Mohammad, Saif  and
      Kiritchenko, Svetlana",
    booktitle = "Proceedings of the Fifth Joint Conference on Lexical and Computational Semantics",
    month = aug,
    year = "2016",
    address = "Berlin, Germany",
    publisher = "Association for Computational Linguistics",
    url = "https://www.aclweb.org/anthology/S16-2021",
    doi = "10.18653/v1/S16-2021",
    pages = "159--169",
}

@inproceedings{ebrahimi-etal-2016-weakly,
    title = "Weakly Supervised Tweet Stance Classification by Relational Bootstrapping",
    author = "Ebrahimi, Javid  and
      Dou, Dejing  and
      Lowd, Daniel",
    booktitle = "Proceedings of the 2016 Conference on Empirical Methods in Natural Language Processing",
    month = nov,
    year = "2016",
    address = "Austin, Texas",
    publisher = "Association for Computational Linguistics",
    url = "https://www.aclweb.org/anthology/D16-1105",
    doi = "10.18653/v1/D16-1105",
    pages = "1012--1017",
}

@inproceedings{ebrahimi-etal-2016-joint,
    title = "A Joint Sentiment-Target-Stance Model for Stance Classification in Tweets",
    author = "Ebrahimi, Javid  and
      Dou, Dejing  and
      Lowd, Daniel",
    booktitle = "Proceedings of {COLING} 2016, the 26th International Conference on Computational Linguistics: Technical Papers",
    month = dec,
    year = "2016",
    address = "Osaka, Japan",
    publisher = "The COLING 2016 Organizing Committee",
    url = "https://www.aclweb.org/anthology/C16-1250",
    pages = "2656--2665",
}

@inproceedings{sridhar-etal-2014-collective,
    title = "Collective Stance Classification of Posts in Online Debate Forums",
    author = "Sridhar, Dhanya  and
      Getoor, Lise  and
      Walker, Marilyn",
    booktitle = "Proceedings of the Joint Workshop on Social Dynamics and Personal Attributes in Social Media",
    month = jun,
    year = "2014",
    address = "Baltimore, Maryland",
    publisher = "Association for Computational Linguistics",
    url = "https://www.aclweb.org/anthology/W14-2715",
    doi = "10.3115/v1/W14-2715",
    pages = "109--117",
}

@inproceedings{ranade-etal-2013-stance,
    title = "Stance Classification in Online Debates by Recognizing Users{'} Intentions",
    author = "Ranade, Sarvesh  and
      Sangal, Rajeev  and
      Mamidi, Radhika",
    booktitle = "Proceedings of the {SIGDIAL} 2013 Conference",
    month = aug,
    year = "2013",
    address = "Metz, France",
    publisher = "Association for Computational Linguistics",
    url = "https://www.aclweb.org/anthology/W13-4008",
    pages = "61--69",
}

@inproceedings{hasan-ng-2013-frame,
    title = "Frame Semantics for Stance Classification",
    author = "Hasan, Kazi Saidul  and
      Ng, Vincent",
    booktitle = "Proceedings of the Seventeenth Conference on Computational Natural Language Learning",
    month = aug,
    year = "2013",
    address = "Sofia, Bulgaria",
    publisher = "Association for Computational Linguistics",
    url = "https://www.aclweb.org/anthology/W13-3514",
    pages = "124--132",
}

@inproceedings{anand-etal-2011-cats,
    title = "Cats Rule and Dogs Drool!: Classifying Stance in Online Debate",
    author = "Anand, Pranav  and
      Walker, Marilyn  and
      Abbott, Rob  and
      Fox Tree, Jean E.  and
      Bowmani, Robeson  and
      Minor, Michael",
    booktitle = "Proceedings of the 2nd Workshop on Computational Approaches to Subjectivity and Sentiment Analysis ({WASSA} 2.011)",
    month = jun,
    year = "2011",
    address = "Portland, Oregon",
    publisher = "Association for Computational Linguistics",
    url = "https://www.aclweb.org/anthology/W11-1701",
    pages = "1--9",
}

@inproceedings{levy-etal-2014-context,
    title = "Context Dependent Claim Detection",
    author = "Levy, Ran  and
      Bilu, Yonatan  and
      Hershcovich, Daniel  and
      Aharoni, Ehud  and
      Slonim, Noam",
    booktitle = "Proceedings of {COLING} 2014, the 25th International Conference on Computational Linguistics: Technical Papers",
    month = aug,
    year = "2014",
    address = "Dublin, Ireland",
    publisher = "Dublin City University and Association for Computational Linguistics",
    url = "https://www.aclweb.org/anthology/C14-1141",
    pages = "1489--1500",
}

@inproceedings{mishra-etal-2020-generating,
    title = "Generating Fact Checking Summaries for Web Claims",
    author = "Mishra, Rahul  and
      Gupta, Dhruv  and
      Leippold, Markus",
    booktitle = "Proceedings of the Sixth Workshop on Noisy User-generated Text (W-NUT 2020)",
    month = nov,
    year = "2020",
    address = "Online",
    publisher = "Association for Computational Linguistics",
    url = "https://www.aclweb.org/anthology/2020.wnut-1.12",
    doi = "10.18653/v1/2020.wnut-1.12",
    pages = "81--90",
}

@inproceedings{medina-serrano-etal-2020-nlp,
    title = "{NLP}-based Feature Extraction for the Detection of {COVID}-19 Misinformation Videos on {Y}ou{T}ube",
    author = "Medina Serrano, Juan Carlos  and
      Papakyriakopoulos, Orestis  and
      Hegelich, Simon",
    booktitle = "Proceedings of the 1st Workshop on {NLP} for {COVID-19} at {ACL} 2020",
    month = jul,
    year = "2020",
    address = "Online",
    publisher = "Association for Computational Linguistics",
    url = "https://www.aclweb.org/anthology/2020.nlpcovid19-acl.17",
}

@inproceedings{hossain-etal-2020-covidlies,
    title = "{COVIDL}ies: Detecting {COVID}-19 Misinformation on Social Media",
    author = "Hossain, Tamanna  and
      Logan IV, Robert L.  and
      Ugarte, Arjuna  and
      Matsubara, Yoshitomo  and
      Young, Sean  and
      Singh, Sameer",
    booktitle = "Proceedings of the 1st Workshop on {NLP} for {COVID}-19 (Part 2) at {EMNLP} 2020",
    month = dec,
    year = "2020",
    address = "Online",
    publisher = "Association for Computational Linguistics",
    url = "https://www.aclweb.org/anthology/2020.nlpcovid19-2.11",
    doi = "10.18653/v1/2020.nlpcovid19-2.11",
}

@inproceedings{nakamura-etal-2020-fakeddit,
    title = "{F}akeddit: A New Multimodal Benchmark Dataset for Fine-grained Fake News Detection",
    author = "Nakamura, Kai  and
      Levy, Sharon  and
      Wang, William Yang",
    booktitle = "Proceedings of the 12th Language Resources and Evaluation Conference",
    month = may,
    year = "2020",
    address = "Marseille, France",
    publisher = "European Language Resources Association",
    url = "https://www.aclweb.org/anthology/2020.lrec-1.755",
    pages = "6149--6157",
    language = "English",
    ISBN = "979-10-95546-34-4",
}

@inproceedings{liu-etal-2020-adapting-open,
    title = "Adapting Open Domain Fact Extraction and Verification to {COVID}-{FACT} through In-Domain Language Modeling",
    author = "Liu, Zhenghao  and
      Xiong, Chenyan  and
      Dai, Zhuyun  and
      Sun, Si  and
      Sun, Maosong  and
      Liu, Zhiyuan",
    booktitle = "Findings of the Association for Computational Linguistics: EMNLP 2020",
    month = nov,
    year = "2020",
    address = "Online",
    publisher = "Association for Computational Linguistics",
    url = "https://www.aclweb.org/anthology/2020.findings-emnlp.216",
    doi = "10.18653/v1/2020.findings-emnlp.216",
    pages = "2395--2400",
}

@inproceedings{tan-etal-2020-detecting,
    title = "Detecting Cross-Modal Inconsistency to Defend Against Neural Fake News",
    author = "Tan, Reuben  and
      Plummer, Bryan  and
      Saenko, Kate",
    booktitle = "Proceedings of the 2020 Conference on Empirical Methods in Natural Language Processing (EMNLP)",
    month = nov,
    year = "2020",
    address = "Online",
    publisher = "Association for Computational Linguistics",
    url = "https://www.aclweb.org/anthology/2020.emnlp-main.163",
    doi = "10.18653/v1/2020.emnlp-main.163",
    pages = "2081--2106",
}

@inproceedings{fan-etal-2020-generating,
    title = "Generating Fact Checking Briefs",
    author = "Fan, Angela  and
      Piktus, Aleksandra  and
      Petroni, Fabio  and
      Wenzek, Guillaume  and
      Saeidi, Marzieh  and
      Vlachos, Andreas  and
      Bordes, Antoine  and
      Riedel, Sebastian",
    booktitle = "Proceedings of the 2020 Conference on Empirical Methods in Natural Language Processing (EMNLP)",
    month = nov,
    year = "2020",
    address = "Online",
    publisher = "Association for Computational Linguistics",
    url = "https://www.aclweb.org/anthology/2020.emnlp-main.580",
    doi = "10.18653/v1/2020.emnlp-main.580",
    pages = "7147--7161",
}

@inproceedings{wadden-etal-2020-fact,
    title = "Fact or Fiction: Verifying Scientific Claims",
    author = "Wadden, David  and
      Lin, Shanchuan  and
      Lo, Kyle  and
      Wang, Lucy Lu  and
      van Zuylen, Madeleine  and
      Cohan, Arman  and
      Hajishirzi, Hannaneh",
    booktitle = "Proceedings of the 2020 Conference on Empirical Methods in Natural Language Processing (EMNLP)",
    month = nov,
    year = "2020",
    address = "Online",
    publisher = "Association for Computational Linguistics",
    url = "https://www.aclweb.org/anthology/2020.emnlp-main.609",
    doi = "10.18653/v1/2020.emnlp-main.609",
    pages = "7534--7550",
}

@inproceedings{vo-lee-2020-facts,
    title = "Where Are the Facts? Searching for Fact-checked Information to Alleviate the Spread of Fake News",
    author = "Vo, Nguyen  and
      Lee, Kyumin",
    booktitle = "Proceedings of the 2020 Conference on Empirical Methods in Natural Language Processing (EMNLP)",
    month = nov,
    year = "2020",
    address = "Online",
    publisher = "Association for Computational Linguistics",
    url = "https://www.aclweb.org/anthology/2020.emnlp-main.621",
    doi = "10.18653/v1/2020.emnlp-main.621",
    pages = "7717--7731",
}

@inproceedings{kim-choi-2020-unsupervised,
    title = "Unsupervised Fact Checking by Counter-Weighted Positive and Negative Evidential Paths in A Knowledge Graph",
    author = "Kim, Jiseong  and
      Choi, Key-sun",
    booktitle = "Proceedings of the 28th International Conference on Computational Linguistics",
    month = dec,
    year = "2020",
    address = "Barcelona, Spain (Online)",
    publisher = "International Committee on Computational Linguistics",
    url = "https://www.aclweb.org/anthology/2020.coling-main.147",
    pages = "1677--1686",
}

@inproceedings{kotonya-toni-2020-explainable-automated,
    title = "Explainable Automated Fact-Checking for Public Health Claims",
    author = "Kotonya, Neema  and
      Toni, Francesca",
    booktitle = "Proceedings of the 2020 Conference on Empirical Methods in Natural Language Processing (EMNLP)",
    month = nov,
    year = "2020",
    address = "Online",
    publisher = "Association for Computational Linguistics",
    url = "https://www.aclweb.org/anthology/2020.emnlp-main.623",
    doi = "10.18653/v1/2020.emnlp-main.623",
    pages = "7740--7754",
}

@inproceedings{yuan-etal-2020-early,
    title = "Early Detection of Fake News by Utilizing the Credibility of News, Publishers, and Users based on Weakly Supervised Learning",
    author = "Yuan, Chunyuan  and
      Ma, Qianwen  and
      Zhou, Wei  and
      Han, Jizhong  and
      Hu, Songlin",
    booktitle = "Proceedings of the 28th International Conference on Computational Linguistics",
    month = dec,
    year = "2020",
    address = "Barcelona, Spain (Online)",
    publisher = "International Committee on Computational Linguistics",
    url = "https://www.aclweb.org/anthology/2020.coling-main.475",
    pages = "5444--5454",
}

@inproceedings{del-tredici-fernandez-2020-words,
    title = "Words are the Window to the Soul: Language-based User Representations for Fake News Detection",
    author = "Del Tredici, Marco  and
      Fern{\'a}ndez, Raquel",
    booktitle = "Proceedings of the 28th International Conference on Computational Linguistics",
    month = dec,
    year = "2020",
    address = "Barcelona, Spain (Online)",
    publisher = "International Committee on Computational Linguistics",
    url = "https://www.aclweb.org/anthology/2020.coling-main.477",
    pages = "5467--5479",
}

@inproceedings{lu-li-2020-gcan,
    title = "{GCAN}: Graph-aware Co-Attention Networks for Explainable Fake News Detection on Social Media",
    author = "Lu, Yi-Ju  and
      Li, Cheng-Te",
    booktitle = "Proceedings of the 58th Annual Meeting of the Association for Computational Linguistics",
    month = jul,
    year = "2020",
    address = "Online",
    publisher = "Association for Computational Linguistics",
    url = "https://www.aclweb.org/anthology/2020.acl-main.48",
    doi = "10.18653/v1/2020.acl-main.48",
    pages = "505--514",
}

@inproceedings{shaar-etal-2020-known,
    title = "That is a Known Lie: Detecting Previously Fact-Checked Claims",
    author = "Shaar, Shaden  and
      Babulkov, Nikolay  and
      Da San Martino, Giovanni  and
      Nakov, Preslav",
    booktitle = "Proceedings of the 58th Annual Meeting of the Association for Computational Linguistics",
    month = jul,
    year = "2020",
    address = "Online",
    publisher = "Association for Computational Linguistics",
    url = "https://www.aclweb.org/anthology/2020.acl-main.332",
    doi = "10.18653/v1/2020.acl-main.332",
    pages = "3607--3618",
}

@inproceedings{maronikolakis-etal-2020-analyzing,
    title = "Analyzing Political Parody in Social Media",
    author = "Maronikolakis, Antonios  and
      S{\'a}nchez Villegas, Danae  and
      Preotiuc-Pietro, Daniel  and
      Aletras, Nikolaos",
    booktitle = "Proceedings of the 58th Annual Meeting of the Association for Computational Linguistics",
    month = jul,
    year = "2020",
    address = "Online",
    publisher = "Association for Computational Linguistics",
    url = "https://www.aclweb.org/anthology/2020.acl-main.403",
    doi = "10.18653/v1/2020.acl-main.403",
    pages = "4373--4384",
}

@inproceedings{zhong-etal-2020-logicalfactchecker,
    title = "{L}ogical{F}act{C}hecker: Leveraging Logical Operations for Fact Checking with Graph Module Network",
    author = "Zhong, Wanjun  and
      Tang, Duyu  and
      Feng, Zhangyin  and
      Duan, Nan  and
      Zhou, Ming  and
      Gong, Ming  and
      Shou, Linjun  and
      Jiang, Daxin  and
      Wang, Jiahai  and
      Yin, Jian",
    booktitle = "Proceedings of the 58th Annual Meeting of the Association for Computational Linguistics",
    month = jul,
    year = "2020",
    address = "Online",
    publisher = "Association for Computational Linguistics",
    url = "https://www.aclweb.org/anthology/2020.acl-main.539",
    doi = "10.18653/v1/2020.acl-main.539",
    pages = "6053--6065",
}

@inproceedings{zhong-etal-2020-reasoning,
    title = "Reasoning Over Semantic-Level Graph for Fact Checking",
    author = "Zhong, Wanjun  and
      Xu, Jingjing  and
      Tang, Duyu  and
      Xu, Zenan  and
      Duan, Nan  and
      Zhou, Ming  and
      Wang, Jiahai  and
      Yin, Jian",
    booktitle = "Proceedings of the 58th Annual Meeting of the Association for Computational Linguistics",
    month = jul,
    year = "2020",
    address = "Online",
    publisher = "Association for Computational Linguistics",
    url = "https://www.aclweb.org/anthology/2020.acl-main.549",
    doi = "10.18653/v1/2020.acl-main.549",
    pages = "6170--6180",
}

@inproceedings{orbach-etal-2020-echo,
    title = "Out of the Echo Chamber: {D}etecting Countering Debate Speeches",
    author = "Orbach, Matan  and
      Bilu, Yonatan  and
      Toledo, Assaf  and
      Lahav, Dan  and
      Jacovi, Michal  and
      Aharonov, Ranit  and
      Slonim, Noam",
    booktitle = "Proceedings of the 58th Annual Meeting of the Association for Computational Linguistics",
    month = jul,
    year = "2020",
    address = "Online",
    publisher = "Association for Computational Linguistics",
    url = "https://www.aclweb.org/anthology/2020.acl-main.633",
    doi = "10.18653/v1/2020.acl-main.633",
    pages = "7073--7086",
}

@inproceedings{vasileva-etal-2019-takes,
    title = "It Takes Nine to Smell a Rat: Neural Multi-Task Learning for Check-Worthiness Prediction",
    author = "Vasileva, Slavena  and
      Atanasova, Pepa  and
      M{\`a}rquez, Llu{\'\i}s  and
      Barr{\'o}n-Cede{\~n}o, Alberto  and
      Nakov, Preslav",
    booktitle = "Proceedings of the International Conference on Recent Advances in Natural Language Processing (RANLP 2019)",
    month = sep,
    year = "2019",
    address = "Varna, Bulgaria",
    publisher = "INCOMA Ltd.",
    url = "https://www.aclweb.org/anthology/R19-1141",
    doi = "10.26615/978-954-452-056-4_141",
    pages = "1229--1239",
}

@inproceedings{hanselowski-etal-2019-richly,
    title = "A Richly Annotated Corpus for Different Tasks in Automated Fact-Checking",
    author = "Hanselowski, Andreas  and
      Stab, Christian  and
      Schulz, Claudia  and
      Li, Zile  and
      Gurevych, Iryna",
    booktitle = "Proceedings of the 23rd Conference on Computational Natural Language Learning (CoNLL)",
    month = nov,
    year = "2019",
    address = "Hong Kong, China",
    publisher = "Association for Computational Linguistics",
    url = "https://www.aclweb.org/anthology/K19-1046",
    doi = "10.18653/v1/K19-1046",
    pages = "493--503",
}

@inproceedings{zlatkova-etal-2019-fact,
    title = "Fact-Checking Meets Fauxtography: Verifying Claims About Images",
    author = "Zlatkova, Dimitrina  and
      Nakov, Preslav  and
      Koychev, Ivan",
    booktitle = "Proceedings of the 2019 Conference on Empirical Methods in Natural Language Processing and the 9th International Joint Conference on Natural Language Processing (EMNLP-IJCNLP)",
    month = nov,
    year = "2019",
    address = "Hong Kong, China",
    publisher = "Association for Computational Linguistics",
    url = "https://www.aclweb.org/anthology/D19-1216",
    doi = "10.18653/v1/D19-1216",
    pages = "2099--2108",
}

@inproceedings{suntwal-etal-2019-importance,
    title = "On the Importance of Delexicalization for Fact Verification",
    author = "Suntwal, Sandeep  and
      Paul, Mithun  and
      Sharp, Rebecca  and
      Surdeanu, Mihai",
    booktitle = "Proceedings of the 2019 Conference on Empirical Methods in Natural Language Processing and the 9th International Joint Conference on Natural Language Processing (EMNLP-IJCNLP)",
    month = nov,
    year = "2019",
    address = "Hong Kong, China",
    publisher = "Association for Computational Linguistics",
    url = "https://www.aclweb.org/anthology/D19-1340",
    doi = "10.18653/v1/D19-1340",
    pages = "3413--3418",
}

@inproceedings{naderi-hirst-2018-automated,
    title = "Automated Fact-Checking of Claims in Argumentative Parliamentary Debates",
    author = "Naderi, Nona  and
      Hirst, Graeme",
    booktitle = "Proceedings of the First Workshop on Fact Extraction and {VER}ification ({FEVER})",
    month = nov,
    year = "2018",
    address = "Brussels, Belgium",
    publisher = "Association for Computational Linguistics",
    url = "https://www.aclweb.org/anthology/W18-5509",
    doi = "10.18653/v1/W18-5509",
    pages = "60--65",
}

@inproceedings{popat-etal-2018-declare,
    title = "{D}e{C}lar{E}: Debunking Fake News and False Claims using Evidence-Aware Deep Learning",
    author = "Popat, Kashyap  and
      Mukherjee, Subhabrata  and
      Yates, Andrew  and
      Weikum, Gerhard",
    booktitle = "Proceedings of the 2018 Conference on Empirical Methods in Natural Language Processing",
    month = oct # "-" # nov,
    year = "2018",
    address = "Brussels, Belgium",
    publisher = "Association for Computational Linguistics",
    url = "https://www.aclweb.org/anthology/D18-1003",
    doi = "10.18653/v1/D18-1003",
    pages = "22--32",
}

@inproceedings{karadzhov-etal-2017-fully,
    title = "Fully Automated Fact Checking Using External Sources",
    author = "Karadzhov, Georgi  and
      Nakov, Preslav  and
      M{\`a}rquez, Llu{\'\i}s  and
      Barr{\'o}n-Cede{\~n}o, Alberto  and
      Koychev, Ivan",
    booktitle = "Proceedings of the International Conference Recent Advances in Natural Language Processing, {RANLP} 2017",
    month = sep,
    year = "2017",
    address = "Varna, Bulgaria",
    publisher = "INCOMA Ltd.",
    url = "https://doi.org/10.26615/978-954-452-049-6_046",
    doi = "10.26615/978-954-452-049-6_046",
    pages = "344--353",
}

@inproceedings{long-etal-2017-fake,
    title = "Fake News Detection Through Multi-Perspective Speaker Profiles",
    author = "Long, Yunfei  and
      Lu, Qin  and
      Xiang, Rong  and
      Li, Minglei  and
      Huang, Chu-Ren",
    booktitle = "Proceedings of the Eighth International Joint Conference on Natural Language Processing (Volume 2: Short Papers)",
    month = nov,
    year = "2017",
    address = "Taipei, Taiwan",
    publisher = "Asian Federation of Natural Language Processing",
    url = "https://www.aclweb.org/anthology/I17-2043",
    pages = "252--256",
}

@inproceedings{thorne-vlachos-2017-extensible,
    title = "An Extensible Framework for Verification of Numerical Claims",
    author = "Thorne, James  and
      Vlachos, Andreas",
    booktitle = "Proceedings of the Software Demonstrations of the 15th Conference of the {E}uropean Chapter of the Association for Computational Linguistics",
    month = apr,
    year = "2017",
    address = "Valencia, Spain",
    publisher = "Association for Computational Linguistics",
    url = "https://www.aclweb.org/anthology/E17-3010",
    pages = "37--40",
}

@inproceedings{vincze-szabo-2020-automatic,
    title = "Automatic Detection of {H}ungarian Clickbait and Entertaining Fake News",
    author = "Vincze, Veronika  and
      Szab{\'o}, Martina Katalin",
    booktitle = "Proceedings of the 3rd International Workshop on Rumours and Deception in Social Media (RDSM)",
    month = dec,
    year = "2020",
    address = "Barcelona, Spain (Online)",
    publisher = "Association for Computational Linguistics",
    url = "https://www.aclweb.org/anthology/2020.rdsm-1.6",
    pages = "58--69",
}

@inproceedings{saadany-etal-2020-fake,
    title = "Fake or Real? A Study of {A}rabic Satirical Fake News",
    author = "Saadany, Hadeel  and
      Orasan, Constantin  and
      Mohamed, Emad",
    booktitle = "Proceedings of the 3rd International Workshop on Rumours and Deception in Social Media (RDSM)",
    month = dec,
    year = "2020",
    address = "Barcelona, Spain (Online)",
    publisher = "Association for Computational Linguistics",
    url = "https://www.aclweb.org/anthology/2020.rdsm-1.7",
    pages = "70--80",
}

@inproceedings{saldanha-etal-2020-understanding,
    title = "Understanding and Explicitly Measuring Linguistic and Stylistic Properties of Deception via Generation and Translation",
    author = "Saldanha, Emily  and
      Garimella, Aparna  and
      Volkova, Svitlana",
    booktitle = "Proceedings of the 13th International Conference on Natural Language Generation",
    month = dec,
    year = "2020",
    address = "Dublin, Ireland",
    publisher = "Association for Computational Linguistics",
    url = "https://www.aclweb.org/anthology/2020.inlg-1.27",
    pages = "216--226",
}

@inproceedings{huguet-cabot-etal-2020-pragmatics,
    title = "{T}he {P}ragmatics behind {P}olitics: {M}odelling {M}etaphor, {F}raming and {E}motion in {P}olitical {D}iscourse",
    author = "Huguet Cabot, Pere-Llu{\'\i}s  and
      Dankers, Verna  and
      Abadi, David  and
      Fischer, Agneta  and
      Shutova, Ekaterina",
    booktitle = "Findings of the Association for Computational Linguistics: EMNLP 2020",
    month = nov,
    year = "2020",
    address = "Online",
    publisher = "Association for Computational Linguistics",
    url = "https://www.aclweb.org/anthology/2020.findings-emnlp.402",
    doi = "10.18653/v1/2020.findings-emnlp.402",
    pages = "4479--4488",
}

@inproceedings{alhindi-etal-2020-fact,
    title = "Fact vs. Opinion: the Role of Argumentation Features in News Classification",
    author = "Alhindi, Tariq  and
      Muresan, Smaranda  and
      Preotiuc-Pietro, Daniel",
    booktitle = "Proceedings of the 28th International Conference on Computational Linguistics",
    month = dec,
    year = "2020",
    address = "Barcelona, Spain (Online)",
    publisher = "International Committee on Computational Linguistics",
    url = "https://www.aclweb.org/anthology/2020.coling-main.540",
    pages = "6139--6149",
}

@inproceedings{srivastava-etal-2019-vernon,
    title = "Vernon-fenwick at {S}em{E}val-2019 Task 4: Hyperpartisan News Detection using Lexical and Semantic Features",
    author = "Srivastava, Vertika  and
      Gupta, Ankita  and
      Prakash, Divya  and
      Sahoo, Sudeep Kumar  and
      R.R, Rohit  and
      Kim, Yeon Hyang",
    booktitle = "Proceedings of the 13th International Workshop on Semantic Evaluation",
    month = jun,
    year = "2019",
    address = "Minneapolis, Minnesota, USA",
    publisher = "Association for Computational Linguistics",
    url = "https://www.aclweb.org/anthology/S19-2189",
    doi = "10.18653/v1/S19-2189",
    pages = "1078--1082",
}

@inproceedings{kiesel-etal-2019-semeval,
    title = "{S}em{E}val-2019 Task 4: Hyperpartisan News Detection",
    author = "Kiesel, Johannes  and
      Mestre, Maria  and
      Shukla, Rishabh  and
      Vincent, Emmanuel  and
      Adineh, Payam  and
      Corney, David  and
      Stein, Benno  and
      Potthast, Martin",
    booktitle = "Proceedings of the 13th International Workshop on Semantic Evaluation",
    month = jun,
    year = "2019",
    address = "Minneapolis, Minnesota, USA",
    publisher = "Association for Computational Linguistics",
    url = "https://www.aclweb.org/anthology/S19-2145",
    doi = "10.18653/v1/S19-2145",
    pages = "829--839",
}

@inproceedings{levi-etal-2019-identifying,
    title = "Identifying Nuances in Fake News vs. Satire: Using Semantic and Linguistic Cues",
    author = "Levi, Or  and
      Hosseini, Pedram  and
      Diab, Mona  and
      Broniatowski, David",
    booktitle = "Proceedings of the Second Workshop on Natural Language Processing for Internet Freedom: Censorship, Disinformation, and Propaganda",
    month = nov,
    year = "2019",
    address = "Hong Kong, China",
    publisher = "Association for Computational Linguistics",
    url = "https://www.aclweb.org/anthology/D19-5004",
    doi = "10.18653/v1/D19-5004",
    pages = "31--35",
}

@inproceedings{rehm-etal-2018-automatic,
    title = "Automatic and Manual Web Annotations in an Infrastructure to handle Fake News and other Online Media Phenomena",
    author = "Rehm, Georg  and
      Moreno-Schneider, Julian  and
      Bourgonje, Peter",
    booktitle = "Proceedings of the Eleventh International Conference on Language Resources and Evaluation ({LREC} 2018)",
    month = may,
    year = "2018",
    address = "Miyazaki, Japan",
    publisher = "European Language Resources Association (ELRA)",
    url = "https://www.aclweb.org/anthology/L18-1384",
}

@inproceedings{baly-etal-2018-predicting,
    title = "Predicting Factuality of Reporting and Bias of News Media Sources",
    author = "Baly, Ramy  and
      Karadzhov, Georgi  and
      Alexandrov, Dimitar  and
      Glass, James  and
      Nakov, Preslav",
    booktitle = "Proceedings of the 2018 Conference on Empirical Methods in Natural Language Processing",
    month = oct # "-" # nov,
    year = "2018",
    address = "Brussels, Belgium",
    publisher = "Association for Computational Linguistics",
    url = "https://www.aclweb.org/anthology/D18-1389",
    doi = "10.18653/v1/D18-1389",
    pages = "3528--3539",
}

@inproceedings{de-sarkar-etal-2018-attending,
    title = "Attending Sentences to detect Satirical Fake News",
    author = "De Sarkar, Sohan  and
      Yang, Fan  and
      Mukherjee, Arjun",
    booktitle = "Proceedings of the 27th International Conference on Computational Linguistics",
    month = aug,
    year = "2018",
    address = "Santa Fe, New Mexico, USA",
    publisher = "Association for Computational Linguistics",
    url = "https://www.aclweb.org/anthology/C18-1285",
    pages = "3371--3380",
}

@inproceedings{li-etal-2017-nlp,
    title = "An {NLP} Analysis of Exaggerated Claims in Science News",
    author = "Li, Yingya  and
      Zhang, Jieke  and
      Yu, Bei",
    booktitle = "Proceedings of the 2017 {EMNLP} Workshop: Natural Language Processing meets Journalism",
    month = sep,
    year = "2017",
    address = "Copenhagen, Denmark",
    publisher = "Association for Computational Linguistics",
    url = "https://www.aclweb.org/anthology/W17-4219",
    doi = "10.18653/v1/W17-4219",
    pages = "106--111",
}

@inproceedings{dusmanu-etal-2017-argument,
    title = "Argument Mining on {T}witter: Arguments, Facts and Sources",
    author = "Dusmanu, Mihai  and
      Cabrio, Elena  and
      Villata, Serena",
    booktitle = "Proceedings of the 2017 Conference on Empirical Methods in Natural Language Processing",
    month = sep,
    year = "2017",
    address = "Copenhagen, Denmark",
    publisher = "Association for Computational Linguistics",
    url = "https://www.aclweb.org/anthology/D17-1245",
    doi = "10.18653/v1/D17-1245",
    pages = "2317--2322",
}

@inproceedings{rashkin-etal-2017-truth,
    title = "Truth of Varying Shades: Analyzing Language in Fake News and Political Fact-Checking",
    author = "Rashkin, Hannah  and
      Choi, Eunsol  and
      Jang, Jin Yea  and
      Volkova, Svitlana  and
      Choi, Yejin",
    booktitle = "Proceedings of the 2017 Conference on Empirical Methods in Natural Language Processing",
    month = sep,
    year = "2017",
    address = "Copenhagen, Denmark",
    publisher = "Association for Computational Linguistics",
    url = "https://www.aclweb.org/anthology/D17-1317",
    doi = "10.18653/v1/D17-1317",
    pages = "2931--2937",
}

@inproceedings{liang-etal-2012-expert,
    title = "Expert Finding for Microblog Misinformation Identification",
    author = "Liang, Chen  and
      Liu, Zhiyuan  and
      Sun, Maosong",
    booktitle = "Proceedings of {COLING} 2012: Posters",
    month = dec,
    year = "2012",
    address = "Mumbai, India",
    publisher = "The COLING 2012 Organizing Committee",
    url = "https://www.aclweb.org/anthology/C12-2069",
    pages = "703--712",
}

@misc{digitalservicesact2020,
  title={{Shaping Europe’s digital future: The Digital Services Act package}},
  author={European Commission},
  month={12},
  year={2020},
  url={https://ec.europa.eu/digital-single-market/en/digital-services-act-package}
}

@inproceedings{aroca-ouellette-rudzicz-2020-losses,
    title = "{O}n {L}osses for {M}odern {L}anguage {M}odels",
    author = "Aroca-Ouellette, St{\'e}phane  and
      Rudzicz, Frank",
    booktitle = "Proceedings of the 2020 Conference on Empirical Methods in Natural Language Processing (EMNLP)",
    month = nov,
    year = "2020",
    address = "Online",
    publisher = "Association for Computational Linguistics",
    url = "https://www.aclweb.org/anthology/2020.emnlp-main.403",
    doi = "10.18653/v1/2020.emnlp-main.403",
    pages = "4970--4981",
}

@article{journals/corr/abs-2002-12327,
  author = {Rogers, Anna and Kovaleva, Olga and Rumshisky, Anna},
  ee = {https://arxiv.org/abs/2002.12327},
  journal = {CoRR},
  title = {A Primer in BERTology: What we know about how BERT works.},
  url = {http://dblp.uni-trier.de/db/journals/corr/corr2002.html#abs-2002-12327},
  volume = {abs/2002.12327},
  year = 2020
}

@misc{francis2016fast,
  title={Fast \& Furious Fact Check Challenge},
  author={Francis, Diane},
  year={2016},
  url={https://www.herox.com/factcheck/5-practise-claims},
}

@article{Bovet2019,
  author = {Bovet, Alexandre and Makse, Hern{\'a}n A.},
  title = {Influence of fake news in Twitter during the 2016 US presidential election},
  journal = {Nature Communications},
  year = 2019,
  volume = 10,
  numer = 1,
  pages = 7,
  da = {2019/01/02},
  doi = {10.1038/s41467-018-07761-2},
  id = {Bovet2019},
  isbn = {2041-1723},
  ty = {JOUR},
  url = {https://doi.org/10.1038/s41467-018-07761-2},
}

@article{juhasz2017political,
  title={The political effects of migration-related fake news, disinformation and conspiracy theories in Europe},
  author={Juh{\'a}sz, Attila and Szicherle, Patrik},
  journal={Friedrich Ebert Stiftung, Political Capital Policy Research \& Consulting Institute, Budapest},
  year={2017}
}

@article{howard2016bots,
  title={Bots, \#strongerin and \#brexit: Computational Propaganda during the UK-EU Referendum},
  author={Howard, Phillip N and Kollany, Bence},
  journal={Social Science Research Network},
  year={2016}
}

@inproceedings{derczynski2019misinformation,
  title={Misinformation on Twitter During the Danish National Election: A Case Study},
  author={Derczynski, Leon and Albert-Lindqvist, Torben Oskar and Bendsen, Marius Ven{\o} and Inie, Nanna and Pedersen, Jens Egholm and Pedersen, Viktor Due},
  booktitle={Proceedings of the conference for Truth and Trust Online},
  year={2019}
}

@article{ncube2019digital,
  title={Digital Media, Fake News and Pro-Movement for Democratic Change (MDC) Alliance Cyber-Propaganda during the 2018 Zimbabwe Election},
  author={Ncube, Lyton},
  journal={African Journalism Studies},
  pages={1--18},
  year={2019},
  publisher={Taylor \& Francis}
}

@article{brennen2020types,
  title={Types, Sources, and Claims of COVID-19 Misinformation},
  author={Brennen, J Scott and Simon, Felix and Howard, Philip N and Nielsen, Rasmus Kleis},
  year={2020}
}

@article{kouzy2020coronavirus,
  title={Coronavirus goes viral: quantifying the COVID-19 misinformation epidemic on Twitter},
  author={Kouzy, Ramez and Abi Jaoude, Joseph and Kraitem, Afif and El Alam, Molly B and Karam, Basil and Adib, Elio and Zarka, Jabra and Traboulsi, Cindy and Akl, Elie W and Baddour, Khalil},
  journal={Cureus},
  volume={12},
  number={3},
  year={2020},
  publisher={Cureus Inc.}
}

@article{cuan2020misinformation,
  title={Misinformation of COVID-19 on the Internet: Infodemiology Study},
  author={Cuan-Baltazar, Jose Yunam and Mu{\~n}oz-Perez, Maria Jos{\'e} and Robledo-Vega, Carolina and P{\'e}rez-Zepeda, Maria Fernanda and Soto-Vega, Elena},
  journal={JMIR Public Health and Surveillance},
  volume={6},
  number={2},
  pages={e18444},
  year={2020},
  publisher={JMIR Publications Inc., Toronto, Canada}
}

@article{waisbord2018truth,
  title={Truth is what happens to news: On journalism, fake news, and post-truth},
  author={Waisbord, Silvio},
  journal={Journalism studies},
  volume={19},
  number={13},
  pages={1866--1878},
  year={2018},
  publisher={Taylor \& Francis}
}

@paper{gephi,
author = {Mathieu Bastian and Sebastien Heymann and Mathieu Jacomy},
title = {Gephi: An Open Source Software for Exploring and Manipulating Networks},
conference = {International AAAI Conference on Weblogs and Social Media},
year = {2009},
url = {http://www.aaai.org/ocs/index.php/ICWSM/09/paper/view/154}
}

@article{oates2016,
  title={Russian media in the digital age: Propaganda rewired},
  author={Oates, Sarah},
  journal={Russian Politics},
  volume={1},
  number={4},
  pages={398--417},
  year={2016},
  publisher={Brill}
}

@inproceedings{goodfellow2015explaining,
  author = {Goodfellow, Ian J. and Shlens, Jonathon and Szegedy, Christian},
  booktitle = {ICLR (Poster)},
  crossref = {conf/iclr/2015},
  editor = {Bengio, Yoshua and LeCun, Yann},
  ee = {http://arxiv.org/abs/1412.6572},
  title = {Explaining and Harnessing Adversarial Examples.},
  url = {http://dblp.uni-trier.de/db/conf/iclr/iclr2015.html#GoodfellowSS14},
  year = 2015
}

@inproceedings{szegedy2013intriguing,
  author    = {Christian Szegedy and
               Wojciech Zaremba and
               Ilya Sutskever and
               Joan Bruna and
               Dumitru Erhan and
               Ian J. Goodfellow and
               Rob Fergus},
  editor    = {Yoshua Bengio and
               Yann LeCun},
  title     = {Intriguing properties of neural networks},
  booktitle = {2nd International Conference on Learning Representations, {ICLR} 2014,
               Banff, AB, Canada, April 14-16, 2014, Conference Track Proceedings},
  year      = {2014},
  url       = {http://arxiv.org/abs/1312.6199},
  bibsource = {dblp computer science bibliography, https://dblp.org}
}

@article{keskar2019ctrl,
  title={Ctrl: A conditional transformer language model for controllable generation},
  author={Keskar, Nitish Shirish and McCann, Bryan and Varshney, Lav R and Xiong, Caiming and Socher, Richard},
  journal={arXiv preprint arXiv:1909.05858},
  year={2019}
}

@inproceedings{kim-allan-2019-fever,
    title = "{FEVER} Breaker{'}s Run of Team {N}b{A}uz{D}r{L}qg",
    author = "Kim, Youngwoo  and
      Allan, James",
    booktitle = "Proceedings of the Second Workshop on Fact Extraction and VERification (FEVER)",
    month = nov,
    year = "2019",
    address = "Hong Kong, China",
    publisher = "Association for Computational Linguistics",
    url = "https://www.aclweb.org/anthology/D19-6615",
    doi = "10.18653/v1/D19-6615",
    pages = "99--104",
}

@inproceedings{hidey-etal-2020-deseption,
    title = "{D}e{S}e{P}tion: Dual Sequence Prediction and Adversarial Examples for Improved Fact-Checking",
    author = "Hidey, Christopher  and
      Chakrabarty, Tuhin  and
      Alhindi, Tariq  and
      Varia, Siddharth  and
      Krstovski, Kriste  and
      Diab, Mona  and
      Muresan, Smaranda",
    booktitle = "Proceedings of the 58th Annual Meeting of the Association for Computational Linguistics",
    month = jul,
    year = "2020",
    address = "Online",
    publisher = "Association for Computational Linguistics",
    url = "https://www.aclweb.org/anthology/2020.acl-main.761",
    doi = "10.18653/v1/2020.acl-main.761",
    pages = "8593--8606",
}

@article{dia2019semantics,
    title={{Semantics Preserving Adversarial Learning}},
    author={Ousmane Amadou Dia and Elnaz Barshan and Reza Babanezhad},
    year={2019},
    eprint={1903.03905},
    archivePrefix={arXiv},
    primaryClass={stat.ML},
    journal={arXiv preprint arXiv:1903.03905},
}

@inproceedings{niewinski-etal-2019-gem,
    title = "{GEM}: Generative Enhanced Model for adversarial attacks",
    author = "Niewinski, Piotr  and
      Pszona, Maria  and
      Janicka, Maria",
    booktitle = "Proceedings of the Second Workshop on Fact Extraction and VERification (FEVER)",
    month = nov,
    year = "2019",
    address = "Hong Kong, China",
    publisher = "Association for Computational Linguistics",
    url = "https://www.aclweb.org/anthology/D19-6604",
    doi = "10.18653/v1/D19-6604",
    pages = "20--26",
}

@inproceedings{ribeiro2018semantically,
  title={{Semantically equivalent adversarial rules for debugging nlp models}},
  author={Ribeiro, Marco Tulio and Singh, Sameer and Guestrin, Carlos},
  booktitle={Proceedings of the 56th Annual Meeting of the Association for Computational Linguistics (Volume 1: Long Papers)},
  pages={856--865},
  year={2018}
}

@inproceedings{ren2019generating,
    title = "Generating Natural Language Adversarial Examples through Probability Weighted Word Saliency",
    author = "Ren, Shuhuai  and
      Deng, Yihe  and
      He, Kun  and
      Che, Wanxiang",
    booktitle = "Proceedings of the 57th Annual Meeting of the Association for Computational Linguistics",
    month = jul,
    year = "2019",
    address = "Florence, Italy",
    publisher = "Association for Computational Linguistics",
    url = "https://www.aclweb.org/anthology/P19-1103",
    doi = "10.18653/v1/P19-1103",
    pages = "1085--1097",
}

@article{gao2019universal,
    title={{Universal Adversarial Perturbation for Text Classification}},
    author={Hang Gao and Tim Oates},
    year={2019},
    eprint={1910.04618},
    archivePrefix={arXiv},
    primaryClass={cs.CL},
    journal={arXiv preprint arXiv:1910.04618}
}

@inproceedings{michel2019evaluation,
  title={{On Evaluation of Adversarial Perturbations for Sequence-to-Sequence Models}},
  author={Michel, Paul and Li, Xian and Neubig, Graham and Pino, Juan Miguel},
  booktitle={Proceedings of NAACL-HLT},
  pages={3103--3114},
  year={2019}
}

@inproceedings{ebrahimi2018hotflip,
  title={{HotFlip: White-Box Adversarial Examples for Text Classification}},
  author={Ebrahimi, Javid and Rao, Anyi and Lowd, Daniel and Dou, Dejing},
  booktitle={Proceedings of the 56th Annual Meeting of the Association for Computational Linguistics (Volume 2: Short Papers)},
  pages={31--36},
  year={2018}
}

@software{jintextfool,
  author       = {Bogdan Kulynych},
  title        = {textfool},
  month        = jul,
  year         = 2017,
  publisher    = {Zenodo},
  version      = {v0.1},
  doi          = {10.5281/zenodo.831638},
  url          = {https://doi.org/10.5281/zenodo.831638}
}

@article{zhang2019adversarial,
  author    = {Wei Emma Zhang and
               Quan Z. Sheng and
               Ahoud Abdulrahmn F. Alhazmi and
               Chenliang Li},
  title     = {Adversarial Attacks on Deep-learning Models in Natural Language Processing:
               {A} Survey},
  journal   = {{ACM} Trans. Intell. Syst. Technol.},
  volume    = {11},
  number    = {3},
  pages     = {24:1--24:41},
  year      = {2020},
  url       = {https://doi.org/10.1145/3374217},
  doi       = {10.1145/3374217},
  bibsource = {dblp computer science bibliography, https://dblp.org}
}

@inproceedings{thorne2019evaluating,
  title={{Evaluating adversarial attacks against multiple fact verification systems}},
  author={Thorne, James and Vlachos, Andreas and Christodoulopoulos, Christos and Mittal, Arpit},
  booktitle={Proceedings of the 2019 Conference on Empirical Methods in Natural Language Processing and the 9th International Joint Conference on Natural Language Processing (EMNLP-IJCNLP)},
  pages={2937--2946},
  year={2019}
}

@article{liu2019roberta,
  title={{Roberta: A robustly optimized bert pretraining approach}},
  author={Liu, Yinhan and Ott, Myle and Goyal, Naman and Du, Jingfei and Joshi, Mandar and Chen, Danqi and Levy, Omer and Lewis, Mike and Zettlemoyer, Luke and Stoyanov, Veselin},
  journal={arXiv preprint arXiv:1907.11692},
  year={2019}
}

@inproceedings{thorne-etal-2018-fever,
    title = "{FEVER}: a Large-scale Dataset for Fact Extraction and {VER}ification",
    author = "Thorne, James  and
      Vlachos, Andreas  and
      Christodoulopoulos, Christos  and
      Mittal, Arpit",
    booktitle = "Proceedings of the 2018 Conference of the North {A}merican Chapter of the Association for Computational Linguistics: Human Language Technologies, Volume 1 (Long Papers)",
    month = jun,
    year = "2018",
    address = "New Orleans, Louisiana",
    publisher = "Association for Computational Linguistics",
    url = "https://www.aclweb.org/anthology/N18-1074",
    doi = "10.18653/v1/N18-1074",
    pages = "809--819",
}

@inproceedings{thorne-etal-2019-fever2,
    title = "The {FEVER}2.0 Shared Task",
    author = "Thorne, James  and
      Vlachos, Andreas  and
      Cocarascu, Oana  and
      Christodoulopoulos, Christos  and
      Mittal, Arpit",
    booktitle = "Proceedings of the Second Workshop on Fact Extraction and VERification (FEVER)",
    month = nov,
    year = "2019",
    address = "Hong Kong, China",
    publisher = "Association for Computational Linguistics",
    url = "https://www.aclweb.org/anthology/D19-6601",
    doi = "10.18653/v1/D19-6601",
    pages = "1--6",
}

@inproceedings{wallace2019universal,
  title={{Universal Adversarial Triggers for Attacking and Analyzing NLP}},
  author={Wallace, Eric and Feng, Shi and Kandpal, Nikhil and Gardner, Matt and Singh, Sameer},
  booktitle={Proceedings of the 2019 Conference on Empirical Methods in Natural Language Processing and the 9th International Joint Conference on Natural Language Processing (EMNLP-IJCNLP)},
  pages={2153--2162},
  year={2019}
}

@article{radford2019language,
  title={Language models are unsupervised multitask learners},
  author={Radford, Alec and Wu, Jeffrey and Child, Rewon and Luan, David and Amodei, Dario and Sutskever, Ilya},
  journal={OpenAI Blog},
  volume={1},
  number={8},
  pages={9},
  year={2019}
}

@article{Wolf2019HuggingFacesTS,
  title={HuggingFace's Transformers: State-of-the-art Natural Language Processing},
  author={Thomas Wolf and Lysandre Debut and Victor Sanh and Julien Chaumond and Clement Delangue and Anthony Moi and Pierric Cistac and Tim Rault and R'emi Louf and Morgan Funtowicz and Jamie Brew},
  journal={ArXiv},
  year={2019},
  volume={abs/1910.03771}
}

@article{ostrowski2020multihop,
    title={{Multi-Hop Fact Checking of Political Claims}},
    author={Wojciech Ostrowski and Arnav Arora and Pepa Atanasova and Isabelle Augenstein},
    year={2020},
    eprint={2009.06401},
    archivePrefix={arXiv},
    primaryClass={cs.CL},
    journal={arXiv preprint arXiv:2009.06401}
}

@inproceedings{yin-roth-2018-twowingos,
    title = "{T}wo{W}ing{OS}: A Two-Wing Optimization Strategy for Evidential Claim Verification",
    author = "Yin, Wenpeng  and
      Roth, Dan",
    booktitle = "Proceedings of the 2018 Conference on Empirical Methods in Natural Language Processing",
    month = oct # "-" # nov,
    year = "2018",
    address = "Brussels, Belgium",
    publisher = "Association for Computational Linguistics",
    url = "https://www.aclweb.org/anthology/D18-1010",
    doi = "10.18653/v1/D18-1010",
    pages = "105--114",
}

@inproceedings{hansen2019neural,
  title={{Neural Check-Worthiness Ranking With Weak Supervision: Finding Sentences for Fact-Checking}},
  author={Hansen, Casper and Hansen, Christian and Alstrup, Stephen and Grue Simonsen, Jakob and Lioma, Christina},
  booktitle={Companion Proceedings of the 2019 World Wide Web Conference},
  pages={994--1000},
  year={2019}
}

@inproceedings{atanasova2018overview,
  author    = {Preslav Nakov and
               Alberto Barr{\'{o}}n{-}Cede{\~{n}}o and
               Tamer Elsayed and
               Reem Suwaileh and
               Llu{\'{\i}}s M{\`{a}}rquez and
               Wajdi Zaghouani and
               Pepa Atanasova and
               Spas Kyuchukov and
               Giovanni Da San Martino},
  editor    = {Patrice Bellot and
               Chiraz Trabelsi and
               Josiane Mothe and
               Fionn Murtagh and
               Jian{-}Yun Nie and
               Laure Soulier and
               Eric SanJuan and
               Linda Cappellato and
               Nicola Ferro},
  title     = {Overview of the {CLEF-2018} CheckThat! Lab on Automatic Identification
               and Verification of Political Claims},
  booktitle = {Experimental {IR} Meets Multilinguality, Multimodality, and Interaction
               - 9th International Conference of the {CLEF} Association, {CLEF} 2018,
               Avignon, France, September 10-14, 2018, Proceedings},
  series    = {Lecture Notes in Computer Science},
  volume    = {11018},
  pages     = {372--387},
  publisher = {Springer},
  year      = {2018},
  url       = {https://doi.org/10.1007/978-3-319-98932-7\_32},
  doi       = {10.1007/978-3-319-98932-7\_32},
  bibsource = {dblp computer science bibliography, https://dblp.org}
}

@inproceedings{baly-etal-2018-integrating,
    title = "Integrating Stance Detection and Fact Checking in a Unified Corpus",
    author = "Baly, Ramy  and
      Mohtarami, Mitra  and
      Glass, James  and
      M{\`a}rquez, Llu{\'\i}s  and
      Moschitti, Alessandro  and
      Nakov, Preslav",
    booktitle = "Proceedings of the 2018 Conference of the North {A}merican Chapter of the Association for Computational Linguistics: Human Language Technologies, Volume 2 (Short Papers)",
    month = jun,
    year = "2018",
    address = "New Orleans, Louisiana",
    publisher = "Association for Computational Linguistics",
    url = "https://www.aclweb.org/anthology/N18-2004",
    doi = "10.18653/v1/N18-2004",
    pages = "21--27",
}

@inproceedings{elkan2008learning,
  title={Learning classifiers from only positive and unlabeled data},
  author={Elkan, Charles and Noto, Keith},
  booktitle={Proceedings of the 14th ACM SIGKDD international conference on Knowledge discovery and data mining},
  pages={213--220},
  year={2008}
}

@inproceedings{kiryo2017positive,
  author    = {Ryuichi Kiryo and
               Gang Niu and
               Marthinus Christoffel du Plessis and
               Masashi Sugiyama},
  editor    = {Isabelle Guyon and
               Ulrike von Luxburg and
               Samy Bengio and
               Hanna M. Wallach and
               Rob Fergus and
               S. V. N. Vishwanathan and
               Roman Garnett},
  title     = {Positive-Unlabeled Learning with Non-Negative Risk Estimator},
  booktitle = {Advances in Neural Information Processing Systems 30: Annual Conference
               on Neural Information Processing Systems 2017, December 4-9, 2017,
               Long Beach, CA, {USA}},
  pages     = {1675--1685},
  year      = {2017},
  url       = {http://papers.nips.cc/paper/6765-positive-unlabeled-learning-with-non-negative-risk-estimator},
  bibsource = {dblp computer science bibliography, https://dblp.org}
}

@inproceedings{li2014spotting,
  title={Spotting fake reviews via collective positive-unlabeled learning},
  author={Li, Huayi and Chen, Zhiyuan and Liu, Bing and Wei, Xiaokai and Shao, Jidong},
  booktitle={2014 IEEE international conference on data mining},
  pages={899--904},
  year={2014},
  organization={IEEE}
}

@inproceedings{ren2014positive,
    title = "Positive Unlabeled Learning for Deceptive Reviews Detection",
    author = "Ren, Yafeng  and
      Ji, Donghong  and
      Zhang, Hongbin",
    booktitle = "Proceedings of the 2014 Conference on Empirical Methods in Natural Language Processing ({EMNLP})",
    month = oct,
    year = "2014",
    address = "Doha, Qatar",
    publisher = "Association for Computational Linguistics",
    url = "https://www.aclweb.org/anthology/D14-1055",
    doi = "10.3115/v1/D14-1055",
    pages = "488--498",
}

@article{bekker2018learning,
  author    = {Jessa Bekker and
               Jesse Davis},
  title     = {Learning from positive and unlabeled data: a survey},
  journal   = {Mach. Learn.},
  volume    = {109},
  number    = {4},
  pages     = {719--760},
  year      = {2020},
  url       = {https://doi.org/10.1007/s10994-020-05877-5},
  doi       = {10.1007/s10994-020-05877-5},
  bibsource = {dblp computer science bibliography, https://dblp.org}
}

@inproceedings{peng2019distantly,
  title={Distantly Supervised Named Entity Recognition using Positive-Unlabeled Learning},
  author={Peng, Minlong and Xing, Xiaoyu and Zhang, Qi and Fu, Jinlan and Huang, Xuan-Jing},
  booktitle={Proceedings of the 57th Annual Meeting of the Association for Computational Linguistics},
  pages={2409--2419},
  year={2019}
}

@inproceedings{du2014analysis,
  author    = {Marthinus Christoffel du Plessis and
               Gang Niu and
               Masashi Sugiyama},
  editor    = {Zoubin Ghahramani and
               Max Welling and
               Corinna Cortes and
               Neil D. Lawrence and
               Kilian Q. Weinberger},
  title     = {Analysis of Learning from Positive and Unlabeled Data},
  booktitle = {Advances in Neural Information Processing Systems 27: Annual Conference
               on Neural Information Processing Systems 2014, December 8-13 2014,
               Montreal, Quebec, Canada},
  pages     = {703--711},
  year      = {2014},
  url       = {http://papers.nips.cc/paper/5509-analysis-of-learning-from-positive-and-unlabeled-data},
  bibsource = {dblp computer science bibliography, https://dblp.org}
}

@inproceedings{denis1998pac,
  title={PAC learning from positive statistical queries},
  author={Denis, Fran{\c{c}}ois},
  booktitle={International Conference on Algorithmic Learning Theory},
  pages={112--126},
  year={1998},
  organization={Springer}
}

@inproceedings{de1999positive,
  title={Positive and unlabeled examples help learning},
  author={De Comit{\'e}, Francesco and Denis, Fran{\c{c}}ois and Gilleron, R{\'e}mi and Letouzey, Fabien},
  booktitle={International Conference on Algorithmic Learning Theory},
  pages={219--230},
  year={1999},
  organization={Springer}
}

@inproceedings{letouzey2000learning,
  title={Learning from positive and unlabeled examples},
  author={Letouzey, Fabien and Denis, Fran{\c{c}}ois and Gilleron, R{\'e}mi},
  booktitle={International Conference on Algorithmic Learning Theory},
  pages={71--85},
  year={2000},
  organization={Springer}
}

@inproceedings{barron2018overview,
  author    = {Alberto Barr{\'{o}}n{-}Cede{\~{n}}o and
               Tamer Elsayed and
               Reem Suwaileh and
               Llu{\'{\i}}s M{\`{a}}rquez and
               Pepa Atanasova and
               Wajdi Zaghouani and
               Spas Kyuchukov and
               Giovanni Da San Martino and
               Preslav Nakov},
  editor    = {Linda Cappellato and
               Nicola Ferro and
               Jian{-}Yun Nie and
               Laure Soulier},
  title     = {Overview of the {CLEF-2018} CheckThat! Lab on Automatic Identification
               and Verification of Political Claims. Task 2: Factuality},
  booktitle = {Working Notes of {CLEF} 2018 - Conference and Labs of the Evaluation
               Forum, Avignon, France, September 10-14, 2018},
  series    = {{CEUR} Workshop Proceedings},
  volume    = {2125},
  publisher = {CEUR-WS.org},
  year      = {2018},
  url       = {http://ceur-ws.org/Vol-2125/invited\_paper\_14.pdf},
  bibsource = {dblp computer science bibliography, https://dblp.org}
}

@inproceedings{elsayed2019overview,
  title={Overview of the CLEF-2019 CheckThat! Lab: Automatic Identification and Verification of Claims},
  author={Elsayed, Tamer and Nakov, Preslav and Barr{\'o}n-Cede{\~n}o, Alberto and Hasanain, Maram and Suwaileh, Reem and Da San Martino, Giovanni and Atanasova, Pepa},
  booktitle={International Conference of the Cross-Language Evaluation Forum for European Languages},
  pages={301--321},
  year={2019},
  organization={Springer}
}

@inproceedings{zubiaga2017exploiting,
  title={{Exploiting Context for Rumour Detection in Social Media}},
  author={Zubiaga, Arkaitz and Liakata, Maria and Procter, Rob},
  booktitle={International Conference on Social Informatics},
  pages={109--123},
  year={2017},
  organization={Springer}
}

@inproceedings{ma2019detect,
  title={Detect rumors on Twitter by promoting information campaigns with generative adversarial learning},
  author={Ma, Jing and Gao, Wei and Wong, Kam-Fai},
  booktitle={The World Wide Web Conference},
  pages={3049--3055},
  year={2019}
}

@inproceedings{redi2019citation,
  title={Citation needed: A taxonomy and algorithmic assessment of Wikipedia's verifiability},
  author={Redi, Miriam and Fetahu, Besnik and Morgan, Jonathan and Taraborelli, Dario},
  booktitle={The World Wide Web Conference},
  pages={1567--1578},
  year={2019}
}

@inproceedings{abulaish2019graph,
  title={A Graph-Theoretic Embedding-Based Approach for Rumor Detection in Twitter},
  author={Abulaish, Muhammad and Kumari, Nikita and Fazil, Mohd and Singh, Basanta},
  booktitle={IEEE/WIC/ACM International Conference on Web Intelligence},
  pages={466--470},
  year={2019}
}

@article{konstantinovskiy2018towards,
  author    = {Lev Konstantinovskiy and
               Oliver Price and
               Mevan Babakar and
               Arkaitz Zubiaga},
  title     = {Towards Automated Factchecking: Developing an Annotation Schema and
               Benchmark for Consistent Automated Claim Detection},
  journal   = {CoRR},
  volume    = {abs/1809.08193},
  year      = {2018},
  url       = {http://arxiv.org/abs/1809.08193},
  archivePrefix = {arXiv},
  eprint    = {1809.08193},
  bibsource = {dblp computer science bibliography, https://dblp.org}
}

@article{zubiaga2016analysing,
  title={{Analysing How People Orient to and Spread Rumours in Social Media by Looking at Conversational Threads}},
  author={Zubiaga, Arkaitz and Liakata, Maria and Procter, Rob and Hoi, Geraldine Wong Sak and Tolmie, Peter},
  journal={PloS one},
  volume={11},
  number={3},
  year={2016},
  publisher={Public Library of Science}
}

@article{hassan2017claimbuster,
  title={ClaimBuster: the first-ever end-to-end fact-checking system},
  author={Hassan, Naeemul and Zhang, Gensheng and Arslan, Fatma and Caraballo, Josue and Jimenez, Damian and Gawsane, Siddhant and Hasan, Shohedul and Joseph, Minumol and Kulkarni, Aaditya and Nayak, Anil Kumar and others},
  journal={Proceedings of the VLDB Endowment},
  volume={10},
  number={12},
  pages={1945--1948},
  year={2017},
  publisher={VLDB Endowment}
}

@article{del2016spreading,
  title={The spreading of misinformation online},
  author={Del Vicario, Michela and Bessi, Alessandro and Zollo, Fabiana and Petroni, Fabio and Scala, Antonio and Caldarelli, Guido and Stanley, H Eugene and Quattrociocchi, Walter},
  journal={Proceedings of the National Academy of Sciences},
  volume={113},
  number={3},
  pages={554--559},
  year={2016},
  publisher={National Acad Sciences}
}

@article{howell2013digital,
  title={Digital wildfires in a hyperconnected world},
  author={Howell, Lee and others},
  journal={WEF report},
  volume={3},
  pages={15--94},
  year={2013}
}

@inproceedings{vlachos2014fact,
  title={Fact checking: Task definition and dataset construction},
  author={Vlachos, Andreas and Riedel, Sebastian},
  booktitle={Proceedings of the ACL 2014 Workshop on Language Technologies and Computational Social Science},
  pages={18--22},
  year={2014}
}

@inproceedings{vaswani2017attention,
  title={{Attention is All You Need}},
  author={Vaswani, Ashish and Shazeer, Noam and Parmar, Niki and Uszkoreit, Jakob and Jones, Llion and Gomez, Aidan N and Kaiser, {\L}ukasz and Polosukhin, Illia},
  booktitle={Advances in Neural Information Processing Systems},
  pages={5998--6008},
  year={2017}
}

@article{kouw2019review,
  title={{A Review of Domain Adaptation Without Target Labels}},
  author={Kouw, Wouter Marco and Loog, Marco},
  journal={IEEE Transactions on Pattern Analysis and Machine Intelligence},
  year={2019},
  publisher={IEEE}
}

@inproceedings{daumeiii:2007:ACLMain,
  address = {Prague, Czech Republic},
  author = {Daum\'e, III, Hal},
  booktitle = {Proceedings of the 45th Annual Meeting of the Association of Computational Linguistics},
  month = {June},
  pages = {256--263},
  publisher = {Association for Computational Linguistics},
  title = {{Frustratingly Easy Domain Adaptation}},
  url = {http://www.aclweb.org/anthology/P/P07/P07-1033},
  year = 2007
}

@inproceedings{conf/cvpr/KulisSD11,
  author = {Kulis, Brian and Saenko, Kate and Darrell, Trevor},
  booktitle = {CVPR},
  crossref = {conf/cvpr/2011},
  ee = {http://doi.ieeecomputersociety.org/10.1109/CVPR.2011.5995702},
  isbn = {978-1-4577-0394-2},
  pages = {1785-1792},
  publisher = {IEEE Computer Society},
  title = {{What You Saw is Not What You Get: Domain Adaptation Using Asymmetric Kernel Transforms}},
  url = {http://dblp.uni-trier.de/db/conf/cvpr/cvpr2011.html#KulisSD11},
  year = 2011
}

@inproceedings{conf/cvpr/DonahueHRSD13,
  author = {Donahue, Jeff and Hoffman, Judy and Rodner, Erik and Saenko, Kate and Darrell, Trevor},
  booktitle = {CVPR},
  crossref = {conf/cvpr/2013},
  ee = {http://doi.ieeecomputersociety.org/10.1109/CVPR.2013.92},
  isbn = {978-0-7695-4989-7},
  pages = {668-675},
  publisher = {IEEE Computer Society},
  title = {{Semi-supervised Domain Adaptation with Instance Constraints}},
  url = {http://dblp.uni-trier.de/db/conf/cvpr/cvpr2013.html#DonahueHRSD13},
  year = 2013
}

@inproceedings{conf/cvpr/YaoPNLM15,
  author = {Yao, Ting and Pan, Yingwei and Ngo, Chong-Wah and Li, Houqiang and Mei, Tao},
  booktitle = {CVPR},
  crossref = {conf/cvpr/2015},
  ee = {http://doi.ieeecomputersociety.org/10.1109/CVPR.2015.7298826},
  isbn = {978-1-4673-6964-0},
  pages = {2142-2150},
  publisher = {IEEE Computer Society},
  title = {{Semi-supervised Domain Adaptation with Subspace Learning for Visual Recognition}},
  url = {http://dblp.uni-trier.de/db/conf/cvpr/cvpr2015.html#YaoPNLM15},
  year = 2015
}

@inproceedings{conf/aaai/SunFS16,
  author = {Sun, Baochen and Feng, Jiashi and Saenko, Kate},
  booktitle = {AAAI},
  crossref = {conf/aaai/2016},
  editor = {Schuurmans, Dale and Wellman, Michael P.},
  ee = {http://www.aaai.org/ocs/index.php/AAAI/AAAI16/paper/view/12443},
  pages = {2058-2065},
  publisher = {AAAI Press},
  title = {{Return of Frustratingly Easy Domain Adaptation}},
  url = {http://dblp.uni-trier.de/db/conf/aaai/aaai2016.html#SunFS16},
  year = 2016
}

@inproceedings{conf/icml/LiptonWS18,
  author = {Lipton, Zachary C. and Wang, Yu-Xiang and Smola, Alexander J.},
  booktitle = {ICML},
  crossref = {conf/icml/2018},
  editor = {Dy, Jennifer G. and Krause, Andreas},
  ee = {http://proceedings.mlr.press/v80/lipton18a.html},
  pages = {3128-3136},
  publisher = {PMLR},
  series = {Proceedings of Machine Learning Research},
  title = {{Detecting and Correcting for Label Shift with Black Box Predictors}},
  url = {http://dblp.uni-trier.de/db/conf/icml/icml2018.html#LiptonWS18},
  volume = 80,
  year = 2018
}

@inproceedings{li2018s,
    title = "What{'}s in a Domain? Learning Domain-Robust Text Representations using Adversarial Training",
    author = "Li, Yitong  and
      Baldwin, Timothy  and
      Cohn, Trevor",
    booktitle = "Proceedings of the 2018 Conference of the North {A}merican Chapter of the Association for Computational Linguistics: Human Language Technologies, Volume 2 (Short Papers)",
    month = jun,
    year = "2018",
    address = "New Orleans, Louisiana",
    publisher = "Association for Computational Linguistics",
    url = "https://www.aclweb.org/anthology/N18-2076",
    doi = "10.18653/v1/N18-2076",
    pages = "474--479",
}

@inproceedings{kim2017domain,
  title={{Domain Attention With an Ensemble of Experts}},
  author={Kim, Young-Bum and Stratos, Karl and Kim, Dongchan},
  booktitle={Proceedings of the 55th Annual Meeting of the Association for Computational Linguistics},
  pages={643--653},
  year={2017}
}

@inproceedings{guo2018multi,
  title={{Multi-Source Domain Adaptation with Mixture of Experts}},
  author={Guo, Jiang and Shah, Darsh and Barzilay, Regina},
  booktitle={EMNLP 2018},
  pages={4694--4703},
  year={2018}
}

@inproceedings{ma2019domain,
  title={{Domain Adaptation with BERT-based Domain Classification and Data Selection}},
  author={Ma, Xiaofei and Xu, Peng and Wang, Zhiguo and Nallapati, Ramesh and Xiang, Bing},
  booktitle={Proceedings of the 2nd Workshop on Deep Learning Approaches for Low-Resource NLP (DeepLo 2019)},
  pages={76--83},
  year={2019}
}

@inproceedings{ganin2015unsupervised,
  title={{Unsupervised Domain Adaptation by Backpropagation}},
  author={Ganin, Yaroslav and Lempitsky, Victor},
  booktitle={International Conference on Machine Learning},
  pages={1180--1189},
  year={2015}
}

@inproceedings{gururangan2020don,
    title = "Don{'}t Stop Pretraining: Adapt Language Models to Domains and Tasks",
    author = "Gururangan, Suchin  and
      Marasovi{\'c}, Ana  and
      Swayamdipta, Swabha  and
      Lo, Kyle  and
      Beltagy, Iz  and
      Downey, Doug  and
      Smith, Noah A.",
    booktitle = "Proceedings of the 58th Annual Meeting of the Association for Computational Linguistics",
    month = jul,
    year = "2020",
    address = "Online",
    publisher = "Association for Computational Linguistics",
    url = "https://www.aclweb.org/anthology/2020.acl-main.740",
    doi = "10.18653/v1/2020.acl-main.740",
    pages = "8342--8360",
}

@inproceedings{han2019unsupervised,
    title = "Unsupervised Domain Adaptation of Contextualized Embeddings for Sequence Labeling",
    author = "Han, Xiaochuang  and
      Eisenstein, Jacob",
    booktitle = "Proceedings of the 2019 Conference on Empirical Methods in Natural Language Processing and the 9th International Joint Conference on Natural Language Processing (EMNLP-IJCNLP)",
    month = nov,
    year = "2019",
    address = "Hong Kong, China",
    publisher = "Association for Computational Linguistics",
    url = "https://www.aclweb.org/anthology/D19-1433",
    doi = "10.18653/v1/D19-1433",
    pages = "4238--4248",
}

@inproceedings{finkel-manning-2009-hierarchical-fixed,
    title = "Hierarchical {B}ayesian {D}omain {A}daptation",
    author = "Finkel, Jenny Rose  and
      Manning, Christopher D.",
    booktitle = "Proceedings of Human Language Technologies: The 2009 Annual Conference of the North {A}merican Chapter of the Association for Computational Linguistics",
    month = jun,
    year = "2009",
    address = "Boulder, Colorado",
    publisher = "Association for Computational Linguistics",
    url = "https://www.aclweb.org/anthology/N09-1068",
    pages = "602--610",
}

@inproceedings{rietzler2019adapt,
    title = "Adapt or Get Left Behind: Domain Adaptation through {BERT} Language Model Finetuning for Aspect-Target Sentiment Classification",
    author = "Rietzler, Alexander  and
      Stabinger, Sebastian  and
      Opitz, Paul  and
      Engl, Stefan",
    booktitle = "Proceedings of the 12th Language Resources and Evaluation Conference",
    month = may,
    year = "2020",
    address = "Marseille, France",
    publisher = "European Language Resources Association",
    url = "https://www.aclweb.org/anthology/2020.lrec-1.607",
    pages = "4933--4941",
    language = "English",
    ISBN = "979-10-95546-34-4",
}

@inproceedings{gui2017part,
  title={{Part-of-Speech Tagging for Twitter with Adversarial Neural Networks}},
  author={Gui, Tao and Zhang, Qi and Huang, Haoran and Peng, Minlong and Huang, Xuan-Jing},
  booktitle={EMNLP 2017},
  pages={2411--2420},
  year={2017}
}

@article{lin2020does,
  title={{Does BERT Need Domain Adaptation for Clinical Negation Detection?}},
  author={Lin, Chen and Bethard, Steven and Dligach, Dmitriy and Sadeque, Farig and Savova, Guergana and Miller, Timothy A},
  journal={Journal of the American Medical Informatics Association},
  volume={27},
  number={4},
  pages={584--591},
  year={2020},
  publisher={Oxford University Press}
}

@inproceedings{blitzer2007biographies,
  title={{Biographies, Bollywood, Boom-Boxes and Blenders: Domain Adaptation for Sentiment Classification}},
  author={Blitzer, John and Dredze, Mark and Pereira, Fernando},
  booktitle={Proceedings of the 45th Annual Meeting of the Association of Computational Linguistics},
  pages={440--447},
  year={2007}
}

@inproceedings{blitzer-etal-2006-domain-fixed,
    title = {{Domain Adaptation with Structural Correspondence Learning}},
    author = "Blitzer, John  and
      McDonald, Ryan  and
      Pereira, Fernando",
    booktitle = "Proceedings of the 2006 Conference on Empirical Methods in Natural Language Processing",
    month = jul,
    year = "2006",
    address = "Sydney, Australia",
    publisher = "Association for Computational Linguistics",
    url = "https://www.aclweb.org/anthology/W06-1615",
    pages = "120--128",
}

@inproceedings{ziser2017neural,
  title={{Neural Structural Correspondence Learning for Domain Adaptation}},
  author={Ziser, Yftah and Reichart, Roi},
  booktitle={Proceedings of the 21st Conference on Computational Natural Language Learning (CoNLL 2017)},
  pages={400--410},
  year={2017}
}

@article{krippendorff2011computing,
  title={Computing Krippendorff's Alpha-Reliability},
  author={Krippendorff, Klaus},
  journal={Computing},
  volume={1},
  number={1},
  pages={25--2011}
}

@article{bojanowski2017enriching,
  title={{Enriching Word Vectors with Subword Information}},
  author={Bojanowski, Piotr and Grave, Edouard and Joulin, Armand and Mikolov, Tomas},
  journal={Transactions of the Association for Computational Linguistics},
  volume={5},
  year={2017},
  issn={2307-387X},
  pages={135--146}
}

@inproceedings{chakrabarty-etal-2018-robust,
    title = "Robust Document Retrieval and Individual Evidence Modeling for Fact Extraction and Verification.",
    author = "Chakrabarty, Tuhin  and
      Alhindi, Tariq  and
      Muresan, Smaranda",
    booktitle = "Proceedings of the First Workshop on Fact Extraction and {VER}ification ({FEVER})",
    month = nov,
    year = "2018",
    address = "Brussels, Belgium",
    publisher = "Association for Computational Linguistics",
    url = "https://www.aclweb.org/anthology/W18-5521",
    doi = "10.18653/v1/W18-5521",
    pages = "127--131",
}

@inproceedings{hanselowski-etal-2018-ukp,
    title = "{UKP}-Athene: Multi-Sentence Textual Entailment for Claim Verification",
    author = "Hanselowski, Andreas  and
      Zhang, Hao  and
      Li, Zile  and
      Sorokin, Daniil  and
      Schiller, Benjamin  and
      Schulz, Claudia  and
      Gurevych, Iryna",
    booktitle = "Proceedings of the First Workshop on Fact Extraction and {VER}ification ({FEVER})",
    month = nov,
    year = "2018",
    address = "Brussels, Belgium",
    publisher = "Association for Computational Linguistics",
    url = "https://www.aclweb.org/anthology/W18-5516",
    doi = "10.18653/v1/W18-5516",
    pages = "103--108",
}

@inproceedings{yoneda-etal-2018-ucl,
    title = "{UCL} Machine Reading Group: Four Factor Framework For Fact Finding ({H}exa{F})",
    author = "Yoneda, Takuma  and
      Mitchell, Jeff  and
      Welbl, Johannes  and
      Stenetorp, Pontus  and
      Riedel, Sebastian",
    booktitle = "Proceedings of the First Workshop on Fact Extraction and {VER}ification ({FEVER})",
    month = nov,
    year = "2018",
    address = "Brussels, Belgium",
    publisher = "Association for Computational Linguistics",
    url = "https://www.aclweb.org/anthology/W18-5515",
    doi = "10.18653/v1/W18-5515",
    pages = "97--102",
}

@inproceedings{malon-2018-team,
    title = "Team Papelo: Transformer Networks at {FEVER}",
    author = "Malon, Christopher",
    booktitle = "Proceedings of the First Workshop on Fact Extraction and {VER}ification ({FEVER})",
    month = nov,
    year = "2018",
    address = "Brussels, Belgium",
    publisher = "Association for Computational Linguistics",
    url = "https://www.aclweb.org/anthology/W18-5517",
    doi = "10.18653/v1/W18-5517",
    pages = "109--113",
}

@inproceedings{karimi-etal-2018-multi,
    title = "Multi-Source Multi-Class Fake News Detection",
    author = "Karimi, Hamid  and
      Roy, Proteek  and
      Saba-Sadiya, Sari  and
      Tang, Jiliang",
    booktitle = "Proceedings of the 27th International Conference on Computational Linguistics",
    month = aug,
    year = "2018",
    address = "Santa Fe, New Mexico, USA",
    publisher = "Association for Computational Linguistics",
    url = "https://www.aclweb.org/anthology/C18-1131",
    pages = "1546--1557",
}

@inproceedings{rajani-etal-2019-explain,
    title = "Explain Yourself! Leveraging Language Models for Commonsense Reasoning",
    author = "Rajani, Nazneen Fatema  and
      McCann, Bryan  and
      Xiong, Caiming  and
      Socher, Richard",
    booktitle = "Proceedings of the 57th Annual Meeting of the Association for Computational Linguistics",
    month = jul,
    year = "2019",
    address = "Florence, Italy",
    publisher = "Association for Computational Linguistics",
    url = "https://www.aclweb.org/anthology/P19-1487",
    doi = "10.18653/v1/P19-1487",
    pages = "4932--4942",
}

@inproceedings{alhindi-etal-2018-evidence,
    title = "Where is Your Evidence: Improving Fact-checking by Justification Modeling",
    author = "Alhindi, Tariq  and
      Petridis, Savvas  and
      Muresan, Smaranda",
    booktitle = "Proceedings of the First Workshop on Fact Extraction and {VER}ification ({FEVER})",
    month = nov,
    year = "2018",
    address = "Brussels, Belgium",
    publisher = "Association for Computational Linguistics",
    url = "https://www.aclweb.org/anthology/W18-5513",
    doi = "10.18653/v1/W18-5513",
    pages = "85--90",
}

@inproceedings{mohtarami-etal-2018-automatic,
    title = "Automatic Stance Detection Using End-to-End Memory Networks",
    author = "Mohtarami, Mitra  and
      Baly, Ramy  and
      Glass, James  and
      Nakov, Preslav  and
      M{\`a}rquez, Llu{\'\i}s  and
      Moschitti, Alessandro",
    booktitle = "Proceedings of the 2018 Conference of the North {A}merican Chapter of the Association for Computational Linguistics: Human Language Technologies, Volume 1 (Long Papers)",
    month = jun,
    year = "2018",
    address = "New Orleans, Louisiana",
    publisher = "Association for Computational Linguistics",
    url = "https://www.aclweb.org/anthology/N18-1070",
    doi = "10.18653/v1/N18-1070",
    pages = "767--776",
}

@article{goyal2017accurate,
  title={{Accurate, Large Minibatch SGD: Training ImageNet in 1 Hour}},
  author={Goyal, Priya and Doll{\'a}r, Piotr and Girshick, Ross and Noordhuis, Pieter and Wesolowski, Lukasz and Kyrola, Aapo and Tulloch, Andrew and Jia, Yangqing and He, Kaiming},
  journal={arXiv preprint arXiv:1706.02677},
  year={2017}
}

@inproceedings{yang2019xfake,
  title={{XFake: Explainable Fake News Detector with Visualizations}},
  author={Yang, Fan and Pentyala, Shiva K and Mohseni, Sina and Du, Mengnan and Yuan, Hao and Linder, Rhema and Ragan, Eric D and Ji, Shuiwang and Hu, Xia},
  booktitle={The World Wide Web Conference},
  pages={3600--3604},
  year={2019}
}

@inproceedings{shu2019defend,
  title={{dEFEND: Explainable Fake News Detection}},
  author={Shu, Kai and Cui, Limeng and Wang, Suhang and Lee, Dongwon and Liu, Huan},
  booktitle={Proceedings of the 25th ACM SIGKDD International Conference on Knowledge Discovery \& Data Mining},
  pages={395--405},
  year={2019}
}

@inproceedings{Adebayo:2018:SCS:3327546.3327621,
 author = {Adebayo, Julius and Gilmer, Justin and Muelly, Michael and Goodfellow, Ian and Hardt, Moritz and Kim, Been},
 title = {{Sanity Checks for Saliency Maps}},
 booktitle = {{Proceedings of the 32nd International Conference on Neural Information Processing Systems}},
 series = {NIPS'18},
 year = {2018},
 location = {Montr\&\#233;al, Canada},
 pages = {9525--9536},
 numpages = {12},
 url = {http://dl.acm.org/citation.cfm?id=3327546.3327621},
 acmid = {3327621},
 publisher = {Curran Associates Inc.},
 address = {USA},
}

@article{hayes2007answering,
  title={{Answering the Call for a Standard Reliability Measure for Coding Data}},
  author={Hayes, Andrew F and Krippendorff, Klaus},
  journal={{Communication Methods and Measures}},
  volume={1},
  number={1},
  pages={77--89},
  year={2007},
  publisher={Taylor \& Francis}
}

@article{loshchilov2017fixing,
  title={{Fixing Weight Decay Regularization in Adam}},
  author={Loshchilov, Ilya and Hutter, Frank},
  journal={{arXiv preprint arXiv:1711.05101}},
  year={2017}
}

@inproceedings{misra2016cross,
  title={{Cross-stitch networks for multi-task learning}},
  author={Misra, Ishan and Shrivastava, Abhinav and Gupta, Abhinav and Hebert, Martial},
  booktitle={{Proceedings of the IEEE Conference on Computer Vision and Pattern Recognition}},
  pages={3994--4003},
  year={2016}
}

@inproceedings{long2017fake,
  title={{Fake News Detection Through Multi-Perspective Speaker Profiles}},
  author={Long, Yunfei and Lu, Qin and Xiang, Rong and Li, Minglei and Huang, Chu-Ren},
  booktitle={{Proceedings of the Eighth International Joint Conference on Natural Language Processing (Volume 2: Short Papers)}},
  pages={252--256},
  year={2017}
}

@inproceedings{Nallapati:2017:SRN:3298483.3298681,
 author = {Nallapati, Ramesh and Zhai, Feifei and Zhou, Bowen},
 title = {{SummaRuNNer: A Recurrent Neural Network Based Sequence Model for Extractive Summarization of Documents}},
 booktitle = {{Proceedings of the Thirty-First AAAI Conference on Artificial Intelligence}},
 series = {AAAI'17},
 year = {2017},
 location = {San Francisco, California, USA},
 pages = {3075--3081},
 numpages = {7},
 url = {http://dl.acm.org/citation.cfm?id=3298483.3298681},
 acmid = {3298681},
 publisher = {AAAI Press},
}

@inproceedings{sanh2019distilbert,
    title={{DistilBERT, a distilled version of BERT: smaller, faster, cheaper and lighter}},
    author={Victor Sanh and Lysandre Debut and Julien Chaumond and Thomas Wolf},
    year={2019},
    booktitle={{Proceedings of the 5th Workshop on Energy Efficient Machine Learning and Cognitive Computing}},
    series={NeurIPS'2019}
    
}

@inproceedings{liu-lapata-2019-text,
    title = {{Text Summarization with Pretrained Encoders}},
    author = "Liu, Yang  and
      Lapata, Mirella",
    booktitle = {{Proceedings of the 2019 Conference on Empirical Methods in Natural Language Processing and the 9th International Joint Conference on Natural Language Processing (EMNLP-IJCNLP)}},
    month = nov,
    year = "2019",
    address = "Hong Kong, China",
    publisher = "Association for Computational Linguistics",
    url = "https://www.aclweb.org/anthology/D19-1387",
    doi = "10.18653/v1/D19-1387",
    pages = "3728--3738",
}

@inproceedings{stammbach-neumann-2019-team,
    title = {{Team {DOMLIN}: Exploiting Evidence Enhancement for the {FEVER} Shared Task}},
    author = "Stammbach, Dominik  and
      Neumann, Guenter",
    booktitle = {{Proceedings of the Second Workshop on Fact Extraction and VERification (FEVER)}},
    month = nov,
    year = "2019",
    address = "Hong Kong, China",
    publisher = "Association for Computational Linguistics",
    url = "https://www.aclweb.org/anthology/D19-6616",
    doi = "10.18653/v1/D19-6616",
    pages = "105--109"
}

@inproceedings{lee2018improving,
  title={{Improving large-scale fact-checking using decomposable attention models and lexical tagging}},
  author={Lee, Nayeon and Wu, Chien-Sheng and Fung, Pascale},
  booktitle={{Proceedings of the 2018 Conference on Empirical Methods in Natural Language Processing}},
  pages={1133--1138},
  year={2018}
}

@inproceedings{Xu2019AdversarialDA,
  title={{Adversarial Domain Adaptation for Stance Detection}},
  author={Brian Xu and Mitra Mohtarami and James R. Glass},
  booktitle={{Proceedings of the Thirty-second Annual Conference on Neural Information Processing Systems
(NeurIPS)–Continual Learning}},
  year={2018}
}

@inproceedings{Ma:2018:DRS:3184558.3188729,
 author = {Ma, Jing and Gao, Wei and Wong, Kam-Fai},
 title = {{Detect Rumor and Stance Jointly by Neural Multi-task Learning}},
 booktitle = {{Companion Proceedings of the The Web Conference 2018}},
 series = {WWW '18},
 year = {2018},
 isbn = {978-1-4503-5640-4},
 location = {Lyon, France},
 pages = {585--593},
 numpages = {9},
 url = {https://doi.org/10.1145/3184558.3188729},
 doi = {10.1145/3184558.3188729},
 acmid = {3188729},
 publisher = {International World Wide Web Conferences Steering Committee},
 address = {Republic and Canton of Geneva, Switzerland}
}

@incollection{NIPS2018_8163,
title = {{e-SNLI: Natural Language Inference with Natural Language Explanations}},
author = {Camburu, Oana-Maria and Rockt\"{a}schel, Tim and Lukasiewicz, Thomas and Blunsom, Phil},
booktitle = {{Advances in Neural Information Processing Systems 31}},
editor = {S. Bengio and H. Wallach and H. Larochelle and K. Grauman and N. Cesa-Bianchi and R. Garnett},
pages = {9539--9549},
year = {2018},
publisher = {Curran Associates, Inc.},
url = {http://papers.nips.cc/paper/8163-e-snli-natural-language-inference-with-natural-language-explanations.pdf}
}

@inproceedings{wagner2019interpretable,
  title={{Interpretable and Fine-Grained Visual Explanations for Convolutional Neural Networks}},
  author={Wagner, J{\"o}rg and K{\"o}hler, Jan Mathias and Gindele, Tobias and Hetzel, Leon and Wiedemer, Jakob Thadd{\"a}us and Behnke, Sven},
  booktitle={2019 IEEE/CVF Conference on Computer Vision and Pattern Recognition (CVPR)},
  pages={9089--9099},
  year={2019},
  organization={IEEE}
}

@incollection{kindermans2019reliability,
  title={The (un) reliability of saliency methods},
  author={Kindermans, Pieter-Jan and Hooker, Sara and Adebayo, Julius and Alber, Maximilian and Sch{\"u}tt, Kristof T and D{\"a}hne, Sven and Erhan, Dumitru and Kim, Been},
  booktitle={Explainable AI: Interpreting, Explaining and Visualizing Deep Learning},
  pages={267--280},
  year={2019},
  publisher={Springer}
}

@inproceedings{jacovi2020towards,
    title = "Towards Faithfully Interpretable {NLP} Systems: How Should We Define and Evaluate Faithfulness?",
    author = "Jacovi, Alon  and
      Goldberg, Yoav",
    booktitle = "Proceedings of the 58th Annual Meeting of the Association for Computational Linguistics",
    month = jul,
    year = "2020",
    address = "Online",
    publisher = "Association for Computational Linguistics",
    url = "https://www.aclweb.org/anthology/2020.acl-main.386",
    doi = "10.18653/v1/2020.acl-main.386",
    pages = "4198--4205",
}

@inproceedings{poerner-etal-2018-evaluating,
    title = {{Evaluating neural network explanation methods using hybrid documents and morphosyntactic agreement}},
    author = {Poerner, Nina  and
      Sch{\"u}tze, Hinrich  and
      Roth, Benjamin},
    booktitle = "Proceedings of the 56th Annual Meeting of the Association for Computational Linguistics (Volume 1: Long Papers)",
    month = jul,
    year = "2018",
    address = "Melbourne, Australia",
    publisher = "Association for Computational Linguistics",
    url = "https://www.aclweb.org/anthology/P18-1032",
    doi = "10.18653/v1/P18-1032",
    pages = "340--350",
}

@article{alvarez2018robustness,
  title={On the robustness of interpretability methods},
  author={Alvarez-Melis, David and Jaakkola, Tommi S},
  journal={arXiv preprint arXiv:1806.08049},
  year={2018}
}

@article{regulation2016regulation,
  title={Regulation (EU) 2016/679 of the European Parliament and of the Council of 27 April 2016 on the protection of natural persons with regard to the processing of personal data and on the free movement of such data, and repealing Directive 95/46},
  author={Regulation, General Data Protection},
  journal={Official Journal of the European Union (OJ)},
  volume={59},
  number={1-88},
  pages={294},
  year={2016}
}

@Inbook{barakat2007rule,
author="Martens, David
and Huysmans, Johan
and Setiono, Rudy
and Vanthienen, Jan
and Baesens, Bart",
title={{Rule Extraction from Support Vector Machines: An Overview of Issues and Application in Credit Scoring}},
bookTitle="Rule Extraction from Support Vector Machines",
year="2008",
publisher="Springer Berlin Heidelberg",
address="Berlin, Heidelberg",
pages="33--63",
isbn="978-3-540-75390-2",
doi="10.1007/978-3-540-75390-2_2",
url="https://doi.org/10.1007/978-3-540-75390-2_2"
}

@inproceedings{zeiler2014visualizing,
  title={{Visualizing and understanding convolutional networks}},
  author={Zeiler, Matthew D and Fergus, Rob},
  booktitle={European conference on computer vision},
  pages={818--833},
  year={2014},
  organization={Springer}
}

@article{castro2009polynomial,
author = {Castro, Javier and G\'{o}mez, Daniel and Tejada, Juan},
title = {{Polynomial Calculation of the Shapley Value Based on Sampling}},
year = {2009},
issue_date = {May 2009},
publisher = {Elsevier Science Ltd.},
address = {GBR},
volume = {36},
number = {5},
issn = {0305-0548},
url = {https://doi.org/10.1016/j.cor.2008.04.004},
doi = {10.1016/j.cor.2008.04.004},
journal = {Comput. Oper. Res.},
month = may,
pages = {1726–1730},
numpages = {5}
}

@inproceedings{springenberg2014striving,
  author    = {Jost Tobias Springenberg and
               Alexey Dosovitskiy and
               Thomas Brox and
               Martin A. Riedmiller},
  editor    = {Yoshua Bengio and
               Yann LeCun},
  title     = {Striving for Simplicity: The All Convolutional Net},
  booktitle = {3rd International Conference on Learning Representations, {ICLR} 2015,
               San Diego, CA, USA, May 7-9, 2015, Workshop Track Proceedings},
  year      = {2015},
  url       = {http://arxiv.org/abs/1412.6806},
  bibsource = {dblp computer science bibliography, https://dblp.org}
}

@article{Kindermans2016InvestigatingTI,
  author    = {Pieter{-}Jan Kindermans and
               Kristof Sch{\"{u}}tt and
               Klaus{-}Robert M{\"{u}}ller and
               Sven D{\"{a}}hne},
  title     = {Investigating the influence of noise and distractors on the interpretation
               of neural networks},
  journal   = {CoRR},
  volume    = {abs/1611.07270},
  year      = {2016},
  url       = {http://arxiv.org/abs/1611.07270},
  archivePrefix = {arXiv},
  eprint    = {1611.07270},
  bibsource = {dblp computer science bibliography, https://dblp.org}
}

@article{vashishth2019attention,
  author    = {Shikhar Vashishth and
               Shyam Upadhyay and
               Gaurav Singh Tomar and
               Manaal Faruqui},
  title     = {Attention Interpretability Across {NLP} Tasks},
  journal   = {CoRR},
  volume    = {abs/1909.11218},
  year      = {2019},
  url       = {http://arxiv.org/abs/1909.11218},
  archivePrefix = {arXiv},
  eprint    = {1909.11218},
  bibsource = {dblp computer science bibliography, https://dblp.org}
}

@inproceedings{Simonyan2013DeepIC,
  author    = {Karen Simonyan and
               Andrea Vedaldi and
               Andrew Zisserman},
  editor    = {Yoshua Bengio and
               Yann LeCun},
  title     = {Deep Inside Convolutional Networks: Visualising Image Classification
               Models and Saliency Maps},
  booktitle = {2nd International Conference on Learning Representations, {ICLR} 2014,
               Banff, AB, Canada, April 14-16, 2014, Workshop Track Proceedings},
  year      = {2014},
  url       = {http://arxiv.org/abs/1312.6034},
  bibsource = {dblp computer science bibliography, https://dblp.org}
}

@inproceedings{sundararajan2017axiomatic,
  title={{Axiomatic attribution for deep networks}},
  author={Sundararajan, Mukund and Taly, Ankur and Yan, Qiqi},
  booktitle={Proceedings of the 34th International Conference on Machine Learning-Volume 70},
  pages={3319--3328},
  year={2017},
  organization={JMLR. org}
}

@inproceedings{lundberg2017unified,
  author    = {Scott M. Lundberg and
               Su{-}In Lee},
  editor    = {Isabelle Guyon and
               Ulrike von Luxburg and
               Samy Bengio and
               Hanna M. Wallach and
               Rob Fergus and
               S. V. N. Vishwanathan and
               Roman Garnett},
  title     = {A Unified Approach to Interpreting Model Predictions},
  booktitle = {Advances in Neural Information Processing Systems 30: Annual Conference
               on Neural Information Processing Systems 2017, December 4-9, 2017,
               Long Beach, CA, {USA}},
  pages     = {4765--4774},
  year      = {2017},
  url       = {http://papers.nips.cc/paper/7062-a-unified-approach-to-interpreting-model-predictions},
  bibsource = {dblp computer science bibliography, https://dblp.org}
}

@inproceedings{bansal2016ask,
author = {Bansal, Trapit and Belanger, David and McCallum, Andrew},
title = {{Ask the GRU: Multi-Task Learning for Deep Text Recommendations}},
year = {2016},
isbn = {9781450340359},
publisher = {Association for Computing Machinery},
address = {New York, NY, USA},
url = {https://doi.org/10.1145/2959100.2959180},
doi = {10.1145/2959100.2959180},
booktitle = {Proceedings of the 10th ACM Conference on Recommender Systems},
pages = {107–114},
numpages = {8},
location = {Boston, Massachusetts, USA},
series = {RecSys ’16}
}

@inproceedings{johansson2004truth,
  title={{The Truth is in There Rule Extraction from Opaque Models Using Genetic Programming}},
  author={Johansson, Ulf and K{\"o}nig, Rikard and Niklasson, Lars},
  booktitle={Proceedings of the Seventeenth International Florida Artificial Intelligence Research Society Conference, FLAIRS 2004},
  year={2004},
  organization={AAAI Press}
}

@inproceedings{ribeiromodel,
  title={{Model-Agnostic Interpretability of Machine Learning}},
  author={Ribeiro, Marco Tulio and EDU, UW and Singh, Sameer and Guestrin, Carlos},
  booktitle={ICML Workshop on Human Interpretability in Machine Learning.},
  year={2016}
}

@article{BARREDOARRIETA202082,
 author = {Alejandro Barredo Arrieta and Natalia D{\'i}az-Rodr{\'i}guez and Javier Del Ser and Adrien Bennetot and Siham Tabik and Alberto Barbado and Salvador Garcia and Sergio Gil-Lopez and Daniel Molina and Richard Benjamins and Raja Chatila and Francisco Herrera},
 doi = {https://doi.org/10.1016/j.inffus.2019.12.012},
 issn = {1566-2535},
 journal = {Information Fusion},
 pages = {82 - 115},
 title = {{Explainable Artificial Intelligence (XAI): Concepts, taxonomies, opportunities and challenges toward responsible AI}},
 url = {http://www.sciencedirect.com/science/article/pii/S1566253519308103},
 volume = {58},
 year = {2020}
}

@article{shapley1953value,
  title={{A value for n-person games}},
  author={Shapley, Lloyd S},
  journal={Contributions to the Theory of Games},
  volume={2},
  number={28},
  pages={307--317},
  year={1953}
}

@InProceedings{guan2019towards,
  title = 	 {{Towards a Deep and Unified Understanding of Deep Neural Models in {NLP}}},
  author = 	 {Guan, Chaoyu and Wang, Xiting and Zhang, Quanshi and Chen, Runjin and He, Di and Xie, Xing},
  booktitle = 	 {Proceedings of the 36th International Conference on Machine Learning},
  pages = 	 {2454--2463},
  year = 	 {2019},
  editor = 	 {Chaudhuri, Kamalika and Salakhutdinov, Ruslan},
  volume = 	 {97},
  series = 	 {Proceedings of Machine Learning Research},
  address = 	 {Long Beach, California, USA},
  month = 	 {09--15 Jun},
  publisher = 	 {PMLR},
  url = 	 {http://proceedings.mlr.press/v97/guan19a.html}
}

@article{hahnloser2000digital,
  title={{Digital selection and analogue amplification coexist in a cortex-inspired silicon circuit}},
  author={Hahnloser, Richard HR and Sarpeshkar, Rahul and Mahowald, Misha A and Douglas, Rodney J and Seung, H Sebastian},
  journal={Nature},
  volume={405},
  number={6789},
  pages={947--951},
  year={2000},
  publisher={Nature Publishing Group}
}

@inproceedings{pennington2014glove,
  author = {Jeffrey Pennington and Richard Socher and Christopher D. Manning},
  booktitle = {Empirical Methods in Natural Language Processing (EMNLP)},
  title = {{GloVe: Global Vectors for Word Representation}},
  year = {2014},
  pages = {1532--1543},
  url = {http://www.aclweb.org/anthology/D14-1162},
}

@incollection{robnik2018perturbation,
  author    = {Marko Robnik{-}Sikonja and
               Marko Bohanec},
  editor    = {Jianlong Zhou and
               Fang Chen},
  title     = {Perturbation-Based Explanations of Prediction Models},
  booktitle = {Human and Machine Learning - Visible, Explainable, Trustworthy and
               Transparent},
  series    = {Human-Computer Interaction Series},
  pages     = {159--175},
  publisher = {Springer},
  year      = {2018},
  url       = {https://doi.org/10.1007/978-3-319-90403-0\_9},
  doi       = {10.1007/978-3-319-90403-0\_9},
  bibsource = {dblp computer science bibliography, https://dblp.org}
}

@article{hochreiter1997long,
  title={{Long short-term memory}},
  author={Hochreiter, Sepp and Schmidhuber, J{\"u}rgen},
  journal={Neural Computation},
  volume={9},
  number={8},
  pages={1735--1780},
  year={1997},
  publisher={MIT Press}
}

@article{fukushima1980neocognitron,
  title={{Neocognitron: A self-organizing neural network model for a mechanism of pattern recognition unaffected by shift in position}},
  author={Fukushima, Kunihiko},
  journal={Biological Cybernetics},
  volume={36},
  number={4},
  pages={193--202},
  year={1980},
  publisher={Springer}
}

@inproceedings{subramanian2020obtaining,
    title = "Obtaining Faithful Interpretations from Compositional Neural Networks",
    author = "Subramanian, Sanjay  and
      Bogin, Ben  and
      Gupta, Nitish  and
      Wolfson, Tomer  and
      Singh, Sameer  and
      Berant, Jonathan  and
      Gardner, Matt",
    booktitle = "Proceedings of the 58th Annual Meeting of the Association for Computational Linguistics",
    month = jul,
    year = "2020",
    address = "Online",
    publisher = "Association for Computational Linguistics",
    url = "https://www.aclweb.org/anthology/2020.acl-main.495",
    doi = "10.18653/v1/2020.acl-main.495",
    pages = "5594--5608",
}

@inproceedings{khashabi-etal-2018-looking,
    title = {{Looking Beyond the Surface: A Challenge Set for Reading Comprehension over Multiple Sentences}},
    author = "Khashabi, Daniel  and
      Chaturvedi, Snigdha  and
      Roth, Michael  and
      Upadhyay, Shyam  and
      Roth, Dan",
    booktitle = "Proceedings of the 2018 Conference of the North {A}merican Chapter of the Association for Computational Linguistics: Human Language Technologies, Volume 1 (Long Papers)",
    month = jun,
    year = "2018",
    address = "New Orleans, Louisiana",
    publisher = "Association for Computational Linguistics",
    url = "https://www.aclweb.org/anthology/N18-1023",
    doi = "10.18653/v1/N18-1023",
    pages = "252--262",
}

@inproceedings{zaidan-etal-2007-using,
    title = {{Using {``}Annotator Rationales{''} to Improve Machine Learning for Text Categorization}},
    author = "Zaidan, Omar  and
      Eisner, Jason  and
      Piatko, Christine",
    booktitle = "Human Language Technologies 2007: The Conference of the North {A}merican Chapter of the Association for Computational Linguistics; Proceedings of the Main Conference",
    month = apr,
    year = "2007",
    address = "Rochester, New York",
    publisher = "Association for Computational Linguistics",
    url = "https://www.aclweb.org/anthology/N07-1033",
    pages = "260--267",
}

@inproceedings{clark2019boolq,
    title = {{{B}ool{Q}: Exploring the Surprising Difficulty of Natural Yes/No Questions}},
    author = "Clark, Christopher  and
      Lee, Kenton  and
      Chang, Ming-Wei  and
      Kwiatkowski, Tom  and
      Collins, Michael  and
      Toutanova, Kristina",
    booktitle = "Proceedings of the 2019 Conference of the North {A}merican Chapter of the Association for Computational Linguistics: Human Language Technologies, Volume 1 (Long and Short Papers)",
    month = jun,
    year = "2019",
    address = "Minneapolis, Minnesota",
    publisher = "Association for Computational Linguistics",
    url = "https://www.aclweb.org/anthology/N19-1300",
    doi = "10.18653/v1/N19-1300",
    pages = "2924--2936",
}

@article{deyoung2019eraser,
  title={{ERASER: A Benchmark to Evaluate Rationalized NLP Models}},
  author={DeYoung, Jay and Jain, Sarthak and Rajani, Nazneen Fatema and Lehman, Eric and Xiong, Caiming and Socher, Richard and Wallace, Byron C},
  booktitle={Proceedings of the conference. Association for Computational Linguistics. Meeting},
  pages={},
  volume={2020},
  year={2020},
}

@article{rich2016machine,
  title={{Machine learning, automated suspicion algorithms, and the fourth amendment}},
  author={Rich, Michael L},
  journal={University of Pennsylvania Law Review},
  pages={871--929},
  year={2016},
  publisher={JSTOR}
}

@article{goodman2017european,
  title={{European Union regulations on algorithmic decision-making and a “right to explanation”}},
  author={Goodman, Bryce and Flaxman, Seth},
  journal={AI magazine},
  volume={38},
  number={3},
  pages={50--57},
  year={2017}
}

@inproceedings{caruana2015intelligible,
author = {Caruana, Rich and Lou, Yin and Gehrke, Johannes and Koch, Paul and Sturm, Marc and Elhadad, Noemie},
title = {{Intelligible Models for HealthCare: Predicting Pneumonia Risk and Hospital 30-Day Readmission}},
year = {2015},
isbn = {9781450336642},
publisher = {Association for Computing Machinery},
address = {New York, NY, USA},
url = {https://doi.org/10.1145/2783258.2788613},
doi = {10.1145/2783258.2788613},
booktitle = {Proceedings of the 21th ACM SIGKDD International Conference on Knowledge Discovery and Data Mining},
pages = {1721–1730},
numpages = {10},
location = {Sydney, NSW, Australia},
series = {KDD ’15}
}

@inproceedings{Lertvittayakumjorn2019HumangroundedEO,
    title = {{Human-grounded Evaluations of Explanation Methods for Text Classification}},
    author = "Lertvittayakumjorn, Piyawat  and
      Toni, Francesca",
    booktitle = "Proceedings of the 2019 Conference on Empirical Methods in Natural Language Processing and the 9th International Joint Conference on Natural Language Processing (EMNLP-IJCNLP)",
    month = nov,
    year = "2019",
    address = "Hong Kong, China",
    publisher = "Association for Computational Linguistics",
    url = "https://www.aclweb.org/anthology/D19-1523",
    doi = "10.18653/v1/D19-1523",
    pages = "5195--5205",
}

@inproceedings{arras-etal-2019-evaluating,
    title = {{Evaluating Recurrent Neural Network Explanations}},
    author = {Arras, Leila  and
      Osman, Ahmed  and
      M{\"u}ller, Klaus-Robert  and
      Samek, Wojciech},
    booktitle = "Proceedings of the 2019 ACL Workshop BlackboxNLP: Analyzing and Interpreting Neural Networks for NLP",
    month = aug,
    year = "2019",
    address = "Florence, Italy",
    publisher = "Association for Computational Linguistics",
    url = "https://www.aclweb.org/anthology/W19-4813",
    doi = "10.18653/v1/W19-4813",
    pages = "113--126",
}

@article{narayanan2018humans,
  title={{How do humans understand explanations from machine learning systems? an evaluation of the human-interpretability of explanation}},
  author={Narayanan, Menaka and Chen, Emily and He, Jeffrey and Kim, Been and Gershman, Sam and Doshi-Velez, Finale},
  journal={arXiv preprint arXiv:1802.00682},
  year={2018}
}

@inproceedings{feng2018pathologies,
    title = {{Pathologies of Neural Models Make Interpretations Difficult}},
    author = "Feng, Shi  and
      Wallace, Eric  and
      Grissom II, Alvin  and
      Iyyer, Mohit  and
      Rodriguez, Pedro  and
      Boyd-Graber, Jordan",
    booktitle = "Proceedings of the 2018 Conference on Empirical Methods in Natural Language Processing",
    month = oct # "-" # nov,
    year = "2018",
    address = "Brussels, Belgium",
    publisher = "Association for Computational Linguistics",
    url = "https://www.aclweb.org/anthology/D18-1407",
    doi = "10.18653/v1/D18-1407",
    pages = "3719--3728",
}

@inproceedings{lei2016rationalizing,
    title = {{Rationalizing Neural Predictions}},
    author = "Lei, Tao  and
      Barzilay, Regina  and
      Jaakkola, Tommi",
    booktitle = "Proceedings of the 2016 Conference on Empirical Methods in Natural Language Processing",
    month = nov,
    year = "2016",
    address = "Austin, Texas",
    publisher = "Association for Computational Linguistics",
    url = "https://www.aclweb.org/anthology/D16-1011",
    doi = "10.18653/v1/D16-1011",
    pages = "107--117",
}

@InProceedings{pmlr-v124-rethmeier20a, 
 title = {TX-Ray: Quantifying and Explaining Model-Knowledge Transfer in (Un-)Supervised NLP}, 
 author = {Rethmeier, Nils and Kumar Saxena, Vageesh and Augenstein, Isabelle}, 
 booktitle = {Proceedings of the 36th Conference on Uncertainty in Artificial Intelligence (UAI)}, 
 pages = {440--449}, 
 year = {2020}, 
 editor = {Jonas Peters and David Sontag}, 
 volume = {124}, 
 series = {Proceedings of Machine Learning Research}, 
 address = {Virtual}, 
 month = {03--06 Aug}, 
 publisher = {PMLR}, 
 url = {http://proceedings.mlr.press/v124/rethmeier20a.html} 
 }

@inproceedings{sterckx2016supervised,
    title = "Supervised Keyphrase Extraction as Positive Unlabeled Learning",
    author = "Sterckx, Lucas  and
      Caragea, Cornelia  and
      Demeester, Thomas  and
      Develder, Chris",
    booktitle = "Proceedings of the 2016 Conference on Empirical Methods in Natural Language Processing",
    month = nov,
    year = "2016",
    address = "Austin, Texas",
    publisher = "Association for Computational Linguistics",
    url = "https://www.aclweb.org/anthology/D16-1198",
    doi = "10.18653/v1/D16-1198",
    pages = "1924--1929",
}

@inproceedings{jaradat2018claimrank,
    title = "{C}laim{R}ank: Detecting Check-Worthy Claims in {A}rabic and {E}nglish",
    author = "Jaradat, Israa  and
      Gencheva, Pepa  and
      Barr{\'o}n-Cede{\~n}o, Alberto  and
      M{\`a}rquez, Llu{\'\i}s  and
      Nakov, Preslav",
    booktitle = "Proceedings of the 2018 Conference of the North {A}merican Chapter of the Association for Computational Linguistics: Demonstrations",
    month = jun,
    year = "2018",
    address = "New Orleans, Louisiana",
    publisher = "Association for Computational Linguistics",
    url = "https://www.aclweb.org/anthology/N18-5006",
    doi = "10.18653/v1/N18-5006",
    pages = "26--30",
}

@inproceedings{gencheva-etal-2017-context-fixed,
    title = "{A Context-Aware Approach for Detecting Worth-Checking Claims in Political Debates}",
    author = "Gencheva, Pepa  and
      Nakov, Preslav  and
      M{\`a}rquez, Llu{\'\i}s  and
      Barr{\'o}n-Cede{\~n}o, Alberto  and
      Koychev, Ivan",
    booktitle = "Proceedings of the International Conference Recent Advances in Natural Language Processing, {RANLP} 2017",
    month = sep,
    year = "2017",
    address = "Varna, Bulgaria",
    publisher = "INCOMA Ltd.",
    url = "https://doi.org/10.26615/978-954-452-049-6_037",
    doi = "10.26615/978-954-452-049-6_037",
    pages = "267--276",
}

@inproceedings{thorne-vlachos-2018-automated-fixed,
    title = "{Automated Fact Checking: Task Formulations, Methods and Future Directions}",
    author = "Thorne, James  and
      Vlachos, Andreas",
    booktitle = "Proceedings of the 27th International Conference on Computational Linguistics",
    month = aug,
    year = "2018",
    address = "Santa Fe, New Mexico, USA",
    publisher = "Association for Computational Linguistics",
    url = "https://www.aclweb.org/anthology/C18-1283",
    pages = "3346--3359",
}

@article{graves2016rise,
  title={The Rise of Fact-Checking Sites in Europe},
  author={Graves, Lucas and Cherubini, Federica},
  year={2016},
  journal={Reuters Institute for the Study of Journalism}
}

\clearpage

\end{document}